\documentclass{scrreport}

\usepackage{floatpag}
\KOMAoptions{bibliography=totoc}
\KOMAoptions{abstract=yes}
\KOMAoptions{twoside=semi}
\floatpagestyle{plain}
\KOMAoption{headsepline}{on}
\KOMAoption{chapterprefix}{on}

\usepackage[utf8]{inputenc}
\usepackage[ngerman,USenglish]{babel}

\usepackage[Sonny]{fncychap}

\usepackage{hyperref}

\usepackage{etoolbox}
\AtBeginEnvironment{abstract}{}

\usepackage{titling}
\usepackage[final]{RWTH-Fak1}

\usepackage[nohyperlinks]{acronym}
\usepackage[chapter]{algorithm}
\usepackage{algorithmicx}
\usepackage{algpseudocode}
\usepackage{amsmath}
\usepackage{amssymb}
\usepackage[backend=biber,style=alphabetic,maxbibnames=10,useprefix=true,backref,backrefstyle=none,datamodel=contribution]{biblatex}
\usepackage{booktabs}
\usepackage[column=C]{cellspace}
\usepackage{csquotes}
\usepackage[inline,shortlabels]{enumitem}
\usepackage{environ}
\usepackage{hyphenat}
\usepackage{ifthen}
\usepackage{listings}
\usepackage{mathtools}
\usepackage{mfirstuc}
\usepackage{mleftright}
\usepackage{amsthm}
\usepackage{thmtools}
\usepackage{rotating}
\usepackage{siunitx}
\usepackage[mode=buildnew]{standalone}
\usepackage{subcaption}
\usepackage{stmaryrd}
\usepackage{tikz}
\usepackage[textsize=scriptsize,disable]{todonotes}
\usepackage{multirow} 
\usepackage{booktabs} 
\usepackage{tabulary} 
\usepackage{placeins}
\usepackage{xcolor}
\usepackage{xspace}
\usepackage{xpatch}
\usepackage{microtype}

\DefineBibliographyStrings{english}{%
  backrefpage = {cited on page},
  backrefpages = {cited on pages}
}

\declaretheorem[numberwithin=chapter]{theorem}

\declaretheorem[sibling=theorem,numberwithin=chapter]{corollary}
\declaretheorem[numberwithin=chapter]{proposition}
\declaretheorem[numberwithin=chapter,style=remark]{remark}
\declaretheorem[numberwithin=chapter,style=definition,qed=${}_\Box$]{definition}
\declaretheorem[numberwithin=chapter,style=definition,qed=${}_\Box$]{example}

\usepackage[createShortEnv,conf={restate, no link to proof,text proof={Proof}}]{proof-at-the-end}

\algnewcommand{\IIf}[1]{\State\algorithmicif\ #1\ \algorithmicthen}
\algnewcommand{\IEndIf}{\unskip\ \algorithmicend\ \algorithmicif}

\renewcommand*{\sectionautorefname}{Section}
\renewcommand*{\chapterautorefname}{Chapter}
\renewcommand*{\subsectionautorefname}{Section}
\renewcommand*{\subsubsectionautorefname}{Section}

\renewcaptionname{USenglish}{\sectionautorefname}{Section}
\renewcaptionname{USenglish}{\chapterautorefname}{Chapter}
\renewcaptionname{USenglish}{\subsectionautorefname}{Section}
\renewcaptionname{USenglish}{\subsubsectionautorefname}{Section}

\renewcaptionname{USenglish}{\listfigurename}{Figures}
\renewcaptionname{USenglish}{\listtablename}{Tables}

\lstdefinelanguage{golog}
{
  keywords={
    while,
    if,
    else,
    concurrent,
    precondition,
    initially,
    effect,
    start_effect,
  },
  morekeywords={[2]
    action,
    bool,
    fluent,
    symbol,
    domain,
  },
  keywordstyle=\color{rwth-blue}\bfseries,
  keywordstyle=[2]\color{rwth-lblack}\bfseries,
}

\lstdefinestyle{golog}{
  language=golog,
  basicstyle=\ttfamily\small,
  numbersep=6pt,
}

\lstMakeShortInline[basicstyle=\ttfamily\footnotesize]|

\usetikzlibrary{automata,arrows,arrows.meta,calc,fit,backgrounds,shapes,
  shapes.geometric,patterns,decorations.markings,
  decorations.pathreplacing,positioning,
  decorations,matrix,intersections}
\tikzset{
  mynode/.style={inner sep=0pt, rectangle,draw=black,thick, rounded corners,
  minimum width=2cm, minimum height=1cm},
	pictocirc/.style={inner sep=3pt,draw=black,fill=black,circle},
	pictorect/.style={inner sep=4pt,draw=black,fill=black,rectangle},
	pictotri/.style={inner sep=2pt,draw=black,fill=black,regular polygon,regular polygon sides=3},
  mysymnode/.style={inner sep=0.5pt, fill=blue!30!white, rectangle,draw=black,thick, minimum width=1cm, minimum height=0.5cm, font=\scriptsize},
  myedge/.style={-{Latex[length=2mm, width=2mm]}},
  mylabel/.style={pos=0.6,font=\scriptsize},
  comp/.style = {rectangle, draw=black,inner sep=1pt,minimum width=40pt,minimum height=40pt,font=\large,text height = -0.3cm},
  plan/.style = {circle, draw=black,inner sep=0pt, minimum size = 20pt},
  diagonal fill/.style 2 args={fill=#2, path picture={
      \fill[#1, sharp corners] (path picture bounding box.south west) -|
  (path picture bounding box.north east) -- cycle;}}
}
\tikzset{lb/.style={font=\scriptsize}}

\author{Till Markus Hofmann}
\actualpostgrade{Master of Science}
\grade{eines Doktors der Naturwissenschaften}
\birthplace{Tübingen, Deutschland}
\title{Towards Bridging the Gap between High-Level Reasoning and Execution on Robots}
\date{20.~September 2023}
\expertI{Prof.~Gerhard Lakemeyer, Ph.D.}
\expertII{Prof.~Yves Lespérance, Ph.D.}

\usepackage[rgb,showlinks]{optional}

\opt{cmyk}{
\definecolor{rwth-blue}{cmyk}{1,.5,0,0}
\definecolor{rwth-lblue}{cmyk}{0.75,0.38,0,0}
\definecolor{rwth-llblue}{cmyk}{0.45,0.14,0,0}
\definecolor{rwth-lllblue}{cmyk}{0.23,0.07,0,0}
\definecolor{rwth-llllblue}{cmyk}{0.09,0.03,0,0}

\definecolor{rwth-black}{cmyk}{0,0,0,1}
\colorlet{rwth-lblack}{rwth-black!75}
\colorlet{rwth-llblack}{rwth-black!50}
\colorlet{rwth-lllblack}{rwth-black!25}
\colorlet{rwth-llllblack}{rwth-black!10}

\definecolor{rwth-magenta}{cmyk}{0,1,.25,0}
\colorlet{rwth-lmagenta}{rwth-magenta!75}
\colorlet{rwth-llmagenta}{rwth-magenta!50}
\colorlet{rwth-lllmagenta}{rwth-magenta!25}
\colorlet{rwth-llllmagenta}{rwth-magenta!10}

\definecolor{rwth-yellow}{cmyk}{0,0,1,0}
\colorlet{rwth-lyellow}{rwth-yellow!75}
\colorlet{rwth-llyellow}{rwth-yellow!50}
\colorlet{rwth-lllyellow}{rwth-yellow!25}
\colorlet{rwth-llllyellow}{rwth-yellow!10}

\definecolor{rwth-petrol}{cmyk}{1,0.3,0.5,0.3}
\colorlet{rwth-lpetrol}{rwth-petrol!75}
\colorlet{rwth-llpetrol}{rwth-petrol!50}
\colorlet{rwth-lllpetrol}{rwth-petrol!25}
\colorlet{rwth-llllpetrol}{rwth-petrol!10}

\definecolor{rwth-turquoise}{cmyk}{1,0,0.4,0}
\colorlet{rwth-lturquoise}{rwth-turquoise!75}
\colorlet{rwth-llturquoise}{rwth-turquoise!50}
\colorlet{rwth-lllturquoise}{rwth-turquoise!25}
\colorlet{rwth-llllturquoise}{rwth-turquoise!10}

\definecolor{rwth-green}{cmyk}{.7,0,1,0}
\colorlet{rwth-lgreen}{rwth-green!75}
\colorlet{rwth-llgreen}{rwth-green!50}
\colorlet{rwth-lllgreen}{rwth-green!25}
\colorlet{rwth-llllgreen}{rwth-green!10}

\definecolor{rwth-grass}{cmyk}{.35,0,1,0}
\colorlet{rwth-lgrass}{rwth-grass!75}
\colorlet{rwth-llgrass}{rwth-grass!50}
\colorlet{rwth-lllgrass}{rwth-grass!25}
\colorlet{rwth-llllgrass}{rwth-grass!10}

\definecolor{rwth-orange}{cmyk}{0,.4,1,0}
\colorlet{rwth-lorange}{rwth-orange!75}
\colorlet{rwth-llorange}{rwth-orange!50}
\colorlet{rwth-lllorange}{rwth-orange!25}
\colorlet{rwth-llllorange}{rwth-orange!10}

\definecolor{rwth-red}{cmyk}{.15,1,1,0}
\colorlet{rwth-lred}{rwth-red!75}
\colorlet{rwth-llred}{rwth-red!50}
\colorlet{rwth-lllred}{rwth-red!25}
\colorlet{rwth-llllred}{rwth-red!10}

\definecolor{rwth-burgundy}{cmyk}{0.25,1,0.7,0.2}
\colorlet{rwth-lburgundy}{rwth-burgundy!75}
\colorlet{rwth-llburgundy}{rwth-burgundy!50}
\colorlet{rwth-lllburgundy}{rwth-burgundy!25}
\colorlet{rwth-llllburgundy}{rwth-burgundy!10}

\definecolor{rwth-violet}{cmyk}{0.7,1,0.35,0.15}
\colorlet{rwth-lviolet}{rwth-violet!75}
\colorlet{rwth-llviolet}{rwth-violet!50}
\colorlet{rwth-lllviolet}{rwth-violet!25}
\colorlet{rwth-llllviolet}{rwth-violet!10}

\definecolor{rwth-purple}{cmyk}{0.6,0.6,0,0}
\colorlet{rwth-lpurple}{rwth-purple!75}
\colorlet{rwth-llpurple}{rwth-purple!50}
\colorlet{rwth-lllpurple}{rwth-purple!25}
\colorlet{rwth-llllpurple}{rwth-purple!10}

\definecolor{rwth-cyan}{cmyk}{1,0,.4,0}\colorlet{rwth-lcyan}{rwth-cyan!50}\colorlet{rwth-llcyan}{rwth-cyan!25}
\definecolor{rwth-teal}{cmyk}{1,.3,.5,.3}\colorlet{rwth-lteal}{rwth-teal!50}\colorlet{rwth-llteal}{rwth-teal!25}
\definecolor{rwth-silver}{cmyk}{.39,.31,.32,.14}
\definecolor{rwth-gold}{cmyk}{.35,.46,.7,.35}
}

\opt{rgb}{
\definecolor{rwth-blue}{RGB}{0,84,159}
\definecolor{rwth-lblue}{RGB}{64,127,183}
\definecolor{rwth-llblue}{RGB}{142,186,229}
\definecolor{rwth-lllblue}{RGB}{199,221,242}
\definecolor{rwth-llllblue}{RGB}{232,241,250}

\definecolor{rwth-black}{RGB}{0,0,0}
\definecolor{rwth-lblack}{RGB}{100,101,103}
\definecolor{rwth-llblack}{RGB}{156,158,159}
\definecolor{rwth-lllblack}{RGB}{208,209,210}
\definecolor{rwth-llllblack}{RGB}{236,237,237}

\definecolor{rwth-magenta}{RGB}{227,0,102}
\definecolor{rwth-lmagenta}{RGB}{233,96,136}
\definecolor{rwth-llmagenta}{RGB}{241,158,177}
\definecolor{rwth-lllmagenta}{RGB}{249,210,218}
\definecolor{rwth-llllmagenta}{RGB}{253,238,240}

\definecolor{rwth-yellow}{RGB}{255,237,0}
\definecolor{rwth-lyellow}{RGB}{255,240,85}
\definecolor{rwth-llyellow}{RGB}{255,245,155}
\definecolor{rwth-lllyellow}{RGB}{255,250,209}
\definecolor{rwth-llllyellow}{RGB}{255,253,,238}

\definecolor{rwth-petrol}{RGB}{0,97,101}
\definecolor{rwth-lpetrol}{RGB}{45,127,131}
\definecolor{rwth-llpetrol}{RGB}{125,164,167}
\definecolor{rwth-lllpetrol}{RGB}{191,208,209}
\definecolor{rwth-llllpetrol}{RGB}{230,236,236}

\definecolor{rwth-turquoise}{RGB}{0,152,161}
\definecolor{rwth-lturquoise}{RGB}{0,177,183}
\definecolor{rwth-llturquoise}{RGB}{137,204,207}
\definecolor{rwth-lllturquoise}{RGB}{202,231,231}
\definecolor{rwth-llllturquoise}{RGB}{235,246,246}

\definecolor{rwth-green}{RGB}{87,171,39}
\definecolor{rwth-lgreen}{RGB}{141,192,96}
\definecolor{rwth-llgreen}{RGB}{184,214,152}
\definecolor{rwth-lllgreen}{RGB}{221,235,206}
\definecolor{rwth-llllgreen}{RGB}{242,247,236}

\definecolor{rwth-grass}{RGB}{189,205,0}
\definecolor{rwth-lgrass}{RGB}{208,217,92}
\definecolor{rwth-llgrass}{RGB}{224,230,154}
\definecolor{rwth-lllgrass}{RGB}{240,43,208}
\definecolor{rwth-llllgrass}{RGB}{249,250,237}

\definecolor{rwth-orange}{RGB}{246,168,0}
\definecolor{rwth-lorange}{RGB}{250,190,80}
\definecolor{rwth-llorange}{RGB}{253,212,143}
\definecolor{rwth-lllorange}{RGB}{254,234,201}
\definecolor{rwth-llllorange}{RGB}{255,247,234}

\definecolor{rwth-red}{RGB}{204,7,30}
\definecolor{rwth-lred}{RGB}{216,92,65}
\definecolor{rwth-llred}{RGB}{230,150,121}
\definecolor{rwth-lllred}{RGB}{243,205,187}
\definecolor{rwth-llllred}{RGB}{250,235,227}

\definecolor{rwth-burgundy}{RGB}{161,16,53}
\definecolor{rwth-lburgundy}{RGB}{182,82,86}
\definecolor{rwth-llburgundy}{RGB}{205,139,135}
\definecolor{rwth-lllburgundy}{RGB}{229,197,192}
\definecolor{rwth-llllburgundy}{RGB}{245,232,229}

\definecolor{rwth-violet}{RGB}{97,33,88}
\definecolor{rwth-lviolet}{RGB}{131,78,117}
\definecolor{rwth-llviolet}{RGB}{168,133,158}
\definecolor{rwth-lllviolet}{RGB}{210,192,205}
\definecolor{rwth-llllviolet}{RGB}{237,229,234}

\definecolor{rwth-purple}{RGB}{122,111,172}
\definecolor{rwth-lpurple}{RGB}{122,111,172}
\definecolor{rwth-llpurple}{RGB}{122,111,172}
\definecolor{rwth-lllpurple}{RGB}{122,111,172}
\definecolor{rwth-llllpurple}{RGB}{122,111,172}

\definecolor{rwth-cyan}{RGB}{0,152,161}
\definecolor{rwth-lcyan}{RGB}{0,177,183}
\definecolor{rwth-llcyan}{RGB}{137,204,207}
\definecolor{rwth-lllcyan}{RGB}{202,231,231}
\definecolor{rwth-llllcyan}{RGB}{235,246,246}

\definecolor{rwth-silver}{cmyk}{.39,.31,.32,.14}
\definecolor{rwth-gold}{cmyk}{.35,.46,.7,.35}
}

\usepackage{amsmath}
\usepackage{amssymb}
\usepackage{mathtools}
\usepackage{relsize}
\usepackage{xspace}
\usepackage{mfirstuc}
\usepackage{ifthen}

\newcommand*{\strips}{STRIPS\xspace}

\newcommand*{\tacos}{TACoS\xspace}
\newcommand*{\gocos}{GoCoS\xspace}
\newcommand*{\gologpp}{\texttt{golog++}\xspace}

\newcommand*{\caesar}{\textsc{Caesar}\xspace}

\newcommand*{\cpp}{C\nolinebreak[4]\hspace{-.05em}\raisebox{.4ex}{\relsize{-3}{\textbf{++}}}\xspace}

\newcommand*{\acfip}[1]{\emph{\aclp{#1}}~(\acsp{#1})}

\newcommand*{\lrs}{\ensuremath{\: \Leftrightarrow \:}}
\newcommand*{\wolog}{without loss of generality\xspace}
\newcommand*{\Wolog}{Without loss of generality\xspace}
\newcommand*{\ctlstar}{CTL\textsuperscript{*}\xspace}

\newcommand*{\sitcalc}{situation calculus\xspace}
\newcommand*{\Sitcalc}{Situation Calculus\xspace}
\newcommand*{\act}[1]{\ensuremath{\mathit{#1}}\xspace}
\providecommand*{\action}[1]{\act{#1}}
\newcommand*{\rfluent}[1]{\ensuremath{\xcapitalisewords{\mathit{#1}}}\xspace}
\newcommand*{\ffluent}[1]{\ensuremath{\mathit{#1}}\xspace}
\newcommand*{\sconst}[1]{\ensuremath{\mathit{#1}}\xspace}
\DeclareMathOperator*{\doopsym}{do}
\newcommand*{\doop}{\ensuremath{\doopsym}\xspace}
\newcommand*{\pick}{\action{pick}}
\newcommand*{\putact}{\action{put}}
\newcommand*{\calibrate}{\action{calibrate}}
\newcommand*{\kitchen}{\sconst{kitchen}}
\newcommand*{\hallway}{\sconst{hallway}}
\newcommand*{\cupobj}{\sconst{cup}}
\newcommand*{\distance}{\ffluent{distance}}
\newcommand*{\connected}{\rfluent{connected}}
\newcommand*{\waitfor}{\action{waitFor}}
\newcommand*{\shelfpick}{\action{getFromShelf}}
\newcommand*{\pay}{\action{pay}}

\DeclareMathOperator*{\sstart}{start}
\DeclareMathOperator*{\actual}{Actual}
\DeclareMathOperator*{\occurs}{Occurs}
\DeclareMathOperator*{\occurst}{Occurs_T}

\newcommand*{\taloc}{\ffluent{loc}}
\DeclareMathOperator*{\labeltrace}{ltrace}

\newcommand*{\sitcalcknow}{\ensuremath{\mathbf{Knows}}\xspace}
\newcommand*{\sitcalcbel}{\ensuremath{\mathbf{Bel}}\xspace}
\newcommand*{\sense}{\action{sense}}
\newcommand*{\sensefun}{\action{read}}
\newcommand*{\sensefluent}{\ensuremath{\mathit{SF}}\xspace}

\newcommand*{\sonar}{\action{sonar}}
\newcommand*{\goto}{\action{goto}}
\newcommand*{\move}{\action{move}}

\newcommand*{\near}{\sconst{near}}
\newcommand*{\midpos}{\sconst{middle}}
\newcommand*{\far}{\sconst{far}}
\newcommand*{\loc}{\rfluent{loc}}
\newcommand*{\robotat}{\rfluent{robotAt}}
\newcommand*{\at}{\robotat}
\newcommand*{\objat}{\ffluent{objAt}}

\DeclareMathOperator*{\sym}{sym}
\DeclareMathOperator*{\untimedop}{\sym}
\newcommand*{\untimed}[1]{\untimedop(#1)}

\DeclareMathOperator*{\doprogsym}{Do}
\newcommand*{\doprog}{\ensuremath{\doprogsym}\xspace}
\DeclareMathOperator*{\gtransop}{Trans}
\newcommand*{\gtrans}{\ensuremath{\gtransop}\xspace}
\DeclareMathOperator*{\gfinalop}{Final}
\newcommand*{\gfinal}{\ensuremath{\gfinalop}\xspace}
\newcommand*{\false}{\textsc{False}\xspace}
\newcommand*{\pconc}{\ensuremath{\gg}\xspace}

\DeclarePairedDelimiter\abs{\lvert}{\rvert}%
\newcommand*{\mi}[1]{\ensuremath{\mathit{#1}}}
\newcommand*{\la}{\langle}
\newcommand*{\ra}{\rangle}

\makeatletter
\newsavebox{\@brx}
\newcommand{\llangle}[1][]{\savebox{\@brx}{\(\m@th{#1\langle}\)}%
  \mathopen{\copy\@brx\mkern2mu\kern-0.9\wd\@brx\usebox{\@brx}}}
\newcommand{\rrangle}[1][]{\savebox{\@brx}{\(\m@th{#1\rangle}\)}%
  \mathclose{\copy\@brx\mkern2mu\kern-0.9\wd\@brx\usebox{\@brx}}}
\makeatother

\newcommand*{\refcmd}[3]{\texorpdfstring{#2}{#3}\xspace}
\newcommand*{\es}{\refcmd{sec:es}{\ensuremath{\mathcal{E \negthinspace S}}}{ES}}
\newcommand*{\esg}{\refcmd{sec:esg}{\ensuremath{\mathcal{E \negthinspace S \negthinspace G}}}{ESG}}
\newcommand*{\tesg}{\refcmd{sec:esg}{\ensuremath{\operatorname{\mathit{t-}} \negthinspace \mathcal{E \negthinspace S \negthinspace G}}}{t-ESG}}
\newcommand*{\ds}{\refcmd{sec:ds}{\ensuremath{\mathcal{D \negthinspace S}}}{DS}}
\newcommand*{\dsg}{\refcmd{sec:dsg}{\ensuremath{\mathcal{D \negthinspace S \negthinspace G}}}{DSG}}
\newcommand*{\uppaal}{\textsc{Uppaal}\xspace}
\newcommand*{\golog}{\textsc{Golog}\xspace}
\newcommand*{\congolog}{\textsc{ConGolog}\xspace}
\newcommand*{\indigolog}{\textsc{IndiGolog}\xspace}
\newcommand*{\readylog}{\textsc{Readylog}\xspace}

\newcommand*{\mtl}{\ac{MTL}\xspace}
\newcommand*{\mtlzi}{$\text{MTL}_{0,\infty}$\xspace}

\newcommand*{\ta}{\ensuremath{A}\xspace}
\newcommand*{\tastates}{\ensuremath{L}\xspace}
\newcommand*{\tafstates}{\ensuremath{\tastates_F}\xspace}
\newcommand*{\tastate}{\ensuremath{l}\xspace}
\newcommand*{\taalph}{\ensuremath{\Sigma}\xspace}
\newcommand*{\taclocks}{\ensuremath{X}\xspace}
\newcommand*{\taswitches}{\ensuremath{E}\xspace}

\newcommand*{\talts}{\ensuremath{\lts_{\ta}}\xspace}
\newcommand*{\taltsstates}{\ensuremath{\ltsstates_{\ta}}\xspace}
\newcommand*{\taltsstate}{\ensuremath{q_{\ta}}\xspace}
\newcommand*{\taltsalph}{\ensuremath{\ltsalph}_{\ta}\xspace}
\newcommand*{\invs}{\ensuremath{I}\xspace}
\newcommand*{\clockvaluation}{\ensuremath{\nu}\xspace}
\newcommand*{\maxconst}{\ensuremath{K}}
\newcommand*{\clockequiv}[1][\maxconst]{\sim_{#1}}
\newcommand*{\regequiv}[1][\maxconst]{\cong_{#1}}
\newcommand*{\regequivclass}[2][\maxconst]{\lbrack #2 \rbrack_{#1}}
\newcommand*{\reglts}[1][\ta]{\ensuremath{\mathcal{R}(#1)}\xspace}
\newcommand*{\regltsstates}{\ensuremath{Q}\xspace}
\newcommand*{\regltsstate}{\ensuremath{q}\xspace}
\newcommand*{\regltsalph}{\ensuremath{\Sigma}\xspace}
\newcommand*{\regltstrans}[1][]{\ensuremath{\xhookrightarrow{#1}}}

\newcommand*{\taltstrans}[2]{\ensuremath{\xrightarrow[#1]{#2}}\xspace}

\DeclareMathOperator{\runs}{Runs}
\newcommand*{\accruns}{\runs_F}
\newcommand*{\finruns}{\runs^*}
\newcommand*{\infruns}{\runs^\omega}
\newcommand*{\accfinruns}{\runs^*_F}
\newcommand*{\accinfruns}{\runs^\omega_F}
\newcommand*{\lang}{\ensuremath{\mathcal{L}}\xspace}

\DeclareMathOperator{\clockconstraints}{\Phi}

\DeclareMathOperator*{\init}{init}
\DeclareMathOperator*{\closure}{cl}

\newcommand*{\controllertraces}{\ensuremath{\traces_{\controller}}\xspace}

\newcommand*{\smid}{\ensuremath{\:\mid\:}}
\newcommand*{\realpos}{\ensuremath{\mathbb{R}_{\geq 0}}\xspace}
\newcommand*{\rationals}{\ensuremath{\mathbb{Q}_{\geq 0}}\xspace}
\newcommand*{\naturals}{\ensuremath{\mathbb{N}}\xspace}
\newcommand*{\fluents}{\ensuremath{\mathcal{F}}\xspace}
\newcommand*{\fluentset}{\ensuremath{F}\xspace}
\newcommand*{\regress}{\ensuremath{\mathcal{R}}\xspace}
\newcommand*{\worlds}{\ensuremath{\mathcal{W}}\xspace}
\newcommand*{\qo}{\ac{qo}\xspace}
\newcommand*{\wqo}{\ac{wqo}\xspace}
\newcommand*{\bqo}{\ac{bqo}\xspace}
\DeclareMathOperator{\poss}{Poss}
\newcommand*{\eqdef}{\ensuremath{:=}}
\newcommand*{\equivspace}{\ensuremath\,\equiv\;}
\newcommand*{\eqspace}{\ =\ }

\DeclareMathOperator{\plays}{plays}
\DeclareMathOperator{\clockconstraint}{g}
\DeclareMathOperator{\reset}{reset}

\DeclareMathOperator{\ztime}{time}
\DeclareMathOperator{\start}{start}

\newcommand*{\zeroclocks}{\ensuremath{\vec{0}}\xspace}
\newcommand*{\clockset}{\ensuremath{\mathcal{C}}\xspace}
\newcommand*{\ata}[1][\phi]{\ensuremath{\mathcal{A}\ifthenelse{\equal{#1}{}}{}{}{_{#1}}}\xspace}
\newcommand*{\ataalphabet}{\ensuremath{\Sigma}\xspace}
\newcommand*{\atalocations}{\ensuremath{L}\xspace}
\newcommand*{\atafinallocations}{\ensuremath{F}\xspace} 
\newcommand*{\atalocformulas}[1][\atalocations]{\ensuremath{\Phi(#1)}}
\newcommand*{\atatrans}{\ensuremath{\eta}\xspace}
\newcommand*{\atastates}{\ensuremath{S_{\ata[]}}\xspace}
\newcommand*{\ataconf}{\ensuremath{G}\xspace}

\newcommand*{\ataconfs}{\ensuremath{\mathcal{G}}\xspace}
\newcommand*{\atalts}[1][]{\ensuremath{\lts_{\ata[#1]}}\xspace}
\newcommand*{\ataltsstates}[1][]{\ensuremath{\ltsstates_{\ata[#1]}}\xspace}
\newcommand*{\ataltsalph}[1][]{\ensuremath{\ltsalph}_{\ata[#1]}\xspace}
\newcommand*{\ataltsstate}[1][]{\ensuremath{q_{\ata[#1]}}\xspace}
\newcommand*{\ataltstrans}[2]{\ensuremath{\xrightarrow[#1]{#2}}\xspace}
\newcommand*{\atatranstime}[1]{\overset{#1}{\rightsquigarrow}}
\newcommand*{\atatranssym}[1]{\xrightarrow{#1}}
\DeclareMathOperator{\suc}{Succ}
\newcommand*{\controller}{\ensuremath{\mi{CR}}\xspace}
\newcommand{\warrow}[1][w]{\ensuremath{\xrightarrow{#1}}}
\newcommand{\wtarrow}[1][d]{\ensuremath{\xrightarrow[#1]{}}}
\newcommand{\wsarrow}[1][p]{\ensuremath{\xrightarrow[#1]{}}}
\newcommand*{\bat}{\ensuremath{\Sigma}\xspace}
\newcommand*{\batactions}[1][\bat]{\ensuremath{A_{#1}}\xspace}
\newcommand*{\pre}{\ensuremath{\text{pre}}}
\newcommand*{\post}{\ensuremath{\text{post}}}
\newcommand*{\final}[1][w]{\ensuremath{\mathcal{F}^{#1}}}
\newcommand*{\sac}[1]{\ensuremath{\action{start}(#1)}\xspace}
\newcommand*{\eac}[1]{\ensuremath{\action{end}(#1)}\xspace}
\newcommand*{\traces}{\ensuremath{\mathcal{Z}}\xspace}
\newcommand*{\inftraces}{\ensuremath{\Pi}\xspace}
\newcommand*{\fintraces}{\traces} 
\newcommand*{\alltraces}{\ensuremath{\mathcal{T}}\xspace}
\DeclareMathOperator{\sub}{sub}
\newcommand*{\syncbisim}{\ensuremath{\approx}\xspace}

\newcommand*{\sboot}{\ensuremath{\mathit{sb}}}
\newcommand*{\sgrasp}{\ensuremath{\mathit{sg}}}

\newcommand*{\eboot}{\ensuremath{\mathit{eb}}}
\newcommand*{\egrasp}{\ensuremath{\mathit{eg}}}

\newcommand*{\symtrace}{\ensuremath{s}\xspace}
\newcommand*{\synclts}{\ensuremath{\mathcal{S}_{\Delta / \phi}}\xspace}
\newcommand*{\syncstates}{\ensuremath{S_{\Delta / \phi}}\xspace}
\newcommand*{\syncstate}{\ensuremath{s}\xspace}
\newcommand*{\synctranstime}[1]{\ensuremath{\xrightarrow{#1}}\xspace}
\newcommand*{\synctranssym}[1]{\ensuremath{\xrightarrow{#1}}\xspace}
\newcommand*{\synctrans}[2][]{\ensuremath{\xrightarrow[#1]{#2}}\xspace}
\newcommand*{\synctransfull}[2][]{\ensuremath{\synctrans[#1]{#2}_{\Delta, \phi}}\xspace}
\newcommand*{\syncclocks}{\ensuremath{C}}
\newcommand*{\syncconfs}{\ensuremath{\mathcal{C}_{\Delta / \phi}}\xspace}
\newcommand*{\syncconf}{C}

\newcommand*{\lts}{\ensuremath{\mathcal{S}}\xspace}
\newcommand*{\ltsstates}{\ensuremath{Q}\xspace}
\newcommand*{\ltsfinalstates}{\ensuremath{F}\xspace}
\newcommand*{\ltsstate}{\ensuremath{q}\xspace}
\newcommand*{\ltsalph}{\ensuremath{\Sigma}\xspace}
\newcommand*{\ltstrans}[1][]{\ensuremath{\xrightarrow{#1}}}
\newcommand*{\abslts}{\ensuremath{\mathcal{W}_{\Delta / \phi}}\xspace}
\newcommand*{\absstates}{\ensuremath{W}\xspace}
\newcommand*{\absstate}{\ensuremath{w}\xspace}
\newcommand*{\abstrans}[2][]{\ensuremath{\lhook\joinrel\synctrans[#1]{#2}}\xspace}
\newcommand*{\abstransfull}[2][]{\ensuremath{\abstrans[#1]{#2}_{\Delta, \phi}}\xspace}
\newcommand*{\abstranstime}[2][]{\abstrans[#1]{#2}}
\newcommand*{\abstranssym}[2][]{\abstrans[#1]{#2}}

\newcommand*{\detregtrans}[2][]{\ensuremath{\xRightarrow[#1]{#2}}\xspace}
\newcommand*{\dettrans}[2][]{\detregtrans[#1]{#2}}
\newcommand*{\dettransfull}[2][]{\ensuremath{\Rightarrow_{\Delta / \phi}}\xspace}
\newcommand*{\detlts}{\ensuremath{\mathcal{D}\abslts}\xspace}
\newcommand*{\detdisq}{\detlts}
\newcommand*{\detstates}{\ensuremath{\mi{DS}_{\Delta / \phi}}\xspace}
\newcommand*{\detstate}{\ensuremath{c}\xspace}

\newcommand*{\wsts}{\ensuremath{\mathcal{W}}\xspace}
\newcommand*{\game}{\ensuremath{\mathbb{G}}\xspace}
\newcommand*{\tracestate}{\ensuremath{\mi{state}_{\detlts}}\xspace}
\newcommand*{\synctracestates}{\ensuremath{\mi{states}_{\synclts}}\xspace}

\newcommand*{\regions}[1][K]{\ensuremath{\text{REG}_{#1}}\xspace}
\DeclareMathOperator{\fract}{fract}
\DeclareMathOperator{\absword}{abs}
\DeclareMathOperator{\reg}{reg}
\DeclareMathOperator{\tw}{tw}
\newcommand*{\twsym}{\ensuremath{\rho}\xspace}
\newcommand*{\twtime}{\ensuremath{\tau}\xspace}

\newcommand*{\powerset}[1]{\ensuremath{\wp(#1)}}

\DeclareMathOperator{\rnext}{next}
\DeclareMathOperator{\rincr}{incr}

\newcommand*{\lpowleq}{\ensuremath{\sqsubseteq}\xspace}
\newcommand*{\syncleq}{\ensuremath{\leq}\xspace}
\newcommand*{\absleq}{\syncleq}
\newcommand*{\detleq}{\ensuremath{\lpowleq}\xspace}

\newcommand*{\tnext}[1]{\ensuremath{\mathbf{X}_{#1}}\,}
\newcommand*{\until}[1]{\ensuremath{\,\mathbf{U}\ifthenelse{\equal{#1}{}}{}{_{#1}}\,}}
\newcommand*{\since}[1]{\ensuremath{\,\mathbf{S}_{#1}}\,}
\newcommand*{\duntil}[1]{\ensuremath{\,\widetilde{\mathbf{U}}\ifthenelse{\equal{#1}{}}{}{_{#1}}\,}}
\newcommand*{\fut}[1]{\ensuremath{\mathbf{F}_{#1}}\if#1{\,}\fi}
\newcommand*{\finally}[1]{\fut{#1}}
\newcommand*{\glob}[1]{\ensuremath{\mathbf{G}_{#1}}\if#1{\,}\fi}
\newcommand*{\globally}[1][]{\glob{#1}}

\newcommand*{\subdefref}[2]{\hyperref[{#2}]{\autoref*{#1}.\ref*{#2}}}

\DeclareMathOperator*{\nilop}{nil}
\newcommand*{\nil}{\ensuremath{\nilop}\xspace}
\newcommand*{\forallobj}[1]{\ensuremath{\forall #1 \mathbf{:} o}}

\newcommand*{\drive}{\action{drive}}
\newcommand*{\grasp}{\action{grasp}}
\newcommand*{\bootcam}{\action{bootCamera}}
\newcommand*{\stopcam}{\action{stopCamera}}

\newcommand*{\camon}{\rfluent{camOn}\xspace}
\newcommand*{\camoff}{\rfluent{camOff}\xspace}
\newcommand*{\holding}{\rfluent{holding}\xspace}
\newcommand*{\grasping}{\rfluent{grasping}\xspace}

\newcommand*{\performing}[1]{\ensuremath{\rfluent{perf}(#1)}\xspace}
\newcommand*{\oat}{\rfluent{objAt}\xspace}
\newcommand*{\rat}{\rfluent{robotAt}\xspace}

\newcommand*{\phibad}{\phi_{\mi{bad}}}
\newcommand*{\moving}{\rfluent{moving}\xspace}
\newcommand*{\raligned}{\rfluent{aligned}\xspace}

\newcommand*{\stdname}{\ensuremath{\mathcal{N}}\xspace}
\newcommand*{\stdobjname}{\ensuremath{\mathcal{N}_O}\xspace}
\newcommand*{\stdactname}{\ensuremath{\mathcal{N}_A}\xspace}
\newcommand*{\stdclockname}{\ensuremath{\mathcal{N}_C}\xspace}
\newcommand*{\stdtimename}{\ensuremath{\mathcal{N}_T}\xspace}

\newcommand*{\primobjs}{\ensuremath{\mathcal{P}_O}\xspace}
\newcommand*{\primactions}{\ensuremath{\mathcal{P}_A}\xspace}
\newcommand*{\primclocks}{\ensuremath{\mathcal{P}_C}\xspace}
\newcommand*{\primterms}{\ensuremath{\mathcal{P}}\xspace}
\newcommand*{\primformulas}{\ensuremath{\mathcal{P}_F}\xspace}
\newcommand*{\denot}[3]{\lvert #1 \rvert^{#2}_{#3}}

\newcommand*{\occ}{\rfluent{occ}}

\newcommand*{\taplan}{\ensuremath{\ta_\sigma}\xspace}
\newcommand*{\taplatform}{\ensuremath{\ta_M}\xspace}

\newcommand*{\tacontext}{\ensuremath{\ta_{\texttt{context}}}\xspace}
\newcommand*{\tachain}{\ensuremath{\ta_{\mathfrak{uc}}}\xspace}
\newcommand*{\taenc}{\ensuremath{\ta_{\mathfrak{enc}}}\xspace}
\newcommand*{\planorder}{\ensuremath{\mathit{PlanOrder}}\xspace}
\newcommand*{\fin}{\ensuremath{l_n}\xspace}
\newcommand*{\clockabs}{\ensuremath{x_\texttt{abs}}\xspace}
\newcommand*{\clockrel}{\ensuremath{x}\xspace}
\newcommand*{\relconstraint}{\ensuremath{\mathfrak{rel}}\xspace}
\newcommand*{\relconstraints}{\ensuremath{C_{\mathfrak{rel}}}\xspace}
\newcommand*{\absconstraint}{\ensuremath{\mathfrak{abs}}\xspace}
\newcommand*{\absconstraints}{\ensuremath{C_\mathfrak{abs}}\xspace}
\newcommand*{\chainconstraint}{\ensuremath{\mathfrak{uc}}\xspace}
\newcommand*{\chainconstraints}{\ensuremath{C_\mathfrak{uc}}\xspace}

\newcommand*{\fancy}{\ensuremath{\mathcal}}

\newcommand{\PlanOrder}{\textit{PlanOrder}}

\newcommand\sgo{\sac{\goto(l_1)}}
\newcommand\ego{\eac{\goto(l_1)}}
\newcommand\spi{\sac{\pick(o_1)}}
\newcommand\epi{\eac{\pick(o_1)}}
\newcommand\mso{\camoff}

\newcommand\msr{\camon}

\newcommand*{\ewarrow}{\ensuremath{\overset{e,w}{\longrightarrow}}}
\newcommand*{\ewfinal}{\ensuremath{\mathcal{F}^{e,w}}}

\newcommand*{\bhl}{BHL\xspace}
\DeclareMathOperator{\exec}{exec}
\newcommand*{\belconnector}{\ensuremath{\mathbf{B}}\xspace}
\newcommand*{\obelconnector}{\ensuremath{\mathbf{O}}\xspace}
\newcommand*{\bel}[2]{\ensuremath{\belconnector\mleft(#1\,\mathbf{:}\,#2\mright)}}

\newcommand*{\know}[1]{\ensuremath{\mathbf{K}#1}}

\newcommand*{\oicomp}{\ensuremath{\approx_{\textrm{oi}}}\xspace}
\newcommand*{\oi}{\ensuremath{\mathit{oi}}\xspace}
\newcommand*{\oisim}[1][w]{\sim_{#1}}
\newcommand*{\whfull}[2]{\ensuremath{\stateset^{#1}_{#2}}}
\newcommand*{\wh}[2][]{\whfull{e_h, w_h#1, z_h#1}{#2}}
\newcommand*{\wlfull}[2]{\ensuremath{\stateset^{#1}_{\ifthenelse{\equal{#2}{\true}}{#2}{m\mleft(#2\mright)}}}}
\newcommand*{\wl}[2][]{\wlfull{e_l, w_l#1, z_l#1}{#2}}
\newcommand*{\oiso}[1][m]{\sim_{#1}}
\newcommand*{\eiso}{\sim_{e}}
\newcommand*{\bisim}[1][m]{\sim_{#1}}

\newcommand*{\lb}{\mleft \lbrack}
\newcommand*{\rb}{\mright \rbrack}

\newcommand*{\norm}{\textsc{Norm}\xspace}
\newcommand*{\eq}{\textsc{Eq}\xspace}
\newcommand*{\bnd}{\textsc{Bnd}\xspace}
\newcommand*{\true}{\textsc{True}\xspace}

\newcommand*{\rigid}{\ensuremath{\mathcal{R}}\xspace}
\newcommand*{\atoms}{\ensuremath{\mathcal{P}}\xspace}

\newcommand*{\states}{\ensuremath{\mathcal{S}}\xspace}
\newcommand*{\stateset}{\ensuremath{S}\xspace}

\newcommand*{\gwhile}{\ensuremath{\;\mathbf{while}\;}}
\newcommand*{\gdo}{\ensuremath{\;\mathbf{do}\;}}
\newcommand*{\gdone}{\ensuremath{\;\mathbf{done}\;}}
\newcommand*{\gif}{\ensuremath{\;\mathbf{if}\;}}
\newcommand*{\gthen}{\ensuremath{\;\mathbf{then}\;}}
\newcommand*{\gelif}{\ensuremath{\;\mathbf{elif}\;}}
\newcommand*{\gelse}{\ensuremath{\;\mathbf{else}\;}}
\newcommand*{\gfi}{\ensuremath{\;\mathbf{fi}\;}}
\newcommand*{\gsearch}{\ensuremath{\Sigma}}
\newcommand*{\sigmah}{\ensuremath{\Sigma_{\mi{goto}}}\xspace}
\newcommand*{\sigmal}{\ensuremath{\Sigma_{\mi{move}}}\xspace}

\hypersetup{
  pdfauthor = {\theauthor},
  pdftitle = {\thetitle},
  pdfsubject = {Ph.D.~Thesis},
  pdfkeywords = {Artificial Intelligence; Knowledge Representation; Cognitive Robotics; Situation Calculus; Golog; Metric Temporal Logic; Verification; Synthesis; Timed Systems; Stochastic Actions; Belief-based Programs}
}

\opt{showlinks}{
  \hypersetup{
    colorlinks,
    linkcolor={rwth-blue},
    citecolor={rwth-blue},
    urlcolor={rwth-blue}
  }
}
\opt{hidelinks}{
  \hypersetup{
    hidelinks
  }
}

\DeclareBibliographyCategory{contributions}
\addtocategory{contributions}{
  haberingUsingPlatformModels2021, hofmannAbstractingNoisyRobot2023a, hofmannConstraintbasedOnlineTransformation2018, hofmannContinualPlanningExecution2015, hofmannContinualPlanningGolog2016, hofmannControllingGologPrograms2022, hofmannControllingTimedAutomata2023, hofmannEnhancingSoftwareHardware2018, hofmannGeneratingMacroActions2017, hofmannInitialResultsGenerating2017, hofmannLogicSpecifyingMetric2018, hofmannMacroOperatorSynthesis2020, hofmannMultiagentGoalReasoning2021, hofmannTACoSToolMTL2021, hofmannUsingAbstractionInterpretable2022, hofmannWinningRoboCupLogistics2019, matarePortableHighlevelAgent2021, niemuellerCLIPSbasedExecutionPDDL2018, niemuellerGoalReasoningCLIPS2019, swobodaUsingPromisesMultiagent2022, viehmannTransformingRoboticPlans2021, matareConstraintbasedPlanTransformation2020,hofmannControllerSynthesisGolog2020}
\nocite{
  haberingUsingPlatformModels2021, hofmannAbstractingNoisyRobot2023a, hofmannConstraintbasedOnlineTransformation2018, hofmannContinualPlanningExecution2015, hofmannContinualPlanningGolog2016, hofmannControllingGologPrograms2022, hofmannControllingTimedAutomata2023, hofmannEnhancingSoftwareHardware2018, hofmannGeneratingMacroActions2017, hofmannInitialResultsGenerating2017, hofmannLogicSpecifyingMetric2018, hofmannMacroOperatorSynthesis2020, hofmannMultiagentGoalReasoning2021, hofmannTACoSToolMTL2021, hofmannUsingAbstractionInterpretable2022, hofmannWinningRoboCupLogistics2019, matarePortableHighlevelAgent2021, niemuellerCLIPSbasedExecutionPDDL2018, niemuellerGoalReasoningCLIPS2019, swobodaUsingPromisesMultiagent2022, viehmannTransformingRoboticPlans2021, matareConstraintbasedPlanTransformation2020,hofmannControllerSynthesisGolog2020}

\begin{document}

\maketitle

\begin{abstract}
\pdfbookmark{Abstract}{abstract}
  When reasoning about actions, e.g., by means of task planning or agent programming with \golog, the robot's actions are typically modeled on an abstract level, where complex actions such as picking up an object are treated as atomic primitives with deterministic effects and preconditions that only depend on the current state.
  However, when executing such an action on a robot it can no longer be seen as a primitive.
  Instead, action execution is a complex task involving multiple steps with additional temporal preconditions and timing constraints.
  Furthermore, the action may be noisy, e.g., producing erroneous sensing results and not always having the desired effects.
  While these aspects are typically ignored in reasoning tasks, they need to be dealt with during execution.
  In this thesis, we propose several approaches towards closing this gap.

  Based on a logic that combines the situation calculus with metric time and metric temporal logic,
  we model the robot platform with timed automata and temporal constraints to describe the connection between the high-level actions and the robot platform.
  We then describe two approaches towards transforming the high-level program.
  First, we view the transformation as a synthesis problem, where the task is to synthesize a controller that executes the program while satisfying the specification, independent of the environment's choices.
  We show that the synthesis problem is decidable, describe an algorithm to construct a controller, and evaluate the approach in two robotics scenarios.
   While this approach supports controlling arbitrary \golog programs against any specification with timing constraints, it does not scale well.
  For this reason, we describe a second approach based on some simplifying assumptions which allow us to view the transformation problem as a reachability problem on timed automata, which can be solved with state-of-the-art tools.
  We demonstrate the effectiveness and scalability of the approach in a number of scenarios.

  Finally, we turn towards noisy sensors and effectors.
  Based on \ds, a probabilistic variant of the situation calculus that allows modeling the agent's degree of belief, we describe an abstraction framework for \golog programs with noisy actions.
  In this framework, a high-level and non-stochastic program is mapped to a more detailed and stochastic low-level program.
  As the high-level program is non-stochastic, we may use non-probabilistic reasoning methods such as task planning or classical \golog program execution.
  At the same time, by mapping the abstract actions to low-level programs, we may still deal with uncertainty during execution.
  We define a suitable notion of bisimulation that guarantees the equivalence between the high-level and low-level programs and demonstrate the approach with an example.
\end{abstract}

\begin{otherlanguage}{ngerman}

\hyphenation{Golog}
\begin{abstract}
\pdfbookmark{Zusammenfassung}{zusammenfassung}
Beim klassichen Schließen über Aktionen, z.~B.\ durch Planung oder Agentenprogrammierung mit \golog, werden die Aktionen des Roboters typischerweise auf einer abstrakten Ebene modelliert, wobei komplexe Aktionen wie das Greifen eines Objekts als atomare Primitive mit deterministischen Effekten und Vorbedingungen behandelt werden, die nur vom aktuellen Zustand abhängen.
Wird eine solche Aktion jedoch von einem Roboter ausgeführt, kann sie nicht mehr als atomar betrachtet werden.
Stattdessen ist jede Aktion eine komplexe Aufgabe, die mehrere Schritte mit zusätzlichen zeitlichen Nebenbedingungen umfasst.
Außerdem kann sie mit Rauschen behaftet sein, z.~B.\ fehlerhafte Sensormessungen liefern und nicht immer die gewünschten Effekte haben.
Während diese Aspekte meist bei Planungsaufgaben ignoriert werden, müssen sie bei der Ausführung berücksichtigt werden.
Diese Arbeit schlägt mehrere Ansätze vor, um diese Lücke zu schließen.

Basierend auf einer Logik, die den Situationskalkül mit metrischer Zeit und metrischer temporaler Logik kombiniert, modellieren wir die Roboterplattform mit Zeitautomaten und zeitlichen Nebenbedingungen um den Zusammenhang zwischen den abstrakten Aktionen und der Roboterplattform zu beschreiben.
Anschließend beschreiben wir zwei Ansätze um das abstrakte Programm zu transformieren.  
Zunächst betrachten wir die Transformation als Syntheseproblem, bei dem die Aufgabe darin besteht, einen Regler zu synthetisieren, der das Programm ausführt und dabei die Spezifikation erfüllt, unabhängig von den Entscheidungen der Umwelt.
Wir zeigen, dass das Syntheseproblem entscheidbar ist, beschreiben einen Algorithmus zur Konstruktion eines Reglers und evaluieren den Ansatz in zwei Robotikszenarien.
Während dieser Ansatz die Steuerung beliebiger \golog-Programme gegen eine Spezifikation mit zeitlichen Nebenbedingungen unterstützt, ist er nicht gut skalierbar.
Deswegen beschreiben wir einen zweiten Ansatz, der auf einigen vereinfachenden Annahmen beruht und uns erlaubt, das Transformationsproblem als ein Erreichbarkeitsproblem auf Zeitautomaten zu betrachten, das mit etablierten Methoden gelöst werden kann.
Wir demonstrieren die Effektivität und Skalierbarkeit des Ansatzes in mehreren Szenarien.

Schließlich wenden wir uns verrauschten Sensoren und Effektoren zu.
Auf der Grundlage von \ds, einer probabilistischen Variante des Situationskalküls, beschreiben wir einen Abstraktionsmechanismus für \golog-Programme.
In diesem System wird ein abstraktes und möglicherweise nicht-stochastisches Programm auf ein detaillierteres und stochastisches Programm abgebildet.
Da das abstrake Programm nicht stochastisch ist, können wir nicht-probabilistische Schlussfolgerungsmethoden wie Planung oder klassische \golog-Programmausführung verwenden.
Indem wir die abstrakten Aktionen auf detaillierte Programme abbilden, können wir dabei mit der Unsicherheit während der Ausführung umgehen.
Wir definieren einen geeigneten Begriff der Bisimulation, der die Äquivalenz der beiden Programme garantiert, und demonstrieren den Ansatz anhand eines Beispiels.
\end{abstract}
\end{otherlanguage}

\thispagestyle{empty}
\renewcommand*{\abstractname}{Acknowledgements}
\begin{abstract}
\pdfbookmark{Acknowledgements}{acknowledgements}
First and foremost, I would like to thank Gerhard Lakemeyer for the great supervision of my thesis.
His input was always invaluable; quite often, a single question by him would give me an entirely new perspective on the problem at hand.
He was also very accommodating and supportive, which allowed me to do both a research visit and an internship during my Ph.D., which was not always easy to organize.
Moreover, he gave me the opportunity to work both on practical problems and theoretical questions, which helped me develop my own research interests.

I am thankful to Yves Lespérance for serving as second examiner and reading as well as reviewing my thesis.
His comments and the discussions during the examination were very helpful.
I also want to thank Erika Ábrahám and Martin Grohe for spending their valuable time for serving on my thesis committee.
Erika Ábrahám also served as my second supervisor and provided great feedback on the general progress of my thesis.

I want to thank Vaishak Belle, who hosted me as a visitor at the University of Edinburgh, provided the main ideas for \autoref{chap:abstraction}, and who was patient with me while I was finalizing our joint work, which took me way too long.
Special thanks also go to Tim Niemueller, who supervised both my Bachelor's and Master's thesis, gave me the opportunity of an internship, and taught me much about programming and writing, which I benefit from to this day.
I am indebted to Stefan Schupp, who helped me understand the intricacies of timed systems,
our countless pair-programming sessions via Zoom
were always a highlight.
I am grateful for the support by Jens Claßen, whose work helped me understand many foundational concepts essential for this thesis.
The countless discussions with Victor Mataré, Stefan Schiffer, and Alexander Ferrein helped shaping key ideas of this thesis, which I value highly.
I also want to thank Tarik Viehmann, whose Master's thesis is the basis of \autoref{chap:transformation-as-reachability-problem}, as well as my other thesis students Daniel Habering, Daniel Swoboda, Mostafa Gomaa, and Matteo Tschesche, whose work helped me better understand various aspects of reasoning about actions on robots.
It was a pleasure working with all the members of the Carologistics RoboCup team, whose tireless efforts lead to great success and also heavily influenced my work.
I am also grateful for the RTG UnRAVeL and all its members; the many talks and seminars helped me improve my research and presentation skills and allowed me to develop a broader understanding of related research.
Special thanks to Joost-Pieter Katoen, Helen Bolke-Hermanns, and Birgit Willms for shaping UnRAVeL into a thriving RTG.

I greatly appreciate the support of my family, who accepted me disappearing for weeks if not months and who were still willing to lend me an ear whenever the need occurred.
Sharing my Ph.D.~journey with my friends and roommates helped me deal with frustration and allowed me to celebrate successes, which I value greatly.
I am also indebted to Susanne Binder, who helped me cope with many challenges during this time.

Last but not least, I owe a thousand \emph{thank you}s to my beloved partner Karin for her continuous support and understanding.
Without her insistence, parts of this thesis would have never seen the light of day.%
\enlargethispage*{\baselineskip}
\end{abstract}
\renewcommand*{\abstractname}{Abstract}

\pdfbookmark{Table of Contents}{contents}
\tableofcontents

\chapter*{Acronyms}
\begin{acronym}[WSTS]
\acro{SSA}{successor state axiom\acroextra{ (introduced in \autoref{sec:sitcalc-bats})}}
\acro{BAT}{basic action theory\acroextra{ (introduced in \autoref{sec:sitcalc-bats} and \autoref{sec:tesg-bat})}}
\acro{LTS}{labeled transition system\acroextra{ (introduced in \autoref{sec:lts})}}
\acroindefinite{LTS}{an}{a}
\acrodef{CTL}{computation tree logic}
\acro{LTL}{Linear Temporal Logic\acroextra{ (introduced in \autoref{sec:temporal-logics})}}
\acroindefinite{LTL}{an}{a}
\acrodef{LTLf}{LTL over finite traces}
\acroindefinite{LTLf}{an}{a}
\acro{MTL}{Metric Temporal Logic\acroextra{ (introduced in \autoref{sec:mtl})}}
\acroindefinite{MTL}{an}{a}
\acrodef{MITL}{Metric Interval Temporal Logic}
\acroindefinite{MITL}{an}{a}
\acro{TA}{timed automaton\acroextra{ (introduced in \autoref{sec:ta})}}
\acroplural{TA}[TAs]{timed automata}
\acro{ATA}{alternating timed automaton\acroextra{ (introduced in \autoref{sec:ata})}}
\acroplural{ATA}[ATAs]{alternating timed automata}
\acroindefinite{ATA}{an}{an}
\acroplural{BAT}[BATs]{basic action theories}
\acro{WSTS}{well-structured transition system\acroextra{ (introduced in \autoref{sec:wsts})}}
\acro{qo}{quasi-ordering\acroextra{ (introduced in \autoref{sec:wsts})}}
\acro{wqo}{well-quasi-ordering\acroextra{ (introduced in \autoref{sec:wsts})}}
\acro{bqo}{better-quasi-ordering\acroextra{ (introduced in \autoref{sec:wsts})}}
\acrodef{HTN}{hierarchical task network}
\acrodef{RCLL}{RoboCup Logistics League}
\acroindefinite{RCLL}{an}{a}
\acrodef{ICP}{iterative closest point}
\acrodef{PDDL}{Planning Domain Definition Language}
\acrodef{IPC}{International Planning Competition}
\acrodef{TAL}{Temporal Action Logic}
\end{acronym}


\addtocontents{lof}{\protect\enlargethispage*{3\baselineskip}}

\listoffigures
\listoftables
\listofalgorithms
\lstlistoflistings

%

\chapter{Introduction}\label{chap:introduction}

Solving a task on a mobile robot involves dozens of sub-tasks that include low-level sensing and control such as recognizing objects or moving a robot arm to a desired pose, search procedures such as finding a path to a goal position, as well as high-level deliberation to decide which actions to pursue to accomplish a given goal.
The latter is the focus of \emph{cognitive robotics}~\parencite{levesqueCognitiveRobotics2008}, which is ``the study of the knowledge representation and reasoning problems faced by an autonomous robot (or agent) in a dynamic and incompletely known  world''~\parencite{levesqueHighlevelRoboticControl1998}.
In contrast to conventional robotics, the goal is high-level robotic control, where the agent operates on some abstract representation of the world and reasons about its capabilities to pursue its goals.
In this context, the robot's capabilities are typically modeled on an abstract level, where complex procedures such as moving to a location or grasping an object are captured by atomic actions.
The implementation of those atomic actions is assumed to be provided by the underlying system and are of no concern to the high-level reasoner.
Furthermore, the action's pre- and postcondition are completely captured by the model.
Hence, as long as the precondition is satisfied, performing the action is guaranteed to have the desired effects.
Details of the robot platform are irrelevant for the high-level behavior.

\begin{figure}[htb]
  \centering
  \begin{subfigure}[c]{0.39\textwidth}
  \begin{center}
    \includegraphics[width=\textwidth,height=0.3\textheight,keepaspectratio]{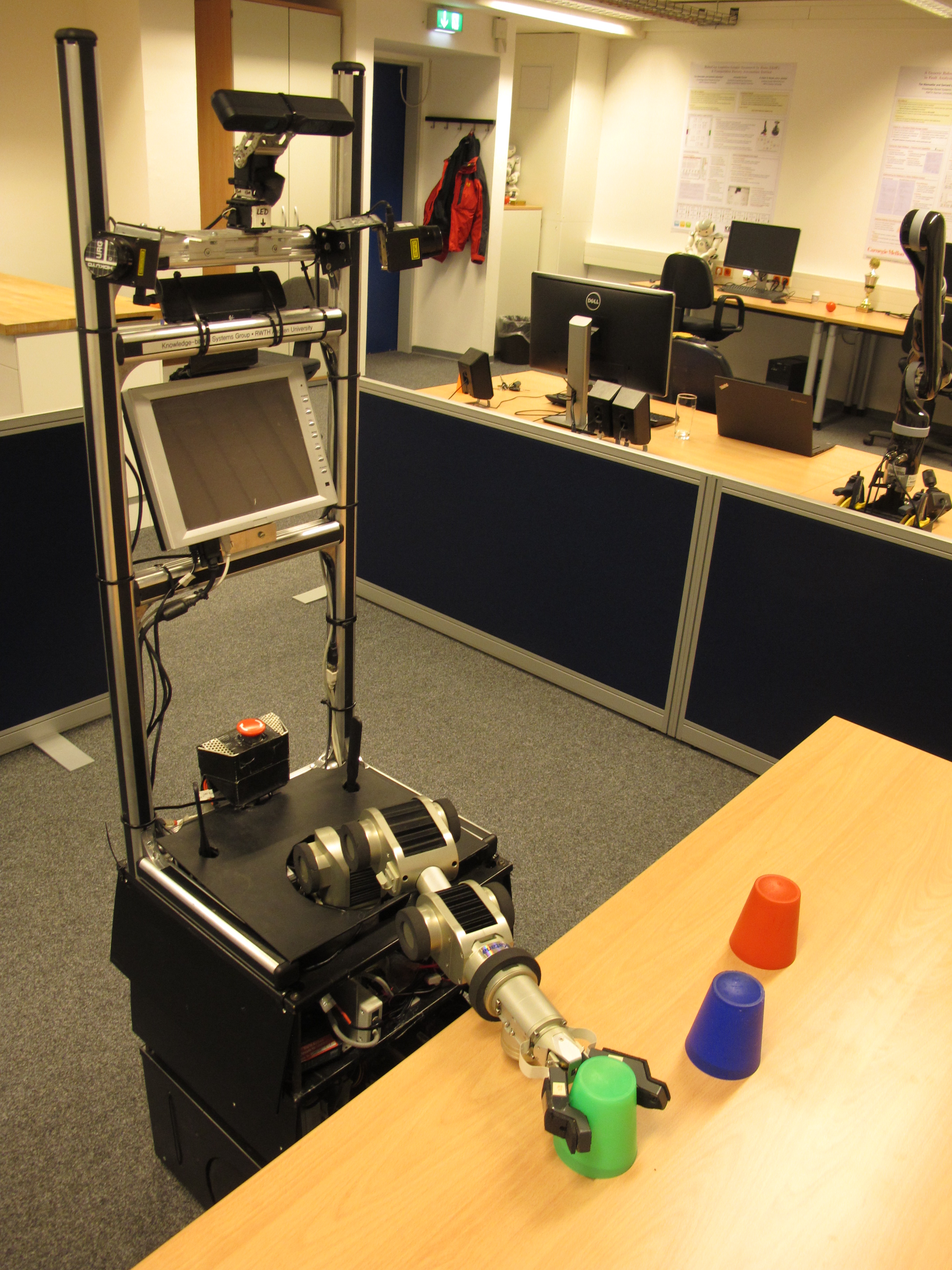}
  \end{center}
  \caption{The domestic service robot \caesar~\parencite{hofmannContinualPlanningGolog2016}.}
  \label{fig:caesar}
  \end{subfigure}
  \hfill
  \begin{subfigure}[c]{0.60\textwidth}
  \begin{center}
    \includegraphics[width=\textwidth,height=0.3\textheight,keepaspectratio]{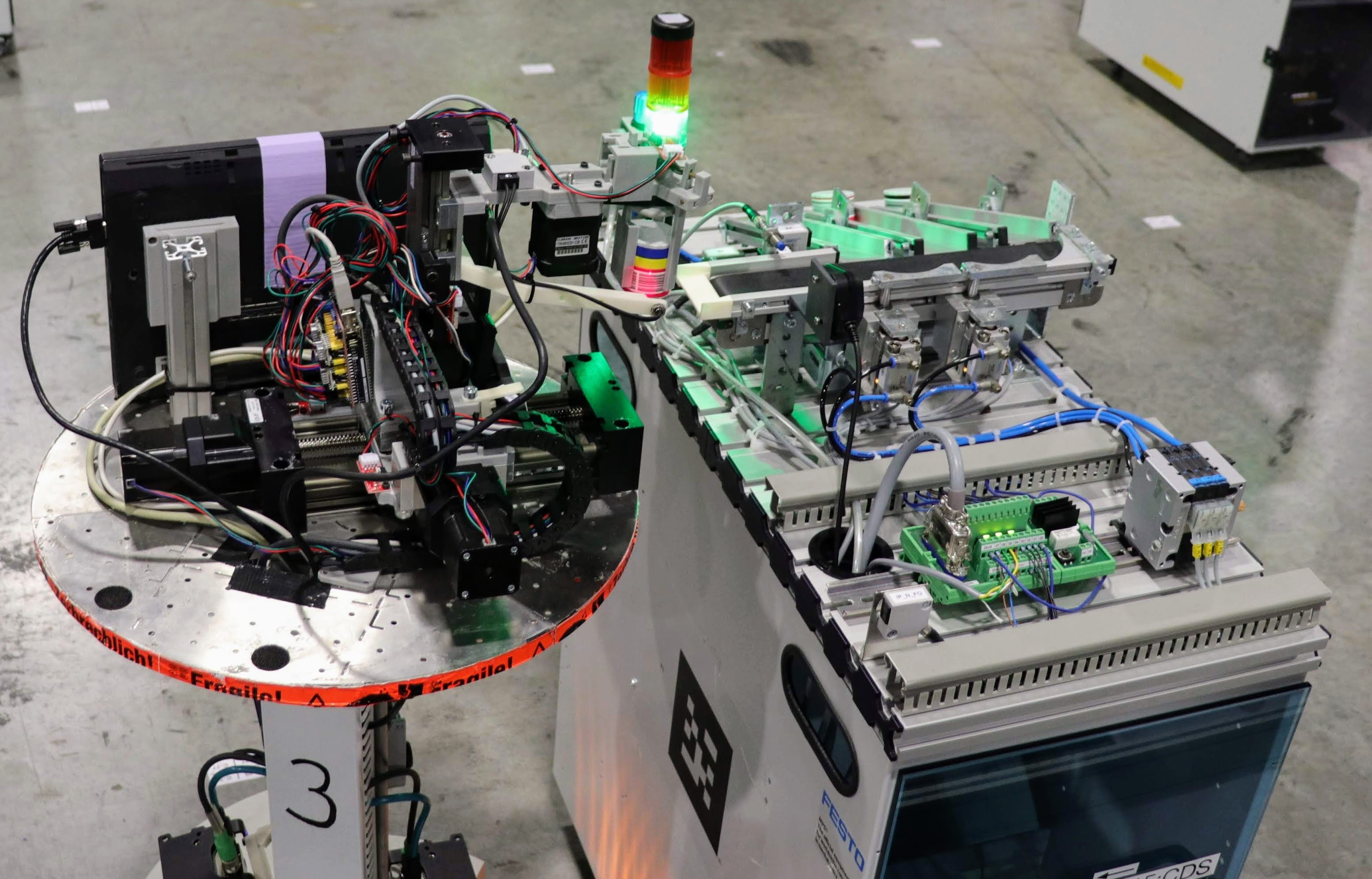}
  \end{center}
  \caption{A logistics robot of the team \emph{Carologistics} as used in the \ac{RCLL}~\parencite{haberingUsingPlatformModels2021}.}
  \label{fig:rcll-robot}
  \end{subfigure}
  \caption[Two different robots with similar capabilities.]{Two different robots with similar capabilities.
    From a high-level perspective, both robots are the same: They can move around and pick up or put down objects.
    However, when considering the actual robot, they clearly differ in many ways.
    For example, while \caesar has an arm that allows it to reach far onto the table, the logistics robot needs to move close to the machine in order to put down its workpiece.
  }
  \label{fig:robots}
\end{figure}

However, when executing an agent program on a real-world robot platform, these assumptions turn out to be unrealistic.
Often, implementing the atomic actions is not trivial and additional constraints need to be satisfied before an action can be executed.
As an example, consider the two robots shown in \autoref{fig:robots}.
From an abstract perspective, both robots have the same capabilities, because each robot is capable of moving between locations and picking up and putting down objects with its robotic arm.
In reality, the two robots differ significantly:
While \caesar has an arm that allows it to reach far onto the table, the logistics robot needs to move close to the machine in order to pick up or put down a workpiece.
On the other hand, the logistics robot has an omni-wheel drive that allows it to move into any direction for fine adjustments, while
\caesar uses conventional wheels and therefore needs to carefully align to a target location before it can pick up an object.
Furthermore, the logistics robot uses a RGB/D camera for object detection, which requires careful handling:
While the camera is essential to detect the target pose, it may interfere with the light barrier of the machine.
Therefore, it should be turned off whenever it is not needed.
On the other hand, the camera needs some time to initialize and therefore needs to be turned on in advance and should not be turned off if it is used again in the near future.
Hence, in addition to state-based action preconditions, we also need to consider \emph{temporal constraints} and \emph{timing constraints} when executing the program.

So far, the most common approach to deal with these kinds of constraints is based on three-layered architectures~\parencite{gatThreelayerArchitectures1998}.
The \emph{controller layer} implements primitive behaviors such as updating the motor speed or recognizing objects that directly involve sensors and actuators.
The middle layer combines primitive behaviors into simple tasks such as navigating to a location or picking up some object, e.g., with a \emph{behavior engine}~\parencite{niemuellerLuabasedBehaviorEngine2009}.
The deliberator on the highest layer is responsible for high-level reasoning, e.g., in the form of a \golog program~\parencite{levesqueGOLOGLogicProgramming1997}.
The layers are clearly separated and communicate through a well-defined interface, i.e., the agent program instructs the behavior engine to execute some action and the behavior engine reports whether the execution was successful.
Yet, in many scenarios, this clear abstraction is impossible, because low-level actions (such as turning on the camera) have an effect on the abstract plan and vice versa.
In the case of the logistics robot, it must turn on its camera well in advance before it is used so the camera is fully initialized in time.
If the low-level framework is not informed of the actions that are planned in the future, it may only start the camera once the high-level component instructs it to grasp an object.
It then needs to initialize the camera, wait until the camera is ready, and only then it may continue with the actual grasp action, leading to unnecessary delays.
As another example, \caesar's arm needs to be calibrated before being used.
During the calibration process, the arm moves around to determine its joint values, therefore it must not calibrate while it is near an obstacle.
In this scenario, the robot must decide to calibrate its arm before it moves to the target location.
However, if the calibration is encapsulated in the underlying framework, this is not possible, because the robot framework does not know which actions the agent plans to do in the future, and therefore does not know if and when it needs to calibrate its arm.

Perhaps even more importantly, it is impossible to give any formal guarantees if those details are hidden in the underlying framework.
In the case of the logistics robot, we may want to verify that the robot turns off any sensors that may possibly interfere with the machines as long as the robot is moving.
Similarly, we may want to ensure that the robot never lifts heavy objects for more than $10$ seconds, as this may overload the robot's arm.
Giving such guarantees is only possible if the low-level system is modeled as part of the robot program.

Another aspect is \emph{uncertainty}: On a real robot, sensing is always subject to noise and action outcomes are never certain.
While probabilistic methods are ubiquitous to solve conventional robotics tasks such as localization, mapping, and navigation~\parencite{thrunProbabilisticRobotics2005}, many approaches to high-level reasoning require non-stochastic actions and noise-free sensors.
In the case of low-level skills, uncertainty can often be abstracted away by the controller layer, e.g., the high-level reasoner does not need to know the exact position of an object as long as the controller layer is able to detect the object precisely enough to grasp it.
However, this form of delegation is not always possible: If the robot fails to grasp an object and drops it to the ground instead, it is impossible for the controller layer to recover (assuming it cannot pick up objects from the ground).
Hence, the high-level reasoner needs to be aware of the possible outcomes so it can react accordingly.

Therefore, rather than hiding away those details in the underlying framework, we propose to incorporate them into the high-level reasoning process.
At the same time, it is also undesirable to directly encode them into the abstract program, for several reasons.
First, designing a suitable program that accomplishes a certain objective is difficult, even on an abstract level.
Incorporating low-level details such as turning on and off the camera repeatedly or considering all possible action outcomes and their probabilities makes this process even more challenging.
Additionally, especially if we include search into the program, the domain should be kept as succinct as possible, as a more complex domain with additional predicates and actions may have a severe impact on the reasoner's performance.
If possible, the abstract program should also be non-stochastic, because reasoning about noisy actions and sensors is infeasible for larger domains.

The goal of this thesis is to close this gap between high-level reasoning and acting.
We focus on two aspects:
\begin{enumerate}
  \item Domains with real-time temporal constraints.
  \item Stochastic domains with noisy sensors and actuators.
\end{enumerate}

\begin{figure}[htb]
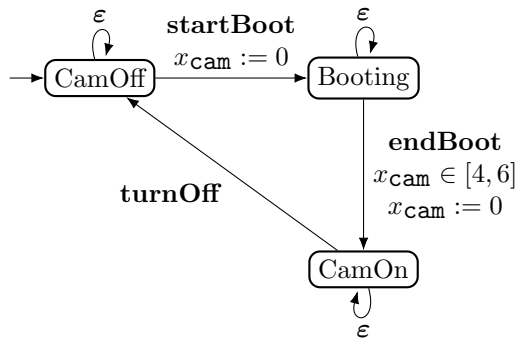

  \centering
  \includestandalone{figures/ex-platform-ta}
  \caption[A \acs*{TA} that models a robot camera.]{A \acl*{TA} that models a robot camera.
    If the camera is off, the robot may start booting the camera, which takes at least \SI{4}{\sec} and at most \SI{6}{\sec}.
    If it is on, the robot may instantaneously turn the camera off again.
  }%
  \label{fig:camera-model}
\end{figure}

For the first aspect, we propose to keep the abstract program as is, but augment it in a separate transformation step.
To deal with the low-level platform components, we propose to model them as \acfp{TA}, as shown in \autoref{fig:camera-model}.
Timed automata are a suitable model for such robot self models:
For one, a state machine is a natural choice, because the physical state of the component can directly be modeled as a state of the state machine.
Second, extending those state machines with time is useful to encode timing constraints into the platform model.
Finally, \acp{TA} offer a good compromise between expressiveness and decidability, as more expressive formalisms such as hybrid automata are often undecidable. 

For the second aspect, we propose to use \emph{abstraction}.
Based on a low-level domain description that contains stochastic actions, we propose to model a program that abstracts away the stochastic aspects of the low-level program.
We then map each action of the high-level domain to a sub-program of the low-level domain.
By doing so, we can write classical \golog programs that do not refer to probabilities, but then translate them to a low-level program that deals with the stochastic aspects of the domain during execution.

\section{Contributions}
The contributions of this thesis are as follows:
\begin{enumerate}
  \item We introduce the logic \tesg, which extends \esg~\parencite{classenLogicNonterminatingGolog2008} with metric time, and therefore allows us to express timing constraints in a variant of the situation calculus.
    Previous work on representing time in the situation calculus usually does so by including the reals and the arithmetic operators $+$ and $\times$ in the logic.
    As we will see, this results in undecidable reasoning tasks, even if the domain is restricted to a finite set of objects.
    To circumvent this issue, we use ideas from timed automata theory and represent time with a finite set of clocks with restricted operators that only allow comparing clock values to fixed rational values and resetting a clock to zero.
    In addition to having a notion of time, \tesg also allows to express temporal properties similar to \ac{MTL}, which can be used to express timing constraints for program executions.
    We show that \tesg basic action theories are compatible to \esg basic action theories, which allows us to use previously established results such as the connection to planning~\parencite{classenIntegrationGologPlanning2007} in \tesg.
    Furthermore, \tesg has the same temporal properties as \ac{MTL}, hence it can be seen as an embedding of \ac{MTL} into the situation calculus.
  \item We investigate the verification of temporal properties in \tesg.
    Intuitively, the verification problem is to decide whether every terminating execution of a given program satisfies a given \ac{MTL} specification.
    We show that verification is decidable, at least if the program is restricted to a finite number of objects and actions.
    To solve the verification problem, we first translate the \ac{MTL} specification to an \acfi{ATA} and then use the synchronous product of the \ac{ATA} and the program to check whether the specification is violated.
    This involves two technical challenges: For one, the resulting transition system is infinitely-branching, because in each step, we have one time successor for each positive real number.
    We show that we can use \emph{regionalization} and a \emph{time-abstract bisimulation} to obtain an equivalent transition system that is only finitely-branching.
    Second, the transition system may contain infinite paths, corresponding to non-terminating executions of the program.
    We demonstrate that the transition system is a \acfi{WSTS} by defining a \ac{wqo} on the states of the transition system, which allows us to stop traversing every path after a finite number of steps.
  \item We also describe \emph{controller synthesis} for \golog programs.
    In addition to the \ac{MTL} specification, we are now given a partition of the actions into \emph{controllable} and \emph{environment} actions.
    The synthesis problem is to determine a controller that selects the right controller actions such that every possible execution satisfies the specification, independent of the choices of the environment.
    In order to solve the control problem, we define a \emph{timed game} on the transition system from above, where a winning strategy for player 1 corresponds to a controller that satisfies the specification.
    We describe an algorithm that first constructs a finite tree from the transition system, labels all nodes bottom-up to identify states that satisfy the specification, and then synthesizes a controller by traversing the tree and selecting those nodes that are labeled as good.
  \item We show how such a program controller can be used to transform an abstract program into a platform-specific controller.
    We do so by modeling the robot platform with timed automata and formulating \ac{MTL} constraints that connect the low-level platform with the abstract program.
    A synthesized controller will then correspond to an execution of the abstract program augmented by platform-specific actions, which guarantee that the specification is satisfied.
    As such a controller considers all possible choices by the environment, it can be synthesized offline.
    We evaluate the approach on two simple scenarios.
  \item As an alternative approach, we also describe an \emph{online transformation method}.
    In this case, we assume that the interpreter has already determined a single sequence of actions to be executed on the robot.
    We show that under some simplifying assumptions, the plan transformation task can be modeled as a \emph{reachability problem on timed automata}.
    This allows us to use the well-established verification tool \uppaal~\parencite{bengtssonUPPAALToolSuite1996} to determine an augmented plan that satisfies the specification.
    We evaluate the approach on a number of problems from the \acl{RCLL}.
  \item Finally, we turn towards noisy sensors and actuators.
    Based on the logic \ds, an epistemic variant of the situation calculus with degrees of belief that allows to model stochastic actions, we describe a form of \emph{abstraction} of basic action theories with noisy actions.
    This allows us to define a high-level basic action theory that abstracts away the details of a lower-level basic action theory, possibly getting rid of probabilistic aspects.
    We do so by defining an appropriate notion of \emph{bisimulation} that guarantees some form of equivalence between the low-level and the high-level theory.
    By doing so, we can use a non-stochastic reasoner on the abstract domain and then translate the resulting actions to low-level programs for execution.
\end{enumerate}

\section{Outline}
This thesis is organized as follows:
\begin{description}
  \item[\autoref{chap:related-work}] discusses related work with a particular focus on the situation calculus, planning in combination with acting, verification and synthesis, as well as abstraction.
  \item[\autoref{chap:foundations}] summarizes the foundations for this thesis.
    In particular, it recaps the situation calculus and its extensions with time, noisy actions, and epistemic operators.
    It also summarizes the foundations of \golog, a programming language building on top of the situation calculus.
    Furthermore, it describes necessary concepts from timed systems, starting with temporal logics and continuing with timed automata.
  \item[\autoref{chap:timed-esg}] introduces the logic \tesg including basic action theories and regression and then shows some properties of the logic, relating it to the logic \esg on the one hand and to \ac{MTL} on the other hand.
  \item[\autoref{chap:synthesis}] defines the \ac{MTL} \emph{verification} and \emph{synthesis} problems for \golog programs and shows that both problems are decidable for terminating programs with a finite number of objects.
    It also describes and evaluates a controller synthesis framework.
  \item[\autoref{chap:transformation-as-reachability-problem}] presents an alternative approach for the program transformation by making some simplifying assumptions.
    It evaluates the approach and demonstrates its capability of transforming a plan in a reasonable time, even when scaled to larger programs.
  \item[\autoref{chap:abstraction}] introduces the logic \dsg that extends \ds with a transition semantics for \golog programs and therefore allows to write programs that include stochastic actions.
    It then defines abstraction of \dsg basic action theories and showcases how abstraction can be used for eliminating the stochastic aspects of a simple robot program.
  \item[\autoref{chap:conclusion}] concludes the thesis with a summary and a discussion of possible future work.
  \item[\autoref{chap:proofs}] contains full proofs for all lemmas, theorems, and corollaries.
  \item[\autoref{chap:contributions}] lists all publications related to this thesis as well as additional unrelated publications by the author.
\end{description}

\chapter{Related Work}\label{chap:related-work}
In this chapter, we discuss previous work related to the goals of this thesis.
We start with action formalisms and the situation calculus, as well as some alternatives.
We continue with planning, which can be seen as a different approach to reasoning about actions, and then focus on the combination of planning and acting.
Next, we discuss verification and synthesis as general concepts and how they were applied in the context of reasoning systems such as \golog.
We conclude the discussion of related work with literature on abstraction.

\section{Action Formalisms}\label{sec:action-formalisms}
A fundamental question in cognitive robotics is how to represent the agent's view of the world, including a description of the current state, as well as the effects that actions may have on that state.
The \emph{situation calculus}~\parencite{mccarthySituationsActionsCausal1963,mccarthyPhilosophicalProblemsStandpoint1969,reiterKnowledgeActionLogical2001} is a logical language based on first-order logic that provides an answer to this question.
In the situation calculus, world states are represented explicitly as first-order terms called \emph{situations}, where \emph{fluents} describe (possibly changing) properties of the world and actions describe how the world changes from one situation to another.
As first argued by \textcite{mccarthyProgramsCommonSense1959} and further discussed by \textcite{mooreRoleLogicKnowledge1982} as well as \textcite{levesqueLogicKnowledgeBases2001}, basing the representation on first-order logic is a reasonable choice:
For one, objects of the world as well their properties and relations to other objects can be directly represented as objects and relations in the logic.
Furthermore, quantification and disjunction allows expressing incomplete knowledge, e.g., we can express that all objects in a bag are green without knowing which objects are in the bag, or we can state that an object is either red or blue without knowing which one it is.

Apart from providing a formalism for describing action and change, \textcite{mccarthyPhilosophicalProblemsStandpoint1969} also identified a fundamental representational challenge, namely the \emph{frame problem}:
When formally describing an action, in addition to stating the changes the action brings to the world, it is also necessary to describe \emph{what does not change}.
As an example, if a robot moves from the kitchen to the living room, the robot's location changes, but the action has no effect on the objects on the dining table or the light in the hallway.
While listing the changes produced by an action is often straightforward, describing everything that remains unchanged is problematic, because of the sheer number of such ``non-effects'' and also because when describing an action, people naturally only remember the effects of an action~\parencite{linSituationCalculus2008}.
Over the years, a number of solutions to the frame problem were discussed, e.g., \parencites{mccarthyApplicationsCircumscriptionFormalizing1986,hanksNonmonotonicLogicTemporal1987,lifschitzFormalTheoriesAction1987}.
One succinct solution to the frame problem was eventually formulated by \textcite{reiterFrameProblemSituation1991}, combining ideas from \textcite{haasCaseDomainSpecificFrame1987}, \textcite{pednaultADLExploringMiddle1989}, and \textcite{schubertMonotonicSolutionFrame1990}.
Intuitively, rather than listing the effects action by action, \citeauthor{reiterFrameProblemSituation1991} proposes the use of \emph{successor state axioms}, where one successor state axiom describes how one fluent may be changed by any action the agent may take.
A complete description of the robot's capabilities then contains one successor state axiom for each fluent of the domain.
Implicitly, these successor state axioms make a completeness assumption, i.e., apart from the explicitly listed effects, there is no other way how a fluent may change its value.
Successor state axioms also allow a form of \emph{regression}~\parencite{waldingerAchievingSeveralGoals1981} in the situation calculus~\parencite{pirriContributionsMetatheorySituation1999}.
Regression modifies a formula containing a situation after a sequence of actions to an equivalent formula only mentioning the initial situation and therefore allows using a standard first-order theorem prover for reasoning in the situation calculus.

Over time, the situation calculus has been extended in numerous directions.
Pertaining to the fact that the agent may only have incomplete knowledge about the world and may need to sense in order gather additional knowledge, the epistemic situation calculus~\parencite{mooreReasoningKnowledgeAction1981,mooreFormalTheoryKnowledge1985} extends the situation calculus with a possible-world semantics known from modal logic~\parencite{kripkeSemanticalAnalysisModal1963,hintikkaKnowledgeBelief1969,garsonModalLogic2021}.
\textcites{scherlFrameProblemKnowledgeProducing1993,scherlKnowledgeActionFrame2003} have extended Reiter's solution to the frame problem to the epistemic situation calculus.
The logic \es~\parencite{lakemeyerSituationsSiSituation2004,lakemeyerSemanticCharacterizationUseful2011} is a variant of the epistemic situation calculus that uses modal operators for actions and knowledge and which defines the meaning of action and knowledge semantically rather than axiomatically.
\textcite{bacchusReasoningNoisySensors1999} extend the epistemic situation calculus with noisy sensors and effectors, where each possible world is assigned a weight, allowing to express that the agent believes a sentence with some degree of belief.
\textcite{belleReasoningProbabilitiesUnbounded2017} provide a modal variant based on \es, which will be the foundation for \autoref{chap:abstraction}.

A different line of work extended the situation calculus with time and concurrency.
Already \textcite{mccarthySituationsActionsCausal1963} proposed a fluent function $\mi{time}$ that gives the value of the time in a situation.
Elaborating on this idea, \textcite{gelfondWhatAreLimitations1991} described an extension where time is a fluent with values from the integers or reals and where each action has some duration.
Similarly, with the goal to describe narratives in the situation calculus, \textcite{millerNarrativesSituationCalculus1994} attached a time to each action and allowed overlapping actions with durations, along with partial ordering of actions that allows to describe two concurrent sequences of actions.
Some approaches also allow true concurrency~\parencite{linConcurrentActionsSituation1992}, where two actions occur in the same situation.
However, this may be problematic, due to  the \emph{precondition interaction problem}~\parencite{pintoReasoningTimeSituation1995}: While two actions may be possible at the same time, executing them both simultaneously may be impossible.
\textcite{pintoTemporalReasoningLogic1993} modeled durative actions with instantaneous \emph{start} and \emph{end} actions and continuous time to describe actions and events, which also allows embedding the event calculus \parencite{shanahanEventCalculusExplained1999}.
By distinguishing actual situations from possible situations, \textcite{pintoReasoningTimeSituation1995} define a total ordering on situations which allows to express linear-time temporal properties within the situation calculus.
Related to time, several approaches~\parencites{millerCaseStudyReasoning1996,reiterNaturalActionsConcurrency1996} also allow to model continuous processes in the situation calculus, where a fluent continuously changes its value between the occurrence of two actions.
The \emph{hybrid situation calculus}~\parencite{batusovHybridTemporalSituation2019} combines the situation calculus with hybrid systems by embedding \emph{hybrid automata}~\parencite{alurHybridAutomataAlgorithmic1993}.
\textcite{finziRepresentingFlexibleTemporal2005} propose a different hybrid approach by combining the situation calculus with temporal constraint reasoning with temporal constraints based on Allen's Interval Algebra~\parencite{allenMaintainingKnowledgeTemporal1983}, which are translated into temporal constraint networks~\parencite{dechterTemporalConstraintNetworks1991,meiriCombiningQualitativeQuantitative1996}.

Reiter's solution to the frame problem and regression-based query evaluation in the situation calculus gave rise to \golog~\parencite{levesqueGOLOGLogicProgramming1997}, an agent programming language based on the situation calculus.
Exploiting a basic action theory that specifies preconditions and effects of the available primitive actions, \golog programs specify high-level agent behavior with constructs such as conditionals and loops known from imperative programming languages as well as non-deterministic constructs such as non-deterministic branching, choice of argument, and iteration.
\congolog~\parencite{degiacomoConGologConcurrentProgramming2000} extends the original \golog with concurrency, interrupts, and exogenous actions.
While the original \golog as well as \congolog executed the program offline, \indigolog~\parencite{degiacomoIndiGologHighlevelProgramming2009} provides an online execution semantics, where the interpreter chooses the next action to execute rather than searching over the whole program to find a complete execution.
\textsc{cc-Golog}~\parencite{grosskreutzOnLineExecutionCcGolog2001,grosskreutzCcGologActionLanguage2003} supports continuous change in \golog.
\textsc{DTGolog}~\parencite{boutilierDecisiontheoreticHighlevelAgent2000} integrates decision-theoretic planning into \golog.
\readylog~\parencite{ferreinLogicbasedRobotControl2008} is aimed at real-time robotic systems and supports decision-theoretic planning as well as continuously changing fluents and passive sensing.
\textcite{lakemeyerSensingOfflineInterpreting1999,reiterKnowledgebasedProgrammingSensing2001,classenFoundationsKnowledgebasedPrograms2006,classenKnowledgebasedProgramsDefaults2016,baierKnowledgebasedProgramsBuilding2022} describe epistemic variants of \golog based on the epistemic situation calculus that support sensing actions.

Apart from the situation calculus, there are other approaches to represent action and change in a logical framework.
The fluent calculus~\parencite{thielscherIntroductionFluentCalculus1998,thielscherSituationCalculusFluent1999} builds on ideas by \textcite{holldoblerNewDeductiveApproach1990} and is based on the situation calculus but solves the frame problem with \emph{state update axioms}.
Instead of having one successor state axiom for each fluent, one state update axiom for each action describes how the action changes the state.
\textsc{FLUX}~\parencite{thielscherFLUXLogicProgramming2005} is an agent programming language similar to \golog but is based on the fluent calculus, where the explicit state representation avoids the need of repeated regression.
The event calculus~\parencites{kowalskiLogicbasedCalculusEvents1986,shanahanEventCalculusExplained1999} is a formalism for representing narratives in the form of events and their effects and is also based on first-order logic.
It uses an explicit notion of time, where each action occurs at some time point and formulas may refer to fluent values not only at some occurrent of an event, but at arbitrary time points.
The event calculus solves the frame problem with circumscription~\parencite{mccarthyCircumscriptionFormNonmonotonic1980}.
\acfi{TAL} is a framework based on sorted first-order logic with a linear discrete time structure and therefore an explicit representation of time.
It uses narratives to specify agent behavior, where a narrative describes fluents that hold at certain points in time, dependencies between fluents, action occurrences, as well as domain constraints.
\ac{TAL} solves the frame problem with circumscription and \emph{persistence statements}, which describe under which conditions a fluent may change its value.
A persistent fluent may only change its value if an action explicitly allows it to change, similar to the situation calculus.
A fluent may also be \emph{durational}, in which case it has a default value that may only be changed by an action or some other constraint.
Dynamic fluents do not have any restrictions and may change their values arbitrarily.




\section{Planning}\label{sec:planning}
\emph{Automated Planning}~\parencites{ghallabAutomatedPlanningTheory2004,geffnerConciseIntroductionModels2013,ghallabAutomatedPlanningActing2016} is a different approach to reasoning about actions.
In classical planning, the task is to find a sequence of actions that achieves some goal, given a description of the initial state and the available actions.
One of the first approaches to planning is the \emph{Stanford Research Institute Problem Solver} (\strips)~\parencite{nilssonSTRIPSNewApproach1971}.
While \strips is also planning system, today it is mostly known as a formal language for planning problems.
In comparison to formalisms such as the situation calculus, the state and action descriptions are typically more constrained.
In \strips, states are represented by a collection of atomic propositions and each action operator is represented by three sets of propositions \emph{Pre}, \emph{Add}, and \emph{Del}: The precondition $\mi{Pre}(o)$ contains all atoms that must be true for the action to be possible, the add list $\mi{Add}(o)$ contains all atoms that $o$ makes true, and $\mi{Del}(o)$ contains all atoms that $o$ makes false.
While some earlier approaches to planning were based on theorem-proving and resolution~\parencite{greenApplicationTheoremProving1969}, this representation allows viewing a planning problem as search on a directed graph~\parencite{newellGPSProgramThat1961,nilssonSTRIPSNewApproach1971}.
In this state-space graph, each node is a state of the world and each edge is an action that changes the state from the source to the target node.
A different representation is used in \textsc{GraphPlan}~\parencite{blumFastPlanningPlanning1997}, where the planning problem is encoded in a \emph{planning graph} that explicitly represents constraints inherent in the planning problem, which resulted in an significant performance increase.
Representing planning as a search problem also allows to utilize heuristic approaches such as A\textsuperscript{*} or best-first search, where a heuristic function is automatically extracted from the planning problem.
In the subsequent years, this view of \emph{planning as heuristic search}~\parencite{bonetPlanningHeuristicSearch2001} resulted in great improvements in planner performance, e.g., with planners such as \textsc{HSP}~\parencite{bonetPlanningHeuristicSearch2001}, \textsc{FF}~\parencite{hoffmannFFPlanningSystem2001}, and \textsc{LAMA}~\parencite{richterLAMAPlannerGuiding2010}, as well as planning frameworks such as \textsc{Fast Downward}~\parencite{helmertFastDownwardPlanning2006}.

Initially, every planner used their own input language and therefore a comparison was difficult.
The \acfi{PDDL}~\parencite{mcdermottPDDLPlanningDomain1998} standardized the language and allowed a comparison of planning systems in the context of the \acfi{IPC}~\parencite{mcdermott1998AIPlanning2000}.
It extends the \strips language with features from \textsc{ADL}~\parencite{pednaultADLExploringMiddle1989} such as quantifiers and conditional effects.
\textsc{PDDL}2.1~\parencite{foxPDDL2ExtensionPDDL2003} extends \textsc{PDDL} with time, numeric properties, and durative actions and is supported by planners such as \textsc{Metric-FF}~\parencite{hoffmannMetricFFPlanningSystem2003} and \textsc{TFD}~\parencite{eyerichUsingContextenhancedAdditive2009}.
\textsc{PDDL}3 \parencite{gereviniPlanConstraintsPreferences2005} further extends the language with strong and soft constraints, where not only the final state needs to satisfy the goal, but intermediate states must satisfy temporal constraints.

Some planning systems combine temporal reasoning with heuristic search, e.g., for temporally extended goals~\parencite{bacchusPlanningTemporallyExtended1998}.
\textsc{TLPlan}~\parencite{bacchusUsingTemporalLogics2000} uses first-order \ac{LTL} for search control, i.e., for guiding the search to improve planning performance.
Similarly, \textsc{TALPlanner}~\parencite{dohertyTALplannerTemporalLogicBased2001} uses a \ac{TAL} narrative to guide a forward-chaining planner.
It can also incorporate sensing results gathered during execution and uses execution monitoring to react to unexpected events~\parencite{dohertyTemporalLogicbasedPlanning2009}.
\textsc{HPlan-P}~\parencite{baierHeuristicSearchApproach2009} combines search guidance with temporally extended preferences.
\textsc{PPlan} \parencite{bienvenuPlanningQualitativeTemporal2006,bienvenuSpecifyingComputingPreferred2011} incorporates user preferences formulated in the temporal logic $\mathcal{LPP}$ with a semantics defined in the \sitcalc.



\section{Planning and Acting}\label{sec:planning-acting}
While planning (and more generally reasoning about actions) can be used to determine a plan that accomplishes a given goal, executing such a plan comes with additional challenges~\parencite{ghallabActorsViewAutomated2014}:
\begin{enumerate*}[label=(\alph*)]
  \item representing actions with preconditions and effects is often too abstract to be useful for acting,
  \item planning is nicely formalized, while acting is much harder to formalize,
  \item acting in an open and dynamic environment requires different information gathering, processing and decision-making capabilities.
\end{enumerate*}
Despite these challenges, deliberating systems that combine planning with acting have been researched extensively, as surveyed by \textcite{ingrandDeliberationAutonomousRobots2017}.

\textsc{Xfrm}~\parencites{beetzDeclarativeGoalsReactive1992,beetzExpressingTransformationsStructured1997} is a reactive planner based on the \emph{Reactive Plan Language} (RPL) \parencite{mcdermottReactivePlanLanguage1991} which provides constructs for sequencing, conditionals, loops, local variables, and subroutines.
During execution, the planner continuously refines handwritten reactive plans by choosing between several alternatives while maximizing the expected utility based on estimations for the plan stability, correctness, execution  time, and completeness.
The \emph{Procedural Reasoning System} (\textsc{PRS})~\parencite{ingrandPRSHighLevel1996,myersProceduralKnowledgeApproach1996} is a  high-level control and supervision framework which uses a plan library to describe partial plans for achieving sub-goals and reacting to certain situations.
During execution, the system selects appropriate plans and procedures to perform the desired tasks while adapting to the current situation.
It can use a planner to anticipate execution paths to avoid plans that may lead to a failure in the future \parencite{despouysPropicePlanUnifiedFramework2000}.
\textsc{IxTeT}~\parencite{ghallabRepresentationControlIxTeT1994} is a temporal planner for robotic systems that generates flexible plans while respecting deadlines and resource constraints.
Each action is modeled with a finite-state automaton~\parencite{chatilaIntegratedPlanningExecution1992}.
It has also been extended with execution monitoring, plan repair, and replanning and uses \textsc{PRS} as procedural executive~\parencite{lemaiInterleavingTemporalPlanning2004}.

The \emph{Reactive Model-Based Programming Language} (RMPL)~\parencite{williamsReactivePlannerModelbased1997,williamsModelbasedProgrammingIntelligent2003} is a framework for constraint-based modeling where the programmer reads from and writes to hidden state variables.
The system's executive then maps between those hidden states and plant sensors as well as control variables based on a component model of the system.
RMPL allows constructs for concurrency, sequencing, iteration, branching, several guarded transitions, and preemption.
It supports different executives with different underlying models.
\emph{RBurton}~\parencite{williamsUnifyingModelbasedReactive1999} supports probabilistic, constraint-based modeling with reactive programming constructs.
The semantics is based on probabilistic hierarchical constraint automata~\parencite{inghamReactiveModelbasedProgramming2001}, which allows the decomposition of the system into automata with location and transition constraints and probabilistic transitions.
\emph{Kirk}~\parencite{kimExecutingReactiveModelbased2001} is based on temporal plan networks, a generalization of a simple temporal networks (STNs), and allows simple temporal constraints combined with serial, parallel, and choice operations.
During execution, the system commits to some alternative for each choice operator and incrementally replans whenever a choice is invalidated by some disturbance.
\emph{Drake}~\parencite{conradDrakeEfficientExecutive2011} extends the model to disjunctive temporal networks (DTNs), which guarantee the feasibility of the plan in the compilation step and therefore avoids online replanning.
Finally, \emph{Pike}~\parencite{levineConcurrentPlanRecognition2014} extends the framework with intent recognition and adaption which permits human-robot collaboration.

\textsc{Plexil}~\parencite{vermaPlanExecutionInterchange2005} is an imperative instruction language where plans are decomposed into nodes for tasks such as executing commands or assigning a variable.
Each node may have a number of constraints and guards, e.g., pre- and  post-conditions, which are evaluated during execution.
\textsc{Idea}~\parencite{finziModelbasedExecutiveControl2004} pursues a different approach by modeling the system components as a distributed multi-agent system, where each agent is responsible for a particular function, e.g., task planning or navigation.
To achieve the main goal, the agents explicitly coordinate based on communication actions modeled as planning actions.

\textsc{KnowRob}~\parencite{tenorthKnowRobKnowledgeProcessing2013,beetzKnowRob2nd2018} is a knowledge representation and reasoning framework based on description logic that combines symbolic reasoning with hybrid, visual, and simulation-based reasoning to predict outcomes of motion plans.
\textsc{Cram}~\parencite{beetzCRAMCognitiveRobot2010} defines high-level programs in the behavior specification language CPL, which is similar to RPL and supports control structures for parallel execution and partial ordering of sub-plans.
\textsc{Cram} uses \textsc{KnowRob} as knowledge processing system that can derive \emph{computable predicates} in addition to the static  facts in the knowledge base.
Additionally, it provides a meta-reasoning system to reason about plans, tasks, and plan execution, and therefore provides methods to analyze the system, e.g., to find flaws in plan execution and transform plans to fix the detected flaws.

In \emph{continual planning}~\parencite{brennerContinualPlanningActing2009}, planning and acting is interleaved such that the agent can gather necessary information during plan execution.
\emph{Assertions} are placeholder actions that guarantee to achieve a certain effect without specifying the necessary actions.
They can be used as placeholders for parts of the plan that requires additional knowledge.
Once all the necessary information has been collected, the agent replans and replaces the assertion by a primitive action sequence.
Continual planning has also been used in \golog~\parencite{hofmannContinualPlanningGolog2016} for interleaved planning and acting.
\textsc{Golex}~\parencite{hahnelGOLEXBridgingGap1998} is an execution framework based on \golog which decomposes primitive high-level actions into a sequence of directives for the low-level robot control system and permanently monitors the execution of the actions.
\textcite{degiacomoExecutionMonitoringHighLevel1998} define formal \emph{Monitor} and \emph{Recover} mechanisms in \golog that can react to exogenous actions by adapting the agent's plan.
\textcite{schifferSelfMaintenanceAutonomousRobots2010} describe an online transformation of a \readylog program by inserting actions to satisfy qualitative temporal platform-specific constraints, under the assumption that agent domain and platform domain are disjunct.

\emph{Task and motion planning} (TAMP) can be seen as a specialized planning and acting approach as it combines high-level task planning with low-level motion planning, e.g., with an interface between task and motion planner to effectively combine off-the-shelf planners \cite{srivastavaCombinedTaskMotion2014}, by extending the FF heuristics for geometric reasoning \cite{garrettFFRobEfficientHeuristic2015}, or by using constraint programming to guide the high-level search with geometric \cite{gravotASyMovPlannerThat2005,dantamIncrementalTaskMotion2016} and temporal constraints \cite{erdemCombiningHighlevelCausal2011}.
%
In a similar fashion, \textcite{erdemSystematicAnalysisLevels2016} integrate general feasibility checks into an ASP-based planner, either by checking constraints directly during search, or by constraining the planner afterwards if a feasibility check is violated.
Alternatively, external predicates can be directly embedded into PDDL-based planners \cite{hertlePlanningSemanticAttachments2012,dornhegeSemanticAttachmentsDomainindependent2012}, ASP-based planners \cite{erdemAnswerSetProgramming2012}, and other reasoners such as \textsc{CCalc} \cite{akerCausalReasoningPlanning2011}.
There, a low-level component sets the value of a symbolic atom used by the task planner, which allows to integrate platform components, but does not allow for temporal constraints.

\section{Verification and Synthesis}\label{sec:related:verification}

Generally speaking, \emph{formal verification} is the process of checking whether a program satisfies some desired properties and is therefore in some sense correct.
In contrast to \emph{testing}, where the program is run on specific inputs and its output is compared against desired outputs, verification shows the correctness of the program on all inputs by means of formal methods.
Early approaches to formal verification (e.g., ~\parencite{mccarthyBasisMathematicalTheory1963,hoareAxiomaticBasisComputer1969,dijkstraHumbleProgrammer1972}) were based on the idea of manually providing a mathematical proof of the program's correctness.
\textcite{dijkstraHumbleProgrammer1972} argued that ``the programmer should let correctness proof and program grow hand in hand'' and thus provide a correctness proof for their program while writing the program.
In contrast, in \emph{computer-aided verification}~\parencite{clarkeComputeraidedVerification1996}, a theorem prover such as Isabelle~\parencite{nipkowIsabelleHOLProof2002} assists the programmer by deriving (parts of) the proof automatically.

Today, \emph{model checking}~\parencite{baierPrinciplesModelChecking2008,clarkeModelCheckingAlgorithmic2009,clarkeIntroductionModelChecking2018} is one of the most commonly used paradigms in formal verification.
In model checking, first introduced by \textcite{clarkeDesignSynthesisSynchronization1982,queilleSpecificationVerificationConcurrent1982}, the user describes the system in an abstract finite-state model (e.g., a Kripke structure \parencite{kripkeCompletenessTheoremModal1959}) and specifies desired properties in a temporal logic such as \acf{LTL}~\parencite{pnueliTemporalLogicPrograms1977}, \acf{CTL}~\parencite{clarkeDesignSynthesisSynchronization1982}, or \ctlstar~\parencite{emersonSometimesNotNever1986}.
The model checker then algorithmically verifies that the model satisfies the specification, usually by exhaustively examining all reachable states, and provides a counterexample otherwise.

Over the years, research in model checking has focused on two major challenges~\parencite{clarkeIntroductionModelChecking2018}: For one, scaling model checking to real-life problems is challenging because the state-space of a program is exponential in the input, resulting in the so-called \emph{state-explosion problem}~\parencite{demriParametricAnalysisStateexplosion2006}.
A number of techniques have been developed to tackle the state-explosion problem~\parencite{dsilvaSurveyAutomatedTechniques2008}.
Rather than using an explicit state representation, \emph{symbolic model checking}~\parencite{burchSymbolicModelChecking1992} uses implicit representations for sets of states, e.g., with binary decision diagrams (BDDs)~\parencite{bryantGraphBasedAlgorithmsBoolean1986} or with propositional formulas, which allow model checking based on SAT \parencite{biereSymbolicModelChecking1999,biereSATBasedModelChecking2018}.

The second challenge is to find suitable models and specification languages that are expressive enough to describe the system or program while maintaining decidability.
Of particular interest for this thesis are extensions with real time.
\emph{Timed automata}~\parencites{alurTheoryTimedAutomata1994,alurTimedAutomata1999,bengtssonTimedAutomataSemantics2004} extend finite-state automata with real time by equipping an automaton with a finite set of clocks whose real-timed values increase uniformly and where a transition may reset clocks and have a guard on clock values.
\uppaal~\parencite{bengtssonUPPAALToolSuite1996,behrmannDevelopingUPPAAL152011} is a tool suite for model-checking timed automata which has seen considerable efforts to improve its performance, e.g., with symbolic model checking based on region zones~\parencite{larsenModelcheckingRealtimeSystems1995}, difference-bounded matrices (DBMs)~\parencite{larsenUppaalStatusDevelopments1997}, and symmetry reduction~\parencite{hendriksAddingSymmetryReduction2004}.
Regarding specifications, several timed temporal logics have been proposed~\parencite{bouyerModelcheckingTimedTemporal2009}: \emph{Timed \ac{CTL}}~\parencite{alurModelcheckingRealtimeSystems1990} extends \ac{CTL} with timing constraints on the temporal modalities  and is therefore a branching-time timed temporal logic.
\ac{MTL}~\parencite{koymansSpecifyingRealtimeProperties1990} extends \ac{LTL} with timing constraints on the temporal modalities, e.g., with formulas such as $\finally{[1,2]} p$ which states that $p$ holds  in some future state at a time point in the interval $[0, 1]$.
As \ac{MTL} extends \ac{LTL}, it is a linear-time timed temporal logic.
For \ac{MTL} (as well as for TCTL) two different semantics exist: In the interval-based semantics, each state of the system is associated with a timed interval which indicates the period of time when the system is in that state.
With this semantics, the satisfiability problem is undecidable~\parencite{alurReallyTemporalLogic1994}.
For this reason, several syntactic restrictions have been proposed~\parencite{ouaknineRecentResultsMetric2008}: \ac{MITL}~\parencite{alurBenefitsRelaxingPunctuality1996} disallows singular intervals in the timing constraints, \mtlzi~\parencite{henzingerItTimeRealtime1998} requires that temporal modalities only restrict time in one direction, i.e., each interval has a left endpoint of $0$ or a right endpoint of $\infty$, and \emph{Bounded \ac{MTL}}~\parencite{bouyerExpressivenessComplexityRealTime2008} requires intervals to have finite length.
Intuitively, in the interval-based semantics, the system is observed at every instant in time.
On the other hand, in the point-based semantics, formulas are interpreted over timed words, which can be understood as a sequence of snapshots of the system.
In the point-based semantics, satisfiability and model checking over finite timed words is decidable~\parencite{ouaknineDecidabilityMetricTemporal2005}.
For a given \ac{MTL} formula $\phi$, the approach constructs an \acfi{ATA} which accepts an input word if and only if the word satisfies the specification $\phi$.
 As we will us a similar construction for synthesis, we describe the approach in greater detail in \autoref{sec:ata}.
If restricted to \emph{Safety \ac{MTL}}, which restricts \ac{MTL} to safety properties by requiring a bounded interval on the until operator (while allowing unbounded dual-until operators), both problems are decidable even over infinite words~\parencite{ouaknineSafetyMetricTemporal2006}.
As the point-based semantics corresponds to the situation calculus, where the system is only observed whenever an action occurs, we will extend \es with MTL-like temporal formulas in \autoref{chap:timed-esg} after describing \ac{MTL} in greater detail in \autoref{sec:mtl}.

Regarding verification of \golog programs, one advantage of \golog is that the program semantics is already defined in a logical framework.
Therefore, there is no need to define a model of the program, but one can instead verify \golog programs directly in the same logical representation that are used for the control of the agent~\parencite{classenPlanningVerificationAgent2013}.
Early work on verification relied on manual proofs: \textcite{degiacomoNonterminatingProcessesSituation1997} describe properties of non-terminating programs in terms of $\mu$-calculus formulas, \textcite{liuHoarestyleProofSystem2002} describe a proof system for \golog programs based on Hoare logic.
\emph{CASL} (Cognitive Agents Specification Language) is a proof system that assists the user to verify properties of multi-agent systems specified in \golog programs.
\textcite{classenLogicNonterminatingGolog2008} describe a system based on characteristic graphs that is able to verify properties of non-terminating \golog programs automatically.
However, as the verification problem is in general undecidable, the system does not always terminate.
Later work identified fragments of \golog that allowed decidable verification: \textcite{classenExploringBoundariesDecidable2014} show that verification of CTL properties is decidable in the two-variable fragment if all successor state axioms are context-free or local-effect and the pick operator is restricted to finite sets, similarly for LTL-like properties~\parencite{zarriessDecidabilityVerifyingLTL2014} and \ctlstar properties \parencite{zarriessVerifyingCTLProperties2014}.
These results have been extended to show that verification is also decidable if the basic action theory is acyclic (i.e., no cyclic dependencies between fluents in the effect descriptors) or flat (quantifier-free effect descriptors)~\parencite{zarriessDecidableVerificationGolog2016}.
Similarly, verification of decision-theoretic programs in \textsc{DTGolog} is decidable for acyclic theories.
\emph{Bounded theories}, where the number of objects described by any situation is bounded, also results in decidable verification of $\mu$-calculus properties~\parencite{degiacomoBoundedSituationCalculus2016}.
Finally, as a negative result, \textcite{liuProjectionProbabilisticEpistemic2022} has shown that verification of PCTL~\parencite{hanssonLogicReasoningTime1994} properties in belief programs based on the logic \ds is undecidable even for context-free successor state axioms, but has also identified a decidable fragment of the logic.
In \autoref{chap:synthesis}, we will investigate the verification of \ac{MTL} properties in \golog programs.

Related to verification is \emph{realizability} as well as \emph{synthesis}, which can both be described with two-player games between the system and the environment.
Given a specification (e.g., \iac{LTL} formula) and a partition of the alphabet into controllable and uncontrollable symbols, both players alternately choose a subset of their symbols.
If a player can always choose symbols such that the resulting word satisfies the specification, the player has a winning strategy.
The \emph{realizability problem}~\parencite{abadiRealizableUnrealizableSpecifications1989} is to determine whether the system has a winning strategy.
The \emph{synthesis problem}~\parencite{pnueliSynthesisReactiveModule1989} is to produce such a winning strategy if it exists.
\ac{LTL} synthesis is known to be \textsc{2ExpTime}-complete~\parencite{pnueliSynthesisReactiveModule1989} and tools such as \textsc{Lily}~\parencite{jobstmannOptimizationsLTLSynthesis2006}, \textsc{Unbeast}~\parencite{ehlersUnbeastSymbolicBounded2011}, and \textsc{Acacia+}\parencite{bohyAcaciaToolLTL2012} apply sophisticated techniques to obtain practical synthesis tools.
Apart from \ac{LTL} synthesis, several approaches synthesize controllers for timed automata.
\textsc{SynthKro} and \textsc{FlySynth} \cite{altisenToolsControllerSynthesis2002} synthesize controllers that remain in or reach a given set of states of a timed automaton.
\textsc{Uppaal-Tiga} \cite{behrmannUPPAALTigaTimePlaying2007} and \textsc{Synthia} \cite{peterSynthiaVerificationSynthesis2011}
control timed automata against a TCTL specification to accomplish reachability or safety.
\textsc{Uppaal-Tiga} has also been extended to models with partial observability \cite{finkbeinerTemplateBasedControllerSynthesis2012}, using pre-defined controller templates.
\textsc{Casaal} \cite{liPracticalControllerSynthesis2017} synthesizes a controller for \mtlzi specifications.
\textcite{bouyerControllerSynthesisMTL2006} show that \ac{MTL} controller synthesis is decidable on finite words for full \ac{MTL} specifications by constructing the synchronous product of the timed automaton and the alternating timed automaton that recognizes the \ac{MTL} specification.
We will use a similar approach in \autoref{chap:synthesis} for synthesizing a controller for a \golog program.

Synthesis has also seen recent interest in the AI community.
For one, \ac{LTL} has been used to describe temporally extended goals for planning~\parencite{bacchusPlanningTemporallyExtended1998,degiacomoAutomataTheoreticApproachPlanning2000,geffnerConciseIntroductionModels2013}, possibly resulting in infinite plans~\parencite{patriziComputingInfinitePlans2011}.
\ac{LTL} can also be used to specify \emph{conformant planning} problems with temporally extended goals~\parencite{calvaneseReasoningActionsPlanning2002}, where the plan is guaranteed to satisfy the goal even if the information about the system is incomplete.
Furthermore, there has been a particular interest in \ac{LTLf}~\parencite{degiacomoLinearTemporalLogic2013}, where the synthesis problem can be solved by transforming the \ac{LTL} specification into a non-deterministic finite automaton, which is subsequently determinized~\parencite{degiacomoSynthesisLTLLDL2015}.
Like the synthesis problem over infinite traces, the \ac{LTLf} synthesis problem is \textsc{2ExpTime}-complete~\parencite{degiacomoLinearTemporalLogic2013}, although \ac{LTLf} synthesis tools usually perform much better than \ac{LTL} synthesis tools.
The \ac{LTL} synthesis problem is also closely related to \emph{Fully Observable Nondeterministic} (FOND) planning~\parencite{camachoNonDeterministicPlanningTemporally2017,camachoFiniteLTLSynthesis2018,degiacomoAutomatatheoreticFoundationsFOND2018} as the nondeterministic effect of an action can be seen as an environment action.
The synthesis approach can also be extended to partially observable environments~\parencite{degiacomoLTLfLDLfSynthesis2016} and to best-effort strategies~\parencite{aminofBestEffortSynthesisDoing2021} without increasing the computational complexity.
So far, these synthesis approaches have focused on discrete time.
In this thesis, we will combine high-level reasoning in \golog with synthesis on real-time traces based on \ac{MTL} to obtain program realizations that satisfy a given \ac{MTL} specification.

\section{Abstraction}\label{sec:related:abstraction}
\textcite{giunchigliaTheoryAbstraction1992} define abstraction generally as a mapping between a ground and an abstract formal system, such that the abstract representation preserves desirable properties while omitting unnecessary details to make it simpler to handle.
Abstraction has been widely used in several fields of AI \cite{saittaAbstractionArtificialIntelligence2013}.
\Ac{HTN} planning systems such as \textsc{SHOP2} \cite{nauSHOP2HTNPlanning2003} decompose tasks into subtasks to accomplish some overall objective, which has also been used in the situation calculus \cite{gabaldonProgrammingHierarchicalTask2002}.
Macro planners such as \textsc{MacroFF}~\cite{boteaMacroFFImprovingAI2005} combine action sequences into macro operators to improve planner performance, e.g., by collecting action traces from plan executions on robots \cite{hofmannInitialResultsGenerating2017}, or by learning them from training problems \cite{chrpaMUMTechniqueMaximising2014}.
Similarly, \textcite{saribaturOmissionbasedAbstractionAnswer2021} use abstraction in Answer Set Programming to reduce the search space, improving solver performance.
\textcite{cuiUniformAbstractionFramework2021} leverage abstraction for generalized planning, i.e., for finding general solutions for a set of similar planning problems.
Abstraction has also been used to analyze causal models \cite{rubensteinCausalConsistencyStructural2017,banihashemiActionsProgramsAbstract2022}. Of particular interest for this work is the notion of \emph{constructive abstraction} \cite{beckersAbstractingCausalModels2019}, where the refinement mapping partitions the low-level variables such that each cell has a unique corresponding high-level variable.
\textcite{holtzenSoundAbstractionDecomposition2018} describe an abstraction framework for probabilistic programs and also describe an algorithm to  generate abstractions.
%
\textsc{REBA}~\cite{sridharanREBARefinementbasedArchitecture2019} is a framework for robot planning that uses abstract and deterministic ASP programs to determine a course of action, which are then translated to POMDPs for execution.
Abstraction has also been used in reinforcement learning to define a hierarchy of MDPs~\cite{cipolloneExploitingMultipleAbstractions2023}, where the lowest-level abstraction accurately captures the environment dynamics, while high-level models abstract away more and more details.
Using such a hierarchy for reinforcement learning increases the sample efficiency of RL algorithms.
Of particular relevance for this thesis is the work by \textcite{banihashemiAbstractionSituationCalculus2017}, who describe a general abstraction framework based on the situation calculus, where a refinement mapping maps a high-level \ac{BAT} to a low-level \ac{BAT} and which is capable of online execution with sensing actions \cite{banihashemiAbstractionAgentsExecuting2018}.
The framework has been used to effectively synthesize plan process controllers in a smart factory scenario~\cite{degiacomoSituationCalculusController2022} and has also been extended to non-deterministic actions~\cite{banihashemiAbstractionNondeterministicSituation2023}.
In contrast to our approach in \autoref{chap:abstraction}, they assume non-probabilistic actions.
On the other hand, \textcite{belleAbstractingProbabilisticModels2020} defines abstraction in a probabilistic but static propositional language and describes a search algorithm to derive such abstractions.
In \autoref{chap:abstraction}, we build on the two approaches to obtain abstraction in a probabilistic and dynamic first-order language with an unbounded domain.


\chapter{Foundations}\label{chap:foundations}

In this chapter, we provide the logical foundations for this thesis by discussing foundational concepts related to \emph{reasoning about actions} as well as \emph{timed systems}.
We start with the \sitcalc in \autoref{sec:sitcalc}, which is a well-established formalism to describe the actions of an agent and the changes that those actions bring to the world.
On the other hand, in \autoref{sec:timed-systems}, we summarize common concepts for \emph{reasoning about time}.
We first give an overview on temporal logics, introduce \emph{timed automata}, which extend finite automata with real time, and then describe \emph{alternating timed automata}, an extension of timed automata that allows to construct an automaton that accepts precisely those words that satisfy a given \ac{MTL} formula.
We will combine the situation calculus with concepts from timed systems in \autoref{chap:timed-esg} and use \aclp{TA} as well as \aclp{ATA} in \autoref{chap:synthesis} for synthesis.
Timed automata will also be used in \autoref{chap:transformation-as-reachability-problem} for plan transformation.

\section{The Situation Calculus}\label{sec:sitcalc}

The \sitcalc is one of the most commonly used formalisms for representing dynamically changing worlds.
It was originally introduced by \textcites{mccarthySituationsActionsCausal1963,mccarthyPhilosophicalProblemsStandpoint1969} and later refined by \textcite{reiterKnowledgeActionLogical2001}.
In the \sitcalc, all changes to the world are the result of \emph{actions}.
The state of the world is represented by a first-order term called \emph{situation}, which is a sequence of actions and can be seen as a history of actions that have occurred so far.
Therefore, a situation not only describes  the current state of the world, but also the actions that have lead to the current state.
The special situation $S_0$ represents the initial situation, which is the empty sequence of actions.
All successor situations are obtained from $S_0$ by a distinguished binary function symbol \doop, where $\doop(\alpha, s)$ describes the situation that results from doing action $\alpha$ in situation $s$.
An action is a $k$-ary function symbol where the $k$ function arguments are the action parameters.
As an example, the action \pick is a unary function symbol, where $\pick(o)$ is the action of picking up the action's (only) parameter $o$.

The current state of the world is described with relations and functions.
Relations whose truth value may change from situation to situation are called \emph{relational fluents}.
Relational fluents take a situation term as their last argument.
As an example, the relational fluent $\holding(o, \doop(\pick(o), S_0))$ describes that a robot is holding the object $o$ in the situation resulting from doing the action $\pick(o)$ in the initial situation $S_0$.
Similarly, functions whose truth value may change from situation to situation are called \emph{functional fluents}.
Analogous to relational fluents, a functional fluent takes a situation term as last argument.
As an example, the functional fluent $\at(\doop(\goto(\kitchen), S_0))$ describes the position of the robot after going into the kitchen with the action $\goto(\kitchen)$.
In addition to relational and functional fluents, \emph{rigid} functions and relations describe unchanging properties of the world, e.g., the rigid function $\distance(l_1, l_2)$ gives the distance between the two locations $l_1$ and $l_2$ and $\connected(l_1, l_2)$ states that $l_1$ and $l_2$ are connected.
In contrast to fluents, rigid functions and relations do not carry a situation term as last argument.

\begin{figure}[htb]
  \centering
  \includestandalone[width=\textwidth]{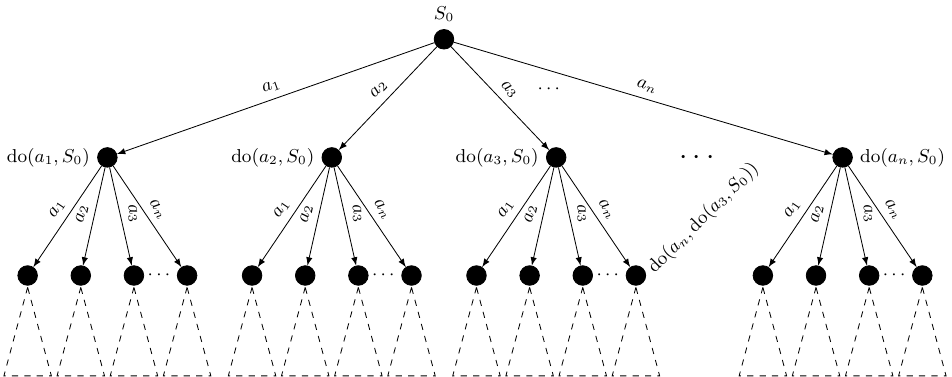}
  \caption[The tree of situations.]{The tree of situations for a model with $n$ actions (adapted from \parencite{reiterKnowledgeActionLogical2001}).}
  \label{fig:situation-tree}
\end{figure}

\subsection{Basic Action Theories}\label{sec:sitcalc-bats}
A \acfi{BAT} describes a domain by axiomatizing its actions.
A \ac{BAT} \bat consists of the following axioms~\parencites{pirriContributionsMetatheorySituation1999,reiterKnowledgeActionLogical2001}:
\begin{description}
  \item[Foundational axioms:]
    The foundational axioms are domain-independent axioms that characterize situations, the successor function $\doop$, and the relation $\sqsubset$, which provides an ordering on situations:%
    \footnote{In the following, free variables will always be implicitly universally quantified from the outside.}
    \begin{align*}
      &\doop(a_1, s_1) = \doop(a_2, s_2) \supset a_1 = a_2 \wedge s_1 = s_2
      \\
      &(\forall P). P(S_0) \wedge (\forall a, s) [P(s) \supset P(\doop(a, s))] \supset (\forall s) P(s)
      \\
      &\neg s \sqsubset S_0
      \\
      &s \sqsubset \doop(a, s') \equiv s \sqsubseteq s'
    \end{align*}
    The first axiom is a unique names axiom for situations: if the resulting situation of doing action $a_1$ in situation $s_1$ is the same as the resulting situation of doing action $a_2$ in situation $s_2$, then $a_1$ and $a_2$ as well as $s_1$ and $s_2$ must be equal.
    Therefore, the history of actions uniquely defines the situation and it is not possible to reach the same situations via two different sequences of actions.
    The second axiom is a second-order induction axiom that limits the situations to the smallest set that contains $S_0$ and that is closed under the function $\doop$.
    The third and fourth axiom axiomatize subhistories: There is no situation that is a subhistory of $S_0$, therefore $S_0$ is the minimal element with respect to $\sqsubset$.
    In the fourth axiom, $s \sqsubseteq s'$ is to be understood as abbreviation for $s \sqsubset s' \vee s = s'$.
    The axiom states that  $s$ is a subhistory of $\doop(a, s')$ if it is a subhistory of $s'$ or if $s$ and $s'$ are the same.
    It therefore axiomatizes $\sqsubset$ as transitive closure of the successor situations as defined by the function $\doop$.

    One consequence of the basic properties of situations is that the situations in any model can be represented by a tree, as shown in \autoref{fig:situation-tree}.
  \item[Initial situation:] The initial situation is defined by a set of first-order sentences $\bat_0$, where $S_0$ is the only term of sort situation mentioned in the sentences of $\bat_0$.
    As the name suggests, these axioms specify the state of the world before any action has been executed.
    As an example, the following axioms state that the robot is initially in the hallway and is not holding any object:
    \begin{align*}
      &\robotat(S_0) = \hallway
      \\
      &\forall  o\; \neg \holding(o, S_0)
    \end{align*}
  \item[Action precondition axioms:] For each action, the \ac{BAT} contains a single axiom that describes the precondition of the action.
    A precondition axiom for action $A$ has the following form:
    \[
      \poss(A(\vec{x}), s) \equivspace \Pi_A(\vec{x}, s)
    \]
    Here, $\Pi_A(\vec{x}, s)$ is a first-order formula with free variables among $\vec{x}, s$.
    As an example, the precondition axiom for the action \goto may look as follows:
    \[
      \poss(\goto(\mi{start}, \mi{goal}), s) \equivspace \at(s) = \mi{start}
    \]
    It states that the robot can move from $\mi{start}$ to $\mi{goal}$ if and only if the robot is currently in location $\mi{start}$.
  \item[Successor state axioms:] For each relational fluent, the \ac{BAT} contains a single \emph{successor state axiom} of the following form:
    \[
      F(\vec{x}, \doop(a, s)) \equivspace \gamma_F^+(\vec{x}, a, s) \vee F(\vec{x}, s) \wedge \neg \gamma_F^-(\vec{x}, a, s)
    \]
    Here, $\gamma_F^+(\vec{x}, a, s)$ and $\gamma_F^-(\vec{x}, a, s)$ are first-order formulas with free variables among $\vec{x}, a, s$.
    Intuitively, a successor state axiom states the following: After doing action $a$ in situation $s$, the relational fluent $F(\vec{x}, \doop(a, s))$ is true if the action $a$ makes it true (expressed with the formula $\gamma_F^+(\vec{x}, a, s)$), or if it was true before and the action $a$ does not cause it to be false (expressed with the formula $\gamma_F^-(\vec{x}, a, s)$). As an example, the successor state axiom for the fluent \holding may look as follows:
    \[
      \holding(o, \doop(a, s)) \equivspace a = \pick(o) \vee \holding(o, s) \wedge a \neq \putact(o)
    \]

    Furthermore, for each functional fluent, the \ac{BAT} contains a single successor state axiom of the following form:
    \[
      f(\vec{x}, \doop(a, s)) = y \equivspace \gamma_f(\vec{x}, y, a, s) \vee f(\vec{x}, s) = y \wedge \neg \exists y' \gamma_f(\vec{x}, y', a, s)
    \]
    Here, $\gamma_f(\vec{x}, y, a, s)$ is a first-order formula with free variables among $\vec{x}, y, a, s$.
    Similar to relational fluents, $\gamma_f(\vec{x}, y, a, s)$ describes how the action $a$ affects the value of the fluent $f(\vec{x}, \doop(a, s))$.
    After doing action $a$, the fluent has the value $y$ if $a$ causes the value (i.e., $\gamma_f(\vec{x}, y, a, s)$ is true), or if the fluent had the value $y$ before (i.e., $f(\vec{x}, s) = y$ is true) and the action does not cause any other value (i.e., $\exists y' \gamma_f(\vec{x}, y', a, s)$ is false).
    As an example, the successor state axiom for the functional fluent $\at$ may look as follows:
    \[
      \at(\doop(a, s)) = l \equivspace a = \goto(l) \vee \at(s) = l \wedge \neg \exists l' a = \goto(l')
    \]
    For functional fluents, one typically requires the \emph{functional fluent consistency property}~\cite{pirriContributionsMetatheorySituation1999}, which states that for every situation, $\gamma_f$ actually defines a value for $f$ and that this value is unique.
  \item[Unique name axioms for actions:] Additionally, the \ac{BAT} contains axioms that ensure that each action has a unique name, i.e., two distinct actions are not equal.
    For each pair of distinct action names $A, B$, the \ac{BAT} contains the following axioms:
    \begin{align*}
      &A(\vec{x}) \neq B(\vec{y})
      \\
      &A(\vec{x}) = A(\vec{y}) \supset \vec{x} = \vec{y}
    \end{align*}
\end{description}

This formulation of a \ac{BAT} contains two assumptions:
\begin{itemize}
  \item The action precondition describes all the necessary and sufficient conditions for an action to be possible.
    In particular, there are no additional conditions not mentioned in the precondition axiom (e.g., the robot's motor being broken) that may render an action impossible.
    This is a solution to the \emph{qualification problem}~\parencite{mccarthyEpistemologicalProblemsArtificial1981}.
  \item The successor state axioms describe all the conditions under which an action may cause a change of a fluent value.
    In other words, there may be no additional action that has an effect on fluent values and that is not described in the \ac{BAT}.
    Also, every change of a fluent value is caused by an action.
    This \emph{causal completeness assumption} is a solution to the \emph{frame problem}~\parencite{mccarthyPhilosophicalProblemsStandpoint1969} and was first described by \textcite{reiterFrameProblemSituation1991}.
\end{itemize}

An important property of \acp{BAT} is \emph{relative satisfiability}~\cite{pirriContributionsMetatheorySituation1999}: If the consistency condition for functional fluents is satisfied, then a \ac{BAT} \bat is satisfiable if and only if the initial situation $\bat_0$ and the unique name axioms $\bat_{\text{una}}$ are satisfiable.
Therefore, given a satisfiable initial situation and unique name axioms, augmenting those with the foundational axioms of the situation calculus as well as with action precondition and successor state axioms may not lead to an unsatisfiable theory.

\subsection{Projection}%
\label{ssub:Projection}
One of the most ubiquitous tasks in the context of the situation calculus is \emph{projection}: Given a \ac{BAT} \bat, a sequence of actions $\sigma = \la a_1, \ldots, a_n \ra$, and a formula $\alpha$, the projection problem is to determine an answer to the following question:
\begin{displayquote}
  Will $\alpha$ hold after executing the action sequence $\sigma$, given the \ac{BAT} \bat?
\end{displayquote}
In the situation calculus, this corresponds to the query:\footnote{
  The short-hand notation $\doop([a_1, \ldots, a_n], S_0)$ stands for $\doop(a_n, \doop(a_{n-1}, \doop(\ldots, \doop(a_1, S_0)\cdots))$.
}
\[
  \bat \overset{?}{\models} \alpha(\doop([a_1, \ldots, a_n], S_0))
\]
One common approach to solve the projection problem is \emph{regression}~\parencite{waldingerAchievingSeveralGoals1981,reiterFrameProblemSituation1991}.
The idea of regression is to reduce a query about the future to a query about the initial situation.
More specifically, in regression, the formula $\alpha$ is transformed into a formula $\alpha'$ such that $\alpha$ holds after the actions $\sigma$ if and only if $\alpha'$ holds in the initial situation.
As the successor state axioms uniquely specify the effects of an action on a fluent, we may replace each fluent occurring in $\alpha$ by the right-hand side $\gamma_F^+(\vec{x}, a, s) \vee F(\vec{x}, s) \wedge \neg \gamma_F^-(\vec{x}, a, s)$ of the successor state axiom, where we substituted $a$ by the action $a_n$.
By doing so, we may get rid of the last action in the sequence, i.e., the resulting formula will hold after the actions $\la a_1, \ldots, a_{n-1} \ra$ (without the last action $a_n$) if and only if the original formula is satisfied after the whole action sequence.
If we apply this operation iteratively, we obtain a formula $a'$ that only mentions the situation $S_0$ and so we only need to check if $\alpha'$ holds in the initial situation.
\textcite{reiterFrameProblemSituation1991} has shown that this form of regression in the situation calculus is sound and complete, i.e., every query can be transformed into a query about the initial situation.
Generally, regression may result in an exponential blowup.
However, if the \ac{BAT} is \emph{context-free}, i.e., each $\gamma^+$ and $\gamma^-$ is independent of the current situation, then regression adds at most linear complexity to the query~\cite{reiterKnowledgeActionLogical2001}.

Nevertheless, regression has some drawbacks~\cite{levesqueCognitiveRobotics2008}:
In a long-lived agent, regressing over thousands of actions is often infeasible.
Additionally, if the agent needs to answer many queries, then regression is impractical, because each query needs to be regressed separately.
Therefore, \emph{progression}~\cite{linHowProgressDatabase1997} has been developed as an alternative approach to projection.
In progression, rather than modifying the formula $\alpha$, we compute a new \ac{BAT} $\bat'$ that represents a new initial situation that corresponds to the situation after the action sequence $\sigma$.
One advantage of progression is that it only needs to be done once and therefore avoids duplicated work if multiple queries need to be answered.
Also, if the agent progresses the \ac{BAT} frequently enough, it does not suffer from a long history of actions.
One drawback of progression is that for the general case, it requires second-order logic to characterize the progressed \ac{BAT}~\parencite{linHowProgressDatabase1997}.
Therefore, several restrictions of \acp{BAT} have been investigated that allow a first-order definition of regression.
As an example, a \emph{local-effect}~\ac{BAT}~\cite{liuTractableReasoningIncomplete2005} restricts a \ac{BAT} such that the effects of an action exclusively depend on the action's parameters, in which case progression is first-order definable \cite{liuTractableReasoningIncomplete2005,vassosProgressionSituationCalculus2007} and in the case of proper\textsuperscript{+} knowledge bases \parencite{lakemeyerEvaluationbasedReasoningDisjunctive2002} even efficiently computable \parencite{liuFirstorderDefinabilityComputability2009,belleProjectionProblemActive2012}.

In \autoref{chap:timed-esg}, we will define a variant of regression in the logic \tesg that allows to regress a formula over a sequence of timed actions, which will be used in the synthesis approach described in \autoref{chap:synthesis}.

\subsection{Time and Durative Actions in the Situation Calculus}%
\label{ssub:sitcalc-time}

When describing real-world systems, time often plays an important role, e.g., because an action may have a certain duration.
For this reason, the situation calculus has been extended with an explicit notion of time, where each situation occurs at a real-valued time point~\cite{pintoTemporalReasoningSituation1994,pintoReasoningTimeSituation1995,reiterKnowledgeActionLogical2001}.
To formalize this, each action has an additional time argument, e.g., $\pick(o, t)$ is the action of picking up object $o$ at time point $t$.
A new function symbol $\ztime$ then specifies the time of occurrence of an action, i.e.,
\[
  \ztime(A(\vec{x}, t)) = t
\]
Using $\ztime(\cdot)$, one can add an axiom that defines the start time of a situation $s$:
\[
  \start(\doop(a, s)) = \ztime(a)
\]
By assuming the standard interpretation for the real numbers and its operands ($+, -, <$, etc.), it is possible to express properties such as ``the action \emph{put} occurs two seconds after the action \emph{pick}'':
\[
  \start(\doop(\putact(o, 4), \doop(\pick(o, 2), S_0))) = \start(\doop(\pick(o, 2), S_0)) + 2
\]
Additionally, it is possible to axiomatize an \emph{actual path of situations}, which is a sequence of situations that have actually occurred~\cite{pintoReasoningTimeSituation1995}:
\begin{align*}
    & \actual(S_0)
    \\
    & \actual(\doop(a, s)) \supset \actual(s) \wedge \poss(a, s)
    \\
    & \actual(\doop(a_1, s)) \wedge \actual(\doop(a_2, s)) \supset a_1 = a_2
\end{align*}
Given an actual path of situations, $\occurs(a, s)$ describes that $a$ is an occurring action and $\occurst(a, t)$ gives the time point $t$ when the action $a$ occurs:
\begin{align*}
    \occurs(a, s) &\equivspace \actual(\doop(a, s))
    \\
    \occurst(a, t) &\equivspace \exists s.\, \occurs(a, s) \wedge \sstart(\doop(a, s)) = t
\end{align*}
This is particularly useful for modeling durative actions.
To incorporate durative actions in the situation calculus, \textcite{pintoTemporalReasoningSituation1994} proposed to split each durative action into two instantaneous actions \emph{start} and \emph{end}, e.g., $\sac{\pick(o, t)}$ and $\eac{\pick(o, t)}$.
The action duration can then be modeled as part of the precondition of the end action, e.g.,
\[
  \poss(\eac{pick(o, t)}) \equivspace \rfluent{picking}(o) \wedge \exists t'.\, \occurst(\sac{pick(o, t')}, t') \wedge t \geq t' + 2
\]

While this extension augments situations and actions with time, fluents are still atemporal.
This may pose a limitation if dealing with continuous fluents, e.g., the functional fluent $\distance(l)$, which describes the distance of the robot to some location $l$.
As any fluent value is evaluated only when an action occurs, it is not directly possible to query for a certain time, or for a situation where the fluent takes a certain value, e.g., $\distance(\kitchen) = 1.5$.
To allow the former, \textcite{soutchanskiExecutionMonitoringHighlevel1999} introduces an auxiliary function $\action{wait}(t)$ that waits until time point $t$ has been reached.
Similarly, \textcite{grosskreutzCcGologActionLanguage2003} introduce an auxiliary action $\waitfor(\phi)$ to wait for a condition $\phi$ to become true.
This allows to query for exact time points where a certain condition is satisfied, e.g., $\waitfor(\distance(\kitchen) = 1.5)$.

\subsubsection{Concurrency}
Using \emph{start} and \emph{end} actions, it is possible to define \emph{interleaved concurrency} of actions, e.g., the action sequence
\[
  \sac{\goto(\kitchen, 2)}, \sac{\calibrate(3)}, \eac{\calibrate(4)}, \eac{\goto(\kitchen, 5)}
\]
expresses that the robot calibrates its arm while it is moving to the kitchen.
However, this does not allow for two (instantaneous) actions to occur simultaneously.\footnote{
  There is a subtle difference between two actions occurring at the same time, e.g., $\doop(\sac{\goto(\kitchen, 2)}, \doop(\sac{\calibrate(2)}, S_0))$ and two actions occurring simultaneously, e.g., $\doop(\{ \sac{\goto(\kitchen, 2)}, \sac{\calibrate(2)} \}, S_0)$.
  In the former, there is a situation where the robot is calibrating its arm but not yet moving, which does not occur in the latter.
}
\textcite{reiterNaturalActionsConcurrency1996} describes an approach to model \emph{true concurrency} in the situation calculus by allowing multiple (possibly infinitely many) actions occurring simultaneously.
In order to do so, the \doop operator does not take a single action but instead a set of concurrent actions as its first argument, e.g., $\doop(\{\sac{\goto(\kitchen, 2)}, \sac{\calibrate(2)}\}, S_0)$ is the resulting situation after starting the actions $\goto(\kitchen)$ and $\calibrate()$ simultaneously.
This form of concurrency brings some complications, as it may be impossible to execute two actions simultaneously even if each action by itself is possible in the current situation, e.g., $\sac{\goto(\kitchen)}$ and $\sac{\goto(\hallway)}$.
In order to deal with this issue, \textcite{reiterNaturalActionsConcurrency1996} proposes to axiomatize \emph{coherent} sets of actions, which allows to exclude any actions that cannot be done simultaneously.

\paragraph{Hybrid Systems in the Situation Calculus}
Instead of using auxiliary actions, \textcite{batusovHybridTemporalSituation2019} extend the situation calculus with \emph{state evolution axioms}, which describe the continuous change of a fluent within a given situation while no action occurs.
This allows to model continuously changing fluents, e.g., $\distance(l)$ without the need to query for a specific value explicitly.
State evolution axioms consist of \emph{temporal change axioms} of the following form:
\[
  \gamma(\vec{x}, s) \wedge \delta_f(\vec{x}, y, t, s) \supset f(\vec{x}, t, s) = y
\]
Here, $\gamma(\vec{x}, s)$ is the \emph{context}, which specifies when the formula $\delta_f(\vec{x}, y, t, s)$ is to be used to determine the value of the fluent $f(\vec{x})$.
The formula $\delta_f(\vec{x}, y, t, s)$ defines how the value $y$ of the fluent $f(\vec{x})$ changes with time $t$ while being in situation $s$.
As $\delta_f$ is an arbitrary formula, it may also encode arbitrary equations, e.g., differential equations.
This allows to embed hybrid systems into the situation calculus~\cite{batusovHybridTemporalSituation2019}.

\subsection{The Epistemic Situation Calculus}
So far, we have assumed that the agent's actions only affect the external world and that the agent knows the truth value of all fluents.
However, especially in robotics, some fluent value may be initially unknown to the agent, e.g., an object may be located in the kitchen, but the agent does not know that fact.
In order to gather additional information, the agent can use \emph{sensing}, e.g., it may use some object detection component to sense whether an object is nearby.
Such sensing actions do not affect the external world, but instead the agent's mental state.
As a sensing action makes a fluent value to be known, it is also called a \emph{knowledge-producing action}.

To formalize (incomplete) knowledge and knowledge-producing actions, \textcite{mooreReasoningKnowledgeAction1981} proposes to adapt the possible-world semantics known from modal logic \parencites{kripkeSemanticalAnalysisModal1963,hintikkaKnowledgeBelief1969,garsonModalLogic2021} to the situation calculus.
Propositional modal logic extends propositional logic with modal operators to express \emph{necessity} and \emph{possibility}, where $\square \phi$ is to be understood as ``it is necessary that $\phi$'' and $\diamond \phi$ as ``it is possible that $\phi$''.
In the possible-world semantics for modal logic, the truth of a sentence is determined by a set of possible worlds $W$ with one element $w$ being the ``real'' world.
A sentence $\square \phi$ is true if it is true in all the worlds of $W$.
Similarly, a sentence $\diamond \phi$ is true if it is true in some world of $W$.

Coming back to the \sitcalc, rather than extending the \sitcalc with modal operators, \textcite{mooreReasoningKnowledgeAction1981} proposes to treat situations as possible worlds.
Given the current situation $s$, a binary relational fluent $K(s', s)$ defines the accessible situations $s'$ from $s$, analogously to the set of possible worlds $W$ and the real world $w$ as described above.
Knowledge can then be defined based on the fluent $K$, where $\sitcalcknow(\phi, s)$ expresses that $\phi$ is known in situation $s$ and is defined as follows~\cite{scherlFrameProblemKnowledgeProducing1993,reiterKnowledgeActionLogical2001}:%
\footnote{
  The notation $\phi[s']$ means the result of restoring the situation argument $s'$ to all fluents mentioned by the formula $\phi$.
}
\[
  \sitcalcknow(\phi, s) \eqdef \forall s'.\, K(s', s) \supset \phi[s']
\]

A situation $s'$ is accessible ($K(s', s)$) if it is considered to be a possible alternative to the current situation.
As $K$ is a relational fluent, it is also defined using successor state axioms.
\textcite{scherlFrameProblemKnowledgeProducing1993} describe a definition for $K$ that extends Reiter's solution to the frame problem to knowledge-producing actions.
To do so, they first distinguish knowledge-producing actions from regular actions and require that each action either affects the external world or the agent's knowledge, but not both.
Given $m$ sensing actions $\sense_{\psi_1}, \ldots \sense_{\psi_m}$ for the formulas $\psi_1, \ldots, \psi_m$ and $n$ sensing actions $\sensefun_{f_1}, \ldots \sensefun_{f_n}$ for functional fluents $f_1, \ldots, f_n$, the successor state axiom for $K$ can be defined as follows~\cite{reiterKnowledgeActionLogical2001}:
\begin{align*}
  K(s', \doop(a, s)) \equivspace
  & \exists s^*.\, s' = \doop(a, s^*) \wedge K(s^*, s) \wedge
  \\
  & \forall \vec{x}_1 \lbrack a = \sense_{\psi_1}(\vec{x}_1) \supset \psi_1(\vec{x}_1, s^*) \equiv \psi_1(\vec{x}_1, s) \rbrack \wedge \ldots \wedge
  \\
  & \forall \vec{x}_m \lbrack a = \sense_{\psi_m}(\vec{x}_m) \supset \psi_m(\vec{x}_m, s^*) \equiv \psi_m(\vec{x}_m, s) \rbrack \wedge
  \\
  & \forall \vec{y}_1 \lbrack a = \sensefun_{f_1}(\vec{y}_1) \supset f_1(\vec{y}_1, s^*) = f_1(\vec{y}_1, s) \rbrack \wedge \ldots \wedge
  \\
  & \forall \vec{y}_n \lbrack a = \sensefun_{f_n}(\vec{y}_n) \supset f_n(\vec{y}_n, s^*) = f_n(\vec{y}_n, s) \rbrack
\end{align*}
This states that $s'$ is accessible from the situation $\doop(a, s)$ if
\begin{enumerate*}[label=(\arabic*)]
  \item $s'$ results from doing the action $a$ in some situation $s^*$ that is accessible from $s$,
  \item if $a$ is a sensing action for the formula $\psi_i(\vec{x}_i)$, then $s^*$ and $s$ must agree on the truth value of $\psi_i(\vec{x}_i)$,
  \item if $a$ is a sensing action for the functional fluent $f_i(\vec{y}_i)$, then $s^*$ and $s$ must agree on the value of $f_i(\vec{y}_i)$.
\end{enumerate*}


\subsubsection{A Modal Variant of the Epistemic \Sitcalc}\label{sec:es}

\textcite{lakemeyerSemanticCharacterizationUseful2011} describe \es, a modal variant of the \sitcalc that is able to express knowledge similar to the epistemic \sitcalc described above.
In \es, situations are part of the semantics but in contrast to the \sitcalc, situations do not appear as terms in the language.
Instead, possible worlds are built into the semantics, where the truth of a sentence is defined given a set of possible worlds $e$ (also called the epistemic state), the actual world $w$, and a sequence of executed actions $z$.
\es uses the modal operators $\lbrack a \rbrack \alpha$ to express that $\alpha$ is true after doing action $a$, $\square \alpha$ to state that $\alpha$ is true after any sequence of actions, and $\sitcalcknow(\alpha)$ to express that $\alpha$ is known.
\textcite{lakemeyerSemanticCharacterizationUseful2011} show that \es is indeed notational variant of the \sitcalc by mapping $\es$ sentences to \sitcalc sentences and then showing that valid sentences of \es can be cast into entailments of the \sitcalc.

The language of \es includes countably many standard names for both objects and actions and therefore fixes the domain of discourse to a countably infinite set.
Standard names can be understood as special constants that satisfy the unique name assumption, i.e., for any distinct standard names $n_i$ and $n_j$, $n_i \neq n_j$ is a valid sentence of \es.
Standard names simplify the interpretation of sentences with quantifiers.
In classical first-order logic, the semantics is usually defined by a structure, which consists of a non-empty domain of discourse $D$ and an interpretation $I$ that defines appropriate functions and relations for the function and predicate symbols.
A quantifier is then evaluated by using a variable assignment, which assigns each free variable to a domain element $d \in D$.
In contrast, standard names allow first-order quantification to be understood substitutionally, where a sentence $\exists x\, \phi(x)$ is true if and only if there is some standard name $n$ such that $\phi(n)$ is true.
As argued by \textcite{lakemeyerSemanticCharacterizationUseful2011}, standard names also considerably simplify proofs, especially when comparing different theories, as there is no need to map the domain of one structure into the domain of another.

Similar to the \sitcalc, the language contains relational and functional rigid as well as relational and functional fluent symbols.
As in the \sitcalc, fluents vary as the result of actions.
In contrast to the \sitcalc, situations do not occur as terms in the language.
Instead, the modal operator $[\cdot]$ is used to express a fluent value after doing some action.
As an example, the formula $[\pick(o)] \holding(o)$ expresses that the robot is holding some object $o$ after doing the action $\pick(o)$.

As in the epistemic \sitcalc, \es allows to model \emph{sensing actions}.
In contrast to the sensing actions described above, sensing actions and regular actions that have an effect of the world are not distinguished.
In fact, in \es, each action is assumed to have a binary sensing result, indicated by the predicate \sensefluent. 

As in the \sitcalc, a domain is axiomatized in a \ac{BAT}.
In \es, a \ac{BAT} consists of the following parts:\footnote{Free variables are implicitly forall-quantified from the outside and $\square$ has lower syntactic precedence than the logical connectives, e.g., $\square \poss(a) \equivspace \pi$ stands for $\forall a.\, \square(\poss(a) \equiv \pi)$.}
\begin{description}
  \item[Initial situation axioms]
    A set of fluent sentences $\Sigma_0$ describing the initial situation, e.g.,
    \[
      \neg \exists o \holding(o) \wedge \robotat = \kitchen
    \]
  \item[Action precondition axiom]
    A single sentence of the following form that specifies the precondition of all actions:
    \[
      \square \poss(a) \equivspace \pi
    \]
    Here, $\pi$ is a fluent formula, i.e., a formula with no $\sitcalcknow$, $\square$, $\poss$, or $\sensefluent$.
    As an example, the precondition axiom for a domain with the two actions \pick and \putact may look as follows:
    \begin{align*}
      \square \poss(a) \equivspace &\exists o.\, a = \pick(o) \wedge \robotat = \objat(o) \wedge \neg \exists o' \holding(o')
      \\
                                   &\vee \exists o.\, a = \putact(o) \wedge \holding(o)
    \end{align*}
  \item[Successor state axioms]
    For each relational fluent, a successor state axiom of the following form:
    \[
      \square [a] F(\vec{x}) \equivspace \gamma_F(\vec{x})
    \]
    Here, $\gamma_F(\vec{x})$ is a fluent formula with free variables among $a, \vec{x}$ and describes the conditions under which the fluent $F(\vec{x})$ becomes true.
    As an example, the successor state axiom for \holding may look as follows:
    \[
      \square [a] \holding(o) \equivspace a = \pick(o) \vee \holding(o) \wedge a \neq \putact(o)
    \]
    Additionally, for each functional fluent, a successor state axiom of the following form:
    \[
      \square [a] f(\vec{x}) = y \equivspace \gamma_f(\vec{x}, y)
    \]
    Here, $\gamma_f(\vec{x}, y)$ is a fluent formula with free variables among $a, \vec{x}, y$ and describes the conditions under which the fluent $f(\vec{x})$ has the value $y$.
    As an example, the successor state axiom for the robot's position \robotat may look as follows:
    \[
      \square [a] \robotat = l \equivspace a = \goto(l) \vee \robotat = l \wedge \neg \exists l'.\, a = \goto(l')
    \]
  \item[Sensing axioms]
    A single sentence that describes the sensing result of each action of the same form as the precondition axiom, i.e.,:
    \[
      \square \sensefluent(a) \equivspace \pi
    \]
    Again, $\pi$ is a fluent formula.
    As an example, the sensing axiom for a domain with the two actions \goto and \sonar, which detects whether the robot is close to a wall, may look as follows:
      \[
        \square \sensefluent(a) \equivspace \exists l\, a = \goto(l) \vee a = \sonar \wedge \distance < 5
      \]
    \item[Unique name axioms for actions]
      The \ac{BAT} also contains axioms that ensure that each action has a unique name.
      Note that as we use standard names, we can just assume that all action names are standard names.
      Alternatively, we can add axioms to ensure unique names, e.g.:
      \begin{align*}
        &\square \pick(o) \neq \putact(o)
        \\
        &\square \pick(o) = \pick(o') \supset o = o'
        \\
        & \ldots
      \end{align*}
\end{description}

Note that apart from sensing axioms, the \ac{BAT} does not contain any special axioms to deal with knowledge, unlike the epistemic \sitcalc described above, where we needed to axiomatize the accessibility relation $K$.
Instead, a model $e, w, z$ satisfies a formula $\sitcalcknow(\alpha)$ if and only if every world $w \in e$ satisfies $\alpha$.
We refer to \parencite{lakemeyerSemanticCharacterizationUseful2011} for the formal definition of the semantics of \es.

\subsection{Noisy Sensors and Effectors in the \Sitcalc}

\begin{figure}[ht]
  \centering
  \includegraphics[width=0.5\textwidth]{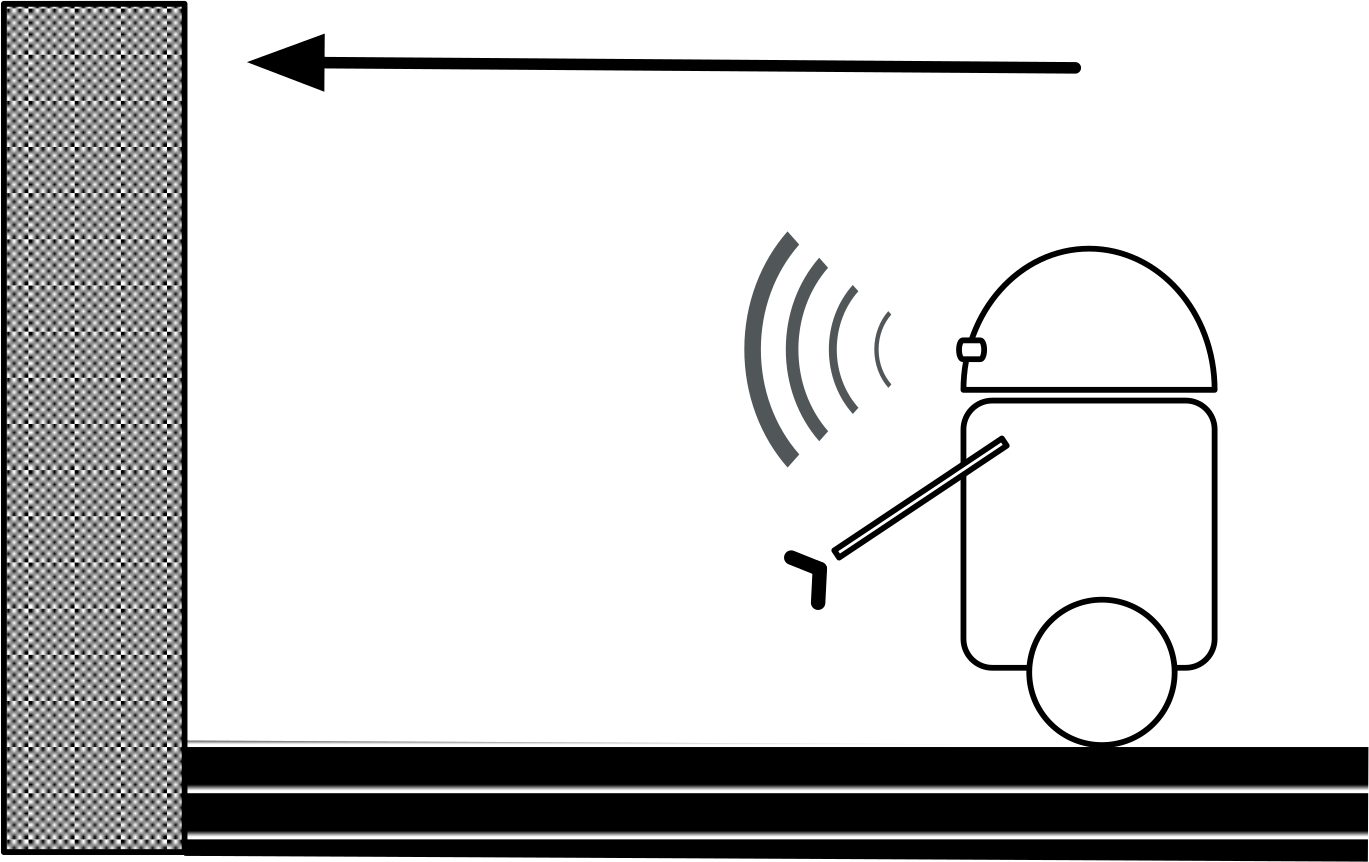}
  \caption[A robot driving towards a wall.]{A robot driving towards a wall~\cite{belleReasoningProbabilitiesUnbounded2017}.
  The robot can measure the distance to the wall with its action \sonar and it can move towards the wall with the action \move.
  Both actions are noisy: the sonar does not measure the exact distance and the \move action may move with further or shorter than intended.
}
  \label{fig:robot-wall}
\end{figure}

The epistemic \sitcalc and its modal variant \es already allow to model incomplete knowledge and therefore sensing actions based on a possible-world semantics.
However, they still assume that a sensor is noiseless and actions are deterministic, i.e., always have the same effect.
Both assumptions are often violated on a real robot.
Consider the simple robot shown in \autoref{fig:robot-wall} that is driving towards a wall and that is equipped with a sonar sensor, which can measure the distance to the wall.
The sonar is imprecise: it measures the correct distance $h$ with a probability of $0.8$ and measures $h$ with an error of $\pm 1$ with probability $0.1$.
Additionally, the action $\move(x)$, which moves the robot by a distance of $x$, is also imprecise, and the robot may instead move by a distance of $x \pm 1$ with a probability of $0.2$, without being able to detect how far it actually moved.
\textcite{bacchusReasoningNoisySensors1999} propose an extension to the epistemic \sitcalc that allows to model such a robot.
To model noisy actions, they propose to augment each action with additional arguments that express the action that was actually executed.
As an example, the noisy $\move(x, y)$ has two arguments: The argument $x$ expresses the nominal distance that the robot intends to move, the argument $y$ expresses the actual distance that the robot really moved.
Here, $x$ is determined by the agent, while $y$ is chosen by the environment.
Similarly, for the sensing action \sonar, the action's arguments are augmented with the measured distance, i.e., $\sonar(5)$ expresses that the robot measured a distance of $h = 5$.
As a second ingredient, \emph{observational indistinguishability axioms} define actions that the agent cannot tell apart.
For example, if the robot cannot detect how far it actually moved, the \ac{BAT} will contain the axiom
\[
  \oi(\move(x, y), a') \equivspace \exists z\, a' = \move(x, z)
\]

Furthermore, to axiomatize the probability of each action outcome, the \ac{BAT} also contains \emph{likelihood axioms}.
To state that the robot moves by the intended distance with probability $0.6$ and with an error of $\pm$ with probability $0.2$, the \ac{BAT} contains the following axiom:
\[
  l(\move(x, y), s) \eqspace
  \begin{cases}
    0.6 & \text{ if } x = y
    \\
    0.2 & \text{ if } \vert x - y \vert = 1
    \\
    0.0 & \text{ else }
  \end{cases}
\]

Additionally, to reason about the likelihood of a fluent having a certain value, \emph{knowledge} from the epistemic situation calculus is extended by \emph{degrees of belief}.
This is done by associating with each situation a \emph{weight}, and then using the normalized weight as degree of belief: for a formula $\phi$, the agent's degree of belief in $\phi$ is the total weight of all the situations where $\phi$ holds, normalized by the total weight of all possible situations.
This allows to express the agent's degree of belief that a certain property holds after doing some action, e.g., we may have:
\begin{equation}\label{eqn:bhl-example}
  \sitcalcbel(h = 3, \doop(\move(1, 1), \doop(\sonar(4), S_0))) = 0.625
\end{equation}
After measuring a distance of $4$ and then moving one step towards the wall, the agent's degree of belief that the robot is at distance $h = 3$ is $0.0625$.

\paragraph{The modal variant \ds}
Based on these concepts to model noisy sensors and effectors, \textcite{belleReasoningProbabilitiesUnbounded2017} extend the modal variant \es of the epistemic \sitcalc with degrees of belief.
Similar to \es, this allows simpler proofs of theoretical questions about knowledge, e.g., whether from $\know(\alpha) \supset (\know(\beta) \vee \know(\gamma))$ it follows that $\know(\alpha) \supset \know(\beta)$ or $\know(\alpha) \supset \know(\gamma)$.
\ds also uses the modal operator $[a]\phi$ to express that $\phi$ holds after doing action $a$.
Similar to above, it uses \emph{observational indistinguishability axioms} and \emph{action likelihood axioms} to model noisy actions.
As an example, the following states that the degree of belief that the robot is at a distance $h = 3$ after first sensing a distance of $4$ and then moving one unit towards the wall:
\[
  [\sonar(4)][\move(1, 1)] \bel{h(3)}{0.625}
\]
\ds has also been extended to support regression~\parencite{liuReasoningBeliefsMetabeliefs2021} and progression~\parencite{liuProgressionBelief2021}.
We will introduce \ds in detail in \autoref{sec:dsg} and use it to define abstractions of \aclp{BAT} in \autoref{chap:abstraction}.

\subsection{\golog} \label{sec:golog}

\golog~\cite{levesqueGOLOGLogicProgramming1997} is a high-level agent programming language based on the \sitcalc that allows to combine hand-crafted high-level programs with automatic reasoning approaches such as planning.
One core idea of \golog is that a developer can provide a program sketch, e.g., that describes some kind of general strategy, and the system then fills in the specifics to find a successful execution, e.g., by choosing a suitable program branch, or by means of search.
\golog combines imperative programming languages such as conditionals and loops with nondeterministic constructs as well as search methods.
The programmer has control over how much of the program they specify manually and how much is left to the system.
They may decide to take complete control over the program execution by only using deterministic instructions in the program.
In the other extreme, they may also write a program that iteratively picks some action nondeterministically until some goal has been accomplished, which corresponds to a classical planning problem.
In practice, most programs are in between the two extremes: The programmer asserts certain control over the search by providing partial programs, while the remaining choices are left to the nondeterministic execution, which picks an appropriate alternative during execution.

In contrast to other programming languages, a \golog program does not consist of low-level machine instructions.
Instead, its primitives consist of primitive actions, which are axiomatized in a \sitcalc \acl{BAT}.


\subsubsection{The \doprog Macro Operator}

The semantics of \golog, as originally proposed by \textcite{levesqueGOLOGLogicProgramming1997}, is defined by a macro operator \doprog, where $\doprog(\delta, s, s')$ intuitively means that $s'$ is a terminating situation of executing the program $\delta$ in situation $s$.
It allows the following program constructs:%
\footnote{
  Similar to above, the notation $a[s]$ means the result of restoring the situation argument $s$ to all fluents mentioned by the action term $a$.
  As an example, if $a$ is the action $\goto(\ffluent{location}(o_1))$, then $a[s]$ is $\goto(\ffluent{location}(o_1, s))$.
}
\begin{description}
  \item[Primitive actions:]
    \[
      \doprog(a, s, s') \eqdef \poss(a[s], s) \wedge s' = \doop(a[s], s)
    \]
    Executing a primitive action $a$ in situation $s$ results in $s'$ if $a$ is possible in situation $s$ and $s'$ is the successor situation of $s$ with respect to $s'$.
  \item[Test actions:]
    \[
      \doprog(\phi?, s, s') \eqdef \phi[s] \wedge s' = s
    \]
    A test $\phi?$ terminates if $\phi$ holds in the current situation $s$.
    A test does not execute any action, therefore, the terminating situation $s'$ is the same as $s$.
  \item[Sequence:]
    \[
      \doprog(\lbrack \delta_1; \delta_2 \rbrack, s, s') \eqdef \exists s^*\, \doprog(\delta_1, s, s^*) \wedge \doprog(\delta_2, s^*, s')
    \]
    Executing a sequence of sub-programs $\delta_1$ and $\delta_2$ in situation $s$ terminates in situation $s'$ if there is some situation $s^*$ such that $\delta_1$ terminates in $s^*$ and $\delta_2$ terminates in $s'$ starting from $s^*$.
  \item[Nondeterministic choice of action:]
    \[
      \doprog(( \delta_1 \vert \delta_2), s, s') \eqdef \doprog(\delta_1, s, s') \vee \doprog(\delta_2, s, s')
    \]
    The program $\delta_1 \vert \delta_2$ nondeterministically chooses between the two subprograms $\delta_1$ and $\delta_2$.
    It terminates in situation $s'$ if any of the two sub-programs terminate in $s'$.
  \item[Nondeterministic choice of arguments:]
    \[
      \doprog(\pi x.\, \delta(x), s, s') \eqdef \exists x.\, \doprog(\delta(x), s, s')
    \]
    The program $\pi x.\, \delta(x)$ nondeterministically picks some argument $x$ and then executes the sub-program $\delta$, where each occurrence of $x$ is substituted by the chosen value for $x$.
    The program terminates in situation $s'$ if there is some $x$ such that $\delta(x)$ terminates in situation $s'$.
    Nondeterministic choice of argument is typically combined with a guard $\phi(x) ?$ to ensure that a suitable argument was chosen, e.g., $\pi o.\, \objat(o) = \robotat ?; \pick(o)$ chooses some object $o$ that is at the same location as the robot and then picks up the object.
  \item[Nondeterministic iteration:]
    \begin{align*}
      \doprog(\delta^*, s, s') \eqdef
      &\forall P.\,
      \big
      \lbrack \forall s_1 P(s_1, s_1) \wedge
      \\
      & \qquad
      \forall s_1, s_2, s_3. \big(P(s_1, s_2) \wedge \doprog(\delta, s_2, s_3) \supset P(s_1, s_3)\big)
      \big\rbrack
      \\
      & \supset P(s, s')
    \end{align*}
    The nondeterministic $\delta^*$ repeats the program $\delta$ for a nondeterministic number of times (including 0).
    Therefore, $\delta^*$ ends in a situation $s'$ if $s'$ is the resulting situation of doing the program $\delta$ in some situation $s^*$ that is also a resulting situation of the iterated program.
    Formally, this corresponds to the transitive closure.
    As the transitive closure is not first-order definable, it is necessary to use second-order quantification $\forall P$ to define $\delta^*$.
    The definition says that $s'$ is the resulting situation of doing $\delta$ in $s$ for zero or more times if $(s, s')$ is in every set such that
    \begin{enumerate}
      \item $(s_1, s_1)$ is in the set for all situations $s_1$,
      \item if $(s_1, s_2)$ is in the set and doing $\delta$ in situation $s_2$ results in $s_3$, then $(s_1, s_3)$ is also in the set.
    \end{enumerate}
\end{description}

With these program constructs, conditionals and loops can defined as macros:
\begin{align*}
  \gif \phi \gthen \delta_1 \gelse \delta_2 \gfi &\eqdef \lbrack \phi?; \delta_1 \rbrack \;\vert\; \lbrack \neg \phi?; \delta_2 \rbrack
  \\
  \gwhile \phi \gdo \delta \gdone &\eqdef \lbrack \lbrack \phi?; \delta \rbrack^*; \neg \phi? \rbrack
\end{align*}

\subsubsection{\congolog}
While using \emph{start} and \emph{end} actions as described above already allows to have some form of concurrent execution of two actions, the original \golog does not allow \emph{concurrent processes}.
With that goal, \congolog~\cite{degiacomoConGologConcurrentProgramming2000} introduces a new construct $(\delta_1 \parallel \delta_2)$, where the two programs $\delta_1$ and $\delta_2$ are executed concurrently.
As before, this is a form of \emph{interleaved} concurrency, i.e., when executing $(\delta_1 \parallel \delta_2)$, either $\delta_1$ or $\delta_2$ takes a single-step transition.
In addition to concurrent execution, \congolog also adds support for prioritized concurrency, concurrent iteration, and interrupts.
Prioritized concurrency $(\delta_1 \pconc \delta_2)$ works similarly to concurrent execution, except that $\delta_2$ may only take a transition if $\delta_1$ cannot.
Concurrent iteration $\delta^{\parallel}$ iterates over the program $\delta$, but in contrast to regular iteration, the instances of $\delta$ are executed concurrently.
Thus, the program $\delta^{\parallel}$ executes like $\nil \smid \delta \smid ( \delta \parallel \delta) \smid (\delta \parallel \delta \parallel \delta) \ldots$.

While the original semantics of \golog programs is defined with the macro operator \doprog, \congolog uses a \emph{transition semantics} with an explicit representation of the program instead.
In the transition semantics, the 4-ary relational symbol $\gtrans(\delta, s, \delta', s')$ is true if the program $\delta$ can take a single-step transition from the situation $s$ into the situation $s'$, where $\delta'$ is the remaining program.
In addition to $\gtrans$, a binary relation symbol $\gfinal(\delta, s)$ says that the program $\delta$ is in a final state in situation $s$, i.e., it may terminate.
For the program constructs of \congolog, $\gtrans$ and $\gfinal$ are defined as follows:
\begin{description}
  \item[Empty program:]
    \begin{align*}
      \gtrans(\nil, s, \delta', s') &\equivspace \false
      \\
      \gfinal(\nil, s) &\equivspace \true
    \end{align*}
    There is no possible transition from the empty program \nil, the program has always terminated.
  \item[Primitive actions:]
    \begin{align*}
      \gtrans(a, s, \delta', s') &\equivspace \poss(a[s], s) \wedge \delta' = \nil \wedge s' = \doop(a[s], s)
      \\
      \gfinal(a, s) &\equivspace \false
    \end{align*}
    The program consisting of the single action $a$ can take a transition step from $s$ to $s'$ if action $a$ is possible in situation $s$.
    The remaining program is the empty program \nil and the resulting situation of doing action $a$ in situation $s$.
    A program consisting of a single action may never be final.
  \item[Test/wait actions:]
    \begin{align*}
      \gtrans(\phi?, s, \delta', s') &\equivspace \phi[s] \wedge \delta' = \nil \wedge s' = s
      \\
      \gfinal(\phi?, s) &\equivspace \false
    \end{align*}
    For a test $\phi?$, the program may transition from situation $s$ to $s'$ if the test condition $\phi$ is satisfied in situation $s$.
    The resulting situation is the same as before, i.e., no action is executed.
    The remaining program after executing a test is the empty program nil and a program consisting of a test action may never be final.
  \item[Sequence:]
    \begin{align*}
      \gtrans(\delta_1; \delta_2, s, \delta', s') &\equivspace \exists \gamma.\, \delta' = (\gamma; \delta_2) \wedge \gtrans(\delta_1, s, \gamma, s')
      \\
                                                  & \qquad \vee \gfinal(\delta_1, s) \wedge \gtrans(\delta_2, s, \delta', s')
      \\
      \gfinal(\delta_1; \delta_2, s) &\equivspace \gfinal(\delta_1) \wedge \gfinal(\delta_2)
    \end{align*}
    For a sequence of actions $\delta_1; \delta_2$, there are two possible transitions:
    \begin{enumerate}
      \item If there is some possible transition for the first sub-program $\delta_1$, then the remaining program is the remaining program after the transition of $\delta_1$ concatenated with the unchanged program $\delta_2$.
        The resulting situation is the situation of the transition for $\delta_1$.
      \item Otherwise, if $\delta_1$ is final and there is an available transition for $\delta_2$, then the remaining program and resulting situation are defined by the possible transition of $\delta_2$. 
    \end{enumerate}
    A sequence of sub-program is final if both sub-programs are final.
  \item[Nondeterministic choice of action:]
    \begin{align*}
      \gtrans(\delta_1 \vert \delta_2, s, \delta', s') &\equivspace \gtrans(\delta_1, s, \delta', s') \vee \gtrans(\delta_2, s, \delta', s')
      \\
      \gfinal(\delta_1 \vert \delta_2, s) &\equivspace \gfinal(\delta_1) \vee \gfinal(\delta_2)
    \end{align*}
    The nondeterministic choice of action $\delta_1 \vert \delta_2$ (also called nondeterministic branching) nondeterministically chooses between $\delta_1$ and $\delta_2$.
    Therefore, the resulting situation and remaining program are defined by the transition of either sub-program, i.e., the program may transition to situation $s$ with the remaining program $\delta'$ if a transition of $\delta_1$ or $\delta_2$ results in $s'$ with remaining program $\delta'$.
    The program is final if any sub-program is final.
  \item[Nondeterministic choice of argument:]
    \begin{align*}
      \gtrans(\pi x.\, \delta(x), s, \delta', s') &\equivspace \exists v.\, \gtrans(\delta^x_v, s, \delta', s')
      \\
      \gfinal(\pi x.\, \delta(x), s) &\equivspace \exists v.\, \gfinal(\delta^x_v, s)
    \end{align*}
    For the nondeterministic choice of argument $\pi x.\; \delta(x)$, the program may make a transition if there is a transition of the program $\delta$ with $x$ substituted by some value $v$.
    It is final if there exists a substitution such that $\delta$ with $x$ substituted by $v$ is final.
  \item[Nondeterministic iteration:]
    \begin{align*}
      \gtrans(\delta^*, s, \delta', s') &\equivspace \exists \gamma.\, (\delta' = \gamma; \delta^*) \wedge \gtrans(\delta, s,  \gamma, s')
      \\
      \gfinal(\delta^*, s) &\equivspace \true
    \end{align*}
    For nondeterministic iteration $\delta^*$ of a sub-program $\delta$, the program may transition to situation $s'$ if there is transition for $\delta$ that results in $s'$.
    The remaining program is the same as the remaining program of the transition of $\delta$, appended by the unmodified iteration $\delta^*$.
    Therefore, after the execution of $\delta$ has completed, the interpreter may choose to execute $\delta$ again.
    At the same time, $\delta^*$ is always final, so the interpreter may also choose to stop iterating.
  \item[Concurrent execution:]
    \begin{align*}
      &\gtrans(\delta_1 \| \delta_2, s, \delta', s')
      \\
      & \quad \equivspace
      \exists \gamma.\, \delta' = (\gamma \| \delta_2) \wedge \gtrans(\delta_1, s, \gamma, s') \vee \exists \gamma.\, \delta' = (\delta_1 \| \gamma) \wedge \gtrans(\delta_2, s, \gamma, s')
      \\
      &\gfinal(\delta_1 \| \delta_2, s) \equivspace \gfinal(\delta_1, s) \wedge \gfinal(\delta_2, s)
    \end{align*}
    For concurrent execution $\delta_1 \| \delta_2$, any transition of the two sub-programs is also a transition of the program.
    The resulting situation and the remaining program are determined by the transition of the chosen sub-program, where the remaining program is augmented with the concurrent execution of the other sub-program, which remains unchanged.
    The concurrent execution of $\delta_1$ and $\delta_2$ is final if both sub-programs are final.
  \item[Synchronized conditional:]
    \begin{align*}
      &\gtrans(\gif \phi \gthen \delta_1 \gelse \delta_2 \gfi, s, \delta', s')
      \\
      &\quad \equivspace
      \phi[s] \wedge \gtrans(\delta, s, \delta', s') \vee \neg \phi[s] \wedge \gtrans(\delta_2, s, \delta', s')
      \\
      &\gfinal(\gif \phi \gthen \delta_1 \gelse \delta_2 \gfi, s)
      \\
      &\quad \equivspace \phi[s] \wedge \gfinal(\delta_1, s) \vee \neg \phi[s] \wedge \gfinal(\delta_2, s)
    \end{align*}
    While combining tests and nondeterministic branching already allows to conditionally execute a sub-program, this is problem if combined with concurrent execution:
    If the interpreter executes the program $(\phi?; \delta_1) \| \delta_2$, it may choose to first test $\phi?$ and then continue with $\delta_2$.
    If $\phi_1$ is affected by $\delta_2$, then it may be false when $\delta_1$ is started.
    To avoid this perhaps surprising behavior, a \emph{synchronized conditional} tests the conditional $\phi$ and then directly executes the sub-program $\delta_1$ in a single transition, thereby avoiding that the interpreter may choose to switch to a different sub-program.
  \item[Synchronized loop:]
    \begin{align*}
      &\gtrans(\gwhile \phi \gdo \delta \gdone, s, \delta', s')
      \\
      &\quad \equivspace
      \exists \gamma.\, (\delta' = \gamma; \gwhile \phi \gdo \delta \gdone) \wedge \phi[s] \wedge \gtrans(\delta, s, \gamma, s')
      \\
      & \gfinal(\gwhile \phi \gdo \delta \gdone, s) \equivspace \neg \phi[s] \vee \gfinal(\delta, s)
    \end{align*}
    Similar to the synchronized conditional, the synchronized loop tests the conditional $\phi$ and then, if $\phi$ is true, directly starts executing the sub-program $\delta$ in a single transition, thereby guaranteeing that $\phi$ is actually true at the beginning of $\delta$.
  \item[Prioritized concurrency:]
    \begin{align*}
      &\gtrans(\delta_1 \pconc \delta_2, s, \delta', s')
      \\
      & \quad \equivspace
      \exists \gamma.\, \delta' = (\gamma \| \delta_2) \wedge \gtrans(\delta_1, s, \gamma, s')
        \\
      & \qquad \vee
      \exists \gamma.\, \delta' = (\delta_1 \pconc \gamma) \wedge \gtrans(\delta_2, s, \gamma, s') \wedge \neg \exists \zeta, s''.\, \gtrans(\delta_1, s, \zeta, s'')
      \\
      &\gfinal(\delta_1 \pconc \delta_2, s) \equivspace \gfinal(\delta_1, s) \wedge \gfinal(\delta_2, s)
    \end{align*}
    Prioritized concurrency $\delta_1 \pconc \delta_2$ works similarly as concurrent execution $\delta_1 \| \delta_2$, except that a transition of $\delta_2$ is only allowed if there is no possible transition of $\delta_1$, i.e., $\delta_1$ is executed with priority over $\delta_2$.
  \item[Concurrent iteration:]
    \begin{align*}
      \gtrans(\delta^{\|}, s, \delta', s') &\equivspace \exists \gamma.\, \delta' = (\gamma \| \delta^{\|}) \wedge \gtrans(\delta, s, \gamma, s')
      \\
      \gfinal(\delta^{\|}, s) &\equivspace \true
    \end{align*}
    For concurrent iteration, the program may transition to situation $s'$ if the sub-program $\delta$ may transition to $s'$.
    The remaining program is the remaining program after the transition step for $\delta$, appended by the (unmodified) concurrent iteration $\delta^{\|}$.
    The interpreter may also choose to stop executing the concurrent iteration, i.e., $\delta^{\|}$ is always final.
\end{description}


\subsubsection{Knowledge-Based \golog Programs with Sensing and Online Execution}
\golog and \congolog programs are interpreted offline, i.e., the interpreter first determines one complete sequence of actions that constitutes a legal execution of the program and only then starts executing the program.
This may be problematic, as a robot may need to first sense some fact about the world before it can determine a legal program execution~\cite{reiterKnowledgebasedProgrammingSensing2001}.
Therefore, \indigolog~\cite{degiacomoIndiGologHighlevelProgramming2009} extends \golog such that the programmer can interleave planning and online execution.
This allows to execute parts of the program, then execute a sensing action, and then decide how to continue the program based on the sensing result.
In order to do so, it extends \congolog with a search operator $\gsearch(\delta)$, which interprets the sub-program $\delta$ offline and determines a legal execution of $\delta$ before continuing.
Any instruction outside of a search operator is interpreted online, i.e., each action is immediately executed.
\indigolog also supports sensing actions, where the value of a fluent is available after executing the action, and which allows to branch on the sensing value during online execution.

To deal with knowledge more generally, \textcite{reiterKnowledgebasedProgrammingSensing2001} describes an extension of \golog to knowledge-based programs.
In a knowledge-based program, test actions $\phi?$ may not only refer to objective formulas, but may also contain explicit references to the agent's knowledge, e.g., the following program picks up the object $\mi{obj}_1$ if it is known to be in the same location as the robot:
\[
  \gif \sitcalcknow(\robotat = \objat(\mi{obj}_1))? \gthen \pick(\mi{obj}_1) \gfi
\]

\textcite{classenFoundationsKnowledgebasedPrograms2006} propose a similar kind of knowledge-based programs, but based on \es rather than the \sitcalc.
Among others, this avoids two limitations of the previous approach:
For one, it also allows to refer to meta beliefs, i.e., knowledge about knowledge.
Second, it also allows quantifying-in~\cite{kaplanQuantifying1968}, which can be used to express ``knowing what'' in contrast to ``knowing that'', e.g., the following expresses that there  is some object that is known to be in same location as the robot:
\[
  \exists o.\, \sitcalcknow(\robotat = \objat(o))
\]

In contrast, ``knowing that'' expresses that it it is known that there is some object at the same location, but not necessarily which object it is:
\[
  \sitcalcknow(\exists o.\, \robotat = \objat(o))
\]

\section{Temporal Logics and Timed Systems}\label{sec:timed-systems}

While the \sitcalc and its variants allow us to model a high-level agent program by specifying the agent's actions with preconditions and effects, it does not provide us with a formalism to naturally specify desired properties of the progression of the program.
As an example, we may require that whenever the robot is carrying a heavy object, it should do so only for a limited time, say two minutes, to protect itself from overheating.
Afterwards, it should not use the arm for thirty seconds so it can completely cool down.
In order to do define such requirements, we will utilize \emph{temporal logics} and we will model the robot's components such as the arm with \emph{timed automata}.

Temporal logics are formal frameworks that allow to describe the progression of a system over time.
They are widely used for model-based verification of programs and reactive systems~\parencite{baierPrinciplesModelChecking2008}, where correctness specifications typically not only specify the desired state at the end of execution, but also pose requirements on the intermediate states.
They allow to model such specifications with modal operators that explicitly refer to different states of execution, e.g., $\tnext{} \phi$ states that in the next state of execution, the requirement $\phi$ must hold.
In addition to a specification language, model-based verification also requires a formalism to describe the underlying system.
One commonly used formalism is a \emph{timed automaton}, which can roughly be seen as a finite-state automaton extended with metric time by means of clocks and timing constraints.

In the following, we first give an overview on the different temporal logics and their properties, before we introduce \acf{MTL} in full detail, as we will later use it to specify constraints on the robot program.
Afterwards, we will introduce \emph{timed automata}, which we will use to model the robot's hardware and low-level software components, e.g., its gripper or its navigation unit.

\subsection{Temporal Logics}\label{sec:temporal-logics}

\begin{figure}[ht]
  \centering
  \tikzset{
  tp/.style={draw,circle},
  every label/.append style={text depth=2pt},
  >=latex,
}

\begin{tikzpicture}[->,align=center,node distance=0.6cm and 1.5cm]
  \node (atomic) {atomic proposition \\ $a$};
  \node[below=of atomic] (next) {next \\ $\tnext{} a$};
  \node[below=of next] (until) {until \\ $a \until{} b$};
  \node[below=of until] (duntil) {dual until \\ $a \duntil{} b$};
  \node[below=of duntil] (finally) {finally \\ $\finally{} a$};
  \node[below=of finally] (globally) {globally \\ $\glob{} a$};

  \node[tp,right=of atomic,label={$a$}] (atomic1) {};
  \node[tp,right=of atomic1,label={any}] (atomic2) {};
  \node[tp,right=of atomic2,label={any}] (atomic3) {};
  \node[tp,right=of atomic3,label={any}] (atomic4) {};
  \node[tp,right=of atomic4,label={any}] (atomic5) {};
  \node[tp,draw=none,right=of atomic5] (atomicdots) {$\ldots$};

  \node[tp,label={any}] (next1) at (atomic1 |- next) {};
  \node[tp,right=of next1,label={$a$}] (next2) {};
  \node[tp,right=of next2,label={any}] (next3) {};
  \node[tp,right=of next3,label={any}] (next4) {};
  \node[tp,right=of next4,label={any}] (next5) {};
  \node[tp,draw=none,right=of next5] (nextdots) {$\ldots$};

  \node[tp,label={any}] (until1) at (atomic1 |- until) {};
  \node[tp,right=of until1,label={$a$}] (until2) {};
  \node[tp,right=of until2,label={$a$}] (until3) {};
  \node[tp,right=of until3,label={$b$}] (until4) {};
  \node[tp,right=of until4,label={any}] (until5) {};
  \node[tp,draw=none,right=of until5] (untildots) {$\ldots$};

  \node[tp,label={any}] (duntil1) at (atomic1 |- duntil) {};
  \node[tp,right=of duntil1,label={$b$}] (duntil2) {};
  \node[tp,right=of duntil2,label={$b$}] (duntil3) {};
  \node[tp,right=of duntil3,label={$a,b$}] (duntil4) {};
  \node[tp,right=of duntil4,label={any}] (duntil5) {};
  \node[tp,draw=none,right=of duntil5] (duntildots) {$\ldots$};

  \node[tp,label={any}] (finally1) at (atomic1 |- finally) {};
  \node[tp,right=of finally1,label={any}] (finally2) {};
  \node[tp,right=of finally2,label={$a$}] (finally3) {};
  \node[tp,right=of finally3,label={any}] (finally4) {};
  \node[tp,right=of finally4,label={any}] (finally5) {};
  \node[tp,draw=none,right=of finally5] (finallydots) {$\ldots$};

  \node[tp,label={any}] (globally1) at (atomic1 |- globally) {};
  \node[tp,right=of globally1,label={$a$}] (globally2) {};
  \node[tp,right=of globally2,label={$a$}] (globally3) {};
  \node[tp,right=of globally3,label={$a$}] (globally4) {};
  \node[tp,right=of globally4,label={$a$}] (globally5) {};
  \node[tp,draw=none,right=of globally5] (globallydots) {$\ldots$};

  \draw (atomic1) -> (atomic2);
  \draw (atomic2) -> (atomic3);
  \draw (atomic3) -> (atomic4);
  \draw (atomic4) -> (atomic5);
  \draw (atomic5) -> (atomicdots);

  \draw (next1) -> (next2);
  \draw (next2) -> (next3);
  \draw (next3) -> (next4);
  \draw (next4) -> (next5);
  \draw (next5) -> (nextdots);

  \draw (until1) -> (until2);
  \draw (until2) -> (until3);
  \draw (until3) -> (until4);
  \draw (until4) -> (until5);
  \draw (until5) -> (untildots);

  \draw (duntil1) -> (duntil2);
  \draw (duntil2) -> (duntil3);
  \draw (duntil3) -> (duntil4);
  \draw (duntil4) -> (duntil5);
  \draw (duntil5) -> (duntildots);

  \draw (finally1) -> (finally2);
  \draw (finally2) -> (finally3);
  \draw (finally3) -> (finally4);
  \draw (finally4) -> (finally5);
  \draw (finally5) -> (finallydots);

  \draw (globally1) -> (globally2);
  \draw (globally2) -> (globally3);
  \draw (globally3) -> (globally4);
  \draw (globally4) -> (globally5);
  \draw (globally5) -> (globallydots);
\end{tikzpicture}
  \caption[\acs*{LTL} Operators.]{\acs*{LTL} operators (adapted from \parencite{baierPrinciplesModelChecking2008}.)
  }
  \label{fig:ltl-operators}
\end{figure}

A variety of temporal formalisms exist (see \parencite{longReviewTemporalLogics1989,alurLogicsModelsReal1992,konurSurveyTemporalLogics2013} for surveys), which can be classified by the following properties (adapted from \parencite{emersonTemporalModalLogic1990}):
\begin{description}
  \item[Discrete versus continuous time:]
    The system may refer to time points either from a discrete domain (e.g., the natural numbers) or a continuous domain (the real numbers).
    When using a discrete notion of time, e.g., in \ac{LTL}~\cite{pnueliTemporalLogicPrograms1977}, the focus is on the order of events and the notion of time is implicit.
    \ac{LTL} temporal operators, as shown in \autoref{fig:ltl-operators},%
      \footnote{
        Usually, temporal operators in \ac{LTL} use a non-strict semantics, where the temporal operators also refer to the current state.
        As an example, $\glob{} a$ usually requires that $a$ also holds in the current state.
        Meanwhile, in \ac{MTL}, a strict semantics is often used, where the current state is excluded from the temporal operators, e.g., $\glob{} a$ does not require that $a$ currently holds, but only in every future state.
        For consistency's sake, we adopt the strict semantics known from \ac{MTL} even for \ac{LTL}.%
      }
    do not have an explicit time parameter, but implicitly refer to the next time point, e.g., $\tnext{} \phi$.
    In contrast, in logics such as \ac{MTL} with real-valued time, temporal operators typically have an interval as parameter, e.g., $\finally{\lbrack 1, 2 \rbrack} \phi$ states that $\phi$ must hold at some point in the time interval $\lbrack 1, 2\rbrack$ from now.
    An alternative is to use temporal formulas with timing constraints as in \ac{MTL} but a discrete model of time based on \emph{digital clocks}~\cite{henzingerWhatDecidableHybrid1998}, where at every state, only a discrete approximation of the real time is recorded.
    This restriction allows to express some interesting but not all timing constraints~\cite{henzingerWhatGoodAre1992}.
  \item[Time points versus time intervals:] 
    When specifying temporal properties, we may either refer to \emph{time points} (e.g., ``two time units from now'') or \emph{intervals} (e.g., ``between event $a$ and event $b$'').
    Depending on the choice of time representation, different modal operators are used.
    In interval-based formalisms such as Allen's interval algebra~\parencite{allenMaintainingKnowledgeTemporal1983}, the operators describe the relation between intervals, e.g., $I_1\;\textbf{meets}\;I_2$ to state that $I_2$ must start at the exact time when $I_1$ ends, or $I_1\;\mathbf{during}\;I_2$ to state that $I_1$ must start after and end before $I_2$.
    \textcite{allenMaintainingKnowledgeTemporal1983} identified 13 different relations that two intervals may have.
    In contrast, temporal logics such as \ac{LTL} or \ac{MTL} refer to time points, i.e., the state of the system at a certain point in time.
    Point-based frameworks are widely used for verification and synthesis and more recently have also been used for conditional planning~\cite{degiacomoSynthesisLTLLDL2015,degiacomoLTLfLDLfSynthesis2016} and for planning with temporally extended goals~\cite{patriziComputingInfinitePlans2011}, while Allen's interval algebra has mainly been applied to temporal planning~\cite{allenGeneralTheoryAction1984,rosuAllenLinearInterval2006}.
  \item[Branching versus linear time:]
    Concerning the progression of a program or a system, two principal views are possible: In \emph{linear} systems, at any point in time, there is only one possible successor of the current state and formulas make assertions about paths.
    The other view is that time is branching: At any point in time, all possible evolutions of the system are considered, resulting in a tree-like structure, where formulas make assertions about states.
    Logics such as \ac{LTL} adopt the former view, hence the name \acl{LTL}, while branching-time logics such as \ac{CTL}~\cite{clarkeDesignSynthesisSynchronization1982,emersonDecisionProceduresExpressiveness1985} adopt the latter view.
    For both \ac{LTL} and \ac{CTL}, there are properties that are expressible in one logic but not the other~\cite{lamportSometimeSometimesNot1980,baierPrinciplesModelChecking2008}.
    However, CTL\textsuperscript{*}~\cite{emersonSometimesNotNever1986} unifies both view points and allows to express all properties that can be expressed in \ac{LTL} and \ac{CTL}.
  \item[Propositional versus first-order:] In propositional formalisms, the non-temporal part of the logic is classical propositional logic, which is the case for most common temporal logics such as \ac{LTL}, \ac{MTL}, or \ac{CTL}.
    However, it is also possible to use first-order logic with functions, predicates, quantifiers, etc., as the underlying logic~\cite{hodkinsonDecidableFragmentsFirstorder2000,calvaneseFirstorderMcalculusGeneric2018,calvaneseVerificationMonitoringFirstorder2022}.
    This allows to define properties such as $\glob{} \forall r.\, \mi{Request}(r) \supset \tnext{} \exists a\, \mi{Serves}(a, r)$, which states that every request $r$ needs to be served by some agent $a$ in the next step.
  \item[Past versus future:]
    In most frameworks, temporal operators are restricted to referring events in the future.
    However, for some properties, it is more natural to express them with additional temporal operators referring to the past.
    Therefore, PLTL~\cite{lichtensteinGlory1985} extends \ac{LTL} with past operators $\mathbf{P}$ (previous) and \since{} (since), which are the duals to $\tnext{}$ (next) and $\until{}$ (until).
    This allows formulas such as $\pick \supset \neg \action{error} \since{} \action{reset}$, stating that if a \pick action occurs, then there must have been no \action{error} since the last \action{reset}.
    Past operators do not add expressiveness to \ac{LTL}~\cite{lichtensteinGlory1985} but can be exponentially more succinct~\cite{markeyTemporalLogicExponentially2003}, i.e., some properties require exponentially larger formulas if restricted to \ac{LTL} without past operators.
    Interestingly, this does not hold for \ac{MTL}, as \ac{MTL} with past operators is strictly more expressive than \ac{MTL} restricted to future operators for infinite words~\cite{bouyerExpressivenessTPTLMTL2005} and finite words~\cite{prabhakarExpressivenessMTLOperators2006}.
  \item[Finite versus infinite traces:]
    When defining properties on a program or a reactive system, we may consider finite executions of the system, i.e., the program eventually terminates, or we may deal with a non-terminating program, where the resulting traces are infinite.
    While verification approaches mostly focus on infinite traces, finite traces are particularly interesting in the context of synthesis.
    In the case of \ac{LTL}, while the synthesis problem is 2EXPTIME-complete in both cases, effective approaches focus on finite traces, as they avoid the need for automata determinization~\cite{degiacomoSynthesisLTLLDL2015}.
    In the case of \ac{MTL}, synthesis on infinite traces is undecidable, while it is decidable on finite traces with some restrictions~\cite{bouyerControllerSynthesisMTL2006}.
\end{description}

In this thesis, we want to use temporal logic to define low-level platform constraints on the high-level program and then synthesize a controller that ensures that the constraints are satisfied.
\ac{MTL} over finite words is a suitable logic for this purpose, for the following reasons:
\begin{itemize}
  \item It allows referring to \emph{continuous time}.
    This is important because many components of a real-world robot require an explicit notion of time, e.g., to state that a camera needs to run for a certain amount of time before the object detection generates reliable results.
  \item It represents time with \emph{time points}.
    While the time domain of \ac{MTL} is continuous, each event occurs at a certain time point.
    This is a natural choice as we can associate each action (and situation) with the time point when it is executed.
  \item It uses \emph{linear time}.
    The main goal is to specify constraints and synthesize a controller that guarantees certain properties for every execution of the program.
    While \citeauthor{vardiBranchingVsLinear2001} argues that ``for the synthesis of reactive systems, one has to consider a branching-time framework, since all possible strategies by the environment need to be considered''~\parencite[][p.~17]{vardiBranchingVsLinear2001}, this makes the assumption that the environment is modeled as part of the specification.
    That is to say, the specification has the form $\mi{Env} \supset \mi{Ctrl}$, where every trace that satisfies the environment specification $\mi{Env}$ must also satisfy the controller specification $\mi{Ctrl}$~\parencite{chatterjeeEnvironmentAssumptionsSynthesis2008}, which lends itself to use the AE-paradigm~\cite{pnueliTemporalLogicPrograms1977} for synthesis.
    However, in our case, the environment is not modeled as part of the specification, but instead with timed automata as well as the abstract input program.
    For this reason, we do not require a formalism that allows expressions about the existence and universality of program executions, i.e., quantification over program branches.
    Additionally, for both discrete time and continuous time formalisms, recent research has focussed on linear-time logics~\cite{bloemCTLSynthesisLTL2017}.
    As we build on top of existing work, in particular \parencite{bouyerControllerSynthesisMTL2006}, a linear-time formalism is the better choice for the purpose of this thesis.
  \item It is restricted to \emph{propositional} specifications.
    While first-order extensions would be interesting, in particular as they allow infinite domains, full first-order temporal logic is undecidable, even when restricted to discrete time~\cite{hodkinsonDecidableFragmentsFirstorder2000}.
    While there are decidable fragments of first-order \ac{LTL}, previous results of \ac{MTL} verification and synthesis have largely focussed on the propositional case.
    For this reason, we will also base our approach on propositional specifications and therefore restrict ourselves to finite domains.
  \item For a similar reason, we will use \ac{MTL} without past operators.
    While having past operators would be helpful to express certain platform constraints, previous results on synthesis have been restricted to \ac{MTL} without past operators and it is not immediately clear how to extend the approach to past operators.
  \item While a logic based on infinite traces would be interesting as it allows expressing properties about non-terminating \golog programs~\parencite{classenLogicNonterminatingGolog2008},  we restrict the formalism to \ac{MTL} on finite traces and thus on terminating programs because \ac{MTL} synthesis on infinite words is undecidable~\cite{bouyerControllerSynthesisMTL2006}.
\end{itemize}


\subsection{\acl*{MTL}}\label{sec:mtl}


\ac{MTL} \parencite{koymansSpecifyingRealtimeProperties1990} is a temporal logic with continuous time and timing constraints on the \emph{Until} modality, therefore allowing temporal constraints with interval restrictions, e.g., $\fut{\leq 2} b$ to say that within the next two time steps, a $b$ event must occur.
Two different semantics have been proposed for \ac{MTL}:
In the \emph{interval-based} semantics~\parencite{koymansSpecifyingRealtimeProperties1990}, each state of the system is associated with a time interval which indicates the period of time when the system is in that state~\parencite{alurBenefitsRelaxingPunctuality1996}.
In this semantics, the system is observed in every instance of time.
Unfortunately, in the interval-based semantics, the satisfiability problem is undecidable~\cite{alurReallyTemporalLogic1994}.
One commonly used alternative is a \emph{point-based semantics}~\cite{ouaknineRecentResultsMetric2008}, sometimes also called \emph{trace semantics}~\cite{alurLogicsModelsReal1992}, in which formulas are interpreted over timed words.
In the point-based semantics, the satisfiability problem is decidable~\cite{ouaknineDecidabilityMetricTemporal2005} and can be checked with \acfp{ATA}, which we will introduce in \autoref{sec:ata}.

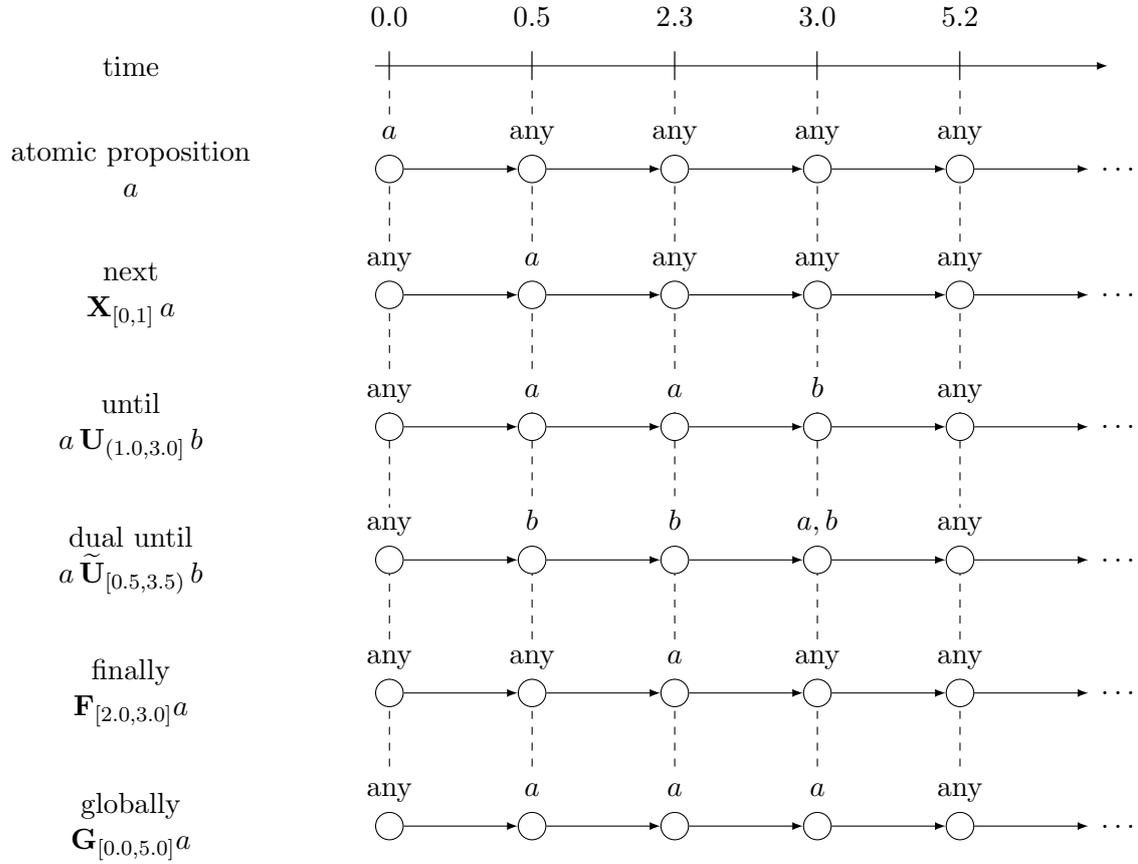
\begin{figure}[ht]
  \centering
  \tikzset{
  every node/.append style={fill=white},
  tp/.style={draw,circle},
  every label/.append style={text depth=2pt,fill=white},
  >=latex,
}

\begin{tikzpicture}[->,align=center,node distance=0.6cm and 1.5cm]
  \node (atomic) {atomic proposition \\ $a$};
  \node[below=of atomic] (next) {next \\ $\tnext{[0,1]} a$};
  \node[below=of next] (until) {until \\ $a \until{(1.0,3.0]} b$};
  \node[below=of until] (duntil) {dual until \\ $a \duntil{[0.5,3.5)} b$};
  \node[below=of duntil] (finally) {finally \\ $\finally{[2.0,3.0]} a$};
  \node[below=of finally] (globally) {globally \\ $\glob{[0.0,5.0]} a$};

  \node[tp,right=of atomic,label={$a$}] (atomic1) {};
  \node[tp,right=of atomic1,label={any}] (atomic2) {};
  \node[tp,right=of atomic2,label={any}] (atomic3) {};
  \node[tp,right=of atomic3,label={any}] (atomic4) {};
  \node[tp,right=of atomic4,label={any}] (atomic5) {};
  \node[tp,draw=none,right=of atomic5] (atomicdots) {$\ldots$};

  \node[tp,label={any}] (next1) at (atomic1 |- next) {};
  \node[tp,right=of next1,label={$a$}] (next2) {};
  \node[tp,right=of next2,label={any}] (next3) {};
  \node[tp,right=of next3,label={any}] (next4) {};
  \node[tp,right=of next4,label={any}] (next5) {};
  \node[tp,draw=none,right=of next5] (nextdots) {$\ldots$};

  \node[tp,label={any}] (until1) at (atomic1 |- until) {};
  \node[tp,right=of until1,label={$a$}] (until2) {};
  \node[tp,right=of until2,label={$a$}] (until3) {};
  \node[tp,right=of until3,label={$b$}] (until4) {};
  \node[tp,right=of until4,label={any}] (until5) {};
  \node[tp,draw=none,right=of until5] (untildots) {$\ldots$};

  \node[tp,label={any}] (duntil1) at (atomic1 |- duntil) {};
  \node[tp,right=of duntil1,label={$b$}] (duntil2) {};
  \node[tp,right=of duntil2,label={$b$}] (duntil3) {};
  \node[tp,right=of duntil3,label={$a,b$}] (duntil4) {};
  \node[tp,right=of duntil4,label={any}] (duntil5) {};
  \node[tp,draw=none,right=of duntil5] (duntildots) {$\ldots$};

  \node[tp,label={any}] (finally1) at (atomic1 |- finally) {};
  \node[tp,right=of finally1,label={any}] (finally2) {};
  \node[tp,right=of finally2,label={$a$}] (finally3) {};
  \node[tp,right=of finally3,label={any}] (finally4) {};
  \node[tp,right=of finally4,label={any}] (finally5) {};
  \node[tp,draw=none,right=of finally5] (finallydots) {$\ldots$};

  \node[tp,label={any}] (globally1) at (atomic1 |- globally) {};
  \node[tp,right=of globally1,label={$a$}] (globally2) {};
  \node[tp,right=of globally2,label={$a$}] (globally3) {};
  \node[tp,right=of globally3,label={$a$}] (globally4) {};
  \node[tp,right=of globally4,label={any}] (globally5) {};
  \node[tp,draw=none,right=of globally5] (globallydots) {$\ldots$};

  \node[above=of atomic] (time) {time};
  \node[label={0.0}] (t1) at (time -| atomic1) {$\vert$};
  \node[label={0.5}] (t2) at (time -| atomic2) {$\vert$};
  \node[label={2.3}] (t3) at (time -| atomic3) {$\vert$};
  \node[label={3.0}] (t4) at (time -| atomic4) {$\vert$};
  \node[label={5.2}] (t5) at (time -| atomic5) {$\vert$};
  \node (tdots) at (time -| atomicdots) {};

  \draw[->] (t1.west) -> (tdots);
  \begin{pgfonlayer}{background}
    \draw[-,dashed] (t1) -- (globally1);
    \draw[-,dashed] (t2) -- (globally2);
    \draw[-,dashed] (t3) -- (globally3);
    \draw[-,dashed] (t4) -- (globally4);
    \draw[-,dashed] (t5) -- (globally5);
  \end{pgfonlayer}

  \draw (atomic1) -> (atomic2);
  \draw (atomic2) -> (atomic3);
  \draw (atomic3) -> (atomic4);
  \draw (atomic4) -> (atomic5);
  \draw (atomic5) -> (atomicdots);

  \draw (next1) -> (next2);
  \draw (next2) -> (next3);
  \draw (next3) -> (next4);
  \draw (next4) -> (next5);
  \draw (next5) -> (nextdots);

  \draw (until1) -> (until2);
  \draw (until2) -> (until3);
  \draw (until3) -> (until4);
  \draw (until4) -> (until5);
  \draw (until5) -> (untildots);

  \draw (duntil1) -> (duntil2);
  \draw (duntil2) -> (duntil3);
  \draw (duntil3) -> (duntil4);
  \draw (duntil4) -> (duntil5);
  \draw (duntil5) -> (duntildots);

  \draw (finally1) -> (finally2);
  \draw (finally2) -> (finally3);
  \draw (finally3) -> (finally4);
  \draw (finally4) -> (finally5);
  \draw (finally5) -> (finallydots);

  \draw (globally1) -> (globally2);
  \draw (globally2) -> (globally3);
  \draw (globally3) -> (globally4);
  \draw (globally4) -> (globally5);
  \draw (globally5) -> (globallydots);
\end{tikzpicture}
  \caption[\acs*{MTL} Operators.]{\acs*{MTL} operators with point-based semantics.
    In comparison to \autoref{fig:ltl-operators}, time is now continuous, but the system is only observed as a sequence of countably many states.
    The temporal operators are now constrained by a time interval, which defines the interval to be considered for the evaluation of the operator.
  }
  \label{fig:mtl-operators}
\end{figure}

%

\mtl formulas are constructed from atomic propositions with the usual boolean operators and the temporal operator $\until{I}$:
\begin{definition}[Formulas of \acs*{MTL}]
  Given a finite set $P$ of atomic propositions, the formulas of \mtl{} are built as follows:
  \[
    \phi ::= p \smid \neg \phi \smid \phi \wedge \phi \smid \phi \until{I} \phi
  \]
  Here, $p \in P$ is an atomic proposition and $I \subseteq \realpos$ is an open, closed, or half-open interval with endpoints in $\naturals \cup \{ \infty \}$.
\end{definition}

As an example, the formula $\mi{cam\_on} \until{[1, 2]} \mi{grasping}(o)$ says that the object $o$ must be grasped in the interval $[1, 2]$ and until then, the camera must be on.

We extend the logic with additional operators by defining them as abbreviations:\footnote{We deviate from the usual notation ($\circ$ for \emph{next}, $\square$ for \emph{globally}, and $\diamond$ for \emph{finally}) to avoid confusion with $\es$ formulas, which use the same symbols.}
\begin{itemize}
  \item $\phi_1 \vee \phi_2 \eqdef \neg (\neg \phi_1 \wedge \neg \phi_2)$ (\emph{disjunction})
  \item $\tnext{I} \phi \eqdef (\bot \until{I} \phi)$ (\emph{next})
  \item $\fut{I} \phi \eqdef (\top \until{I} \phi)$ (\emph{finally})
  \item $\glob{I}\phi \eqdef \neg \fut{I} \neg \phi$ (\emph{globally})
  \item $\phi_1 \duntil{I} \phi_2 \eqdef \neg (( \neg \phi_1) \until{I} (\neg \phi_2))$ (\emph{dual until})
\end{itemize}
We also use the operators $<, \leq, =, \geq, >$ to denote intervals, e.g., $\geq 5$ for the interval $[5, \infty)$.
We may omit the interval $I$ if $I = [0, \infty)$, e.g., $\phi \until{} \psi$ is short for $\phi \until{[0,\infty)} \psi$.
\autoref{fig:mtl-operators} shows an overview of the \ac{MTL} temporal operators.
Using the disjunction and the dual-until operators, it is possible to rewrite every \ac{MTL} formula into an equivalent formula in \emph{positive normal form}, where negation is only applied to the atomic propositions $P$.

\ac{MTL} formulas are interpreted over timed words, which consist of a sequence states described by atomic propositions along with a time stamp:
\begin{definition}[Timed Words]\label{def:timed-words}
	A timed word $\twsym$ over a finite set of atomic propositions $P$ is a finite sequence
  \[
    \twsym = \left(\twsym_0,\twtime_0\right)\left(\twsym_1,\twtime_1\right)\ldots\left(\twsym_n,\twtime_n\right)
  \]
  where $\twsym_i \subseteq P$ and $\twtime_i \in \realpos$ such that $\twtime_0 = 0$ and the sequence $\left(\tau_i\right)_i$ is monotonically non-decreasing.
  We also write $\lvert \twsym \rvert$ for the length of $\twsym$.
  The set of timed words over $P$ is denoted as $\mi{TP}^*$.
\end{definition}

In contrast to the usual definition, we expect each symbol $\twsym_i$ to be a subset (rather than a single element) of the alphabet $P$.
We do this because we want to use a state-based setting, where each symbol $\twsym_i$ describes the state of the system with a set of propositions that are true in the state, analogous to how each situation in the situation calculus can be described by the set of fluents satisfied in the situation.


We can now formally define when a timed word $\rho$ satisfies an \ac{MTL} formula $\phi$:
\begin{definition}[Point-based Semantics of \acs*{MTL}]\label{def:mtl-semantics}
  Given a timed word $\rho = \left(\twsym_0, \twtime_0\right) \ldots \left(\twsym_n, \twtime_n\right)$ over
  alphabet $P$ and an \mtl{} formula $\phi$, $\rho, i \models \phi$ is defined
  as follows:
  \begin{enumerate}
    \item $\rho, i \models p$ iff $p \in \rho_i$,
    \item $\rho, i \models \neg \phi$ iff $\rho, i \not\models \phi$,
    \item $\rho, i \models \phi_1 \wedge \phi_2$ iff $\rho_i \models \phi_1$ and $\rho_i \models \phi_2$, and
    \item \label{def:mtl-semantics:until}
      $\rho, i \models \phi_1 \until{I} \phi_2$ iff there exists $j$ such that
      \begin{enumerate}
        \item $i < j < \abs{\rho}$,
        \item $\rho, j \models \phi_2$,
        \item $\tau_j - \tau_i \in I$,
        \item and $\rho, k \models \phi_1$ for all $k$ with $i < k < j$.
      \end{enumerate}
  \end{enumerate}
  For \iac{MTL} formula $\phi$, we also write $\rho \models \phi$ for $\rho, 0 \models \phi$ and we define the language of $\phi$ as $\mathcal{L}(\phi) = \{ \rho \mid \rho \models \phi \}$.
\end{definition}

The formula $\phi_1 \until{I} \phi_2$ states that the formula $\phi_2$ is satisfied at a point in the future within the interval $I$ and at every point before that, the formula $\phi_1$ is satisfied.
Note that we use strict-until, i.e., in \autoref{def:mtl-semantics:until}, we require that $i < j$ rather than $i \leq j$.
However, weak-until can be expressed with strict-until ($ \phi \textbf{U}_I^{\text{weak}} \psi \eqdef \psi \vee \phi \until{I} \psi)$, while strict-until cannot be expressed with weak-until \cite{henzingerItTimeRealtime1998}.
Using strict-until also allows us to define \emph{finally} as $\fut{I} \phi \eqdef (\top \until{I} \phi)$, because strict-until does not refer to the current state but to states strictly in the future.

In \autoref{sec:ata}, we will explain how the satisfiability of an \ac{MTL} formula can be checked with \acp{ATA}.
However, before we can describe \acp{ATA}, we first need to introduce \acp{LTS}, clocks, and timed automata.

\subsection{Labeled Transition Systems}\label{sec:lts}

We start by introducing \acfip{LTS}, which lay the foundation for both \aclp{TA} and \aclp{ATA}.
We mostly follow the notation used by \textcite{alurTimedAutomata1999}.

An \ac{LTS} models a discrete system by a state-transition graph whose transitions are labeled with symbols:
\begin{definition}[\Acl{LTS}]
  A \acfi{LTS} $\lts$ is a tuple $ \left( \ltsstates, \ltsstate_0, \ltsfinalstates, \ltsalph, \ltstrans \right) $, where
  \begin{itemize}
    \item $\ltsstates$ is a set of states,
    \item $\ltsstate_0 \subseteq \ltsstates$ is the initial state,
    \item $\ltsfinalstates \subseteq \ltsstates$ is a set of final states,
    \item $\ltsalph$ is a set of labels (also called events),
    \item $\operatorname{\ltstrans} \subseteq \ltsstates \times \ltsalph \times Q$ is a set of transitions.
      \qedhere
  \end{itemize}
\end{definition}

We may omit \ltsfinalstates from the tuple if $\ltsfinalstates = \ltsstates$.
We also write $\ltsstate \ltstrans[a] \ltsstate'$ for the transition $\left(\ltsstate, a, \ltsstate'\right) \in \operatorname{\ltstrans}$.
The system starts in the initial state $\ltsstate_0$ and it can switch from state $\ltsstate$ to state $\ltsstate'$ if $a$ is read. 
We also write $\ltsstate \ltstrans \ltsstate'$ if there is an $a \in \ltsalph$ such that $\ltsstate \ltstrans[a] \ltsstate'$.
A \emph{path} from $\ltsstate$ to $\ltsstate'$ is a sequence of transitions $\ltsstate \ltstrans \ltsstate_1 \ltstrans \ldots \ltstrans \ltsstate'$.
A path is \emph{infinite} if it consists of infinitely many transitions, and \emph{finite} otherwise.
A \emph{run} on \ta over a word $\sigma = \sigma_0 \sigma_1 \ldots$ is a path $\ltsstate_0 \ltstrans[\sigma_0] \ltsstate_1 \ltstrans[\sigma_1] \ldots$ starting in the initial state $\ltsstate_0$.
We denote the set of all finite runs of \ta with $\finruns(\ta)$, the set of all infinite runs of \ta with $\infruns(\ta)$ and the set of all runs of \ta with $\runs(\ta) \eqdef \finruns(\ta) \cup \infruns(\ta)$.
A finite run is \emph{accepting} if it ends in an accepting state $\ltsstate \in \ltsfinalstates$.
An infinite run is accepting if it visits at least one accepting state $\ltsstate \in \ltsfinalstates$ infinitely often (\emph{Büchi condition}).
We denote the set of accepting runs of a \ac{TA} \ta with $\accruns(\ta)$, the set of accepting finite runs with $\accfinruns(\ta)$, and the set of accepting infinite runs with $\accinfruns(\ta)$.
We write $\ltsstate \ltstrans^* \ltsstate'$ if there is a finite path from $\ltsstate$ to $\ltsstate'$.
We say that $\ltsstate'$ is reachable from $\ltsstate$ if $\ltsstate \ltstrans^* \ltsstate'$ and we call $\ltsstate'$ reachable if it is reachable from some initial state.

\begin{example}[\ac{LTS}]
  Consider the following \ac{LTS} \lts:
  \begin{center}
    \includestandalone{figures/lts}
  \end{center}


  It allows the following accepting runs:
  \begin{alignat*}{3}
    a b a b a &\in \accfinruns(\lts)
    &\qquad
    b c b a a a b a &\in \accfinruns(\lts)
    &\qquad
    b c b b c &\in \accfinruns(\lts)
    \\
    a^\omega &\in \accinfruns(\lts)
    &\qquad
    a b a^\omega &\in \accinfruns(\lts)
    &\qquad
    (b c a b)^\omega &\in \accinfruns(\lts)
  \end{alignat*}
\end{example}

We will later use \acp{LTS} to define the semantics of \acp{TA} and \acp{ATA} by associating a \ac{TA} (\ac{ATA}) with an \ac{LTS} that describes the transitions of the automaton.

\subsection{Clocks}\label{sec:clocks}
Timing constraints in timed transition systems are expressed with the help of a real-valued variable called \emph{clock}.
Clocks are used both for \acp{TA} and \acp{ATA} and we will use a similar notion in \autoref{chap:timed-esg} for the logic \tesg.
Any of these systems has a finite set of clocks, whose values increase with the same rate, and which may be reset to zero when following a transition.
Other operators, such as setting the clock to an arbitrary value or setting a clock to another clock's value is not possible.
Furthermore, clocks can be used as constraints for transitions, where the transition is only possible if the clock constraint is satisfied.
Clock constraints are defined by the following grammar:
\begin{definition}[Clock constraint]
  Let $\taclocks$ be a set of clocks.
  The set $\clockconstraints(\taclocks)$ of \emph{clock constraints} $\clockconstraint$ is defined by the grammar:
  \[
    \clockconstraint ::= x < c \smid x \leq c \smid x = c \smid x \geq c \smid x > c \smid \clockconstraint \wedge \clockconstraint
  \]
  where $x \in \taclocks$ is a clock and $c \in \mathbb{Q}$ is a constant.
\end{definition}

\begin{example}
  The clock constraint $x_1 < 3 \wedge x_2 \geq 5$ expresses that the value of the clock $x_1$ must be strictly smaller than $3$ and the value of the clock $x_2$ must be at least $5$.
\end{example}
To evaluate clock constraints, we use \emph{clock valuations}.
A clock valuation assigns a real value to each clock:
\begin{restatable}[Clock valuation]{definition}{defclockvaluation}
\label{def:clock-valuation}
  A \emph{clock valuation} \clockvaluation for a set of clocks \taclocks is a mapping $\clockvaluation: \taclocks \rightarrow \realpos$.
  For some $\delta \in \realpos$, $\clockvaluation + \delta$ denotes the clock valuation which maps every clock $x$ to the value $\clockvaluation(x) + \delta$.
  For $Y \subseteq \taclocks$, $\clockvaluation \lbrack Y \eqdef 0 \rbrack$ denotes the clock valuation for $\taclocks$ which assigns $0$ to each $x \in Y$ and agrees with $\clockvaluation$ over the rest of the clocks, i.e.,
  \[
    \clockvaluation \lbrack Y \eqdef 0 \rbrack (x) =
    \begin{cases}
      0 & \text{ if } x \in Y
      \\
      \clockvaluation(x) & \text{ else }
    \end{cases}
  \]
  For a set of clocks $\taclocks$, we also write $\vec{0}$ for the clock valuation that sets every clock value to $0$, i.e., $\vec{0}(x) = 0$ for every $x \in \taclocks$.
  We will also sometimes denote a clock valuation as a set $\syncclocks$ of pairs, where $(c, r) \in \syncclocks$ if $\clockvaluation(c) = r$.
  If the set of clocks is clear from context, we may also denote a clock valuation of $n$ clocks as a vector $v \in \realpos^n$, e.g., for $\taclocks = \{ c_1, c_2 \}$, the vector $(0.1, 0.2)$ denotes the clock valuation \clockvaluation with $\clockvaluation(c_1) = 0.1$ and $\clockvaluation(c_2) = 0.2$.
\end{restatable}
We can now define when a clock valuation satisfies some clock constraint:
\begin{definition}[Clock constraint satisfaction]
  Given a clock valuation \clockvaluation for \taclocks and a clock constraint $\clockconstraint \in \clockconstraints(\taclocks)$, the satisfaction of the clock constraint $\clockconstraint$ by the clock valuation \clockvaluation, denoted by $\clockvaluation \models \clockconstraint$, is defined as follows:
  \begin{enumerate}
    \item $\clockvaluation \models x \bowtie c$ iff $\clockvaluation(x) \bowtie c$ for $\operatorname{\bowtie} \in \{ <, \leq, =, \geq, > \}$,
    \item $\clockvaluation \models \clockconstraint_1 \wedge \clockconstraint_2$ iff $\clockvaluation \models \clockconstraint_1$ and $\clockvaluation \models \clockconstraint_2$.
      \qedhere
  \end{enumerate}
\end{definition}

\begin{example}
  Let $\clockvaluation_1$ be a clock valuation for $\{ x_1, x_2 \}$ with $\clockvaluation_1(x_1) = 2.5$ and $\clockvaluation_1(x_2) = 7.2$.
  Clearly, $\clockvaluation_1 \models x_1 < 3 \wedge x_2 \geq 5$.
  On the other hand, $\clockvaluation_1 \not\models x_1 = 3 \wedge x_2 \geq 5$, because $\clockvaluation_1(x_1) = 2.5 \neq 3$.
\end{example}

Clocks and clock constraints completely capture the time aspect of a timed transition system.
By adding clock constraints to transitions, we can constrain when a transition may happen, depending on the time progression of the system.
If we do this for finite automata, we obtain \aclp{TA}, which we introduce in the following section.

\subsection{Timed Automata}\label{sec:ta}

A \acfi{TA}~\parencite{alurTheoryTimedAutomata1994,alurTimedAutomata1999} extends a finite automaton with clocks and timing constraints and therefore allows modeling a real-time system.
\Iac{TA} has a finite set of \emph{clocks}~(\autoref{sec:clocks}), which are used for timing constraints on transitions and locations.
In particular, a \ac{TA} transition is not only labeled with a symbol (event), but may also have a clock constraint, which restricts the transition to certain clock valuations, and may reset some of the \ac{TA} clocks.
Similarly, a clock constraint on a location constrains when the system may enter and stay in that location.
While a \ac{TA} uses real-time clocks, it still resembles a \emph{discrete system} in the sense that it consists of (a finite number of) locations with discrete transitions between the locations.
We proceed with the formal definition and then provide an example:
\begin{definition}[\Ac{TA}]
  A \acfi{TA} is a tuple $\ta = \left(\tastates, \tastate_0, \tastates_F, \taalph, \taclocks, \invs, \taswitches \right)$ where
  \begin{itemize}
    \item $\tastates$ is a finite set of locations,
    \item $\tastate_0$ is the initial location,
    \item $\tastates_F \subseteq \tastates$ is a set of final locations,
    \item $\taalph$ is a finite set of labels
    \item $\taclocks$ is a finite set of clocks,
    \item $\invs$ is a mapping that labels each location $\tastate$ with some clock constraint from $\clockconstraints(\taclocks)$,
    \item $\taswitches \subseteq \tastates \times \taalph \times \clockconstraints(\taclocks) \times 2^\taclocks \times \tastates$ is a set of \emph{switches}, where a switch $\left(\tastate, a, \varphi, Y, \tastate'\right)$ describes the switch from location $\tastate$ to location $\tastate'$ with label $a$, clock constraints $\varphi$ and clock resets $Y$.
      \qedhere
  \end{itemize}
\end{definition}
The following example illustrates how a simple finite automaton can be extended to a \ac{TA}:
\begin{example}[\ac{TA}]\label{ex:ta}
  The following visualizes a \ac{TA} with two locations $l_1$ and $l_2$ and two events $a$ and $b$:
  \begin{center}
    \includestandalone{figures/ta}
  \end{center}
  Formally, it is a \ac{TA} $\ta = \left(\tastates, \tastate_1, \tastates_F, \taalph, \taclocks, \invs, \taswitches \right)$, where
  \begin{itemize}
    \item $\tastates = \{ \tastate_1, \tastate_2 \}$,
    \item $\tastates_F = \{ \tastate_1 \}$,
    \item the initial location is $l_1$,
    \item $\taalph = \{ a, b\}$,
    \item $\taclocks = \{ x_1, x_2 \}$,
    \item $\invs(\tastate_1) = x_1 \leq 1$ and $I(\tastate_2) = x_2 < 2$,
    \item $\taswitches = \{
        \left(\tastate_1, a, x_2 \geq 1, \emptyset, \tastate_2\right),
        \left(\tastate_2, a, x_1 > 0, \{ x_1 \}, \tastate_2\right),
        \left(\tastate_2, b, \top, \{ x_2 \}, \tastate_1\right)
      \}$.
  \end{itemize}
  It consists of two locations, the starting location $l_1$ and a second location $l_2$.
  It may transition between the locations on the events $a$ and $b$, while it may also stay in $l_2$ if an $a$ event occurs.
  The clock constraints restrict the automaton such that it effectively stays in $l_1$ for exactly one time unit and such that it may read the symbol $a$ repeatedly in location $l_2$, but only with some time delay greater than zero.
  Also, it may do so only as long as the value of the clock $x_2$ is smaller than $2$.
\end{example}

We continue with the semantics by defining the \emph{language} of a \ac{TA}, which defines the timed words that are accepted by a \ac{TA}.
In order to do so, we we first build \iac{LTS} corresponding to a \ac{TA}.
In the \ac{LTS}, each location consists of a \ac{TA} location and a clock valuation and the transitions consist of a symbol and a time step built from the \ac{TA} transitions.
Thus, the \ac{LTS} can be considered as a time expansion of the \ac{TA}, where each possible state (i.e., location and clock valuation) of the \ac{TA} is considered to be a separate state.
Formally:
\begin{definition}\label{def:ta-lts}
  Let $\ta = (\tastates, \tastate_0, \tastates_F, \taalph, \taclocks, \invs, \taswitches)$ be a \ac{TA}.
  The corresponding \ac{LTS} $\talts = \left(\taltsstates, \taltsstates^0, \taltsstates^F, \taltsalph, \taltstrans{}{}\right)$ is defined as follows:
  \begin{itemize}
    \item A state $\taltsstate \in \taltsstates$ of $\talts$ is a pair $\left(\tastate, \clockvaluation\right)$ such that
      \begin{enumerate}
        \item \tastate is a location of \ta,
        \item \clockvaluation is a clock valuation for the clocks \taclocks of \ta, and
        \item $\clockvaluation \models \invs(\tastate)$,
      \end{enumerate}
    \item The initial state $(\tastate_0, \zeroclocks)$ consists of the initial location $l_0$ and a clock valuation \zeroclocks where all clocks are zero-initialized,
    \item The final states $\taltsstates^F = \{ (\tastate, \clockvaluation) \mid \tastate \in \tastates_F \}$ are those states that contain a final location of \ta,
    \item The labels $\taltsalph$ consist of the labels of the \ac{TA} and time increments, i.e., $\taltsalph = \taalph \times \mathbb{R}$.
    \item A transition $\left(\tastate, \clockvaluation\right) \taltstrans{\delta}{a} \left(\tastate', \clockvaluation'\right)$ consists of two steps:%
      \footnote{
        Sometimes (e.g., \parencite{alurTimedAutomata1999}), a transition is split into two separate transitions, one for each step, which allows multiple consecutive time transitions, but since time transitions are additive~\cite{alurTimedAutomata1999}, they allow the same switches and thus having separate transitions results in the same timed words.
      }
      \begin{enumerate}
        \item \emph{Elapse of time:} All clocks are incremented by some time increment $\delta \in \realpos$ that satisfies the invariant of the source location, i.e., $\clockvaluation^* = \clockvaluation + \delta$ and for all $0 \leq \delta' \leq \delta$, $\clockvaluation + \delta' \models \invs(\tastate)$,
        \item \emph{Switch of location:} The location changes based on a switch $\left(\tastate, a, \varphi, Y, \tastate'\right) \in \taswitches$, where the clock constraint $\varphi$ must be satisfied by the incremented clocks $\clockvaluation^*$ and $Y$ specifies which clocks are reset after the transition, i.e., $\clockvaluation^* \models \varphi$ and $\clockvaluation' = \clockvaluation^*[Y \eqdef 0]$.
          \qedhere
      \end{enumerate}
  \end{itemize}
\end{definition}
The following example illustrates such \iac{LTS}:
\begin{example}[\ac{TA} \ac{LTS}]\label{ex:ta-lts}
  \begin{figure}[tbh]
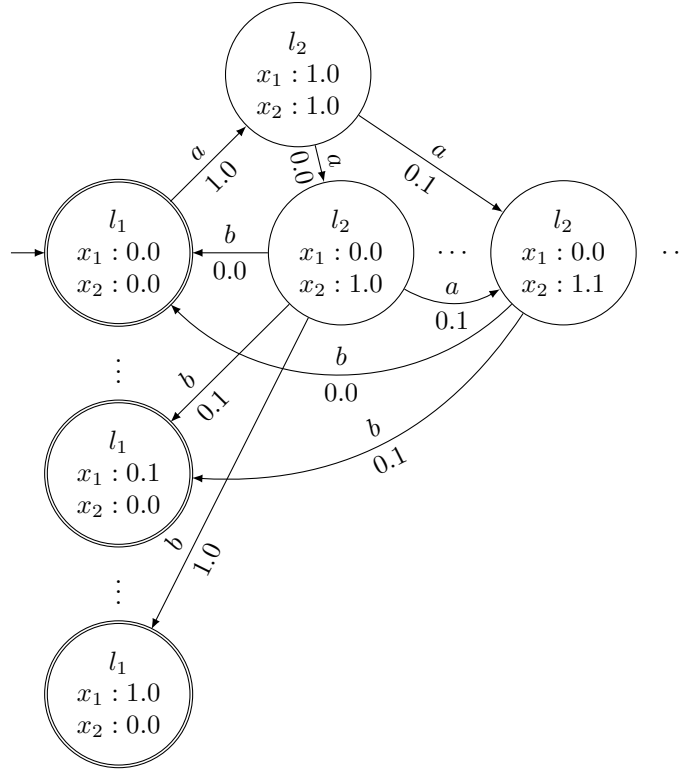

    \centering
    \includestandalone{figures/ta-lts}
    \caption[The \acs*{LTS} corresponding to a \acs*{TA}.]{The \acs*{LTS} corresponding to the \acs*{TA} \ta from \autoref{ex:ta}.}
    \label{fig:ta-lts}
  \end{figure}
  \autoref{fig:ta-lts} shows the \ac{LTS} corresponding to the \ac{TA} from \autoref{ex:ta}.
  The \ac{LTS} has uncountably infinitely many states, e.g., because it allows a transition
  \[
    \left(l_2, \{ x_1: 0.0, x_2: 1.0 \}\right) \taltstrans{\delta}{b} \left(l_1, \{ x_1: \delta, x_2: 0.0 \}\right)
  \]
  for every $\delta \in \lbrack 0, 1 \rbrack \subseteq \realpos$.
\end{example}

Using the \ac{LTS} corresponding to a \ac{TA}, we can now define the language of the \ac{TA}:
\begin{definition}[Language of a \ac{TA}]\label{def:ta-language}
  Given a (finite or infinite) run
  \[
    p = \left(\tastate_0, \clockvaluation_0\right) \taltstrans{\delta_1}{a_1} \left(\tastate_1, \clockvaluation_1\right) \taltstrans{\delta_2}{a_2} \ldots \taltstrans{\delta_n}{a_n} \left(\tastate_{n+1}, \clockvaluation_{n+1}\right) \ldots \in \runs(\ta)
  \]
  on a \ac{TA} \ta.
  The \emph{timed word induced by $p$} is the timed word
  \[
    \tw(p) = \left(a_1, \delta_1\right) \left(a_2, \delta_1 + \delta_2\right) \ldots (a_n, \sum_1^n \delta_i) \ldots
  \]
  The \emph{language of finite words of \ta} is the set
  \[
    \lang^*(\ta) = \{ \tw(p) \mid p \in \accfinruns(\ta) \}
  \]
  The \emph{language of infinite words of \ta} is the set
  \[
    \lang^\omega(\ta) = \{ \tw(p) \mid p \in \accinfruns(\ta) \}
  \]
  We also write $\lang(\ta)$ for the union $\lang(\ta) = \lang^*(\ta) \cup \lang^\omega(\ta)$.
\end{definition}
This allows us to define the language of the \ac{TA} from \autoref{ex:ta}:
\begin{example}[Language of a \ac{TA}]
  The language of the \ac{TA} shown in \autoref{ex:ta} contains the following finite words:
  \begin{gather*}
      \left(\left(a, 1.0\right) \left(a, 1.0\right) \left(b, 1.0\right)\right)
      \\
      \left(\left(a, 1.0\right) \left(a, 1.0\right) \left(b, 1.1\right)\right)
      \\
      \vdots
      \\
      \left(\left(a, 1.0\right) \left(a, 1.0\right) \left(b, 2.0\right)\right)
      \\
      \left(\left(a, 1.0\right) \left(a, 1.0\right) \left(b, 1.0\right)\left(a, 2.0\right) \left(a, 2.0\right) \left(b, 2.0\right)\right)
      \\
      \left(\left(a, 1.0\right) \left(a, 1.0\right) \left(b, 1.0\right)\left(a, 2.0\right) \left(a, 2.1\right) \left(b, 2.1\right)\right)
      \\
      \left(\left(a, 1.0\right) \left(a, 1.0\right) \left(b, 1.0\right)\left(a, 2.0\right) \left(a, 2.1\right) \left(b, 2.1\right)\left(a, 3.1\right) \left(a, 3.2\right) \left(b, 3.3\right)\right)
      \qedhere
  \end{gather*}
\end{example}

\subsubsection{Regionalization}
As the \ac{LTS} corresponding to a \ac{TA} has infinitely many states, it is not possible to directly analyze it, e.g., for checking whether a certain state can be reached or whether the language of the automaton is empty.
A common technique to solve such problems is \emph{regionalization}~\cite{alurTheoryTimedAutomata1994}, which involves constructing a discrete and finite quotient of the system.
The construction is based on an equivalence relation on the state space, where two states are considered to be equivalent if they agree on the integral parts of all clock values and on the ordering of the fractional parts of all clock values.
The integral parts are needed to check whether a given clock constraint is satisfied, the ordering of the fractional parts is needed to determine which clock will change its integral part first.
For this construction to work, we usually assume that all numeric constants mentioned in clock constraints are integral.
For a given \ac{TA} with rational clock constraints, we may multiply all constraints by the least common multiple to obtain \iac{TA} that only uses integral constraints.
Also, for a given \ac{TA}, the largest integer mentioned in any clock constraint is known and finite.
As these clock constraints are the only way to distinguish two clock values, any clock values larger than the largest integer can not be distinguished.
Therefore, for the equivalence relation, we only need to distinguish clock values less than (or equal to) the largest integer and we may consider all clock values above the maximal integer to be equivalent.
Formally, the equivalence relation is defined as follows:
\begin{restatable}[Clock Regions]{definition}{clockregions}\label{def:regionalization}
  Given a maximal constant $\maxconst$, let $V = [0, \maxconst] \cup \{ \top \}$.
  We define the \emph{region equivalence} as the equivalence relation $\clockequiv$ on $V$ such that $u \clockequiv v$ if
  \begin{itemize}
    \item $u = v = \top$, or
    \item $u, v \neq \top$, $\lceil u \rceil = \lceil v\rceil$, and $\lfloor u \rfloor = \lfloor v \rfloor$.
  \end{itemize}
  A \emph{region} is an equivalence class of $\operatorname{\clockequiv}$.
  The corresponding set of equivalence classes is $\regions = \{ r_0, r_1, \ldots, r_{2 \maxconst + 1}\}$, where $r_{2i} = \{ i \}$ for $i \leq \maxconst$, $r_{2i + 1} = (i, i + 1)$ for $i < \maxconst$, and $r_{2 \maxconst + 1} = \{ \top \}$.

  We define the  \emph{fractional part} $\fract(v)$ of $v \in V$ as follows:
  \[
    \fract(v) \eqdef
    \begin{cases}
      v - \lfloor v \rfloor & \text{ if } v \in [0, \maxconst]
      \\
      0 & \text{ if } v = \top
    \end{cases}
  \]

  We extend region equivalence to clock valuations of $n$ clocks.
  Let $\clockvaluation, \clockvaluation' \in \realpos^n$.
  We say $\clockvaluation$ and $\clockvaluation'$ are region-equivalent, written $\clockvaluation \regequiv \clockvaluation'$ iff
  \begin{enumerate}
    \item for every $i$, $\clockvaluation_i \clockequiv \clockvaluation'_i$,
    \item for every $i, j$, $\fract(\clockvaluation_i) \leq \fract(\clockvaluation_j)$ iff $\fract(\clockvaluation'_i) \leq \fract(\clockvaluation'_j)$.
  \end{enumerate}
  We denote the equivalence class of a clock valuation \clockvaluation induced by $\regequiv$ with $\regequivclass {\clockvaluation}$.
\end{restatable}

We illustrate clock regions with the following example:
\begin{example}[Regionalization]
  \begin{figure}[ht]
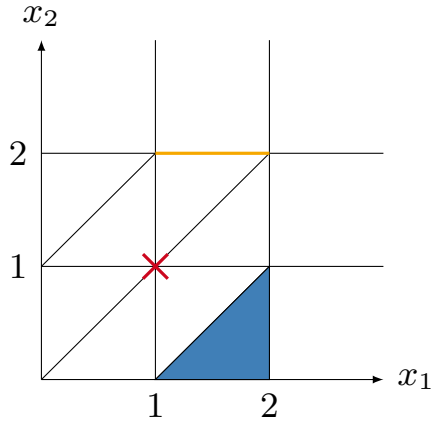

    \centering
    \includestandalone{figures/regionalization}
    \caption[The regions for a system with two clocks.]{The regions for a system with two clocks $x_1$ and $x_2$ and a maximal constant of $\maxconst = 2$.
      Highlighted are examples for a \textcolor{rwth-red}{corner point}, a \textcolor{rwth-orange}{line segment}, and an \textcolor{rwth-blue}{open region}.
      Adapted from \cite{alurTheoryTimedAutomata1994}.
    }
    \label{fig:regionalization}
  \end{figure}
  \autoref{fig:regionalization} shows all clock regions for the \ac{TA} from \autoref{ex:ta}, which is a system with two clocks and a maximal constant $\maxconst = 2$.
  The regions consist of
  \begin{itemize}
    \item $9$ corner points, e.g., $x_1 = x_2 = 1$,
    \item $22$ line segments, e.g, $1 < x_1 < 2 \wedge x_2 = 2$, and
    \item $13$ open regions, e.g., $x_1 \in (1, 2) \wedge x_2 \in (0, 1) \wedge \fract(x_1) < \fract(x_2)$.
      \qedhere
  \end{itemize}
\end{example}
Based on the region equivalence relation, we can define \emph{region automata}, where each state of the automaton is a region, i.e., an equivalence class of the region equivalent relation:
\begin{definition}[Region Automaton]
  Given an \ac{LTS} $\lts = \left(\taltsstates, \taltsstates^0, \taltsalph, \taltstrans{}{}\right)$ associated with some \ac{TA} \ta over alphabet $\taalph$ and with maximal constant $\maxconst$.
  The \emph{region automaton} $\reglts$ is an \ac{LTS} $\reglts = \left(\regltsstates, \regltsstate_0, \regltsalph, \regltstrans\right)$ defined as follows:
  \begin{itemize}
    \item $\regltsstates = \left\{ \left(\tastate, \regequivclass{\clockvaluation}\right) \mid \left(\tastate, \clockvaluation\right) \in \ltsstates \right\}$,
    \item $\regltsstate_0 = (\tastate_0, \regequivclass{\vec{0}})$,
    \item The labels \taalph of \reglts are the labels of the \ac{TA} \ta,
    \item $\left(\tastate_1, \regequivclass{\clockvaluation_1}\right) \regltstrans[a] \left(\tastate_2, \regequivclass{\clockvaluation_2}\right)$ if there is some $\delta \in \realpos$ such that $\left(\tastate_1, \clockvaluation_1\right) \taltstrans{\delta}{a} \left(\tastate_2, \clockvaluation_2\right)$.
      \qedhere
  \end{itemize}
\end{definition}
It can be shown~\parencite{alurTheoryTimedAutomata1994} that runs on the region automaton correspond to runs on the \acl{TA} and vice versa.
This allows to use the region automaton as basis for various problems, e.g., deciding language emptiness or reachability.
We finish the discussion of regionalization with an example:
\begin{example}[Region Automaton]
  \autoref{fig:region-lts} shows the region automaton $\reglts$ of the \ac{TA} from \autoref{ex:ta}.
  Note that in contrast to the corresponding \ac{LTS} \lts shown in \autoref{fig:ta-lts}, \reglts has only finitely many states.
  \begin{figure}[ht]
    \centering
    \includestandalone{figures/region-lts}
    \caption[A region automaton.]{The region automaton $\reglts$ for the \ac{TA} from \autoref{ex:ta}, which regionalizes the \ac{LTS} from \autoref{fig:ta-lts}.
      \todo[inline]{Explain where representatives come from}
    }
    \label{fig:region-lts}
  \end{figure}
\end{example}

\subsubsection{Decidable and Undecidable Extensions}
The properties of \acp{TA} as well as various extensions have been studied extensively.
Here, we summarize some relevant properties and we refer to \parencite{alurDecisionProblemsTimed2004} for a survey.
Already \textcite{alurTheoryTimedAutomata1994} have shown that deciding the language emptiness of a given \ac{TA} is \textsc{Pspace}-complete and that deciding whether a \ac{TA} accepts all timed words is undecidable.
As a direct corollary, the language inclusion problem of deciding whether the language of a \ac{TA} $\ta_1$ is a subset of the language of a second \ac{TA} $\ta_2$, i.e., $\lang(\ta_1) \subseteq \lang(\ta_2)$, is also undecidable.
Subsequent work has focused on restricting or extending \acp{TA} syntactically.

One possible restriction is the number of clocks.
For \acp{TA} with one clock, the reachability problem is \textsc{Nlogspace}-complete, while it is \textsc{NP}-hard for automata with two clocks~\parencite{laroussinieModelCheckingTimed2004}.
For automata with at least three clocks, the reachability problem is \textsc{Pspace}-complete~\parencite{courcoubetisMinimumMaximumDelay1992}.
For language inclusion and universality, the problem is already undecidable with two clocks~\parencite{alurTheoryTimedAutomata1994}, but it is decidable if the automaton only has a single clock~\parencite{ouaknineLanguageInclusionProblem2004}.
The latter result is particularly interesting, as the proof is based on converting the problem to a reachability problem on an infinite state space of the two automata and then, in addition to regionalization, using \aclp{wqo} to guarantee termination.
We will use a similar technique in \autoref{sec:wsts} for verification and synthesis of timed \golog programs.

A second way to extend \aclp{TA} is to allow more expressive guards.
\emph{Diagonal clock constraints} of the form $x - y \bowtie c$ allow the comparison of the difference of two clock values.
However, they do not increase the expressiveness of \acp{TA}~\parencite{alurTheoryTimedAutomata1994,berardCharacterizationExpressivePower1998} and therefore can be seen as syntactic sugar.
On the other hand, allowing \emph{additive clock constraints} of the form $x + y \bowtie c$ renders the emptiness problem undecidable if the automaton has at least four clocks~\parencite{alurTheoryTimedAutomata1994,berardTimedAutomataAdditive2000}.
If restricted to two clocks, the problem remains decidable~\parencite{berardTimedAutomataAdditive2000}.

Third, \textcite{bouyerUpdatableTimedAutomata2004} have studied updatable timed automata, which allow setting clocks to values other than zero.
Updates of the form $x \eqdef x + 1$ render the emptiness problem \textsc{Pspace}-complete with diagonal-free clock constraints and undecidable otherwise.
Setting a clock to the value of another clock with updates of the form $x \eqdef y$ does not increase expressiveness, hence the emptiness problem remains \textsc{Pspace}-complete.
As a third example, allowing decrements of the form $x \eqdef x - 1$ make \aclp{TA} Turing-complete.
\emph{Event-recording automata}~\parencite{alurEventclockAutomataDeterminizable1999} restrict clock resets such that each clock is associated with an event and therefore tracks the time since the last occurrence of the event.
With this restriction, the language inclusion problem is decidable.

Finally, \emph{hybrid automata}~\parencite{alurHybridAutomataAlgorithmic1993,raskinIntroductionHybridAutomata2005} can be seen as a generalization of timed automata, where clocks are replaced by variables.
The values of variables continuously change over time and are governed by a set of differential equations called \emph{flow functions} that depend on the current state.
Additionally, a variable may be assigned to a new value on a discrete jump step of the automaton.
For general hybrid automata, the reachability problem is undecidable~\parencite{alurHybridAutomataAlgorithmic1993}.
In \emph{linear hybrid automata}~\parencite{alurAutomaticSymbolicVerification1996}, the constraints are restricted to linear constraints on the first derivatives.
In this case, the reachability problem is semi-decidable~\parencite{alurAutomaticSymbolicVerification1996}.
Finally, a hybrid automaton is called \emph{initialized} if each variable is reinitialized (i.e., assigned to value in a given interval with constant bounds) whenever its flow function changes  and \emph{rectangular} if all constraints are restricted to rectangular sets, i.e., Cartesian products of intervals with fixed rational endpoints~\parencite{henzingerWhatDecidableHybrid1998,abrahamModelingAnalysisHybrid2012} .
For initialized rectangular hybrid automata, reachability is \textsc{Pspace}-complete, while it is undecidable if the automaton is not uninitialized or non-rectangular~\parencite{henzingerWhatDecidableHybrid1998}.

As this discussion shows, the boundary of decidability has been well-studied and sometimes, simple extensions already result in undecidability.
Therefore, when we extend the logic \esg with time in \autoref{chap:timed-esg}, we will use a syntactic restriction based on clock formulas, similar to clock constraints in timed automata.
This will allow us to use regionalization for the verification and synthesis problems and ensure that those problems remain decidable.

\subsection{Alternating Timed Automata}\label{sec:ata}

Nondeterminism plays an important role in formal systems, e.g., in the form of nondeterministic finite automata~\parencite{rabinFiniteAutomataTheir1959} or nondeterministic Turing machines~\parencite{hopcroftFormalLanguagesTheir1969}.
In such nondeterministic machines, the transition rule allows to switch from one configuration to several different successor configurations.
Timed automata are also nondeterministic, as the automaton may have multiple switches in the same location with the same input symbol.
In all of those systems, the interpretation of such a nondeterministic transition is that the system may take one of the several alternatives and the machine accepts an input if some successor leads to an accepting configuration.
In that sense, they can be considered to be \emph{existential branches}.
\emph{Alternation}~\parencite{chandraAlternation1981} generalizes this idea by adding \emph{universal branches}, e.g., in the form of an alternating Turing machine.
In a universal branch, an input is only accepted if all successors lead to  an accepting  configuration.
Sometimes, the alternation leads to more expressive formalisms, e.g., in the form of alternating pushdown automata~\parencite{chandraAlternation1981}.

These observations motivate the generalization of timed automata to \emph{alternating timed automata}~\parencite{lasotaAlternatingTimedAutomata2005} to obtain a more expressive yet decidable formalism.
As language inclusion and universality are already undecidable for timed automata with at least two clocks (see above), alternating timed automata are usually restricted to a single clock.
Indeed, \textcite{lasotaAlternatingTimedAutomata2005} have shown that the emptiness problem for alternating timed automata with one clock is decidable.
As alternating timed automata are closed under boolean operations, the universality problem is also decidable.
Moreover, they have shown that there are languages recognizable by alternating timed automata with one clock that are not recognizable by timed automata with any number of clocks.

While this is interesting from a theoretical point of view, more directly relevant for this thesis are results by \textcite{ouaknineDecidabilityMetricTemporal2005}, who have shown that given \iac{MTL} formula $\phi$, one can construct \iac{ATA} \ata that accepts precisely those words that satisfy $\phi$.
In the following, we summarize the construction from \parencite{ouaknineDecidabilityMetricTemporal2005,ouaknineDecidabilityComplexityMetric2007}.


We start with location formulas, which specify the target configurations of a transition:
\begin{definition}[\ac{ATA} location formula]
Let $L$ be a finite set of locations. The set of formulas $\atalocformulas$ is generated by the following grammar:
  \[
    \varphi ::= \top \smid \bot \smid \varphi_1 \wedge \varphi_2 \smid \varphi_1 \vee \varphi_2 \smid l \smid x \bowtie k \smid x.\varphi
  \]
where $k \in \mathbb{N}$, $\operatorname{\bowtie} \in \{ <, \leq, =, \geq, > \}$, and $l \in L$.
\end{definition}
Intuitively, the configuration after doing a transition is defined by the minimal model of a location formula.
If $\phi$ is a location $l$, then the target configuration simply consists of the single location $l$.
The disjunction $\phi_1 \vee \phi_2$ corresponds to \emph{existential branching}, as the target configuration may be model of $\phi_1$ or $\phi_2$.
Similarly, conjunctions of the form $\phi_1 \wedge \phi_2$ correspond to \emph{universal branching}.
Finally, we also allow clock constraints and clock resets that use the implicit clock $x$ of the automaton.
We can now define the automaton:
\begin{definition}[\ac{ATA}]
	An \acfi{ATA} is a tuple \newline $\ata[] = \left(\atalocations, l_0, F, \ataalphabet, \atatrans\right)$ where
    \begin{itemize}
      \item $\atalocations$ is a finite set of locations,
      \item $l_0$ is the initial location,
      \item $F \subseteq \atalocations$ is a set of accepting locations,
      \item $\ataalphabet$ is a finite alphabet, and
      \item $\atatrans: \atalocations \times \ataalphabet \rightarrow \atalocformulas$ is the transition function.
    \end{itemize}

  An \ac{ATA} has an implicit single clock $x$.
  A state of \ata[] is a pair $(l, \clockvaluation)$, where $l \in \atalocations$ is the location and $\clockvaluation \in \realpos$ is a \emph{clock valuation} of the clock $x$.
  We denote the set of all possible states with $\atastates = \atalocations \times \realpos$.
  A \emph{configuration} $\ataconf$ of \ata[] is a finite set of states $\ataconf \subseteq \atastates$.
  The initial configuration is $\ataconf_0 = \{(l_0, 0)\}$ and we denote the set of all configurations with $\ataconfs$.
  A configuration $\ataconf$ is accepting if $l \in F$ for all $(l, u) \in \ataconf$.
\end{definition}

Before defining the semantics, we provide an example for \iac{ATA} that recognizes a simple language:
\begin{example}[\ac{ATA} for time-bounded response \parencite{ouaknineDecidabilityMetricTemporal2005}]\label{ex:ata}
  The time-bounded response property \emph{for every $a$-event, there is a $b$-event exactly one time unit later} can be expressed by the following \ac{ATA}:
  \begin{itemize}
    \item The alphabet consists of the two events, i.e., $\ataalphabet = \{ a, b \}$.
    \item There is one location $l_0$ to say that for every $a$-event, a $b$-event has occurred and thus no $b$-event is pending, and one location $l_1$ to say a $b$-event is pending, i.e., $L = \{ l_0, l_1 \}$.
    \item Accept as long as no $b$-event pending, i.e., $F = \{ l_0 \}$.
    \item The transition function \atatrans is given by the following table:
      \begin{center}
        \begin{tabular}{ccc}
          \toprule
          & $a$ & $b$
          \\
          \cmidrule(lr){2-3}
          $l_0$ & $l_0 \wedge x.\, l_1$ & $l_0$
          \\
          $l_1$ & $l_1$ & $(x = 1) \vee l_1$
          \\
          \bottomrule
        \end{tabular}
      \end{center}
      where
      \begin{itemize}
        \item $\atatrans(l_0, a) = l_0 \wedge x.\, l_1$ to say that if an $a$-event occurs in $l_0$, then reset the clock $x$ and go to location $l_1$, as a $b$-event is pending.
          At the same time, stay in $l_0$, as another $a$-event may occur.
        \item $\atatrans(l_0, b) = l_0$ as no $b$-event is pending and no $a$-event has occurred.
        \item $\atatrans(l_1, a) = l_1$ to say that a $b$-event is still pending,
        \item $\atatrans(l_1, b) = (x = 1) \vee l_1$ to say that if $x = 1$, then the pending $b$-event has occurred at the correct time, i.e., one time unit after the corresponding $a$-event that reset the clock.
          Otherwise, the $b$-event is still pending.
          \qedhere
      \end{itemize}
  \end{itemize}
\end{example}

Regarding the semantics, we first define when a location formula is satisfied:
\begin{definition}[Truth of location formulas]
Given a set of states $M \subseteq\atastates$ and a clock valuation $\clockvaluation \in \realpos$, the truth of a formula $\varphi \in \atalocformulas$ is defined as follows:
\begin{enumerate}
  \item $M, \clockvaluation \models \top$,
  \item $M, \clockvaluation \not\models \bot$,
  \item $M, \clockvaluation \models l$ iff $\left(l, \clockvaluation\right) \in M$,
  \item $M, \clockvaluation \models x \bowtie k$ iff $\clockvaluation \bowtie k$,
  \item $M, \clockvaluation \models x.\varphi$ iff $M, 0 \models \varphi$,
  \item $M, \clockvaluation \models \varphi_1 \wedge \varphi_2$ iff $M, \clockvaluation \models \varphi_1$ and $M, \clockvaluation \models \varphi_2$,
  \item $M, \clockvaluation \models \varphi_1 \vee \varphi_2$ iff $M, \clockvaluation \models \varphi_1$ or $M, \clockvaluation \models \varphi_2$.
\end{enumerate}
  The set of states $M$ is a \emph{minimal model of $\varphi$ with respect to \clockvaluation} if $M, \clockvaluation \models \varphi$ and there is no proper subset $N \subsetneq M$ with $N, \clockvaluation \models \varphi$.
\end{definition}
As pointed out by \textcite{ouaknineDecidabilityMetricTemporal2005}, the minimal models can be directly read off from a location formula:
\begin{remark}[\cite{ouaknineDecidabilityMetricTemporal2005}]\label{remark:ata-dnf}
  A location formula atom is a term of the form $l, x.\,l$, or $x \bowtie k$.
  Every location formula $\varphi \in \loc(X)$ can be rewritten in disjunctive normal form as $\varphi = \bigvee_i \bigwedge A_i$, where each $A_i$ is a set of atoms.
  Given a location formula $\varphi$ in disjunctive normal form, the minimal models can be read off as follows:
  For each set of atoms $A_i$ and clock valuation $\clockvaluation \in \realpos$, let $A[\clockvaluation] \subseteq \atastates$ denote the set of states $A[\clockvaluation] \eqdef \{ (l, v) \mid l \in A \} \cup \{ (l, 0) \mid x.\,l \in A \}$ (note that $A[\clockvaluation]$ may not contain clock constraints or clock resets, as these are not valid states of the \ac{ATA}).
  Then each minimal model $M$ of $\varphi$ has the form $M = A_i[\clockvaluation]$ for some $i$, where $v$ satisfies all the clock constraints in $A_i$.
\end{remark}

We demonstrate models and minimal models by continuing the example from \autoref{ex:ata}:
\begin{example}[Models of location formulas]
  Let $\varphi_1 = l_0 \wedge x. l_1$ and $\varphi_2 = \left(x = 1\right) \vee l_1$ be the location formulas from \autoref{ex:ata}.
  Let
  \begin{align*}
    M_1 &= \{ (l_0, 0.5), (l_1, 0.0) \} & \clockvaluation_1 &= \{ x: 0.5 \}
    \\
    M_2 &= \{ (l_0, 0.5), (l_1, 1.0) \} & \clockvaluation_2 &= \{ x: 1.0 \}
    \\
    M_3 &= \{ (l_0, 0.5), (l_1, 0.0), (l_1, 1.0) \} & &
    \\
    M_4 &= \emptyset
  \end{align*}
  Then:
  \begin{itemize}
    \item $M_1, \clockvaluation_1 \models \varphi_1$ because
      \begin{enumerate}
        \item $(l_0, 0.5) \in M_1$ and thus $M_1, \clockvaluation_1 \models l_0$, and
        \item $(l_1, 0.0) \in M_1$, thus $M_1, 0 \models l_1$ and therefore $M_1, \clockvaluation_1 \models x.l_1$.
      \end{enumerate}
      Furthermore, $M_1$ is a minimal model of $\varphi_1$ with respect to $\clockvaluation_1$.
    \item $M_2, \clockvaluation_2 \not\models \varphi_1$ because $\left(l_1, 0.0\right) \not\in M_2$ and thus $M_2, \clockvaluation_2 \not\models x.l_1$.
    \item $M_3, \clockvaluation_1 \models \varphi_1$ for the same reasons as $M_1, \clockvaluation_1 \models \varphi_1$.
      However, as $M_1 \subsetneq M_3$, $M_3$ is not a minimal model of $\varphi_1$ with respect to $\clockvaluation_1$.
    \item $M_4, \clockvaluation_2 \models \varphi_2$ because $\clockvaluation_2 \models \left(x = 1\right)$.
      Clearly, $M_4$ is also a minimal model of $\varphi_2$ with respect to $\clockvaluation_2$.
  \end{itemize}
  Following \autoref{remark:ata-dnf}, we can read off the minimal models as follows:
  \begin{itemize}
    \item For $\phi_1$, there is no disjunct and therefore there is a single set of atoms $A_1 = \{ l_0, x.\, l_1 \}$.
      For $\clockvaluation_1$, we obtain $A_1[\clockvaluation_1] = \{ (l_0, 0.5), (l_1, 0.0) \}$ and therefore, $M = \{ (l_0, 0.5), (l_1, 0.0) \}$.
      Similarly, for $\clockvaluation_2$, we obtain $A_1[\clockvaluation_2] = \{ (l_0, 0.5), (l_1, 1.0) \}$ and hence $M = \{ (l_0, 0.5), (l_1, 1.0) \}$.
    \item The location formula $\phi_2$ is already in disjunctive normal form, where $A_1 = \{ (x = 1) \}$ and $A_2 = \{ l_1 \}$.
      For $\clockvaluation_1$, there is no minimal model, because $\clockvaluation_1$ does not satisfy the clock constraint $(x = 1) \in A_1$.
      For $\clockvaluation_2$, we obtain $A_1[\clockvaluation_2] = \{ \}$ because $A_1$ contains no \ac{ATA} location.
      As $\clockvaluation_2$ satisfies the only clock constraint $(x = 1)$ in $A_1$, the unique minimal model of $\varphi_2$ with respect to $\clockvaluation_2$ is $M = \emptyset$.
      \qedhere
  \end{itemize}
\end{example}

Similar to \acp{TA}, the language accepted by an \ac{ATA} is defined in terms of \iac{LTS}:
\begin{definition}[\ac{ATA} \ac{LTS}]\label{def:ata-lts}
  Let $\ata[] = (\atalocations, l_0, F, \ataalphabet, \atatrans)$ be an \ac{ATA}.
  The corresponding \ac{LTS} $\atalts = \left(\ataltsstates, \ataltsstate^0, \ataltsstates^F, \ataltsalph, \ataltstrans{}{}\right)$ is defined as follows:
  \begin{itemize}
    \item A state $\ataltsstate$ of $\atalts$ is a configuration of \ata[], i.e., $\ataltsstates = \ataconfs$.
    \item The initial state $\ataltsstate^0$ of \atalts is the initial configuration, i.e., $\ataltsstate^0 = \left( l_0, 0\right)$.
    \item A state $\ataltsstate$ is accepting if each contained location is accepting, i.e., for $\ataltsstate = \{ \left(l_1, t_1\right), \left(l_2, t_2\right), \ldots, \left(l_k, t_k\right) \}$, $\ataltsstate \in \ataltsstates^F$ iff $l_i \in F$ for every $i$.
    \item The labels $\ataltsalph$ consist of symbols of \ata[] and time increments, i.e., $\ataltsalph = \ataalphabet \times \realpos$,
    \item A transition $\ataconf \ataltstrans{\delta}{a} \ataconf'$ consists of two steps:
      \begin{enumerate}
        \item \emph{Elapse of time:} All clock valuations are incremented by some time increment $\delta \in \realpos$, i.e.,
          \[
            \ataconf^* = \left\{ \left(l, \clockvaluation + \delta\right) \mid \left(l, \clockvaluation\right) \in \ataconf \right\}
          \]
        \item \emph{Switch of location:} The location changes instantaneously based on the transition function $\atatrans$.
          A target configuration $\ataconf'$ contains for each state $\left(l_i, \clockvaluation_i\right) \in \ataconf^*$ a minimal model of $\atatrans(l_i, a)$ with respect to $\clockvaluation_i$,  i.e.,
          \[
            \ataconf' = \bigcup_{ \left(l_i, \clockvaluation_i\right) \in \ataconf^*} \{ M_i \mid M_i \text{ is some minimal model of $\atatrans(l_i, a)$ with respect to $\clockvaluation_i$ } \}
          \]
      \end{enumerate}
      As each $\atatrans(l_i, a)$ may have more than one minimal model with respect to $\clockvaluation_i$, there may also be multiple target configurations for a given start configuration $\ataconf$ and symbol $a$.
      \qedhere
  \end{itemize}
\end{definition}

We demonstrate such \iac{LTS} by continuing the running example:
\begin{example}[\ac{ATA} \ac{LTS}]
  \begin{figure}[ht]
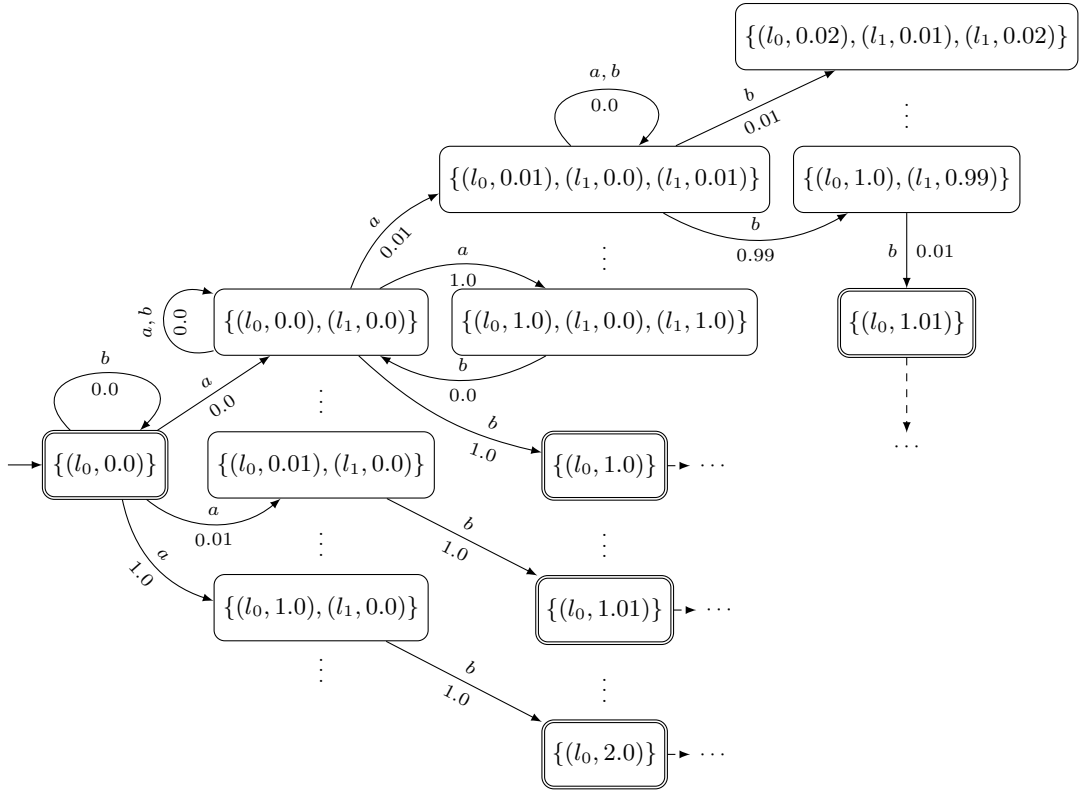

    \centering
    \includestandalone{figures/ata-lts}
    \caption[The \acs*{LTS} corresponding to an \acs*{ATA}.]{The \ac{LTS} corresponding to the \ac{ATA} from \autoref{ex:ata}.}
    \label{fig:ata-lts}
  \end{figure}
  \autoref{fig:ata-lts} shows the \ac{LTS} corresponding to the \ac{ATA} from \autoref{ex:ata}.
  The \ac{LTS} has uncountably infinitely many states, e.g., because it allows a transition
  \[
    \left\{ \left(l_0, 0.0\right) \right\} \ataltstrans{\delta}{a} \left\{ \left(l_0, \delta\right), \left(l_1, 0.0\right) \right\}
  \]
  for every $\delta \in \realpos$.
\end{example}

Using the \ac{LTS} corresponding to the \ac{ATA}, we can now define the language of the \ac{ATA}.
\begin{definition}[Language of an \ac{ATA}] \label{def:ata-language}
  Given a finite run
  \[
    p = \ataltsstate^{(0)} \ataltstrans{\delta_1}{a_1} \ataltsstate^{(1)} \ataltstrans{\delta_2}{a_2} \ldots \ataltstrans{\delta_n}{a_n} \ataltsstate^{(n+1)} \in \runs(\atalts)
  \]
  on the \ac{LTS} \atalts.
  The \emph{timed word induced by $p$} is the timed word
  \[
    \tw(p) = \left(a_1, \delta_1\right) \left(a_2, \delta_1 + \delta_2\right) \ldots (a_n, \sum_1^n \delta_i)
  \]
  The \emph{language of \ata[]} is the set
  \[
    \lang^*(\ata[]) = \{ \tw(p) \mid p \in \accfinruns(\atalts) \}
    \qedhere
  \]
\end{definition}
Note the similarity to \autoref{def:ta-language}: Both for \acp{TA} and \acp{ATA}, the language is defined by the accepting runs of the corresponding \ac{LTS}.
For the sake of simplicity, we only define the language over finite words for \acp{ATA}.
However, the definition may be extended to infinite words analogously to the language of infinite words of \iac{TA}.

We turn back to the running example:
\begin{example}[Language of an \ac{ATA}]
  In \autoref{fig:ata-lts}, we can see that the language of the \ac{ATA} from \autoref{ex:ata} contains the following words:
  \begin{gather*}
    \left(\left(a, 0.0\right)\left(b, 1.0\right)\right)
    \\
    \left(\left(a, 0.01\right)\left(b, 1.01\right)\right)
    \\
    \left(\left(a, 0.0\right)\left(a, 0.0\right)\left(b, 1.0\right)\right)
    \\
    \left(\left(a, 0.0\right)\left(a, 0.01\right)\left(b, 1.0\right)\left(b, 1.01\right)\right)
    \qedhere
  \end{gather*}
\end{example}

A fundamental result is that both language emptiness and language inclusion is decidable for \acp{ATA}:
\begin{theorem}[\parencite{ouaknineDecidabilityMetricTemporal2005}]\label{thm:ata-decidability}
  Let $\mathcal{A}$ be \iac{ATA} and $\mathcal{B}$ be \iac{TA}.
  Then the language emptiness problem $\lang^*(\mathcal{A}) = \emptyset$ and the language inclusion problem $\lang^*(\mathcal{A}) \subseteq \lang^*(\mathcal{B})$ are both decidable.
\end{theorem}
We omit the details of the proof and instead continue with the construction of \iac{ATA} for \ac{MTL} formulas.

\subsubsection{Constructing an \ac{ATA} for an \ac{MTL} formula}

In the previous section, we have seen how \acp{ATA} generally work:
In \iac{ATA}, the transitions are defined by a transition function that maps locations to location formulas.
The successor configuration of a transition is then a minimal model of the location formula.
This allows both \emph{existential branching}, where there are multiple successor configurations that can be understood as alternatives, and \emph{universal branching}, where the successor configuration consists of multiple \ac{ATA} states.
In \iac{ATA}, existential branching is realized with disjunctions and universal branching is realized with conjunctions in the location formula.

We now summarize how such \iac{ATA} can be used to recognize the language of a given \ac{MTL} formula $\phi$, as described by \textcite{ouaknineDecidabilityMetricTemporal2005}.
Intuitively, the approach works as follows: For a given \ac{MTL} formula $\phi$ in positive normal form, we construct an \ac{ATA} where each location of the \ac{ATA} is a sub-formula of $\phi$ whose outermost connective is one of the temporal operators $\until{}$ or  $\duntil{}$.
For $\until{}$ sub-formulas, such a location represents a sub-formula of $\phi$ that has not been satisfied yet.
On the other hand, for $\duntil{}$ sub-formulas, such a location represents a sub-formula of $\phi$ that has so far been satisfied.
The automaton is constructed in such a way that it is in an accepting configuration if and only if the word read so far satisfied the formula $\phi$.
Therefore, the accepting locations of the automaton are the sub-formulas with outermost connective $\duntil{}$.

Formally, we first define the closure of a formula:
\begin{definition}[Closure of an \acs*{MTL} formula]
  Given an \ac{MTL} formula $\varphi$, the \emph{closure} of $\varphi$, written $\closure(\varphi)$, is the set of sub-formulas of $\varphi$ whose outermost connective is $\until{}$ or $\duntil{}$.
\end{definition}
We can now define the \ac{ATA} \ata for a given \ac{MTL} formula $\phi$:%
\footnote{
  In contrast to \textcite{ouaknineDecidabilityMetricTemporal2005}, we assume a state-based setting over a set of atomic propositions $P$, where each symbol in the timed word is a subset of $P$.
  We have modified the construction accordingly.
}
\begin{definition}[\acs*{MTL} \acs*{ATA}]\label{def:mtl-ata}
  Given an \ac{MTL} formula $\varphi$ over atomic propositions $P$, the corresponding \ac{ATA} $\ata = \left(\atalocations, \varphi_i, F, \ataalphabet, \atatrans\right)$ for $\varphi$ is defined as follows:
  \begin{itemize}
    \item $\ataalphabet = 2^P$,
    \item $\atalocations = \closure(\varphi) \cup \{ \varphi_i \}$,
    \item $F = \{ \psi \in \closure(\varphi) \mid \text{ the outermost connective of $\psi$ is $\duntil{}$} \}$
    \item The transition function \atatrans is defined as follows:
      \begin{align*}
        \atatrans(\varphi_i, a) &= \init(\varphi, a)
        \\
        \atatrans(\psi_1 \until{I} \psi_2, a) &= (\init(\psi_2, a) \wedge x \in I) \vee (\init(\psi_1, a) \wedge (\psi_1 \until{I} \psi_2))
        \\
        \atatrans(\psi_1 \duntil{I} \psi_2, a) &= (\init(\psi_2, a) \vee x \notin I) \wedge (\init(\psi_1, a) \vee (\psi_1 \duntil{I} \psi_2))
      \end{align*}
      where $\init$ is a helper function defined as follows:
      \begin{align*}
        \init(\psi, a) &= x.\psi \text{ if } \psi \in \closure(\varphi)
        \\
        \init(\psi_1 \wedge \psi_2, a) &= \init(\psi_1, a) \wedge \init(\psi_2, a)
        \\
        \init(\psi_1 \vee \psi_2, a) &= \init(\psi_1, a) \vee \init(\psi_2, a)
        \\
        \init(b, a) &=
        \begin{cases}
          \top & \text{ if } b \in a
          \\
          \bot & \text{ else }
        \end{cases}
        \text{ for } b \in \Sigma
        \\
        \init(\neg b, a) &= \neg \init(b, a)
        \qedhere
      \end{align*}
  \end{itemize}
\end{definition}

The $\init$ helper function works as follows: For an until formula $\psi_1 \until{I} \psi_2$ or a dual-until formula $\psi_1 \duntil{I} \psi_2$, the clock $x$ is reset, which allows tracking whether the formula is satisfied within the interval $I$.
For the boolean connectors and negation, $\init$ is defined recursively.
Finally, for atomic propositions,  $\init(b, a)$ is always true if the symbol $b$ is contained in the read symbol $a$ and false otherwise.
The transition function \atatrans intuitively works as follows: For an until formula $\psi = \psi_1 \until{I} \psi_2$, if the clock $x$ currently satisfies the constraints defined by the interval $I$, then it suffices to satisfy $\psi_2$ to satisfy $\psi$, as defined by the first disjunct $(\init(\psi_2, a) \wedge x \in I)$.
Otherwise, as indicated by the second disjunct $\init(\psi_1, a) \wedge (\psi_1 \until{I} \psi_2)$, the sub-formula $\psi_1$ must be satisfied and $\psi$ must be satisfied at some point in the future.
For a dual-until formula $\psi = \psi_1 \duntil{I} \psi_2$, the transition function works similarly, except that due to the duality of the operator, all operators are inverted, i.e., containment $\in$ is replaced by non-containment $\not\in$, conjunctions are replaced by disjunctions, and disjunctions are replaced by conjunctions.

We demonstrate the construction with an example:
\begin{example}[\ac{ATA} constructed from an \ac{MTL} formula]
	Given the \ac{MTL} formula
  \[
    \phibad = \top \until{\leq 1} \left(\neg \camon \land \grasping\right)
  \]
  the corresponding \ac{ATA} \ata[\phibad] constructed according to \autoref{def:mtl-ata} looks as follows:
  \begin{itemize}
    \item The alphabet is the power set of the set of atomic propositions $P = \left\{ \camon, \grasping \right\}$, i.e., $\ataalphabet = \left\{ \emptyset, \{ \camon \}, \{ \grasping \}, \{ \camon, \grasping \} \right\}$.
    \item The initial location is the location $\phibad^i$.
    \item There are two locations, the initial location $\phibad^i$ and one location for (the only) sub-formula with outermost connective $\until{}$, which is $\phibad$ itself, i.e., $\atalocations = \{ \phibad^i, \phibad \}$.
    \item There are no final locations, as there are no sub-formulas with outermost connective $\duntil{}$, i.e., $F = \emptyset$.
    \item The transition function $\atatrans$ is defined as follows:
      \begin{align*}
        \atatrans(\phibad^i, a) &= \phibad \text{ for every $ a \in \ataalphabet $ }
        \\
        \atatrans(\phibad, \{ \}) &= \phibad
        \\
        \atatrans(\phibad, \{ \camon \}) &= \phibad
        \\
        \atatrans(\phibad, \{\grasping \}) &= x \leq 1 \vee \phibad
        \\
        \atatrans(\phibad, \{\camon, \grasping \}) &= \phibad
      \end{align*}
  \end{itemize}
  We can see that as long as the input symbol is anything other than $\{ \grasping \}$, the automaton simply stays in the current location $\phibad$.
  When reading $\{ \grasping \}$, the automaton checks whether the timing constraint is satisfied, i.e., whether $x \leq 1$.
  If this is the case, the resulting configuration is the empty configuration $\emptyset$ and so the automaton accepts the input (and will continue to do so independent of the subsequent input symbols).
\end{example}

It can be shown that the automaton \ata indeed accepts the same language as $\phi$:
\begin{theorem}[\cite{ouaknineDecidabilityMetricTemporal2005}] \label{thm:mtl-ata}
  Given an \ac{MTL} formula $\varphi$, the \ac{ATA} \ata constructed from $\varphi$ according to \autoref{def:mtl-ata} accepts the same language as $\varphi$, i.e., $\lang^*(\varphi) = \lang^*(\ata)$.
\end{theorem}
We omit the details of the proof and instead turn towards the satisfiability and model checking problems.
Given \iac{MTL} formula $\phi$, the satisfiability  problem is to check whether there exists a timed word that satisfies $\phi$.
The model checking problem asks for a given \ac{TA} \ta and \iac{MTL} formula $\phi$, whether every word accepted by \ta satisfied $\phi$.
With \autoref{thm:ata-decidability} and \autoref{thm:mtl-ata}, it immediately follows:
\begin{corollary}[\cite{ouaknineDecidabilityMetricTemporal2005}] \label{thm:mtl-decidability}
  The satisfiability and model checking problems for \ac{MTL} over finite words are both decidable.
\end{corollary}
\textcite{ouaknineDecidabilityMetricTemporal2005} have also shown the complexity of the two problems:
\begin{theorem}[\parencite{ouaknineDecidabilityMetricTemporal2005,ouaknineDecidabilityComplexityMetric2007}]\label{thm:mtl-complexity}
  The satisfiability and model-checking problems for \ac{MTL} over finite words have non-primitive recursive complexity.
\end{theorem}

On the other hand, \textcite{ouaknineMetricTemporalLogic2006} have shown that for infinite words, both problems are undecidable:
\begin{theorem}[\parencite{ouaknineMetricTemporalLogic2006}]
  The satisfiability and model checking problems for \ac{MTL} over infinite words are both undecidable.
\end{theorem}

As we will see in the next chapter, this allows us to construct \iac{ATA} that tracks \ac{MTL} properties of \golog programs, which can be used for verification and synthesis.
As \ac{MTL} is decidable for finite words but undecidable for infinite words, we will restrict those properties to finite traces of the program.

\chapter{Timed \esg}\label{chap:timed-esg}


In \autoref{chap:foundations}, we have seen how the \sitcalc can be used to model a robot in a \acl{BAT}, where the robot's actions are modeled with preconditions and effects.
We have also seen how the \sitcalc can be extended with a notion of time, where each action occurs at a certain time, formulas such as $\ztime(\goto(l))$ refer to the time when an action occurs, and timing constraints can be used by assuming the standard interpretation for the real numbers and its operands.
However, as we will see later, this results in an undecidable logic, even if we restrict the domain to a finite number of objects.
The reason for this is that we can use these extensions to model more expressive variants of timed automata, e.g., timed automata that allow addition in clock constraints, which are undecidable~\cite{alurTheoryTimedAutomata1994,alurDecisionProblemsTimed2004}.
To avoid this problem, we instead propose the logic \tesg, which incorporates time, but separates situation formulas and clock formulas syntactically.
Similar to timed automata, the logic contains \emph{clocks} that allow the specification of restricted timing constraints.
As in timed automata, those timing constraints allow comparing clocks to rational constants but not to other clocks.
To combine clocks with actions, the logic adds clock constraints to actions, which describe the timing constraint of the action.
Additionally, each action may reset a subset of the program's clocks to zero.
This is very similar to how clocks are handled in timed automata.
It allows specifying timing constraints while avoiding undecidability, at least for finite domains.

In addition to incorporating time into the logic, \tesg also allows \emph{trace formulas} similar to \ac{MTL} formulas, which describe temporal properties of a program execution, e.g., the robot is not grasping any object in the next \SI{10}{\sec}:
\[
  \neg \finally{[0, 10]} \grasping
\]

In this chapter, we first describe \tesg{}, summarizing its syntax in \autoref{sec:tesg-syntax} and semantics including a transition semantics for \golog programs in \autoref{sec:tesg-semantics}.
We show in \autoref{sec:tesg-bat} how \acp{BAT} may be specified in \tesg, before we describe a variant of \emph{regression} that allows to reduce a query about the world after a sequence of actions to a query about the initial state in \autoref{sec:tesg-regression}.
In the remainder of this chapter, we analyze some properties of the logic.
First, as \tesg can be seen as a combination of the situation calculus variant \esg and the temporal logic \ac{MTL}, we compare \tesg with \esg in \autoref{sec:tesg-esg} and with \ac{MTL} in \autoref{sec:tesg-mtl}.
In \autoref{sec:tesg-ta}, we show that \aclp{TA} can be modeled in \tesg.
We close with some remarks on why we chose to model time with clocks and we demonstrate that more commonly used alternatives quickly result in undecidable verification and synthesis problems.


\section{Syntax}\label{sec:tesg-syntax}


The logic \tesg extends \esg~\cite{classenLogicNonterminatingGolog2008,classenPlanningVerificationAgent2013}, which is based on \es~\cite{lakemeyerSemanticCharacterizationUseful2011}, with clocks and timing constraints.
It therefore uses a possible-world semantics where situations are part of the semantics rather than appearing as terms in the language, as is the case in the \sitcalc.
As in \esg, we use the modal operator $\lbrack \cdot \rbrack$ to express what is true in a situation after executing some program, e.g., $\lbrack \delta \rbrack \alpha$ states that $\alpha$ is true after any successful execution of the program $\delta$.
Also similar to \es and \esg, the logic uses a sorted language.
The language of \tesg has four sorts: \emph{object}, \emph{action}, \emph{clock}, and \emph{time}.
Another feature inherited from \es{}~\cite{lakemeyerSemanticCharacterizationUseful2011} and $\mathcal{OL}$~\cite{levesqueLogicKnowledgeBases2001} is the use of countably infinite sets of \emph{standard names} for those sorts.
Standard names are treated like constants but additionally serve as \emph{unique identifiers} by asserting that each standard name is distinct from any other name.
They are also intended to be isomorphic with the set of all objects (or actions and clocks respectively) of the domain.
In other words, standard names can be thought of as constants that satisfy the unique name assumption and domain closure for objects.
One advantage of using standard names is that quantifiers can be understood substitutionally when defining the semantics.

To incorporate time into the language, \tesg extends the language with \emph{clock formulas}, which define constraints on clock values.
Similar to clock constraints in \acp{TA}, a clock constraint in \tesg allows to compare a clock value to a rational number and clocks may be reset to zero.
Other operators on clock values, e.g., arithmetic operators such as $+$ or $\cdot$, are not allowed.
Also, while each action will occur at a certain point in time, the time point is not explicitly specified in an action term, in contrast to similar approaches described in \autoref{sec:sitcalc}, but in line with how evolving time is treated in \acp{TA}.
Intuitively, the reason is as follows: while clock constraints allow to specify some constraint on the execution time point, the exact time point cannot be controlled but is determined by the environment.
For this reason, only clock constraints may refer to any notion of time, action terms and situation formulas may not.

Formally, the language is defined as follows:
\begin{definition}[Symbols of \tesg{}]
The symbols of the language are from the following vocabulary:
\begin{enumerate}
  \item variables of sort object $x_1, x_2, \ldots$, action $a, a_1, a_2, \ldots$, and clock $c, c_1, c_2, \ldots$, 
  \item standard names of sort object $\stdobjname = \{ o_1, o_2, \ldots \}$, action $\stdactname = \{ p_1, p_2, \ldots \}$, clock $\stdclockname = \{ q_1, q_2, \ldots \}$, and time $\stdtimename = \rationals = \{0, \frac{1}{2}, \frac{2}{3}, 1, \ldots\}$,
  \item fluent object function symbols of arity $k$: $f_1^k, f_2^k, \ldots$,
  \item rigid function symbols of arity $k$ for sorts object, action, and clock: $g_1^k, g_2^k, \ldots$,
  \item fluent predicate symbols of arity $k$: $\mathcal{F}^k = \{ F_1^k, F_2^k, \ldots \}$, e.g., $\mi{Holding(o)}$; we assume this list contains the distinguished predicates $\poss$ for action preconditions, $\reset$ for clock resets, and $\clockconstraint$ for clock constraints,
  \item rigid predicate symbols of arity $k$: $\mathcal{G}^k = \{ G_1^k, G_2^k, \ldots \}$,
  \item open, closed, and half-closed intervals, e.g., $[1, 2]$, with natural numbers as interval endpoints,
  \item connectives and other symbols: $<$ $\leq$, $=$, $\geq$, $>$, $\wedge$, $\neg$, $\forall$, $\square$, $\until{}$, round parentheses, single and double square brackets, period, and comma.
    \qedhere
\end{enumerate}
\end{definition}

We write $\mathcal{F}$ for the set of all fluent predicate symbols $\mathcal{F} \eqdef \bigcup_{k \in \mathbb{N}_0} \mathcal{F}^k$ and we denote all standard names as $\stdname \eqdef \stdobjname \cup \stdactname \cup \stdclockname$.
Furthermore, we assume that all action and clock function symbols are rigid.

Using the symbols defined above, we can define the terms of the language:
\begin{definition}[Terms of \tesg{}]
  The set of terms of \tesg{} is the least set such that
  \begin{enumerate}
    \item every variable is a term of the corresponding sort,
    \item every standard name is a term of the corresponding sort,
    \item if $t_1,\ldots,t_k$ are terms and $f$ is a $k$-ary function symbol, then $f(t_1,\ldots,t_k)$ is a term of the corresponding sort.
      \qedhere
  \end{enumerate}
\end{definition}

We call a function term \emph{primitive} if it is of the form $f(n_1,\ldots,n_k)$, with $n_i$ being standard names.
We denote the set of primitive terms as \primobjs (objects), \primactions (actions), and \primclocks (clocks)  and we denote the set of all primitive terms as $\primterms \eqdef \primobjs \cup \primactions \cup \primclocks$.
Furthermore, a term is is called \emph{rigid} if it only consists of rigid function symbols and standard names.

We continue by defining the formulas of the language:
\begin{definition}[Formulas]\label{def:tesg-formulas}
  The \emph{formulas of \tesg{}}, consisting of \emph{situation formulas}, \emph{clock formulas}, and \emph{trace formulas} are the least set such that
  \begin{enumerate}
    \item if $t_1,\ldots,t_k$ are terms and $P$ is a $k$-ary predicate symbol, then $P(t_1,\ldots,t_k)$ is a situation formula,
    \item if $t_1$ and $t_2$ are terms, then $(t_1 = t_2)$ is a situation formula,
    \item if $c$ is a clock term, $r, r' \in \stdtimename$, and $\operatorname{\bowtie} \in \{ <, \leq, =, \geq, > \}$, then $c \bowtie r$ and $r \bowtie r'$ are clock formulas, 
    \item if $\alpha$ and $\beta$ are situation formulas, $x$ is a variable, and $\delta$ is a program expression (defined below), $\phi$ is a trace formula, then $\alpha \wedge \beta$, $\neg \alpha$, $\forall x.\, \alpha$, $\square\alpha$, $[\delta]\alpha$, $\llbracket \delta \rrbracket \phi$, and $\llbracket \delta \rrbracket^{< \infty} \phi$ are situation formulas.
    \item if $\alpha$ is a clock formula, it is also a situation formula,
    \item if $\alpha$ is a situation formula, it is also a trace formula,
    \item if $\phi$ and $\psi$ are trace formulas, $x$ is a variable, and $I$ is an open, closed, or half-closed interval, then $\phi \wedge \psi$, $\neg \phi$, $\forall x.\,\phi$, 
      and $\phi \until{I} \psi$ are also trace formulas.
      \qedhere
  \end{enumerate}
\end{definition}

Note that we restrict the usage of clocks:
Clock formulas may only compare clock values to rational numbers and not to other clock values.%
\footnote{
  We do allow the formula $c_1 = c_2$ for clock terms $c_1$ and $c_2$.
  However, this formula compares the clock \emph{names} rather than their \emph{values}.
}
Also, we do not allow other arithmetic operators such as $+$ or $\cdot$ to be used on clock values.
This is similar to how clocks are handled in \aclp{TA}.

As usual, we define $\exists$ and $\vee$ as abbreviations, i.e., $\exists x\, \alpha \eqdef \neg \forall x\, \neg \alpha$ and $\alpha \vee \beta \eqdef \neg \left(\neg \alpha \wedge \neg \beta\right)$.
We also write $\top \eqdef \forall x(x = x)$ for the formula that is always true and $\bot \eqdef \neg \top$ or its negation.
For the temporal operators, we define $\finally{I} \phi \eqdef \top \until{I} \phi$, $\glob{I} \phi \eqdef \neg \finally{I} \neg \phi$, and $\tnext{I} \phi \eqdef \bot \until{I} \phi$.
A predicate symbol with standard names as arguments is called a \emph{primitive formula}, and we denote the set of primitive formulas as \primformulas.
If $F \in \fluents^0$ is a nullary fluent predicate, we may also omit the parentheses and write $F$ instead of $F()$.
We read $\square \alpha$ as ``$\alpha$ holds after executing any sequence of actions'', $[\delta] \alpha$ as ``$\alpha$ holds after the execution of program $\delta$'',
$\llbracket \delta \rrbracket \alpha$ as ``$\alpha$ holds during every execution of program $\delta$'',
$\llbracket \delta \rrbracket^{< \infty} \alpha$ as ``$\alpha$ holds during every \emph{terminating} execution of program $\delta$'',
and $c < r$ as ``the value of clock c is less than r'' (analogously for $\leq, =, \geq, >$).
We also use intervals to denote clock constraints, e.g., we write $c_1 \in (2, 3]$ for $c_1 > 2 \wedge c_1 \leq 3$ and $c_2 \in [1, \infty)$ for $c_2 \geq 1$.
Furthermore, we may omit the interval $I$ if $I$ is the unbounded interval $I = [0, \infty)$, e.g., $\phi \until{} \psi$ is short for $\phi \until{[0,\infty)} \psi$.

Free variables are implicitly understood to be quantified from the outside.
For a formula $\alpha$ with free variable $x$, we may also write $\alpha^x_t$ for the formula that results from replacing each occurrence of $x$ with $t$.
In order to reduce the number of parentheses, we assign a precedence to each connective, as shown in \autoref{tab:operator-precedence}.
\begin{table}[tbh]
  \centering
  \caption{Operator precedence in the logic \tesg.}
  \label{tab:operator-precedence}
  \begin{tabular}{*{15}{Cc}}
    \toprule
    Precedence & 1 & 2 & 3 & 4 & 5 & 6 & 7 & 8 & 9 & 10 & 11 & 12 & 13 & 14
    \\
    Operator & $\lbrack \cdot \rbrack$ & $\neg$ & $\tnext{}$ & $\globally{}$ & $\finally{}$ & $\until{}$ & $\wedge$ & $\vee$ & $\forall$ & $\exists$ & $\supset$ & $\equiv$ & $\llbracket \cdot \rrbracket$ & $\square$
    \\
    \bottomrule
  \end{tabular}
\end{table}
Lower precedence means that the operator binds tighter (as if by parentheses).
We demonstrate operator precedence with some examples:
\begin{alignat*}{3}
        &\square &[a] &\rat(l) & \equivspace & \exists l'\, a = \eac{\drive(l', l)}
        \\
        &&&&&\vee \rat(l) \wedge \neg \exists l''\, a = \sac{\drive(l, l'')}
  \intertext{This is the same as the following formula:}
        \forall a\Big[&\square\big\{ (&[a] &\rat(l)) & \equivspace & \big(\exists l'\, (a = \eac{\drive(l', l)}
        \\
                      &&&&&\vee (\rat(l) \wedge \neg \exists l''\, a = \sac{\drive(l, l')}))\big) \big\} \Big]
\end{alignat*}
As a second example, consider the following formula:
\begin{align*}
  \llbracket &\delta \rrbracket \finally{} \rat(l) \wedge \neg \glob{[0, 1]} \grasping \until{} \camon
  \intertext{This is the same as the following formula:}
  \llbracket &\delta \rrbracket \{ (\finally{} \rat(l)) \wedge ((\neg \glob{[0, 1]} \grasping) \until{} \camon)\}
\end{align*}

Sometimes, we may want to restrict formulas:
\begin{definition}[Static, Fluent, and Time-Invariant Formulas]
  We distinguish the following formulas:
  \begin{description}
    \item[Static Formulas] A formula $\alpha$ is called \emph{static} if it contains no $[\cdot]$, $\llbracket \cdot \rrbracket$, or $\square$ operators.
    \item[Time-Invariant Formulas] A formula $\alpha$ is called \emph{time-invariant}
      if it does contain any clock terms and does not mention the distinguished predicate symbol $\clockconstraint$.
    \item[Fluent Formulas] A formula $\alpha$ is called \emph{fluent} if it is static, time-invariant, and does not mention the distinguished predicate symbol $\poss$.
  \end{description}
  Furthermore, given a pair $(\fluents, \clockset)$ of fluents $\fluents$ and clocks $\clockset$, a \emph{formula over $(\fluents, \clockset)$} is a formula that only mentions fluents from \fluents and clocks from \clockset.
\end{definition}



We are now ready to define the syntax of \golog{} program expressions referred to by the operators $[\delta]$ and $\llbracket \delta \rrbracket$:%
\footnote{Note that although the definitions of formulas (\autoref{def:tesg-formulas}) and programs (\autoref{def:tesg-programs}) mutually depend on each other, they are still well-defined:
  Programs only allow static situation formulas and static situation formulas may not refer to programs.
  Technically, we would first need to define static situation formulas, then programs, and then all formulas.
For the sake of presentation, we omit this separation.}

\begin{definition}[Program Expressions]\label{def:tesg-programs}
  \[
    \delta ::= t \smid \alpha? \smid \delta_1 ; \delta_2 \smid \delta_1 \vert \delta_2
    \smid \delta_1 \| \delta_2 \smid \delta^*
  \]
  where $t$ is an action term and $\alpha$ is a static situation formula. A
  program expression consists of actions $t$, tests $\alpha?$, sequences
  $\delta_1;\delta_2$, nondeterministic
  branching $\delta_1 \vert \delta_2$,
  interleaved concurrency
  $\delta_1 \| \delta_2$, and nondeterministic iteration $\delta^*$.%
  \footnote{
    We leave out the pick operator $\pi x.\, \delta$, as we later restrict the domain to be finite, where pick can be expressed with nondeterministic branching.
  }
\end{definition}

We also use the abbreviation $\nil \eqdef \top?$ for the empty program that always succeeds.
Moreover, we define conditionals and loops as macros:
\begin{align*}
  \gif \alpha \gthen \delta_1 \gelse \delta_2 \gfi &\eqdef (\alpha?; \delta_1) \vert (\neg \alpha?; \delta_2)
  \\
  \gwhile \alpha \gdo \delta \gdone &\eqdef (\alpha?; \delta)^*; \neg \alpha?
\end{align*}

We remark that the above program constructs are a proper subset of the original \congolog~\cite{degiacomoConGologConcurrentProgramming2000}.
We have left out other constructs such as prioritized concurrency for simplicity.

\section{Semantics}\label{sec:tesg-semantics}


We continue with the semantics of \tesg.
Similar to \es and \esg, the semantics of \tesg are based on a possible-world semantics where situations do not occur in the language but are instead part of the semantics.
In particular, in \tesg, a timed trace, which is a sequence of action-time pairs, specifies the actions and their time points that have occurred.
A world then specifies which primitive formulas are true, not only initially, but after any (finite) timed trace.

We start with the definition of timed traces, which are similar to timed words in \ac{MTL} (see \autoref{def:timed-words}):
\begin{definition}[Timed Traces]\label{def:timed-traces}
A \emph{timed trace} is a finite or infinite sequence of alternating time points and action standard names:
\[
  z = t_1 \cdot p_1 \cdot t_2 \cdot p_2 \cdot \ldots
\]
where $p_i \in \stdactname$ and $t_i \in \realpos$ such that
the sequence $\left(t_i\right)_i$ is monotonically non-decreasing.
We also write $\lvert z \rvert$ for the length of $z$.
We call a trace \emph{rational} if it only contains rational time points $t_i \in \rationals$.
We denote the set of finite timed traces with \traces, the set of infinite traces with \inftraces, and the set of all traces with \alltraces.
\end{definition}


As we are often only interested in traces starting with a time step and ending with an action step, we also write $(p_1, t_1)(p_2, t_2)\cdots(p_k, t_k)$ for the timed trace $t_1 \cdot p_1 \cdot t_2 \cdot p_2 \cdot \ldots \cdot t_k \cdot p_k$ that starts with the time step $t_1$ and ends with the action step $p_k$.
For a (finite or infinite) trace $\tau = (p_1, t_1) (p_2, t_2) \cdots$, we write $\tau^{(i)} = (p_1, t_1) \cdots (p_i, t_i)$ for the finite prefix of $\tau$ that contains the first $i$ time-symbol pairs.
For a finite timed trace $z = \left(p_1,t_1\right) \cdots \left(p_k,t_k\right)$, we define the time point of the last action in $z$ as $\ztime(z) \eqdef t_k$ if $z \neq \la\ra$ and $\ztime(\la\ra) \eqdef 0$ otherwise.

In comparison to a timed word (\autoref{def:timed-words}), a timed trace does not contain the atomic propositions that are true at some point, but instead the action that occurs at each time point.
Therefore, in contrast to a timed word, it does not directly express which state properties are true at a certain point in time. 
To relate an action sequence to a certain state, we continue with the definition of \emph{worlds}.
Intuitively, a world $w$ determines the truth of fluent predicates and functions, not just initially, but after any (timed) sequence of actions:
\begin{definition}[World]\label{def:tesg-world}
  A world $w$ is a mapping that maps
   \begin{enumerate}
     \item each primitive object term to a co-referring object standard name after every possible trace, i.e.,
       \[
         \primobjs \times \traces \rightarrow \stdobjname
       \]
     \item each primitive action term to a co-referring action standard name after every possible trace, i.e.,
       \[
         \primactions \times \traces \rightarrow \stdactname
       \]
     \item each primitive clock term to a co-referring clock standard name after every possible trace, i.e.,
       \[
         \primclocks \times \traces \rightarrow \stdclockname
       \]
     \item each primitive formula to a truth value $0$ or $1$ after every possible trace, i.e.,
       \[
         \primformulas \times \fintraces \rightarrow \{ 0, 1 \}
       \]
     \item each clock standard name to a clock value from the reals, i.e.,
       \[
         \stdclockname \times \traces \rightarrow \realpos
       \]
   \end{enumerate}
   satisfying the following constraints:
   \begin{description}
     \item[Rigidity:]
       If $R$ is a rigid function or predicate symbol, then for all $z$ and $z'$ in \traces:
       \[
         w[R(n_1,\ldots,n_k),z]=w[R(n_1,\ldots,n_k),z']
       \]
     \item[Unique names for actions and clocks:]
       If $g(n_1, \ldots, n_k)$ and $g'(n_1', \ldots, n_l')$ are two distinct primitive action terms or primitive clock terms, then for all $z$ and $z'$ in \traces:
       \[
         w[ g(n_1, \ldots, n_k), z ] \neq w[ g'(n_1', \ldots, n_l'), z' ]
       \]
     \item[Clock initialization:]
       All clock values are initialized to $0$, i.e., for every clock standard name $c \in \stdclockname$:
       \[
         w[c, \la\ra] = 0
       \]
     \item[Time progression:]
       The clock values increase according to the time increments  determined by a trace, i.e., for every clock $c \in \stdclockname$, $z = t_0 p_0 \cdots t_k p_k \in \traces$, and every time step $t_{k+1} \in \realpos$:
       \[
         w[c, z \cdot t_{k+1}] = w[c, z] + t_{k+1} - \ztime(z)
       \]
       Also, a clock value may not be changed by an action, unless the action resets the clock.
       Hence, for every clock $c \in \stdclockname$, $z = t_0 p_0 \cdots t_k \in \traces$, and every action step $p_k$:
       \[
         w[c, z \cdot p_k] =
         \begin{cases}
           0 & \text{ if } w[\reset(c), z \cdot p_k] = 1
           \\
           w[c, z] & \text{ else }
         \end{cases}
       \]
   \end{description}
  The set of all worlds is denoted by $\worlds$.
\end{definition}

A world maps each primitive term to some co-referring standard name of the corresponding sort.
Additionally, it defines for each primitive formula $\alpha$ whether the formula is true after any trace $z$ by mapping the pair $(\alpha, z)$ to a truth value, where $0$ stands for false and $1$ for true.

We continue with term denotation, which extends co-referring standard names from primitive terms to arbitrary terms:
\begin{definition}[Denotation of terms]\label{def:tesg-denotation}
  Given a ground term $t$, a world $w$, and a trace $z \in \mathcal{Z}$, we define $\denot{t}{z}{w}$ by:
  \begin{enumerate}
    \item if $t \in \mathcal{N}$, then $\denot{t}{z}{w} = t$,
    \item if $t = f(t_1,\ldots,t_k)$ then $\denot{t}{z}{w} = w[f(n_1,\ldots,n_k), z]$, where $n_i = \denot{t_i}{z}{w}$
      \qedhere
  \end{enumerate}
\end{definition}

We can now define the transitions that a program may take in a given world $w$.
The program transition semantics is similar to the transition semantics in \congolog~(\autoref{sec:sitcalc}) and \esg~\cite{classenLogicNonterminatingGolog2008,classenPlanningVerificationAgent2013}.
It extends the transition semantics of \esg with time and clocks.
Here, a program configuration is a tuple $(z, \delta)$ consisting of a timed trace $z$ and the remaining program $\delta$.
A program may take a transition $(z, \delta) \warrow (z', \delta')$ if it can take a single action that results in the new configuration.
In some places, the transition semantics refers to the truth of clock and situation formulas (see Definition~\ref{def:tesg-truth} below).%
\footnote{Similar to above, although they mutually depend on each other, the semantics is well-defined, as the transition semantics only refers to static and clock formulas which in turn may not contain programs.}%
\begin{definition}[Program Transition Semantics]\label{def:tesg-trans}
  Program transitions consist of a time step and an action step, each defined as the least set satisfying the following conditions:
  \begin{description}
    \item[Time step:] For each $d \in \realpos$, a time step that increments the time by $d$:
      \[
        \la z, \delta \ra \wtarrow \la z \cdot t, \delta \ra \text{ where } t = \ztime(z) + d
      \]
      A time step increments the time by the increment $d$.
      Other than that, no changes occur. In particular, no action occurs.
    \item[Action step:]
      ~
      \begin{enumerate}
        \item \label{def:tesg-trans:action}
          \[
            \la z, a \ra \wsarrow \la z', \nil \ra
          \]
          if $z' = z \cdot p$ with $p = \denot{a}{z}{w}$. 
          Intuitively, the program may take a single transition step with action $p$ whenever $p$ is the co-referring standard name of the primitive program $a$.
          In the resulting configuration, the program trace is appended with the new action $p$ and the remaining program is the empty program \nil.\footnote{
            Note that following \parencite{classenPlanningVerificationAgent2013}, we do not require the precondition of $a$ to be satisfied.
            However, such a check can easily be done for a program $\delta$ by replacing each occurrence of a primitive action $a$ in $\delta$ by $\poss(a)?; a$.
            We refer to \parencite{classenPlanningVerificationAgent2013} for a more detailed discussion of this augmentation.
          }
        \item
          \[
            \la z,\delta_1;\delta_2 \ra \wsarrow \la z \cdot p, \gamma;\delta_2 \ra \text{ if } \la z, \delta_1 \ra \wsarrow \la z \cdot p, \gamma \ra
          \]
          For a sequence $\delta_1; \delta_2$ of sub-programs $\delta_1$ and $\delta_2$, if there is a possible transition of $\delta_1$ to the remaining program $\gamma$, then the resulting configuration is the same as the resulting configuration of making the transition in the sub-program $\delta_1$, but where the remaining program $\gamma$ is concatenated with the (unchanged) sub-program $\delta_2$.
        \item
          \[
            \la z,\delta_1;\delta_2 \ra \wsarrow \la z \cdot p, \delta' \ra \text{ if }\la z, \delta_1 \ra \in \mathcal{F}^w \text{ and } \la z, \delta_2 \ra \wsarrow \la z \cdot p, \delta' \ra
          \]
          For a sequence $\delta_1; \delta_2$ of sub-programs $\delta_1$ and $\delta_2$, if $\delta_1$ is final in the current configuration, then the resulting configurations are the same as the configurations resulting from following the transitions of the second sub-program $\delta_2$.
        \item
          \[
            \la z,\delta_1 \vert \delta_2 \ra \wsarrow \la z \cdot p, \delta' \ra \text{ if } \la z, \delta_1 \ra \wsarrow \la z \cdot p, \delta' \ra \text{ or }\la z, \delta_2 \ra \wsarrow \la z \cdot p, \delta' \ra
          \]
          For non-deterministic branching $\delta_1 \vert \delta_2$ of the two sub-programs $\delta_1$ and $\delta_2$, we may follow the transitions of the first or the second sub-program.
        \item
          \[
            \la z,\delta^* \ra \wsarrow \la z \cdot p, \gamma; \delta^* \ra \text{ if } \la z,\delta \ra \wsarrow \la z \cdot p, \gamma \ra
          \]
          For non-deterministic iteration $\delta^*$ of the sub-program $\delta$, the resulting configuration of doing a single step is the same as following a single step of $\delta$, but with the resulting program concatenated with $\delta^*$ to allow further iterations later on.
        \item
          \begin{align*}
            \la z,\delta_1 \| \delta_2 \ra \wsarrow \la z \cdot p, \delta' \| \delta_2 \ra &\text{ if } \la z, \delta_1 \ra \wsarrow \la z \cdot p, \delta' \ra
            \\
            \la z,\delta_1 \| \delta_2 \ra \wsarrow \la z \cdot p, \delta_1 \| \delta' \ra &\text{ if } \la z, \delta_2\ra \wsarrow \la z \cdot p, \delta' \ra
          \end{align*}
          For interleaved concurrency $\delta_1 \| \delta_2$, we may follow the transition steps of $\delta_1$ or $\delta_2$ similarly to non-deterministic choice, but where the remaining program consists of the remaining program of the program that we followed, concurrently executed with the unchanged other sub-program.
      \end{enumerate}
  \end{description}
  The transition relation $\warrow$ between configurations is then defined as the combination of a time and an action step,
  where
  \[
    \la z, \delta \ra \warrow \la z', \delta' \ra
  \]
  iff there exists $d \in \realpos, p \in \stdactname$ and $z^*, \delta^*$ such that
  \[
    \la z, \delta \ra \wtarrow \la z^*, \delta^* \ra \wsarrow \la z', \delta' \ra
  \]
  We also write $\warrow^*$ for the reflexive and transitive closure of $\warrow$.

  The set of final configurations \final{} is the smallest set that satisfies the following conditions:
  \begin{enumerate}
    \item
      The program $\alpha?$ is final if $\alpha$ is true in the current situation:
      \[
        \la z,\alpha? \ra \in \final \text{ if } w, z \models \alpha
      \]
    \item
      The sequence $\delta_1; \delta_2$ is final if both sub-programs $\delta_1$ and $\delta_2$ are final:
      \[
        \la z,\delta_1;\delta_2 \ra \in \final \text{ if } \la z,\delta_1 \ra \in \final \text{ and } \la z,\delta_2 \ra \in \final
      \]
    \item
      The non-deterministic branching $\delta_1 \vert \delta_2$ is final if $\delta_1$ or $\delta_2$ is final:
      \[
        \la z,\delta_1 \vert \delta_2 \ra \in \final \text{ if } \la z,\delta_1 \ra \in \final \text{ or } \la z,\delta_2 \ra \in \final
      \]
    \item
      Non-deterministic iteration $\delta^*$ is always final, as we may choose to stop iterating:
      \[
        \la z,\delta^* \ra \in \final
      \]
    \item
      Interleaved concurrency $\delta_1 \| \delta_2$ is final if both sub-programs $\delta_1$ and $\delta_2$ are final:
      \[
        \la z,\delta_1 \| \delta_2 \ra \in \final \text{ if } \la z,\delta_1 \ra \in \final \text{ and } \la z,\delta_2 \ra \in \final
        \qedhere
      \]
  \end{enumerate}
\end{definition}

As every action step takes exactly one action and tests $\alpha?$ do not result in transitions but instead are checked in the final configurations, tests in \tesg correspond to \emph{synchronized conditionals} in the \congolog transition semantics as described in \autoref{sec:golog}.

By following the transitions, we obtain \emph{program traces}:
\begin{definition}[Program Traces]\label{def:program-trace}
  Given a world $w$ and a finite trace $z$, the \emph{program traces} of a program expression $\delta$ starting in $z$ are defined as follows:
  \begin{align*}
    \| \delta \|^z_w = &\{ z' \in \traces \mid \la z, \delta \ra \warrow^* \la z \cdot z', \delta' \ra \text{ and } \la z \cdot z',  \delta' \in \final \} \cup
    \\
                       &\{ \pi \in \inftraces \mid \la  z, \delta \ra \warrow \la z \cdot \pi^{(i)}, \delta_1 \ra \warrow \la z \cdot \pi^{(2)}, \delta_2 \ra \warrow \ldots
                    \\ &\quad \text{ where for all $i$, }  \la z \cdot \pi^{(i)}, \delta_i \ra \not\in \final \}
                    \qedhere
  \end{align*}
\end{definition}
Intuitively, the program traces of program $\delta$ are those finite traces that end in a final configuration, as well as those infinite traces that never visit a final configuration.
We also omit $z$ if $z = \la\ra$, i.e., $\|\delta\|_w$ denotes $\|\delta\|^{\la\ra}_w$.



Using the program transition semantics, we can now define the truth of a formula:
\begin{definition}[Truth of Formulas]\label{def:tesg-truth}
  Given a world $w \in \mathcal{W}$ and a formula $\alpha$, we define $w \models \alpha$ as $w,\langle\rangle \models \alpha$ and where $w, z \models \alpha$ is defined as follows for every $z \in \mathcal{Z}$:
  \begin{enumerate}
    \item $w, z \models F(t_1,\ldots,t_k)$ iff $w[F(n_1,\ldots,n_k),z] = 1$, where $n_i=\denot{t_i}{z}{w}$,
    \item $w, z \models (t_1 = t_2)$ iff $n_1$ and $n_2$ are identical, where $n_i=\denot{t_i}{z}{w}$,
    \item $w, z \models r \bowtie r'$ iff $r \bowtie r'$ and $r, r' \in \stdtimename$, 
    \item $w, z \models c \bowtie r$ iff $w[c, z] \bowtie r$ and $c \in \stdclockname$, $r \in \stdtimename$, 
    \item $w, z \models \alpha \wedge \beta$ iff $w, z \models \alpha$ and
      $w, z \models \beta$,
    \item $w, z \models \neg \alpha$ iff $w, z \not\models \alpha$,
    \item $w, z \models \forall x.\, \alpha$ iff $w, z \models \alpha^x_n$ for
      every standard name of the right sort,
    \item $w, z \models \square \alpha$ iff $w, z\cdot z' \models \alpha$ for all $z' \in \mathcal{Z}$,
    \item $w, z \models [\delta]\alpha$ iff $w, z \cdot z' \models \alpha$ for all $z' \in \|\delta\|^z_w$,
    \item $w, z \models \llbracket \delta \rrbracket \phi$ iff for all $\tau \in \|\delta\|^z_w$, $w, z, \tau \models \phi$,
    \item $w, z \models \llbracket \delta \rrbracket^{< \infty} \phi$ iff for all finite $z' \in \|\delta\|^z_w$, $w, z, z' \models \phi$.
  \end{enumerate}

  The truth of trace formulas $\phi$ is defined as follows for $w \in \mathcal{W}$, $z \in \fintraces, \tau \in \alltraces$:
  \begin{enumerate}
    \item $w, z, \tau \models \alpha$ iff $w, z \models \alpha$ and $\alpha$ is
      a situation formula;
    \item $w, z, \tau \models \phi \wedge \psi$ iff $w, z, \tau \models \phi$
      and $w, z, \tau \models \psi$;
    \item $w, z, \tau \models \neg \phi$ iff $w, z, \tau \not\models \phi$;
    \item $w, z, \tau \models \forall x.\, \phi$ iff
      $w, z, \tau \models \phi^x_n$ for all $n \in \mathcal{N}_x$;
    \item $w, z, \tau \models \phi \until{I} \psi$ iff there is a $\tau' \in \alltraces$ and $z_1 \in \traces$ with $z_1 = (t_1, p_1) \cdots (t_k, p_k) \neq \la\ra$ such that
      \begin{enumerate}
        \item $\tau = z_1 \cdot \tau'$,
        \item $w, z \cdot z_1, \tau' \models \psi$,
        \item $\ztime(z_1) \in \ztime(z) + I$,
        \item for all $z_2 = (t_i, p_i) \cdots (t_j, p_j)$ with $z_1 = z_2 \cdot z_3$, $z_2 \neq \la\ra$, and $z_3 \neq \la\ra$:
          \\ $w, z \cdot z_2, z_3 \cdot \tau' \models \phi$.
          \qedhere
      \end{enumerate}
  \end{enumerate}
\end{definition}



A situation formula $\alpha$ is also called \emph{satisfiable} if there is some world $w$ such that $w \models \alpha$.
For a set of sentences $\Sigma$ and a situation formula $\alpha$, we also say $\Sigma$ entails $\alpha$, written $\Sigma \models \alpha$, if for every $w$ with $w \models \beta$ for every $\beta \in \Sigma$, it follows that $w \models \alpha$.
We say that $\alpha$ is \emph{valid}, denoted with $\models \alpha$, if $\{ \} \models \alpha$.
Similarly, for a trace formula $\phi$, we say that $\phi$ is satisfiable if there is some world and some trace $\tau \in \traces$ such that $w, \la\ra, \tau \models \phi$.
Finally, $\phi$ is valid, denoted with $\models \phi$, if $w, \la\ra, \tau \models \phi$ for every $w \in \worlds$ and $\tau \in \alltraces$.

Note that we make an important restriction for evaluating trace formulas $\phi \until{I} \psi$: We only consider traces $z_1$ that end with an action, i.e., we do not evaluate the trace $z_1$ after a time increment $t_k$ but before action $p_k$.
As an example, consider the trace $z = \la 1, p \ra$ and the world $w$ with $w[F, \la\ra] = 0$ and $w[F, z] = 1$.
The world satisfies $w \models \glob{[1,1]} F$ even though $w[F, \la 1 \ra ] = 0$.
Hence, we only observe the world when the agent does some action and we cannot express any properties about the states in between.
This is related to the difference of point-based and continuous semantics of \ac{MTL}~\parencite{dsouzaExpressivenessMTLPointwise2007}, where observations are also restricted to time points at which an event occurs.
In particular, restricting the evaluation to action occurrences will be important to show the relationship between \ac{MTL} and \tesg in \autoref{sec:tesg-mtl}.



\section{Basic Action Theories}\label{sec:tesg-bat}

Equipped with the logic \tesg that allows to express temporal constraints, we continue by describing how we can model specific application domains in the logic.
As usual, this is done in a \acf{BAT}, which needs to specify the following properties of the domain:
\begin{enumerate}
  \item The initial state of the world;
  \item The preconditions of the actions that the agent may take;
  \item The effects of the agent's actions, i.e., what the world looks like after taking some action.
\end{enumerate}

We follow the usual solution to the qualification problem that the action precondition describes all the necessary and sufficient conditions for an action to be possible.
We also follow the causal completeness assumption by \textcite{reiterFrameProblemSituation1991} to solve the frame problem by assuming that there are no additional actions that have an effect on fluent values other than those described in the \ac{BAT}.
Using these assumptions, we obtain the following definition for a \acl{BAT}:
\begin{definition}[Basic Action Theory]\label{def:bat}
  Given a finite set of fluents $\mathcal{F}$ and a finite set of clocks $\clockset$, a set $\bat \subseteq \tesg$ of sentences is called a \acf{BAT} over $(\mathcal{F},\clockset)$ iff $\bat = \bat_0 \cup \bat_\pre \cup \bat_g \cup \bat_\post$, where $\bat$ mentions only fluents in $\mathcal{F}$, clocks in $\clockset$, and
  \begin{enumerate}
    \item $\bat_0$ is any set of fluent sentences describing the initial situation,
    \item $\bat_\pre$ consists of a single sentence of the form $\square \poss(a) \equiv \pi_a$, where $\pi_a$ is a fluent situation formula with free variable $a$ that specifies the precondition of all actions,
    \item $\bat_g$ consists of a single sentence of the form $\square \clockconstraint(a) \equivspace g_a$ describing the clock constraints for action $a$, where $g_a$ is a static situation formula that does not mention $\poss$ and which may only contain numbers from \naturals,\footnote{If we want to use rationals for clock constraints in \bat, we can multiply all occurring numbers in \bat by the largest common divisor, thereby scaling them to natural numbers.} and
    \item $\bat_\post$ is a set of sentences describing the effects of actions:
      \begin{itemize}
        \item one for each fluent predicate $F \in \mathcal{F}$ (including the distinguished predicate $\reset$), of the form $\square[a]F(\vec{x}) \equivspace \gamma_F(\vec{x})$, where $\gamma_F(\vec{x})$ is a fluent situation formula with free variables among $a$ and $\vec{x}$,
        \item one for each functional fluent $f \in \mathcal{F}$ of the form $\square [a] f(\vec{x}) = y \equivspace \gamma_f(\vec{x}, y)$, where $\gamma_f(\vec{x}, y)$ is a fluent situation formula with free variables among $a$, $\vec{x}$, and $y$.
      \end{itemize}
      Each such sentence is also called the \acfi{SSA} for $F$ (respectively $f$).
      \qedhere
  \end{enumerate}
\end{definition}

Apart from dealing with the qualification problem and the frame problem, we also need to consider clock constraints and clock resets in our \ac{BAT}.
For doing so, we extended the definition of a \ac{BAT} in \es~(see \autoref{sec:es}) in two ways:
\begin{enumerate}
  \item For each action $a$, the \ac{BAT} contains a clock formula $g_a$ that describes the clock constraints of the action, similar to how the precondition axiom defines the precondition of the action.
  \item Additionally, the \ac{BAT} contains a sentence that describes the conditions for resetting a clock to zero after doing some action.
\end{enumerate}

A \ac{BAT} defines a (possibly infinite) set of actions that the robot may perform, which we denote with \batactions.

We can now define \emph{programs}:

\begin{definition}[Program]
  A program is a pair $\Delta = (\bat, \delta)$ consisting of a \ac{BAT} \bat over $(\fluents, \clockset)$ and a program expression $\delta$ that only mentions fluents from $\fluents$ and clocks from $\clockset$.
\end{definition}

We will later refer to the reachable subprograms of some program $\delta$:
\begin{definition}[Reachable Subprograms]
  Given a program $(\bat, \delta)$, 
  we define the \emph{reachable subprograms $\sub(\delta)$ of $\delta$}: 
\[
  \sub(\delta) = \{ \delta' \smid \exists w \models \bat, z \in \traces \text{ such that } \la\la\ra, \delta\ra \warrow^* \la z, \delta' \ra\}
  \qedhere
\]
\end{definition}

\paragraph{Durative Actions}
Often, we want to model actions that have a certain \emph{duration}, e.g., a grasp action that takes \SI{15}{\sec}.
We follow the usual approach~\parencite{pintoReasoningTimeSituation1995} to model these with \emph{start} and \emph{end} actions, e.g., a durative \grasp may be modeled with the two primitive actions $\sac{\grasp}$ and $\eac{\grasp}$.
As our logic can measure time with clocks, we may use clocks to constrain the duration of an action, e.g., we may add the clock constraint $g(\eac{\grasp}) \equiv c_\mi{grasp} = 15$ to our \ac{BAT}, which requires that the clock value of the clock $c_\mi{grasp}$ has the value $15$ for the action $\eac{\grasp}$ to be possible.
If the clock is reset with each $\sac{\grasp}$, this clock constraint encodes that the action always takes exactly \SI{15}{\sec}.
The following \ac{BAT} demonstrates a more complete example of durative actions.

\begin{example} \label{ex:bat}
  The following \ac{BAT} describes a simple robot that is able to drive from location to location and that can grasp objects that are placed on machines.
  It is also equipped with a camera that it can turn on and off.
  For the sake of simplicity, the robot cannot put down objects.
  The \ac{BAT} \bat consists of the following axioms:
  \begin{description}
    \item[Initial situation:]
      \begin{align*}
        \bat_0 = \{ 
                    &\rat(l) \equiv \left(l = m_1\right),
                 \\ &\oat(o,l) \equiv \left(o = o_1\right) \wedge \left(l = m_2\right),
                 \\ &\neg \holding(\mi{obj}),
                 \\ &\neg \performing{a},
                 \\ & \neg \camon,
                 \\ &c(\drive(l, m_1)) = q_1 \wedge c(\drive(l, m_2)) = q_2,
                 \\ &c(\drive(l, m_3)) = q_3 \wedge c(\grasp(m, o)) = q_4,
                 \\ &c(\bootcam) = q_5 \wedge c(\stopcam) = q_6
               \}
      \end{align*}
      Here, $m_1, m_2, m_3, o_1$ are object standard names and $q_1, \ldots, q_6$ are clock standard names.
      Initially, the robot's (only) location is the machine $m_1$.
      There is a single object $o_1$, which is at the machine $m_2$.
      The robot is currently not holding anything, it is not performing any action, and its camera is turned off.
      For convenience, we also use a unary function $c$ of sort clock that assigns a clock to each action.
      This way, we can use more descriptive terms for clocks, e.g., $c(\drive(m_1, m_2))$ in place for $q_1$.
      Somewhat arbitrarily, we only distinguish the action clocks of the \drive action by the target location, i.e., for each possible goal location we only track the time for the last action that went to that location.
      Similarly for \grasp, we only use a single clock, independent of the action's arguments.
    \item[Precondition axiom:]
      \begin{alignat*}{3}
        \square \poss(a) &\equivspace  & \exists l_1, l_2 .\, a &= \sac{\drive(l_1, l_2)} \wedge \rat(l_1)
                      \\ && \vee \exists l, p .\, a &= \sac{\grasp(l, p)} \wedge \rat(l) \wedge \oat(p, l)
                      \\ && \vee a &= \sac{\bootcam} \wedge \neg \camon
                      \\ && \vee a &= \sac{\stopcam} \wedge \camon
                      \\ && \vee \exists a'.\, a &= \eac{a'} \wedge \performing{a'}
      \end{alignat*}
      The robot has four available durative actions: \drive, \grasp, \bootcam, and \stopcam.
      It can start driving from location $l_1$ to location $l_2$ if it is currently at location $l_1$.
      Also, it can start grasping an object if it is at the same location as the object.
      Furthermore, it can start booting its camera if the camera is currently turned off and it can stop the camera if it is currently turned on.
      Finally, to end any durative action, it needs to be performing the action.
    \item[Clock constraint axiom:]
      \begin{alignat*}{4}
        \square \clockconstraint(a) & \equivspace & \exists p.\, a &= \sac{p}
                                 \\ && \vee\; \exists l_1, l_2.\, a &= \eac{\drive(l_1, l_2)} &\wedge& &c(\drive(l_1, l_2)) &\in [1, 2]
                                 \\ && \vee\; \exists l, o.\, a &= \eac{\grasp(l, o)} &\wedge& &c(\grasp(l, o)) &= 2
                                 \\ && \vee\; a &= \eac{\bootcam} &\wedge& &c(\bootcam) &= 1
                                 \\ && \vee\; a &= \eac{\stopcam} &\wedge& &c(\stopcam) &= 0
      \end{alignat*}
      There are no clock constraints on any of the start actions.
      For the end actions, the clock constraints restrict the duration of the action: For a \drive action, the corresponding clock value must be in the interval $[1, 2]$, i.e., any \drive action takes between \SI{1}{\sec} and \SI{2}{\sec}.
      Similarly, a clock constraint for the \emph{end} action of \grasp restricts the action to take exactly \SI{2}{\sec}.
      Finally, booting the camera also always takes exactly \SI{1}{\sec}, while stopping the camera is instantaneous, i.e., it takes \SI{0}{\sec}.
    \item[Successor state axioms:]
      \begin{alignat*}{3}
        &\square [a] &&\rat(l) & \equivspace & \exists l'.\, a = \eac{\drive(l', l)}
        \\
        &&&&&\vee \rat(l) \wedge \neg \exists l'.\, a = \sac{\drive(l, l')}
        \\
        &\square [a] &&\objat(o, l) & \equivspace & \objat(o, l) \wedge \neg a = \sac{\grasp(l, o)}
        \\
        &\square [a] &&\holding(o) & \equivspace & \exists l\, a = \eac{\grasp(l, o)} \vee \holding(o)
        \\
        &\square [a] &&\camon & \equivspace & a = \eac{\bootcam} \vee \camon \wedge a \neq \sac{\stopcam}
        \\
        &\square [a] &&\performing{p} & \equivspace & a = \sac{p} \vee \performing{p} \wedge a \neq \eac{p}
      \end{alignat*}
      The first \ac{SSA} states that the robot is at location $l$ after ending a \drive action to $l$, or if it was at $l$ before and does not start driving anywhere else.
      Note that this also means that the robot does not have any position while it is driving.
      The next \ac{SSA} specifies the object location and states that the object $o$ is at location $l$ if it was there before and the robot does not start grasping the object.
      Note that as we do not have a \action{put} action that puts down the object somewhere, once the robot has picked up an object, the object cannot be at any location later on.
      Therefore, after the robot has grasped an object, it will always be holding the object.
      Similarly, the camera is on if the robot ends booting the camera and it will never be off again.
      Finally, the robot is performing any durative action if it starts doing the action, or if it has been performing the action before and does not end it.
    \item[Clock resets:]
      \begin{align*}
        \square [a] \reset(c) \equivspace \exists p.\, a = \sac{p} \wedge c = c(p)
      \end{align*}
      A clock $c$ is reset if the robot starts some action and $c$ is the clock assigned to the action.
      \qedhere
  \end{description}
\end{example}

\section{Regression}\label{sec:tesg-regression}

One of the fundamental reasoning tasks of a knowledge-based agent is to determine whether some formula $\alpha$ holds after executing a sequence of ground actions, given a \ac{BAT} \bat, i.e., to decide whether the following holds:
\[
  \bat \models [g_1(\vec{t}_1);g_2(\vec{t}_2); \ldots; g_n(\vec{t}_n)] \alpha
\]
More generally, for a given \golog program $(\bat, \delta)$, we may want to know whether $\alpha$ holds after executing the program:
\[
  \bat \models [\delta] \alpha
\]
This reasoning task is called \emph{projection}.
One particular way to do projection is \emph{regression}.
We explain the idea of regression with a simple example.
Let us consider the following query:
\[
  [\drive(\hallway, \kitchen)]\objat(\cupobj, \kitchen)
\]
That is, we want to know whether the cup is in the kitchen after executing the action $\drive(\hallway, \kitchen)$ (for the sake of this example, we assume that the action \drive is non-durative).
Now, assume that the \ac{BAT} \bat contains the following successor state axiom:
\begin{align*}
  \square [a] \objat(o, l) \equivspace &\exists l'.\, a = \drive(l', l) \wedge \holding(o)
  \\
  &\quad \vee \objat(o, l) \wedge (\neg \holding(o) \vee \neg \exists l',l''.\, a = \drive(l', l''))
\end{align*}
This successor state axiom states that the object $o$ is at location $l$ if the robot drove to $l$ with the last action while holding $o$, or if $o$ was at $l$ before and the robot did not drive it anywhere else.
To answer our query, we can substitute $a$ by $\drive(\hallway, \kitchen)$, $o$ by $\cupobj$, and $l$ by $\kitchen$ so we obtain:
\begin{align*}
  \square &[\drive(\hallway, \kitchen)] \objat(\cupobj, \kitchen) \equivspace
  \\ &\exists l'.\, \drive(\hallway, \kitchen) = \drive(l', \kitchen) \wedge \holding(o)
  \\
  &\quad \vee \objat(\cupobj, \kitchen) \wedge (\neg \holding(\cupobj) \vee \neg \exists l',l''.\, a = \drive(l', l''))
\end{align*}
We can then substitute the left-hand side $[\drive(\hallway, \kitchen)] \objat(\cupobj, \kitchen)$ of the equivalence by the right-hand side in the original query.
We obtain:
\begin{align*}
  &\exists l'.\, \drive(\hallway, \kitchen) = \drive(l', \kitchen) \wedge \holding(o)
  \\
  &\quad \vee \objat(\cupobj, \kitchen) \wedge (\neg \holding(\cupobj) \vee \neg \exists l',l''.\, a = \drive(l', l''))
\end{align*}

After some simplification, this is equivalent to:

\begin{equation}
  \holding(o) \vee \objat(\cupobj, \kitchen)
\end{equation}

Therefore, $\objat(\cupobj, \kitchen)$ is true after action $\drive(\hallway, \kitchen)$ if and only if in the initial situation, the robot is holding the cup or the cup is already in the kitchen.
Note that this query no longer contains any action terms and was reduced to a query about the initial situation.
All we need to do answer the query is to check if the regressed formula is satisfied by the initial situation, i.e., whether the following holds:
\[
  \bat_0 \models  \holding(o) \vee \objat(\cupobj, \kitchen)
\]

In our setting, a slight complication arises:
In \tesg, a trace consists of alternating time points and actions, where each time point may be any real number.
Therefore, to regress a clock formula, e.g., $c < 5$, we might want to subtract the time increment from the clock formula, e.g., for a time increment of $2$, the regressed clock formula becomes $c < 3$.
However, we do not allow real numbers in formulas, so we may not simply use arbitrary time points as a term for regression.
Also, we cannot regress a formula that contains a program term, e.g., $[\delta] \alpha$:
In \tesg, actions do not have a time argument, but instead the time points are determined by the program transition semantics and restricted by clock constraints.
Hence, we would need to consider all possible time successors, which in general are uncountably many.

For these reasons, we restrict the regressable formulas to static formulas and define regression only for rational traces, i.e., traces that only contain rational time points.
We will later see that this restricted regression operator is sufficient for our purposes.
With these considerations in mind, we define regression as follows:
\begin{definition}[Regression]\label{def:regression}
  Let \bat be a \ac{BAT}, $\alpha$ be a static formula and $z$ be a rational trace.
  The regression operator $\regress[z, \alpha]$ is defined inductively:
 \begin{enumerate}
   \item $\regress[z, (t_1 = t_2)] \eqdef (t_1 = t_2)$;
   \item $\regress[z, \alpha \wedge \beta] \eqdef \regress[z, \alpha] \wedge \regress[z, \beta]$;
   \item $\regress[z, \neg \alpha] \eqdef \neg \regress[z, \alpha]$;
   \item $\regress[z, \forall x.\, \alpha] \eqdef \forall x.\, \regress[z, \alpha]$;
   \item $\regress[z, G(\vec{t})] \eqdef {\gamma_G}^{\vec{x}}_{\vec{t}}$ for rigid predicates $G$;
   \item $\regress[z, \poss(t)] \eqdef {\pi_a}^a_t$;
   \item $\regress[z, g(t)] \eqdef {g_a}^a_t$;
   \item $\regress[z, \reset(t)] \eqdef {\gamma_c}^c_t$;
   \item $\regress[z, F(\vec{t})]$ for relational fluents $F$ is defined inductively by:
     \begin{enumerate}
       \item $\regress[\la\ra, F(\vec{t})] \eqdef F(\vec{t})$;
       \item $\regress[z \cdot t, F(\vec{t})] \eqdef \regress[z, F(\vec{t})]$ if $t \in \stdtimename$;
       \item $\regress[z \cdot t, F(\vec{t})] \eqdef \regress[z, {\gamma_F(\vec{t})}^a_t]$ if $t \in \stdactname$;
     \end{enumerate}
   \item $\regress[z, f(\vec{t} = t')]$ for functional fluents $f$ is defined inductively by:
     \begin{enumerate}
       \item $\regress[\la\ra, f(\vec{t} = t')] = f(\vec{t} = t')$;
       \item $\regress[z \cdot t, f(\vec{t}) = t'] = \regress[z, f(\vec{t}) = t']$ if $t \in \stdtimename$;
       \item $\regress[z \cdot t, f(\vec{t}) = t'] = \regress[z, {\gamma_f(\vec{t}, t')}^a_t]$ if $t \in \stdactname$;
     \end{enumerate}
   \item $\regress[z, c \bowtie r]$ for clocks $c$ is defined inductively by:
     \begin{enumerate}
       \item $\regress[\la\ra, c \bowtie r] = 0 \bowtie r$;
       \item $\regress[z \cdot p, c \bowtie r] = \regress[z \cdot p, \reset(c)] \wedge 0 \bowtie r \vee \neg \regress[z \cdot p, \reset(c)] \wedge \regress[z, c \bowtie r]$ if $p \in \stdactname$;
       \item $\regress[z \cdot t, c \bowtie r] = \regress[z, c \bowtie r']$ if $t \in \stdtimename$ and where $r' = r - (t - \ztime(z))$.
         \qedhere
     \end{enumerate}
 \end{enumerate}
\end{definition}

Note in particular the regression rule for clock formulas $c \bowtie r$: For an action step $p$, if the clock $c$ is reset by $p$, then the regressed formula is equivalent to $0 \bowtie r$, where the clock $c$ was replaced by the constant $0$.
Otherwise, the clock formula is unchanged.
For a time increment $t \in \stdtimename$, the regression operator subtracts the time increment from the right-hand side $r$ of the clock formula.

To show the correctness of the regression operator, we first define a world $w_\bat$ for a given world $w$ and \ac{BAT} \bat:
\begin{definition}
  Let $w$ be a world and \bat be a \ac{BAT} over $(\fluents, \clockset)$.
  Then $w_\bat$ is a world satisfying the following conditions:
  \begin{enumerate}
    \item For any functional $g \not\in \fluents$ and  relational $G \not\in \fluents$:
      \begin{align*}
        w_\bat[g(\vec{n}), z] &= w[g(\vec{n}), z]
        \\
        w_\bat[G(\vec{n}), z] &= w[G(\vec{n}), z]
      \end{align*}
    \item For any functional $f \in \fluents$:
      \begin{align*}
        w_\bat[f(\vec{n}), \la\ra] &= w[f(\vec{n}), \la\ra]
        \\
        w_\bat[f(\vec{n}), z \cdot t] &= w_\bat[f(\vec{n}), z] \text{ for } t \in \realpos
        \\
        w_\bat[f(\vec{n}), z \cdot p] &= n \text{ iff } w_\bat, z\models {\gamma_f(\vec{n}, n)}^a_p \text{ for } p \in \stdactname
      \end{align*}
    \item For any relational $F \in \fluents$:
      \begin{align*}
        w_\bat[F(\vec{n}), \la\ra] &= w[F(\vec{n}), \la\ra]
        \\
        w_\bat[F(\vec{n}), z \cdot t] &= w_\bat[F(\vec{n}), z \cdot t] \text{ for } t \in \realpos
        \\
        w_\bat[F(\vec{n}), z \cdot p] &= 1 \text{ iff } w_\bat, z \models {\gamma_F(\vec{n})}^a_p \text{ for } p \in \stdactname
      \end{align*}
    \item For $\poss$: $w_\bat[\poss(p), z] = 1$ iff $w_\bat, z  \models {\pi_a}^a_p$;
    \item For clock constraints $g$: $w_\bat[g(p), z] = 1$  iff $w_\bat, z \models {g_a}^a_p$.
      \qedhere
  \end{enumerate}
\end{definition}
Hence, the world $w_\bat$ is like $w$ except that is satisfies the \ac{BAT} \bat.
We can show the following for $w_\bat$:
\begin{lemmaE}
  Let \bat be a \ac{BAT} and a $w$ be a world.
  $w_\bat$ exists and is uniquely defined.
\end{lemmaE}
\begin{proofE}
  The argument is similar to \cite[Lemma 3]{lakemeyerSemanticCharacterizationUseful2011}:
  Clearly, $w_\bat$ exists.
  The uniqueness follows from the fact that $\pi$ is a fluent situation formula and that for all fluents
  in $\fluents$, once their initial values are fixed, then the values after any number of actions are uniquely determined by $\bat_\text{post}$.
\end{proofE}

Finally, we can show that given a world  $w$  and a rational trace $z$, we can indeed use regression to determine whether a fluent sentence holds after $z$:
\begin{theoremE}
  Let \bat be a \ac{BAT}, $\alpha$ a regressable sentence, $w$ a world with $w \models \bat_0$, and $z$ a rational trace.
  Then $\regress[z, \alpha]$ is a fluent sentence that satisfies
  \[
    w_\bat, z \models \alpha \text{ iff } w \models \regress[z, \alpha]
  \]
\end{theoremE}
\begin{proofE}
  By induction on the length of $z$ and structural sub-induction on $\alpha$.
  \\
  \textbf{Base case.}
  Let $z = \la\ra$.
  Note that the $\alpha$ is unchanged by the regression operator \regress unless $\alpha$ is a clock formula  $c \bowtie r$.
  In this case $\regress[\la\ra, c \bowtie r] = 0 \bowtie r$.
  By definition of worlds (\autoref{def:tesg-world}), $w[\la\ra, c] = 0$ for every clock standard name.
  The claim directly follows.
  \\
  \textbf{Induction step.}
  We distinguish two cases:
  \begin{enumerate}
    \item Let $z = z' \cdot t$ for some $t \in \stdtimename$.
      By induction, $w_\bat, z' \models \alpha$ iff $w \models \regress[z', \alpha]$.
      Again, the regression operator \regress leaves any fluent formula unchanged.
      Also, by definition of $w_\bat$, for every functional fluent $f \in \fluents$, $w_\bat[z' \cdot t , f(\vec{n})] = w_\bat[z' , f(\vec{n})]$ and for every relational fluent $F$, $w_\bat[z' \cdot t , F(\vec{n})] = w_\bat[z' , F(\vec{n})]$, i.e., the value of fluents does not change with a time step $t$.
      Therefore, the truth of a fluent formula $\alpha$ does not change and the claim directly follows.
      \\
      Now, consider a clock formula $c \bowtie r$ and let $d = t - \ztime(z')$.
      By definition of \regress, $\regress[z, c \bowtie r] = \regress[z', c \bowtie r']$ with $r' = r - d$.
      Also, by \autoref{def:tesg-world}, $w[z, c] = w[z', c] + d$  and therefore, $w_\bat, z \models c \bowtie r$ iff $w_\bat, z' \models c \bowtie (r - d)$ iff $w_\bat, z' \models c \bowtie r'$.
      By induction, $w_\bat, z' \models c \bowtie r'$ iff $w \models \regress[z', c \bowtie r']$ and so the claim follows.
    \item Let $z = z' \cdot p$ for some $p \in \stdactname$.
      For a relational fluent $F$, $\regress[z, F(\vec{t})] = \regress[z', {\gamma_F(\vec{t})}^a_p]$.
      By definition of $w_\bat$, $w_\bat, z \models F(\vec{t})$ iff $w_\bat, z' \models {\gamma_F(\vec{t})}^a_p$.
      By induction, $w_\bat, z' \models  {\gamma_F(\vec{t})}^a_p$ iff $w \models \regress[z', {\gamma_F(\vec{t})}^a_p]$, and so the claim follows.
      For a functional fluent $f$, the proof is analogous.
      Finally, we consider a clock formula  $c \bowtie r$.
      We distinguish two cases:
      \begin{enumerate}
        \item Let $w_\bat, z \models \reset(c)$ and therefore, $w \models \regress[z, \reset(c)]$, as shown above.
          From $w \models \regress[z, \reset(c)]$, it follows that $w \models \regress[z, c \bowtie r]$ iff $w \models \regress[z', 0 \bowtie r]$, which is equivalent to $w \models 0 \bowtie r$.
          It directly follows that $w_\bat, z \models c \bowtie r$ iff $w \models 0 \bowtie  r$.
          By \autoref{def:tesg-world}, $w_\bat[z, c] = 0$ and so $w_\bat, z \models c \bowtie r$ iff $w_\bat, z \models 0 \bowtie r$.
          It follows that $w_\bat, z \models c \bowtie r$ iff $w \models \regress[z, c \bowtie r]$.
        \item Let $w_\bat, z \models \neg \reset(c)$ and therefore, $w \models \neg \regress[z, \reset(c)]$, as shown above.
          From $w \models \neg \regress[z, \reset(c)]$, it follows that $\regress[z, c \bowtie r]$ iff $w \models \regress[z', c \bowtie r]$.
          On the other hand, by \autoref{def:tesg-world}, $w_\bat[z, c] = w_\bat[z', c]$.
          By induction, $w_\bat, z' \models c \bowtie r$ iff $w \models \regress[z', c \bowtie r]$ and so the claim follows.
          \qedhere
      \end{enumerate}
  \end{enumerate}
\end{proofE}

We will see in \autoref{chap:synthesis} that this notion of regression is sufficient for our purposes: As we will restrict the domain to a finite set of objects, considering a single world $w$ is not a restriction.
More importantly, we will see that considering rational traces is sufficient, as we can use \emph{regionalization} to reduce the time successors to a finite number at any point of the program execution.

\section{Complete-Information and Finite-Domain Basic Action Theories}\label{sec:complete-finite-bat}

We conclude the discussion of \acp{BAT} with two restrictions to \acp{BAT}:
\begin{description}
  \item[Complete information:] Generally, a \ac{BAT} allows to model incomplete information, e.g., it may entail $\bat \models \alpha \vee \beta$, but neither $\bat \models \alpha$ nor $\bat \models \beta$ holds.
    Complete information restricts a \ac{BAT} such that for every sentence $\alpha$, it either entails $\alpha$ or $\neg \alpha$.
  \item[Finite domain:] Generally, a \ac{BAT} may refer to infinitely many objects, e.g., by stating $\forall o.\, \objat(o, m_1)$, which states that for every standard name $n$ (of which there are infinitely many), $\objat(n, o_1)$ is true.
    A \emph{finite \ac{BAT}} restricts a \ac{BAT} to finitely many objects and therefore to a finite domain of discourse.
\end{description}


We start with complete information:
\begin{definition}[\Acp{BAT} with complete information]
  A \ac{BAT} \bat over $(\fluents, \clockset)$ is called a \emph{\ac{BAT} with complete information} if
  \begin{enumerate}
    \item for every primitive formula $\alpha$ over $(\fluents, \clockset)$, either $\bat_0  \models \alpha$ or $\bat_0 \models \neg \alpha$ holds,
    \item for every functional fluent $f \in \fluents$ and standard names $\vec{n}$,  there is some standard name $n$ such that $\bat_0 \models f(\vec{n}) = n$.
      \qedhere
  \end{enumerate}
\end{definition}

\begin{theoremE}\label{thm:complete-bat-equiv-worlds}
  Let $\Delta = (\bat, \delta)$ be a program and \bat a \ac{BAT} over $(\fluents, \clockset)$ with complete information.
  Let $w, w'$ be two worlds with $w \models \bat$ and $w' \models \bat$.
  Then, for every trace $z \in \traces$ and every formula $\alpha$ over $(\fluents, \clockset)$:
  \[
    w, z \models \alpha \text{ iff } w', z \models \alpha
  \]
\end{theoremE}
\begin{proofE}
  We first consider static $\alpha$:
  As \bat has complete information, it is clear that for every $F(\vec{n}) \in \primformulas$, $w[F(\vec{n}), \la\ra] = w'[F(\vec{n}), \la\ra]$ and for every $f(\vec{n}) \in \primterms$, $w[f(\vec{n}), \la\ra] = w'[f(\vec{n}), \la\ra]$.
  Now, once the initial value of each fluent has been fixed, the values after any actions $z$ is uniquely determined by $\bat_\text{post}$.
  As $w$ and $w'$ agree on each initial fluent value, they also agree on each fluent value after any actions.

  Now, for a program $\delta$, as $w$ and $w'$ agree on every static formula after any actions and $\delta$ may only contain static formulas, it directly follows that $z' \in \| \delta \|^z_w$ iff $z' \in \| \delta \|^z_{w'}$.
  As $w$ and $w'$ permit the same traces of $\delta$, it can be shown that $w, z \models [\delta] \beta$ iff $w, z' \models [\delta] \beta$ and also $w, z \models \llbracket \delta \rrbracket \phi$ iff $w', z \models \llbracket \delta \rrbracket \phi$.
  Finally, for $\alpha = \square \beta$, the claim directly follows from the fact that $w$ and $w'$ agree on any formula after any actions.
\end{proofE}

Therefore, $w$ and $w'$ are identical with respect to \bat, because they agree on every formula over $(\fluents, \clockset)$.
Hence, for a \ac{BAT} with complete information, it is sufficient to consider a single model $w \models \bat$.

We continue with \acp{BAT} that refer to a finite domain of objects.
The idea of a finite-domain \ac{BAT} is to restrict all quantifiers to objects of a certain type and then fix each object type to a finite set of objects.
In our case, the three types are \emph{objects}, \emph{actions}, and \emph{clocks}.
Formally, a finite-domain \ac{BAT} is defined as follows:
\begin{definition}[Finite-domain \ac{BAT}]
  We call a \acl{BAT} $\bat$ a \emph{finite-domain \ac{BAT} with domain $D$} if it satisfies the following conditions:
  \begin{enumerate}
    \item \label{def:fd-bat:domain}
      $\bat_0$ contains axioms
      \begin{itemize}
      \item $\forall x\; \tau_o(x) \equiv (x = o_1 \vee x = o_2 \vee \ldots \vee x = o_k)$,
      \item $\forall x\; \tau_a(x) \equiv (x = a_1 \vee x = a_2 \vee \ldots \vee x = a_l)$, and
      \item $\forall x\; \tau_c(x) \equiv (x = c_1 \vee x = c_2 \vee \ldots \vee x = c_m)$
      \end{itemize}
      where each $\tau_i$ is a rigid predicate of sort object, action, and clock, respectively, and each $o_i \in D$, $a_i \in D$, and $c_i \in D$ is a standard name of the corresponding sort.
    \item \label{def:fd-bat:quantifiers}
      Except for the axioms from \autoref{def:fd-bat:domain}, each $\forall$ quantifier in $\bat$ occurs as $\forall x.\, \tau_i(x) \supset \phi(x)$.
      \qedhere
  \end{enumerate}
\end{definition}
We call a program $\Delta = (\bat, \delta)$ a \emph{finite-domain program} if \bat is a finite-domain \ac{BAT}.
We also write $\exists x\mathbf{:}i.\,\phi$ for $\exists x.\, \tau_i(x) \wedge \phi$ and $\forall x\mathbf{:}i.\,\phi$ for $\forall x.\, \tau_i(x) \supset \phi$.
Furthermore, we abbreviate $\forallobj{x} \forallobj{y}\, \phi$ with $\forallobj{x,y}\, \phi$, similarly for the other types and existential quantification.
Since a finite-domain \ac{BAT} restricts the domain of discourse to be finite, quantifiers can be understood as abbreviations:
\[
  \forall x\mathbf{:}\tau_o. \phi \eqdef \bigwedge_{i=1}^k \phi^x_{o_i}
  \qquad
  \forall x\mathbf{:}\tau_a. \phi \eqdef \bigwedge_{i=1}^l \phi^x_{a_i}
  \qquad
  \forall x\mathbf{:}\tau_c. \phi \eqdef \bigwedge_{i=1}^m \phi^x_{c_i}
\]

\begin{example} \label{ex:fd-bat}
  Coming back to the \ac{BAT} from \autoref{ex:bat}, we can change it to be a finite-domain \ac{BAT} by adding the following sentences to $\bat_0$:
  \begin{align*}
       & \forall x. \tau_o(x) \equiv (x = m_1 \vee x = m_2 \vee x = o_1)
    \\ & \forall x. \tau_a(x) \equiv (x = \sac{\drive(m_1, m_2)} \vee x = \eac{\drive(m_1, m_2)}
    \\ &   \quad \qquad \vee x = \sac{\drive(m_2, m_1)} \vee x = \eac{\drive(m_2, m_1)}
    \\ &   \quad \qquad \vee x = \sac{\grasp(m_1, o_1)} \vee x = \eac{\grasp(m_1, o_1)}
    \\ &   \quad \qquad \vee x = \sac{\grasp(m_2, o_1)} \vee x = \eac{\grasp(m_2, o_1)}
    \\ &   \quad \qquad \vee x = \sac{\bootcam} \vee x = \eac{\bootcam}
    \\ &   \quad \qquad \vee x = \sac{\stopcam} \vee x = \eac{\stopcam}
    \\ & \forall x. \tau_c(x) \equiv (x = q_1 \vee x = q_2 \vee x = q_3 \vee x = q_4 \vee x = q_5 \vee x = q_6)
  \end{align*}
  This fixes the domain to the three objects $m_1, m_2, o_1$ and the corresponding actions, and restricts the clocks to the set $\{ q_1, \ldots, q_6 \}$.
\end{example}

We can now define an equivalence relation between worlds, where two worlds are equivalent if they initially agree on all fluents:
\newcommand*{\equivbat}{\ensuremath{\equiv_{\fluents}}\xspace}
\begin{definition}
  We define the equivalence relation \equivbat of worlds with respect to a set of fluents $\fluents$ and domain $D$ as
    $w \equivbat w'$ iff
    \begin{enumerate}
      \item $w[F(\vec{n}), \la\ra] = w'[F(\vec{n}), \la\ra]$ for every relational fluent $F \in \fluents$ and every $\vec{n} \in D$ of the right sort, and
      \item $w[f(\vec{n}), \la\ra] = w'[f(\vec{n}), \la\ra]$ for every functional fluent  $f \in \fluents$ and every $\vec{n} \in D$ of the right sort.
        \qedhere
    \end{enumerate}
\end{definition}
We use $[w]$ to denote the equivalence class $[w] = \{ w' \in \worlds \mid w \equivbat w' \}$. 
As a finite-domain \ac{BAT} only refers to finitely many fluents, the following is immediate:
\begin{remark}\label{thm:finite-w-equivalence-classes}
  For a finite-domain \ac{BAT} $\bat$ over $(\fluents, \clockset)$, \equivbat has finitely many equivalence classes.
\end{remark}

Furthermore, for each equivalence class, we only need to consider one world:
\begin{theoremE}[]\label{thm:fd-bat-single-world}
  Let $\bat$ be a finite-domain \ac{BAT} over $(\fluents, \clockset)$ with domain $D$ and $w, w'$ be two worlds with $w \models \bat$, $w' \models \bat$, and $w \equivbat w'$.
  Then, for every formula $\alpha$ over $(\fluents, \clockset)$:
  \[
    w \models \alpha \text{ iff } w' \models \alpha.
  \]
\end{theoremE}
\begin{proofE}
  From $w \equivbat w'$, it is clear that for every primitive formula $\phi$, $w \models \phi$ iff $w' \models \phi$.
  Hence, we can construct a \ac{BAT} $\bat'$ that is like $\bat$ but which additionally contains the following sentences as part of $\bat_0'$:
  \begin{itemize}
    \item $F(\vec{n})$ for every relational fluent $F \in \fluents$ and all $\vec{n} \in D$ with $w \models F(\vec{n})$,
    \item $\neg F(\vec{n})$ for every relational fluent $F \in \fluents$ and $\vec{n} \in D$ with $w \models \neg F(\vec{n})$,
    \item $f(\vec{n}) = n$ for every functional fluent $f \in \fluents$ and $\vec{n} \in D$ with $w \models f(\vec{n}) = n$.
  \end{itemize}
  Clearly, $w \models \bat'$ and $w' \models \bat'$.
  Furthermore, by construction, $\bat'$ is a \ac{BAT} with complete information.
  Therefore, by \autoref{thm:complete-bat-equiv-worlds}, $w \models \alpha$ iff $w' \models \alpha$.
\end{proofE}

Thus, for the sake of simplicity, for a given \ac{BAT} $\bat$ with complete information and finite domain, we may assume that we are given a single world $w$ with $w \models \bat$.

\section{\tesg and \esg}\label{sec:tesg-esg}

As \tesg can be seen as extension of \esg, it is helpful to discuss their differences.
While both \esg and \tesg allow to express temporal properties of program traces, \tesg extends the temporal operators with timing constraints.
We first summarize the logic \esg before we compare the two logics in \autoref{sec:esg-tesg-valid-sentences} and \autoref{ssub:tesg-esg-bats}.

\subsection{The Logic \esg}

The language of \esg is similar to the language of \tesg except that it does not mention any clocks or timing constraints:
\begin{definition}[Language of \esg]
  Every \tesg term of sort object or action is a term of \esg.
  A program expression of \tesg is also a program expression of \esg if it does not mention any clock terms or any of the distinguished predicates $\clockconstraint$ and $\reset$.
  The language of \esg consists of those \tesg formulas that
  \begin{enumerate}
    \item only mention \esg terms and \esg program expressions,
    \item only mention the until operator $\until{}$ with an unbounded interval $I = [0, \infty)$.
      \qedhere
  \end{enumerate}
\end{definition}

As before, we call a function term \emph{primitive} if it is of the form $f(n_1,\ldots,n_k)$, with $f,n_i$ being standard names.
We denote the set of primitive terms as \primobjs (objects) and \primactions (actions), and we denote the set of all primitive terms as $\primterms \eqdef \primobjs \cup \primactions$.
Furthermore, a term is is called \emph{rigid} if it only consists of rigid function symbols and standard names.

In contrast to \tesg traces, traces of \esg do not mention any time points, but only actions:
\begin{definition}[\esg Traces]
  An \esg trace is a finite or infinite sequence of action standard names
  $z = p_1 \cdot p_2 \cdot \ldots$, where $p_i \in \stdactname$.
  We denote the set of all finite \esg traces with $\fintraces_{\esg}$, the set of all infinite \esg traces with $\inftraces_{\esg}$, and the set of all \esg traces with $\alltraces_{\esg}$.
\end{definition}

A world of \esg is similar to a world of \tesg, except that it mentions \esg traces and does not define any clock values and therefore does not need to satisfy any clock value constraints:
\begin{definition}[\esg Worlds]
  A world of \esg is a mapping that maps
  \begin{enumerate}
     \item 
       $\primobjs \times \fintraces_\esg \rightarrow \stdobjname$,
     \item 
         $\primactions \times \fintraces_\esg \rightarrow \stdactname$,
     \item 
         $\primformulas \times \fintraces_\esg \rightarrow \{ 0, 1 \}$.
  \end{enumerate}
  satisfying the following constraints:
   \begin{description}
     \item[Rigidity:]
       If $R$ is a rigid function or predicate symbol, then for all $z$ and $z'$ in \traces:
       \[
         w[R(n_1,\ldots,n_k),z]=w[R(n_1,\ldots,n_k),z']
       \]
     \item[Unique names for actions:]
       If $g(n_1, \ldots, n_k)$ and $g'(n_1', \ldots, n_l')$ are two distinct primitive action terms, then for all $z$ and $z'$ in $\fintraces_\esg$:
       \[
         w[ g(n_1, \ldots, n_k), z ] \neq w[ g'(n_1', \ldots, n_l'), z' ]
       \]
   \end{description}
  The set of all worlds of \esg is denoted by $\worlds_\esg$.
\end{definition}

Terms of \esg are denoted in the same way as in \tesg: 
\begin{definition}[Denotation of \esg terms]\label{def:esg-denotation}
  Given a ground term $t$, a world $w \in \worlds_\esg$, and a trace $z \in \mathcal{Z}_\esg$, we define $\denot{t}{z}{w}$ by:
  \begin{enumerate}
    \item if $t \in \mathcal{N}$, then $\denot{t}{z}{w} = t$,
    \item if $t = f(t_1,\ldots,t_k)$ then $\denot{t}{z}{w} = w[f(n_1,\ldots,n_k), z]$, where $n_i = \denot{t_i}{z}{w}$
      \qedhere
  \end{enumerate}
\end{definition}

While the program transition semantics of \tesg consists of time and action steps, all transitions in the \esg program transition semantics are action steps:
\begin{definition}[\esg Program Transition Semantics]
  The transition relation $\warrow$ among configurations, given a world $w$, is the least set satisfying the following conditions:
  \begin{enumerate}
    \item $\la z, a \ra \warrow \la z', \nil \ra$ if $z' = z \cdot p$ and $p = \vert a \vert^z_w$
    \item $\la z,\delta_1;\delta_2 \ra \warrow \la z \cdot p, \gamma;\delta_2 \ra$ if $\la z, \delta_1 \ra \warrow \la z \cdot p, \gamma \ra$
    \item $\la z,\delta_1;\delta_2 \ra \warrow \la z \cdot p, \delta' \ra$ if $\la z, \delta_1 \ra \in \mathcal{F}^w \text{ and } \la z, \delta_2 \ra \warrow \la z \cdot p, \delta' \ra$
    \item $\la z,\delta_1 \vert \delta_2 \ra \warrow \la z \cdot p, \delta' \ra$ if  $\la z, \delta_1 \ra \warrow \la z \cdot p, \delta' \ra \text{ or }\la z, \delta_2 \ra \warrow \la z \cdot p, \delta' \ra$
    \item $\la z,\delta^* \ra \warrow \la z \cdot p, \gamma; \delta^* \ra$ if $\la z,\delta \ra \warrow \la z \cdot p, \gamma \ra$
    \item $\la z,\delta_1 \| \delta_2 \ra \warrow \la z \cdot p, \delta' \| \delta_2 \ra$ if $\la z, \delta_1 \ra \warrow \la z \cdot p, \delta' \ra$

    \item $\la z,\delta_1 \| \delta_2 \ra \warrow \la z \cdot p, \delta_1 \| \delta' \ra$ if $\la z, \delta_2\ra \warrow \la z \cdot p, \delta' \ra$
  \end{enumerate}
  The set of final configurations \final{} is the smallest set that satisfies the following conditions:
  \begin{enumerate}
    \item $\la z,\alpha? \ra \in \final$ if $w, z \models \alpha$
    \item $\la z,\delta_1;\delta_2 \ra \in \final$ if $\la z,\delta_1 \ra \in \final \text{ and } \la z,\delta_2 \ra \in \final$
    \item $\la z,\delta_1 \vert \delta_2 \ra \in \final$ if $\la z,\delta_1 \ra \in \final \text{ or } \la z,\delta_2 \ra \in \final$
    \item $\la z,\delta^* \ra \in \final$
    \item $\la z,\delta_1 \| \delta_2 \ra \in \final$ if $\la z,\delta_1 \ra \in \final \text{ and } \la z,\delta_2 \ra \in \final$
      \qedhere
  \end{enumerate}
\end{definition}

As before, program traces are obtained from the transition semantics by following the program transitions:
\begin{definition}[Program Traces]\label{def:esg-program-trace}
  Given a world $w \in \worlds_\esg$ and a finite trace $z \in \fintraces_\esg$, the \emph{program traces} of a program expression $\delta$ starting in $z$ are defined as follows:
  \begin{align*}
    \| \delta \|^z_w = &\{ z' \in \fintraces \mid \la z, \delta \ra \warrow^* \la z \cdot z', \delta' \ra \text{ and } \la z \cdot z',  \delta' \in \final \} \cup
    \\
                       &\{ \pi \in \inftraces \mid \la  z, \delta \ra \warrow \la z \cdot \pi^{(i)}, \delta_1 \ra \warrow \la z \cdot \pi^{(2)}, \delta_2 \ra \warrow \ldots
                    \\ &\quad \text{ where for all $i$, }  \la z \cdot \pi^{(i)}, \delta_i \ra \not\in \final \}
    \qedhere
  \end{align*}
\end{definition}

We can now define the truth of \esg formulas.
The definition is the same as truth of \tesg formulas (\autoref{def:tesg-truth}) except for the until operator $\until{}$, where we do not need to consider any timing constraints:
\begin{definition}[Truth of \esg Formulas]\label{def:esg-truth}
  Given a world $w \in \worlds_\esg$ and a formula $\alpha$, we define $w \models \alpha$ as $w,\langle\rangle \models \alpha$ and where $w, z \models \alpha$ is defined as follows for every $z \in \fintraces_\esg$:
  \begin{enumerate}
    \item $w, z \models F(t_1,\ldots,t_k)$ iff $w[F(n_1,\ldots,n_k),z] = 1$, where $n_i=\denot{t_i}{z}{w}$,
    \item $w, z \models (t_1 = t_2)$ iff $n_1$ and $n_2$ are identical, where $n_i=\denot{t_i}{z}{w}$,
    \item $w, z \models \alpha \wedge \beta$ iff $w, z \models \alpha$ and $w, z \models \beta$,
    \item $w, z \models \neg \alpha$ iff $w, z \not\models \alpha$,
    \item $w, z \models \forall x.\, \alpha$ iff $w, z \models \alpha^x_n$ for every standard name of the right sort,
    \item $w, z \models \square \alpha$ iff $w, z\cdot z' \models \alpha$ for all $z' \in \fintraces_\esg$,
    \item $w, z \models [\delta]\alpha$ iff $w, z \cdot z' \models \alpha$ for all $z' \in \|\delta\|^z_w$,
    \item $w, z \models \llbracket \delta \rrbracket \phi$ iff for all $\tau \in \|\delta\|^z_w$, $w, z, \tau \models \phi$,
    \item $w, z \models \llbracket \delta \rrbracket^{< \infty} \phi$ iff for all finite $z' \in \|\delta\|^z_w$, $w, z, z' \models \phi$.\todo{Complete proof}
  \end{enumerate}

  The truth of trace formulas $\phi$ is defined as follows for $w \in \worlds_\esg$, $z \in \fintraces_\esg, \tau \in \alltraces_\esg$:
  \begin{enumerate}
    \item $w, z, \tau \models \alpha$ iff $w, z \models \alpha$ and $\alpha$ is
      a situation formula;
    \item $w, z, \tau \models \phi \wedge \psi$ iff $w, z, \tau \models \phi$
      and $w, z, \tau \models \psi$;
    \item $w, z, \tau \models \neg \phi$ iff $w, z, \tau \not\models \phi$;
    \item $w, z, \tau \models \forall x.\, \phi$ iff
      $w, z, \tau \models \phi^x_n$ for all $n \in \mathcal{N}_x$;
    \item $w, z, \tau \models \phi \until{} \psi$ iff there is a $\tau' \in \alltraces_\esg$ and $z_1 \in \traces_\esg$ with $z_1 = p_1 \cdots p_k \neq \la\ra$ such that%
      \footnote{
        Note that in contrast to the original \esg semantics, we assume strict-until in order to be compatible with \tesg.
        However, as weak-until can be expressed with strict-until, this is no restriction.
      }
      \begin{enumerate}
        \item $\tau = z_1 \cdot \tau'$,
        \item $w, z \cdot z_1, \tau' \models \psi$,
        \item for all $z_2 = p_i \cdots p_j$ with $z_1 = z_2 \cdot z_3$, $z_2 \neq \la\ra$, and $z_3 \neq \la\ra$: $w, z \cdot z_2, z_3 \cdot \tau' \models \phi$.
          \qedhere
      \end{enumerate}
  \end{enumerate}
\end{definition}

To distinguish the semantics of \esg and \tesg, we will also write $\models_\esg$ and $\models_{\tesg}$ do denote truth in \esg and \tesg respectively.


\subsection{Valid Sentences of \esg and \tesg}\label{sec:esg-tesg-valid-sentences}
As a \tesg trace is not a valid \esg trace, we first translate a \tesg trace to an \esg trace by omitting the time points, resulting in a \emph{symbolic trace}:
\begin{definition}[Symbolic Trace]
  Let $z = \left(p_0, t_0\right)\left(p_1, t_1\right)\cdots\left(p_n, t_n\right) \in \traces_{\tesg}$.
  Then the corresponding \emph{symbolic trace} $\untimed{z} \in \traces_{\esg}$ is the trace $\untimed{z} \eqdef p_0 p_1 \cdots p_n$.
\end{definition}
%

%


Similarly, we cannot directly use a \tesg world as an \esg world.
This time, we do the translation in the other direction, i.e., given a world of \esg, we define the corresponding \emph{time-extended world} of \tesg:
\begin{definition}[Time-extended world]
  Let $w \in \worlds_{\esg}$ be a world of \esg.
  We construct the corresponding \emph{time-extended world} $w_t \in \worlds_{\tesg}$ from $w$.
  For every $z_t \in \traces_{\tesg}$, we define $w_t$ as follows:
  \begin{enumerate}
    \item For every primitive formula $P(n_1, \ldots, n_k)$:
      \[
        w_t[P(n_1, \ldots, n_k), z_t] \eqdef w[P(n_1, \ldots, n_k), \untimed{z_t}]
      \]
    \item For every primitive term $f(n_1, \ldots, n_k)$:
      \[
        w_t[f(n_1, \ldots, n_k), z_t] \eqdef w[f(n_1, \ldots, n_k), \untimed{z_t}]
      \]
    \item For every clock standard name $c \in \stdclockname$, we set $w[c, z_t]$ to some value that satisfies the time progression criteria from \autoref{def:tesg-world}.\footnote{Note that if we compare \esg and \tesg, clock values are irrelevant because we cannot refer to their values in the language of \esg.
      However, for a complete definition of the world of \tesg, we need to define the clock values somehow.}
  \end{enumerate}
  The \tesg world $w_t$ agrees with the \esg world $w$ on every primitive formula and primitive term after every sequence of actions, independent of the time point of each action.
\end{definition}

We start by comparing the valid sentences of both logics.
First, for a time-extended world $w_t$, we can show that it assigns the same value to each action or object term:
\begin{lemmaE}\label{lma:time-extended-worlds-same-denotations}
  Let $w \in \worlds_{\esg}$ and $w_t \in \worlds_{\tesg}$ the corresponding time-extended world.
  For every timed trace $z_t \in \traces_{\tesg}$, untimed trace $z \in \traces_{\esg}$ with $z = \untimed{z_t}$, and every action or object term $p$, the following holds:
  \[
    \lvert p \rvert^z_w = \lvert p \rvert^{z_t}_{w_t}
  \]
\end{lemmaE}
\begin{proofE}
  By structural induction on $p$:
  \begin{itemize}
    \item Let $p \in \stdname$.
      Clearly, $\lvert p \rvert^z_w = \lvert p \rvert^{z_t}_{w_t} = p$.
    \item Let $p = f(p_1, \ldots, p_k)$.
      By definition of $\lvert \cdot \rvert$, $\lvert p \rvert^z_w = w[f(n_1, \ldots, n_k), z]$ with $n_i = \lvert p_i \rvert^z_w$.
      By induction, for each $p_i$, it follows that $\lvert p_i \rvert^{z_t}_{w_t} = \lvert p_i \rvert^z_w$.
      Furthermore, by definition of $w_t$, $w_t[f(n_1, \ldots, n_k), z_t] = w[f(n_1, \ldots, n_k), z]$.
      Therefore, $\lvert p \rvert^{z_t}_{w_t} = \lvert p \rvert^z_w$.
  \end{itemize}
  It follows that $\lvert p \rvert^z_w = \lvert p \rvert^{z_t}_{w_t}$.
\end{proofE}

The time-extended world $w_t$ also satisfies the same static sentences:
\begin{lemmaE}\label{lma:tesg-esg-time-extended-world-equivalence-static}
  Let $w \in \worlds_{\esg}$ and $w_t \in \worlds_{\tesg}$ the corresponding time-extended world.
  Let $\alpha$ be a static sentence of \esg.
  Then for every timed trace $z_t \in \traces_{\tesg}$ and untimed trace $z \in \traces_{\esg}$ with $z = \untimed{z_t}$, the following holds:
  \[
    w, z \models_{\esg} \alpha \text{ iff } w_t, z_t \models_{\tesg} \alpha
  \]
\end{lemmaE}
\begin{proofE}
  By structural induction on $\alpha$:
  \begin{itemize}
    \item Let $\alpha = P(p_1, \ldots, p_k)$.
      Let $\lvert p_i \rvert^z_w = n_i$ for each $p_i$.
      For each $p_i$, it follows by \autoref{lma:time-extended-worlds-same-denotations} that $\lvert p_i \rvert^{z_t}_{w_t} = \lvert p_i \rvert^z_w$ and thus $\lvert p_i \rvert^{z_t}_{w_t} = n_i$.
      Furthermore, by definition of $w_t$, it follows that $w_t[P(n_1, \ldots, n_k), z_t] = w[P(n_1, \ldots, n_k), z]$.
      Thus, $w, z \models \alpha$ iff $w_t, z_t \models \alpha$.
    \item Let $\alpha = (p_1 = p_2)$, where $p_1$ and $p_2$ are terms.
      By \autoref{lma:time-extended-worlds-same-denotations}, $\lvert p_i \rvert^{z_t}_{w_t} = \lvert p_i \rvert^z_w = n_i$ for some standard name $n_i$.
      The semantics of \esg and \tesg do not differ with respect to equality, i.e., $w, z \models (n_1 = n_2)$ and $w_t, z_t \models (n_1 = n_2)$  iff $n_1$ and $n_2$ are identical.
      Thus, $w, z \models \alpha \text{ iff } w_t, z_t \models \alpha$.
    \item Let $\alpha = \alpha_1 \wedge \alpha_2$.
      By induction, $w_t, z_t \models \alpha_1$ iff $w, z \models \alpha_1$ and $w_t \models \alpha_2$ iff $w, z \models \alpha_2$.
      Again, the semantics of conjunction do not differ, thus $w, z \models \alpha_1 \wedge \alpha_2$ iff $w_t, z_t \models \alpha_1 \wedge \alpha_2$.
    \item Let $\alpha = \neg \beta$.
      By induction, $w, z \models \beta$ iff $w_t, z_t \models \beta$.
      It follows directly that $w, z \models \neg \beta$ iff $w_t, z_t \models \neg \beta$.
    \item Let $\alpha = \forall x.\, \beta$.
      By induction, for each $n \in \stdname_x$ of the same sort as $x$, $w, z \models \beta^x_n$ iff $w_t, z_t \models \beta^x_n$.
      As \esg and \tesg have the same standard names, it follows that $w, z \models \forall x.\, \beta$ iff $w_t, z_t \models \forall x.\, \beta$.
      \qedhere
  \end{itemize}
\end{proofE}

Also, the time-extended world $w_t$ allows the same symbolic program traces:
\begin{lemmaE}\label{lma:tesg-esg-same-traces}
  ~
  \begin{enumerate}
    \item
      Let $\la z, \delta \ra \warrow \la z_1, \delta_1 \ra \warrow \ldots \warrow \la z_n, \delta_n \ra$ be a finite sequence of transitions starting in $\la z, \delta \ra$.
      Then there is a finite sequence of transitions $\la z_t, \delta \ra \warrow[w_t] \la z_{t,1}, \delta_1 \ra \warrow[w_t] \ldots \warrow[w_t] \la z_{t,n}, \delta_n \ra$ starting in $\la z_t, \delta \ra$ such that $\untimed{z_{t,i}} = z_i$ and $\la z_{t,n}, \delta_n \ra \in \final[w_t]$ iff $\la z_n, \delta_n \ra \in \final$.
    \item
      Let $\la z_t, \delta \ra \warrow[w_t] \la z_{t,1}, \delta_1 \ra \warrow[w_t] \ldots \warrow[w_t] \la z_{t,n}, \delta_n \ra$ be a finite sequence of transitions starting in $\la z_t, \delta \ra$.
      Then there is a finite sequence of transitions $\la z, \delta \ra \warrow \la z_1, \delta_1 \ra \warrow \ldots \warrow \la z_n, \delta_n \ra$ starting in $\la z, \delta \ra$ such that $\untimed{z_{t,i}} = z_i$ and $\la z_n, \delta_n \ra \in \final$ iff $\la z_{t,n}, \delta_n \ra \in \final[w_t]$.
  \end{enumerate}
\end{lemmaE}
\begin{proofE}
  \begin{enumerate}
    \item
      Let $\la z, \delta \ra \warrow \la z_1, \delta_1 \ra \warrow \ldots \warrow \la z_n, \delta_n \ra$ such that $\la z_n, \delta_n \ra \in \final$.
      We first show by induction on the number of transitions $i$ that $\la z_t, \delta \ra \warrow[w_t] \la z_{t,1}, \delta_1 \ra \warrow[w_t] \ldots \warrow[w_t] \la z_{t,n}, \delta_n \ra$ with $\untimed{z_{t,i}} = z_i$:
      \\
      \textbf{Base case.}
      Let $i = 0$, i.e., there are no transitions and therefore $\la z_n, \delta_n \ra = \la z, \delta \ra$.
      As $\warrow[w_t]^*$ is defined as reflexive and transitive closure of $\warrow[w_t]$, it is clear that $\la z_t, \delta \ra \warrow[w_t]^* \la z_t, \delta \ra$.
      \\
      \textbf{Induction step.}
      Let $\la z, \delta \ra \warrow[w] \la z_1, \delta_1 \ra \warrow[w] \ldots \warrow[w] \la z_i, \delta_i \ra$.
      By induction, $\la z_t, \delta \ra \warrow[w_t] \la z_{t,1}, \delta_1 \ra \warrow[w_t] \ldots \warrow[w_t] \la z_{t,i}, \delta_i \ra$.
      Now, assume $\la z_i, \delta_i \ra \warrow \la z_i \cdot p_{i+1}, \delta_{i+1} \ra$.
      By \autoref{def:tesg-trans}, $\la z_{t,i}, \delta_i \ra \wtarrow[d] \la z_{t,i} \cdot t, \delta_i \ra$ for every $d \in \realpos$ and $t = \ztime(z_{t,i}) + d$.
      Also, note that $\untimed{z_{t,i} \cdot t} = \untimed{z_{t,i}} = z_i$.
      As the rules of the \esg transition semantics exactly correspond to the action step of the \tesg transition semantics, it directly follows that $\la z_{t,i} \cdot t, \delta_i \ra \warrow[w_t] \la z_{t,i} \cdot t \cdot p_{i+1}, \delta_{i+1} \ra$  and $\untimed{z_{t,i} \cdot t \cdot p_{i+1}} = z_{i+1}$ for each possible transition type.

      Next, we show by structural induction on $\delta_n$ that $\la z_n, \delta_n \ra \in \final$ iff $\la z_{t,n}, \delta_n \ra \in \final[w_t]$:
      \begin{itemize}
        \item Let $\delta_n = \alpha?$ for some static situation formula $\alpha$.
          Then $\la z_n, \delta_n \ra \in \final$ iff $w, z_n \models \alpha$.
          By \autoref{lma:tesg-esg-time-extended-world-equivalence-static}, $w, z_n \models \alpha$ iff $w_t, z_{t,n} \models \alpha$.
          Thus, by \autoref{def:tesg-trans}, $\la z_n, \delta_n \ra \in \final$ iff $\la z_{t,n}, \delta_n \ra \in \final[w_t]$.
        \item For all other cases, the claim follows by induction and from the fact that the final configurations of \esg and \tesg are defined in the same way.
      \end{itemize}
    \item
      Let $\la z_t, \delta \ra \warrow[w_t] \la z_{t,1}, \delta_1 \ra \warrow[w_t] \ldots \warrow[w_t] \la z_{t,n}, \delta_n \ra$ such that $\la z_{t,n}, \delta_n \ra \in \final[w_t]$.
      We first show by induction on the number of transitions $i$ that $\la z, \delta \ra \warrow \la z_1, \delta_1 \ra \warrow \ldots \warrow \la z_n, \delta_n \ra$:
      \\
      \textbf{Base case.}
      Let $i = 0$, i.e., there are no transitions and therefore $\la z_{t,n}, \delta_n \ra = \la z_t, \delta \ra$.
      As $\warrow^*$ is defined as reflexive and transitive closure of $\warrow$, it is clear that $\la z, \delta \ra \warrow^* \la z, \delta \ra$.
      \\
      \textbf{Induction step.}
      Let $\la z_t, \delta \ra \warrow[w_t] \la z_{t,1}, \delta_1 \ra \warrow[w_t] \ldots \warrow[w_t] \la z_i, \delta_i \ra$.
      By induction, $\la z, \delta \ra \warrow \la z_{1}, \delta_1 \ra \warrow \ldots \warrow \la z_{i}, \delta_i \ra$.
      Now, assume $\la z_{t,i}, \delta_i \ra \warrow \la z_{t,i} \cdot t \cdot p_{i+1}, \delta_{i+1} \ra$.
      Note that $\untimed{z_{t,i} \cdot t} = \untimed{z_{t,i}} = z_i$.
      As the rules of the \esg transition semantics exactly correspond to the action step of the \tesg transition semantics, it directly follows that $\la z_{i}, \delta_i \ra \warrow \la z_{i} \cdot p_{i+1}, \delta_{i+1} \ra$ for each possible transition type.

      Next, we show that by structural induction on $\delta_n$ that $\la z_{t,n}, \delta_n \ra \in \final[w_t]$ iff $\la z_n, \delta_n \ra \in \final$:
      \begin{itemize}
        \item Let $\delta_n = \alpha?$ for some static situation formula $\alpha$.
          Then $\la z_{t,n}, \delta_n \ra \in \final[w_t]$ iff $w, z_{t,n} \models \alpha$.
          By \autoref{lma:tesg-esg-time-extended-world-equivalence-static}, $w, z_n \models \alpha$ iff $w, z_{t,n} \models \alpha$.
          Thus, $\la z_n, \delta_n \ra \in \final$ iff $\la z_{t,n}, \delta_n \ra \in \final[w_t]$.
        \item For all other cases, the claim follows by induction and from the fact that the final configurations of \esg and \tesg are defined in the same way.
          \qedhere
      \end{itemize}
  \end{enumerate}
\end{proofE}

Combining these results, if $w_t$ is the time-extended world of some $w \in \worlds_{\esg}$ and $z$ is the symbol trace of some $z_t \in \traces_{\tesg}$, then both worlds satisfy the same formulas after $z$ or $z_t$ respectively:
\begin{lemmaE}\label{lma:tesg-esg-time-extended-world-equivalence}
  Let $w \in \worlds_{\esg}$ and $w_t \in \worlds_{\tesg}$ the corresponding time-extended world.
  Let $\alpha$ be a sentence of \esg.
  Then for every timed trace $z_t \in \traces_{\tesg}$ and untimed trace $z \in \traces_{\esg}$ with $z = \untimed{z_t}$, the following holds:
  \[
    w, z \models_{\esg} \alpha \text{ iff } w_t, z_t \models_{\tesg} \alpha
  \]
\end{lemmaE}
\begin{proofE}
  By structural induction on $\alpha$:
  \begin{itemize}
    \item For static $\alpha$, the claim was already shown in \autoref{lma:tesg-esg-time-extended-world-equivalence-static}.
    \item Let $\alpha = \square \beta$.
      \\
      \textbf{ $\Rightarrow$:}
      By contraposition.
      Assume $w_t, z_t \not\models \square \beta$.
      Thus, there is a $z_t'$ such that $w_t, z_t \cdot z_t' \not\models \alpha$.
      By induction, $w, \untimed{z_t \cdot z_t'} \not\models \alpha$.
      Therefore, $w, z_t \not\models \square \alpha$.
      \\
      \textbf{ $\Leftarrow$:}
      By contraposition.
      Assume $w, z \not\models \square \beta$.
      It follows that there is a $z'$ such that $w, z \cdot z' \not \models \beta$.
      Let $z_t' \in \traces_{\tesg}$ be any \esg trace with $\untimed{z_t'} = z'$
      By induction, $w_t, z_t \cdot z_t' \not \models \beta$ and therefore $w_t, z_t \not\models \square \beta$.
    \item Let $\alpha = \lbrack \delta \rbrack \beta$.
      \\
      \textbf{ $\Rightarrow$:}
      By contraposition.
      Assume $w_t, z_t \not\models \lbrack \delta \rbrack \beta$.
      Let $z_t' \in \|\delta\|^{z_t}_{w_t}$ such that $w_t, z_t \cdot z_t' \models \neg \beta$.
      There is a finite sequence of transitions $\la z_t, \delta \ra \warrow[w_t] \la z_{t,1}, \delta_1 \ra \warrow[w_t] \ldots \warrow[w_t] \la z_t', \delta' \ra$ such that $\la z_t', \delta' \ra \in \final[w_t]$.
      By \autoref{lma:tesg-esg-same-traces}, there is a finite sequence of transitions $\la z, \delta \ra \warrow \la z_1, \delta_1 \ra \warrow \ldots \warrow \la z', \delta' \ra$ starting in $\la z, \delta \ra$ such that $\untimed{z_{t,i}} = z_i$ and $\la z', \delta' \ra \in \final$.
      By induction, $w, z, z' \models \neg\beta$ and so $w, z \not\models \lbrack \delta \rbrack \beta$.
      \\
      \textbf{ $\Leftarrow$:}
      By contraposition.
      Assume $w, z \not\models \lbrack \delta \rbrack \beta$.
      Let $z' \in \|\delta\|^z_w$ such that $w, z \cdot z' \models \neg \beta$.
      There is a finite sequence of transitions $\la z, \delta \ra \warrow \la z_1, \delta_1 \ra \warrow \ldots \warrow \la z', \delta' \ra$ such that $\la z', \delta' \ra \in \final$.
      By \autoref{lma:tesg-esg-same-traces}, there is a finite sequence of transitions $\la z_t, \delta \ra \warrow[w_t] \la z_{t,1}, \delta_1 \ra \warrow[w_t] \ldots \warrow[w_t] \la z_t', \delta' \ra$ starting in $\la z_t, \delta \ra$ such that $\untimed{z_{t,i}} = z_i$ and $\la z_t', \delta' \ra \in \final[z_t]$.
      By induction, $w_t, z_t \cdot z_t' \models \neg\beta$ and so $w_t, z_t \not\models \lbrack \delta \rbrack \beta$.
    \item Let $\alpha = \llbracket \delta \rrbracket \phi$.
      \\
      We first show by structural sub-induction on $\phi$ that for every $z_t'$ and $\tau_t'$ with $\tau = z_t' \cdot \tau_t'$,
      it follows that $w_t, z_t \cdot z_t', \tau_t' \models \phi$ iff $w, z \cdot z', \tau' \models \phi$, where $z' = \sym(z_t')$ and $\tau' = \sym(\tau_t')$.
      The only interesting case is $\phi = \psi_1 \until{} \psi_2$.
      \\
      \textbf{ $\Rightarrow$:}
      By contraposition.
      Assume $w_t, z_t, \tau_t \not\models \psi_1 \until{} \psi_2$.
      We have two cases:
      \begin{enumerate}
        \item
          For every $z_{t,1}$ and $\tau_t''$ with $\tau_t' = z_{t,1} \cdot \tau''$, we have $w_t, z_t \cdot z_t' \cdot z_{t,1}, \tau_t'' \models \neg \psi_2$.
          By sub-induction, $w, z \cdot z' \cdot \sym(z_{t,1}), \tau'' \models \neg \psi_2$ and so $w, z \cdot z', \tau' \not\models \psi_1 \until{} \psi_2$.
        \item There is a $z_{t,1}$ and $\tau_t''$ with $\tau_t' = z_{t,1} \cdot \tau''$ and $w_t, z_t \cdot z_t' \cdot z_{t,1}, \tau_t'' \models \psi_2$.
          However, there is some $z_{t,2}, z_{t,3}$ with $z_{t,1} = z_{t,2} \cdot z_{t,3}$  and such that $w_t, z_t \cdot z_t' \cdot z_{t,2}, z_{t,3} \cdot \tau_t'' \not\models \psi_1$.
          By sub-induction, $w, z \cdot z' \cdot \sym(z_{t,2}), \sym(z_{t,3}) \cdot \tau'' \not\models \psi_1$ and so $w,  z \cdot z', \tau' \not\models \psi_1 \until{} \psi_2$.
      \end{enumerate}
      \textbf{ $\Leftarrow$:}
      Assume $w_t, z_t, \tau_t \models \psi_1 \until{} \psi_2$.
      Then:
      \begin{enumerate}
        \item There is a $z_{t,1}$ and $\tau_t''$ with $\tau_t' = z_{t,1} \cdot \tau''$ and $w_t, z_t \cdot z_t' \cdot z_{t,1}, \tau_t'' \models \psi_2$.
          By sub-induction, $w, z \cdot z' \cdot \sym(z_{t,1}), \sym(\tau_t'') \models \psi_2$.
        \item For every $z_{t,2}, z_{t,3}$ with $z_{t,1} = z_{t,2} \cdot z_{t,3}$, it follows that $w_t, z_t \cdot z_t' \cdot z_{t,2}, z_{t,3} \cdot \tau_t'' \models \psi_1$.
          By sub-induction, for every such $z_{t,2}$ and $z_{t,3}$, it follows that $w, z \cdot z' \cdot \sym(z_{t,2}), \sym(z_{t,3} \cdot \tau_t'') \models \psi_1$.
      \end{enumerate}
      Therefore, $w, z, \tau \models \psi_1 \until{} \psi_2$.

      It remains to be shown that for every $\tau_t \in \alltraces_{\tesg}$ and $\tau \in \alltraces_{\esg}$ with $\sym(\tau_t) = \tau$, it holds that $\tau_t \in \| \delta \|^{z_t}_{w_t}$ iff $\tau \in \| \delta \|^z_w$.
      \\
      \textbf{ $\Rightarrow$:}
      By contradiction.
      Suppose $\tau_t \in \| \delta \|^{z_t}_{w_t}$ but $\tau \not\in \| \delta \|^z_w$.
      We consider two cases:
      \begin{enumerate}
        \item The trace $\tau$ and therefore also $\tau_t$ is finite.
          But from $\tau_t \in \| \delta \|^{z_t}_{w_t}$, it follows that there is a finite number of transitions $\la z_t, \delta \ra \warrow[w_t] \ldots \warrow[w_t] \la z_t \cdot \tau_t, \delta' \ra$ such that $\la z_t \cdot \tau_t, \delta' \ra \in \final[w_t]$.
          But then, by \autoref{lma:tesg-esg-same-traces}, there is a finite number of transitions $\la z, \delta \ra \warrow[w] \ldots \warrow[w] \la z \cdot \tau, \delta' \ra$ such that $\la z \cdot \tau, \delta' \ra \in \final[w]$ and so $\tau \in \| \delta \|^z_w$.
        \item The trace $\tau$ and therefore also $\tau_t$ is infinite.
          Let $\tau_t = (a_1, t_1) (a_2, t_2) \cdots$ and for every $n \in \naturals$, let $z_{t,n} = (a_1, t_1) \cdots (a_n, t_n)$ and let $z_n = \sym(z_{t,n})$.
          There are again two cases:
          \begin{enumerate}
            \item For some $i \in \naturals$, we have  $\la z, \delta \ra \warrow[w] \ldots \warrow[w] \la z \cdot z_i, \delta_i \ra$, but there is no $\delta_{i+1}$ with $\la z_i, \delta_i \ra \warrow[w] \la z_{i+1}, \delta_{i+1} \ra$, i.e., there is no possible transition after $i$ steps that agrees with $\tau$.
              However, as $\tau_t \in \| \delta \|^{z_t}_{w_t}$, we have $\la z_t, \delta \ra \warrow[w_t] \ldots \warrow[w_t] \la z_t \cdot z_{t,i}, \delta_i \ra \warrow[w_t] \la z_{t,i+1}, \delta_{i+1} \ra$.
              With  \autoref{lma:tesg-esg-same-traces}, it follows that $\la z, \delta \ra \warrow[w] \ldots \warrow[w] \la z \cdot z_i, \delta_i \ra \warrow[w] \la z \cdot z_{i+1}, \delta_{i+1} \ra$, leading to a contradiction.
            \item For some $i \in \naturals$, we have $\la z, \delta \ra \warrow[w] \ldots \warrow[w] \la z \cdot z_i, \delta_i \ra$ and $\la  z \cdot z_i, \delta_i \ra \in \final$.
              By \autoref{lma:tesg-esg-same-traces}, $\la z_t, \delta \ra \warrow[w_t] \ldots \warrow[w_t] \la z_t \cdot z_{t,i}, \delta_i \ra$ and $\la z_t \cdot z_{t,i}, \delta_i \ra \in \final[w_t]$.
              However, as $\tau_t$ is infinite, by \autoref{def:program-trace}, $\tau_t \not\in \| \delta \|^{z_t}_{w_t}$, in contradiction to the assumption.
          \end{enumerate}
      \end{enumerate}
      \textbf{ $\Leftarrow$:}
      By contradiction.
      Suppose $\tau \in \| \delta \|^z_w$ but $\tau_t \not\in \| \delta \|^{z_t}_{w_t}$.
      We consider two cases:
      \begin{enumerate}
        \item The trace $\tau_t$ and therefore also $\tau$ is finite.
          But from $\tau \in \| \delta \|^z_w$, it follows that there is a finite number of transitions $\la z, \delta \ra \warrow[w] \ldots \warrow[w] \la z \cdot \tau, \delta' \ra$ such that $\la z \cdot \tau, \delta' \ra \in \final[w]$.
          But then, by \autoref{lma:tesg-esg-same-traces}, there is a finite number of transitions $\la z_t, \delta \ra \warrow[w_t] \ldots \warrow[w_t] \la z_t \cdot \tau_t, \delta' \ra$ such that $\la z_t \cdot \tau_t, \delta' \ra \in \final[w_t]$ and so $\tau_t \in \| \delta \|^{z_t}_{w_t}$.
        \item The trace $\tau_t$ and therefore also $\tau$ is infinite.
          Let $\tau_t = (a_1, t_1) (a_2, t_2) \cdots$ and for every $n \in \naturals$, let $z_{t,n} = (a_1, t_1) \cdots (a_n, t_n)$ and let $z_n = \sym(z_{t,n})$.
          There are again two cases:
          \begin{enumerate}
            \item For some $i \in \naturals$, we have  $\la z_t, \delta \ra \warrow[w_t] \ldots \warrow[w_t] \la z_t \cdot z_{t,i}, \delta_i \ra$, but there is no $\delta_{i+1}$ with $\la z_{t,i}, \delta_i \ra \warrow[w_t] \la z_{t,i+1}, \delta_{i+1} \ra$, i.e., there is no possible transition after $i$ steps that agrees with $\tau_t$.
              However, as $\tau \in \| \delta \|^{z}_{w}$, we have $\la z, \delta \ra \warrow[w] \ldots \warrow[w] \la z \cdot z_{i}, \delta_i \ra \warrow[w] \la z_{i+1}, \delta_{i+1} \ra$.
              With  \autoref{lma:tesg-esg-same-traces}, it follows that $\la z_t, \delta \ra \warrow[w_t] \ldots \warrow[w_t] \la z_t \cdot z_{t,i}, \delta_i \ra \warrow[w_t] \la z_t \cdot z_{t,i+1}, \delta_{i+1} \ra$, leading to a contradiction.
            \item For some $i \in \naturals$, we have $\la z_t, \delta \ra \warrow[w_t] \ldots \warrow[w_t] \la z_t \cdot z_{t,i}, \delta_i \ra$ and $\la  z_t \cdot z_{t,i}, \delta_i \ra \in \final[w_t]$.
              By \autoref{lma:tesg-esg-same-traces}, $\la z, \delta \ra \warrow[w] \ldots \warrow[w] \la z \cdot z_{i}, \delta_i \ra$ and $\la z \cdot z_{i}, \delta_i \ra \in \final[w]$.
              However, as $\tau$ is infinite, by \autoref{def:program-trace}, $\tau \not\in \| \delta \|^{z}_{w}$, in contradiction to the assumption.
          \end{enumerate}
      \end{enumerate}
      Summarizing, for every $\tau_t \in \alltraces_{\tesg}$ and $\tau \in \alltraces_{\esg}$ with $\sym(\tau_t) = \tau$, it holds that $\tau_t \in \| \delta \|^{z_t}_{w_t}$ iff $\tau \in \| \delta \|^z_w$.
      Also, for every such $\tau$ and $\tau_t$, we have shown that $w_t, z_t, \tau_t \models \phi$ iff $w, z, \tau \models \phi$.
      Therefore, $w_t, z_t \models \llbracket \delta \rrbracket \phi$ iff $w, z \models \llbracket \delta \rrbracket \phi$.
      \qedhere
  \end{itemize}
\end{proofE}
Note that restricting $w_t$ to be a time-extended world is a real restriction, as not every world $w \in \worlds_{\tesg}$ is a time-extended world of some $w \in \worlds_{\esg}$.

If we only consider fluent formulas and therefore only consider the initial situation, then the valid sentences of both logics are the same:
\begin{theoremE}\label{thm:tesg-esg-static-formula-equivalence}
  For every fluent sentence $\alpha$ of \esg:
  \[
    \models_{\esg} \alpha \text{ iff } \models_{\tesg} \alpha
  \]
\end{theoremE}
\begin{proofE}
  \begin{description}
    \item[$\mathbf{ \Rightarrow }$:]
      By contraposition.
      Assume $\not\models_{\tesg} \alpha$, so there is a world $w_t \in \worlds_{\tesg}$ with $w_t \not\models_{\tesg} \alpha$.
      Note that for static situation formulas $\alpha$, the truth of $\alpha$ does not depend on any future states.
      Thus, wlog, for every  $z \neq \la\ra$, assume that $w_t[P(\vec{n}), z] = 0$ for arbitrary primitive formulas $P(\vec{n})$.
      Furthermore, for every  $z \neq \la\ra$, assume that $w_t[f(\vec{n}), z] = n_f$ for arbitrary primitive terms $f(\vec{n})$ and where $n_f$ is some standard name of the right sort.
      Now, let $w \in \worlds_{\esg}$ be a world such that for every $z_t \in \traces_{\tesg}$, $w[P(\vec{n}), \untimed{z_t}] = w_t[P(\vec{n}), z_t]$ and $w[f(\vec{n}), \untimed{z_t}] = w[f(\vec{n}), z_t]$.
      Clearly, $w_t$ is the time-extended world of $w$.
      By \autoref{lma:tesg-esg-time-extended-world-equivalence}, it follows that $w \not\models_{\esg} \alpha$.
    \item[$\mathbf{ \Leftarrow }$:]
      By contraposition.
      Assume $\not\models_{\esg} \alpha$, so there is a world $w \in \worlds_{\esg}$ with $w \not\models_{\esg} \alpha$.
      Let $w_t$ be the time-extended world of $w$.
      By \autoref{lma:tesg-esg-time-extended-world-equivalence}, $w_t \not\models_{\tesg} \alpha$.
      \qedhere
  \end{description}
\end{proofE}


More generally, if we consider arbitrary sentences that may include programs and trace formulas, we can show that every valid sentence of \tesg is also a valid sentence of \esg:
\begin{theoremE}\label{thm:tesg-validity-entails-esg-validity}
  Let $\alpha$ be a sentence of \esg.
  If $\models_{\tesg} \alpha$, then also $\models_{\esg} \alpha$.
\end{theoremE}
\begin{proofE}
  By contraposition.
  Assume $\not\models_{\esg} \alpha$.
  We show that $\not\models_{\tesg} \alpha$.
  As $\not\models_{\esg} \alpha$, there is an \esg world $w$ such that $w \not\models_{\esg} \alpha$.
  Let $w_t$ be the time-extended world of $w$.
  By \autoref{lma:tesg-esg-time-extended-world-equivalence}, $w_t \not \models \alpha$, and therefore $\not\models_{\tesg} \alpha$.
\end{proofE}
However, the other direction is not true.
There are valid sentences of \esg that are not valid in \tesg:
\begin{theoremE}[][normal]\label{thm:valid-esg-does-not-entail-valid-tesg}
  Let $\alpha$ be a sentence of \esg.
  From $\models_{\esg} \alpha$, it does not follow that $\models_{\tesg} \alpha$.
\end{theoremE}
\begin{proofE}[normal]
  By counter example.
  Let $p$ be an action standard name and $F$ a fluent predicate symbol of arity $0$.
  Let $\alpha = [p] F \vee [p] \neg F$.

  We first show that $\models_{\esg} \alpha$:
  Let $w$ be an arbitrary $\esg$ world.
  Clearly, $\|p\|_w = \{ \la p \ra \}$.
  By definition of world $w$, we have two cases:
  \begin{enumerate}
    \item $w[F, \la p \ra] = 1$.
      Then $w \models [p] F$ and therefore $w \models \alpha$.
    \item $w[F, \la p \ra] = 0$.
      Then $w \models [p] \neg F$ and therefore $w \models \alpha$.
  \end{enumerate}
  Thus, for an arbitrary \esg world $w$, it follows that $w \models \alpha$ and therefore $\models_{\esg} \alpha$.
  Now, we show that $\not\models_{\tesg} \alpha$.
  The idea here is that in \tesg, the truth value of a fluent may depend on time because a world assigns a truth value to a fluent for each possible timed trace.
  Thus, we may define a world where neither of the disjuncts is necessarily true.
  Let $w$ be a \tesg world
  such that
  \begin{align*}
    w[F, \underbrace{\la 0, p \ra}_{=: z_0}] &= 0
    \\
    w[F, \underbrace{\la 1, p \ra}_{=: z_1}] &= 1
  \end{align*}
  Clearly, $\{ z_0, z_1 \} \subseteq \|p\|_w$.
  From $w[F, \la 0, p \ra] = 0$ it follows that $w, z_0 \not\models F$.
  At the same time, from $w[F, \la 1, p \ra] = 1$, it follows that $w, z_1 \not\models \neg F$.
  Thus, $w \not\models \alpha$ and therefore $\not\models_{\tesg} \alpha$.
\end{proofE}

In a sense, this is a negative result, as we cannot directly use existing methods for \esg and apply them to problems in \tesg.
However, as we will see in the next section, this changes if we consider \aclp{BAT}.

\subsection{\esg Basic Action Theories in \tesg}%
\label{ssub:tesg-esg-bats}

We have seen that there are valid sentences of \esg that are not valid in \tesg.
In this section, we consider \iac{BAT} and investigate which sentences are entailed by a \ac{BAT}.
We first show that for a \tesg \ac{BAT}, if two traces only differ in the time points but contain the same action steps, then they entail the same static time-invariant formulas:

\begin{lemmaE}\label{lma:tesg-bat-static-time-invariance}
  Let $\bat$ be a \tesg \ac{BAT} over $(\fluentset, \clockset)$ and let $w \in \worlds_{\tesg}$ such that $w \models \bat$.
  Let $t$ be a term only mentioning fluent function symbols from $\fluents$ and let $\alpha$ be a static and time-invariant sentence over $\fluents$.
  For every pair of traces $z, z' \in \traces_{\tesg}$ with $\sym(z) = \sym(z')$, the following holds:
  \begin{enumerate}
    \item $\lvert t \rvert^z_w = \lvert t \rvert^{z'}_w$
    \item $w, z \models \alpha \text{ iff } w, z' \models \alpha$
  \end{enumerate}
\end{lemmaE}
\begin{proofE}
  First, let $\alpha'$ be the formula obtained from $\alpha$ by replacing each occurrence of $\poss(t)$ with $\pi_a\vert^a_t$.
  This is possible because $\pi_a$ is a fluent situation formula and therefore may not mention $\poss$.
  Also, because $w \models \bat$ and therefore $w \models \square \poss(a) \equiv \pi_a$, it is clear that $w, z \models \alpha'$ iff  $w, z \models \alpha$ and similarly, $w, z' \models \alpha'$ iff $w,  z' \models \alpha$.

  Let $l = \lvert z \rvert = \lvert z' \rvert$ be the length of $z$ and $z'$.
  For every $i \leq l$, let $z^{(i)}$ ($z'^{(i)}$) denote the prefix of $z$ ($z'$ respectively) with length $i$.
  We show both claims by induction on the length $l$.
  \\
  \textbf{Base case.}
  Let $l = 0$ and thus $z = z' = \la\ra$.
  As $z = z'$, both claims immediately follow.
  \\
  \textbf{Induction step.}
  Assume $\lvert z \rvert = \lvert z' \rvert = l + 1$.
  \begin{enumerate}
    \item
      We first show for each term $t$ that $\lvert t \rvert^z_w = \lvert t \rvert^{z'}_w$ by structural sub-induction on $t$:
      \begin{itemize}
        \item Let $t = n$ for some standard name $n \in \stdname$.
          Clearly, $\lvert t \rvert^z_w = \lvert t \rvert^{z'}_w = n$.
        \item Let $t = f(n_1, \ldots, n_k)$ for some rigid function symbol $f$.
          As $f$ is rigid, it immediately follows that $\lvert t \rvert^z_w = \lvert t \rvert^{z'}_w$.
        \item Let $t = f(t_1, \ldots, t_k)$ for some fluent function symbol from $\fluents$.
          By sub-induction, $\lvert t_i \rvert^z_w = \lvert t_i \rvert^{z'}_w$ for each $t_i$.
          There must be a \ac{SSA} for $f$ of the form $\square [a] f(\vec{x}) = y \equiv \gamma_f(\vec{x}, y)$, where $\gamma_f(\vec{x}, y)$ is a fluent situation formula.
          It follows by induction that $w, z^{(l)} \models \gamma_f(t_1, \ldots, t_k, y)$ iff $w, z'^{(l)} \models \gamma_f(t_1, \ldots, t_k, y)$.
          Thus, $w, z \models f(t_1, \ldots, t_k) = y$ iff $w, z' \models f(t_1, \ldots, t_k) = y$.
          Therefore, $\lvert f(t_1, \ldots, t_k) \rvert^z_w = \lvert f(t_1, \ldots, t_k) \rvert^{z'}_w$.
      \end{itemize}
\item We show by structural sub-induction on $\alpha'$ that $w, z \models \alpha' \text{ iff } w, z' \models \alpha'$:
  \begin{itemize}
    \item Let $\alpha' = F(t_1, \ldots, t_k)$, where $F \in \fluents$ is a $k$-ary fluent predicate symbol.
      From above, it follows for each $i$ that $\lvert t_i \rvert^z_w = \lvert t_i \rvert^{z'}_w$.
      \Wolog, $\lvert t_i \rvert^z_w = n_i$.
      By \autoref{def:bat}, there is a \ac{SSA} for $F$ of the form $\square [a] F(\vec{x}) \equivspace \gamma_F(\vec{x})$, where $\gamma_F(\vec{x})$ is a fluent situation formula.
      By induction, $w, z^{(l)} \models \gamma_F(n_1, \ldots, n_k)$ iff $w, z'^{(l)} \models \gamma_F(n_1, \ldots, n_k)$
      Therefore, $w, z \models \alpha'$ iff $w, z' \models \alpha'$.
    \item Let $\alpha' = (t_1 = t_2)$.
      It follows from the above that $\lvert t_1 \rvert^z_w = \lvert t_1 \rvert^{z'}_w$ and $\lvert t_2 \rvert^z_w = \lvert t_2 \rvert^{z'}_w$.
      Thus, $w, z \models \alpha'$ iff $w, z' \models \alpha'$.
    \item Let $\alpha' = \beta_1 \wedge \beta_2$.
      By sub-induction, $w, z \models \beta_1$ iff $w, z' \models \beta_2$ and $w, z \models \beta_2$ iff $w, z' \models \beta_2$.
      Thus, $w, z \models \alpha'$ iff $w, z' \models \alpha'$.
    \item Let $\alpha' = \neg \beta$.
      By sub-induction, $w, z \not\models \beta$ iff $w, z' \not\models \beta$.
      Thus, $w, z \models \alpha'$ iff $w, z' \models \alpha'$.
    \item Let $\alpha' = \forall x.\, \beta$.
      By sub-induction, for each standard name $n \in \stdname_x$ of the corresponding type, $w, z \models \beta^x_n$ iff $w, z' \models \beta^x_n$.
      Thus, $w, z \models \alpha'$ iff $w, z' \models \alpha'$.
  \end{itemize}
  \end{enumerate}
  Therefore, $w, z \models \alpha'$ iff $w, z' \models \alpha'$ and hence also $w, z \models \alpha$ iff $w, z' \models \alpha$.
\end{proofE}

Intuitively, this is true because action effects are time-independent, as clock formulas may only be used in clock constraints for actions.

Now, we consider an \esg \ac{BAT} $\bat$.
First, note that we can extend $\bat$ to a \tesg \ac{BAT} $\bat'$ by adding the (vacuous) clock constraint axiom $\square \clockconstraint(a) \equiv \top$.\footnote{
  The additional axiom is necessary as every \tesg \ac{BAT} must contain a clock constraint axiom.
  However, as we excluded the distinguished symbol $\clockconstraint$ from the language of \esg, the axiom does not have any effect.
}
We can now show that such a \ac{BAT} entails the  same formulas in \esg and \tesg:
\begin{theoremE}\label{thm:tesg-esg-bat-equivalence}
  Let $\Sigma$ be an \esg \ac{BAT}, $\bat' = \bat \cup \{ \square \clockconstraint(a) \equiv \top \}$ the corresponding \tesg \ac{BAT}, and $\alpha$ a sentence of \esg.
  Then the following holds:
  \[
    \Sigma \models_{\esg} \alpha \text{ iff } \bat' \models_{\tesg} \alpha
  \]
\end{theoremE}
\begin{proofE}
  ~
  \\
  \textbf{ $\Rightarrow$: }
  By contraposition.
  Assume $\alpha$ is an \esg sentence such that $\bat' \not\models_{\tesg} \alpha$.
  Thus, there is a world $w_t \in \worlds_{\tesg}$ with $w_t \models \bat'$ but $w_t \not\models \alpha$.
  We show that $\bat \not\models_{\esg} \alpha$.
  As $w_t \models \bat'$, it follows with \autoref{lma:tesg-bat-static-time-invariance} for every pair of traces $z, z'$ with $\sym(z) = \sym(z')$, every $k$-ary relational fluent symbol $F \in \fluents$, and every $k$-ary relational functional symbol $f \in \fluents$:
  \begin{align*}
    w_t[F(n_1, \ldots, n_k), z] &= w_t[F(n_1, \ldots, n_k), z']
    \\
    w_t[f(n_1, \ldots, n_k), z] &= w_t[f(n_1, \ldots, n_k), z']
  \end{align*}
  Thus, $w$ is the time-extended world of some $w \in \worlds_{\esg}$.
  By \autoref{lma:tesg-esg-time-extended-world-equivalence}, it follows that $w \models \bat$ but $w \not\models \alpha$.
  Thus, $\bat \not\models \alpha$.
  \\
  \textbf{ $\Leftarrow$: }
  Let $\bat' \models_{\tesg} \alpha$.
  Note that $\bat = \{ \sigma_1, \sigma_2, \ldots, \sigma_n \}$ is a finite set of sentences.
  Thus, $\bigwedge \bat = \sigma_1 \wedge \sigma_2 \wedge \ldots \wedge \sigma_n$ is a sentence of $\esg$.
  It follows that $\models_{\tesg} \bigwedge \bat \supset \alpha$.
  By \autoref{thm:tesg-validity-entails-esg-validity}, $\models_{\esg} \bigwedge \bat \supset \alpha$ and therefore $\bat \models_{\esg} \alpha$.
\end{proofE}

The idea here is the same as in \autoref{lma:tesg-bat-static-time-invariance}: The \ac{BAT} uniquely defines the effects of each action, which are independent of time.
Similarly, the action precondition may not depend on time and the \ac{BAT} may not contain any clock constraints.

Therefore, as long as we reason about a \ac{BAT}, we may use previously established results about \esg \acp{BAT} and apply them to \tesg.
This becomes even more relevant with the following result that relates \esg to \es:
\begin{theoremE}[\parencite{classenLogicNonterminatingGolog2008}][normal]
  Let $\alpha$ be a sentence of \es without epistemic operators.
  Then $\models_{\es} \alpha$ iff $\models_{\esg} \alpha$.
\end{theoremE}

It immediately follows:
\begin{corollary}\label{thm:es-tesg-bat-equivalence}
  Let $\Sigma$ be an \es \ac{BAT}, $\bat' = \bat \cup \{ \square \clockconstraint(a) \equiv \top \}$,  and $\alpha$ be a sentence of \esg.
  Then the following holds:
  \[
    \bat \models_{\es} \alpha \text{ iff } \bat' \models_{\tesg} \alpha
  \]
\end{corollary}

Hence, previously established methods on the objective fragment of \es may directly be applied in \tesg.
As an example, \textcite{classenSemanticsADLProgression2006} have provided a semantics for task planning based on \es, which allows to use a PDDL planner in \golog.
With \autoref{thm:es-tesg-bat-equivalence}, we may also use the same semantics for planning in \tesg and therefore incorporate a planner in a \tesg program.

\section{\acl*{MTL} and \tesg}%
\label{sec:tesg-mtl}

In the previous section, we have compared \tesg and \esg, which focused on the comparison of time-invariant properties, as \esg does not include metric time in the logic.
To complete the picture, we now compare \tesg to \ac{MTL} and therefore investigate timing properties in \tesg.

Similarly to the above, we first need to translate between the two logics.
More specifically, given a finite set of fluents \fluents, we provide a translation from a \tesg trace to a timed word of \ac{MTL} for a fixed world $w$ of \tesg:
\begin{definition}\label{def:rho-i}
  Let $w \in \worlds_{\tesg}$ and $\tau \in \traces$ with $\tau = (p_1, t_1) (p_2, t_2) \cdots$.
  The \emph{timed word $\rho$ corresponding to $(w, \tau)$} is a timed word of \ac{MTL} such that for each $i$, $\rho_i$ is defined as follows:
  \[
    \rho_i \eqdef \{ F(\vec{n}) \in \primformulas \mid w[F(\vec{n}), \tau^{(i)}] = 1 \}
    \qedhere
  \]
\end{definition}

We can show a connection of \tesg and \ac{MTL} with respect to such a fixed world and trace:
\begin{lemmaE}\label{lma:mtl-equivalence}
  Let $\phi$ be an \ac{MTL} sentence over alphabet $\fluents^0$.
  Let $\tau \in \traces$ be a (possibly infinite) trace, $z^{(i)}$ be the prefix of $\tau$ with length $i$, and $\tau^{(i)}$ be the suffix of $\tau$ such that $\tau = z^{(i)} \cdot \tau^{(i)}$.
  Let $w \in \worlds$ be a world of \tesg and $\rho$ be the timed word corresponding to $(w, \tau)$.
  Then the following holds for each $i \in \naturals_0$:
  \[
    w, z^{(i)}, \tau^{(i)} \models_{\tesg} \phi \text{ iff } \rho, i \models_{\mathrm{MTL}} \phi
  \]
\end{lemmaE}
\begin{proofE}
  By structural induction on $\phi$:
  \begin{itemize}
    \item Let $\phi = F$ with $F \in \mathcal{F}^0$.
      \begin{align*}
        &w, z^{(i)}, \tau^{(i)} \models F
        \\
        \lrs &w[F, z^{(i)}] = 1 & \text{ by \autoref{def:tesg-truth}}
        \\
        \lrs &F \in \rho_i & \text{ by \autoref{def:rho-i}}
        \\
        \lrs &\rho, i \models \phi & \text{ by \autoref{def:mtl-semantics}}
      \end{align*}
    \item Let $\phi = \neg \psi$.
      \begin{align*}
        & w, z^{(i)}, \tau^{(i)} \models \phi
        \\
        \lrs & w, z^{(i)}, \tau^{(i)} \not\models \psi & \text{ by \autoref{def:tesg-truth} }
        \\
        \lrs & \rho, i \not\models \psi & \text{ by induction }
        \\
        \lrs & \rho, i \models \phi & \text{ by \autoref{def:mtl-semantics} }
      \end{align*}
    \item Let $\phi = \psi_1 \wedge \psi_2$.
      \begin{align*}
        & w, z^{(i)}, \tau^{(i)} \models \phi
        \\
        \lrs & w, z^{(i)}, \tau^{(i)} \models \psi_1 &\text { and } &w, z^{(i)}, \tau^{(i)} \models \psi_2 & \text{ by \autoref{def:tesg-truth} }
        \\
        \lrs & \rho, i \models \psi_1 &\text{ and } &\rho, i \models \psi_2 & \text{ by induction }
        \\
        \lrs & \rho, i \models \phi & & & \text{ by \autoref{def:mtl-semantics} }
      \end{align*}
    \item Let $\phi = \psi_1 \until{I} \psi_2$.
      \\
      \textbf{ $\Rightarrow$: }
      Assume $w, z^{(i)}, \tau^{(i)} \models \phi$ and thus $w, z^{(i)}, \tau^{(i)} \models \psi_1 \until{I} \psi_2$.
      Then, there is a $\tau' \in \inftraces$ and $z_1 \in \traces$ with $z_1 = (t_{i+1}, p_{i+1}) \cdots (t_k, p_k) \neq \la\ra$ such that
      \begin{enumerate}
        \item $\tau^{(i)} = z_1 \cdot \tau'$,
        \item $w, z^{(i)} \cdot z_1, \tau' \models \psi_2$,
        \item $\ztime(z_1)  \in \ztime(z^{(i)}) + I$,
        \item for each $z_2, z_3$ with $z_2 = (t_{i+1}, p_{i+1}) \cdots (t_m, p_m)$, $m < k$, and $z_1 = z_2 \cdot z_3$:
          \\
          $w, z^{(i)} \cdot z_2, z_3 \cdot \tau' \models \psi_1$ and thus $w, z^{(m)}, \tau^{(m)} \models \psi_1$ for each $m$ with $i < m < k$.
      \end{enumerate}
      Let $j = k$.
      Note that $z^{(i)} \cdot z_1 = z^{(k)}$ and $\tau' = \tau^{(k)}$.
      It follows:
      \begin{enumerate}
        \item $i < j < \lvert \rho \rvert$, because $k > i$ and $\lvert \rho \rvert > j$ by definition of $\rho$;
        \item from $w, z^{(k)}, \tau^{(k)} \models \psi_2$, it follows by induction that $\rho, j \models \psi_2$;
        \item $\ztime(z_1) = t_k$ and $\ztime(z^{(i)}) = t_i$ and therefore,
          from $\ztime(z_1) \in \ztime(z^{(i)}) + I$, it follows that $t_k \in t_i + I$ and thus $t_j - t_i \in I$;
        \item  $w, z^{(m)}, \tau^{(m)} \models \psi_1$ for each $m$ with $i < m < k$ and thus, by induction, for each $m$ with $i < m < j$, $\rho, m \models \psi_1$.
      \end{enumerate}
      Thus: $\rho, i \models \phi$.
      \\
      \textbf{ $\Leftarrow$: }
      Assume $\rho, i \models \phi$.
      Thus, there is a $j$ such that
      \begin{enumerate}
        \item $i < j < \lvert \rho \rvert$,
        \item $\rho, j \models \psi_2$,
        \item $t_j - t_i \in I$,
        \item for each $m$ with $i < m < j$: $\rho, m \models \psi_1$.
      \end{enumerate}
      Let $z_1 = \left(t_{i+1}, p_{i+1}\right) \cdots \left(t_j, p_j\right)$ and let $\tau' = \tau^{(j)}$.
      It follows:
      \begin{enumerate}
        \item $\tau^{(i)} = z_1 \cdot \tau'$,
        \item from $\rho, j \models \psi_2$, it follows by induction that $w, z \cdot z_1, \tau' \models \psi_2$,
        \item $\ztime(z_1) = t_j$ and $\ztime(z^{(i)}) = t_i$ and therefore, from $t_j - t_i \in I$, it follows that $\ztime(z^{(i)}) - \ztime(z_1) \in I$,
        \item $\rho, m \models \psi_1$ for each $m$ with $i < m < j$ and thus, by induction, $w, z^{(m)}, \tau^{(m)} \models \psi_1$.
      \end{enumerate}
      Thus, by \autoref{def:tesg-truth}, $w, z^{(i)}, \tau^{(i)} \models \phi$.
      \qedhere
  \end{itemize}
\end{proofE}

This directly leads to the following result regarding valid sentences of \ac{MTL} and \tesg:
\begin{theoremE}\label{thm:mtl-equivalence}
  For an arbitrary \ac{MTL} sentence $\phi$: 
  \[
    \models_{\mathrm{MTL}} \phi \text{ iff } \models_{\tesg} \phi
  \]
\end{theoremE}
\begin{proofE}
  Without loss of generality, assume that $P \subseteq \mathcal{F}^0$, i.e., each atomic proposition occurring in the \ac{MTL} formula is a $0$-ary fluent of \tesg.
  \\
  \textbf{ $\Rightarrow$: }
  By contraposition.
  Assume $\not\models_{\tesg} \phi$.
  Then there is a world $w$ and a trace $\tau$ such that $w, \la\ra, \tau \not\models_{\tesg} \phi$.
  Let $\rho$ be a timed word as defined in \autoref{def:rho-i}.
  Then, by \autoref{lma:mtl-equivalence}, $\rho \not\models_{\mathrm{MTL}} \phi$.
  \\
  \textbf{ $\Leftarrow$: }
  By contraposition.
  Assume $\not\models_{\mathrm{MTL}} \phi$.
  Then there is a timed word
  \[
    \rho = \left(\rho_0, 0\right)  \left(\rho_1, t_1\right) \left(\rho_2, t_2\right)\cdots
  \]
  such that $\rho \not\models_{\mathrm{MTL}} \phi$.
  Let $a \in \stdactname$ be some action standard name and $\tau = \left(a, t_1\right) \left(a, t_2\right) \cdots$ (i.e., the trace $\tau$ consists of a single repeating action $a \in \stdactname$ and the same time points as $\rho$).
  Let $z^{(i)}$ denote the finite prefix of $\tau$ consisting of $i$ time-action pairs, i.e., $z^{(i)} = \left(a, t_1\right) \cdots \left(a, t_i\right)$.
  Let $w \in \worlds$ such that for each $p \in P$ and each $i \in \naturals_0$, $w[p, z^{(i)}] = 1 \text{ iff } p \in \rho_i$.
  By \autoref{lma:mtl-equivalence}, $w, \la\ra, \tau \not\models_{\tesg} \phi$.
\end{proofE}

Hence, concerning timing properties, \ac{MTL} and \tesg have the same valid sentences.
Therefore, we can apply methods for \ac{MTL} on \tesg problems.
This will become important in \autoref{chap:synthesis}, as we will use the translation of \ac{MTL} to \acp{ATA}, as described in \autoref{sec:mtl}, to check the satisfaction of a trace formula $\phi$.

\section{Timed Automata in \tesg}\label{sec:tesg-ta}

Next, we look at the relationship of \aclp{TA} and \tesg.
Timed automata play an important role as they are one of the most commonly used models for timed systems.
Moreover, as motivated above, we intend to use \aclp{TA} for robot self models to describe the behavior of different components of a robot, e.g., its camera or gripper.
Finally, in \autoref{chap:transformation-as-reachability-problem}, we will use \aclp{TA} to transform an abstract plan into a timed action sequence that is executable on the robot platform.

Given a timed automaton, we can construct a \tesg \ac{BAT} that simulates the automaton:
\begin{definition}[Timed Automaton in \tesg]
Let $\ta = \left(\tastates, \tastate_0, \tastates_F, \taalph, \taclocks, \invs, \taswitches\right)$ be a \ac{TA}.
We assume that for any $\left(l, \sigma, \varphi, Y, l'\right) \in \taswitches$, if $c \in Y$, then $\invs(l')$ does not mention $c$.%
\footnote{
  If $c \in Y$, its value will always be $0$ after the switch.
  Therefore, the invariant cannot guard the incoming transition and we may just move it to all outgoing transitions.
}
We define the corresponding \ac{BAT} $\bat_{\ta}$ over $\left(\{ \taloc, \occ \}, \taclocks \right)$ as follows:
\begin{itemize}
  \item In the initial situation, the \ac{TA} is in the initial location and no action has occurred, i.e., $\bat_0$ is defined as follows:
    \begin{align*}
      &\taloc = l_0
      \\
      &\forall \sigma.\, \neg \occ(\sigma)
    \end{align*}
  \item The switch action is possible if the \ac{TA} is currently in the starting location of the switch:
    \[
      \poss(a) \equivspace \bigvee_{\left(l, \sigma, \varphi, Y, l'\right) \in E} a = s(l, \sigma, \varphi, Y, l') \wedge \taloc = l
    \]
  \item The clock constraint of the switch action makes sure that the clock constraint of the switch is satisfied, as well as that the invariants of both the starting and the target location are satisfied:
    \[
      \clockconstraint(a) \equivspace \bigvee_{\left(l, \sigma, \varphi, Y, l'\right) \in E} a = s(l, \sigma, \varphi, Y, l') \wedge \varphi \wedge I(l) \wedge I(l')
    \]
  \item The switch action changes the location to the target location of the switch, i.e., $\bat_\post$ contains the \ac{SSA}:
    \[
      \square \lbrack a \rbrack loc = l^* \equivspace \bigvee_{\left(l, \sigma, \varphi, Y, l'\right) \in E} a = s(l, \sigma, \varphi, Y, l^*) \wedge l^* = l'
    \]
  \item The switch action also sets $\occ(\sigma)$ for the symbol $\sigma$ that has just occurred, , i.e., $\bat_\post$ contains the \ac{SSA}:
    \[
      \square \lbrack a \rbrack \occ(\sigma^*) \equivspace \bigvee_{\left(l, \sigma, \varphi, Y, l'\right) \in E} a = s(l, \sigma, \varphi, Y, l') \wedge \sigma^* = \sigma
    \]
  \item The switch action resets a clock iff the corresponding \ac{TA} switch resets the clock:
    \[
      \square \lbrack a \rbrack \reset(c) \equivspace \bigvee_{ \left(l, \sigma, \varphi, Y, l'\right) \in E } a = s(l, \sigma, \varphi, Y, l') \wedge \bigvee_{y \in Y} c = y
    \]
\end{itemize}

We simulate the \ac{TA} \ta with the following program:
\[
  \delta_\ta \eqdef \left(\pi a.\, \poss(a) \wedge \clockconstraint(a)?; a\right)^* ;\; \mi{final}(\taloc)?
\]
where
\[
  \mi{final}(\taloc) \eqdef \bigvee_{l \in \tastates_f} \taloc = l
  \qedhere
\]
\end{definition}

\newcommand*{\tabat}{\ensuremath{\bat_{\ta}}\xspace}
\newcommand*{\wta}{\ensuremath{w_{\tabat}}\xspace}

Note that a trace $z \in \| \delta_\ta \|_w$ of the program $\delta_\ta$ consists of \ac{TA} switches rather than labels from the alphabet $\taalph$.
As we want to relate the program traces with timed words accepted by the \ac{TA}, we first need to translate a program trace to a timed word.
We define the \emph{label trace} of a trace $z$ as follows:
\begin{definition}[Label trace]\label{def:label-trace}
  Given a \ac{TA} \ta, a world $w \models \tabat$, and a trace $z = (s_1, t_1)(s_2, t_2) \ldots \in \|\delta_\ta\|_w$.
  The \emph{label trace} $\labeltrace(z)$ of $z$ is the sequence $\labeltrace(z) \eqdef (a_1, t_1)(a_2, t_2) \ldots$ such that for each $i$:
  \[
    a_i =
    \begin{cases}
      a & \text{ if } s_i = (l, a, \clockconstraint, Y, l') \text{ for some switch $(l, a, \clockconstraint, Y, l')$ of \ta }
      \\
      s_i & \text{ otherwise }
    \end{cases}
    \qedhere
  \]
\end{definition}

If some action $s_i$ in the trace $z$ is a switch $(l, a, g, Y, l')$ of the \ac{TA}, then the corresponding symbol in the label trace is the action $a$.
Otherwise, for any other symbol, the symbol remains unchanged.
This will be useful later, as we want to compose a \ac{TA} with another program, where we should only substitute the actions that correspond to \ac{TA} switches.


We can now show that our program indeed allows exactly those finite traces that correspond to a finite timed word accepted by the \ac{TA}:
\begin{theoremE}\label{thm:tesg-ta-equivalence}
  Let \ta be a \ac{TA}, $\tabat$ the corresponding \ac{BAT}, and $w \models \tabat$.
  Then the following holds:
  \[
    \rho \in \lang^*(\ta) \text{ iff } \rho = \labeltrace(z) \text{ for some finite } z \in \| \delta_\ta \|_w
  \]
\end{theoremE}
\begin{proofE}
  Let $\talts$ the \ac{LTS} corresponding to \ta, $l_0 \in \tastates_0$ some initial location of \ta, and $w$ be a world such that $w \models \tabat$.
  We show by induction on the number of transitions:
  \[
    \left(l_0, \clockvaluation_0\right) \taltstrans{d_1}{\sigma_1} \left(l_1, \clockvaluation_1\right) \taltstrans{d_2}{\sigma_2} \ldots \taltstrans{d_n}{\sigma_n} \left(l_n, \clockvaluation_n\right) \text{ iff } \la \la\ra, \delta \ra \warrow[w] \la z_1, \delta_1 \ra \warrow[w] \ldots \warrow[w] \la z_n, \delta_n \ra
  \]
  such that
  \begin{enumerate}
    \item $w, z_n \models \forall o. \occ(o) \equiv \exists l, \varphi, Y, l'.\, o = s(l, \sigma_n, \varphi, Y, l')$,
    \item $\ztime(z_n) = \sum_{i=1}^n d_i$,
    \item $w, z_n \models (\taloc = l_n)$
    \item $w, z_n \models \mi{final}(\taloc)$ iff $l_n \in \tastates_F$
    \item $w, z_n \models (c_i = r)$ iff $\clockvaluation_n(c_i) = r$ for every $c_i \in \taclocks$
    \item $\delta_{n} = \left(\pi a.\, \poss(a) \wedge \clockconstraint(a)?; a\right)^* ;\; \mi{final}(\taloc)?$
  \end{enumerate}
  \textbf{Base case.}
  \\
  Let $n = 0$, i.e., there is no transition and therefore, $z_0 = \la\ra$.
  It follows:
  \begin{enumerate}
    \item $w, z_0 \models \forall o.\, \neg \occ(o)$ and therefore, as there is no $\sigma_0$, we have $w, z_0 \models \forall o. \occ(o) \equiv \exists l, \varphi, Y, l'.\, o = s(l, \sigma_0, \varphi, Y, l')$,
    \item $\ztime(z_0) = \ztime(\la\ra) = 0 = \sum_{i=1}^0 d_i$,
    \item $w, z_0 \models (\taloc = l_0)$ by definition of $\bat_0$,
    \item $w, z_0 \models \mi{final}(\taloc)$ iff $w, z_0 \models \mi{final}(l_0)$ (by definition of $\bat_0$) iff $l_0 \in \tafstates$ (by definition of $\mi{final}$),
    \item $w, z_0 \models (c_i = 0)$ and $\clockvaluation_0(c_i) = 0$ for every $c_i \in \taclocks$,
    \item $\delta_0 = \delta = \left(\pi a.\, \poss(a) \wedge \clockconstraint(a)?; a\right)^* ;\; \mi{final}(\taloc)?$ by definition of $\delta$.
  \end{enumerate}
  \textbf{Induction step.}
  Assume:
  \begin{align*}
    \left(l_0, \clockvaluation_0\right) \taltstrans{d_0}{\sigma_0} \left(l_1, \clockvaluation_1\right) &\taltstrans{d_1}{\sigma_1} \ldots \taltstrans{d_n}{\sigma_n} \left(l_n, \clockvaluation_n\right)
    \\
    \la \la\ra, \delta \ra \warrow \la z_1, \delta_1 \ra &\warrow \ldots \warrow \la z_n, \delta_n \ra
  \end{align*}
  We need to show that $\left(l_n, \clockvaluation_n\right) \taltstrans{d_n}{\sigma_n} \left(l_{n+1}, \clockvaluation_{n+1}\right)$ iff $\la z_n, \delta_n \ra \warrow[w] \la z_{n+1}, \delta_{n+1} \ra$ such that $z_{n+1} = z_n \cdot a$ and $a = s(l_n, \sigma_{n+1}, \varphi, Y, l_{n+1})$ for some $\varphi, Y$.
  By induction, $w, z_n \models (\taloc = l_n)$.
  Also, again by induction, $w, z_n \models (c_i = r)$ iff $\clockvaluation_n(c_i) = r$ for every $c_i \in \taclocks$.
  We first show that $\left(l_n, \clockvaluation_n\right) \taltstrans{d_{n+1}}{\sigma_{n+1}} \left(l_{n+1}, \clockvaluation_{n+1}\right)$ iff $\la z_n, \delta_n \ra \warrow \la z_{n+1}, \delta_{n+1} \ra$.
  \\
  \textbf{$\Rightarrow$:}
  Assume $\left(l_n, \clockvaluation_n\right) \taltstrans{d_{n+1}}{\sigma_{n+1}} \left(l_{n+1}, \clockvaluation_{n+1}\right)$.
  Thus, by \autoref{def:ta-lts}, there is a switch $\left(l_n, \sigma_{n+1}, \varphi, Y, l_{n+1}\right)$ such that for $\clockvaluation^* = \clockvaluation_n + d_{n+1}$:
  \begin{itemize}
    \item $\clockvaluation^* \models \varphi$,
    \item $\clockvaluation_{n+1} = \clockvaluation^*[Y := 0]$,
    \item $\clockvaluation_n + d \models I(l_n)$ for each $0 \leq d \leq d_{n+1}$, and
    \item $\clockvaluation_{n+1} \models I(l_{n+1})$.
  \end{itemize}
  Let $a = s(l_n, \sigma_{n+1}, \varphi, Y, l_{n+1})$, $t_{n+1} = t_n + d_n$, and $z_{n+1} = z_n \cdot t_{n+1} \cdot a$.
  As $w, z_n \models (\taloc = l_n)$, it directly follows that $w, z_n \models \poss(a)$.
  Also, $w, z_n \cdot t_{n+1} \models \varphi \wedge I(l_n) \wedge I(l_{n+1})$ (using the assumption that $I(l_{n+1})$ does not mention any $c \in Y$).
  Therefore, $w, z_n \cdot t_{n+1} \models \clockconstraint(a)$.
  Thus, by \autoref{def:tesg-trans}, $\la z_n, \delta_n \ra \warrow \la z_{n+1}, \delta_{n+1} \ra$.
  \\
  \textbf{$\Leftarrow$:}
  Assume $\la z_n, \delta_n \ra \warrow \la z_{n+1}, \delta_{n+1} \ra$ with $z_{n+1} = z_n \cdot t_{n+1} \cdot p_{n+1}$.
  By definition of \tabat and \autoref{def:tesg-trans}, there is an action $a = s(l_n, \sigma_{n+1}, \varphi, Y, l_{n+1}) = p_{n+1}$ such that:
  \begin{itemize}
    \item $w, t_{n+1} \models \poss(a) \wedge \clockconstraint(a)$,
    \item $w, z_{n+1} \models (\taloc = l_{n+1})$,
  \end{itemize}
  Let $d_{n+1} = t_{n+1} - \ztime(z_n)$.
  By definition of \tabat, there is a switch $(l_n, \sigma_{n+1}, \varphi, Y, l_{n+1}) \in \taswitches$.
  Let $\clockvaluation^* = \clockvaluation_n + d_{n+1}$.
  We show that $\left(l_n, \clockvaluation_n\right) \taltstrans{d_{n+1}}{\sigma_{n+1}} \left(l_{n+1}, \clockvaluation_{n+1}\right)$:
  From $w, t_{n+1} \models \clockconstraint(a)$, it follows that $w, t_{n+1} \models \varphi \wedge \invs(l_n) \wedge \invs(l_{n+1})$.
  By induction, $w, z_n \models (c_i = r)$ iff $\clockvaluation_n(c_i) = r$.
  Thus, $\clockvaluation^* \models \varphi \wedge \invs(l_n) \wedge \invs(l_{n+1})$.
  By assumption, $\invs(l_{n+1})$ does not mention any clocks from $Y$, and thus, it follows that $\clockvaluation_{n+1} = \clockvaluation^*[Y \eqdef 0] \models \invs(l_{n+1})$.
  Thus, $\left(l_n, \clockvaluation_n\right) \taltstrans{d_{n+1}}{\sigma_{n+1}} \left(l_{n+1}, \clockvaluation_{n+1}\right)$.

  \bigskip

  \noindent Therefore, $\left(l_n, \clockvaluation_n\right) \taltstrans{d_{n+1}}{\sigma_{n+1}} \left(l_{n+1}, \clockvaluation_{n+1}\right)$ iff $\la z_n, \delta_n \ra \warrow \la z_{n+1}, \delta_{n+1} \ra$.
  Additionally, it follows:
  \begin{enumerate}
    \item $w, z_{n+1} \models \forall o. \occ(o) \equiv \exists l, \varphi, Y, l'.\, o = s(l, \sigma_{n+1}, \varphi, Y, l')$ by definition of the \ac{SSA} of $\occ$;
    \item $\ztime(z_{n+1}) = z_n + d_{n+1}$.
      By induction, $z_n = \sum_{i=1}^n d_i$.
      Thus, $\ztime(z_{n+1}) = \sum_{i=1}^{n+1} d_i$;
    \item $w, z_{n+1} \models (\taloc = l_{n+1})$ by definition of the \ac{SSA} of $\taloc$;
    \item $w, z_{n+1} \models \mi{final}(\taloc)$ iff $w, z_{n+1} \models \mi{final}(l_{n+1})$ (by previous item) iff $l_{n+1} \in \tafstates$.
    \item By induction, for each $c_i \in \taclocks$, $w, z_{n} \models (c_i = r)$ iff $\clockvaluation_{n}(c_i)$.
      It follows:
      \begin{itemize}
        \item If $c_i \in Y$, then $\clockvaluation_{n+1}(c_i) = 0$.
          Also, by definition of the \ac{SSA} of $\reset$, $w, z_{n+1} \models \reset(c_i)$.
          Thus, by \autoref{def:tesg-world}, $w, z_{n+1} \models (c_i = 0)$.
        \item Otherwise, $\clockvaluation_{n+1}(c_i) = \clockvaluation_n + d_{n+1}$.
          By definition of the \ac{SSA} of $\reset$, $w, z_{n+1} \models \neg \reset(c_i)$.
          Thus, by \autoref{def:tesg-world}, $w[c_i, z_{n+1}] = w[c_i, z_n \cdot t_{n+1}] = w[c_i, z_n] + t_{n+1} - \ztime(z_n) = w[c_i, z_n] + d_{n+1}$.
          By induction, $w[c_i, z_n] = \clockvaluation_n(c_i)$.
          Thus, $w[c_i, z_{n+1}] = \clockvaluation_{n+1}(c_i)$.
      \end{itemize}
    \item  $\delta_{n+1} = \left(\pi a.\, \poss(a) \wedge \clockconstraint(a)?; a\right)^* ;\; \mi{final}(\taloc)?$ follows directly by \autoref{def:tesg-trans}.
      \qedhere
  \end{enumerate}
\end{proofE}

For infinite words, the construction does not work, because the acceptance conditions for \acp{TA} and non-terminating programs differ: While a \ac{TA} accepts a word if the corresponding run visits a final state infinitely often (Büchi condition), an infinite run of a program is accepted if it never visits a final configuration.
While it may be possible to adapt the construction to also work for infinite runs, we do not investigate this here, as we are only interested in finite traces later on.


\section{Avoiding Undecidability with Clocks}\label{sec:tesg-decidability}
Before we conclude the discussion of \tesg and its properties, we motivate in this section why we deviated from the common approach to include time in the situation calculus, as sketched in \autoref{sec:sitcalc}.
Usually, time is added to the situation calculus (and its variants such as \esg) by adding a time argument to each action and by having a special fluent function $\ztime(a)$ that gives the time point of executing action $a$.
In \esg, this may modeled with a \ac{SSA} as follows:
\[
  \square [a] \ztime(a') = t \equivspace a = \sac{a', t}
\]

This can then be used in a precondition axiom of the corresponding end action:
\begin{align*}
  \square \poss(a) \equivspace &\exists l,l', t_e.\, a = \eac{\drive(l, l'), t_e}
                            \\ & \quad \wedge \performing{\drive(l, l')} \wedge t_e \geq \ztime(\drive(l, l'))  + 2
  \\
                               &\vee \ldots
\end{align*}
In words, it is possible to end the action $\drive(l, l')$ if the robot is currently performing the action and it has started the action at least two time steps ago.

Alternatively, if we want to avoid to have an explicit time argument for each action (which is problematic if we want to use \realpos as time domain), we may also instead extend the denotation of terms (\autoref{def:tesg-denotation}) as follows:
\begin{enumerate}
  \item The special function $\mi{now}$ denotes the current time,
    \\
    i.e., if $z = (a_1, t_1) \cdots (a_k, t_k)$, then $\vert \mi{now} \vert^z_w = t_k$.
  \item For actions $a$, $\ztime(a)$ denotes the last occurrence of $a$.
    Formally:
    \[
      \vert \ztime(a) \vert ^z_w = \max\{ t_a \mid (a, t_a) \in z \}
    \]
\end{enumerate}

By doing so, we do not need a \ac{SSA} for $\ztime$ and we can define the precondition axiom of the end action of \drive as follows:
\begin{align*}
  \square \poss(a) \equivspace &\exists l,l'.\, a = \eac{\drive(l, l)} \\
                               & \quad \wedge \performing{\drive(l, l')} \wedge \mi{now} \geq \ztime(\sac{\drive(l, l')}) + 2
\end{align*}
This is the approach taken in an earlier version of \tesg~\cite{hofmannLogicSpecifyingMetric2018}.

In both approaches, we need fluent time functions, we must be able to do basic arithmetic operations such as $+$ and $-$, and we need to compare time fluents.
In the following, we show that reasoning in such a logic is undecidable, even if the objects and actions (but not time) are restricted to finite domains, as described in \autoref{sec:complete-finite-bat}.

For the sake of the argument, we extend \tesg to $\tesg^*$ as follows:
\begin{itemize}
  \item We add fluent and rigid functions of type \emph{time}, in particular $+,-$ with the intended semantics.
  \item We include the binary predicate $<$ with the intended semantics for terms of type time.
\end{itemize}

A \ac{BAT} in $\tesg^*$ is like a \ac{BAT} in \tesg, except that it may also include \acp{SSA} for fluent time functions.
Similar to \autoref{sec:complete-finite-bat}, we call a $\tesg^*$ \ac{BAT} \emph{finite-domain} if all quantifiers of objects and actions are restricted to a finite domain.

We show that reasoning in $\tesg^*$ is undecidable, even with a finite domain of objects and actions.
More specifically, we define a program $\Delta = (\bat, \delta)$ with a finite number of actions and objects such that deciding whether $\delta$ terminates is undecidable.
We do so by reducing the \emph{halting problem for two-counter machines}:
\begin{definition}[Two-Counter Machines~\parencite{minskyComputationFiniteInfinite1967,bouyerUpdatableTimedAutomata2004}]
  A \emph{two-counter machine} is a finite set of labeled instructions over two counters $c_1$ and $c_2$.
  There are two types of instructions:
  \begin{enumerate}
    \item An \emph{incrementation instruction} of counter $x \in \{ c_1, c_2 \}$:
      \[
        p:\: x \eqdef x + 1; \;\mathbf{goto}\; q
      \]
      The instruction increments counter $x$ by one and then goes to the next instruction $q$.
    \item A \emph{decrementation instruction} of counter $x \in \{ c_1, c_2 \}$:
      \[
        p:\: \textbf{if}\; x > 0
        \begin{cases}
          \mathbf{then}\; x \eqdef x - 1; \; \mathbf{goto}\; q
          \\
          \mathbf{else}\; \mathbf{goto}\; r
        \end{cases}
      \]
      The instruction branches on $x$: If $x$ is larger than $0$, then it decrements $x$ and goes to instruction $q$.
      Otherwise, it does not change $x$ and directly goes to instruction $r$.
  \end{enumerate}
  The machine starts with instruction $s_0$ and with  counter values $c_1 = c_2 = 0$ and stops at a special instruction $\textbf{HALT}$.
  The \emph{halting problem} for a two-counter machine is to decide whether a machine reaches the instruction $\textbf{HALT}$.
\end{definition}

Two-counter machines are useful to show undecidability by reducing a given problem to the halting problem for two-counter machines:
\begin{theorem}[\parencite{minskyComputationFiniteInfinite1967}]
  The halting problem for two-counter machines is undecidable.
\end{theorem}

We can define a $\tesg^*$ \ac{BAT} that models a two-counter machine as follows:
Let $\mi{Incrs} = \{ (p, c, q)_i \}_i$ be the finite set of increment instructions, where $p$ is the instruction label, $c \in \{ c_1, c_2 \}$ is the counter to be incremented, and $q$ is the next instruction label.
Similarly, let $\mi{Decrs} = \{ (p, c, q, r)_i \}_i$ be the finite set of decrement instructions, where $p$ is the instruction label, $c \in \{ c_1, c_2 \}$ is the counter to be decremented, $q$ is the jump instruction if the condition is true, and $r$ is the jump instruction otherwise.
We define a \ac{BAT} $\bat_\mathcal{M}$ corresponding to a two-counter machine $\mathcal{M}$ as follows:
\begin{itemize}
  \item There are four  fluents:
    \begin{itemize}
      \item The unary relational fluent $\rfluent{next}$ describes the next instruction.
      \item The nullary functional fluents $c_1$ and $c_2$ of sort time track the counter values.
      \item The nullary relational fluent \rfluent{Halt} is true if the machine halts.
    \end{itemize}
  \item Each instruction label $p, q, r, \ldots$ (including $s_0$ and $\ffluent{halt}$) is an action.
    Initially, both counters are zero and the next instruction is the action with label $s_0$:
    \[
      \bat_0 = \{ c_1 = 0, c_2 = 0, \rfluent{next}(a) \equiv a = s_0 \}
    \]
  \item An action is possible iff it is the next instruction:
    \[
      \square \poss(a) \equivspace \rfluent{next}(a)
    \]
  \item For each $i$, the counter $c_i$ is incremented if the instruction is an increment of $c_i$, decremented if the instruction is a decrement of $c_i$ and $c_i > 0$, and unchanged otherwise:
    \begin{align*}
      \square [a] c_i = n \equivspace &\bigvee_{\mathclap{(p, c, q) \in \mi{Incrs}}} a = p \wedge c = c_i \wedge  n = c_i + 1
                                   \\ & \vee \bigvee_{\mathclap{(p, c, q, r) \in \mi{Decrs}}} a = p \wedge c = c_i
                                      \wedge (c_i > 0 \wedge n = c_i - 1 \vee c_i = 0 \wedge n = c_i)
                                    \\
                                      & \vee c_i = n \wedge \bigwedge_{\mathclap{(p, c, q) \in \mi{Incrs}}} (a \neq p  \vee c \neq c_i) \wedge \bigwedge_{\mathclap{(p, c, q, r) \in \mi{Decrs}}} (a \neq p \vee c \neq c_i)
    \end{align*}
  \item The next instruction is as specified by the current instruction:
    \begin{align*}
      \square [a] \rfluent{next}(i) \equivspace &\bigvee_{\mathclap{(p, c, q) \in \mi{Incrs}}} a = p \wedge i = q
      \\
                                            &\quad \vee \bigvee_{\mathclap{(p, c, q, r) \in \mi{Decrs}}} a = p \wedge (c > 0 \wedge i = q \vee c = 0 \wedge i = r)
    \end{align*}
  \item The predicate \rfluent{halt} is true if and only if the last action was the special instruction $\action{halt}$:
    \[
      \square [a] \rfluent{halt} \equivspace a = \action{halt}
    \]
\end{itemize}

The program $\delta$ nondeterministically picks an instruction, checks if it is possible, and then executes it until it reaches \rfluent{halt}:
\[
  \gwhile \neg \rfluent{halt} \gdo
  \pi a;\; \poss(a)?; a
  \gdone
\]
As there is only a single instruction that is possible at any point in time, the program just executes the instructions as defined by the two-counter machine.

Finally, to check whether the program halts, we can use the following query:
\[
  \bat_\mathcal{M} \models \neg [ \delta ] \bot
\]
If the two-counter machine $\mathcal{M}$ does not halt, then $[\delta] \bot$ is satisfied because there is no finite execution of $\delta$.
If there is no finite execution, then $[\delta] \alpha$ is vacuously true for any $\alpha$, including $\alpha = \bot$.
This results in the following proposition:
\begin{proposition}
  The two-counter machine $\mathcal{M}$ halts iff $\bat_\mathcal{M} \models \neg [\delta] \bot$.
\end{proposition}

Hence, in order to allow reasoning about time, we may not just add time functions along with the standard operators $+,-$ and the relation $<$ to the logic, as this immediately results in an undecidable projection problem (and therefore also undecidable verification and synthesis problems, which will be introduced in \autoref{chap:synthesis}), even if we restrict the domain to a finite number of objects and actions.
If we also restrict the time to a finite domain, then the problem will likely disappear.
However, this is not suitable for our purposes, as it precludes the reals as time domain and essentially restricts the expressible temporal properties to \ac{LTL}.
Furthermore, as the boundary of decidability has been researched extensively for timed automata and their extensions (e.g., \parencite{alurTheoryTimedAutomata1994,henzingerWhatDecidableHybrid1998,berardTimedAutomataAdditive2000,bouyerUpdatableTimedAutomata2004}), it is reasonable to restrict the logic syntactically such that it allows precisely those timing constraints that are allowed in timed automata.

\section{Discussion}

In this chapter, we have introduced \tesg, a variant of the situation calculus that allows to formulate temporal real-time constraints on the program execution.
The logic is based on \esg, which already allowed to express temporal properties of program execution traces similar to \ac{LTL}.
In comparison to \esg, \tesg program traces consist of alternating time and action steps, corresponding to a fixed time of occurrence for each action, allowing us to express temporal properties referring to metric time, akin to \ac{MTL}.
Additionally, the logic incorporates \emph{clocks} and \emph{clock constraints} on actions, which model timing constraints similar to how precondition axioms model state constraints.
The logic does not allow arbitrary arithmetic operations on clocks.
Instead, clocks can only be compared to fixed rational numbers and may be reset to zero by an action.
This is necessary because allowing standard arithmetic on the reals results in undecidable reasoning problems, even on finite domains.

We have also introduced a notion of regression that reduces a query about the state after some timed trace to a query about the initial state.
The regression operator is restricted to rational traces, because real numbers are not contained in the language of the logic.
However, as we will see in the next chapter, this suffices to determine whether some formula is satisfied after every possible execution of some given program, because every program trace is bisimilar to some program trace that only mentions rational time steps.

We have also introduced \emph{finite-domain \acp{BAT}}, where the number of actions and objects is restricted to a finite set.
As such a finite-domain \ac{BAT} only allows finitely many initial situations and each fluent value is uniquely determined by $\bat_{\text{post}}$ once the initial values are fixed, we may assume that a finite-domain \ac{BAT} \bat is a \ac{BAT} with complete information.
If the initial situation of a finite-domain \ac{BAT} \bat is not completely determined, we may consider one model $w \models \bat$ for each equivalence class of \equivbat, e.g., to determine a realization of a program, or to determine a control strategy.
When executing the program or controller, we then only need to determine which of the possible initial situations is the true initial situation and use the corresponding realization or controller.
Clearly, this will not perform well in practice, as we may obtain an exponential number of equivalence classes.
However, if we are only concerned with the decidability of the synthesis problem over finite domains, we may assume complete information without loss of generality.

Regarding the properties of the logic, we have seen that \esg \acp{BAT} can be used with \tesg, as a \ac{BAT} entails the same sentences in \esg and \tesg.
Furthermore, valid \ac{MTL} sentences are valid trace formulas in \tesg and vice versa.
Therefore, \tesg can be seen as a combination of \esg and \ac{MTL} that preserves the properties of the two logics.
Finally, we have demonstrated that \tesg is expressive enough to model \aclp{TA}.
Therefore, the logic is a well-suited foundation for the following two chapters.

\chapter{Program Transformation as Synthesis}\label{chap:synthesis}

As motivated in \autoref{chap:introduction}, the goal of this thesis is to transform an abstract program based on a self model of the robot such that it satisfies additional constraints given as \iac{MTL} specification.
In the previous chapter, we have introduced \tesg, which already allows us to define \golog programs with metric time, based on real-valued clocks.
In this chapter, we describe a first approach to the program transformation, which is based on \emph{synthesis}.
In synthesis, based on a partition of the available actions into controllable and environment actions, the task is to determine a controller that executes the given program such that each resulting trace satisfies the specification, no matter what the environment does.
Closely related to the synthesis problem is \emph{verification}.
In verification, the task is to check whether a \golog program is guaranteed to satisfy a specification.
In our case, the specification is again \iac{MTL} formula that describes \emph{undesired behavior}.
Therefore, in verification, we need to check whether there is an execution trace that satisfies the specification, in which case the program is \emph{unsafe}.
In our setting, synthesis is a direct extension of verification: While verification checks for any unsafe execution trace, synthesis checks whether it is possible to avoid those traces by choosing the right actions.

In the following, we first provide a formal definition of the verification problem in \autoref{sec:verification-problem} and the synthesis problem in \autoref{sec:control-problem}.
We continue with an approach that solves both the verification and synthesis problems.
As we have shown in \autoref{sec:tesg-ta}, we can model \iac{TA} in \tesg.
Therefore, for the sake of simplicity, we assume that we are given a \golog program $\Delta = (\bat, \delta)$ that contains both the abstract program $\delta_h$ and the self model of the robot in the form of a sub-program $\delta_l$, which may be composed as parallel programs, i.e., $\delta \eqdef \delta_h \| \delta_l$.
While this gets rid of the separation of the abstract program and the robot self model that we have before argued for, this is purely for the sake of the theoretical treatment of the transformation.
For practical purposes, the abstract program and the self model may be implemented separately.

Furthermore, we assume that $\Delta$ is a finite-domain program with complete information, i.e., the domain of discourse only contains finitely many objects and actions and the initial situation is completely determined.
As argued in \autoref{chap:timed-esg}, assuming complete information is not a restriction for finite-domain programs, because there are only finitely many alternatives, which we may just consider one by one.
Also, we need to assume that the domain of discourse is finite because \ac{MTL} is propositional and therefore only allows finitely many objects.

The transformation procedure is inspired by \ac{MTL} controller synthesis for timed automata~\parencite{bouyerControllerSynthesisMTL2006} and works as follows:
In a first step, the \ac{MTL} formula is translated into \iac{ATA}, as described in \autoref{sec:ata}.
In \autoref{sec:synchronous-products}, we compute the \emph{synchronous product} of the program and the \ac{ATA}, which is \iac{LTS} that describes the parallel execution of the program and the \ac{ATA}.
In principle, we can use this \ac{LTS} to check whether the program is safe and whether a control strategy exists.
However, the \ac{LTS} is both infinitely branching and may contain infinite paths.
Therefore, in \autoref{sec:regionalization}, we first reduce the \ac{LTS} to a finitely-branching \ac{LTS} by using \emph{regionalization} and we show that this \ac{LTS} is equivalent to the original \ac{LTS} in the sense of \emph{time-abstract bisimulation}.
Next, we define a determinized version of the \ac{LTS} in \autoref{sec:determinization} and we show in \autoref{sec:wsts} that the time-abstract \ac{LTS} is a \acfi{WSTS}, which allows us to stop on every infinite path after a finite number of steps.
As the resulting \ac{LTS} is finite, we directly obtain that the verification problem is decidable.
To solve the synthesis problem, we play a variant of a \emph{timed game} on the finite \ac{LTS} in \autoref{sec:timed-game}, which allows us to determine a control strategy.
After obtaining these theoretical results, we also describe and evaluate an implementation of the approach in \autoref{sec:synthesis-evaluation}.
We summarize and discuss the synthesis approach in \autoref{sec:synthesis-discussion}.


\section{The \acs*{MTL} Verification Problem for \golog Programs}\label{sec:verification-problem}

We start with the \emph{verification problem} of checking whether a \golog program violates an \mtl specification of undesired behavior $\phi$.
Formally, the verification problem is defined as follows:
\begin{definition}[Verification Problem]
  Let $\Delta = (\bat, \delta)$ be a finite-domain program and $\phi$ a trace formula.
  The \emph{\ac{MTL} verification problem for \golog programs} is to decide whether $\bat \models \llbracket \delta \rrbracket \neg \phi$.
\end{definition}
In other words, the goal is to check whether it can be guaranteed that every possible execution of $\delta$ avoids the undesired behavior specified by $\phi$.

As the program $\delta$ may be non-terminating and therefore allow infinite traces and because \ac{MTL} is undecidable over infinite words, we immediately obtain the following corollary from \autoref{thm:mtl-equivalence}:
\begin{corollary}\label{crly:infinite-traces-undecidability}
  The verification problem for finite-domain \golog programs is undecidable.
\end{corollary}

Hence, we will only consider finite program traces of the program $\delta$:
\begin{definition}[Verification problem over finite traces]
  Let $\Delta = (\delta, \bat)$ be a finite-domain program $\phi$ be a trace formula.
  The \emph{\ac{MTL} verification problem for \golog programs over finite traces} is to decide whether $\bat \models \llbracket \delta \rrbracket^{< \infty} \neg \phi$.
\end{definition}
Note that we do not require that the program $\Delta$ only produces finite traces.
However, we only put restrictions on finite executions, i.e., we only require that $\phi$ is satisfied if the program terminates.
Otherwise, on infinite runs, we do not pose any restrictions on the execution of the program.
By doing so, we avoid undecidability (\autoref{crly:infinite-traces-undecidability}) while still allowing possibly non-terminating (sub-)programs, e.g., loops in the robot self model in the form of \iac{TA}.

\section{The \acs*{MTL} Control Problem for \golog Programs}\label{sec:control-problem}

Related to and extending verification is the \emph{control problem}.
In verification, we merely check whether the specification is guaranteed to be satisfied.
However, if good behavior can not be guaranteed, verification will simply return ``no''.
In comparison, in controller synthesis, the answer is not a simple ``yes'' or ``no''.
Instead, the goal is to determine a \emph{controller}, which has additional control over the execution of the program.
If a certain execution trace violated the specification, then the controller may avoid this by executing a different path.
In controller synthesis, all available actions are partitioned into \emph{controllable actions} and \emph{environment actions}.
While the controller may decide which controller action to execute, the environment actions are not under its control.
Therefore, a controller needs to find a control strategy that selects the right controller actions such that no matter which environment actions are executed, the specification is not violated.
In our case, both controller and environment actions are restricted by the program: Both controller and environment may only choose actions that are possible according to the current program configuration.

Typically, the agent can control the \emph{start} but not the \emph{end} of a durative action.
Therefore, start actions are usually controller actions while all end actions are environment actions.
Furthermore, we may model exogenous events such as an incoming request as additional environment actions.

Intuitively, a controller defines for every possible execution state of the program which action(s) to execute next.
Formally, a controller is defined as follows:
\begin{definition}[Controller]\label{def:controller}
Let $\Delta = (\delta, \bat)$ be a program and $\batactions = A_E \dot\cup A_C$ be a partition of possible actions.
A controller $\controller$ is a partial function $\traces \times \sub(\delta) \rightarrow \batactions \times \realpos$ that maps a configuration to a set of timed actions, i.e., $\controller(z, \rho) = \{ (a_i, t_i) \}_{i \in I}$ such that
\begin{enumerate}[align=left,label=(C\arabic*)]
  \item For each $i$, $\la z, \rho \ra \warrow \la z \cdot (a_i, t_i), \rho_i \ra$ for some $\rho_i$;
  \item For each $a_e \in A_E$, if $\la z, \rho \ra \warrow \la z \cdot (a_e, t), \rho' \ra$, then
    \begin{itemize}
      \item $(a_e, t) \in \controller(z, \rho)$, or
      \item there is $a_c \in A_C$ and $t_c < t$ such that $(a_c, t_c) \in \controller(z, \rho)$;
    \end{itemize}
  \item $\controller(z, \rho) = \emptyset$ implies $\la z, \rho \ra \in \mathcal{F}^w$.
    \qedhere
\end{enumerate}
\end{definition}

As we can see, a controller must satisfy certain restrictions:
\begin{enumerate}
  \item For each selected action, there must must be some program transition of the remaining program $\rho$ in the world $w$, i.e., the controller may only select actions that are actually possible to execute according to the program.
  \item For each environment action that is possible in the current state, the controller must either select this environment action, or it must select a controller action that occurs strictly before the environment action.
    Note that this is slightly different from the usual definition of \emph{non-restrictiveness} (e.g., \cite{dsouzaTimedControlSynthesis2002}): In the standard definition, the controller must allow any environment action, independent of the time of occurrence.
    However, in our setting, this is very restrictive, as the controller may effectively never interfere, unless there is currently no possible environment action.
    Instead, in the modified definition above, the controller may interfere, as long as its action occurs before the environment action.
  \item The controller must be \emph{non-blocking}: if it decides to select no action, then the program must be in a final configuration.
\end{enumerate}

A controller \controller restricts the traces of a program $\|\delta\|$ to a subset \controllertraces, which results by following the action selection iteratively.
Formally, the controller traces are defined as follows:
\begin{definition}[Controller Trace]
  Let $\Delta = (\delta, \bat)$ be a program and \controller be a controller for $\Delta$.
  Then the controller traces \controllertraces of \controller is the set of traces with $(a_1, t_1) \cdots (a_n, t_n) \in \controllertraces$ if and only if there are $\la z_0, \rho_0 \ra, \ldots, \la z_{n}, \rho_{n} \ra$  such that
  \begin{enumerate}
    \item $z_i = (a_1, t_1) \cdots (a_i, t_i)$ (where $z_0 = \la\ra$) and $\rho_0 = \delta$,
    \item for each $i$, $\la z_i, \rho_i \ra \warrow \la z_{i+1}, \rho_{i+1} \ra$ and $(a_{i+1}, t_{i+1}) \in \controller(z_i, \rho_i)$,
    \item $\controller(z_n, \rho_n) = \emptyset$.
      \qedhere
  \end{enumerate}
\end{definition}

The goal of controller synthesis is to determine a controller that avoids \emph{undesired behavior} $\phi$, where $\phi$ is a \iac{MTL} formula.
The \emph{control problem} is defined as follows:
\begin{definition}[Control Problem]\label{def:control-problem}
  Let $\Delta$ be a finite-domain program, $\batactions = A_E \dot\cup A_C$ a partition  of possible actions, and $\phi$ a fluent trace formula.
  The \emph{control problem} is to determine a controller \controller such that for each finite controller trace $\psi \in \controllertraces$: $w, \la\ra, \psi \models \neg \phi$.
\end{definition}

We assume that the specification $\phi$ does not mention any function symbols.
We follow the usual convention to specify the required behavior in terms of undesired behavior.
However, this is not a restriction: given a specification for \emph{desired behavior} $\theta$, a controller that controls against the undesired behavior $\phi = \neg \theta$ will guarantee that every controller trace will satisfy $\neg \phi = \neg \neg \theta \equiv \theta$.
Similar to verification, we only require the specification $\phi$ to be avoided on finite traces, as determining a controller on infinite traces is undecidable.

We continue with a simple example for a control problem, based on the \ac{BAT} from \autoref{sec:tesg-bat}:
\newcommand*{\phione}{\ensuremath{\top \until{} (\neg \camon \wedge \grasping)}}
\newcommand*{\phitwo}{\ensuremath{\top \until{}(\neg \camon \wedge \top \until{[0, 2]} \grasping)}}
\newcommand*{\phithree}{\ensuremath{\top \until{[0, 2]} \grasping}}
\begin{example}[Control Problem] \label{ex:control-problem}
  Consider the \ac{BAT} from \autoref{ex:bat} with the following program:
  \begin{align*}
    \delta_h \eqdef {}&\sac{\drive(m_1, m_2)};  \eac{\drive(m_1, m_2)};
    \\
    {} & \quad \sac{\grasp(m_2, o_1)}; \eac{\grasp(m_2, o_1)}
    \\
    \delta_m \eqdef {}&\sac{\bootcam}; \eac{\bootcam}
    \\
    \delta \eqdef {}&\delta_h \| \delta_m
  \end{align*}
  In the high-level program $\delta_h$, the robot first drives to machine $m_2$ and then grasps the object $o_1$.
  At the same time, the maintenance program $\delta_m$ simply boots the camera.
  The main program $\delta$ executes both programs concurrently.

  In this simple scenario, the controller needs to determine the order of execution and the exact time points of the actions such that the following specification of undesired behavior is avoided:
  \begin{align*}
    \phi \eqdef \finally{}(\neg \camon \wedge \grasping) \vee \finally{}(\neg \camon \wedge \finally{[0, 2]} \grasping)
  \end{align*}
  The first disjunct of $\phi$ states that it is bad behavior if there is some future state in which the robot is grasping an object while the camera is turned off.
  The second disjunct is similar but enforces that the camera must have been turned on for at least \SI{2}{\sec}.
  It states that it is bad behavior if there is some future state where the camera is turned off and there is a later state within \SI{2}{\sec} where the robot is grasping an object.
  Note that the second disjunct does not entail the first, as we use strict semantics, and thus the second disjunct does not say anything about the state in which $\neg \camon$ was observed, but only about subsequent states.
  Overall, the specification guarantees that the camera is ready to use whenever the robot intends to grasp an object.
\end{example}

\section{Synchronous Products}\label{sec:synchronous-products}


To synthesize a controller that satisfies the above criteria and that guarantees that the specification is not violated, we need to explore the state space of the program to find paths that end in a final program configuration while not violating the specification.
In order to do so, we first construct the \ac{ATA} corresponding to the \ac{MTL} specification, as described in \autoref{sec:ata}.
The resulting automaton checks the satisfaction of the specification and accepts any timed word that violates the specification:
\begin{example}[\ac{ATA} for the specification $\phi$]
  We start with the specification $\phi$ from \autoref{ex:control-problem}:
  \begin{align*}
    \phi \eqdef \finally{}(\neg \camon \wedge \grasping) \vee \finally{}(\neg \camon \wedge \finally{[0, 2]} \grasping)
  \end{align*}
  After translating all abbreviations, we obtain the equivalent formula:
  \[
    \phi = \top \until{} (\neg \camon \wedge \grasping) \vee \top \until{}(\neg \camon \wedge \top \until{[0, 2]} \grasping)
  \]
  Following the construction from \autoref{def:mtl-ata}, we obtain an \ac{ATA} $\ata = \left(\ataalphabet, \atalocations, \phi_i, \atafinallocations, \atatrans\right)$ which tracks the satisfaction of $\phi$, where:
  \begin{itemize}
    \item The alphabet consists of all subsets of $\{ \camon, \grasping \}$, i.e.,
      \[
        \ataalphabet = \powerset{\{ \camon, \grasping \}} = \{ \emptyset, \{ \camon \}, \{ \grasping \}, \{ \camon, \grasping \}\}
      \]
    \item The locations \atalocations consist of initial location $l_0$ and the closure $\closure(\phi)$ of $\phi$, i.e., the set of subformulas whose outermost connective is $\until{}$ or $\duntil{}$:
      \begin{align*}
        \atalocations = \{
          &l_0,
          \\
          &\phi_1 \eqdef \phione,
          \\
          &\phi_2 \eqdef \phitwo,
          \\
          &\phi_3 \eqdef \phithree
        \}
      \end{align*}
    \item As \atalocations does not contain any location whose outermost connector is \duntil{}, there is no final location: $\atafinallocations = \emptyset$.
    \item The transition function $\atatrans$ is defined as follows:
      \begin{center}
        \begin{tabular}{cccccc}
          \toprule
          & $\{\}$ & $\{ \camon \}$ & $\{ \grasping \}$ & $\{ \camon, \grasping \}$
          \\
          \cmidrule(lr){2-5}
          $l_0$ & $\phi_1 \vee \phi_2$ & $\phi_1 \vee \phi_2$ & $\phi_1 \vee \phi_2$ & $\phi_1 \vee \phi_2$
          \\
          $\phi_1$ & $\phi_1$ & $\phi_1$ & $\top$ & $\phi_1$
          \\
          $\phi_2$ & $\phi_2 \vee x.\phi_3$ & $\phi_2$ & $\phi_2 \vee x.\phi_3$ & $\phi_2$
          \\
          $\phi_3$ & $\phi_3$ & $\phi_3$ & $\left(x \geq 0 \wedge x \leq 2\right) \vee \phi_3$ & $\left(x \geq 0 \wedge x \leq 2\right) \vee \phi_3$
          \\
          \bottomrule
        \end{tabular}
      \end{center}
      We can make the following observations:
      \begin{itemize}
        \item From the initial location $l_0$, we can go into the locations $\phi_1 = \phione$ or $\phi_2 = \phitwo$ independent of the input symbols.
          This is because $\phi = \phi_1 \vee \phi_2$ and both $\phi_1$ and $\phi_2$ have the outermost connective $\until{}$.
          Note that we ignore the input symbol, as we use strict semantics for \until{} and therefore only consider states strictly in the future.
          Thus, the satisfied fluents in the initial situation do not have an influence on $\phi_1$ or $\phi_2$.
        \item In location $\phi_1 = \phione$, we always stay in $\phi_1$ unless we read $\{ \grasping \}$, in which case the successor configuration is the empty configuration $\{ \}$, which is the unique minimal model of $\top$.
          This is because if $\grasping$ is true and $\camon$ is false, then $\neg \camon \wedge \grasping$ is satisfied and therefore the specification has been violated.
        \item Concerning the location $\phi_2 = \phitwo$, we can see that for any input that satisfies $\neg \camon$, we can either stay in $\phi_2$ or reset the clock $x$ and switch to $\phi_3$.
          This is because if $\camon$ is false, then we only need to satisfy $\top \until{[0,2]} \grasping$ to satisfy $\phi_2$.
          This is done by resetting the clock so we can later check that the bounds $x \geq 0 \wedge x \leq 2$ are satisfied.
          However, it could also be the case that $\camon$ is currently false but $\grasping$ is not satisfied in the next two time units.
          For this reason, we may also just stay in the location $\phi_2$ in case it is satisfied later on.
        \item Finally, for $\phi_3 = \phithree$, we can see that for any input satisfying $\grasping$, the bound $x \geq 0 \wedge x \leq 2$ is checked.
          As $x$ was reset when transitioning from $\phi_2$, this keeps track of the time difference of the two states where $\neg \camon$ was satisfied and where $\grasping$ was satisfied.
          If $x \geq 0 \wedge x \leq 2$, then $\phi_3$ is satisfied and the next configuration is the empty configuration.
          Otherwise, we stay in $\phi_3$.\footnote{Note that in this particular case, the lower bound $x \geq 0$ is vacuously true.
            Furthermore, if the bound is not satisfied, then we can see that it will also never be satisfied later on, and thus we could simplify $\left(x \geq 0 \wedge x \leq 2\right) \vee \phi_3$ to $x \leq 2$.
          However, these simplifications are difficult to apply generally.}
          \qedhere
      \end{itemize}
  \end{itemize}
\end{example}

Next, we build the \emph{synchronous product} of the program and the \ac{ATA}.
The synchronous product follows all possible program transitions and the corresponding \ac{ATA} transitions and therefore contains all possible program executions while tracking the specification:

\begin{definition}[Synchronous Product]\label{def:sync-product}
  Let $\Delta = (\bat, \delta)$ be a finite-domain program and $\atalts = (\ataconfs, \ataconf_0, \ataconfs_F, \ataltsalph, \ataltstrans{}{})$ the \ac{LTS} corresponding to the \ac{ATA} \ata.
  The synchronous product $\synclts = (\syncstates, \syncstate_0, \syncstates^F, \batactions \cup \realpos, \synctrans{})$ is \iac{LTS} defined as follows:
\begin{itemize}
  \item The states consist of triples\footnote{We usually write $(\la z, \rho \ra, \ataconf)$ for a state $(z, \rho, \ataconf)$ to distinguish the program component from the \ac{ATA} component.} $(\la z, \rho \ra, \ataconf)$, where $z \in \traces$ is the trace of actions executed so far, $\rho \in \sub(\delta)$ is the remaining program, and $\ataconf$ is the current \ac{ATA} configuration.
    Additionally, there is a distinguished initial state $(\delta^i, \ataconf_0)$:
    \[
      \syncstates = (\traces \times \sub(\delta) \times \ataconfs) \cup \{ (\delta^i, \ataconf_0) \}
    \]
  \item The initial state is the pair $(\delta^i, \ataconf_0)$, which consists of the distinguished symbol $\delta^i$ and the initial \ac{ATA} configuration.
    \[
      \syncstate_0 = (\delta^i, \ataconf_0)
    \]
  \item For the initial state, there is an unlabeled transition to a state that consists of the initial program configuration $\la \la\ra, \delta \ra$ and the \ac{ATA} configuration \ataconf corresponding to the initial situation, i.e., the \ac{ATA} configuration that results from reading all primitive fluents that are true in the initial situation:
    \[
      (\delta^i, \ataconf_0) \synctranstime{} (\la \la\ra, \delta \ra, \ataconf)
      \text{ if }\ataconf_0 \atatranssym{F} \ataconf \text{ with } F = \{ P(\vec{n}) \in \primformulas \mid w[P(\vec{n}), \la\ra] = 1 \}
    \]
  \item A time transition labeled with $t \in \realpos$ progresses time of both the program and the \ac{ATA}:
    \[
      (\la z, \rho \ra, \ataconf) \synctranstime{t} (\la z^*, \rho^* \ra, \ataconf^*) \text{ if }\la z, \rho \ra \wtarrow[t] \la z^*, \rho^* \ra \text{ and }\ataconf \atatranstime{t} \ataconf^*
    \]
  \item A symbol transition with action $a \in \batactions$ corresponds to a symbol transition of the program with the same action.
    The successor \ac{ATA} configuration is the configuration resulting from reading all primitive fluents that are true in the situation after executing action $a$:
    \[
      (\la z^*, \rho^* \ra, \ataconf^*) \synctranssym{a} (\la z', \rho' \ra, \ataconf')
      \text{ if }
      \la z^*, \rho^* \ra \wsarrow[a] \la z', \rho' \ra \text{ and } \ataconf^* \atatranssym{F} \ataconf'
    \]
    with $F = \{ P(\vec{n}) \in \primformulas \mid w[P(\vec{n}), z'] = 1 \}$.
  \item A state is final if the program is in a final configuration and the \ac{ATA} is accepting:
    \[
      (\la z, \rho \ra, \ataconf) \in \syncstates^F \text{ iff } \la z, \rho \ra \in \final \text{ and } \ataconf \in \ataconfs_F
      \qedhere
    \]
\end{itemize}
\end{definition}

As the name suggests, the synchronous product \synclts synchronously follows the transitions of the program $\Delta$ and the \ac{ATA} \ata.
For each time transition, it simply progresses both the program and the \ac{ATA}.
For symbol transitions, it first computes the resulting program configuration and then uses the primitive fluents that are satisfied in the resulting program transition to determine the next \ac{ATA} configuration.
As there may be multiple resulting \ac{ATA} configurations for one symbol transition, the \ac{LTS} is nondeterministic, i.e., from a single state, there may be multiple symbol transitions with the same input symbol to different successor states.
Also note the distinguished initial state $(\delta^i, \ataconf_0)$, which is similar to the distinguished initial location $l_0$ of the \ac{ATA} \ata.
It is necessary to initialize the \ac{ATA} with the fluents that are satisfied in the initial situation of the program.

We usually omit the subscript $\Delta / \phi$ if $\Delta$ and $\phi$ are clear from the context.
For a state $\syncstate = (\la z, \rho \ra, \ataconf) \in \syncstates$, we also write $\syncclocks(\syncstate)$ for the set that contains all configurations from \ataconf and all clock valuations from $z$, i.e.,
\[
  \syncclocks(\syncstate) \eqdef \ataconf \cup \{ (c, v) \mid c \in \clockset, w[c, z] = v \}
\]
We also write $\syncconfs = \ataconfs \cup (\clockset \times \realpos)$ for the set of all such configurations.
Furthermore, we may also write $\clockvaluation_{\syncstate}$ for the corresponding clock valuation with $\clockvaluation_{\syncstate}(c) = v$ for each $(c, v) \in \syncclocks(\syncstate)$.
The set $\syncclocks(\syncstate)$ completely captures the time component of the state: it contains all clock valuations of the \ac{ATA} as well as all clock valuations of the program.
We will later use $\syncclocks(\syncstate)$ to define clock regions for the states of \synclts.

\begin{example}[Synchronous Product] \label{ex:sync-product}
  \begin{figure}[tb]
    \centering
    \includestandalone[width=\textwidth]{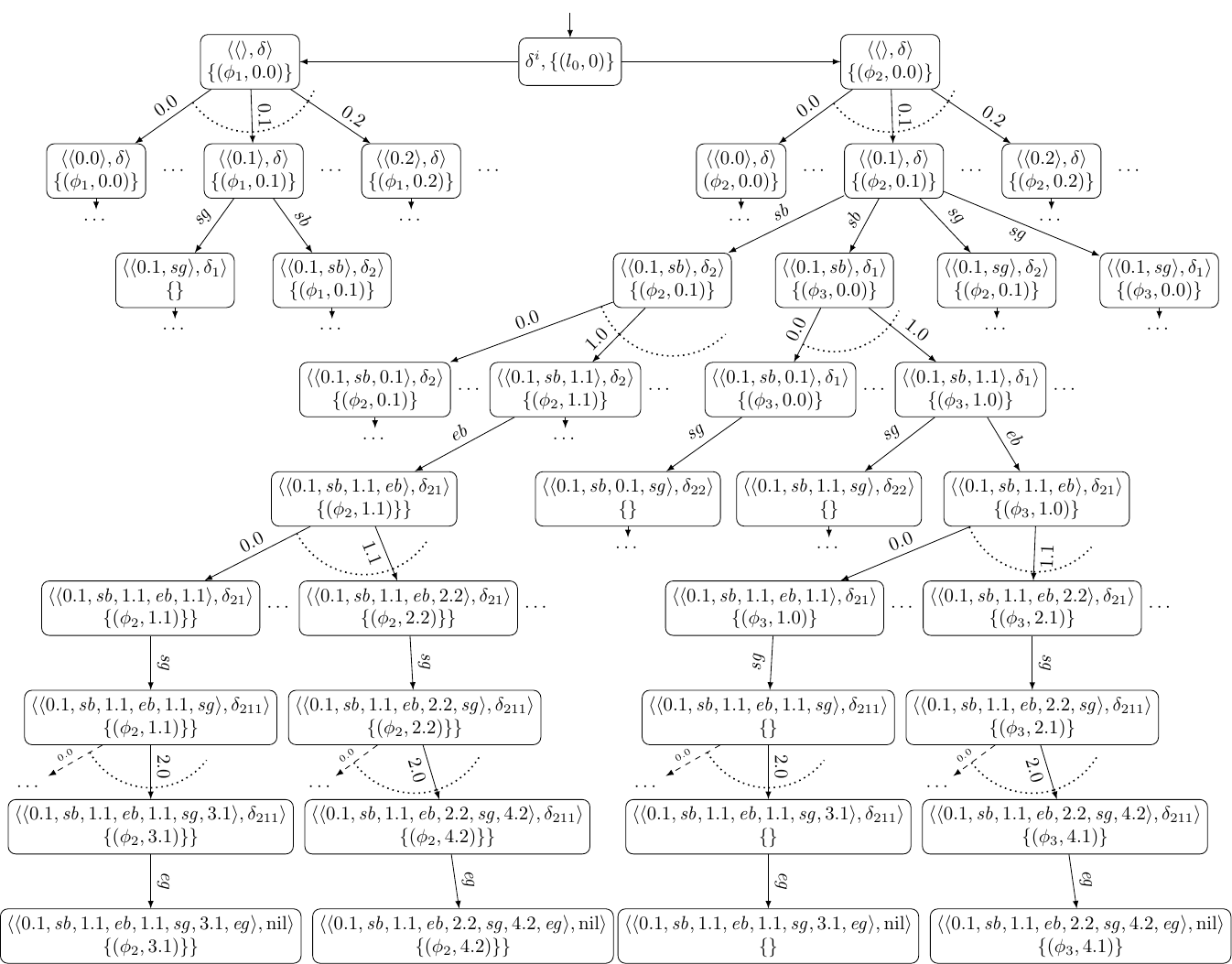}
    \caption[The synchronous product.]{
      The synchronous product of the program and \ac{ATA} from \autoref{ex:control-problem}, where actions are abbreviated as follows:
      $\sboot \eqdef \sac{\bootcam}$, $\eboot \eqdef \eac{\bootcam}$, $\sgrasp \eqdef \sac{\grasp(m_2, o_1)}$, $\egrasp \eqdef \eac{\grasp(m_2, o_1)}$.
      Omitted time successors are indicated by dotted arcs.
    }
    \label{fig:sync-product}
  \end{figure}

  \autoref{fig:sync-product} shows the synchronous product for the control problem from \autoref{ex:control-problem}.
  Note that the \ac{LTS} has uncountably many states and is infinitely branching, because there is a time transition for each time increment $t \in \realpos$.
\end{example}

As the synchronous product \synclts contains all program executions and tracks whether the specification has been satisfied, it could in principle be used to determine a controller: if we can define a mapping that steers the program execution away from those states where both the program and the \ac{ATA} are accepting, then we can guarantee that the specification will never be violated.
However, there are two issues:
\begin{enumerate}
  \item The resulting tree is infinitely branching.
    In fact, for each time step, there is a succinct time successor for each $r \in \realpos$ and therefore uncountably many successors.
    Hence, for a particular state, we cannot directly iterate over all successors to check whether a controller step exists.
  \item The tree may contain infinite paths, e.g., if the program contains a non-terminating while loop.
    As we are only interested in finite traces of the program and hence in terminating executions, we can just ignore those paths.
    However, in order to do so, we need to detect those infinite paths.
\end{enumerate}

We start with the first issue by applying \emph{regionalization} to the synchronous product in \autoref{sec:regionalization}.
For the second issue, we will show in \autoref{sec:wsts} that we can define a suitable \acfi*{wqo} such that the resulting transition system is a \acfi*{WSTS}, where it is known that the subcovering problem is decidable.

\section{Regionalization} \label{sec:regionalization}


As we have seen above, there are uncountably many distinct alternatives for each time step in the synchronous product \synclts.
However, as can be seen in \autoref{fig:sync-product}, many of those time successors are very similar.
In fact, many states consist of the same remaining program, program traces that contain the same action but slightly different time points, and \ac{ATA} configurations with the same location but slightly different clock valuations.
Formally, this similarity is captured by \emph{time-abstract bisimulations}:
\begin{definition}[Time-Abstract Bisimulation] \label{def:time-abstract-bisimulation}
  An equivalence relation $R \subseteq \syncstates \times \syncstates$ is a \emph{time-abstract bisimulation} on \synclts if $(s_1, s_2) \in R$ with $s_1 = (\la z_1, \rho_1 \ra, G_1)$ and $s_2 = (\la z_2, \rho_2 \ra, G_2)$ implies:
  \begin{enumerate}
    \item for every fluent situation formula $\alpha$: $w, z_1 \models \alpha$ iff $w, z_2 \models \alpha$,
    \item $\rho_1 = \rho_2$,
    \item for every $a \in \stdactname$ and every $t \in \realpos$, $s_1 \synctrans[t]{a} s_1'$ implies there is a $s_2'$ and $t'$ such that $s_2 \synctrans[t']{a} s_2'$ and $(s_1', s_2') \in R$ (and vice versa).
      \qedhere
  \end{enumerate}
\end{definition}

The first condition states that the world is in the same state in both cases.
The second condition states that the remaining program of both states must be the same; if two states differ in the remaining program, they may have different successor states and therefore cannot be considered to be bisimilar.
Third, whenever the program allows to execute a timed action $(a, t)$ in $\syncstate_1$ resulting in some state $\syncstate_1'$, then there must be a timed action $(a, t')$ (possibly at a different time point) that is executable in $\syncstate_2$ and that results in a state $\syncstate_2'$ that is again bisimilar with $\syncstate_1'$.


A well-known concept to define a time-abstract bisimulation is \emph{regionalization}~\cite{alurTheoryTimedAutomata1994}:
Based on the fact that clock constraints may only mention natural numbers and therefore cannot distinguish two clock valuations with the same integer component, two clock values with the same integer part but possible different non-zero fractional part can be considered to be equivalent.
Integer clock values need to be treated separately as they can be distinguished from clock values with non-zero fractional part with strict inequality, e.g., $c > 2$ can distinguish the clock values $c = 2.0$ and $c = 2.1$.
For multiple clocks, we also need to consider the ordering of clocks defined by their fractional parts.
To see why, consider two clock constraints $c_1 \leq 1$ and $c_2 > 2$.
In a state with the clock valuation $\clockvaluation_1$ with $\clockvaluation_1(c_1) = 0.4$ and $\clockvaluation_2(c_2) = 1.7$, the time successor $\clockvaluation_1 + 0.4$ satisfies both clock constraints.
However, for the clock valuation $\clockvaluation_2$ with $\clockvaluation_2(c_1) = 0.6$ and $\clockvaluation_2(c_2) = 1.4$, no time successor satisfies both constraints.
Therefore, the two states should not be considered to be equivalent.
For this reason, we must keep track of the ordering of the fractional parts of the clock valuations.

Additionally, for a given program and \ac{MTL} specification, the maximal constant appearing anywhere in the program or specification is known and is fixed to a value $K \in \naturals$.
Hence, if a clock reaches a value $r > K$, then no clock constraint may distinguish any of the time successors.
Therefore, we only need to consider finitely many clock regions and we may put all clock valuations exceeding $K$ into the same region.




With these considerations, we can use clock regions from \autoref{def:regionalization} to obtain a time-abstraction bisimulation:
\clockregions*

The clock value $\top$ represents any clock value greater than the maximal constant $K$.
For convenience, we may write $u > \maxconst$ if $u = \top$ and we define the fractional part $\fract(\top)$ of $\top$ to be always $0$.
Also, note that the number of regions (i.e., the equivalence classes of $\clockequiv$) is finite.
We demonstrate clock regions in our setting with an example:
\begin{example}[Clock Regions]
  First, for $\maxconst = 3$, we obtain the following region equivalences:
  \begin{alignat*}{3}
    0.0 &\clockequiv[3] 0.0 \quad & 1.0 &\clockequiv[3] 1.0 \quad & 0.0 &\not\clockequiv[3] 1.0
    \\
    0.5 &\clockequiv[3] 0.8 \quad & 1.1 &\clockequiv[3] 1.9 \quad & \top &\clockequiv[3] \top
    \\
    1.0 &\not\clockequiv[3] 1.1 & \quad 0.0 &\not\clockequiv[3] \top \quad & 1.4 &\not\clockequiv[3] 2.5
  \end{alignat*}
  For clock valuations of 3 clocks and again with $\maxconst = 3$, we obtain the following region equivalences:
  \begin{alignat*}{2}
    (0.5, 0.2, 2.0) &\regequiv[3] (0.8, 0.3, 2.0)
    \qquad & (0.5, 0.2, 2.0) &\not\regequiv[3] (0.5, 0.6, 2.0)
    \\
    (0.1, 0.2, 2.2) &\not\regequiv[3] (0.8, 0.8, 2.6)
    \qquad & (0.0, 0.5, 2.2) &\not\regequiv[3] (1.0, 0.4, 2.6)
    \\
    (0.1, \top, 2.2) &\not\regequiv[3] (0.4, 3.0, 2.6)
    \qquad & (0.1, \top, 2.2) &\regequiv[3] (0.4, \top, 2.6)
  \end{alignat*}
\end{example}

Clock regions can be used to conjoin states in the synchronous product \synclts such that each state has a finite number of successors.
One way to do so is to replace each clock value by the respective clock region such that the node represents all nodes where each clock is in the same equivalence class.
This approach is commonly taken for \acp{TA}, as shown in \autoref{fig:region-lts}.
Here, we take a slightly different approach adapted from \parencite{ouaknineDecidabilityComplexityMetric2007}: Instead of replacing each clock value by the corresponding equivalence class, we directly determine a representative of each equivalence class such that we have exactly one time successor for each equivalence class.
To do so, we first define the \emph{region increment}, a canonical time increment that uniquely represents all time increments leading to the next region, as well as the \emph{time successor}, which is the clock valuation corresponding to a region increment:

\begin{definition}[Region Increments and Time Successors] \label{def:region-increment}
  Let $\syncclocks = \syncclocks(\syncstate)$ for some state $\syncstate \in \syncstates$ of \synclts.
  If $C$ is non-empty, let $\mu = \max\{\fract(v) \mid (s, v) \in C\}$ be the maximal fractional part of the clock values appearing in $C$.
  We define the \emph{region increment} $\rincr(C) \in \realpos$ as follows:
  \begin{itemize}
    \item if $v = \top$ for every $(s, v) \in C$, then $\rincr(C) = 0$,
    \item if $(s, v) \in C$ for some integer clock value $v \in [0, K]$, then $\rincr(C) = \frac{1-\mu}{2}$,
    \item otherwise, $\rincr(C) = 1 - \mu$.
  \end{itemize}
  We define the \emph{time successor} of $C$ to be the configuration $\rnext(C) = C + \rincr(C)$.
  We inductively define the $n$-increment $\rincr^n(C)$ of $C$ and the $n$th successor $\rnext^n(C)$ of $C$:
  \begin{align*}
    \rincr^0(C) &\eqdef 0
    \\
    \rincr^n(C) &\eqdef \rincr(\rnext^{n-1}(C)) + \rincr^{n-1}(C)
    \\
    \rnext^n(C) &\eqdef C + \rincr^n(C)
  \end{align*}

  Furthermore, we define the set of all possible increments $\rincr^*(C)$  and the set of all possible time successors $\rnext^*(C)$ as follows:
  \begin{align*}
    \rincr^*(C) &\eqdef \bigcup_{n \in \naturals} \rincr^n(C)
    \\
    \rnext^*(C) &\eqdef \bigcup_{n \in \naturals} \rnext^n(C)
  \end{align*}

  For a state $\syncstate \in \synclts$, we also write $\rincr(\syncstate)$ for $\rincr(\syncclocks(\syncstate))$, i.e., for the time increment of all clocks defined by the state $\syncstate$.
  Similarly, for a set of states $\detstate \subseteq \syncstates$, we write $\rincr(\detstate)$ for the time increment defined by the union of all clock valuations in \detstate.
\end{definition}

The following example shows time successors for some clock valuations that may occur in the synchronous product \synclts from \autoref{ex:sync-product}:
\begin{example}[Time Successors] \label{ex:time-successors}
  ~\\
  Consider the state $\syncstate = (\la \la 0.5, \sac{\bootcam}\ra, \delta_2 \ra, \{ (\phi_2, 0.5) \})$ with set of clocks $C = \{ c_b, c_{\phi_2} \}$ with valuations $\clockvaluation(c_b) = 0$ and $\clockvaluation(c_{\phi_2}) = \frac{1}{2}$.
  Assume that the maximal constant is $K = 2$.
  The following table shows the clock regions increments $\rincr^i(C)$ and their corresponding time successors, i.e., the clock valuations after each increment:
  \begin{center}
    \begin{tabular}{*7{Cc}}
      \toprule
      $i$ & increment & \shortstack{acc.\ increment \\ $\rincr^i(C)$} & $\clockvaluation(c_b)$ & \shortstack{region index \\ of $c_b$ } & $\clockvaluation(c_{\phi_2})$ & \shortstack{region index \\ of $c_{\phi_2}$ }
      \\
      \midrule
      $0$ & & $0$ & $0$ & $0$ & $\frac{1}{2}$ & $1$
      \\
      $1$ & $\frac{1}{4}$ & $\frac{1}{4}$ & $\frac{1}{4}$ & $1$ & $\frac{3}{4}$ & $1$
      \\
      $2$ & $\frac{1}{4}$ & $\frac{1}{2}$ & $\frac{1}{2}$ & $1$ & $1$ & $2$
      \\
      $3$ & $\frac{1}{4}$ & $\frac{3}{4}$ & $\frac{3}{4}$ & $1$ & $\frac{5}{4}$ & $3$
      \\
      $4$ & $\frac{1}{4}$ & $1$ & $1$ & $2$ & $\frac{3}{2}$ & $3$
      \\
      $5$ & $\frac{1}{4}$ & $\frac{5}{4}$ & $\frac{5}{4}$ & $3$ & $\frac{7}{4}$ & $3$
      \\
      $6$ & $\frac{1}{4}$ & $\frac{3}{2}$ & $\frac{3}{2}$ & $3$ & $2$ & $4$
      \\
      $7$ & $\frac{1}{4}$ & $\frac{7}{4}$ & $\frac{7}{4}$ & $3$ & $\top$ & $5$
      \\
      $8$ & $\frac{1}{4}$ & $2$ & $2$ & $4$ & $\top$ & $5$
      \\
      $9$ & $\frac{1}{2}$ & $\frac{5}{2}$ & $\top$ & $5$ & $\top$ & $5$
      \\
      \bottomrule
    \end{tabular}
  \end{center}
  The table shows for each $i$ the increment, the accumulated increment $\rincr^i(C)$ (which is the sum over all previous increments), as well as the clock valuation and clock region for clock after each increment.
  In the first row, we can see that the maximal fractional part is $\mu = \frac{1}{2}$.
  As the clock $c_b$ has an integer value, the first region increment is $\rincr^1(C) = \frac{1 - \mu}{2} = \frac{1}{4}$.
  For the next increment, there is no clock with an integer value and the maximal fractional part is $\mu = \frac{3}{4}$.
  Thus, the second region increment is $\rincr(\rnext^1(C)) = 1 - \mu = \frac{1}{4}$ and therefore $\rincr^2(C) = \rincr(\rnext^1(C)) + \rincr^1(C) = \frac{1}{2}$, and so on.
  Eventually, both clock values reach a value larger than the maximal constant $K = 2$ and therefore have the value $\top$, which corresponds to the maximal region $2K+1 = 5$.
  Note that in the last row and different to all other rows, the increment is $\frac{1}{2}$.
  This is because both clocks have integer values and the maximal fractional part is therefore $\mu = 0$, resulting in an increment of $\rincr(\rnext^8(C)) = \frac{1}{2}$.

  As the following example shows, the increments may vary from step to step, not only in the last step: Consider again the clock set $C = \{ c_b, c_{\phi_2} \}$ as above, but this time with initial values of $\clockvaluation(c_b) = 0$ and $\clockvaluation(c_{\phi_2}) = \frac{3}{4}$.
  The resulting increments look as follows:
  \begin{center}
    \begin{tabular}{*7{Cc}}
      \toprule
      $i$ & increment & \shortstack{acc.\ increment \\ $\rincr^i(C)$} & $\clockvaluation(c_b)$ & \shortstack{region index \\ of $c_b$ } & $\clockvaluation(c_{\phi_2})$ & \shortstack{region index \\ of $c_{\phi_2}$ }
      \\
      \midrule
      $0$ & & $0$ & $0$ & $0$ & $\frac{3}{4}$ & $1$
      \\
      $1$ & $\frac{1}{8}$ & $\frac{1}{8}$ & $\frac{1}{8}$ & $1$ & $\frac{7}{8}$ & $1$
      \\
      $2$ & $\frac{1}{8}$ & $\frac{1}{4}$ & $\frac{1}{4}$ & $1$ & $1$ & $2$
      \\
      $3$ & $\frac{3}{8}$ & $\frac{5}{8}$ & $\frac{5}{8}$ & $1$ & $\frac{11}{8}$ & $3$
      \\
      $4$ & $\frac{3}{8}$ & $1$ & $1$ & $2$ & $\frac{7}{4}$ & $3$
      \\
      $5$ & $\frac{1}{8}$ & $\frac{9}{8}$& $\frac{9}{8}$ & $3$ & $\frac{15}{8}$ & $3$
      \\
      $6$ & $\frac{1}{8}$ & $\frac{5}{4}$& $\frac{5}{4}$ & $3$ & $2$ & $4$
      \\
      $7$ & $\frac{3}{8}$ & $\frac{13}{8}$& $\frac{13}{8}$ & $3$ & $\top$ & $5$
      \\
      $8$ & $\frac{3}{8}$ & $2$ & $2$ & $4$ & $\top$ & $5$
      \\
      $9$ & $\frac{1}{2}$ & $\frac{5}{2}$ & $\top$ & $5$ & $\top$ & $5$
      \\
      \bottomrule
    \end{tabular}
  \end{center}
  As we can see, the increment in each step alters between $\frac{1}{8}$ and $\frac{3}{8}$, depending on the clock valuations and the maximal fractional part $\mu$.
\end{example}

In the following, we will apply regionalization to the \ac{LTS} \synclts.
As a first step, we note that every time successor of a reachable state of \synclts is also a state of \synclts:
\begin{lemmaE} \label{lma:next-defines-successors}
  Let $\syncstate$ be a state reachable in \synclts.
  Then for every $i$, there is a unique $\syncstate'$ such that $\syncstate \synctranstime{\rincr^i(\syncstate)} \syncstate'$ and $\syncclocks(\syncstate') = \rnext^i(\syncclocks(\syncstate))$.
\end{lemmaE}
\begin{proofE}
  By definition, $(\la z, \rho \ra, \ataconf) \synctranstime{d} (\la z^*, \rho^* \ra, \ataconf^*)$ if $\la z, \rho \ra \wtarrow[d] \la z^*, \rho^* \ra$ and $\ataconf \atatranstime{d} \ataconf^*$.
  By \autoref{def:tesg-trans}, for every $d$: $\la z, \rho \ra \wtarrow[d] \la z \cdot t, \rho \ra$ and such that $t = \ztime(z) + d$.
  In particular, there are no restrictions on $d$.
  Similarly, by \autoref{def:ata-lts}, there is a transition $\ataconf \atatranstime{d} \ataconf^*$ for every $d$.
  Therefore, there is some time transition for $\rincr^i(\syncstate)$.
  It remains to be shown that the successor state is indeed $\rnext^i(\syncstate)$.
  By \autoref{def:tesg-world}, for every $c \in \clockset$, $w[c, z \cdot t] = w[c, z] + t - \ztime(z) = w[c, z] + d$.
  Furthermore, by \autoref{def:ata-lts}, $\ataconf \atatranstime{d} \ataconf^*$ if $\ataconf^* = \{ l, v + d \mid (l, v) \in \ataconf \}$.
  Therefore, $\syncclocks(\syncstate') = \syncclocks(\syncstate) + \rincr^i(\syncstate) = \rnext^i(\syncclocks(\syncstate))$.
  As for uniqueness, note that both \tesg and \ac{ATA} time transitions lead to a unique successor.
\end{proofE}

As there is a unique time successor for every $i$, it directly follows that the set of time successors of a state $\syncstate \in \synclts$ is finite.
This will later allow us to restrict \synclts to a finitely branching \ac{LTS} that still represents all possible paths in \synclts.
Also, as the time successor for each $i$ is unique, we can write $\rnext^i(\syncstate)$ for the unique $\syncstate'$ with $\syncstate \synctranstime{\rincr^i(\syncstate)} \syncstate'$.

We can now define an equivalence relation $\operatorname{\syncbisim}$ that formally captures the equivalence of states with respect to clock regions:
\begin{definition} \label{def:sync-bisim}
  We define the equivalence relation $\operatorname{\syncbisim} \subseteq \syncstates \times \syncstates$ such that for $\syncstate_1 = (\la z_1, \rho_1 \ra, \ataconf_1) \in \syncstates$ and $\syncstate_2 = (\la z_2, \rho_2 \ra, \ataconf_2) \in \syncstates$, $\syncstate_1 \syncbisim \syncstate_2$ iff
  \begin{enumerate}
    \item for every $F(\vec{n}) \in \primformulas$, $w[F(\vec{n}), z_1] = w[F(\vec{n}), z_2]$,
    \item for every $t \in \primterms$, $w[t, z_1] = w[t, z_2]$,
    \item $\rho_1 = \rho_2$,
    \item there is a bijection $f: \syncclocks(\syncstate_1) \rightarrow \syncclocks(\syncstate_2)$ such that:
      \begin{enumerate}
        \item $f(s, u) = (t, v)$ implies $s = t$ and $u \clockequiv v$;
        \item If $f(s, u) = (t, v)$ and $f(s', u') = (t', v')$, then $\fract(u) \leq \fract(u')$ iff $\fract(v) \leq \fract(v')$.
          \qedhere
      \end{enumerate}
  \end{enumerate}
\end{definition}

Intuitively, two states $\syncstate_1$ and $\syncstate_2$ are equivalent if they have
\begin{enumerate*}[label=(\arabic*)]
  \item the same relational fluent values,
  \item the same functional fluent values,
  \item the same remaining programs, and
  \item region-equivalent clock valuations.
\end{enumerate*}

We continue by showing that $\operatorname{\syncbisim}$ is indeed a time-abstract bisimulation.
We first show that equivalent states satisfy the same static formulas:
\begin{lemmaE}\label{lma:sync-states-satisfy-same-static-formulas}
  Let $\syncstate_1, \syncstate_2 \in \syncstates$ with $\syncstate_1 = (\la z_1, \rho_1 \ra, \ataconf_1)$, $\syncstate_2 = (\la z_2, \rho_2 \ra, \ataconf_2)$, and $\syncstate_1 \syncbisim \syncstate_2$.
  Let $\alpha$ be a static formula.
  Then $w, z_1 \models \alpha$ iff $w, z_2 \models \alpha$.
\end{lemmaE}
\begin{proofE}
  By structural induction on $\alpha$.
  \begin{itemize}
    \item Let $\alpha = F(\vec{n})$ be a primitive formula.
      By definition, $w[F(\vec{n}), z_1] = w[F(\vec{n}), z_2]$ and so  $w, z_1 \models \alpha$ iff $w, z_2 \models \alpha$.
    \item Let $\alpha = c \bowtie n$ be a clock formula, where $c$ is a clock term and $n \in \naturals$ is some constant.
      By definition of \syncbisim, $\lvert c \rvert^{z_1}_w = \lvert c \rvert^{z_2}_w = q$ for some $q \in \stdclockname$.
      As $s_1 \syncbisim s_2$, there is a bijection $f$ with $f(q, u) = (q, v)$ and $u \clockequiv v$.
      From $u \clockequiv v$, it directly follows that $u \bowtie n$ iff $v \bowtie n$ and therefore $w, z_1 \models c \bowtie n$ iff $w, z_2 \models c \bowtie n$.
    \item Let $\alpha = \beta \wedge \gamma$.
      It directly follows by induction that $w, z_1 \models \beta$ iff $w, z_2 \models \beta$ and $w, z_1 \models \gamma$ iff $w, z_2 \models \gamma$.
      Therefore, $w, z_1 \models \beta \wedge \gamma$  iff $w, z_2 \models \beta \wedge \gamma$.
    \item Let $\alpha = \neg \beta$.
      It directly follows by induction that $w, z_1 \models \beta$ iff $w, z_2 \models \beta$.
      Therefore, $w, z_1 \models \neg\beta$  iff $w, z_2 \models \neg \beta$.
    \item Let $\alpha = \forall x.\, \beta$.
      By induction, for each $n \in \stdname_x$ of the same sort as $x$, $w, z_1 \models \beta^x_n$ iff $w, z_2 \models \beta^x_n$.
      Therefore,  $w, z_1 \models \forall x.\, \beta$ iff $w, z_2 \models \forall x.\, \beta$.
      \qedhere
  \end{itemize}
\end{proofE}

For the next step, we use a well-known result that for any two region-equivalent clock valuations $\clockvaluation_1$ and $\clockvaluation_2$ and for an arbitrary increment $t$ of $\clockvaluation_1$, there is some increment $t'$ of $\clockvaluation_2$ such that the two resulting clock valuations are again region-equivalent:
\begin{proposition}[\parencite{alurTheoryTimedAutomata1994,ouaknineDecidabilityMetricTemporal2005}]\label{prop:reg-successor}
  Let $\clockvaluation_1$ and $\clockvaluation_2$ be two clock valuations over a set of clocks \clockset such that $\clockvaluation_1 \regequiv \clockvaluation_2$.
  Then for all $t \in \realpos$ there exists a $t' \in \realpos$ such that $\clockvaluation_1 + t \regequiv \clockvaluation_2 + t'$.
\end{proposition}

We are now ready to show that $\operatorname{\syncbisim}$ is a time-abstract bisimulation:
\begin{theoremE} \label{thm:time-abstract-bisimulation}
  The equivalence relation $\operatorname{\syncbisim}$ is a time-abstract bisimulation on \synclts.
\end{theoremE}
\begin{proofE}
   Assume $\syncstate_1 = (\la z_1, \rho_1 \ra, \ataconf_1)$, $\syncstate_2 = (\la z_2, \rho_2 \ra, \ataconf_2) \in \syncstates$, $\syncstate_1 \syncbisim \syncstate_2$ and such that $f: \syncclocks(\syncstate_1) \rightarrow \syncclocks(\syncstate_2)$ is a bijection witnessing $\syncstate_1 \syncbisim \syncstate_2$.
  \begin{enumerate}
    \item By \autoref{lma:sync-states-satisfy-same-static-formulas}, it directly follows for every fluent situation formula  $\alpha$ that $w, z_1 \models \alpha$ iff $w, z_2 \models \alpha$.
    \item By definition of $\syncbisim$, $\rho_1 = \rho_2$.
    \item
      \begin{description}
        \item[Time step:]
          Assume $s_1 \synctranstime{d_1} s_1'$.
          Let $t_1 = \ztime(z_1) + d_1$.
          By \autoref{def:tesg-trans}, $\la z_1, \rho_1 \ra \wtarrow[d_1] \la z_1', \rho_1' \ra$ with $z_1' = z_1 \cdot t_1$ and $\rho_1' = \rho_1$.
          Furthermore, by \autoref{def:sync-product}, there is a $\ataconf_1'$ such that $\ataconf_1 \overset{d_1}{\rightsquigarrow} \ataconf_1'$.
          From $s_1 \syncbisim s_2$, it follows that $\clockvaluation_{s_1} \regequiv \clockvaluation_{s_2}$.
          By \autoref{prop:reg-successor}, there is a $d_2 \in \realpos$ and a clock valuation $\clockvaluation_{s_2}$ such that $\clockvaluation_{s_1} + d_1 \regequiv \clockvaluation_{s_2} + d_2$.
          With $t_2 = \ztime(z_2) + d_2$, $z_2' = z_2 \cdot t_2$, and $\rho_2' = \rho_2$ we obtain $\la z_2, \rho \ra \wtarrow[d_2] \la z_2', \rho_2 \ra$.
          With $\ataconf_2' = \ataconf_2 + d_2$, we obtain $\ataconf_2 \overset{d_2}{\rightsquigarrow} \ataconf_2'$.
          We construct the bijection $f': \syncclocks(\syncstate_1') \rightarrow \syncclocks(\syncstate_2')$ as follows:
          As $\syncstate_1 \syncbisim \syncstate_2$, there is a bijection $f: \syncclocks(\syncstate_1) \rightarrow \syncclocks(\syncstate_2)$ satisfying the criteria from \autoref{def:sync-bisim}.
          Note that for each $(c_1', v_1') \in \syncclocks(\syncstate_1')$, there is a $(c_1, v_1) \in \syncclocks(\syncstate_1)$ such that $(c_1', v_1') = (c_1, v_1 + d_1)$.
          Similarly, for each $(c_2', v_2') \in \syncclocks(\syncstate_2')$, there is a $(c_2, v_2) \in \syncclocks(\syncstate_2)$ such that $(c_2', v_2') = (c_2, v_2 + d_2)$.
          Also, from $\clockvaluation_{s_1} + d_1 \regequiv \clockvaluation_{s_2} + d_2$, it follows that $v_1 + d_1 \clockequiv v_2 + d_2$.
          Therefore, let $f'$ be a bijection such that for each $(c_1, v_1) \in \syncclocks(\syncstate_1)$, $f'(c_1, v_1 + d_1) = (c_2, v_2 + d_2)$ if $f(c_1, v_1) = (c_2, v_2)$.
          Clearly, $f'$ is a witness for $\syncstate_1' \syncbisim \syncstate_2'$.
        \item[Action step:]
          Assume $\syncstate_1 \synctranssym{p} \syncstate_1'$.
          Thus, $\la z_1, \rho_1 \ra \wsarrow \la z_1', \rho_1' \ra$ and $\ataconf_1 \atatranssym{F_1} \ataconf_1'$ with  $f \in F_1$ iff $w[f, z_1'] = 1$.
          We first show for an arbitrary program $\delta$ that $\la z_1, \delta \ra \in \final$ iff $\la z_2, \delta \ra \in \final$.
          We do so by structural induction on $\delta$:
          \begin{itemize}
            \item Let $\delta = \alpha?$, where $\alpha$ is a static formula.
              As $s_1 \syncbisim s_2$, it directly follows from \autoref{lma:sync-states-satisfy-same-static-formulas} that $w, z_1 \models \alpha$ iff $w, z_2 \models \alpha$.
            \item For all other cases, the claim follows by structural induction and the transition rules from \autoref{def:tesg-trans}.
          \end{itemize}
          Next, we show that for every program $\delta$, $\la z_1, \delta \ra \wsarrow \la z_1 \cdot p, \delta' \ra$ implies $\la z_2, \delta \ra \wsarrow \la z_2 \cdot p, \delta' \ra$ for some $\delta'$.
          We do so by structural induction on $\delta$:
          \begin{itemize}
            \item Let $\delta = a$ for some primitive action term $a$.
              Therefore, $p = \lvert a \rvert^{z_1}_w$ and $\la z_1, \delta \ra \wsarrow \la z_1 \cdot p, \nil \ra$.
              With \autoref{lma:tesg-bat-static-time-invariance} and $\symtrace(z_1) = \symtrace(z_2)$, it also follows that $\lvert a \rvert^{z_2}_w = \lvert a \rvert^{z_1}_w$ and thus also $\lvert a \rvert^{z_2}_w = \lvert a \rvert^{z_1}_w$.
              Therefore, $\la z_2, \delta \ra \wsarrow \la z_2 \cdot p, \nil \ra$.
            \item Let $\delta = \delta_1; \delta_2$.
              If $\la z_1, \delta_1 \ra \not\in \final$, then also $\la z_2, \delta_1 \ra \not\in \final$.
              By induction, from $\la z_1, \delta_1 \ra \wsarrow \la z_1 \cdot p, \delta' \ra$, it follows that $\la z_2, \delta_1 \ra \wsarrow \la z_2 \cdot p, \delta' \ra$.
              Otherwise, if $\la z_1, \delta_1 \ra \in \final$, then also $\la z_2, \delta_1 \ra \in \final$.
              Again, by induction, from $\la z_1, \delta_2 \ra \wsarrow \la z_1 \cdot p, \delta' \ra$, it follows that $\la z_2, \delta_2 \ra \wsarrow \la z_2 \cdot p, \delta' \ra$.
            \item Let $\delta = \delta_1 \vert \delta_2$.
              If $\la z_1, \delta \ra \wsarrow \la z_1 \cdot p, \delta' \ra$, then $\la z_1, \delta_1 \ra \wsarrow \la z_1 \cdot p, \delta_1' \ra$ and $\delta' = \delta_1'$ or $\la z_2, \delta_2 \ra \wsarrow \la z_2 \cdot p, \delta_2' \ra$ and $\delta' = \delta_2'$.
              If $\la z_1, \delta_1 \ra \wsarrow \la z_1 \cdot p, \delta_1' \ra$, then, by induction, $\la z_2, \delta_1 \ra \wsarrow \la z_2 \cdot p, \delta_1' \ra$.
              Similarly, if $\la z_1, \delta_2 \ra \wsarrow \la z_1 \cdot p, \delta_2' \ra$, then, by induction, $\la z_2, \delta_2 \ra \wsarrow \la z_2 \cdot p, \delta_2' \ra$.
              Hence, $\la z_2, \delta \ra \wsarrow \la z_2 \cdot p, \delta' \ra$.
            \item Let $\delta = \delta_1 \| \delta_2$.
              If $\la z_1, \delta \ra \wsarrow \la z_1 \cdot p, \delta' \ra$, then $\la z_1, \delta_1 \ra \wsarrow \la z_1 \cdot p, \delta_1' \ra$ and $\delta' = \delta_1' \| \delta_2$ or $\la z_2, \delta_2 \ra \wsarrow \la z_2 \cdot p, \delta_2' \ra$ and $\delta' = \delta_2' \| \delta_2$.
              If $\la z_1, \delta_1 \ra \wsarrow \la z_1 \cdot p, \delta_1' \ra$, then, by induction, $\la z_2, \delta_1 \ra \wsarrow \la z_2  \cdot p, \delta_1' \ra$.
              Similarly, if $\la z_1, \delta_2 \ra \wsarrow \la z_1 \cdot p, \delta_2' \ra$, then, by induction, $\la z_2, \delta_2 \ra \wsarrow \la z_2  \cdot p, \delta_2' \ra$.
              Hence, $\la z_2, \delta \ra \wsarrow \la z_2 \cdot p, \delta' \ra$.
            \item Let $\delta = \delta_1^*$.
              If $\la z_1, \delta \ra \wsarrow \la z_1 \cdot p, \delta' \ra$, then $\la z_1, \delta_1 \ra \wsarrow \la z_1 \cdot p, \delta_1' \ra$ and $\delta' = \delta_1'; \delta^*$.
              Then, by induction, $\la z_2, \delta_1 \ra \wsarrow \la z_2 \cdot p, \delta_1' \ra$ and so $\la z_2, \delta_1^* \ra \wsarrow \la z_2 \cdot p, \delta_1'; \delta^* \ra$.
          \end{itemize}
          Therefore, from $\la z_1, \rho_1 \ra \wsarrow \la z_1', \rho_1' \ra$, it follows that $\la z_2, \rho_2 \ra \wsarrow \la z_2', \rho_2' \ra$ with $\sym(z_1') = \sym(z_2')$ and $\rho_1' = \rho_2'$.
          Furthermore, with $\sym(z_1') = \sym(z_2')$ and \autoref{lma:tesg-bat-static-time-invariance}, it follows that $w[\phi, z_1'] = w[\phi, z_2']$ for every $\phi \in \primformulas$.
          Next, we show that $\ataconf_2 \atatranssym{F} \ataconf_2'$ with  $\phi \in F$ iff $w[\phi, z_2'] = 1$.
          First, note that $F_1 = F_2$ because $w[\phi, z_1'] = w[\phi, z_2']$ for each $\phi \in \primformulas$.
          By \autoref{remark:ata-dnf}, we know that $\ataconf_1' = \bigcup_i A_i[v_1^{(i)} + d_1]$, where for each $i$, the set of atoms $A_i$ is a clause in the disjunctive normal form for $\atatrans(s_1^{(i)}, F_1)$.
          Let $\ataconf_2' = \bigcup_i A_i[v_2^{(i)} + d_2]$.
          With $\clockvaluation_{s_1} + d_1 \regequiv \clockvaluation_{s_2} + d_2$, $G_1 \overset{F_1}{\rightarrow} G_1'$, and $F_1 = F_2$, we obtain $\ataconf_2 \overset{F_2}{\rightarrow} \ataconf_2'$.
          Therefore, $s_2 \synctranssym{p} s_2'$.
          \\
          It remains to be shown that $\syncstate_1' \syncbisim \syncstate_2'$:
          We have already established that $\sym(z_1') = \sym(z_2')$ and $\rho_1' = \rho_2'$.
          We define the bijection $f': \syncclocks(\syncstate_1') \rightarrow \syncclocks(\syncstate_2)'$ as follows:
          We can write $C_1'$ as $C_1' = \{ (c_{1'}^{(i)}, u_{1'}^{(i)}) \}_i$, $\ataconf_1'$ as $\ataconf_1' = \{ (s_{1'}^{(i)}, v_{1'}^{(i)}) \}_i$, $C_2' = \{ (c_{2'}^{(i)}, u_{2'}^{(i)}) \}_i$, and $\ataconf_2'$ as $\ataconf_2' = \{ (s_{2'}^{(i)}, v_{2'}^{(i)}) \}_i$.
          Note that for every $c \in \clockset$, $w, z_1' \models \reset(c)$ iff $w, z_2' \models \reset(c)$.
          Therefore, $(c, 0) \in C_1'$ iff $(c, 0) \in C_2'$.
          If $w, z_1' \not\models \reset(c)$, then $w[c, z_1'] = w[c, z_1] + d_1$ and $w[c, z_2'] = w[c, z_2] + d_2$.
          Thus, for each $c \in \clockset$, we can set $f'(c_i, \clockvaluation_{s_1}(c_i) + d_1) = (c_i, \clockvaluation_{s_2}(c_i) + d_2)$ if $(c_i, \clockvaluation_{s_1}(c_i) + d_1) \in C_1')$ and $f'(c_i, 0) = (c_i, 0)$ otherwise.
          Similarly, for each $(s_{1'}^{(i)}, u_{1'}^{(i)}) \in \ataconf_1'$, let $f'(s_{1'}^{(i)}, u_{1'}^{(i)}) = (s_{2'}^{(i)}, u_{2'}^{(i)})$ if $(s_{1'}^{(i)}, u_{1'}^{(i)}) \in \ataconf_{1}'$ and $f'(s_{1'}^{(i)}, 0) = (s_{1'}^{(i)}, 0)$ otherwise.
          As $\clockvaluation_{s_1} + d_1 \regequiv \clockvaluation_{s_2} + d_2$, $f'$ is a bijection that witnesses $\syncstate_1' \syncbisim \syncstate_2'$.
          \qedhere
      \end{description}
  \end{enumerate}
\end{proofE}

Using the time-abstract bisimulation $\operatorname{\syncbisim}$, we can now show that the region increments indeed capture all time successors:
\begin{lemmaE}\label{lma:time-abstract-bisimulation}
  Let $\syncstate \synctranstime{t} \syncstate^*$.
  Then $\syncstate^* \syncbisim \syncstate'$ for some $\syncstate' \in \rnext^*(\syncstate)$.
\end{lemmaE}
\begin{proofE}
  \begin{enumerate}
    \item If $t = \rincr^i(s)$ for some $i$, then $s^* = \rnext^i(s)$ and therefore $s^* \in \rnext^*(s)$.
    \item Assume there is an $i$ such that $\rincr^i(\syncstate) < t < \rincr^{i+1}(\syncstate)$.
      Let $\syncstate' = (\la z', \rho' \ra, \ataconf') = \rnext^i(\syncstate)$, $C = \ataconf' \cup G_{z'}$, and $\mu = \max\{\fract(v) \mid (c, v) \in C \}$.
      We distinguish two cases:
      \begin{enumerate}
        \item Assume there is some $(c, v) \in C$ for some integer clock value $v \in [0, K]$.
          Then for every $\varepsilon \in \realpos$ with $0 < \varepsilon < 1 - \mu$, $\rnext^i(\syncstate) + \varepsilon \syncbisim \rnext^{i+1}(\syncstate)$:
          Let $f$ be a bijection $f: C \rightarrow C + \varepsilon$ such that $f(c, v) = (c, v + \varepsilon)$.
          Clearly, for every  $(c, v) \in \rnext^i(\syncstate)$, $\lfloor v + \varepsilon \rfloor = \lfloor v + 1 - \mu \rfloor$ and $\lceil v + \varepsilon \rceil = \lceil v + 1 - \mu \rceil$, therefore $v +  \varepsilon \clockequiv v + 1 - \mu$.
          Furthermore, for every  $(c, v) \in C$: $\fract(v + \varepsilon) = \fract(v) + \varepsilon$ and $\fract(v + 1 - \mu) = \fract(v) + 1 - \mu$.
          Therefore, for every $(s_1, v_1), (s_2, v_2) \in C$: $\fract(v_1 + \varepsilon) \leq \fract(v_2 + \varepsilon)$ iff $\fract(v_1 + 1 - \mu) \leq \fract(v_2 + 1 - \mu)$.
          Thus, $f$ satisfies the criteria of \autoref{def:sync-bisim}.
          Now, let $\varepsilon^* = t - \rincr^i(s)$.
          As $\rincr^{i+1}(s) = \rincr^i(s) + \frac{1-\mu}{2}$, we obtain $0 < \varepsilon^* < 1 - \mu$.
          It follows that $s^* \syncbisim \rnext^{i+1}(\syncstate)$.
        \item Otherwise, for every $\varepsilon \in \realpos$ with $0 < \varepsilon < 1 - \mu$, $\rnext^i(\syncstate) + \varepsilon \syncbisim \rnext^i(\syncstate)$:
          Let $f$ be a bijection $f: C \rightarrow C + \varepsilon$ such that $f(c, v) = (c, v + \varepsilon)$.
          Clearly, for every  $(c, v) \in C$, $\lfloor v + \varepsilon \rfloor = \lfloor v \rfloor$ and $\lceil v + \varepsilon \rceil = \lceil v \rceil$, therefore $v \clockequiv v +  \varepsilon$.
          Furthermore, for every  $(c, v) \in C$: $\fract(v + \varepsilon) = \fract(v) + \varepsilon$.
          Therefore, for every $(s_1, v_1), (s_2, v_2) \in C$: $\fract(v_1 + \varepsilon) \leq \fract(v_2 + \varepsilon)$ iff $\fract(v_1) \leq \fract(v_2)$.
          Thus, $f$ satisfies the criteria of \autoref{def:sync-bisim}.
          Now, let $\varepsilon^* = t - \rincr^i(s)$.
          As $\rincr^{i+1}(s) = \rincr^i(s) + 1-\mu$, we obtain $0 < \varepsilon^* < 1 - \mu$.
          It follows that $s^* \syncbisim \rnext^i(\syncstate)$.
      \end{enumerate}
    \item Otherwise, $t > \max \{ \rincr^*(s) \}$.
      But then all clock values $v$ in $s$ must satisfy $v > K$.
      Note that by definition of $\rnext{}$, there is some $\syncstate' \in \rnext^*(s)$ such that all clocks $c$ in $s'$ satisfy $c > K$.
      It directly follows that $s' \syncbisim s^*$.
      \qedhere
  \end{enumerate}
\end{proofE}

Based on region increments, we can define the \emph{discrete quotient} \abslts of the synchronous product \synclts:
\begin{definition}[Discrete Quotient]
  Let $\synclts = (\syncstates, \syncstate_0, \syncstates^F, \batactions \cup \realpos, \synctransfull{})$ be a synchronous product.
  The \emph{discrete quotient} of $\synclts$ is a \ac{LTS} $\abslts = (\absstates, \absstate_0, \absstates_F, \batactions \cup \rationals, \abstransfull{})$ such that 
  \begin{itemize}
    \item $\absstate_0 = \syncstate_0$,
    \item $\absstate \abstrans{\delta} \absstate^*$ iff
        $\delta \in \rincr^*(\absstate)$ and
        $\absstate \synctranstime{\delta} \absstate^*$ for $\delta \in \rationals$,
    \item $\absstate^* \abstrans{a} \absstate'$ iff $\absstate^* \synctranssym{a} \absstate'$ for $a \in \batactions$,
    \item $\absstates \subseteq \syncstates$ is the smallest set such that $\absstate_0 \in \absstates$ and if $\absstate \abstrans{} \absstate'$, then $\absstate' \in \absstates$.
    \item $\absstate \in \absstates_F$ if $\absstate \in \syncstates^F$
      \qedhere
  \end{itemize}
\end{definition}

As before, we may omit the subscript $\Delta / \phi$ if $\Delta$ and $\phi$ are clear from the context.
We also write $\absstate \abstrans[t]{a} \absstate'$ if there is a state $\absstate^*$ such that $\absstate \abstrans{t} \absstate^* \abstrans{a} \absstate'$.
The discrete quotient \abslts is like \synclts, except that it only contains those time successors that correspond to some region increment.
Thus, in contrast to \synclts, it is finitely-branching.
The following example shows the discrete quotient of the running example:
\begin{example}[Discrete Quotient]
  \begin{figure}[tb]
    \centering
    \includestandalone[width=\textwidth]{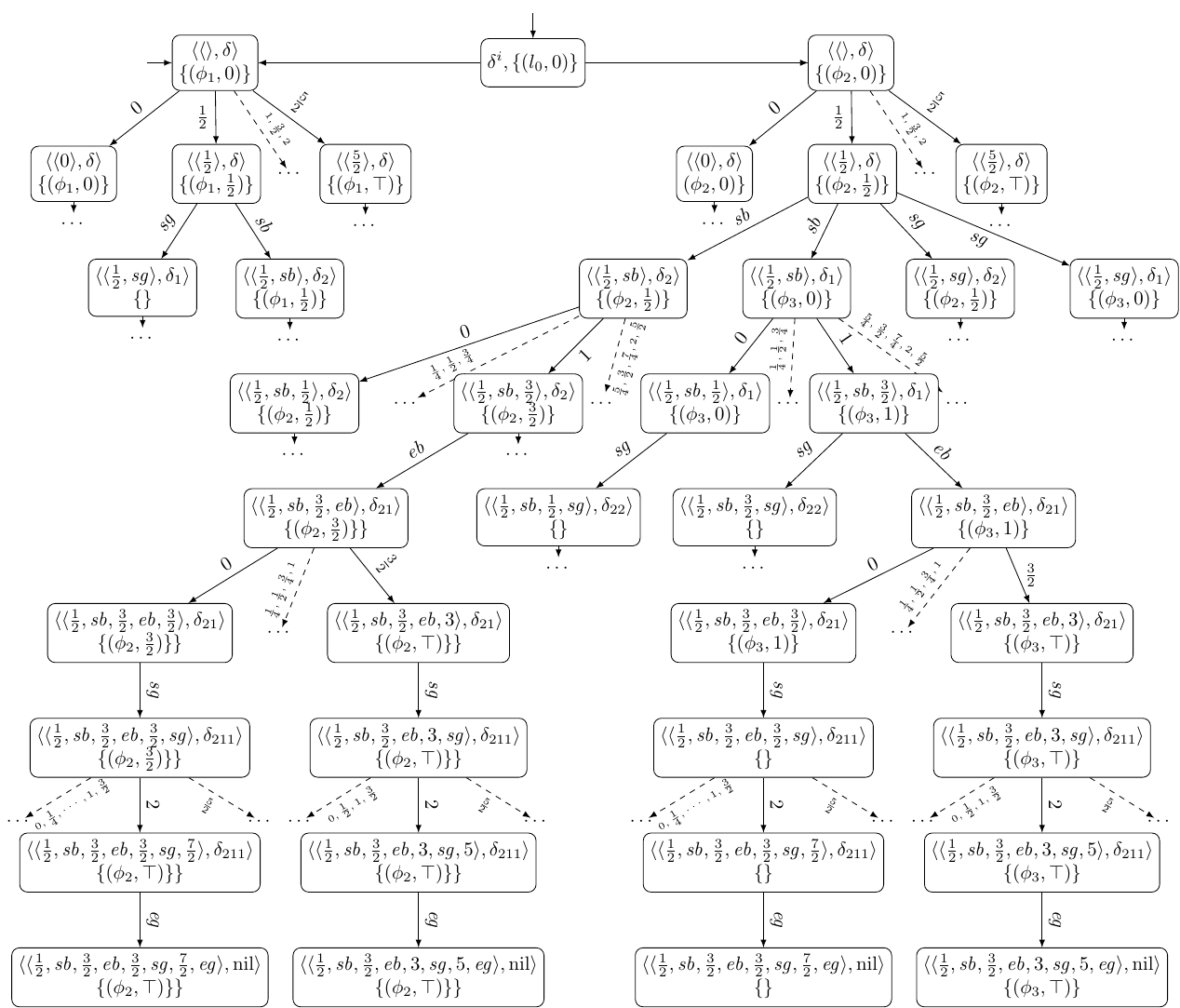}
    \caption[The discrete quotient.]{
      The discrete quotient of the synchronous product from \autoref{fig:sync-product}.
    }
    \label{fig:reg-sync-product}
  \end{figure}
  \autoref{fig:reg-sync-product} shows the discrete quotient \abslts of the synchronous product \synclts from \autoref{ex:sync-product}, using the same abbreviations as in \autoref{fig:sync-product}.
  We can see that \abslts has a similar structure as \synclts, but it only contains some of the time successors of \synclts.
  In fact, each state of \abslts only has finitely many time successors, which correspond to the region increments from \autoref{def:region-increment}.
\end{example}

With \autoref{lma:next-defines-successors}, it is easy to see that \abslts is contained in \synclts:
\begin{remark}\label{lma:absstate-reachable-in-synclts}
  As \abslts follows the transitions of \synclts and only restricts the time successors to a subset of the time successors in \synclts, every state reachable in \abslts is also reachable in \synclts.
\end{remark}

On the other hand, for each reachable state \syncstate of the synchronous product \synclts, the discrete quotient \abslts indeed contains a reachable state that is bisimilar to \syncstate:
\begin{lemmaE}\label{lma:syncstate-bisim-state-reachable-in-abslts}
  If a state $\syncstate$ is reachable from $\syncstate_0$ in \synclts, then there is a state $\absstate$ reachable from $\absstate_0$ in \abslts such that $\syncstate \syncbisim \absstate$.
  Furthermore, in each such state \absstate, all clock have rational values, i.e., $\clockvaluation_{\absstate}(c) \in \rationals$ for each clock $c$.
\end{lemmaE}
\begin{proofE}
  By induction on the number of transitions from $\syncstate_0$ to $\syncstate$.
  \\
  \textbf{Base case.}
  Assume $\syncstate = \syncstate_0$.
  Note that $\syncstate_0 \syncbisim \absstate_0$.
  Furthermore, $\absstate_0$ is trivially reachable from $\absstate_0$.
  Also, by definition of \synclts, for each clock $c$, $\clockvaluation_{\absstate_0}(c) = 0 \in \rationals$.
  \\
  \textbf{Induction step.}
  \\
  Assume $\syncstate_1$ is reachable in \synclts.
  By induction, there is a $\absstate_1$ such that $\absstate_1 \syncbisim \syncstate_1$ and $\absstate_1$ is reachable in \abslts.
  We distinguish time and action steps:
  \begin{description}
    \item[Time step:]
      Assume there is a transition $\syncstate_1 \synctranstime{t} \syncstate_2$.
      As $\absstate_1$ is reachable in \abslts, by \autoref{lma:absstate-reachable-in-synclts}, it is also reachable in \synclts.
      As $\absstate_1 \syncbisim \syncstate_1$ and because $\operatorname{\syncbisim}$ is a time-abstract bisimulation by \autoref{thm:time-abstract-bisimulation}, it follows from $\syncstate_1 \synctranstime{t} \syncstate_2$ that there is a $\absstate_2 \syncbisim \syncstate_2$ such that $\absstate_1 \synctranstime{t'} \absstate_2$ for some $t'$.
      By \autoref{lma:time-abstract-bisimulation}, there is a $\absstate_2' = \rnext^i(\absstate_1)$ such that $\absstate_2 \syncbisim \absstate_2'$ for some $i$.
      As $\absstate_2 \syncbisim \syncstate_2$ and $\absstate_2' \syncbisim \absstate_2$, it follows that $\absstate_2' \syncbisim \syncstate_2$.
      Now, let $t_i = \rincr^i(\absstate_1)$.
      By \autoref{lma:next-defines-successors}, $\absstate_1 \synctranstime{t_i} \absstate_2'$ and therefore, by definition of \abslts, $\absstate_1 \abstrans{t_i} \absstate_2'$.
      Finally, for each $c$, $\clockvaluation_{\absstate_1}(c) \in \rationals$ by induction, $t_i \in \rationals$ by \autoref{def:region-increment}, and therefore, $\clockvaluation_{\absstate_2}(c) \in \rationals$.
    \item[Action step:]
      Assume there is a transition $\syncstate_1 \synctranssym{a} \syncstate_2$.
      As $\absstate_1$ is reachable in \abslts, by \autoref{lma:absstate-reachable-in-synclts}, it is also reachable in \synclts.
      As $\absstate_1 \syncbisim \syncstate_1$ and because $\operatorname{\syncbisim}$ is a time-abstract bisimulation by \autoref{thm:time-abstract-bisimulation}, it follows with $\syncstate_1 \synctranssym{a} \syncstate_2$ that there is a $\absstate_2$ such that $\absstate_1 \synctranssym{a} \absstate_2$.
      By definition of \abslts, $\absstate_1 \abstrans{a} \absstate_2$.
      Finally, for each $c$, $\clockvaluation_{\absstate_1}(c) \in \rationals$ by induction and therefore, $\clockvaluation_{\absstate_2}(c) = 0 \in \rationals$ if the clock is reset and  $\clockvaluation_{\absstate_2}(c) = \clockvaluation_{\absstate_1}(c) \in \rationals$ otherwise.
      \qedhere
  \end{description}
\end{proofE}

Therefore, it is sufficient to consider \abslts for the synthesis problem: As we will see later, for any accepting path in \synclts, there is a path of bisimilar states in \abslts.
Furthermore, as each state in \abslts only contains rational clock configurations, we may use regression as defined in \autoref{def:regression} to determine the satisfied fluents and therefore the \ac{ATA} successor of each state.

However, there is a remaining issue: As we can see in \autoref{fig:det-reg-sync-product}, the \ac{LTS} is nondeterministic because a node may have multiple successors for the same action.
This is because the \ac{ATA} may be nondeterministic, as a location formula may have multiple distinct minimal models.
Our resulting controller should avoid bad states for all of those paths.
In order to solve this issue, we determinize the \ac{LTS} \abslts in the next section.

\section{Determinization} \label{sec:determinization}

As we have seen in the previous section, the discrete quotient \abslts as well as the synchronous product \synclts are nondeterministic.
For one, this is because an \ac{ATA} configuration may have multiple symbol successors for the same input symbols.
Additionally, the program may also be a source of nondeterminism, e.g., if the two subprograms of nondeterministic branching $\delta_1 \vert \delta_2$ start with the same action.
In order to determine a controller, we need to determinize the \ac{LTS} \abslts.
However, we cannot directly apply a power set construction on \abslts:
Different paths in \abslts with the same input may contain states with different clock valuations because a clock may be reset in one path while it is not reset in the other.
\autoref{fig:reg-sync-product} shows an example: If we follow the paths with input $\la \frac{1}{2}, \sboot, 1, \eboot \ra$, we end up in two states, where the first state contains the \ac{ATA} configuration $\{ (\phi_2, \frac{1}{2}) \}$ and the second state contains the \ac{ATA} configuration $\{ (\phi_3, 1) \}$ with different clock values than the first state.
As we could see in \autoref{ex:time-successors}, the region increment depends on the clock values.
Thus, the two paths may be incompatible in the sense that the two states may have different time successors.
For this reason, we cannot directly determinize \abslts but need to define the deterministic discrete quotient based on the synchronous product \synclts instead:
\begin{definition}[Deterministic Discrete Quotient]\label{def:detdiscquotient}
  Let $\synclts = (\syncstates, \syncstate_0, \syncstates^F, \batactions \cup \rationals, \synctrans{})$ be a synchronous product.
  The deterministic discrete quotient $\detdisq = (\detstates, \detstate_0, \detstates^F, \batactions \cup \rationals, \dettransfull{})$ of \synclts is defined as follows:
  \begin{itemize}
    \item $\detstate_0 = \{ \syncstate_0 \}$,
    \item $\detstate \detregtrans[t]{a} \detstate'$ iff $t \in \rincr^*(\detstate)$, $\detstate' = \{ \syncstate' \mid \exists \syncstate \in \detstate: \syncstate \synctrans[t]{a} \syncstate' \}$ and $\detstate' \neq \emptyset$,
    \item $\detstate \in \detstates^F$ if there is a $\syncstate \in \detstate$ such that $\syncstate \in \syncstates^F$,
    \item $\detstates \subseteq \powerset{\syncstates}$ is the smallest set such that $\detstate_0 \in \detstates$ and if $\detstate \dettrans{} \detstate'$, then $\detstate' \in \detstates$.
      \qedhere
  \end{itemize}
\end{definition}
As usual, we may omit the subscript $\Delta / \phi$ if $\Delta$ and $\phi$ are clear from the context.

In order to show the similarity of \abslts and \detlts, we first need the following observation:
\begin{theoremEnd}{remark} \label{rmk:time-successors-subset}
  Let $\clockvaluation_1, \clockvaluation_2$ be two clock valuations such that $\clockvaluation_1 \subseteq \clockvaluation_2$.
  Let $\clockvaluation_1' = \rnext^i(\clockvaluation_1)$.
  Then there is a $\clockvaluation_2' = \rnext^*(\clockvaluation_2)$ and $\clockvaluation_2^* \subseteq \clockvaluation_2'$ such that $\clockvaluation_1' \syncbisim \clockvaluation_2^*$.
\end{theoremEnd}

The two \acp{LTS} \abslts and \detlts only contain bisimilar paths:
\begin{lemmaE} \label{lma:det-path-existence}
  Let \synclts be a synchronous product, $\abslts$ the corresponding discrete quotient and $\detlts$ the corresponding deterministic discrete quotient.
  \begin{enumerate}
    \item
      Let $\absstate_0 \abstrans[t_1]{a_1} \absstate_1 \abstrans[t_2]{a_2} \ldots \abstrans[t_n]{a_n} \absstate_n$ be a path in $\abslts$.
      Then there is a path $\detstate_0 \dettrans[t_1']{a_1} \detstate_1 \dettrans[t_2']{a_2} \ldots \dettrans[t_n']{a_n} \detstate_n$ in \detlts such that for each $i$, there is a $\syncstate_i \in \detstate_i$ with $\syncstate_i \syncbisim \absstate_i$.
    \item Let $\detstate_0 \dettrans[t_1']{a_1} \detstate_1 \dettrans[t_2']{a_2} \ldots \dettrans[t_n']{a_n} \detstate_n$ be a path in \detlts.
      Then there exists a path $\absstate_0 \abstrans[t_1]{a_1} \absstate_1 \abstrans[t_2]{a_2} \ldots \abstrans[t_n]{a_n} \absstate_n$ in \abslts such that for each $i$, there is a $\syncstate_i \in \detstate_i$ with $\syncstate_i \syncbisim \absstate_i$.
  \end{enumerate}
\end{lemmaE}
\begin{proofE}
  ~
  \begin{enumerate}
    \item
      By induction on the length $n$.
      \\
      \textbf{Base case.}
      Assume $n = 0$.
      The claim follows with $\detstate_0 = \{ \syncstate_0 \}$ and $\absstate_0 = \syncstate_0$.
      \\
      \textbf{Induction step.}
      Assume $\absstate_0 \abstrans[t_1]{a_1} \ldots \abstrans[t_{n+1}]{a_{n+1}} \absstate_{n+1}$
      By induction, there is a path $\detstate_0 \dettrans[t_1']{a_1} \ldots \dettrans[t_n']{a_n} \detstate_n$ such that $\syncstate_n \in \detstate_n$ and $\syncstate_n \syncbisim \absstate_n$.
      From $\absstate_{n} \abstrans[t_{n+1}]{a_{n+1}} \absstate_{n+1}$, it follows that $\absstate_{n} \synctranstime{t_{n+1}} \absstate^* \synctranssym{a_{n+1}} \absstate_{n+1}$ is a pair of transitions in \synclts.
      By \autoref{rmk:time-successors-subset}, there is a $\detstate^* = \rnext^*(\detstate)$ such that $\absstate^* \syncbisim \syncstate^*$ for some $\syncstate^* \in \detstate^*$.
      Furthermore, by definition of \detlts, there is a $\detstate_{n+1}$ such that $\syncstate^* \synctranssym{a_{n+1}} \syncstate_{n+1}$ for some $\syncstate_{n+1} \in \detstate_{n+1}$ and such that $\syncstate_{n+1} \syncbisim \absstate_{n+1}$.
      Therefore, $\detstate_n \dettrans[t_{n+1}]{a_{n+1}} \detstate_{n+1}$ with $\syncstate_{n+1} \in \detstate_{n+1}$ and $\syncstate_{n+1} \syncbisim \absstate_{n+1}$.
    \item
      By induction on the length $n$.
      \\
      \textbf{Base case.}
      Assume $n = 0$.
      The claim follows with $\detstate_0 = \{ \syncstate_0 \}$ and $\absstate_0 = \syncstate_0$.
      \\
      \textbf{Induction step.}
      Assume $\detstate_0 \dettrans[t_1']{a_1} \ldots \dettrans[t_n']{a_{n+1}} \detstate_{n+1}$.
      By induction, $\absstate_0 \abstrans[t_1]{a_1} \ldots \abstrans[t_{n}]{a_{n}} \absstate_{n}$ such that $\absstate_{n} \syncbisim \syncstate_n$ for some $\syncstate_n \in \detstate_n$.
      From $\detstate_n \dettrans[t_{n+1}']{a_{n+1}'} \detstate_{n+1}$, it follows that $\syncstate_n \synctrans[t_{n+1}']{a_{n+1}}\syncstate_{n+1}$ for some $\syncstate_{n+1}$.
      As $\absstate_n \syncbisim \syncstate_n$ and because $\operatorname{\syncbisim}$ is a time-abstract bisimulation, it follows with \autoref{def:sync-bisim} that there is some $t_{n+1}^*$ and $\absstate_{n+1}$ such that $\absstate_n \synctrans[t_{n+1}]{a_{n+1}} \absstate_{n+1}$.
      Hence, with \autoref{lma:syncstate-bisim-state-reachable-in-abslts}, $\absstate_n \abstrans[t_{n+1}]{a_{n+1}} \absstate_{n+1}$. \todo{Needs \autoref{lma:syncstate-bisim-state-reachable-in-abslts} to be about paths}
      \qedhere
  \end{enumerate}
\end{proofE}

Therefore, even though \detlts is not directly constructed from \abslts, it can still be considered to be the deterministic version of \abslts.
\begin{example}[Deterministic Discrete Quotient]
  \begin{figure}[tb]
    \centering
    \includestandalone[width=\textwidth]{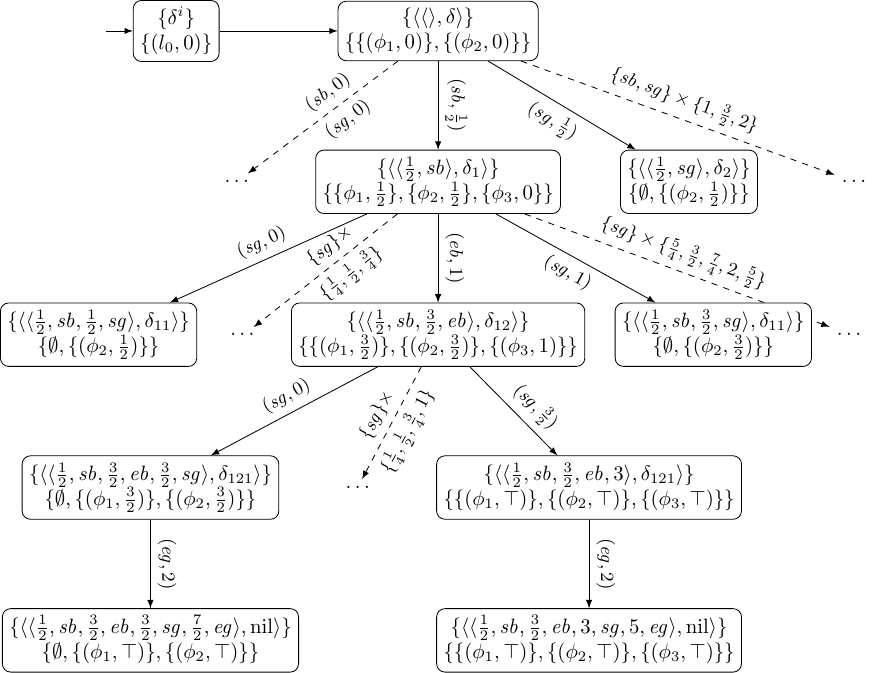}
    \caption[The deterministic discrete quotient.]{
      The deterministic discrete quotient of the program and \ac{ATA} from \autoref{ex:control-problem}.
      To improve readability, each node has been split in two parts, the program configurations and the \ac{ATA} configurations.
      The actual state is the cross product of the two parts.
    }
    \label{fig:det-reg-sync-product}
  \end{figure}
  \autoref{fig:det-reg-sync-product} shows the deterministic discrete quotient of the synchronous product from \autoref{ex:sync-product}.
  Each state is a set of states of the synchronous product.
  Time and action successor transitions are joined such that each state has a unique successor for each possible timed action $(a, t)$.
\end{example}

%

We conclude by showing the path equivalence of the program transition semantics and the three \acp{LTS} introduced above:
\begin{theoremE}\label{lma:path-trace-equivalence}
  Let $\Delta = (\delta, \bat)$ be a program over a finite-domain \ac{BAT} \bat, $w \in \worlds$ a world with $w \models \bat$, and $\phi$ a fluent trace formula that does not mention any function symbols.
  The following statements are equivalent:
  \begin{enumerate}
    \item \label{lma:path-trace-equivalence:trace}
      There is a finite trace $z \in \|\delta\|_w$ satisfying $\phi$. 
    \item \label{lma:path-trace-equivalence:sync}
      There is an accepting run $\syncstate_0 \synctrans{}^* \syncstate_f \in \accfinruns(\synclts)$ in \synclts.
    \item \label{lma:path-trace-equivalence:abs}
      There is an accepting run $\absstate_0 \abstrans{}^* \absstate_f  \in \accfinruns(\abslts)$ in \abslts.
    \item \label{lma:path-trace-equivalence:det}
      There is an accepting run $\detstate_0 \detregtrans{}^* \detstate_f \in \accfinruns(\detlts)$ in \detdisq.
  \end{enumerate}
\end{theoremE}
\begin{proofE}
  \begin{description}
    \item[$(\ref{lma:path-trace-equivalence:trace}) \Leftrightarrow (\ref{lma:path-trace-equivalence:sync})$:]
      Note that a fluent trace formula not mentioning any function symbols is an \ac{MTL} formula over alphabet \primformulas.
      Let $\rho = \{ F(\vec{n}) \in \primformulas \mid w[F(\vec{n}), z^{(i)}] = 1 \}$.
      By \autoref{lma:mtl-equivalence}, $w, z \models \phi$ iff $\rho \models_{\mathrm{MTL}} \phi$.
      By \autoref{thm:mtl-ata}, $\rho \models_{\mathrm{MTL}} \phi$ iff $\ata$ accepts $\phi$.
      By \autoref{def:ata-lts}, \ata accepts $\phi$ iff there is a finite run $\ataconf_0 \atatranstime{d_1} \ataconf^*_1 \atatranssym{s_1} \ataconf'_1 \atatranstime{d_2} \ldots \atatranssym{s_n} \ataconf_n$ such that $\ataconf_n$ is accepting.
      By definition of \synclts, such a run on \atalts exists iff such a run exists on \synclts.
    \item[$(\ref{lma:path-trace-equivalence:sync}) \Rightarrow (\ref{lma:path-trace-equivalence:abs})$:]
      Let $\syncstate_f = (\la z_f, \rho_f \ra, \ataconf_f)$ be an accepting state of \synclts.
      By \autoref{lma:syncstate-bisim-state-reachable-in-abslts}, it follows that there is a $\absstate_f = (\la z_f', \rho_f' \ra, \ataconf_f') \in \absstates$ that is reachable in \abslts and such that $\absstate_f \syncbisim \syncstate_f$.
      By \autoref{lma:sync-states-satisfy-same-static-formulas}, $\la z_f', \rho_f' \ra \in \final$.
      Furthermore, as $\syncstate_f$ and therefore $\ataconf_f$ is accepting, every location in the \ac{ATA} configuration $\ataconf_f$ must be accepting.
      By definition of $\operatorname{\syncbisim}$, there is a bijection $f: \ataconf_f \rightarrow \ataconf_f'$ and such that $f(s, u) = f(t, v)$ implies that $s = t$.
      Therefore, every location in $\ataconf_f'$ must be accepting.
    \item[$(\ref{lma:path-trace-equivalence:abs}) \Rightarrow (\ref{lma:path-trace-equivalence:sync})$:]
      Follows directly from \autoref{lma:absstate-reachable-in-synclts}.
    \item[$(\ref{lma:path-trace-equivalence:abs}) \Rightarrow (\ref{lma:path-trace-equivalence:det})$:]
      Let $\absstate_f$ be an accepting state in \abslts.
      By \autoref{lma:det-path-existence}, there is a path in \detlts ending in a state $\detstate_f$ such that $\absstate_f \syncbisim \syncstate_f$ for some $\syncstate_f \in \detstate_f$.
      Furthermore, as $\absstate_f$ is accepting, it follows that $\syncstate_f$ is accepting and hence, by definition of \detlts, $\detstate_f$ is accepting.
    \item[$(\ref{lma:path-trace-equivalence:det}) \Rightarrow (\ref{lma:path-trace-equivalence:sync})$:]
      Follows immediately by construction of \detdisq.
      \qedhere
  \end{description}
\end{proofE}

Therefore, for verifying \iac{MTL} property on a program $\delta$, it suffices to consider the \ac{LTS} \detlts.
While the program is generally infinitely branching because it may have a time successor for each $r \in \realpos$, we reduced the problem to checking a finitely-branching \ac{LTS}.
However, there is a remaining problem: While the \ac{LTS} is finitely branching, it may still contain infinite paths.
As we are only interested in finite traces, we may simply ignore paths that do not end in a final configuration.
Furthermore, as the domain is restricted to be finite, every infinite path will eventually reach a state with the same remaining program and the same satisfied fluents as a previous state on the path.
Thus, if we just consider program configurations, we may stop whenever we reach a configuration that satisfies the same fluents and has the same remaining program.
However, this approach does not work for \ac{ATA} configurations: As the number of \ac{ATA} states may always increase and may not have an upper bound, it is possible that we never see a repeating \ac{ATA} configuration.
To solve this problem, we will use \emph{well-structured transition systems} that allow us to define a stop criterion even in such transition systems with infinite paths.

\section{Well-Structured Transition Systems} \label{sec:wsts}

Generally, when analyzing a transition system with infinite paths, it is necessary to evaluate all states of the path and therefore infinitely many states, which is infeasible.
However, under certain conditions, it is sufficient to stop following a path: As an example, consider the \emph{reachability problem}, where the task is to check whether some subset $V \subseteq W$ of the system's states $W$ is reachable.
Now, assume that we can define an ordering $(W, \leq)$ of the states $W$ such that $w_1 \leq w_2$ implies that every path starting in $w_2$ that reaches $V$ is also a valid path from $w_1$ that also reaches $V$.
Assuming we have already visited $w_1$, we can stop when we reach $w_2$: If $V$ is reachable from $w_2$, then it is also reachable from $w_1$, so it is sufficient to check $w_1$.
If the ordering $(W, \leq$) is a \acfi{wqo}, where every infinite sequence $w_1, w_2, \ldots$ contains a pair with $w_i \leq w_j$ and $i < j$, then every infinite path in the transition system must eventually reach a state where we can stop expanding.
This is the (simplified) intuition for \emph{well-structured transition systems}~\parencite{finkelReductionCoveringInfinite1990,abdullaGeneralDecidabilityTheorems1996,finkelWellstructuredTransitionSystems2001}:

\begin{definition}[Well-Structured Transition System \parencite{ouaknineDecidabilityComplexityMetric2007}]
  A \acfi{WSTS} is a triple $\mathcal{W} = (W, \preccurlyeq, \rightarrow)$, where $(W, \rightarrow)$ is a finitely-branching transition system equipped with a \ac{wqo} $\preccurlyeq$ such that:
  \begin{enumerate}
    \item $\preccurlyeq$ is a decidable relation,
    \item $\suc(w) \eqdef \{ w' \mid w \rightarrow w' \}$ is computable for each $w \in W$,
    \item $\preccurlyeq$ is \emph{downward compatible}: if $w, v \in W$ with $w \preccurlyeq v$, then for any transition $v \rightarrow v'$, there exists a matching sequence of transitions $w \rightarrow^* w'$ with  $w' \preccurlyeq v'$.
      \qedhere
  \end{enumerate}
\end{definition}

In the following, we construct a suitable \ac{wqo} for the \ac{LTS} \detlts that allows us to only consider a finite subset of the states of \detlts.
In particular, it will allow us to apply the following result:
\begin{theorem}[\parencite{finkelWellstructuredTransitionSystems2001,ouaknineDecidabilityComplexityMetric2007}] \label{thm:wsts-decidability}
  Let \wsts be a \ac{WSTS}.
  Let $V \subseteq W$ be a downward-closed decidable subset of $W$.
  Then, given a state $u \in W$, it is decidable whether there is a sequence of transitions starting at $u$ and ending in $V$.
\end{theorem}

Before we can define the \ac{wqo} on the states \detstates of \detlts, we introduce some basic notions about \acp{wqo}.
We start with the \emph{monotone domination order}, which allows to compare finite sequences of symbols from some set $S$, based on a \ac{qo} on $S$.
\begin{definition}[Monotone Domination Order]\label{def:mon-dom-order}
  Let $(S, \leq)$ be a \qo.
  The \emph{monotone domination order} is the \qo $(S^*, \leq^*)$ over the set $S^*$ of finite words over $S$ such that $x_1, \ldots, x_m \leq^* y_1, \ldots, y_n$ iff there is a strictly monotone injection $h: \{ 1, \ldots, m \} \rightarrow \{ 1, \ldots, n \}$ such that $x_i \leq y_{h(i)}$ for all $1 \leq i \leq m$.
\end{definition}
The monotone domination order will allow us to define an ordering on \ac{ATA} configurations by means of an \emph{abstraction function} that encodes a set of clock valuations as a sequence of symbols.
But before we introduce the abstraction function, we provide an example for a monotone domination order on the alphabet $\{ a, b, \ldots, z \}$:

\begin{example}[Monotone Domination Order]
  Let $L = \{a, b, \ldots, z \}$ and let $(L, =)$ be the \ac{qo} where $l_1 = l_2$ iff $l_1$ and $l_2$ are identical.
  The induced monotone domination order $(L^*, =^*)$ compares finite sequences of letters, where:
  \begin{enumerate}
    \item $\la a, c, d \ra =^* \la a, b, c, d \ra$ with the injection $h(1) = 1, h(2) = 3, h(3) = 4$, which maps the letters as follows:
      \begin{center}
        \begin{tikzpicture}[->,>=latex]
          \node (a1) {$a$};
          \node[right=of a1] (c1) {$c$};
          \node[right=of c1] (d1) {$d$};
          \node[below=of a1] (a2) {$a$};
          \node[right=of a2] (b2) {$b$};
          \node[right=of b2] (c2) {$c$};
          \node[right=of c2] (d2) {$d$};

          \draw (a1) -- (a2);
          \draw (c1) -- (c2);
          \draw (d1) -- (d2);
        \end{tikzpicture}
      \end{center}
    \item $\la e \ra =^* \la e, e \ra$ with the injection $h(1) = 1$ (alternatively, $h(1) = 2$), which maps the letters as follows:
      \begin{center}
        \begin{tikzpicture}[->,>=latex]
          \node (e1) {$e$};
          \node[below=of e1] (e2) {$e$};
          \node[right=of e2] (e21) {$e$};

          \draw (e1) -- (e2);
        \end{tikzpicture}
      \end{center}
    \item $\la e, e \ra \neq^* \la e \ra$ because there is no injection from $\{ 1, 2 \}$ to $\{ 1 \}$.
      Note that this means that $(L^*, =^*)$ is not symmetric even though $(L, =)$ is symmetric.
    \item $\la a, b \ra \neq^* \la b, a \ra$.
      The first letter $a$ of the first sequence must be mapped to $a$ in the second sequence.
      However, after doing so, it is impossible to map $b$ to $b$ with a strictly monotone injection.

  \end{enumerate}
  As a second example, consider the canonical \ac{qo} $(L, \leq)$  of $L$, where $l_1 \leq l_2$ if $l_1$ occurs before $l_2$ in the alphabet.
  Now, we obtain the following:
  \begin{enumerate}
    \item $\la a, b \ra \leq^* \la c, d \ra$ with the monotone injection $h(1) = 1, h(2) = 2$, which maps the letters as follows:
      \begin{center}
        \begin{tikzpicture}[->,>=latex]
          \node (a1) {$a$};
          \node[right=of a1] (b1) {$b$};
          \node[below=of a1] (c2) {$c$};
          \node[right=of c2] (d2) {$d$};

          \draw (a1) -- (c2);
          \draw (b1) -- (d2);
        \end{tikzpicture}
      \end{center}
    \item $\la a, c, a \ra \leq^* \la d, a, d, a \ra$ with the injection $h(1) = 1, h(2) = 3, h(3) = 4$, which maps the letters as follows:
      \begin{center}
        \begin{tikzpicture}[->,>=latex]
          \node (a1) {$a$};
          \node[right=of a1] (c1) {$c$};
          \node[right=of c1] (a11) {$a$};
          \node[below=of a1] (d2) {$d$};
          \node[right=of d2] (a2) {$a$};
          \node[right=of a2] (d21) {$d$};
          \node[right=of d21] (a21) {$a$};

          \draw (a1) -- (d2);
          \draw (c1) -- (d21);
          \draw (a11) -- (a21);
        \end{tikzpicture}
      \end{center}
      \qedhere
  \end{enumerate}
\end{example}

Based on monotone domination orders, we can define an abstraction function to obtain a canonical representation of all the clock values of a state \syncstate of \synclts:
\begin{definition}[Abstraction Function]
  Let $\Lambda = \powerset{(\clockset \cup L) \times \regions}$ be an alphabet that consists of sets of name-index pairs, where each name is either a clock name of the program or a location name of the \ac{ATA}, and each index is a region index of \regions.
  Let \syncstate be a state of \synclts.
  We partition $\syncconf = \syncclocks(\syncstate) \in \syncconfs$ into a sequence of subsets $\syncconf_1, \ldots, \syncconf_n$ such that for every $1 \leq i \leq j \leq n$, for every pair $(l_i, c_i) \in \syncconf_i$ and every pair $(l_j, c_j) \in \syncconf_j$, the following holds: $i \leq j$ iff $\fract(c_i) \leq \fract(c_j)$.
  For each $\syncconf_i$, let $\absword(\syncconf_i) = \{ (l, \reg(c)) \mid (l, c) \in \syncconf_i \} \in \Lambda$.
  Then, the \emph{abstraction function} $H: \syncconfs \rightarrow \Lambda^*$ with $H(\syncconf) \eqdef (\absword(\syncconf_1), \ldots, \absword(\syncconf_n))$ defines a canonical representation $H(\syncconf) \in \Lambda^*$ of $\syncconf$.
\end{definition}

We illustrate the abstraction function with some examples:
\begin{example}[Abstraction Function]\label{def:abstraction-function}
  Clocks with the same fractional part are assigned to the same partition, independent of the integer part:
  \begin{align*}
    C &= \{ (c_1, 0.5), (c_2, 1.5) \}
    \\
    H(C) &= ( \{ (c_1, 1), (c_2, 3) \})
  \end{align*}
  The partitions are ordered by the fractional parts of the clock valuation, independent of the integer part:
  \begin{align*}
    C &= \{ (c_1, 0.5), (c_3, 0.6), (c_2, 1.5)  \}
    \\
    H(C) &= ( \{ (c_1, 1), (c_2, 3) \}, \{ (c_3, 1) \})
  \end{align*}
  As an \ac{ATA} configuration may contain the same clock with multiple values, it may also appear in $H(C)$ in multiple places:
  \begin{align*}
    C &= \{ (c_1, 0.0), (c_2, 0.5), (c_3, 0.6), (x, 0.5), (x, 2.0) \}
    \\
    H(C) &= ( \{ (c_1, 0), (x, 4) \}, \{ (c_2, 1), (x, 1) \}, \{ (c_3, 1) \} )
  \end{align*}
  Assume $K = 2$.
  Clocks with a valuation greater than $K$ are assigned to the first partition because $\fract(\top) = 0$ by definition:
  \begin{align*}
    C &= \{ (c_1, 0.0), (c_2, 0.5), (c_3, 2.6), (x, 0.5), (x, 2.0) \}
    \\
    H(C) &= ( \{ (c_1, 0), (x, 4), (c_3, \top) \}, \{ (c_2, 1), (x, 1) \} )
    \qedhere
  \end{align*}
\end{example}

The abstraction function $H$ induces an order $(\syncconfs, \leq_H)$ on the clock valuations \syncconfs of a state $\syncstate \in \syncstates$, where $\syncconf \leq_H \syncconf'$ iff $H(\syncconf) \subseteq^* H(\syncconf')$ and where $(\Lambda^*, \subseteq^*)$ is the monotone domination ordering induced by $(\Lambda, \subseteq)$ according to \autoref{def:mon-dom-order}:

\begin{example}[Ordering on \syncconfs]
  Consider the following clock valuations:
  \begin{align*}
    C_1 &\eqdef \{ (c_1, 0.3), (c_2, 0.5), (c_3, 1.3) \}
    \\
    C_2 &\eqdef \{ (c_1, 0.2), (c_2, 0.6) \}
    \\
    C_3 &\eqdef \{ (c_1, 0.2), (c_2, 0.2) \}
  \end{align*}
  The abstraction function $H$ defines the following abstracted configurations:
  \begin{align*}
    H(C_1) &= ( \{ (c_1, 1), (c_3, 3) \}, \{ (c_2, 1) \})
    \\
    H(C_2) &= ( \{ (c_1, 1) \}, \{ (c_2, 1) \})
    \\
    H(C_3) &= ( \{ (c_1, 1), (c_2, 1) \})
  \end{align*}
  We can compare $C_1$, $C_2$, and $C_3$ with the ordering $(\syncconfs, \leq_H)$ induced by the abstraction function $H$:
  \begin{itemize}
    \item $C_2 \leq_H C_1$ because $\{ (c_1, 1) \} \subseteq \{ (c_1, 1), (c_3, 3) \}$ and $\{ (c_2, 1) \} \subseteq \{ (c_2, 1) \}$ and therefore, the monotone injection $h$ with $h(1) = 1$ and $h(2) = 2$ satisfies the criteria of \autoref{def:mon-dom-order}.
    \item $C_2 \not\leq_H C_3$.
      Note that $\{ (c_1, 1) \} \subseteq \{ (c_1, 1), (c_2, 1) \}$ and $\{ (c_2, 1) \} \subseteq \{ (c_1, 1), (c_2, 1) \}$.
      However, the resulting injection $h$ with $h(1) = 1$ and $h(2) = 1$ is not strictly monotonically increasing and no other injection satisfying all criteria exists.
      Therefore, even if two configurations contain the same clock region values, they are not comparable if the fractional parts of both configurations are not in the same order.
      In $C_1$, the fractional part of $c_2$ is larger than the fractional part of $c_1$, while they are both the same in $C_2$.
      \qedhere
  \end{itemize}
\end{example}

To define the order on the states \detstates of \detlts, we need two more notions.
First, the states \syncstates of \synclts are tuples of program configurations and \ac{ATA} configurations.
To order those, we will need the \emph{Cartesian product of orders}:
\begin{definition}[Cartesian Product of Orders]
  Let $(A, \leq_A)$ and $(B, \leq_B)$ be \acp{qo}.
  The Cartesian product $(A \times B, \leq_{A \times B})$ of $(A, \leq_A)$ and $(B, \leq_B)$ is a \qo such that
  $(a, b) \leq_{A \times B} (a', b')$ iff $a \leq_A a'$ and $b \leq_B b'$.
\end{definition}
Second, the states \detstates of the deterministic version \detlts of \abslts are sets of states of \abslts.
These may be ordered with the power set order:
\begin{definition}[Powerset Order \parencite{marconeFineAnalysisQuasiOrderings2001,abdullaWellBetterQuasiOrdered2010}]\label{def:powerset-order}
  Let $(S, \leq)$ be a \qo.
  The power set ordering $(\powerset{S}, \lpowleq)$ induced by $(S, \leq)$ is a \qo  such that for every $X, Y \in \powerset{S}$:\footnote{
    Following the notation by \textcite{marconeFineAnalysisQuasiOrderings2001}, this would be written as $(\powerset{S}, \lpowleq^\forall_\exists)$ to distinguish it from the more common ordering $(\powerset{S}, \lpowleq^\exists_\forall)$.
    We omit the sub- and superscript as we are only interested in the former ordering.
  }
  \[
    X \lpowleq Y \text{ iff } \forall y \in Y \exists x \in X: x \leq y
    \qedhere
  \]
\end{definition}
We can now define an ordering on the states of \synclts:
\begin{definition}[Ordering \syncleq on \synclts]
  The ordering $(\syncstates, \syncleq)$ between states of \synclts is defined as follows:
  Let $\syncstate = (\la z, \rho \ra, \ataconf)$ and $\syncstate' = (\la z', \rho' \ra, \ataconf')$.
  Then $\syncstate \syncleq \syncstate'$ iff
  \begin{enumerate}
    \item $w[F(\vec{n}), z] = w[F(\vec{n}), z']$ for every $F(\vec{n}) \in \primformulas$,
    \item $\rho = \rho'$, and
    \item $H(\syncclocks(\syncstate)) \leq^* H(\syncclocks(\syncstate'))$.
      \qedhere
  \end{enumerate}
\end{definition}

The ordering compares two states $\syncstate$ and $\syncstate'$ by comparing
\begin{enumerate*}[label=(\arabic*)]
  \item the satisfied fluents,
  \item the remaining program,
  \item and the clock valuations using the monotone domination order from above.
\end{enumerate*}
If two states satisfy different fluents (i.e., the world states are not identical), then the states are incomparable.
Similarly, if the remaining programs differ, then the states are also incomparable.
If both the world state and the remaining program are the same, then the states are compared using the canonical representation of the clock valuations.

\begin{example}[Ordering $(\syncstates, \syncleq)$]
  Consider the following states:
  \begin{align*}
    \syncstate_1 &= (\la  z_1, \rho_1 \ra, \ataconf_1)
    \\
                 &= \left(\la \la (\sac{\drive(m_1, m_2)}, 0.9), (\eac{\drive(m_1, m_2)}, 2.8) \ra, \nil \ra, \{ (\phi_3, 1.8) \}\right)
    \\
    \syncstate_2 &= (\la z_2, \rho_2 \ra, \ataconf_2)
    \\
                 &= \left(\la \la (\sac{\drive(m_1, m_2)}, 1.2), (\eac{\drive(m_1, m_2)}, 2.95) \ra, \nil \ra, \{ \}\right)
  \end{align*}
  Both states consist of the same actions but at different time points and both have the empty program as remaining program.
  The first state $\syncstate_1$ has an \ac{ATA} configuration $\{ (\phi_3, 1.9) \}$, while the second state $\syncstate_2$ has an empty \ac{ATA} configuration.
  First, note that $\bat \models \square c(\drive(m_1, m_2)) = q_2$, i.e., $q_2$ is the clock that keeps track of the time since $\drive(m_1, m_2)$ has started and therefore is reset by the action $\sac{\drive(m_1, m_2)}$.
  Hence:
  \begin{align*}
    w[z_1, q_2] &= 1.9
    \\
    w[z_2, q_2] &= 1.75
  \end{align*}
  That is, the clock $q_2$ has the value $1.9$ after $z_1$ and the value $1.75$ after $z_2$.
  The other clocks are never reset, and so for every $q_i \in \{ q_1, q_3, q_4, q_5, q_6 \}$:
  \begin{align*}
    w[z_1, q_i] &= 2.8
    \\
    w[z_2, q_i] &= 2.95
  \end{align*}
  Assuming $K = 2$ as before, the abstracted configurations look as follows:
  \begin{align*}
    H(\syncclocks(\syncstate_1)) &= (\{ (q_1, \top), (q_3, \top), (q_4, \top), (q_5, \top), (q_6, \top) \}, \{ (q_2, 3), (\phi_3, 3) \},
    \\
    H(\syncclocks(\syncstate_2)) &= (\{ (q_1, \top), (q_3, \top), (q_4, \top), (q_5, \top), (q_6, \top) \}, \{ (q_2, 3) \})
  \end{align*}
  It follows that $\syncstate_2 \syncleq \syncstate_1$:
  \begin{enumerate}
    \item $w[\phi, z_1] = w[\phi, z_2]$ for every $\phi \in \primformulas$.
      Both states satisfy the same fluents, because they consist of the same actions, just at different time points.
    \item Both states have the empty program as remaining program: $\rho_1 = \rho_2 = \nil$.
    \item $H(\syncclocks(\syncstate_2)) \leq^* H(\syncclocks(\syncstate_1))$ because $\{ (q_2, 3) \} \subseteq \{ (q_2, 3), (\phi_3, 3) \}$ and so we can map $\{ (q_2, 3) \}$ to $\{ (q_2, 3), (\phi_3, 3) \}$:
      \begin{center}
        \begin{tikzpicture}[->,>=latex, node distance=1.0cm and 0.5cm]
          \node (qs1) {$\{ (q_1, \top), (q_3, \top), (q_4, \top), (q_5, \top), (q_6, \top) \}$};
          \node[right=of qs1] (q2phi1) {$\{ (q_2, 3), (\phi_3, 3) \}$};
          \node[above=of qs1] (qs2) {$\{ (q_1, \top), (q_3, \top), (q_4, \top), (q_5, \top), (q_6, \top) \}$};
          \node[above=of q2phi1] (q22) {$\{ (q_2, 3) \}$};

          \node[left=0cm of qs1] (leftpar1) {$\big($};
          \node[left=of leftpar1] {$H(\syncclocks(\syncstate_1))=$};
          \node[right=0cm of q2phi1] {$\big)$};

          \node[left=0cm of qs2] (leftpar2) {$\big($};
          \node[left=of leftpar2] {$H(\syncclocks(\syncstate_2))=$};
          \node[right=0cm of q22] {$\big)$};

          \draw (qs2) -- (qs1);
          \draw (q22) -- (q2phi1);
        \end{tikzpicture}
      \qedhere
      \end{center}
  \end{enumerate}
\end{example}

Finally, the ordering $(\syncstates, \syncleq)$ induces a power set order $(\detstates, \detleq$), following \autoref{def:powerset-order}.
We now want to show that $(\detstates, \detleq)$ is \iac{wqo}.
While a \ac{wqo} is sufficient for our purposes, it is often easier to show that an ordering is a \ac{bqo}~\cite{nash-williamsWellquasiorderingInfiniteTrees1965}.
As we are only interested in the fact that each \ac{bqo} is also a \ac{wqo} and because we may construct a \ac{bqo} as follows, we omit the definition of \acp{bqo} and instead summarize some known results about the composition of \acp{bqo}:
\begin{proposition}\label{prop:bqos}
  ~
  \begin{enumerate}
    \item \label{prop:bqos:bqo-is-wqo}
      Each \bqo is a \wqo \cite{abdullaBetterBetterWell2000}.
    \item \label{prop:bqos:equality}
      If $S$ is finite, then $(S, =)$ is a \bqo \cite{abdullaBetterBetterWell2000}.
    \item \label{prop:bqos:finite-powerset}
      If $S$ is finite, $(2^S, \subseteq)$ is a \bqo \cite{abdullaBetterBetterWell2000}.
    \item \label{prop:bqos:mondom}
      If $(S, \leq)$ is a \bqo, then $(S^*, \leq^*)$ is a \bqo \cite{abdullaBetterBetterWell2000}.
    \item \label{prop:bqos:cartesian}
      If $(A, \leq_A)$ and $(B, \leq_B)$ are \acp{bqo}, then $(A \times B, \leq_{A \times B})$ is a \bqo \parencite{nash-williamsWellquasiorderingInfiniteTrees1965}.
    \item \label{prop:bqos:powerset}
      If $(S, \leq)$ is a \bqo, then $(2^S, \sqsubseteq)$ is a \bqo \parencite{nash-williamsWellquasiorderingInfiniteTrees1965}.
    \item \label{prop:bqos:subset}
      If $(S, \leq)$ is a \bqo and $S' \subseteq S$, then $(S', \leq)$ is a \bqo \parencite{laverFraisseOrderType1971}.\footnote{
        This is a special case of the \emph{homomorphism property}~\parencite[][p.~93]{laverFraisseOrderType1971}.}
  \end{enumerate}
\end{proposition}

With this, we can show that the ordering $(\detstates, \detleq)$ is indeed a \bqo:
\begin{lemmaE}\label{lma:detleq-bqo}
  ~
  \begin{enumerate}
    \item \label{lma:detleq-wqo:mondom}
      The monotone domination ordering $(\Lambda^*, \preccurlyeq)$ induced by the \qo $(\Lambda, \subseteq)$ is a \bqo.
    \item The ordering $(\syncstates, \syncleq)$ is a \bqo.
    \item The ordering $(\detstates, \detleq)$ is a \bqo.
  \end{enumerate}
\end{lemmaE}
\begin{proofE}
  ~
  \begin{enumerate}
    \item $(S \cup L) \times \regions$ is finite, thus, by \subdefref{prop:bqos}{prop:bqos:finite-powerset}, $(\Lambda, \subseteq)$ is a \bqo.
      By \subdefref{prop:bqos}{prop:bqos:mondom}, $(\Lambda^*, \preccurlyeq)$ is a \bqo.
    \item As $\fluentset$ and $\sub(\delta)$ are finite sets, we directly obtain with \subdefref{prop:bqos}{prop:bqos:equality} that $(\fluentset, =)$ and $(\sub(\delta), =)$ are \acp{bqo}.
      By \autoref{lma:detleq-wqo:mondom}, $(\Lambda^* \preccurlyeq)$ is a \bqo.
      Finally, note that $(\syncstates, \syncleq)$ is the Cartesian product of the three \acp{bqo} above.
      By \subdefref{prop:bqos}{prop:bqos:cartesian}, $(\syncstates, \syncleq)$ is a \bqo.
    \item
      As $(\syncstates, \syncleq)$ is a \bqo, it follows by \subdefref{prop:bqos}{prop:bqos:powerset} that $(\powerset{\syncstates}, \lpowleq)$ is a \bqo.
      As $\detstates \subseteq \powerset{\syncstates}$, it follows with \subdefref{prop:bqos}{prop:bqos:subset} that $(\detstates, \lpowleq)$ is a \bqo.
      \qedhere
  \end{enumerate}
\end{proofE}

We have now defined a \wqo on the states of \detlts, which brings us a step towards showing that \detlts is indeed a \ac{WSTS}.
In addition to being a \wqo, a \ac{WSTS} also requires the ordering to be \emph{downward compatible}:

\pagebreak

\begin{lemmaE} \label{lma:downward-compat}
  ~
  \begin{enumerate}
    \item \label{lma:downward-compat:abs}
      The transition relation \abstrans{} of \abslts is downward-compatible with respect to $\absleq$, i.e., for $\absstate_1, \absstate_2 \in \absstates$ with $\absstate_1 \absleq \absstate_2$, $\absstate_2 \abstrans{} \absstate_2'$ implies that there is a $\absstate_1' \absleq \absstate_2'$ such that $\absstate_1 \abstrans{} \absstate_1'$.
    \item \label{lma:downward-compat:det}
      The transition relation \dettrans{} of \detlts is downward-compatible with respect to $\detleq$, i.e., for $\detstate_1, \detstate_2 \in \detstates$ with $\detstate_1 \detleq \detstate_2$, $\detstate_2 \dettrans{} \detstate_2'$ implies that there is a $\detstate_1' \detleq \detstate_2'$ such that $\detstate_1 \dettrans{} \detstate_1'$.
  \end{enumerate}
\end{lemmaE}
\begin{proofE}
  ~
  \begin{enumerate}
    \item
      Let $\absstate_1 = (\la z_1, \rho \ra, \ataconf_1)$ and $\absstate_2 = (\la z_2, \rho_2 \ra, \ataconf_2)$.
      First, note that $\absstate_1 \absleq \absstate_2$ implies $\fluentset(z_1) = \fluentset(z_2)$, $\rho_2 = \rho$, and that there is a state $\absstate_2^\downarrow = (\la z_2, \rho \ra, \ataconf_2^\downarrow) \syncbisim \absstate_1$ such that
      $\ataconf_2^\downarrow \subseteq \ataconf_2$.
      We distinguish the type of transition:
      \begin{description}
        \item[Time step:] Assume $\absstate_2 \abstranstime{d} \absstate_2'$.
          Then $\absstate_2' = (\la z_2 \cdot d, \rho \ra, \ataconf_2 + d)$.
          With $\ataconf_2'^{\downarrow} = \ataconf_2^\downarrow + d$, we obtain $\ataconf_2^\downarrow \abstranstime{d} \ataconf_2'^{\downarrow}$ and $\ataconf_2'^{\downarrow} \subseteq \ataconf_2'$.
          As $\syncbisim$ is a time-abstract bisimulation, there exists a $\ataconf_1'$ and a $d'$ such that $\ataconf_1 \abstranstime{d'} \ataconf_1'$ and $\absstate_1' \syncbisim \absstate_2'^{\downarrow}$ for $\absstate_1' = (\la z_1 \cdot d', \rho \ra, \ataconf_1')$ and $\absstate_2'^{\downarrow} = (\la z_2 \cdot d, \rho \ra, \ataconf_2'^{\downarrow})$.
          With $\absstate_1' \syncbisim \absstate_2'^\downarrow$ and $\ataconf_2'^\downarrow \subseteq \ataconf_2'$, we obtain $\absstate_1' \absleq \absstate_2'$.
        \item[Action step:]
          Assume $\absstate_2 \abstranssym{p} \absstate_2'$.
          Then $\absstate_2' = (\la z_2 \cdot p, \rho' \ra, \ataconf_2')$ and such that $\ataconf_2 \atatranssym{F} \ataconf_2'$, where $F = F(z_2 \cdot p)$.
          By \autoref{def:ata-lts}, the successors of a configuration under symbol steps are computed pointwise.
          With that and because $\ataconf_2^\downarrow \subseteq \ataconf_2$, there is a $\ataconf_2'^\downarrow$ such that $\ataconf_2^\downarrow \atatranssym{F} \ataconf_2'^\downarrow$.
          As $\absstate_1 \syncbisim \absstate_2^\downarrow$ and because $\syncbisim$ is a time-abstract bisimulation, there exists a $\ataconf_1'$ with $\ataconf_1 \atatranssym{F} \ataconf_1'$ and such that $\absstate_1' \syncbisim \absstate_2'^\downarrow$ for $\absstate_1' = (\la z_1 \cdot p, \rho'\ra, \ataconf_1')$ and $\absstate_2'^\downarrow = (\la z_2 \cdot p, \rho' \ra, \ataconf_2'^\downarrow)$.
          With $\absstate_1' \syncbisim \absstate_2'^\downarrow$ and $\ataconf_2'^\downarrow \subseteq \ataconf_2'$, we obtain $\absstate_1' \absleq \absstate_2'$.
      \end{description}
    \item Assume $\detstate_2 \dettrans[t]{p} \detstate_2'$ and $\detstate_1 \detleq \detstate_2$.
      Let $\absstate_2 \in \detstate_2$, $\absstate_2^* \in \absstates$, and $\absstate_2' \in \detstate_2'$ such that $\absstate_2 \abstranstime{t} \absstate_2^* \abstranssym{p} \absstate_2'$.
      By definition of \detleq, $\detstate_1 \detleq \detstate_2$ implies that for each $\absstate_2 \in \detstate_2$, there is a $\absstate_1 \in \detstate_1$ with $\absstate_1 \absleq \absstate_2$.
      With \autoref{lma:downward-compat:abs}, there is a $\absstate_1^* \absleq \absstate_2^*$ and a $\absstate_1' \absleq \absstate_2'$ such that $\absstate_1 \abstranstime{t} \absstate_1^* \abstranssym{p} \absstate_1'$.
      Therefore, the set $\detstate_1' = \{ \absstate_1' \mid \exists \absstate_1 \in \detstate_1: \absstate_1 \abstrans[t]{p} \absstate_1' \}$ is not empty, and so $\detstate_1 \dettrans[t]{p} \detstate_1'$.
      Furthermore, as such a $\absstate_1' \in \detstate_1'$ with $\absstate_1' \absleq \absstate_2'$ exists for each $\absstate_2' \in \detstate_2$, it follows that $\detstate_1' \detleq \detstate_2'$.
      \qedhere
  \end{enumerate}
\end{proofE}

\todo[inline]{Need to use completeness somewhere?}



We can finally show that \detdisq is indeed a \ac{WSTS}:
\begin{theoremE}\label{thm:detlts-is-wsts}
  The \ac{LTS} $\detdisq$ with the \wqo $(\detstates, \detleq)$ is a \ac{WSTS}.
\end{theoremE}
\begin{proofE}
  ~
  \begin{enumerate}
    \item As $\fluentset$ and $\sub(\delta)$ are both finite sets and because $H$ is decidable on rational states, the relation $\absleq$ and therefore also the relation $\detleq$ is decidable.
    \item First, $\warrow$ is computable for programs over finite-domain \acp{BAT}:
      As each path in \detleq only contains rational time steps, we may use regression (\autoref{def:regression}) to determine the set of satisfied fluents in every state \detstate of \detleq.
      Regression reduces the query to a propositional query of the form $\bat_0 \models \alpha$, which is decidable.
      Furthermore, \ac{ATA} successors are also computable.
      Therefore, $\suc$ is computable.
    \item As shown in \autoref{lma:downward-compat}, $(\detstates, \detleq)$ is downward-compatible.
      \qedhere
  \end{enumerate}
\end{proofE}

With this \ac{WSTS} and with the observation that all accepting states of \detlts are downward-closed with respect to \detleq, we can apply \autoref{thm:wsts-decidability} to obtain:
\begin{corollaryE}\label{thm:mtl-verification}
  The \ac{MTL} verification problem for finite-domain \golog programs over finite traces is decidable.
\end{corollaryE}
\begin{proofE}
  By \autoref{lma:path-trace-equivalence}, the program violates the specification $\phi$ iff there is an accepting path in \detlts.
  Let $V \subseteq \detstates$ be the accepting states of \detlts.
  Clearly, $V$ is downward-closed with respect to $\detleq$: Assume $\detstate \in V$ and $\detstate' \detleq \detstate$.
  From $\detstate' \detleq \detstate$, it directly follows that there is a $\syncstate' \in \detstate'$ for every $\syncstate \in \detstate$.
  As $\detstate$ is accepting, there is some accepting $\syncstate \in \detstate$ and therefore, by definition of \syncleq, there is also an accepting $\syncstate' \in \detstate'$.
  So $\detstate'$ is accepting and therefore $\detstate' \in V$.
  By \autoref{thm:wsts-decidability}, it is decidable whether there is a sequence of actions ending in $V$.
\end{proofE}
With \autoref{thm:mtl-complexity} and \autoref{thm:mtl-equivalence}, we directly obtain:
\begin{corollary}\label{thm:verification-complexity}
  The \ac{MTL} verification problem for finite-domain \golog programs over finite traces has non-primitive recursive complexity.
\end{corollary}

\section{Timed Games}\label{sec:timed-game}
In the previous section, we have shown that the \mtl verification problem for \golog programs is decidable.
However, our goal is to synthesize a controller that controls the program execution such that the specification is satisfied.
Note that these problems are closely related: If we can verify that a certain behavior $\phi$ is not observable when executing the program, then \emph{any} control strategy is valid.
On the other hand, once we have determined a controller, it should also be possible to verify that every controller trace adheres to the specification.
Nevertheless, we cannot directly obtain a controller from verification: For verification, we merely check if a certain set of states is reachable; for synthesis, we need to determine a mapping that steers the execution away from this state set.
Therefore, for controller synthesis, we use a variant of \emph{downward closed games}~\cite{abdullaAlgorithmicAnalysisPrograms2000,abdullaDecidingMonotonicGames2003}.
The idea is similar to the verification approach and uses the same \ac{LTS} \detlts: We first build the synchronous product of the program execution and the \ac{ATA}, then we regionalize and determinize the \ac{LTS}.
We can use the resulting \ac{LTS} to determine a control strategy, where the \wqo on \detlts again allows us to stop after a finite number of steps on each path.
To determine the controller, we define a \emph{timed Golog game}, which is a variant of a two-player game on Golog programs.
Intuitively, the game works as follows: Player 1 (the \emph{controller}) selects a set of actions (satisfying certain criteria that guarantee the conditions from \autoref{def:controller}).
The second player (the \emph{environment}) then replies by selecting one action from this set.
If player 1 can guarantee that player 2 can never select an action that ends in a violating state (i.e., an execution of the program that satisfies the undesired behavior $\phi$), then the game is \emph{winning for player 1}.
Otherwise, it is winning for player 2.
If player 1 is winning, then we can extract a control strategy from the player's turns.

Before we describe the algorithm in detail, we first define timed Golog games:
\begin{definition}[Timed Golog Game]
  A \emph{timed Golog game} is a tuple $\game = (\Delta, \phi, A_C \dot\cup A_E)$, where $\Delta = (\bat, \delta)$ is a \golog program over $(\fluents, \clockset)$, $A_C \dot\cup A_E = \batactions$ is a partition of the actions into controller and environment actions, and $\phi$ is a fluent trace formula.

The game is played between the controller $C$ and the environment $E$.
A play $z = (a_1, t_1) (a_2, t_2) \cdots (a_n, t_n) \in \traces$ is built up as follows: Player $C$ chooses a \emph{valid} subset (defined below) of timed actions $U = \{ (a, t)_i \}_i$ that are possible in the initial situation.
Player $E$ responds by choosing one action $(a, t) \in U$.
Player $C$ continues by choosing again a valid subset of timed actions that are possible after executing the first action, to which player $E$ responds by choosing one action, and so on, until a final state has been reached and $E$ chooses the empty set.

Let $z \in \fintraces$ and $\rho \in \sub(\delta)$.
A set of timed actions $U \subseteq \batactions \times \realpos$ is \emph{valid} in configuration $\la z, \rho \ra$ if
\begin{enumerate}
  \item $(a, t) \in U$ implies that $\la z, \rho \ra \warrow \la z \cdot (a, t) \ra, \rho' \ra$ for some $\rho'$,
  \item For each $a_e \in A_E$, if $\la z, \rho \ra \warrow \la z \cdot (a_e, t), \rho' \ra$, then
    \begin{itemize}
      \item $(a_e, t) \in U$, or
      \item there is $a_c \in A_C$ and $t_c < t$ such that $(a_c, t_c) \in U$;
    \end{itemize}
  \item $U = \emptyset$ implies $\la z, \rho \ra \in \mathcal{F}^w$.
\end{enumerate}

A \emph{strategy} for player $C$ is a partial function $f: \traces \times \sub(\delta) \rightarrow \powerset{\batactions \times \realpos}$ such that
\begin{enumerate}
  \item $f$ is defined on $\la \la\ra, \delta \ra$,
  \item if
    \begin{enumerate}
      \item $f$ is defined on $\la z, \rho \ra$,
      \item $(a, t) \in f(\la z, \rho \ra)$, and
      \item $\la z, \rho \ra \warrow \la z \cdot (a, t), \rho' \ra$,
    \end{enumerate}
    then $f$ is defined on $\la z \cdot (a, t), \rho' \ra$,
  \item if $f$ is defined on $\la z, \rho \ra$, then $f(\la z, \rho \ra)$ is valid with respect to $\la z, \rho \ra$.
\end{enumerate}

The set of plays of $f$, denoted by $\plays(f)$, is the set of traces that are consistent with the strategy $f$.
Formally, $z = (a_1, t_1) \cdots (a_n, t_n) \in \plays(f)$ iff
\begin{enumerate}
  \item $\la \la\ra, \delta \ra \warrow \la z^{(1)}, \rho_1 \ra \warrow \cdots \warrow \la z^{(n)}, \rho_n \ra$ for some $\rho_1, \ldots, \rho_n$,
  \item $f(z, \rho_n \ra) = \emptyset$, and
  \item $(a_{i+1}, t_{i+1}) \in f(\la z^{(i)}, \rho_i \ra)$.
\end{enumerate}
A strategy $f$ is winning with respect to undesired behavior $\phi$ iff for every $z \in \plays(f)$: $w, \la\ra, z \models \neg \phi$.
\end{definition}

A timed Golog game indeed captures controller synthesis:
\begin{proposition} \label{prop:controller-winning-strategy-equivalence}
  Let $\Delta$ be a program and $\batactions = A_C \dot\cup A_E$ a partition of the actions into controller and environment actions.
  Then there exists a controller \controller for program $\Delta$ against undesired behavior $\phi$ iff $C$ has a winning strategy in the timed \golog game $(\Delta, \phi, A_C \dot\cup A_E)$.
\end{proposition}

\newcommand*{\open}{\ensuremath{\mathit{Open}}}
\newcommand*{\nodes}{\ensuremath{V}}
\newcommand*{\edges}{\ensuremath{E}}
\newcommand*{\nodelabel}{\ensuremath{\mathrm{label}}}

\begin{algorithm}[tbh]
\begin{algorithmic}
  \Function{GetAncestors}{$\detstate, E$}
  \State $A \gets \{ \detstate \}$
  \ForAll{$\detstate': (\detstate', (a, t), \detstate) \in E$}
    \State $A \gets A \cup \Call{GetAncestors}{\detstate', E}$
  \EndFor
  \State \Return $A$
  \EndFunction
\end{algorithmic}
\caption[Auxiliary function for \autoref{alg:build-tree}]{Auxiliary function for \autoref{alg:build-tree} to find node ancestors.}
\label{alg:get-ancestors}
\end{algorithm}

\begin{algorithm}[tbh]
\begin{algorithmic}
  \Procedure{BuildTree}{$\detstate_0, \dettrans{}$}
  \State $E \gets \emptyset$
  \State $\open \gets \{ \detstate_0 \}$
  \While{$\open \neq \emptyset$}
    \State $\detstate \gets \Call{pop}{\open}$
    \If{$\detstate$ is \emph{bad}}
      \State mark $\detstate$ as \emph{unsuccessful}
    \ElsIf{$\exists \detstate' \in \Call{GetAncestors}{\detstate, E}: \detstate' \detleq \detstate$}
      \State mark $\detstate$ as \emph{successful}
      \Comment{Any bad state reachable from $\detstate$ is reachable from $\detstate'$}
    \ElsIf{$\suc(\detstate, \dettrans{}) = \emptyset$}
      \State mark $\detstate$ as \emph{dead}
      \Comment{Not bad and no successors}
    \Else
      \ForAll{$\detstate', (a, t)$ with $\detstate \dettrans[t]{a} \detstate'$}
        \State $\open \gets \open \cup \{ \detstate' \}$
        \State $\edges \gets \edges \cup \{ (\detstate, (a, t), \detstate') \}$
      \EndFor
    \EndIf
  \EndWhile
  \State \Return $(\detstate_0, \edges)$
  \EndProcedure
\end{algorithmic}
\caption[The algorithm \textsc{BuildTree}.]{
  The algorithm \textsc{BuildTree} that builds a finite tree from the \ac{LTS} \detlts.
}
\label{alg:build-tree}
\end{algorithm}

\begin{algorithm}[tb]
\begin{algorithmic}
  \Function{IsValid}{$U, \detstate, E, A_C$}
  \State $\mi{Enabled} \gets \{ (a, t) \mid (\detstate, (a, t), \detstate') \in E \}$
  \State $\mi{Enabled}_{\mi{Ctl}} \gets \{ (a, t) \in \mi{Enabled} \mid a \in A_C \}$
  \State $\mi{Enabled}_{\mi{Env}} \gets \{ (a, t) \in \mi{Enabled} \mid a \not\in A_C \}$
  \If{$U = \emptyset$}
    \State \Return{$\detstate \text{ is final } \wedge \mi{Enabled}_{\mi{Env}} = \emptyset$}
  \EndIf
  \If{$\mi{Enabled}_{\mi{Env}} \subseteq U$}
    \State \Return $\top$
    \Comment{Choosing all env actions is always valid}
  \EndIf
  \State $t_c \gets \min\{ t \mid (a, t) \in \mi{Enabled}_{\mi{Ctl}} \cap U \}$
  \State $t_e \gets \min\{ t \mid (a, t) \in \mi{Enabled}_{\mi{Env}} \setminus U \}$
  \State \Return $t_c < t_e$
  \Comment{First chosen ctl must be before first non-chosen env}
  \EndFunction
  \Function{IsGoodChoice}{$U, \detstate, E$}
  \ForAll{$(a, t) \in U$}
    \State $\detstate' \gets \suc(\detstate, E, (a, t))$
    \If{$\detstate'.\nodelabel = \bot$}
      \State \Return $\bot$
    \EndIf
  \EndFor
  \State \Return $\top$
  \EndFunction
\end{algorithmic}
\caption[Auxiliary functions \autoref{alg:traverse-tree}.]{Auxiliary functions for the tree traversal in \autoref{alg:traverse-tree} to determine valid and good controller choices.}
\label{alg:traverse-tree-helpers}
\end{algorithm}

\begin{algorithm}[tb]
\begin{algorithmic}
  \Procedure{Visit}{$\detstate, E, A_C$}
  \If{$\detstate$ is \emph{unsuccessful}}
    $\detstate.\nodelabel \gets \bot$
  \ElsIf{$\detstate$ is \emph{successful}}
    $\detstate.\nodelabel \gets \top$
  \ElsIf{$\detstate$ is \emph{dead}}
    $\detstate.\nodelabel \gets \top$
  \Else
    \State $\detstate.\nodelabel \gets \bot$
    \ForAll{$U: \Call{IsValid}{U, \detstate, E, A_C}$}
      \Comment{Check all valid choices of player $E$}
      \If{$\Call{IsGoodChoice}{U, \detstate, E}$}
        \State $\detstate.\nodelabel \gets \top$
       \EndIf
    \EndFor
  \EndIf
  \EndProcedure
  \Procedure{TraverseTree}{$\detstate, E, A_C$}
  \Comment{Post-order traversal starting in $c$}
  \ForAll{$\detstate' \in \suc(\detstate, E)$}
    \State $\Call{TraverseTree}{\detstate', E, A_C}$
  \EndFor
  \State $\Call{Visit}{\detstate, E, A_C}$
  \EndProcedure
\end{algorithmic}
\caption[The procedure \textsc{TraverseTree}.]{The procedure \textsc{TraverseTree} (and its sub-procedures) that traverses and labels the tree bottom-up.}
\label{alg:traverse-tree}
\end{algorithm}

\begin{algorithm}[tb]
\begin{algorithmic}
  \Procedure{CheckForController}{$\detstate_0, \dettrans{}, A_C$}
  \State $(n_0, E) \gets \Call{BuildTree}{\detstate_0, \dettrans{}}$
  \State $\Call{TraverseTree}{n_0, E, A_C}$
  \State \Return $n_0.\nodelabel$
  \EndProcedure
\end{algorithmic}
\caption[The algorithm \textsc{CheckForController}.]{The algorithm \textsc{CheckForController} which builds \detlts and checks whether a controller exists.}
\label{alg:check-for-controller}
\end{algorithm}

We can also apply a strategy on the \ac{LTS} \detlts.
To do so, we introduce some additional notions:
\begin{itemize}
  \item If a state $\detstate$ of \detlts is accepting, we may also call it \emph{bad}.
  \item For a finite trace $z = (a_1, t_1) \cdots (a_n, t_n) \in \fintraces$, let $\synctracestates(z)$ denote the set of states $\{ \syncstate \mid \syncstate_0 \synctrans[t_1]{a_1} \ldots \synctrans[t_n]{a_n} \syncstate \}$.
  \item We write $\suc(c, \dettrans{}) \eqdef \{ c' \mid \exists a,t:\, c \dettrans[t]{a} c' \}$ for the set of successors of a state $c$ in \detlts.
  \item We call a strategy $f$ \emph{maximal} with respect to \synclts if for every finite play $z \in \plays(f)$ and for every state $\syncstate' \in \syncstates$ with $\syncstate' \syncbisim \syncstate$ for some $\syncstate \in \synctracestates(z)$, there is a play $z' \in \plays(f)$ such that $\syncstate' \in \synctracestates(z')$.
    In other words, if the strategy $f$ ends in a state $\syncstate$ and there is a state $\syncstate'$ in \synclts that is bisimilar to $\syncstate$, then there is some play that ends in $\syncstate'$.
  \item For a maximal strategy $f$ and every play $z = (a_1, t_1) \cdots (a_n, t_n) \in \plays(f)$, let $\tracestate(z) \in \detstates$ denote the unique state $\detstate_f$ of the path $\detstate_0 \dettrans[t_1]{a_1} \ldots \dettrans[t_n]{a_n} \detstate_f$.
  \item Finally, we call a maximal strategy $f$ \emph{safe} in \detlts iff for every finite play $z \in \plays(f)$, $\tracestate(z)$ is not bad.
\end{itemize}

We may restrict strategies to maximal strategies without loss of generality:
\begin{lemmaE} \label{lma:maximal-strategies}
  Let \game be a timed \golog game.
  There is a winning strategy in \game iff there is a winning maximal strategy in \game.
\end{lemmaE}
\begin{proofE}
  Clearly, every winning maximal strategy in \game is also a winning strategy in \game.
  Now, suppose \game has a winning strategy $f$ but no winning maximal strategy.
  If no winning maximal strategy exists, there must be a play $z \in \plays(f)$ and a state $\syncstate \in \synctracestates(z)$ such that for some $\syncstate' \syncbisim \syncstate$, $\syncstate'$ is bad.
  But then, since $\syncstate \syncbisim \syncstate'$, $\syncstate$ must be bad and therefore, $w, \la\ra, z \models \phi$.
  Contradiction to the assumption that $f$ is winning.
\end{proofE}

We can now show that if we want to determine a winning strategy for the timed game $\game$, it is sufficient to determine a safe strategy on \detlts:
\begin{lemmaE} \label{lma:winning-strategy-safe-strategy-equivalence}
  There is a winning strategy in the timed \golog game $\game = (\Delta, \phi, A_C \dot\cup A_E)$ iff there is a safe strategy in \detlts.
\end{lemmaE}
\begin{proofE}
  \textbf{$\Rightarrow$:}
  Assume $f$ is a winning strategy in \game.
  By \autoref{lma:maximal-strategies}, we can assume \wolog that $f$ is maximal.
  We show that $f$ is a safe strategy in \detlts:
  Suppose $f$ is not safe, thus there is a play $z \in \plays(f)$ such that $\tracestate(z)$ is bad and therefore, $\tracestate(z)$ is accepting.
  By definition of \detlts, $w, \la\ra, z \models \phi$.
  Therefore, $f$ is not a winning strategy in \game, in contradiction to the assumption.
  \\
  \textbf{$\Leftarrow$:}
  Assume $f$ is a safe strategy in \detlts (and therefore also a maximal strategy).
  We show that $f$ is a winning strategy in \game.
  Suppose $f$ is not winning.
  Then there is a play $z \in \plays(f)$ such that $w, \la\ra, z \models \phi$.
  By definition of \detlts, $\tracestate(z)$ is accepting and therefore bad.
  Therefore, $f$ is not a safe strategy in \detlts, in contradiction to the assumption.
\end{proofE}

We have already seen that \detlts is finitely branching and we have used a \ac{WSTS} to show that the verification problem is decidable.
We did so by using a \wqo on the states \detstates of \detlts, which allowed us to stop on every path in \detlts after a finite number of steps.
We use a similar idea to determine a safe strategy on \detlts:
Whenever we encounter a state $\detstate$ with $\detstate' \detleq \detstate$ for some ancestor $\detstate'$, then we can mark \detstate as successful and stop expanding the path, because the current path will not lead to a bad state.
This idea leads to the following procedure to determine a safe strategy:
\begin{enumerate}
  \item Build a tree from \detlts and stop at \detstate whenever there is an ancestor $\detstate'$ with $\detstate' \detleq \detstate$.
    As $(\detstates, \detleq)$ is a \wqo, we can always stop after a finite number of steps on each path.
    As \detlts is also finitely branching, the resulting tree is finite and the algorithm always terminates.
    The resulting algorithm is shown in \autoref{alg:build-tree}.
  \item Label the tree bottom-up: If the controller can guarantee that only good children are reachable from a node, then label the node as \emph{good}, otherwise label it as \emph{bad}.
    The resulting algorithm is shown in \autoref{alg:traverse-tree}.
\end{enumerate}
Combining the two steps, we obtain \autoref{alg:check-for-controller}, which returns $\top$ if a safe controller exists and $\bot$ otherwise:
\begin{lemmaE} \label{lma:alg-safe-strategy-equivalence}
  There exists a safe strategy on \detlts with controller actions $A_C$ iff \autoref{alg:check-for-controller} returns $\top$ on input $(\detstate_0, \dettransfull{}, A_C)$.
\end{lemmaE}
\begin{proofE}
  \textbf{$\Rightarrow$:}
  Assume \autoref{alg:check-for-controller} returns $\bot$ and therefore $\detstate_0$ is labeled with $\bot$.
  It is easy to see that for every node labeled with $\bot$, either the node is unsuccessful and thus bad, or for every valid choice of actions $U$, the environment can choose one action that leads to a node labeled with $\bot$.
  Therefore, no valid strategy may exist.
  \\
  \textbf{$\Leftarrow$:}
  Let $T$ be the tree constructed by \autoref{alg:check-for-controller} and $T'$ the sub-tree that is obtained from $T$ by removing the nodes labeled with $\bot$.
  We can build a finite tree $T_{\mi{strat}}$ that satisfies the following condition:
  If $\detstate$ is a node of $T_{\mi{strat}}$ that is not \emph{good}, then the set of edges $U$ of $T_{\mi{strat}}$ starting in $\detstate$ is a subset of edges in $T'$ starting in $\detstate$ and such that $U$ is a valid choice of actions for player $C$.
\end{proofE}

As a safe strategy on \detlts exists if and only if a winning strategy exists in the timed game, we can conclude:
\begin{theoremE}\label{thm:synthesis}
  \autoref{alg:check-for-controller} returns $\top$ on input $(\detstate_0, \dettransfull{}, A_C)$ iff there exists a controller for program $\Delta$ against undesired behavior $\phi$ with controllable actions $A_C$.
\end{theoremE}
\begin{proofE}
  The claim directly follows from \autoref{prop:controller-winning-strategy-equivalence}, \autoref{lma:winning-strategy-safe-strategy-equivalence}, and \autoref{lma:alg-safe-strategy-equivalence}.
\end{proofE}
This provides us a decidable procedure for the synthesis problem, hence:
\begin{corollary}
  The control problem for finite-domain \golog programs over finite traces is decidable.
\end{corollary}

\begin{example}
  \begin{figure}[tb]
    \centering
    \includestandalone[width=\textwidth]{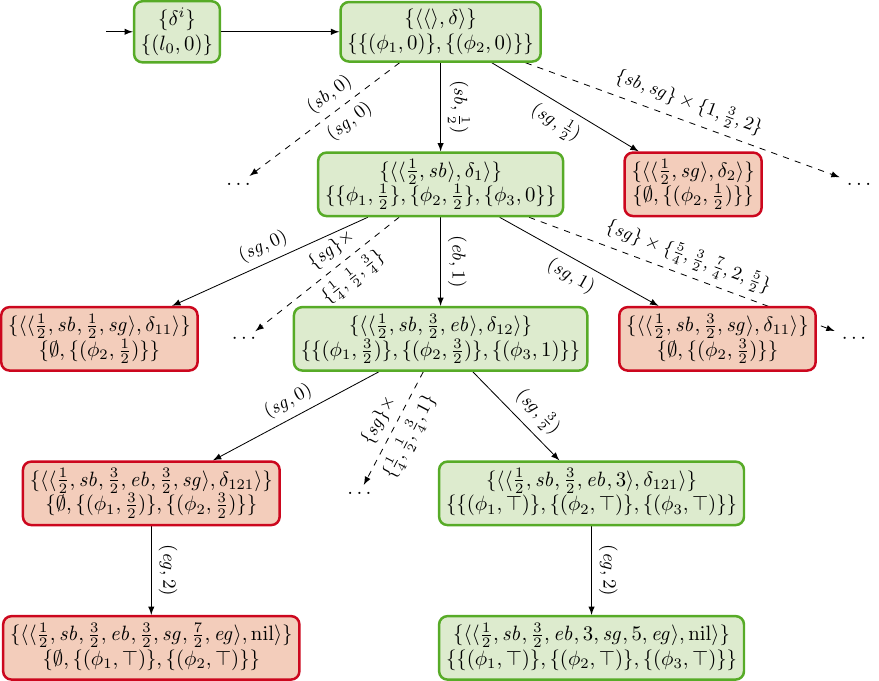}
    \caption[The labeled deterministic discrete quotient.]{The labeled deterministic discrete quotient from \autoref{fig:det-reg-sync-product}, where $\top$-labeled nodes are shown in green and $\bot$-labeled nodes are shown in red.}
    \label{fig:det-reg-sync-product-labeled}
  \end{figure}
  \autoref{fig:det-reg-sync-product-labeled} shows the result of playing the timed game on the deterministic discrete quotient of the running example from \autoref{ex:control-problem}.
  As the initial node is labeled with $\top$, by \autoref{thm:synthesis}, there exists a controller for the program against the undesired behavior $\phi = \finally{}(\neg \camon \wedge \grasping) \vee \finally{}(\neg \camon \wedge \finally{[0, 2]} \grasping)$.
\end{example}

\subsection{Extracting a Controller}%
\label{ssub:controller-extraction}

Usually, we actually want to generate a controller, not just decide whether a controller exists.
We can extract a controller from the labeled search tree from \autoref{alg:check-for-controller}:
First, we traverse \detlts and choose every action that leads to a node that is labeled with $\top$.
As each time step in \detlts is a representative of an equivalence class of \syncbisim, each such action is a representative for a set of timed actions with equivalent time steps.
These time steps are a convex set that can be directly computed from the region indices, which can be represented as clock constraints.
Therefore, it is usually convenient to represent the controller as \iac{TA}.
An example will be shown in the next section, where we evaluate an implementation of the approach in several scenarios.

\section{Evaluation}\label{sec:synthesis-evaluation}

We have implemented the synthesis approach in our tool \gocos (\emph{Golog Controller Synthesis}) by extending our own \tacos~\cite{hofmannTACoSToolMTL2021,hofmannControllingTimedAutomata2023} to \golog programs.
While \tacos implements the synthesis approach described by \parencite{bouyerControllerSynthesisMTL2006} and therefore controls a \ac{TA} against \ac{MTL} specification, \gocos works in a similar way, but uses \golog programs instead of \acp{TA} as execution model, following the approach described in the previous sections.
\gocos is implemented in \cpp and uses \gologpp \cite{matareGologIntegrativeSystem2018,matarePortableHighlevelAgent2021} as the underlying \golog framework.
It provides a \cpp API which allows integrating the synthesis method into other frameworks, e.g., a \golog execution engine.
Additionally, it supports human-readable text input in the form of protobuf messages for \ac{MTL} formulas and \gologpp programs for program input.

The implementation differs from the theoretical framework described above in several aspects:
\begin{enumerate}
  \item Rather than storing concrete candidates in the search tree, it directly stores a canonical word representation based on the abstraction function (\autoref{def:abstraction-function}).
    Time increments are directly represented by the index of the region increment (\autoref{def:region-increment}) rather than by the absolute value.
  \item As nodes in the search tree may be reachable via different paths,
    rather than computing the same sub-tree multiple times, the nodes with the same node label are merged.
    The resulting structure is a \emph{search graph} rather than a \emph{search tree}.
    This significantly reduces the number of nodes and therefore increases the performance of the tool.
  \item Node labels are determined on-the-fly, i.e., while the search graph is expanded.
    If a node's label can already be determined, all its successor nodes are closed and not further expanded.
    Therefore, the timed game described in \autoref{sec:timed-game} is solved while constructing the deterministic discrete quotient from \autoref{sec:determinization}.
    This allows pruning parts of the search graph and therefore further reduces the size of the search graph.
  \item As \gocos extends \tacos, it can also control \acp{TA} against a specification.
    However, it currently cannot combine \acp{TA} with \golog programs.
    Therefore, for the sake of this evaluation, we assume that the actions of the self model are also directly included in the main program.
  \item The executor of \gologpp assumes all actions to be durative actions.
    When executing the program, each durative action is implicitly split into a \emph{start} and an \emph{end} action.
    The precondition of the actions apply to the \emph{start} action, while the precondition of the end action only checks that there has been a corresponding \emph{end} action.
    The clock constraint of the \emph{end} action enforces the action duration specified in the program.
    The \emph{start} effects are applied after the \emph{start} action, while the other effects are applied after the \emph{end} action.
    Therefore, the explicit encoding of durative actions as shown in \autoref{ex:bat} is not necessary and done implicitly by the program interpreter.
  \item Due to limitations of \gologpp, clocks are restricted to measure the duration of durative actions.
    Rather than having a fixed set of clocks, a clock is only added once the respective \emph{start} action occurs.
    As there must always be at least one clock, a special clock named \texttt{golog} is used if no other clock has been used yet.
\end{enumerate}


We have evaluated the implementation in two variants of a scenario with a mobile robot that transports objects:
\begin{description}
  \item[Camera] In this scenario, the robot needs to turn on its camera before it can grasp an object.
    As the camera needs some time to initialize, it needs to be running for a certain time before the robot can grasp any object.
  \item[Household] In this scenario, the robot moves around between different locations to collect objects.
    Before it can grasp an object, it needs to align precisely to the target location.
    This scenario is loosely inspired from \parencite{hofmannContinualPlanningGolog2016}.
\end{description}

\subsection{Camera}

\begin{lstlisting}[float,label={lst:camera-fluents},style=golog,caption={
  The object and fluent definitions of the robot camera example.
}]
symbol domain Location = { m1, m2 }
symbol domain Object = { obj1 }
bool fluent robot_at(Location l) {
  initially: (m1) = true;
}
bool fluent obj_at(Object obj, Location l) {
  initially: (obj1, m2) = true;
}
bool fluent holding(Object obj) {
  initially: (obj1) = false;
}
bool fluent grasping() {
  initially: () = false;
}
bool fluent camera_on() {
  initially: () = false;
}
\end{lstlisting}

\begin{lstlisting}[float,label={lst:camera-actions},style=golog,caption={
  The actions of the robot in the robot camera example.
}]
action drive(Location from, Location to) {
  duration: [1, 2]
  precondition: robot_at(from)
  start_effect: robot_at(from) = false;
  effect: robot_at(to) = true;
}
action grasp(Location from, Object obj) {
  duration: [1, 1]
  precondition: robot_at(from) & obj_at(obj, from)
  start_effect: grasping() = true;
  effect:
    grasping() = false;
    obj_at(obj, from) = false;
    holding(obj) = true;
}
action boot_camera() {
  duration: [1, 1]
  precondition: !camera_on()
  effect: camera_on() = true;
}
action shutdown_camera() {
  duration: [1, 1]
  precondition: camera_on()
  start_effect: camera_on() = false;
}
\end{lstlisting}

\begin{lstlisting}[float,label={lst:camera-program},style=golog,caption={
  [The main program in the robot camera example.]The main program in the robot camera example, where the camera may boot and shutdown exactly once.
}]
procedure main() {
  concurrent {
    { drive(m1, m2); grasp(m2, obj1); }
    { boot_camera(); shutdown_camera(); }
}}
\end{lstlisting}

\begin{lstlisting}[float,label={lst:camera-program-looped},style=golog,caption={
  [The main program in the extended robot camera example.]The main program in the robot camera example, where the camera may be booted and shut down in a loop until the robot has reached its goal.
}]
procedure main() {
  concurrent {
    { drive(machine1, machine2); grasp(machine2, obj1); }
    while (!holding(obj1)) { boot_camera(); shutdown_camera(); }
}}
\end{lstlisting}

We first consider and extend the running example, as introduced in \autoref{ex:bat}.
In this scenario, the robot is able to move between locations and it may grasp an object from a location.
As shown in \autoref{lst:camera-fluents}, the world is described with the following objects and fluent predicates:
\begin{itemize}
  \item There are two locations |m1| and |m2| of type |Location|.
    Furthermore, there is a single object |obj1|.
  \item The predicate |robot_at(l)| is true if the robot is currently in location $l$.
    Initially, the robot is at |m1|.
  \item The predicate |obj_at(obj, l)| describes an object's location.
    Initially, the (only) object |obj1| is at the location |m2|.
  \item The 0-ary fluent |grasping()| is true if the robot is currently grasping an object.
  \item The 0-ary fluent |camera_on()| is true if the robot's camera is turned on.
\end{itemize}
As shown in \autoref{lst:camera-actions}, the robot has the following high-level durative actions available:
\begin{itemize}
  \item It may |drive| from one location to another.
    While the robot is driving, it is at no location.
    After finishing the action, the robot is at the location specified by the action parameter |to|.
  \item It may |grasp| an object, which is only possible if the robot and the object are at the same location.
    While the robot is grasping an object, the fluent |grasping()| is true.
    Afterwards, the object is no longer at the location, but instead the robot is |holding| the object.
\end{itemize}
Additionally, the robot can also perform two low-level durative actions:
\begin{itemize}
  \item It may boot its camera with the action |boot_camera()|.
  \item It may also shut down the camera again with the action |shutdown_camera()|.
\end{itemize}
We consider two variants of the main program.
In both variants, the robot first drives to |m2| and then grasps the object.
The variants differ in how they model the camera:
\begin{enumerate}
  \item In the simple version, shown in \autoref{lst:camera-program}, the robot may boot and shut down the camera exactly once.
  \item In the looped version in \autoref{lst:camera-program-looped}, the robot may repeatedly boot and shut down the camera until it has reached its goal (i.e., it is holding |obj1|).
\end{enumerate}
We partition the actions into controllable and environment actions as follows: Each \emph{start} action of a durative action is under the agent's control, while each \emph{end} action  is under the environment's control.
Therefore, the agent can decide when it wants to start executing an action, but the environment determines how long the action takes (within the duration constraints given by the \ac{BAT}).
Finally, the undesired behavior is modeled with the following \ac{MTL} formula:
\[
  \finally{} (\neg \camon \wedge \grasping) \vee \finally{} (\neg \camon \wedge \finally{[0, 1]} \grasping)
\]
Therefore, it is undesired behavior if at any point in time the camera is off while the robot is grasping the object, or if the camera is off and the robot will be grasping an object within the next \SI{1}{\sec}.
In other words, the camera must have been running for at least \SI{1}{\sec} before the robot may grasp.

\begin{figure}[ht]
  \centering
  \includegraphics[width=\textwidth]{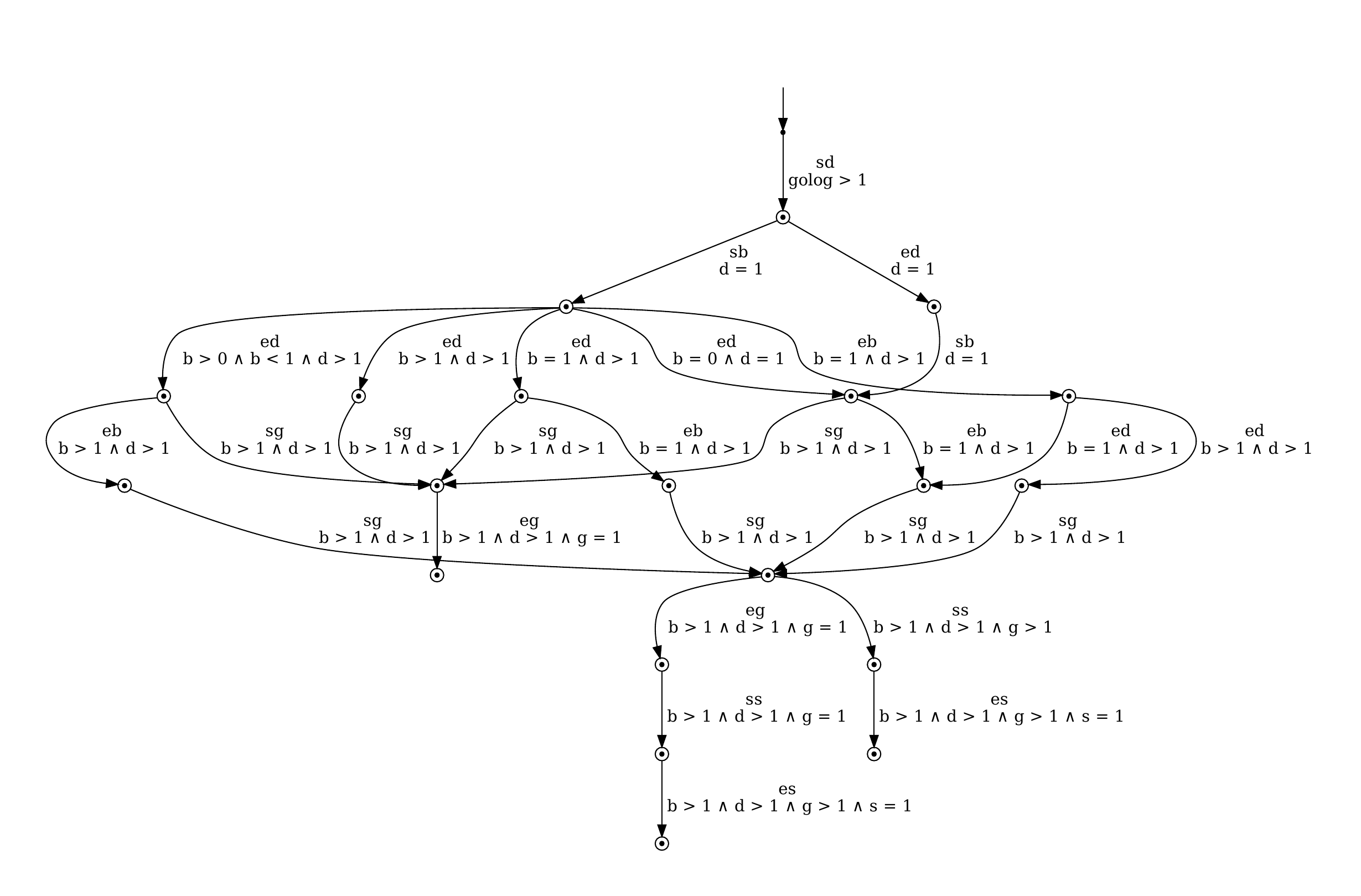}
  \caption[A controller for the robot scenario with a camera  boot time of \SI{1}{\sec}.]{A controller for the robot scenario with a camera  boot time of \SI{1}{\sec}. Actions and clock names are abbreviated with \texttt{d:=drive(m1, m2)}, \texttt{g:=grasp(m2, obj1)}, \texttt{b:=boot\_camera()}, and \texttt{s:=shutdown\_camera()}, and where the prefixes \texttt{s} and \texttt{e} indicate the corresponding start and end actions.}
  \label{fig:robot-controller-1}
\end{figure}

\autoref{fig:robot-controller-1} shows the resulting controller.
We can see that the controller first starts the first action of the program |drive(m1, m2)| and then, depending on how long the drive action takes, either starts booting the camera before or after the drive action ends.
After the action |boot_camera()| ends, the controller waits until the clock constraint |boot_camera() > 1| is satisfied and then starts executing |grasp(m2, obj1)|.
After the grasp action ends, the controller continues by shutting down the camera.

We can observe multiple interesting aspects about the synthesized controller:
\begin{enumerate}
  \item Whenever the controller waits for an action to end, there is one successor for each possible duration of the action.
    This is because the end action and therefore the action duration is not under the agent's control.
    Therefore, it needs to consider every possible end action.
    This is different for the start actions: As these are under the agent's control, it may select one of the multiple possible actions.
  \item The controller also contains paths that seem to be invalid because they violate the specification, e.g.,
    \begin{multline*}
      \circ \taltstrans{\mi{golog} > 1}{\sac{\drive(m1, m2)}} \circ \taltstrans{\drive(m1, m2) = 1}{\sac{\mi{boot\_camera}()}} \circ \\
      \taltstrans{\mi{boot\_camera}() = 1 \wedge \drive(m1,  m2) > 1}{\eac{\drive(m1, m2)}} \circ \taltstrans{\mi{boot\_camera} > 1 \wedge \drive(m1,  m2) > 1}{\sac{\grasp(m2, o)}} \circ \dashrightarrow \circ
    \end{multline*}
    However, none of those paths end in a final configuration of the program.
    In fact, this path may only occur if an expected \emph{end} action does not occur.
    In this example, the action $\mi{boot\_camera}()$ has not ended and the corresponding end action may no longer occur because the duration constraint cannot be satisfied: The duration of the action is $[1, 1]$, but the corresponding clock value satisfies $\mi{boot\_camera} > 1$.
    As this path may no longer end in a final configuration, the controller may execute \emph{any} action.
    Note that none of these actions are necessary and the controller may also decide to do nothing, i.e., the shown controller is not minimal.
\end{enumerate}

\begin{table}[ht]
  \centering
  \caption[Evaluation of the camera scenario.]{Evaluation of the camera scenario.
    In \emph{Camera (simple)}, the robot may turn and off its camera exactly once, as shown in \autoref{lst:camera-program}.
    \emph{Camera (scaled)} is the same program, but with a specification that requires more time for the camera to initialize.
    \emph{Camera (looped)} uses the program from \autoref{lst:camera-program-looped}, where the robot may turn on and off the camera repeatedly.
    Each configuration was run five times.
    The table shows the mean of the total CPU time in seconds, the number of all nodes and explored nodes in the search graph, and the size of the controller.
  }
  \label{tab:gocos-robot-evaluation}
  \begin{tabular*}{\textwidth}{@{\extracolsep{\fill}} l  r r r r r r @{}}
\toprule
{Scenario} & scale $k$& CPU (s) & nodes & expl & ctrl
\tabularnewline\midrule
Camera (simple) & 1 & \num{3} & \num{160} & \num{100} & \num{22}
 \tabularnewline 
Camera (scaled) & 2 & \num{4} & \num{217} & \num{145} & \num{22}
 \tabularnewline 
 & 3 & \num{6} & \num{278} & \num{174} & \num{22}
 \tabularnewline 
 & 4 & \num{19} & \num{628} & \num{473} & \num{26}
 \tabularnewline 
 & 5 & \num{25} & \num{747} & \num{559} & \num{23}
 \tabularnewline 
 & 6 & \num{49} & \num{1357} & \num{999} & \num{24}
 \tabularnewline 
 & 7 & \num{87} & \num{2016} & \num{1444} & \num{25}
 \tabularnewline 
 & 8 & \num{90} & \num{1847} & \num{1395} & \num{24}
 \tabularnewline 
 & 9 & \num{160} & \num{3017} & \num{2532} & \num{26}
 \tabularnewline 
 & 10 & \num{100} & \num{1935} & \num{1499} & \num{25}
 \tabularnewline 
Camera (looped) & 2 & \num{42} & \num{377} & \num{216} & \num{37}
 \tabularnewline 
 & 3 & \num{2849} & \num{747} & \num{397} & \num{49}
 \tabularnewline 
\bottomrule \end{tabular*}
\end{table}

We consider multiple variants of this scenario that differ in the time that the camera needs to be running before it may be used.
We do so by introducing a parameter $k \in \naturals$ in the specification:
\[
  \finally{} (\neg \camon \wedge \grasping) \vee \finally{} (\neg \camon \wedge \finally{[0, k]} \grasping)
\]
For evaluation, we measured
\begin{enumerate}
  \item the total CPU time in seconds,
  \item the number of nodes in the search graph,
  \item the number of explored nodes in the search graph, i.e., nodes that have not been pruned,
  \item the size of the resulting controller.
\end{enumerate}

\begin{figure}[ht]
  \centering
  \includegraphics[width=\textwidth]{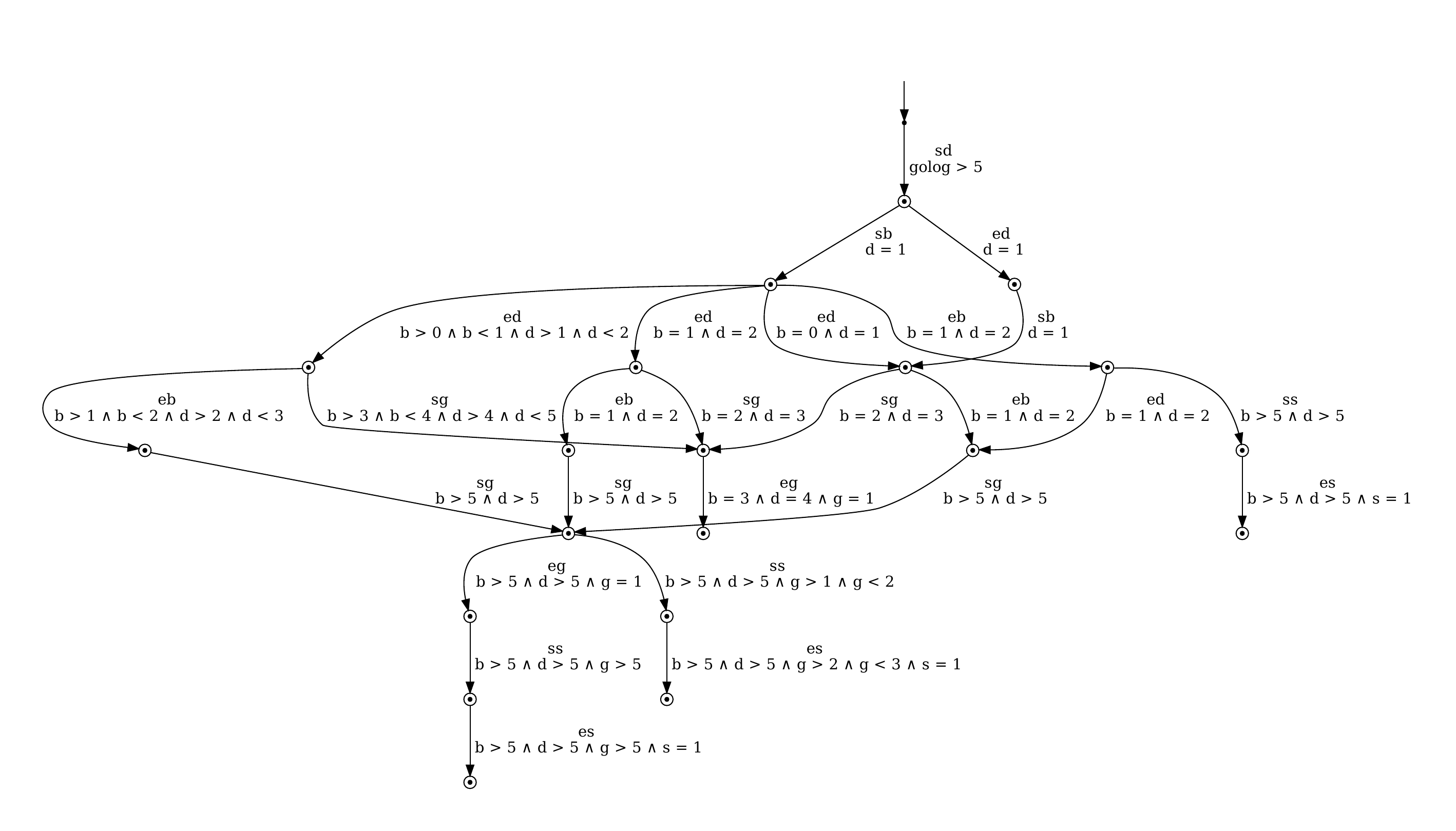}
  \caption{A controller for the robot scenario with a camera  boot time of \SI{5}{\sec}.}
  \label{fig:robot-controller-5}
\end{figure}

\autoref{tab:gocos-robot-evaluation} shows the results for the different settings.
We can see that for small $k$, the tool can synthesize a controller in a reasonably short time (e.g., in \SI{19 \pm 4}{\sec} for $k = 4$).
While the number of total nodes and explored nodes increase with $k$, the size of the resulting controller does not vary much.
Comparing the controllers for $k = 1$ (\autoref{fig:robot-controller-1}) and $k = 5$ (\autoref{fig:robot-controller-5}) provides some explanation.
With increasing $k$, we require more time between booting the camera and using it.
However, the action durations are unchanged.
Therefore, the branching due to the environment's actions, which is caused by the \emph{end} actions, does not differ significantly between the two scenarios.

Coming back to \autoref{tab:gocos-robot-evaluation}, we can also see that if the robot may turn on and off its camera arbitrarily often, then the synthesis does not scale well anymore.
Synthesizing a controller for $k = 3$ already takes a mean time of \SI{2849}{\sec}.
For any larger $k$, the tool does not terminate within a reasonable time.
Interestingly, the number of nodes in the search graph does not increase as much as the running time.
This suggests that many nodes can be reached via many different paths, e.g., by executing the camera loop once or twice.
Further analysis of this problem may help to reduce the total running time for larger $k$.

\subsection{Household}

\begin{lstlisting}[float,label={lst:household-fluents},style=golog,caption={
  The object and fluent definitions of the household example.
}]
symbol domain Location = {lroom, sink, table}
symbol domain Object = {cup1}
bool fluent robot_at(Location l) {
  initially: (lroom) = true;
}
bool fluent moving() {
  initially: () = false;
}
bool fluent grasping() {
  initially: () = false;
}
bool fluent cup_at(Object o, Location l) {
  initially: (cup1, table) = true;
}
bool fluent aligned(Location l) {
  initially: (table) = false;
}

\end{lstlisting}
\begin{lstlisting}[float,label={lst:household-actions},style=golog,caption={
  The actions of the robot in the household example.
}]
action move(Location from, Location to) {
  precondition: robot_at(from)
  start_effect:
    moving() = true;
    robot_at(from) = false;
  effect:
    moving() = false;
    robot_at(to) = true;
}
action grasp(Location l, Object o) {
  precondition: robot_at(l) & cup_at(o, l)
  start_effect: grasping() = true;
  effect:
    grasping() = false;
    cup_at(o, l) = false;
}
action align(Location l) {
  precondition: robot_at(l)
  effect: aligned(l) = true;
}
action back_off(Location l) {
  precondition: aligned(l)
  effect: aligned(l) = false;
}

\end{lstlisting}
\begin{lstlisting}[float,label={lst:household-program},style=golog,caption={
  The main program in the household example.
}]
procedure main() {
  concurrent {
    { move(lroom, table); grasp(table, cup1); move(table, sink); }
    if (!robot_at(sink)) { align(table); back_off(table); }
  }
}
\end{lstlisting}

As in the previous scenario, the robot is able to move between locations and it may grasp an object from a location.
In contrast to the previous scenario, we are not considering the robot's camera, but instead we require the robot to fine-align to each location before it grasps an object.
Furthermore, we do not fix the action's durations, each action may take arbitrarily long.
As shown in \autoref{lst:household-fluents}, the world is described with the following objects and fluent predicates:
\begin{itemize}
  \item There are three locations |lroom|, |sink|, and |table| and a single object |cup1|.
  \item The predicate |robot_at(l)| is true if the robot is currently in location $l$.
    Initially, the robot is in the living room, i.e., |robot_at(lroom)| is true.
  \item The 0-ary  fluent |moving()| describes whether the robot is currently moving to a different location.
  \item The 0-ary fluent |grasping()| describes whether the robot is currently doing a |grasp| action, i.e., trying to grasp an object.
  \item The binary fluent |cup_at(c, l)| states that the cup |c| is at location |l|.
    Initially, |cup1| is at |table|.
  \item The unary fluent |aligned(l)| describes whether the robot is fine-aligned to location |l|.
    Initially, it is not aligned anywhere.
\end{itemize}
As shown in \autoref{lst:household-actions}, the robot has the following high-level durative actions available:
\begin{itemize}
  \item It may |move| from one location to another, similarly  to |drive| in the previous scenario.
    While the robot is moving, the fluent |moving()| is true.
  \item It may |grasp| an object from a location.
    As before, it may do so only if the robot and the object are at the same location.
    While the robot is grasping an object, the fluent |grasping()| is true.
    Afterwards, the object is no longer at the location.
\end{itemize}
Additionally, the robot can also perform two low-level durative actions:
\begin{itemize}
  \item It may |align(l)| to location |l|, which has the effect that |aligned(l)| is true.
  \item Once it has aligned to a location, it may |back_off| again, which sets |aligned(l)| to false and allows the robot to move freely again.
\end{itemize}

As before, each \emph{start} action is controllable, while each \emph{end} action is under the environment's control.
None of the durative actions has a specified duration, therefore the environment may choose an arbitrary time for each \emph{end} action.
In this scenario, we use two clocks: The clock |golog| keeps track of the time since the last action and is reset on any action, the clock |align(table)| is only reset when the action |align(table)| \emph{ends}, i.e., it measures the time since the robot has aligned to the location |table|.

The undesired behavior is modeled with the following parameterized \ac{MTL} formula:
\[
  \finally{}\big[ \moving \wedge \raligned \vee \neg \raligned \wedge \grasping \vee \neg \raligned \wedge \finally{[0, k]} \grasping \big]
\]
This specification says that it is undesired behavior if at any point the robot is moving while it is still aligned to some location, or if it is currently grasping an object but not aligned, or if it is currently not aligned and it will be grasping within the next $k$ seconds, where $k \in \naturals$ again is a parameter used for scaling.

The main program is shown in \autoref{lst:household-program}.
In the main program, the robot first moves to |table|, then grasps |cup1|, and then moves to the sink.
Concurrently, the robot aligns to |table|.
The main task for the controller is to determine the action sequence and action time points for each of those actions, such that the resulting traces are guaranteed to satisfy the specification.

\begin{table}[ht]
  \centering
  \caption[Evaluation of the Household scenario.]{Evaluation of the Household scenario.
    Each configuration was run five times.
    The table shows the mean of the total CPU time in seconds, the number of all nodes and explored nodes in the search graph, and the size of the controller.
  }
  \label{tab:gocos-household-evaluation}
  \begin{tabular*}{\textwidth}{@{\extracolsep{\fill}} l  r r r r r r @{}}
\toprule
{Scenario} & scale $k$& CPU (s) & nodes & expl & ctrl
\tabularnewline\midrule
Household & 1 & \num{10} & \num{299} & \num{164} & \num{26}
 \tabularnewline 
 & 2 & \num{47} & \num{847} & \num{408} & \num{38}
 \tabularnewline 
 & 3 & \num{243} & \num{2565} & \num{1195} & \num{59}
 \tabularnewline 
 & 4 & \num{334} & \num{3550} & \num{1406} & \num{69}
 \tabularnewline 
\bottomrule \end{tabular*}
\end{table}

\begin{sidewaysfigure}
  \centering
  \includegraphics[width=\textwidth,height=\textheight,keepaspectratio]{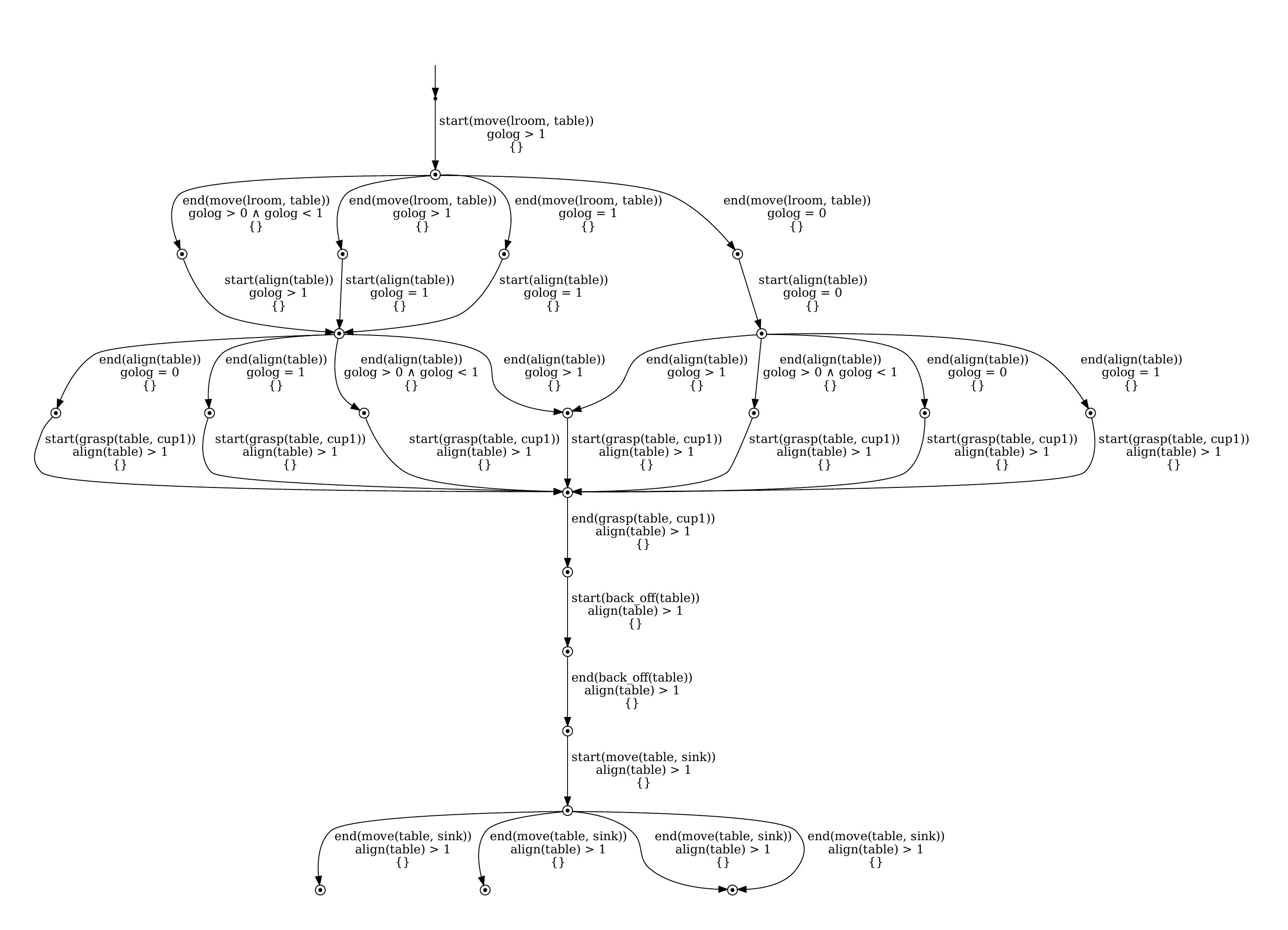}
  \caption{Controller for the household scenario with an align time of \SI{1}{\sec}.}
  \label{fig:household-controller-1}
\end{sidewaysfigure}

\autoref{tab:gocos-household-evaluation} shows the evaluation results.
In comparison to the previous scenario (\autoref{tab:gocos-robot-evaluation}), we can see that this scenario scales much worse.
With a required minimum time of $k = 4$ between finishing to align and starting to grasp, the mean time to synthesize a controller is already $\SI{334}{\sec}$.
Similarly, the number of nodes in the search graph also grows much more quickly.
One reason for this poor scaling behavior is that we do not have any restrictions on any of the action durations.
Therefore, we obtain a high number of time successors for each possible state.
In the worst case, each of those successors needs to be explored separately to determine whether a controller exists.
This can also be seen in \autoref{fig:household-controller-1}, which shows the resulting controller for $k = 1$.
In the top half of the controller, we see many nodes and transitions that only differ in some clock value.
As each of those nodes needs to be explored separately, the size of the search graph increases quickly for larger $k$.

\section{Discussion}\label{sec:synthesis-discussion}

In this chapter, we have viewed the transformation problem as a \emph{synthesis problem}, where some of the program's action are under the agent's control, while the other actions are controlled by the environment.
In this setting, the synthesis problem is to determine a controller that executes the \golog program such that every resulting execution trace satisfies the given specification, no matter how the environment acts.
We have seen that the synthesis problem is decidable for \golog programs over finite domains if we only consider finite program traces and it is undecidable for infinite program traces.
The decidability proof is constructive and results in a controller that executes the program.
We have described an implementation of the approach based on the synthesis tool \tacos.
The tool is able to synthesize controllers in several settings, but it does not scale well with larger problem instances.
The \ac{MTL} satisfiability problem and therefore also the synthesis problem has non-primitive recursive complexity, so it is not surprising that it does not scale well.
On the other hand, as discussed in \autoref{sec:ta}, deciding language emptiness of \acp{TA} is also \textsc{Pspace}-complete, yet tools such as \uppaal are able to verify properties  on larger instances.
It is conceivable that this is in part due to the considerable efforts put into improving the performance of state-of-the-art tools, e.g., with symbolic model checking~\parencite{larsenModelcheckingRealtimeSystems1995}.
While \tacos has seen some efforts (e.g., search node re-usage ~\cite{hofmannTACoSToolMTL2021}) towards performance improvement, many state-of-the-art techniques such as symbolic model checking are also applicable to \tacos but have not been implemented yet.
Therefore, for future work, it may be interesting to apply those methods to \tacos, both for \ac{TA} and \golog controller synthesis.

A different approach towards better scalability would be to consider less expressive fragments of \ac{MTL}, e.g., MITL~\parencite{alurBenefitsRelaxingPunctuality1996}, where intervals must be non-singular, \mtlzi~\parencite{alurBenefitsRelaxingPunctuality1996,henzingerItTimeRealtime1998}, where every time bound has a lower bound of $0$ or an upper bound of $\infty$, Safety MTL~\parencite{ouaknineSafetyMetricTemporal2006}, where the until operator \until{I} may only occur with bounded intervals $I$, and time-bounded MTL~\parencite{ouaknineTimeBoundedVerification2009}, where the time horizon is fixed a priori.
These variants of \ac{MTL} and the complexity of the respective model checking problems are discussed in \parencite{ouaknineRecentResultsMetric2008}).
Restricting the logic to a subset of \ac{MTL} such as Safety \ac{MTL} would also allow us to verify properties and synthesize controllers for non-terminating programs.
As an example, the \ac{TA} control problem for Safety \ac{MTL} is decidable, even over infinite words~\parencite{bouyerControllerSynthesisMTL2006}.


\chapter{Plan Transformation as Reachability Analysis}\label{chap:transformation-as-reachability-problem}

In the previous chapter, we have described a transformation approach based on \ac{MTL} synthesis.
We have seen that this approach is quite general, the resulting controller controls an arbitrary \golog program such that each trace is guaranteed to satisfy the specification.
This controller works against every possible environment, which may control some of the actions of the program.
In particular, it may determine the duration of durative actions by controlling the corresponding \emph{end} action of each durative action.
Also, the approach allows full \ac{MTL} and nondeterministic expressions in the \golog program.
However, it does not scale well with larger problem instances.

For this reason, we describe a second, simpler approach in this chapter.
To simplify the problem, we make the following assumptions:
\begin{enumerate}
  \item Instead of a program, we consider a single plan, i.e., a sequence of actions.
  \item We do not distinguish between controller and environment actions anymore, i.e., the interpreter is in control of every action and there is no devilish nondeterminism controlled by the environment.
  \item The task is to insert additional actions into the sequence to satisfy the specification.
    The original plan is not modified, but only augmented by additional actions.
  \item In addition to determining necessary platform actions, we also need to determine the execution time point of all actions (both plan and platform actions).
  \item Furthermore, we restrict the constraint language.
    Instead of allowing full \ac{MTL}, we only consider a fragment that is useful for our application.
    Also, the constraints are on actions rather than on fluents.
  \item Plan and platform actions operate on a disjoint domain.
    Hence, we do not need to deal with preconditions and effects of the plan actions and can use them as \ac{MTL} symbols in the constraints.
\end{enumerate}

In the following, we show that based on these assumptions, we can reduce the transformation problem to a \emph{reachability problem} on \acp{TA}.
We do this by first constructing a \ac{TA} that corresponds to the abstract plan.
In the next step, we do a parallel composition of the plan \ac{TA} and the platform model, which is also given as a \ac{TA}.
The resulting automaton is then processed such that it only permits transitions that do not violate the constraints.
Hence, we only need to determine a path that reaches a final state of the automaton.
By construction, this will correspond to an execution of the abstract plan with additional platform actions that satisfies the specification.

After describing the procedure and showing its correctness, we evaluate the approach based on a benchmark from the \acf{RCLL}.


\section{The Transformation Problem}

We start by defining the transformation problem.
As a first input, we are given a plan $\sigma = \la a_1, a_2, \ldots, a_n \ra$ that consists of a sequence of actions.
Additionally, we are given a self model of the robot in the form of a \ac{TA} \taplatform.\footnote{
  For reasons that will become apparent later on, we assume that \taplatform contains self-looping $\varepsilon$ transitions for each location, i.e., for every location $l$, there is a switch $(l, \varepsilon, \top, \emptyset, l) \in E$.
}
We will restrict the constraint language to contain formulas over action symbols rather than fluents of the domain.
Therefore, for the purpose of plan transformation, the definition of the action's preconditions and effects is irrelevant.
In a first step, we construct a \ac{BAT} that captures both $\sigma$ and \taplatform.
We start with the \ac{BAT} $\bat_{\taplatform} = \bat_{\taplatform}^{\text{pre}} \cup \bat_{\taplatform}^{\text{post}} \cup \bat_{\taplatform}^{0}$ constructed from \taplatform according to \autoref{sec:tesg-ta}.
Next, we augment $\bat_{\taplatform}$ to include $\sigma$ to obtain a combined \ac{BAT} \bat as follows:
\begin{itemize}
  \item As $\sigma$ is a valid plan, we augment the precondition axiom to also always allow every plan action.
    Let $\bat_{\taplatform}^{\text{pre}} = \{ \square \poss(a) \equiv \pi_{\taplatform} \}$ be the precondition axiom of $\bat_{\taplatform}$.
    We define the new precondition axiom $\bat_{\text{pre}}$ as follows:
    \[
      \square \poss(a) \equivspace \pi_{\taplatform} \vee \bigvee_i a = a_i
    \]
  \item As we want to refer to action occurrences in the constraint language, we add the following successor state axioms to the successor state axioms $\bat_{\taplatform}^{\text{post}}$ of $\bat_{\taplatform}$:
    \begin{align*}
      \square [a] \occ(a') &\equivspace a' = a
      \\
      \square [a] \planorder(i) &\equivspace a = a_i \vee \planorder(i) \wedge \bigwedge_j a \neq a_j
    \end{align*}
    The fluent $\occ(a)$ is true iff $a$ is the action that is currently occurring (more precisely, $a$ is the action that resulted in the current situation).
    For the sake of brevity, we we will also just write $a$ for $\occ(a)$.
    The fluent $\planorder(i)$ is true iff the $i$th action $a_i$ of $\sigma$ was the last high-level action.
    Therefore, $\planorder(i)$ allows to index the actions of the high-level plan, which is useful to express timing constraints between actions.
  \item Initially, no action has occurred, therefore:
    \[
      \bat_0  = \bat_{\taplatform}^0 \cup \{ \neg \occ(a) \wedge \neg \planorder(i) \}
    \]
\end{itemize}

As the initial situation is completely determined, we assume in the following that $w$ is some world with $w \models \bat$.
With this \ac{BAT}, we can define a program $\delta \eqdef (\sigma \| \delta_M)$ that executes both the high-level plan $\sigma$ and the platform program $\delta_M$ corresponding to the \ac{TA}, as defined in \autoref{sec:tesg-ta}.

Based on this \ac{BAT}, we can define our constraint language, which consists of three types of constraints:
\begin{enumerate}
  \item Absolute timing constraints for the $i$th action; the action must occur within a certain interval $I$ after the start of the plan:
    \[
      \absconstraint(i,I) \eqdef \fut{I} \PlanOrder(i)
    \]
    We denote the set of absolute timing constraints as $\absconstraints$.
    We also write $\absconstraint(i) \eqdef I_i$ for the interval $I_i$ of the constraint $\absconstraint(i, I_i)$.
  \item Relative timing constraints between the $i$th and $j$th action of the plan, requiring that action $j$ occurs after action $i$ within the interval $I$:
    \begin{align*}
      \relconstraint(i,j,I) \eqdef \textbf{F}\big[
        \PlanOrder(i) \wedge\textbf{F}_I \textit{PlanOrder}(j)
      \big]
    \end{align*}
    We denote the set of all relative timing constraints as \relconstraints.
  \item Constraints that require additional platform actions in so-called \emph{chaining constraints}:\todo{Rename to chain constraint?}
    \begin{align*}
      \chainconstraint(\langle \langle &\beta_1,I_1\rangle,\ldots,\langle \beta_n,I_n\rangle\rangle,\alpha_1,\alpha_2) \eqdef
      \\
                    \glob{}\Big[
        &(\alpha_1\wedge \neg\alpha_1\until{}\alpha_2)                        \\
        &\supset \beta_1 \wedge \neg \alpha_2 \wedge \left(\beta_1 \wedge \neg \alpha_2\right) \until{I_1} \big[\beta_2 \wedge \neg \alpha_2 \wedge \left(\beta_2 \wedge \neg \alpha_2\right) \until{I_2} \left(\cdots \until{I_n} \alpha_2\right)\big]
    \Big]
    \end{align*}
    Here, $\alpha_1$ and $\alpha_2$ are fluent formulas only mentioning $\occ(a_i)$ (where $a_i \in \sigma$ is some action of the plan), each $\beta_i$ is a fluent formula only mentioning locations $L = \{ l_i \}_i$ of $\ta_M$, and each $I_i$ is an interval.
    Intuitively, a chaining constraint $\chainconstraint(\langle \langle \beta_1,I_1\rangle,\ldots,\langle \beta_n,I_n\rangle\rangle,\alpha_1,\alpha_2)$ requires that between every occurrence of $\alpha_1$ and $\alpha_2$, the constraints $\beta_i$ are satisfied subsequently, i.e., at the beginning of the sequence, $\beta_1$ must be satisfied until the system eventually and within interval $I_1$ switches to a state satisfying $\beta_{2}$, and so on.
    This allows requiring certain platform actions matching $\beta_1, \ldots, \beta_n$ between two plan actions matching $\alpha_1$ and $\alpha_2$.
    As not every $\beta_i$ must necessitate a change in the platform state, we assume that \taplatform contains $\varepsilon$ transitions, which allow switching from a state satisfying $\beta_i$ to a state satisfying $\beta_{i+1}$ without an actual change in the platform state.
\end{enumerate}


We can now define the transformation problem:
\begin{definition}[Transformation Problem]
  Given an untimed sequence of actions $\sigma = \la a_1, a_2, \ldots, a_n \ra$, a \ac{TA} \taplatform, and a set of of constraints $\Phi = \{ \phi_1, \ldots, \phi_k \}$.
  Let \bat be a \ac{BAT} constructed from \taplatform and $\sigma$ as described above and let  $w \models \bat$.
  The \emph{transformation problem} is to determine a trace $z = \la (a_1, t_1) (a_2, t_2) \cdots (a_l, t_l) \ra$ such that the following holds:
  \begin{enumerate}
    \item The trace $z$ is a valid trace of the parallel execution of the plan and the platform, i.e., $z \in \|(\sigma \| \delta_M)\|_w$.
    \item The trace $z$ satisfies the constraints, i.e., $w, \la\ra, z \models \bigwedge \Phi$.
      \qedhere
  \end{enumerate}
\end{definition}

\begin{example}[Transformation Problem]\label{ex:ta-transformation-problem}
  Consider the following high-level plan:
  \[
    \sigma = \la \sac{\goto(l_1)}, \eac{\goto(l_1)}, \sac{\pick(o_1)}, \eac{\pick(o_1)} \ra
  \]

  In addition to the high-level plan, we are given a self model of the robot.
  Here, we only consider the robot's camera, which is shown again in \autoref{fig:camera-model-example}.

  \begin{figure}[thb]
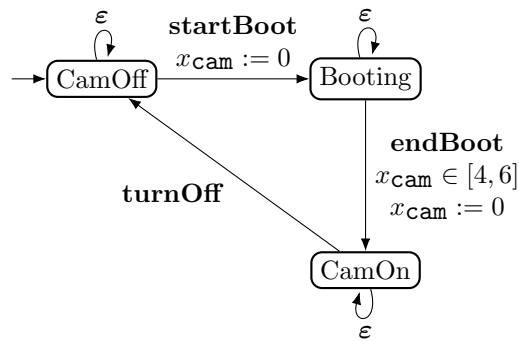

    \centering
    \includestandalone{figures/ex-platform-ta}
    \caption[A \acs*{TA} that models a robot camera.]{A \ac{TA} that models a robot camera.
      If the camera is off, the robot may start booting the camera, which takes at least \SI{4}{\sec} and at most \SI{6}{\sec}.
      If it is on, the robot may instantaneously turn the camera off again.
    }%
    \label{fig:camera-model-example}
  \end{figure}

  \begin{align*}
    \intertext{
      We may know that the first durative action $\goto$ takes between \SI{30}{\sec} and \SI{45}{\sec}, which can be encoded by requiring that action $2$ occurs within the interval $[30, 45]$ after action $1$:
    }
    \gamma_1 &\eqdef \relconstraint(1, 2, [30, 45])
    \intertext{Similarly, \pick may take between \SI{15}{\sec} and \SI{20}{\sec}, corresponding to the following relative timing constraint:}
    \gamma_2 &\eqdef \relconstraint(3, 4, [15, 20])
    \intertext{We also require that the robot starts with \pick immediately after it has arrived:}
    \gamma_3 &\eqdef \relconstraint(2, 3, [0, 0])
    \intertext{
      Regarding platform constraints, we require that the robot's camera is \texttt{off} while it is moving.
      It may turn on its camera in the last \SI{4}{\sec} of a \goto action:
    }
    \gamma_4 &\eqdef \chainconstraint(\langle \langle \mso,[0,\infty)\rangle,\langle \top,[0,4]\rangle\rangle,\sgo,\ego)
    \intertext{Additionally, the camera must be on all the time while the robot is picking up an object:}
    \gamma_5 &\eqdef \chainconstraint(\langle \langle \msr,[0,\infty)\rangle\rangle,\spi,\epi)
  \end{align*}
  To transform the plan $\sigma$, we need to determine the execution time point for each action and we may need to insert additinal platform actions.
  In our case, the following sequence is a realization of the plan that satisfies all constraints:
  \begin{multline*}
    (\sgo, 0), (\sac{\bootcam}, 26), (\ego, 30), \\ (\eac{\bootcam}, 30), (\spi, 30), (\epi, 45)
  \end{multline*}
  The robot starts moving right away.
  \SI{26}{\sec} after the start, it starts booting the camera.
  It finishes the \goto action at time \SI{30}{\sec} and also immediately finishes booting the camera, before it continues picking up the object without further delay.
\end{example}

\section{Plan Encoding}%
\label{sec:plan-encoding}

As a first step of the transformation procedure, we encode the high-level plan $\sigma$ into \iac{TA} \taplan.
The resulting \ac{TA} will accept every timed word that corresponds to the high-level plan augmented with execution time points.
In addition to considering the high-level plan, we also encode the relative timing constraints \relconstraints and the absolute timing constraints \absconstraints into the \ac{TA}.
We do this by inserting appropriate clocks and clock constraints that restrict transitions in the \ac{TA} such that they satisfy the constraints.
Therefore, each timed word accepted by \taplan corresponds to a timed execution of the high-level plan that satisfies all timing constraints.

\begin{definition}[Plan \ac{TA}]
  Given a high-level plan $\sigma = \la a_1, \ldots, a_n \ra$, we construct the corresponding \ac{TA} $\taplan = (L, l_0, L_F, \taalph, X, I, E)$ as follows:
\begin{enumerate}
  \item There is one location $l_i$ for each action $a_i$ of the plan: $L = \{ l_0, \ldots, l_n \}$
  \item The initial location is $l_0$.
  \item The only final location is the last location, i.e., $L_F = \{ l_n \}$.
  \item The alphabet consists of the actions of the plan: $\taalph = \{ a_1, \ldots, a_n \}$
  \item There is one clock \clockabs for absolute timing constraints and one clock $\clockrel_{i,j}$ for each pair of actions to track relative timing constraints:
    \[
      X = \{ \clockabs \} \cup \bigcup_{1 \leq i < j \leq n} \clockrel_{i, j}
    \]
  \item There are no location invariants: $I(l) = \top$ for each $l \in L$.
  \item There is one switch for each plan action $a_i$ that switches from $l_i$ to $l_{i+1}$:
    \[
      E = \bigcup_{1 \leq i \leq n + 1} (l_{i-1}, a_i, \Psi_i, X_i, l_i)
    \]
    where
    \begin{itemize}
      \item The clock constraint is a conjunction of the absolute clock constraint for action $a_i$ and all relative clock constraints mentioning $a_i$ as endpoint:%
        \footnote{Recall that an interval can be written as clock constraint, e.g., $x \in (a, b]$ becomes $x > a \wedge x \leq b$.}
        \[
          \Psi_i = \clockabs \in \absconstraint(i) \wedge \bigwedge_{\relconstraint(k, i, I) \in \relconstraints} \clockrel_{k, i} \in I
        \]
      \item The switch resets all clocks that track constraints with $a_i$ as starting point:
        \[
          X_i = \bigcup_{i < j \leq n} \clockrel_{i, j}
          \qedhere
        \]
    \end{itemize}
\end{enumerate}
\end{definition}

We demonstrate the construction on the running example:
\begin{example}[Plan \ac{TA}]
  \begin{figure}[ht]
    \centering
    \includestandalone{figures/ex-plan-ta}
    \caption[The plan encoding.]{The \ac{TA} \taplan that encodes the plan $\sigma$ and all timing constraints \relconstraints and \absconstraints from \autoref{ex:ta-transformation-problem}.}
    \label{fig:ta-encode-plans}
  \end{figure}
  \autoref{fig:ta-encode-plans} shows the encoding of the plan from \autoref{ex:ta-transformation-problem}.
\end{example}

It follows from construction that \taplan accepts a timed word if and only if the timed word is a trace of the program that satisfies all timing constraints in \relconstraints and \absconstraints:
\begin{theoremE}\label{thm:plan-encoding}
  \[
    z \in \lang(\taplan) \text{ iff } z \in \big\| \sigma \big\|_w \text{ and } w, \la\ra, z \models \relconstraints \wedge \absconstraints
  \]
\end{theoremE}
\begin{proofE}
  \textbf{$\Rightarrow$:}
  \\
  Let $z \in \lang(\taplan)$ and let $r = (l_0, \clockvaluation_0) \taltstrans{a_1}{t_1} (l_1, \clockvaluation_1) \taltstrans{a_2}{t_2} \ldots \taltstrans{a_n}{t_n} (l_n, \clockvaluation_n) \in \accfinruns(\taplan)$ be the corresponding run on $\taplan$ with $\tw(r) = z$.
  First, it directly follows from the construction of $\taplan$ that $z \in \lang(\taplan)$ implies $z \in \big\| P \big\|_w$: For every location $l_i$ of $\taplan$, the only possible transition is to $l_{i+1}$.
  Furthermore, as $l_0$ is the initial location and \fin the only final location, every timed word $z \in \lang(\taplan)$ must start with $a_1$ and end with $a_n$.
  It remains to be shown that $w, \la\ra, z \models \relconstraints \wedge \absconstraints$.
  \begin{enumerate}
    \item Let $\relconstraint(i, j, I) = \finally{} [\planorder(i) \wedge \finally{I} \planorder(j)] \in \relconstraints$.
      Note that there is a unique prefix $z_i = (a_1, 0) \cdots (a_i, t_i) $ of $z$ such that $w, z_i \models \planorder(i)$.
      Let $z = z_i \cdot z'$ and let $r =(l_0, \clockvaluation_0) \taltstrans{a_1}{t_1} (l_1, \clockvaluation_1) \taltstrans{a_2}{t_2} \ldots \taltstrans{a_i}{t_i} (l_i, \clockvaluation_i)$ the corresponding prefix of $r$.
      By definition of $\taplan$, $\clockvaluation_i(x_{i,j}) = 0$.
      Similarly, there is a unique $z_j = (a_1, 0) \cdots (a_j, t_j)$ such that $z = z_j \cdot z''$ and $w, z_j, z'' \models \planorder(j)$.
      It remains to be shown that $\sum_{k=i+1}^j \in I$.
      By definition of $\taplan$, the switch from $l_{j-1}$ to $l_j$ has the guard $x_{i,j} \in I$ and therefore, $\clockvaluation_j(x_{i,j}) \in I$.
      As $\clockvaluation_i(x_{i,j}) = 0$ and because $x_{i,j}$ is not reset with any other transition, it follows that $\sum_{k=i+1}^j \in I$.
      Therefore, $w, z_i, z' \models \finally{I} \planorder(j)$ and hence, $w, \la\ra, z \models \relconstraint(i, j, I)$.
    \item Let $\absconstraint(i, I) = \finally{I} \planorder(i) \in \absconstraints$.
      As before, there is a unique prefix $z_i = (a_1, 0) \cdots (a_i, t_i) $ of $z$ such that $w, z_i \models \planorder(i)$.
      Note that the switch for action $a_i$ has a clock constraint $\clockabs \in I$ and $\clockabs$ is never reset in any $\taplan$ switch.
      Therefore, $\sum_{k=1}^i t_k \in I$.
      It directly follows that $w, \la\ra, z \models \absconstraint(i, I)$.
  \end{enumerate}
  \textbf{$\Leftarrow$:}
  \\
  Let $z = (a_1, t_1) \cdots (a_n, t_n) \in \big\| P \big\|_w$ and $w, \la\ra, z \models \relconstraints \wedge \absconstraints$ and let $z_i = (a_1, t_1) \cdots (a_i, t_i)$ the prefix of $z$ with length $i$.
  We show by induction on $i$ that there is a run $r = (l_0, \clockvaluation_0) \taltstrans{a_1}{t_1} (l_1, \clockvaluation_1) \taltstrans{a_2}{t_2} \ldots \taltstrans{a_i}{t_i} (l_i, \clockvaluation_i) \in \accfinruns(\taplan)$
  \\
  \textbf{Base case.}
  For $i = 0$, it follows immediately that $\la\ra \in \accfinruns(\taplan)$.
  \\
  \textbf{Induction step.}
  By induction, $(l_0, \clockvaluation_0) \taltstrans{a_1}{t_1} (l_1, \clockvaluation_1) \taltstrans{a_2}{t_2} \ldots \taltstrans{a_i}{t_i} (l_i, \clockvaluation_i) \in \accfinruns(\taplan)$.
  By definition of $\taplan$, there is a switch $(l_i, a_{i+1}, \Psi_{i+1}, X_{i+1}, l_{i+1}) \in E$.
  As the invariant of $l_{i+1}$ is $I(l_{i+1}) = \top$, it is always satisfied.
  It remains to be shown that $\clockvaluation_i \models \Psi_{i+1}$.
  By definition, $\Psi_i = \clockabs \in \absconstraint(i) \wedge \bigwedge_{\relconstraint(k, i, I) \in \relconstraints} \clockrel_{k, i} \in I$.
  \begin{itemize}
    \item As $w, \la\ra, z \models \absconstraint(i)$, it follows that $\sum_{k=1}^{i} \in I$.
      Furthermore, the clock $\clockabs$ is never reset.
      Thus, $\clockvaluation_i \models \absconstraint(i)$.
    \item Let $\relconstraint(k, i, I) = \finally{}[\planorder(k) \wedge \finally{I} \planorder(i)] \in \relconstraints$.
      As $w, \la\ra \models \relconstraint(k, i, I)$ and because $w, z' \models \planorder(k)$ iff $z' = z_k$, it immediately follows that $\sum_{j=k+1}^{i} \in I$.
      By definition of $\taplan$, $x_{k,i}$ is only reset in $l_k$.
      Therefore, $\clockvaluation_i(x_{k,i}) = \sum_{j=k+1}^{i} \in I$.
  \end{itemize}
  It follows that $\clockvaluation_i \models \Psi_{i+1}$.
  \\
  Finally, $l_i$ is accepting iff $a_i$ is the last action of the plan.
  Therefore, $(l_0, \clockvaluation_0) \taltstrans{a_1}{t_1} (l_1, \clockvaluation_1) \taltstrans{a_2}{t_2} \ldots \taltstrans{a_n}{t_n} (l_n, \clockvaluation_n) \in \accfinruns(\taplan)$ is accepting, hence $z \in \lang(\taplan)$.
\end{proofE}

Therefore, we can encode a high-level plan into \iac{TA} \taplan such that \taplan only accepts words that correspond to an execution of the plan that satisfies all relative timing constraints \relconstraints and absolute timing constraints \absconstraints.
However, we have not considered the platform constraints \chainconstraints that require additional platform actions.
In the next section, we will extend the encoding to also consider those platform constraints.

\section{Platform Encoding}%
\label{sec:platform-encoding}

So far, we have encoded the high-level plan $\sigma$ into \iac{TA} \taplan that accepts exactly those words that correspond to an execution of the plan that satisfies all timing constraints.
However, we have not yet considered the robot self model \taplatform and the corresponding  constraints \chainconstraints.
In contrast to \relconstraints and \absconstraints, these are not merely timing constraints, but may require additional platform actions.

\begin{algorithm}[tb]
  \caption[{The algorithm \textsc{TransformPlan}.}]{The algorithm \textsc{TransformPlan} which converts a plan and a platform model into a product automaton that satisfies all constraints.}
  \label{alg:transform-plan}
  \begin{algorithmic}[1]
  \Procedure{TransformPlan}{$P, \taplatform, \relconstraints, \absconstraints, \chainconstraints$}
  \State $\taplan \gets \Call{EncodePlan}{P, \relconstraints, \absconstraints}$
  \State $\ta \gets \taplan \times \taplatform$
  \ForAll{$\gamma \in \chainconstraints$} \label{alg:transform-plan:chain-loop}
  \ForAll{$(s, e) \in \Call{GetActivations}{\gamma, P}$}
  \State $\ta \gets \Call{EnforceUC}{\ta, s, e, \gamma}$
  \EndFor
  \EndFor
  \EndProcedure
  \Function{EnforceUC}{$\ta, s, e, \gamma = \chainconstraint(\la \beta_1, I_1 \ra, \ldots, \la \beta_n, I_n \ra, \alpha_1, \alpha_2)$} \label{alg:transform-plan:enforce-uc}
  \State $\tacontext \gets \Call{GetSubTA}{\ta, s, e}$
  \ForAll{$i \in \{ 1, \ldots, n \}$}
  \State $S_i \gets \Call{Copy}{\tacontext}$ \Comment{Each $l$ is renamed to $l^{(i)}$}
  \ForAll{$l \in \mi{Locations}(S_i)$}
  \If{$l \not \in \beta_i$}
  \Call{Delete}{$l,S_i$} \label{alg:transform-plan:delete}
  \EndIf
  \EndFor
  \EndFor
  \State $\tachain \gets \Call{Combine}{\tacontext, S_1, \ldots, S_n, I_1, \ldots, I_n, \gamma}$ \Comment{See \autoref{alg:transform-plan:helpers}}
  \State $\ta \gets \Call{Replace}{\ta, \tacontext, \tachain, I_n}$
  \State \Return $\ta$
  \EndFunction
\end{algorithmic}
\end{algorithm}

In the following, we will extend the construction to incorporate the robot self model and the corresponding constraints.
Before we  can describe the procedure, we must introduce some auxiliary functions:
\begin{itemize}
  \item For a given constraint $\gamma = \chainconstraint(B, \alpha_1, \alpha_2)$, $\textsc{GetActivations}(\gamma, \sigma)$ returns a set of pairs $(s, e)$ such that the plan action $a_s$ with index $s$ satisfies $\alpha_1$, the plan action $a_e$ with index $e$ satisfies $\alpha_2$, and no action between $s$ and $e$ satisfies $\alpha_1$.
  \item $\textsc{GetSubTA}(\ta, s, e)$ returns the \ac{TA} that only contains locations starting with the plan action with index $s$ and ending with the plan action with index $e$ (exclusive).%
    \footnote{
      Technically, $\textsc{GetSubTA}(\ta, s, e)$ is not necessarily a \ac{TA} because it may not have an initial location.
      As we will later recombine this automaton with the original automaton, we ignore this detail.}
  \item $\textsc{Copy}$ copies a \ac{TA} (or a part of a \ac{TA}) and renames each location so all locations have a unique name.
  \item The union $\ta_1 \cup \ta_2$ is a \ac{TA} that contains all locations, invariants, and switches of both $\ta_1$ and $\ta_2$, assuming that the two \acp{TA} do not share any location names.
    Note that the resulting automaton has two disconnected components.
    Formally, for two \acp{TA} $\ta_1 = (L_1, l_0, \Sigma_1, X_1, I_1, E_1)$ and $\ta_2 = (L_2, l_0, \Sigma_2, X_2, I_2, E_2)$, the union $\ta = \ta_1 \cup \ta_2$ is the \ac{TA} $\ta = (L_1 \cup L_2, l_0, \Sigma_1 \cup \Sigma_2, X_1 \cup X_2, I_1 \cup I_2, E_1 \cup E_2)$.
  \item The difference $\ta_1 \setminus \ta_2$ is a \ac{TA} that is like $\ta_1$ except that all locations of $\ta_2$ and the corresponding switches and invariants are removed.
    Formally, for two \acp{TA} $\ta_1 = (L_1, l_0, \Sigma_1, X_1, I_1, E_1)$ and $\ta_2 = (L_2, l_0, \Sigma_2, X_2, I_2, E_2)$, the difference $\ta = \ta_1 \setminus \ta_2$ is the \ac{TA} $\ta = (L_1 \setminus L_2, l_0, \Sigma_1, X_1, I_1 \setminus I_2, E_1 \setminus \{ (l, a, g, Y, l') \mid l \in L_2 \vee l' \in L_2 \})$.
\end{itemize}

\begin{algorithm}[tb]
  \caption[Auxiliary functions for \autoref{alg:transform-plan}]{
    Auxiliary functions for \autoref{alg:transform-plan}.
    The function \textsc{Combine} combines the automata $S_1, \ldots, S_n$ into one automaton while enforcing the timing constraints specified by the intervals $I_1, \ldots, I_n$.
    The function \textsc{Replace} replaces the activation context \tacontext in the original $\ta$ by the newly constructed $\tachain$.
  }
  \label{alg:transform-plan:helpers}
  \begin{algorithmic}[1]
  \Function{Combine}{$\tacontext, S_1, \ldots, S_n, I_1, \ldots, I_n, \gamma$}
  \State $\tachain \gets \cup_i S_i$
  \ForAll{$i < n$}
  \ForAll{$l \in L_{S_i}$} 
  \ForAll{$l' \in L_{S_{i+1}}$} 
  \ForAll{$(l, a, g, X, l') \in E_{\tacontext}$}
  \State Add $(l, a, g \wedge x_\gamma \in I_{i}, X \cup \{ x_\gamma \}, l')$ to $E_{\tachain}$ \label{alg:transform-plan:guard}
  \EndFor
  \EndFor
  \EndFor
  \EndFor
  \EndFunction
  \Function{Replace}{$\ta, \tacontext, \tachain, I_1$}
  \State $\ta \gets \ta \setminus \tacontext$
  \For{$(l, a, g, X, l') \in E_\ta$ with $l' \in L_{\tacontext}$} \Comment{Incoming transition of \tacontext}
  \State Add $(l, a, g, X \cup \{ x_\gamma \}, l')$ to $E_\ta$
  \EndFor
  \For{$(l, a, g, X, l') \in E_\ta$ with $l \in L_{\tacontext}$} \Comment{Outgoing transition of \tacontext}
  \State Add $(l, a, g \wedge x_\gamma \in I_n, X, l')$ to $E_\ta$ \label{alg:transform-plan:context-outgoing}
  \EndFor
  \EndFunction
\end{algorithmic}
\end{algorithm}

The algorithm is shown in \autoref{alg:transform-plan}.
We start with the plan encoding \taplan and construct a product automaton $\ta = \taplan \times \taplatform$ that combines the high-level plan with the platform automaton.
This product automaton allows us to insert arbitrary platform actions while still executing the high-level plan.
Next, for each constraint $\gamma_i = \chainconstraint(\la \beta_1, I_1 \ra, \ldots, \la \beta_n, I_n \ra, \alpha_1, \alpha_2) \in \chainconstraints$, we compute its \emph{activation scope}, i.e., the plan actions that match $\alpha_1$ and $\alpha_2$ correspondingly.
We call  the corresponding part of the \ac{TA} the \emph{context} \tacontext of the activation.
For each such activation, we must modify \tacontext so $\gamma_i$ is guaranteed to be satisfied.
This is done in the function \textsc{EnforceUc} (\autoref{alg:transform-plan}, line~\ref{alg:transform-plan:enforce-uc}), which works as follows:
For each $\la \beta_j, I_j \ra$, we copy the states and transitions that are within the activation scope into a new sub-automaton $S_j$.
In the next step, we remove all locations of $S_j$ that do not satisfy $\beta_j$.
Therefore, we obtain $n$ sub-automata $S_1, \ldots, S_n$, where each $S_j$ tracks the satisfaction of $\beta_j$.
Next, the function \textsc{Combine} in \autoref{alg:transform-plan:helpers} combines $S_1, \ldots, S_n$ such that it is possible to switch from $S_i$ to $S_{i+1}$ if there is a switch with the same action between the corresponding locations in the original automaton (\autoref{alg:transform-plan:helpers}, line~\ref{alg:transform-plan:guard}) .
Hence, by construction, every accepted word by the resulting automaton must transition through locations that subsequently satisfy $\beta_1, \ldots \beta_n$.
Finally, to take care of the timing constraints $I_1, \ldots, I_n$ of $\gamma_i$, we introduce a new clock $x_\gamma$ that is reset between each transition from $S_j$ to $S_{j+1}$ and where each incoming transition of $S_{j+1}$ has an additional clock constraint $x_\gamma \in I_j$ that guarantees that the system stays in the states specified by $\beta_j$ for some duration restricted by $I_j$.
By replacing the original context \tacontext by the newly constructed automaton, we obtain a \ac{TA} that only accepts those words that satisfy the $\gamma_i$ within the activation scope.
The corresponding function \textsc{Replace} is shown in \autoref{alg:traverse-tree-helpers}.
After we iteratively apply this construction for every activation of every constraints $\chainconstraint \in \chainconstraints$, we obtain an \ac{TA} that only accepts words that satisfy all constraints.

\begin{example}[Platform Encoding]
 \begin{sidewaysfigure}
   \thisfloatpagestyle{plain}
   \centering
   \includestandalone[width=\textwidth]{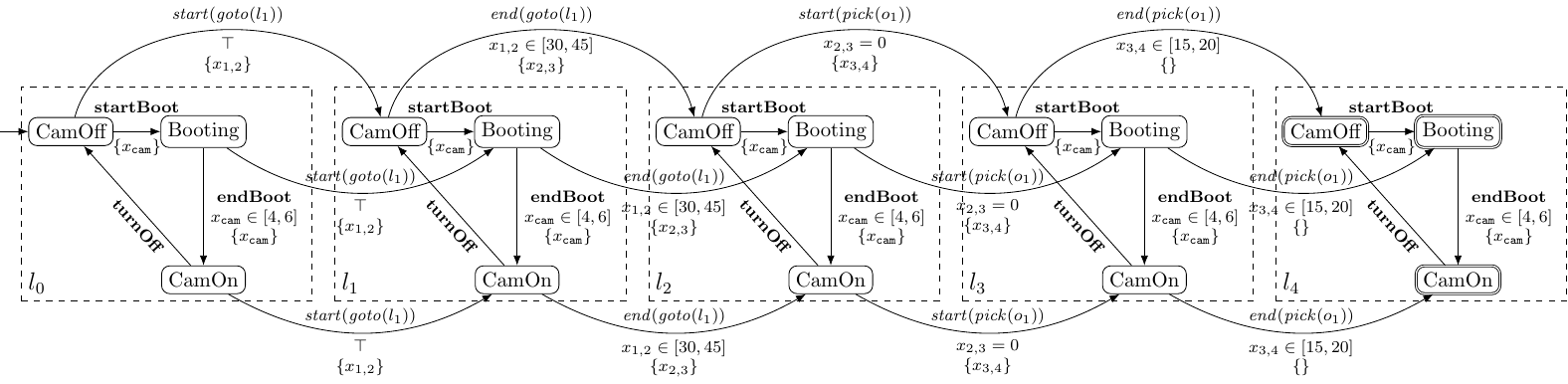}
   \caption[The product automaton.]{The product automaton $\taplan \times \taplatform$. The $\varepsilon$ transitions from \taplatform are omitted.}
   \label{fig:ex-product-automaton}
 \end{sidewaysfigure}
 \autoref{fig:ex-product-automaton} shows the product automaton $\taplan \times \taplatform$ before any of the chaining constraints \chainconstraints have been considered.
 By construction, it allows every timed word that is also accepted by the plan automaton \taplan and additionally allows any platform actions from \taplatform.
 \begin{sidewaysfigure}
   \centering
   \includestandalone[width=\textwidth]{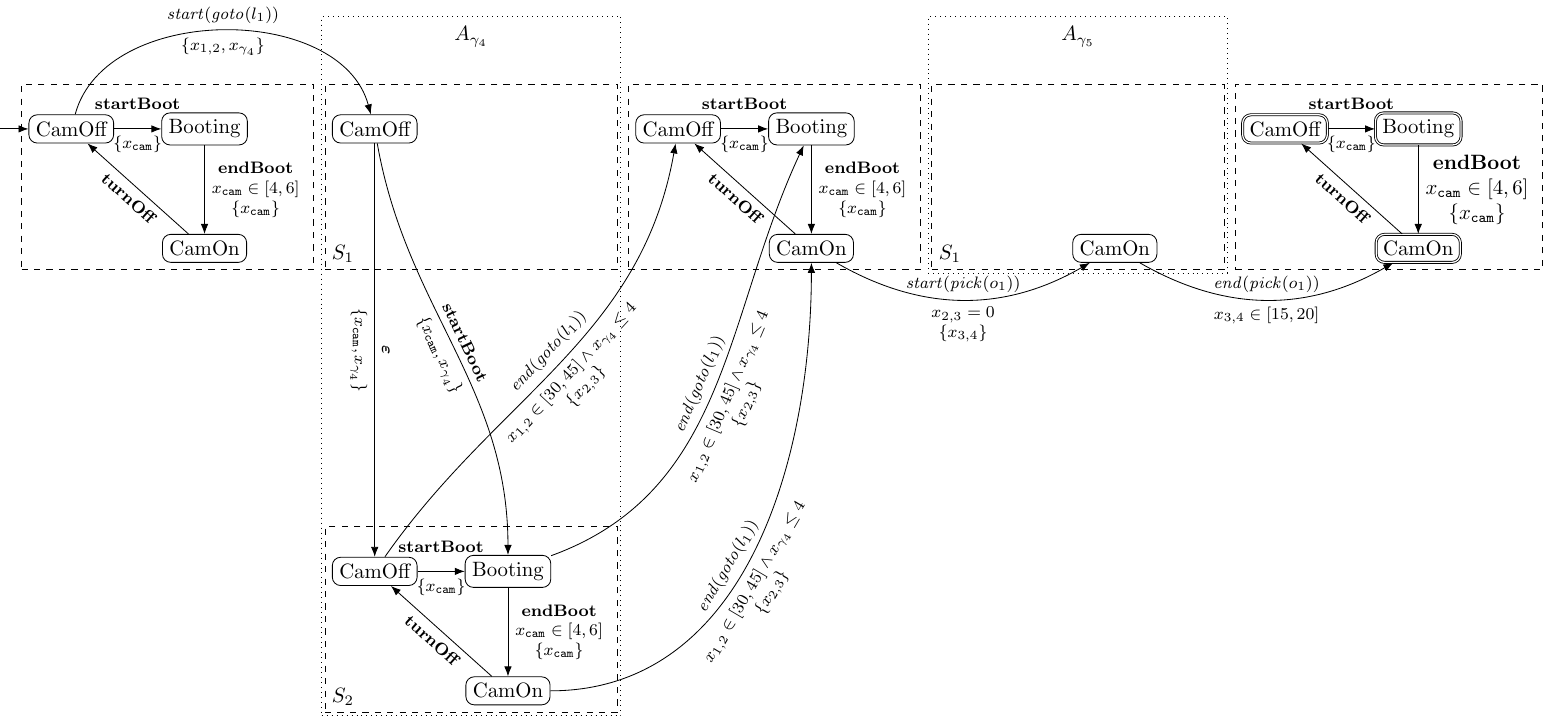}
   \caption[Constraint encoding.]{Encoding of the constraints $\gamma_4$ and $\gamma_5$.}
   \label{fig:ex-platform-encoding}
 \end{sidewaysfigure}
 Starting from the product automaton, \autoref{alg:transform-plan} restricts transitions by removing locations and adding clock constraints such that the resulting automaton only accepts words that satisfy the constraints $\gamma_4$ and $\gamma_5$.
 The result is shown in \autoref{fig:ex-platform-encoding}.

 For $\gamma_4 = \chainconstraint(\langle \langle \mso,[0,\infty)\rangle,\langle \top,[0,4]\rangle\rangle,\sgo,\ego)$, the activation scope are the locations starting with the incoming action \sgo and ending with the outgoing action \ego.
 The activation scope is replaced by the new automaton $\ta_{\gamma_4}$.
 As $\gamma_4$ contains two state constraints $\langle \mso,[0,\infty)\rangle$ and $\langle \top,[0,4]\rangle$, $\ta_{\gamma_4}$ consists of the two sub-automata $S_1$  and $S_2$, as shown in \autoref{fig:ex-platform-encoding}.
 The automaton $S_1$ enforces the state constraint $\mso$ and therefore consists of the single location $\camoff$.
 As the second state constraint $\top$ allows every location, $S_2$ contains all locations of the original \taplatform.
 Finally, the timing constraints are enforced with a new clock $x_{\gamma_4}$, which is reset on the incoming transitions of $S_1$ and $S_2$.
 While the first state constraint does not have any timing constraints, the second state  constraint states that the automaton must stay in any location of $S_2$ for at most $4$ time units, which is enforced by the clock constraint $x_{\gamma_4} \leq 4$ on each outgoing transition of $S_2$.

 For $\gamma_5 = \chainconstraint(\langle \langle \msr,[0,\infty)\rangle\rangle,\spi,\epi)$, the construction works similarly.
 As there is only a single state constraint in the chain, $\ta_{\gamma_5}$ also consists of a single sub-automaton $S_1$, which must match $\msr$ and therefore consists of the single location \camon.
\end{example}

We can show that the resulting \ac{TA} allows only those traces that correspond to executions of the high-level plan and the platform automaton that satisfy all constraints:%
\footnote{
  Recall that we encode \acp{TA} transitions in \tesg by a sequence of \emph{switches}.
  Given such a sequence $z$, $\labeltrace(z)$ is the corresponding sequence of action labels (\autoref{def:label-trace}).
}
\begin{theoremE}\label{thm:platform-encoding}
  \[
    \rho \in \lang(\taenc) \text{ iff } \rho = \labeltrace(z) \text{ for some } z \in \big\| (\sigma \| \delta_M) \|_w \text{ with } z \models \relconstraints \wedge \absconstraints \wedge \chainconstraints
  \]
\end{theoremE}
\begin{proofE}
  ~\\
  \textbf{ $\Rightarrow$: }
  \\
  Assume $\chainconstraints = \{ \gamma_1, \ldots, \gamma_n \}$.
  Let $\taenc^{(i)}$ be the \ac{TA} constructed in \textsc{TransformPlan} after iterating over the first $i$ constraints $\gamma_i \in \chainconstraints$ (line \ref{alg:transform-plan:chain-loop}).
  We show the following by induction on the number of constraints $n$:
  If $\rho \in \lang(\taenc^{(i)})$, then there is a $z \in \big\| (P \| \delta_M) \|_w$ such that $\rho = \labeltrace(z)$ and $w, \la\ra, z \models \relconstraints \wedge \absconstraints \wedge \gamma_1 \wedge \ldots \wedge \gamma_i$.
  \\
  \textbf{Base case.}
  Let $i = 0$. 
  By construction, $\taenc^{(0)} = \taplan \times \ta_M$.
  Therefore, $\rho \in \lang(\taplan \times \ta_M)$.
  Notice that $\rho$ consists of interleaved symbols from $\taplan$ and $\ta_M$.
  As $\taplan$ and $\ta_M$ do not share any symbol or clock names\todo{Require this}, it is clear that there is some $z \in \| (P \| \delta_M) \|$ such that $\rho = \labeltrace(z)$.
  Also, with \autoref{thm:plan-encoding}, $w, \la\ra, z \models \relconstraints \wedge \absconstraints$.
  \\
  \textbf{Induction step.}
  \\
  Let $\rho \in \lang(\taenc^{(i)})$ and let $r = (l_0, \clockvaluation_0) \taltstrans{a_1}{t_1} (l_1, \clockvaluation_1) \taltstrans{a_2}{t_2} \ldots \taltstrans{a_k}{t_k} (l_k, \clockvaluation_k) \in \accfinruns(\taenc^{(i)})$ be the corresponding run with $\tw(r) = \rho$.
  First, notice that $\rho \in \lang(\taenc^{(i-1)})$:
  $\taenc^{(i)}$ is constructed from $\taenc^{(i-1)}$ by replacing $\tacontext$ with a new sub-automaton.
  In the sub-automaton, each added $S_j$ is a copy of $\tacontext$ with some locations removed.
  Therefore, each transition within some $S_j$ is also possible in \tacontext and therefore in $\taenc^{(i-1)}$ (which contains \tacontext).
  Furthermore, each switch added by \textsc{Combine} and \textsc{Replace} is like a switch of $\tacontext$ but with additional clock constraints for and resets of $x_\gamma$.
  Hence, each of those modifications only restrict $\taenc^{(i)}$ in comparison to $\taenc^{(i-1)}$, therefore $\rho \in \lang(\taenc^{(i-1)})$.
  By induction, there is a $z \in \big\| (P \| \delta_M) \|_w$ such that $\rho = \labeltrace(z)$ and $w, \la\ra, z \models \relconstraints \wedge \absconstraints \wedge \gamma_1 \wedge \ldots \wedge \gamma_{i-1}$.

  It remains to be shown that $w, \la\ra, z \models \gamma_i$:
  Assume $\gamma_i = \chainconstraint(\la \la \beta_1, I_1 \ra, \ldots, \la \beta_m, I_m \ra\ra)$ and $z = z' \cdot z''$ such that $w, z', z'' \models \alpha_1 \wedge \neg \alpha_1 \until{} \alpha_2$.
  By construction of $\taenc^{(i)}$, each run must pass through $S_1, \ldots, S_m$ as constructed in \textsc{EnforceUC}.
  We can split $r$ according to $z'$ and $z''$, i.e.,
  \begin{multline*}
    r = (l_0, \clockvaluation_0) \taltstrans{\sigma_{1}}{t_{1}} \ldots \taltstrans{\sigma_s}{t_s} 
    (l_{1,1}, \clockvaluation_{1,1}) \taltstrans{\sigma_{1,1}}{t_{1,1}}
    \ldots \taltstrans{\sigma_{1,k_1}}{t_{1,k_1}} \\
    (l_{1,k_1}, \clockvaluation_{1,k_1}) \taltstrans{\sigma_{2,1}}{t_{2,1}} \ldots \taltstrans{\sigma_{m,k_m}}{t_{m,k_m}} (l_{m,k_m}, \clockvaluation_{m, k_m}) \taltstrans{\sigma_e}{t_e} (l_e, \clockvaluation_e) \taltstrans{}{} \ldots
  \end{multline*}
  such that $\labeltrace((\sigma_1, t_1) \cdots (\sigma_s, t_s)) = z'$, $\labeltrace((\sigma_{1,1}, t_{1,1}) \cdots) = z''$, each $l_{i,j}$ is a location of $S_i$, and $l_{1,1}$ is the start and $l_e$ the end of the activation.
  Therefore, $l_{1,1}$ satisfies $\alpha_1$ and $l_e$ satisfies $\alpha_2$.
  Clearly, for each $i \leq m$ and each $j \leq k_i$, the location $l_{i,j}$ matches $\beta_i$ (otherwise, the location would have been deleted in line \ref{alg:transform-plan:delete}).
  Furthermore, $l_{i,j}$ may not match $\alpha_2$, as \textsc{GetActivations} returns the smallest scope that does not match $\alpha_2$ except in the endpoint $l_e$.
  Also, note that each $(\sigma_{i,j}, t_{i,j})$ corresponds to a timed action $(a_{i,j}, t_{i,j})$.
  Hence, we can write $z''$ as follows:
  \[
    z'' = (a_{1,1}, t_{1,1}) \cdot \ldots \cdot (a_{1,k_1}, t_{1, k_1}) \cdot \ldots \cdot (a_m, t_{m, k_m}) \cdot (a_e, t_e) \cdot \ldots
  \]
  It directly follows for each $i \leq m$ and $j \leq k_i$  that $w, z' \cdot \la (a_{1,1}, t_{1,1}) \cdot \ldots \cdot (a_{i,j}, t_{i, j}) \ra \models \beta_i \wedge \neg \alpha_2$.
  Next, notice that by construction, $x_\gamma$ is reset when entering each $S_i$ (i.e., on action $(a_{i,1}, t_{i,1})$).
  Hence, $\clockvaluation_{i,k_i}(x_\gamma) = \sum_{j=2}^{k_i} t_{i,j}$.
  Also, the switch from $S_i$ to $S_{i+1}$ has the guard $x_\gamma \in I_{i}$ (Algorithm~\ref{alg:transform-plan:helpers}, line~\ref{alg:transform-plan:guard}).
  Therefore, $\clockvaluation_{i,k_i}(x_\gamma) + t_{i+1,1} = \sum_{j=1}^{k_i} t_{i,j} + t_{i+1,1} \in I_i$.
  It follows that for each $j < m$, $w, z' \cdot \la (a_{1,1}, t_{1,1}) \cdot \ldots \cdot (a_{j,1}, t_{j,1}) \ra, \la  (a_{j,2}, t_{j,2}), \cdots, (a_{m,k_m}, t_{m,k_m}) \ra \models (\beta_j \wedge \neg \alpha_2) \until{I_j} (\beta_{j+1} \wedge \neg \alpha_2)$.
  In a similar way, each transition leaving $S_m$ has a clock constraint $x_\gamma \in I_m$ (Algorithm~\ref{alg:transform-plan:helpers}, line~\ref{alg:transform-plan:context-outgoing}) and so $\clockvaluation_{m,k_m} + t_e = \sum_{j=1}^{k_m} t_{m,j} + t_e \in I_m$.
  In conclusion, $w, z', z'' \models \beta_1 \wedge \neg \alpha_2 \wedge \left(\beta_1 \wedge \neg \alpha_2\right) \until{I_1} \big[\beta_2 \wedge \neg \alpha_2 \wedge \left(\beta_2 \wedge \neg \alpha_2\right) \until{I_2} \left(\cdots \until{I_n} \alpha_2\right)\big]$ for each $z', z''$ with $z = z' \cdot z''$ and $w, z', z'' \models \alpha_1 \wedge \neg\alpha_1\until{}\alpha_2$.
  Therefore, $w, \la\ra, z \models \gamma_i$.

  \medskip

  \noindent \textbf{$\Leftarrow$:}
  \\
  Assume $\chainconstraints = \{ \gamma_1, \ldots, \gamma_n \}$.
  Let $\taenc^{(i)}$ be the \ac{TA} constructed in \textsc{TransformPlan} after iterating over the first $i$ constraints $\gamma_i \in \chainconstraints$ (line \ref{alg:transform-plan:chain-loop}).
  We show the following by induction on the number of constraints $n$:
  If $z \in \big\| (P \| \delta_M) \|_w$ with $w, \la\ra, z \models \relconstraints \wedge \absconstraints \wedge \gamma_1 \wedge \ldots \wedge \gamma_i$, then there is $\rho \in \lang(\taenc^{(i)})$ such that $\rho = \labeltrace(z)$.
  \\
  \textbf{Base case.}
  Let $i = 0$. 
  By construction, $\taenc^{(0)} = \taplan \times \ta_M$.
  Therefore, $\rho \in \lang(\taplan \times \ta_M)$.
  As $z \in \| (P \| \delta_M) \|$ and $\taplan$ and $\ta_M$ do not share any symbols or clock names, it directly follows that $\labeltrace(z) \in \lang(\taplan \times \ta_M)$.
  \\
  \textbf{Induction step.}
  Let $z \in \big\| (P \| \delta_M) \|_w$ such that $w, \la\ra, z \models \relconstraints \wedge \absconstraints \wedge \gamma_1 \wedge \ldots \wedge \gamma_i$.
  Note that $w, \la\ra, z \models \relconstraints \wedge \absconstraints \wedge \gamma_1 \wedge \ldots \wedge \gamma_{i-1}$.
  Therefore, by induction, there is a $\rho \in \lang(\taenc^{(i-1)})$ such that $\rho \in \labeltrace(z)$.
  Let $r = (l_0, \clockvaluation_0) \taltstrans{a_1}{t_1} (l_1, \clockvaluation_1) \taltstrans{a_2}{t_2} \ldots \taltstrans{a_k}{t_k} (l_k, \clockvaluation_k) \in \accfinruns(\taenc^{(i)})$ be the corresponding run with $\tw(r) = \rho$.
  Assume $\gamma_i = \chainconstraint(\la \la \beta_1, I_1 \ra, \ldots, \la \beta_m, I_m \ra\ra)$ and $z = z' \cdot z''$ such that $w, z', z'' \models \alpha_1 \wedge \neg \alpha_1 \until{} \alpha_2$.
  As $w, \la\ra, z \models \gamma_i$, it directly follows that $w, z', z'' \models \beta_1 \wedge \neg \alpha_2 \wedge \left(\beta_1 \wedge \neg \alpha_2\right) \until{I_1} \big[\beta_2 \wedge \neg \alpha_2 \wedge \left(\beta_2 \wedge \neg \alpha_2\right) \until{I_2} \left(\cdots \until{I_n} \alpha_2\right)\big]$.
  We can write $z''$ as
  \[
    z'' = (a_{1,1}, t_{1,1}) \cdot \ldots \cdot (a_{1,k_1}, t_{1, k_1}) \cdot \ldots \cdot (a_m, t_{m, k_m}) \cdot (a_e, t_e) \cdot \ldots
  \]
  such that $w, z \cdot \la (a_{1,1}, t_{1,1}) \cdot \ldots \cdot (a_{i,j}, t_{i, j}) \ra \models \beta_i \wedge \neg \alpha_2$ for each $i \leq m$ and $j \leq k_i$.
  We can split $r$ accordingly, i.e.,
  \begin{multline*}
    r = (l_0, \clockvaluation_0) \taltstrans{\sigma_{1}}{t_{1}} \ldots \taltstrans{\sigma_s}{t_s} 
    (l_{1,1}, \clockvaluation_{1,1}) \taltstrans{\sigma_{1,1}}{t_{1,1}}
    \ldots \taltstrans{\sigma_{1,k_1}}{t_{1,k_1}} \\
    (l_{1,k_1}, \clockvaluation_{1,k_1}) \taltstrans{\sigma_{2,1}}{t_{2,1}} \ldots \taltstrans{\sigma_{m,k_m}}{t_{m,k_m}} (l_{m,k_m}, \clockvaluation_{m, k_m}) \taltstrans{\sigma_e}{t_e} (l_e, \clockvaluation_e) \taltstrans{}{} \ldots
  \end{multline*}
  We need to show that $r$ is an accepting run in $\taenc^{(i)}$.
  First, note that $(l_{i,j}, \clockvaluation_{i,j}) \taltstrans{a_{i,j}}{t_{i,j}} (l_{i,j+1}, \clockvaluation_{i,j+1})$ is a valid transition within $S_i$: From $\rho \in \taenc^{(i-1)}$, it follows that it is a valid transition in $\taenc^{(i-1)}$.
  Furthermore, $w, z' \cdot \la (a_{1,1}, t_{1,1}) \cdot \ldots \cdot (a_{i,j}, t_{i, j}) \ra \models \beta_i$ and therefore, the location $l_{i,j+1}$ is a location of $S_i$ (i.e., it is not deleted in Algorithm~\ref{alg:transform-plan}, line~\ref{alg:transform-plan:delete}).
  As the locations and switches of $\taenc^{(i)}$ are copied from $\taenc^{(i-1)}$, the transition is possible in $\taenc^i$.
  Next, notice that for the transitions $(l_{j, k_j}, \clockvaluation_{j, k_j}) \taltstrans{\sigma_{j+1,1}}{t_{j+1,1}} (l_{j+1,1}, \clockvaluation_{j+1,1})$ switching from $S_j$ to $S_{j+1}$, the only difference to $\taenc^{(i-1)}$ is an additional guard $x_\gamma \in I_{j}$.
  From $w, z' \cdot \la (a_{1,1}, t_{1,1}) \cdot \ldots \cdot (a_{j,1}, t_{j,1}) \ra, \la  (a_{j,2}, t_{j,2}), \cdots, (a_{m,k_m}, t_{m,k_m}) \ra \models (\beta_j \wedge \neg \alpha_2) \until{I_j} (\beta_{j+1} \wedge \neg \alpha_2)$, it follows that $\sum_{j=2}^{k_i} t_{i,j} + t_{i+1,1} \in I_i$.
  As $x_\gamma$ is reset on the transition $(l_{j,k_j}, \clockvaluation_{j,k_j}) \taltstrans{\sigma_{j+1,1}}{t_{j+1,1}} (l_{j+1,1}, \clockvaluation_{j+1,1})$ and because it is not reset in any transition within $S_j$, it follows that $\clockvaluation_{j,k_j}(x_\gamma) = \sum_{j=2}^{k_i} t_{i,j}$ and therefore, $\clockvaluation_{j,k_j}(x_\gamma) + t_{j+1} \in I_j$, i.e., the guard $x_\gamma \in I_j$ is satisfied and the transition is valid.
  Finally, the same holds for the transition $(l_{m,k_m}, \clockvaluation_{m,k_m}) \taltstrans{\sigma_e}{t_e} (l_e, \clockvaluation_e)$: As $x_\gamma$ is reset on the incoming transition to $(l_{m,1}, \clockvaluation_{m,1})$ and not reset afterwards, $\clockvaluation_{m,k_m} = \sum_{j=2}^{k_m} t_{m,j}$.
  From $w, z' \cdot \la (a_{1,1}, t_{1,1}) \cdot \ldots \cdot (a_{j,1}, t_{j,1}) \ra, \la  (a_{j,2}, t_{j,2}), \cdots, (a_{m,k_m}, t_{m,k_m}) \ra \models (\beta_j \wedge \neg \alpha_2) \until{I_j} (\beta_{j+1} \wedge \neg \alpha_2)$ , it follows that $\clockvaluation_{m,k_m} + t_e \in I_m$.
  Therefore, every transition of $r$ is valid in $\taenc^{(i)}$ and so $r \in \accfinruns(\taenc^{(i)})$.
  Finally, as $r$ is an accepting run in $\taenc^{(i-1)}$, it is also an accepting run in $\taenc^{(i-1)}$.
  Therefore, $\rho \in \lang(\taenc^{(i)})$.
\end{proofE}

\section{Evaluation}

We have implemented the approach in the tool \texttt{taptenc}.
The tool constructs a \ac{TA} \taenc that encodes the plan, the robot self model, and the constraints, as described above.
It then uses \textsc{Uppaal}~\cite{bengtssonUPPAALToolSuite1996} to determine a trace of \taenc that reaches a final state.
With \autoref{thm:platform-encoding}, this trace corresponds to an execution of the plan and the robot platform that satisfies all constraints.

We have evaluated the approach in a scenario inspired by the \ac{RCLL}.
In the following, we first describe the robot's high-level actions and plans constructed from those actions, as well as the corresponding timing constraints, before we describe the robot self model in the \ac{RCLL} setting.

\subsection{High-Level Actions}
The robot has the following high-level durative actions available:
\begin{description}
  \item[$\goto(m, m')$:] Move from machine $m$ to machine $m'$.
  \item[$\pick(o, m)$:] Pick up an object $o$ from the machine $m$.
  \item[$\shelfpick(o, m)$:] Fetch a workpiece $o$ from the shelf of machine $m$.
  \item[$\putact(o, m)$:] Put the object $o$ onto machine $m$.
  \item[$\pay(o, m)$:] Use object $o$ to pay for additional material at machine $m$.
\end{description}
As these actions are durative, there is a corresponding \emph{start} and \emph{end} action for each.
A high-level plan may look as follows:
\begin{align*}
  &\sac{\goto(\mi{start}, \mi{CS}_1)}, \eac{\goto(\mi{start}, \mi{CS}_1)},
  \\
  &\sac{\shelfpick(\mi{CC}_1, \mi{CS}_1)}, \eac{\shelfpick(\mi{CC}_1, \mi{CS}_1)},
  \\
  &\sac{\putact(\mi{CC}_1, \mi{CS}_1)}, \eac{\putact(\mi{CC}_1, \mi{CS}_1)},
  \\
  &\sac{\goto(\mi{CS}_1, \mi{BS}_1)}, \eac{\goto(\mi{CS}_1, \mi{BS}_1)},
  \\
  &\sac{\pick(\mi{BS}, \mi{WS}_1)}, \eac{\pick(\mi{BS}, \mi{WS}_1)}
\end{align*}

In this example, the robot first moves to the machine $\mi{CS}_1$, picks up a workpiece $\mi{CC}_1$ from the shelf of the machine, and then puts it into the machine.
In the next step, the robot moves to the machine $\mi{BS}$ and picks up a workpiece from the machine.
It may later use this workpiece to continue the production process, e.g., by moving to another machine, and so on.
Here, we are not particularly concerned with what the plan achieves, but focus on the constraints between the high-level actions and the robot platform.

We have the following timing constraints:
\begin{itemize}
  \item Each \pick takes between \SI{15}{\sec} and \SI{20}{\sec}.
    Therefore, for each durative pick action, we add a timing constraint $\relconstraint(i,j, [15, 20])$, where $i$ and $j$ are action indices of the \emph{start} and \emph{end} action of the corresponding \pick action.
  \item Similarly, \goto may take between \SI{30}{\sec} and \SI{45}{\sec}.
  \item The robot should not stall for more than \SI{30}{\sec}, because the user may think it is broken.
    Therefore, for each \emph{end} action with index $i$, we add the constraint $\relconstraint(i, i+1, [0, 30])$.
  \item The  robot should start executing the high-level plan after at most \SI{30}{\sec}, i.e., action $1$ should occur in the interval $[0, 30]$ after the start, which can be formalized with the constraint $\absconstraint(1, [0, 30])$.
\end{itemize}

\subsection{Robot Self Model}\label{sec:robot-self-model}

The self model of the robot consists of the robot's perception unit, its gripper, and its communication unit.

\paragraph{Perception Unit}
\begin{figure}[htb]
\begin{center}
  \includestandalone[width=\textwidth]{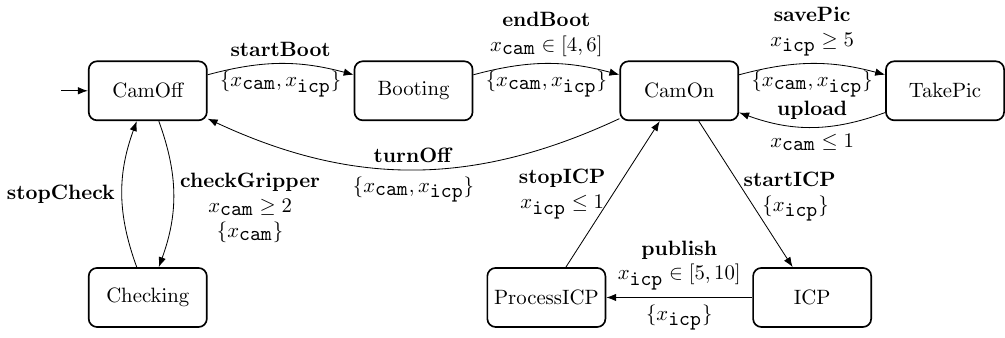}
\end{center}
\caption[A \acs*{TA} that models the robot's perception unit.]{A \ac{TA} model of robot's perception unit $\fancy A_{\fancy M_\texttt{perc}}$, which is an extended version of the model shown in \autoref{fig:camera-model-example}.}%
\label{fig:perception-ta}
\end{figure}

\autoref{fig:perception-ta} shows the robot's perception unit.
Similar to the model shown in \autoref{fig:camera-model-example}, the camera is initially off and needs some time before it can be used.
When the camera is on, it may be used for object detection based on an \acfi{ICP} algorithm, which compares the RGB/D image of the camera with a pre-recorded model and by doing so computes the precise position of a target object.
While the details of this algorithm are not relevant here, an important aspect is that it takes some time for first processing the input data and then processing the result.
This process is modeled with the two locations \mi{ICP} and \mi{ProcessICP} and the corresponding actions.
Additionally, after the robot has successfully computed the precise object position, it may take a picture of the object, e.g., for training a neural network for object detection.
This picture is then uploaded to a central storage.
Finally, the robot's gripper is also equipped with an infrared sensor that can detect whether there is an object in the gripper.
As the camera interferes with the infrared sensor, it must be turned off while the robot is checking its gripper sensor.

\paragraph{Gripper}

\begin{figure}[htb]
\begin{center}
  \includestandalone{figures/platform-ta-calib}
\end{center}
\caption[A \acs*{TA} that models the robot's axis calibration unit.]{The \ac{TA} $\fancy A_{\fancy M_\texttt{calib}}$ that models the robot's axis calibration module.}%
\label{fig:calib-ta}
\end{figure}

\autoref{fig:calib-ta} shows the self model of the robot's gripper.
Initially, the gripper is uncalibrated and needs to be calibrated before usage.
Whenever the gripper is used, it becomes less precise.
After being used twice, it is again uncalibrated.

\paragraph{Communication Unit}

\begin{figure}[htb]
\begin{center}
  \includestandalone{figures/platform-ta-comm}
\end{center}
\caption[A \acs*{TA} that models of robot's machine instruction module.] {The \ac{TA} $\fancy A_{\fancy M_\texttt{comm}}$ that models the robot's communication module.}%
\label{fig:comm-ta}
\end{figure}

\autoref{fig:comm-ta} shows the communication unit of the robot, which is the third component of the robot self model.
Whenever the robot intends to use a machine for some processing step, it needs to instruct the machine by sending a command.
After it has sent the command, it must received an acknowledgement of the instruction before it can continue.

\subsection{Platform Constraints}

\newcommand*{\actconstraint}[2]{\ensuremath{\Psi^{\text{#1}}_{\text{#2}}}}

As a final final step, we need to connect the platform models with the high-level plan actions by formulating platform constraints.
Before doing so, we define some notational devices:
\begin{align*}
  \actconstraint{start}{A}(\vec{y}) &\eqdef \exists \vec{x} \occ(\sac{\mi{A}(\vec{x}, \vec{y})})
  \\
  \actconstraint{end}{A}(\vec{y}) &\eqdef \exists \vec{x} \occ(\eac{\mi{A}(\vec{x}, \vec{y})})
  \\
  \actconstraint{op}{grasp} &\eqdef \actconstraint{op}{pick} \vee \actconstraint{op}{getFromShelf}
  \\
  \actconstraint{op}{release} &\eqdef \actconstraint{op}{put} \vee \actconstraint{op}{pay}
  \\
  \actconstraint{op}{manip} &\eqdef \actconstraint{start}{grasp} \vee \actconstraint{start}{release}
\end{align*}

The formula $\actconstraint{op}{A}(\vec{y})$ (where $\mi{op} \in \{ \mi{start}, \mi{end} \}$) allows us to specify an action occurrence of a start or end action for the durative action $A$, where only some of the action parameters are fixed by $\vec{y}$ and all other action parameters may be set arbitrarily.
This is helpful because we often only want to specify that some action instance, e.g., a \pick, occurs, independent of what the action's parameters are.
As an example, $\actconstraint{start}{goto}$ matches any start action for the durative \goto, independent of the action's parameters.
Furthermore, we also define $\actconstraint{op}{grasp}$ as the occurrence of any grasping action, i.e., \mi{pick} or \mi{getFromShelf} actions, similarly for \mi{release} and \mi{manip}.

We start with the constraints for the perception unit:
\begin{align*}
  \gamma^1_\text{perc} &\eqdef \chainconstraint\left( \la\la \mi{ICP}, [0, \infty) \ra, \la \mi{ProcessICP}, [0, 0] \ra, \la \neg \mi{ICP}, [10, 10] \ra \ra, \actconstraint{start}{manip}, \actconstraint{end}{manip}\right)
  \\
  \gamma^2_\text{perc} &\eqdef \chainconstraint\left( \la\la \top, [0, \infty) \ra, \la \mi{TakePic}, [0, \infty) \ra, \la \top, [0, \infty) \ra \ra, \actconstraint{start}{manip}, \actconstraint{end}{manip} \right)
  \\
  \gamma^3_\text{perc} &\eqdef \chainconstraint\left( \la \la \mi{CamOff} \vee \mi{Checking}, [0, \infty) \ra \ra, \actconstraint{start}{goto}, \actconstraint{end}{goto} \right)
  \\
  \gamma^4_\text{perc} &\eqdef \chainconstraint\left( \la \la \mi{Checking}, [0, \infty) \ra \ra, \actconstraint{start}{goto}, \neg \actconstraint{start}{goto} \right)
  \\
  \gamma^5_\text{perc} &\eqdef \chainconstraint\left( \la \la \mi{Checking}, [0, \infty) \ra \ra, \actconstraint{end}{goto}, \neg \actconstraint{end}{goto} \right)
\end{align*}
The constraints require the following:
\begin{enumerate}
  \item During any manipulation action, the robot must run \ac{ICP}, immediately process the results, and then keep \ac{ICP} off for exactly \SI{10}{\sec}.
  \item During any manipulation action, the robot must also take a picture of the object at some point.
  \item While the robot is moving (i.e., while it is performing a \goto action), the camera should only be used to check the gripper.
    In particular, it must not boot the camera, run \ac{ICP}, or take a picture.
  \item Whenever the robot starts moving, it must check whether there is an object in the gripper.
  \item Similarly, at the end of each \goto, the robot must check the gripper again.
\end{enumerate}

Next, we also require certain states of $\fancy A_{\fancy M_\texttt{calib}}$, which models the gripper and its calibration:
\begin{align*}
  \gamma^1_\text{calib} &\eqdef \chainconstraint\left( \la \la \neg \mi{Usage1} \wedge \neg \mi{Usage2}, [0, \infty) \ra \ra, \actconstraint{start}{goto}, \actconstraint{end}{goto} \right)
  \\
  \gamma^2_\text{calib} &\eqdef \chainconstraint\left( \la \la \neg \mi{Calibrate}, [0, \infty) \ra \ra, \actconstraint{end}{grasp}, \actconstraint{start}{release} \right)
  \\
  \gamma^3_\text{calib} &\eqdef \chainconstraint\left( \la \la \top, [0, \infty) \ra, \la \mi{Precise}, [0, \infty) \ra, \la \top, [0, \infty) \ra \ra, \actconstraint{end}{grasp}, \actconstraint{end}{pay} \right)
  \\
  \gamma^4_\text{calib} &\eqdef \chainconstraint\left( \la \la \mi{Usage1} \vee \mi{Usage2}, [0, \infty) \ra \ra, \actconstraint{start}{manip}, \actconstraint{end}{manip} \right)
\end{align*}
In words, we require:
\begin{enumerate}
  \item The gripper must not be used while the robot is moving, because any manipulation task is dangerous while the robot is moving.
  \item The gripper must not calibrate between a \emph{grasp} and a \emph{release} action.
    After any grasp action, the robot is holding an object, which would be dropped if the gripper was recalibrated.
  \item Whenever the robot performs a \emph{pay} action, it must do so with a precisely calibrated gripper.
    This is because the payment operation is quite brittle and must be performed with utmost care.
  \item For any manipulation action, the robot actually needs to use the gripper.
    Without this constraint, never switching the location in the gripper model $\fancy A_{\fancy M_\texttt{calib}}$ (and thus never actually using the gripper) would be feasible, which obviously is not the intended behavior.
\end{enumerate}

Finally, we turn towards machine communication.
As we may need to communicate with multiple machines, we will use the \ac{TA} $\fancy A_{\fancy M_\texttt{comm}}$ multiple times, once for each machine.
We add an index $i$ to each \ac{TA} location to refer to the $i$th machine $m_i$.
We have two constraints for each machine $m_i$:
\begin{align*}
  \gamma^1_{\text{comm},i} &\eqdef \chainconstraint\left( \la \la \mi{Idle}_i \vee \mi{Prepare}_i, [0, \infty) \ra, \la \mi{Prepared}_i, [0, \infty) \ra \ra, \actconstraint{end}{put}(m_i), \actconstraint{start}{pick}(m_i) \right)
  \\
  \gamma^2_{\text{comm},i} &\eqdef \chainconstraint\left( \la \la \mi{Idle}_i, [0, \infty) \ra \ra, \actconstraint{end}{pick}(m_i), \actconstraint{end}{put}(m_i) \right)
\end{align*}
This requires the following machine communication:
\begin{enumerate}
  \item After the robot put down any workpiece into machine $m_i$, it needs to prepare the machine so the machine starts processing the workpiece.
    As the robot should not pick up the workpiece before it has been processed, it needs to do so before it picks it up again.
  \item Otherwise, after picking up the workpiece and before putting the next workpiece into the machine, it must not send any instructions.
    As there is no workpiece in the machine, sending any instruction would break it.
\end{enumerate}

\subsection{Results}

\begin{table}[htb]
  \centering
  \begin{tabular}{l r r r r r r}
    \toprule
    & \multicolumn{5}{c}{Time (\si{\second})}
    \\
    \cmidrule(lr){2-6}
    Platform TA & \texttt{trans} & \texttt{load\_ta} & \texttt{reach} & \texttt{tracer} & total & \# locations
    \\
    \midrule
    \texttt{perc} & 0.32 &  0.11 &  0.08 &  0.03 &  0.54 & 655 \\
    \texttt{calib} & 0.07 &  0.04 &  0.03 & 0.01 &  0.15 & 271 \\
    \texttt{comm} & 0.02&  0.01 &  0.01 &  0.01 &   0.05 & 69 \\
    \texttt{perc} + \texttt{calib} & 0.63 &  0.85 &  0.58 &  0.14 &  2.2 & 2660 \\
    \: + 1x \texttt{comm} & 1.2 &  2.4&  1.6 &  0.26 & 5.46 & 4566 \\
    \: + 2x \texttt{comm} & 2.0 &  4.0 &  2.5 &  0.38 & 8.88 & 5645 \\
    \: + 3x \texttt{comm} & 4.2&  8.7&  4.9&  0.63& 18.43 & 8600 \\
    \: + 4x \texttt{comm} & 13.5&  18.1&  9.0&  1.1& 41.7 & 13883 \\
    \bottomrule
   \end{tabular}
   \caption[Average execution times.]{Average execution times of five runs on plans of length $50$. \texttt{trans}: building the encoding and decoding, \texttt{load\_ta}: required preprocessing step of \texttt{verifyta}, \texttt{reach}: reachability analysis, \texttt{tracer}: computation of a concrete trace.}
 \label{fig:merges}
\end{table}

As a first benchmark, we fixed the plan length to 50 actions and considered multiple combinations of the three components described above.
The results are shown in \autoref{fig:merges}.
We can see that if we only consider the perception unit \texttt{perc}, then it takes a total execution time of \SI{0.54}{\sec} to compute the transformed plan.
Roughly half of the time (\SI{0.32}{\sec}) is needed to construct the \ac{TA}.
When extending the model, e.g., to a perception unit, a gripper, and 4 communication units, the constructed \ac{TA} has $13883$ locations and the average execution time of the transformation is \SI{41.7}{\sec}.
Interestingly, the reachability analysis itself only takes \SI{9.0}{\sec}, less than the time needed to construct the automaton and also less than loading the model into the verification tool.

\begin{table}[tb]
  \centering
  \begin{tabular}{r rrr rrr}
    \toprule
    & \multicolumn{3}{c}{Time (\si{\second})}
    & \multicolumn{3}{c}{\# locations} \\
    \cmidrule(rl){2-4} \cmidrule(rl){5-7}
    \shortstack{Plan \\ length} & \texttt{perc} & \texttt{calib} & \shortstack{\texttt{perc} \\ + \texttt{calib}}
                & \texttt{perc} & \texttt{calib} & \shortstack{\texttt{perc} \\ + \texttt{calib}}
    \\
    \midrule
    50 & .6 & .1 & 2.1
       & 662 & 269 & 2574
    \\
    100 & 2.0 & .5 & 7.7
        & 1325 & 527 & 5513
    \\
    150 & 4.9 & .1 & 15.5
        & 1978 &  769 & 8297
    \\
    300 & 19.2 & 2.9 & 53.1
        & 3953  & 1538 & 16476
    \\
    \bottomrule
  \end{tabular}
  \caption[Average total transformation time and encoding size.]{Average total transformation time and encoding size of five runs on plans with varying lengths.}
  \label{fig:scaling-plan-length}
\end{table}

In a second benchmark, we investigated how the approach scales with increasing plan length, as shown in \autoref{fig:scaling-plan-length}.
We can see that with increasing plan length, the execution time also increases significantly.
However, even for plans with $150$ actions, the transformation of a plan based on a self model consisting of the perception unit and the gripper takes \SI{15.5}{\sec} in average.
Depending on the application, this may be an acceptable execution time, especially for such a large plan.

\section{Discussion} \label{sec:transformation:discussion}

In this chapter, we have considered a second approach towards the transformation problem that makes some simplifying assumptions.
Most importantly, we now only consider a plan (i.e., a sequence of actions) rather than arbitrary \golog programs.
Second, we do not partition the actions into controllable and environment actions, but instead assume that all actions are controllable by the agent.
This allows us to model the transformation problem as a \emph{reachability problem} on \aclp{TA}.
We did so by constructing a \acl{TA} such that every run on the \acl{TA} corresponds to an execution of the plan with additional platform actions.
We constructed the automaton in such a way that each accepting run satisfies the specification.
In contrast to the first approach, this approach scales well with larger problem instances and large robot self models.

There are several reasons why the second approach performs better than the first.
First, the simplifying assumptions make the problem significantly easier.
As an example, we do not need to consider all possible ways the environment may act, but instead we only need to find a single run that reaches a final state.
Therefore, we do not need to branch on every possible environment action, which significantly reduces the considered search space.
However, the simplifying assumptions are not the only reason for the better performance:
As the approach constructs a \acl{TA} and then solves a reachability problem on the constructed automaton, we were able to use the well-established verification tool \uppaal~\cite{bengtssonUPPAALToolSuite1996,behrmannDevelopingUPPAAL152011}, which has seen considerable efforts to improve its performance, e.g., with symbolic model checking \cite{larsenModelcheckingRealtimeSystems1995}, control structure analysis ~\cite{larsenUppaalStatusDevelopments1997}, and symmetry reduction \cite{hendriksAddingSymmetryReduction2004}.
In contrast, the synthesis method from \autoref{chap:synthesis} is not based on \uppaal, but instead on the newly developed tool \tacos.
While \tacos has seen some efforts (e.g., search node re-usage ~\cite{hofmannTACoSToolMTL2021}) towards performance improvement, many state-of-the-art techniques such as symbolic model checking are also applicable to \tacos but have not been implemented yet.


\chapter{Abstracting Noisy Robot Programs}\label{chap:abstraction}

In the previous chapters, we have described several methods to transform an abstract program to a realizable program on a specific robot platform based on a self model of the robot and temporal constraints in the form of \ac{MTL} formulas.
This allows us to specify timing constraints that must be satisfied during the execution of the program.
The focus was \emph{metric time}: We extended the logic \esg to \tesg by means of timed traces and clock constraints and we used \aclp{TA} for the robot self  models.

In this chapter, we turn towards a different aspect. We consider \emph{uncertainty} in robot programs in the form of noisy sensors and effectors.
In robotics applications, uncertainty is ubiquitous: A robot sensor is almost never exact and actions rarely have the desired effect with certainty.
Instead, a robot sensor typically has some noise such that it measures a value close but not equal to the real value.
Similarly, an action may have several possible outcomes, each of which has some likelihood.

While expressing noisy sensors and effectors in a \acl{BAT} is desirable and often necessary to describe a robot, we ideally want to ignore probabilistic aspects when programming a robot, for several reasons:
\begin{enumerate}
  \item Correctly designing a probabilistic domain and writing a probabilistic program is challenging, because we need to consider all possible outcomes and their probabilities.
  \item Reasoning about probabilities is hard:
    Plan existence in a probabilistic planning domain is undecidable~\parencite{littmanComputationalComplexityProbabilistic1998}.
    Similarly, in the context of the situation calculus, verifying some property of a belief program is undecidable, even if all fluents are nullary and the successor state axioms are context-free~\parencite{liuProjectionProbabilisticEpistemic2022}.
  \item Understanding how such a system operates is difficult:
    A probabilistic plan (or similarly, a belief program) typically contains many conditional branches to deal with the different outcomes.
    Also, as we will demonstrate later, analyzing an execution trace of a \golog program with noisy actions is cumbersome, because it is cluttered with noise and sensing actions.
\end{enumerate}

Hence, we need to incorporate noisy actions into the domain, but at the same time, we want to ignore them for writing a program.
In order to accomplish this, we propose to use \emph{abstraction}. 
Generally speaking, abstraction is the ``process of mapping a representation of a problem onto a new representation'' \cite{giunchigliaTheoryAbstraction1992}.
In the context of intelligent agents, abstraction typically serves three purposes \cite{belleAbstractingProbabilisticModels2020}:
\begin{enumerate}
  \item It provides a way to structure knowledge.
  \item It allows reasoning about larger problems by abstracting the problem domain, resulting in a smaller search space.
  \item It may provide more meaningful explanations and is therefore critical for \emph{explainable AI}.
\end{enumerate}

Based on \cite{banihashemiAbstractionSituationCalculus2017}, abstraction in our context works as follows:
In addition to the low-level \ac{BAT} that describes the robot in detail, including its noisy sensors and effectors, we define a second, high-level \ac{BAT} that abstracts aways all those details and may be non-stochastic.
We use a \emph{refinement mapping} that connects the high-level with the low-level \ac{BAT} by mapping each high-level proposition to a low-level formula and each high-level action to a low-level program.
To establish the equivalence between the two programs, we define a suitable notion of \emph{bisimulation}~\parencite{milnerAlgebraicDefinitionSimulation1971}, which is a mapping from high-level to low-level states and which intuitively requires the following:
\begin{enumerate}
  \item If the high-level state satisfies some formula $\alpha$, then the low-level state satisfies the refined formula $m(\alpha)$ (and vice versa).
  \item If the agent can execute some action $a$ in the high-level state, then it can execute the refined program $m(a)$ in the low-level state (and vice versa) and the resulting states are again bisimilar.
\end{enumerate}

Our starting point is the logic \ds~\parencite{belleReasoningProbabilitiesUnbounded2017}, a modal variant of the situation calculus with probabilistic belief.
In \autoref{sec:dsg}, we extend \ds by defining a transition semantics for noisy \golog programs.
Based on this transition semantics, we then propose a notion of abstraction of noisy programs, building on top of abstraction of probabilistic static models \cite{belleAbstractingProbabilisticModels2020} and non-stochastic dynamic models in the classical situation calculus~\cite{banihashemiAbstractionSituationCalculus2017}.
We do so by defining a notion of bisimulation of probabilistic dynamic systems in \autoref{sec:bisimulation} and we show that the notions of sound and complete abstraction carry over.
We also demonstrate how this abstraction framework can be used to define a high-level domain, where noisy actions are abstracted away and thus, no probabilistic reasoning is necessary.

\section{The Logic \dsg}\label{sec:dsg}

In \autoref{sec:sitcalc} as well as in \autoref{chap:timed-esg}, we have seen multiple variants of the situation calculus that allows modeling a robot by means of a \acl{BAT}.
While \tesg in \autoref{chap:timed-esg} focuses on modeling time in the situation calculus and assumes that the agent has complete knowledge, we now look at a different aspect, namely stochastic domains with incomplete knowledge.
In \autoref{sec:sitcalc}, we have described \es, which is a modal variant of the situation calculus that allows expressing the agent's knowledge.
This is done by means of \emph{epistemic states}, which are sets of worlds that the agent assumes to be possible.
In this setting, some formula is \emph{known} if it is true in all worlds in the epistemic state.
Building on top of \es, we have also summarized the logic \ds~\parencite{belleReasoningProbabilitiesUnbounded2017}, which extends \es by \emph{degrees of belief}.
Rather than knowing or not knowing some fact with certainty, the epistemic state assigns some probability to each possible world and therefore allows modeling uncertain beliefs.
In this section, we introduce \dsg, which extends \ds by a transition semantics for \golog programs, analogous to how \esg~\cite{classenLogicNonterminatingGolog2008,classenPlanningVerificationAgent2013} extends \es~\cite{lakemeyerSituationsSiSituation2004,lakemeyerSemanticCharacterizationUseful2011}.

\subsection{Syntax}

\dsg extends \ds with a transition semantics for \golog similar to the transition semantics in \esg and \tesg.
In the same way as \ds and similar to standard names, the logic uses a countably infinite set of \emph{rigid designators} \rigid with the unique name assumption and which allows to define quantification substitutionally.
Also similar to \ds, \es, and \esg, it uses a possible-world semantics, where situations are part of the semantics rather than appearing as terms in the language.
As before, we use the modal operator $[\cdot]$ to refer to the state after executing some program, e.g., $[ \delta ] \alpha$ states that $\alpha$ is true after every possible execution of the program $\delta$.
Additionally, we use the modal operator $\belconnector$ to describe the agent's \emph{belief}, e.g., $\bel{\loc(2)}{0.5}$ states that the agent believes with degree $0.5$ to be in location $2$.
Apart from belief, the language is similar to the language of \esg and \tesg, but excluding their temporal operators.
We summarize the language below and start with the logic's symbols:
\begin{definition}[Symbols of \dsg{}]
The symbols of the language are from the following vocabulary:
\begin{enumerate}
  \item infinitely many variables $x, y, \ldots, u, v, \ldots, a, a_1, \ldots$;
  \item rigid function symbols of every arity, e.g., $\mi{near}$, $\goto(x, y)$;
  \item fluent predicates of every arity, such as $\mi{\at(l)}$; we assume that this list contains the following distinguished predicates:
    \begin{itemize}
      \item $\poss$ to denote the executability of an action;
      \item $\oi$ to denote that two actions are indistinguishable from the agent's viewpoint; and
      \item $l$ that takes an action as its first argument and the action's likelihood as its second argument;
    \end{itemize}
  \item connectives and other symbols: $=$, $\wedge$, $\neg$, $\forall$, $\square$, $[\cdot]$,
    $\belconnector$.
    \qedhere
\end{enumerate}
\end{definition}
Note that in contrast to \autoref{chap:timed-esg}, for the sake of simplicity and analogous to \ds, we do not include fluent function symbols.
The terms of the language are built from variables and rigid function symbols:
\begin{definition}[Terms of \dsg{}]
  The set of terms of \dsg{} is the least set such that
  \begin{enumerate}
    \item every variable is a term,
    \item if $t_1, \ldots, t_k$ are terms and $f$ is a $k$-ary function symbol, then $f(t_1, \ldots, t_k)$ is a term.
      \qedhere
  \end{enumerate}
\end{definition}
We let $\rigid$ denote the set of all ground rigid terms and we assume that they contain the rational numbers, i.e., $\mathbb{Q} \subseteq \rigid$.
In contrast to \tesg, we do not distinguish several sorts and instead allow every ground rigid term as action term.
We can now define the formulas of the language:
\begin{definition}[Formulas] \label{def:dsg-formulas}
  The \emph{formulas of \dsg{}} are the least set such that
  \begin{enumerate}
    \item if $t_1,\ldots,t_k$ are terms and $P$ is a $k$-ary predicate symbol,
      then $P(t_1,\ldots,t_k)$ is a formula,
    \item if $t_1$ and $t_2$ are terms,
      then $(t_1 = t_2)$ is a formula,
    \item if $\alpha$ and $\beta$ are formulas,
      $x$ is a variable, $\delta$ is a program (defined below),%
      \footnote{
        Analogously to \tesg, although the definitions of formulas (\autoref{def:dsg-formulas}) and programs (\autoref{def:dsg-programs}) mutually depend on each other, they are still well-defined:
        Programs only allow static situation formulas and static situation formulas may not refer to programs.
        Technically, we would first need to define static situation formulas, then programs, and then all formulas.
        For the sake of presentation, we omit this separation.
      }
      and
      $r \in \mathbb{Q}$,
      then $\alpha \wedge \beta$, $\neg \alpha$, $\forall x.\, \alpha$,
      $\square\alpha$, $[\delta]\alpha$,
      and $\bel{\alpha}{r}$ are formulas.
      \qedhere
  \end{enumerate}
\end{definition}
We read $\square \alpha$ as ``$\alpha$ holds after executing any sequence of
actions'', $[\delta] \alpha$ as ``$\alpha$ holds after the execution of program $\delta$''
and $\bel{\alpha}{r}$ as ``$\alpha$ is believed with probability $r$''.%
\footnote{The original version of the logic also has an only-knowing modal operator \obelconnector, which captures the idea that something and only that thing is known. For the sake of simplicity, we ignore this operator in our presentation.}
We also write $\know{\alpha}$ for $\bel{\alpha}{1}$, to be read as ``$\alpha$ is known''.%
\footnote{We use ``knowledge'' and ``belief'' interchangeably, but do not require that knowledge be true in the real world (i.e., weak S5).}
We use \true as abbreviation for $\forall x \left(x = x\right)$ to denote truth. 
Free variables are implicitly understood to be quantified from the outside.
\begin{table}[htb]
  \centering
  \caption{Operator precedence in the logic \dsg.}
  \label{tab:dsg-operator-precedence}
  \begin{tabular}{*{13}{Cc}}
    \toprule
    Precedence & 1 & 2 & 3 & 4 & 5 & 6 & 7 & 8 & 9 & 10 & 11 & 12
    \\
    Operator & $\lbrack \cdot \rbrack$ & $\neg$ & $\belconnector$ & $\know$ & $\wedge$ & $\vee$ & $\forall$ & $\exists$ & $\supset$ & $\equiv$ & $\llbracket \cdot \rrbracket$ & $\square$
    \\
    \bottomrule
  \end{tabular}
\end{table}
As in \tesg, we assign a precedence to each connective, which is shown in \autoref{tab:dsg-operator-precedence}.
For a formula $\alpha$, we write $\alpha^x_r$ for the formula resulting from $\alpha$ by substituting every occurrence of $x$ with $r$.
For a finite set of formulas $\Sigma = \{ \alpha_1, \ldots, \alpha_n \}$, we may just write $\Sigma$ for the conjunction $\alpha_1 \wedge \ldots \wedge \alpha_n$, e.g., $\know{\Sigma}$ for $\know{(\alpha_1 \wedge \cdots \wedge \alpha_n)}$.

A predicate symbol with terms from $\rigid$ as arguments is called a \emph{ground atom}, and we denote the set of all ground atoms with $\atoms$.
Furthermore, a formula is called \emph{bounded} if it contains no $\square$ operator,
\emph{static} if it contains no $[\cdot]$ or $\square$ operators,
\emph{objective} if it contains no \belconnector or \know{},
and
{\em fluent} if it is static and does not mention $\poss$, \belconnector, or \know{}.

Finally, we define the syntax of \golog{} program expressions referred to by the operator $[\delta]$.

\begin{definition}[Program Expressions] \label{def:dsg-programs}
  \[
    \delta ::= t \smid \alpha? \smid \delta_1 ; \delta_2 \smid \delta_1 \vert \delta_2
    \smid \pi x.\, \delta
    \smid \delta^*
  \]
  where $t$ is a ground rigid term and $\alpha$ is a static formula. A program expression consists of actions $t$, tests $\alpha?$, sequences $\delta_1;\delta_2$, nondeterministic branching $\delta_1 \vert \delta_2$, nondeterministic choice of argument $\pi x.\, \delta$,
  and nondeterministic iteration $\delta^*$.
\end{definition}

In contrast to \tesg, we do not allow interleaved concurrency $\delta_1 \| \delta_2$,
\footnote{The reason will become apparent later on. Intuitively, if we allow interleaved concurrency, then the low-level program could pause the execution of a high-level action and continue with a different high-level action, possibly leading to different effects.
  This significantly complicates the formal treatment relating the probabilities of high-level worlds to their low-level counterparts.
}
but we include the nondeterministic pick operator $\pi x.\, \delta$.
We also use $\nil$ as abbreviation for $\true?$, the empty program that always succeeds.
Similar to formulas, $\delta^x_r$ denotes the program expression resulting from $\delta$ by substituting every $x$ with $r$.
Furthermore, we define $\gif \ldots \gfi$ and $\gwhile \ldots \gdone$ as syntactic sugar as follows:
\begin{align*}
  \gif \phi \gthen \delta_1 \gelse \delta_2 \gfi
  &\eqdef
  (\phi?; \delta_1) \mid (\neg \phi?; \delta_2)
  \\
  \gif \phi_1 \gthen \delta_1 \gelif \phi_2 \gthen \delta_2 \gfi
  &\eqdef
  (\phi_1?; \delta_1) \mid (\neg \phi_1 \wedge \phi_2?; \delta_2)
  \\
  \gwhile \phi \gdo \delta \gdone
  &\eqdef
  (\phi?; \delta)^*; \neg \phi?
\end{align*}

\subsection{Semantics}

In the same way as \es and its extensions, \dsg uses a possible-world semantics, where a world defines the state of the world not only initially but after any sequence of actions.
Here, a sequence of actions consists of only action symbols and in contrast to \tesg does not contain timesteps.
Additionally, an \emph{epistemic state} describes the agent's belief.
Here, an epistemic state is a distribution that assigns a weight to each possible world.
Based on the epistemic state, the operator \belconnector describes the degree of belief.
To capture noisy actions and sensors, \emph{likelihood axioms} describe the possible outcomes of an action and \emph{observational indistinguishability} defines which states of the world the agent may tell apart.
Both likelihood of possible outcomes and observational indistinguishability are built into the worlds using distinguished symbols and then modelled using \emph{basic action theories}, as described in \autoref{sec:dsg-bat}.

We start with traces, which are sequences of (action) terms and which we will use to describe possible executions of a program.
As we do not distinguish sorts, every sequence of ground rigid terms can be considered as trace:
\begin{definition}[Trace]
  A trace $z = \la a_1, \ldots, a_n \ra$ is a finite sequence of \rigid.
  We denote the set of traces as \traces and the empty trace with $\la\ra$.
\end{definition}
A world defines the truth of each ground atom from $\atoms$ not only initially but after any sequence of actions:
\begin{definition}[World]
  A world is a mapping $w : \atoms \times \traces \rightarrow \{0, 1\}$.
  The set of all worlds is denoted as \worlds.
\end{definition}

We require that every world $w \in \worlds$ defines the following distinguished predicates:
\begin{itemize}
  \item a unary predicate $\poss$ which defines possible actions,
  \item a binary predicate $l$ that behaves like a function (i.e., there is exactly one $q \in \mathbb{Q}$ such that $w[l(a, q), z] = 1$ for any $a, z$), 
  \item a binary predicate $\oi \subseteq \rigid \times \rigid$ to be understood as equivalence relation which describes the observational indistinguishability of traces.
\end{itemize}

We call a pair $\left(w, z\right) \in \worlds \times \traces$ a \emph{state}, we denote the set of all states with \states, and we use $\stateset, \stateset_i, \ldots \subseteq \states$ to denote sets of states.

Given a state $(w, z)$, the predicate $l(a, q)$ states that the action likelihood of action $a$ in state $(w, z)$ is equal to $q$.
We extend $l$ to $l^*$ to define the likelihood of an action sequence:
\begin{definition}[Action Sequence Likelihood]
  The action sequence likelihood $l^*: \worlds \times \mathcal{Z} \rightarrow \mathbb{Q}^{\geq 0}$ is defined inductively:
  \begin{itemize}
    \item $l^*\mleft( w, \la\ra \mright) = 1$ for every $w \in \worlds$,
    \item $l^*\mleft(w, z \cdot r \mright) = l^*\mleft(w, z\mright) \times q$ where $w \mleft[ l(r, q), z \mright] = 1$.
      \qedhere
  \end{itemize}
\end{definition}
Next, to deal with partially observable states, we define:
\begin{definition}[Observational indistinguishability]
  ~
  \begin{enumerate}
    \item
      Given a world $w \in \worlds$, we define the relation $\oisim \subset \mathcal{Z} \times \mathcal{Z}$ inductively:
      \begin{itemize}
        \item $\la\ra \oisim z'$ iff $z' = \la\ra$
        \item $z \cdot r \oisim z'$ iff $z' = z^* \cdot r^*$, $z \oisim z^*$, and $w \mleft[ \oi(r,r^*), z \mright] = 1$
      \end{itemize}
    \item
      We say $w$ is \emph{observationally indistinguishable} from $w'$, written $w \oicomp w'$ iff for all $a, a' \in \rigid$, $z \in \traces$:
      \[
        w \mleft[ \mi{oi}(a,a'),z \mright] = w' \mleft[ \mi{oi}(a,a'),z \mright]
      \]
    \item
      For $w, w' \in \worlds$, $z, z' \in \mathcal{Z}$, we say $\left(w, z\right)$ is \emph{observationally indistinguishable} from $\left(w', z'\right)$,  written $\left(w, z\right) \oicomp \left(w', z'\right)$, iff $w \oicomp w'$ and $z \oisim z'$.
      \qedhere
  \end{enumerate}
\end{definition}
Intuitively, $z \oisim z'$ means that the agent cannot distinguish whether it executed $z$ or $z'$.
For states, $\left(w, z\right) \oicomp \left(w', z'\right)$ is to be understood as ``if the agent believes to be in state $\left(w, z\right)$, it may also actually be in state $\left(w', z'\right)$'', i.e., it cannot distinguish the possible worlds $w, w'$ and traces $z, z'$.
As \oicomp is an equivalence relation, the set of its equivalence classes on a set of states \stateset induces a partition, which we denote with $\stateset / \oicomp$.

As another notational device, we extend the executability of an action to traces:
\begin{definition}[Executable trace]
  For a trace $z$, we define the formula $\exec(z)$ inductively:
  \begin{itemize}
    \item For $z = \la\ra$, $\exec(z) \eqdef \true$.
    \item For $z = a \cdot z'$, $\exec(z) \eqdef \poss(a) \wedge [a] \exec(z')$.
      \qedhere
  \end{itemize}
\end{definition}
The first item states that the empty action sequence is always executable.
The second item inductively states that a sequence $z = a \cdot z'$ is executable if $a$ is currently possible (i.e., if $\poss(a)$ is true) and $z'$ is executable after doing action $a$ (i.e., if $[a] \exec(z')$ is true).


Based on observational indistinguishability and executability, we can now define \emph{compatible states}:
\begin{definition}[Compatible States]\label{def:compatible-states}
  Given an epistemic state $e$, a world $w$, a trace $z$, and a formula $\alpha$, we define the states $\stateset^{e,w,z}_\alpha$ compatible to $\left(e, w, z\right)$ wrt to $\alpha$:
  \[
    \stateset^{e,w,z}_\alpha = \{ (w', z') \mid (w', z') \oicomp (w, z), e, w' \models \exec(z') \wedge [z']\alpha \}
    \qedhere
  \]
\end{definition}
We may write $\stateset_\alpha$ for $\stateset^{e,w,z}_\alpha$ if $e,w,z$ are clear from the context.
Intuitively, the set $\stateset^{e,w,z}_\alpha$ consists of the states $(w', z')$ that are indistinguishable from the actual state $(w, z)$, where each such state $(w', z')$ consists of a possible world $w'$, a trace $z'$ that is executable $w'$, and such that the formula $\alpha$ is satisfied in $(w', z')$.
We will later use compatible states to define the semantics of the belief operator $\belconnector$.

Before defining the semantics of belief, we first need \emph{epistemic states}, which assign probabilities to worlds:
\begin{definition}[Epistemic state]
  A distribution is a mapping $\worlds \rightarrow \mathbb{R}^{\geq 0}$.
  An epistemic state is any set of distributions.
\end{definition}
As in \bhl and \ds, it is possible to permit the agent to entertain any set of initial distributions.
As an example, the initial theory could say that $\bel{p}{0.5} \vee \bel{p}{0.6}$, which says that the agent is not sure about the distribution of $p$.
In this case, there would be at least two distributions in the epistemic state $e$, one satisfying $\bel{p}{0.5}$ and one satisfying $\bel{p}{0.6}$.
As another example, if we say $\bel{p \vee q}{1}$, then this says that the disjunction is believed with probability $1$, but it does not specify the probability of $p$ or $q$, resulting in infinitely many distributions that are compatible with this constraint.
Thus, not committing to a single distribution results in higher expressivity in the representation of uncertainty.

In order to compute the belief in some formula $\alpha$, we will need to determine the normalized weight of a set of worlds $\mathcal{V}$ in relation to the set of all worlds $\mathcal{W}$ according to a distribution $d$.
While summing over uncountably many worlds is impossible, \textcite{belleFirstorderLogicProbability2016} have shown that if the set of worlds with non-zero weights is countable, we may obtain a well-defined notion of normalization:
\begin{definition}[Normalization]
  ~ \\
  For any distribution $d$ and any set $\mathcal{V} = \left\{\left(w_1, z_1\right), \left(w_2, z_2\right), \ldots \right\}$, we define:
  \begin{enumerate}
    \item $\bnd\mleft(d, \mathcal{V}, r\mright)$ iff there is no $k$ such that
      \[
        \sum_{i=1}^k d\mleft(w_i\mright) \times l^*\mleft(w_i, z_i\mright) > r
      \]
    \item $\eq\mleft(d, \mathcal{V}, r\mright)$ iff $\bnd\mleft(d, \mathcal{V}, r\mright)$ and there is no $r' < r$ such that $\bnd\mleft(d, \mathcal{V}, r'\mright)$ holds.
    \item For any $\mathcal{U} \subseteq \mathcal{V}$: $\norm\mleft(d, \mathcal{U}, \mathcal{V}, r\mright)$ iff $\exists b \neq 0$ such that $\eq\mleft(d, \mathcal{U}, b \times r\mright)$ and $\eq\mleft(d, \mathcal{V}, b\mright)$.
      \qedhere
  \end{enumerate}
\end{definition}
Given $\norm(d, \mathcal{V}, r)$, $r$ can be understood as the normalization of the weights of worlds in $\mathcal{V}$ in relation to the set of all worlds $\mathcal{W}$ with respect to distribution $d$.
The conditions \bnd and \eq are auxiliary conditions to define \norm, where $\bnd(d, \mathcal{V}, r)$ states that the weight of worlds in $\mathcal{V}$ is bounded by $b$ and $\eq(d, \mathcal{V}, r)$ expresses that the weight of worlds in $\mathcal{V}$ is equal to $b$.
\textcite{belleFirstorderLogicProbability2016} have shown that although the set of worlds $\mathcal{W}$ is in general uncountable, this leads to a well-defined summation over the weights of worlds.

To simplify notation,
we also write $\norm(d, \mathcal{U}, \mathcal{V}) = r$ for $\norm(d, \mathcal{U}, \mathcal{V}, r)$.
Furthermore, we write $\norm(d_1, \mathcal{U}_1, \mathcal{V}_1) = \norm(d_2, \mathcal{U}_2, \mathcal{V}_2)$ if there is an $r$ such that
$\norm(d_1, \mathcal{U}_1, \mathcal{V}_1, r)$ and $\norm(d_2, \mathcal{U}_2, \mathcal{V}_2, r)$.
Finally, we write
\[
  \norm(d, \mathcal{U}_1, \mathcal{V}) + \norm(d, \mathcal{U}_2, \mathcal{V}) = r
\]
if
$\norm(d, \mathcal{U}_1, \mathcal{V}, r_1)$,
$\norm(d, \mathcal{U}_2, \mathcal{V}, r_2)$, and $r = r_1 + r_2$.

We continue with the program transition semantics, which defines the traces resulting from executing some program.
The transition semantics is defined in terms of \emph{configurations} $\la z, \delta \ra$, where $z$ is a trace describing the actions executed so far and $\delta$ is the remaining program.
In some places, the transition semantics refers to the truth of formulas (see \autoref{def:dsg-truth} below).%
\footnote{As above, although they depend on each other, the semantics is well-defined, as the transition semantics only refers to static formulas which may not contain programs.}%

\begin{definition}[Program Transition Semantics]\label{def:dsg-trans}
  The transition relation \ewarrow{} among configurations, given
  an epistemic state $e$ and a world $w$, is the least set satisfying
  \begin{enumerate}
    \item $\la z, a\ra \ewarrow \la z \cdot a, \nil \ra$ if $w, z \models \poss(a)$
    \item $\la z, \delta_1;\delta_2 \ra \ewarrow
      \la z \cdot a, \gamma;\delta_2 \ra$,
      if $\la z,\delta_1 \ra \ewarrow \la z \cdot a, \gamma \ra$,
    \item $\la z, \delta_1;\delta_2 \ra \ewarrow \la z \cdot a, \delta' \ra$
      if $\la z, \delta_1 \ra \in \ewfinal$ and
      $\la z, \delta_2 \ra \ewarrow \la z \cdot a, \delta' \ra$
    \item $\la z, \delta_1 \vert \delta_2 \ra \ewarrow \la z \cdot a, \delta' \ra$
      if $\la z, \delta_1 \ra \ewarrow \la z \cdot a, \delta' \ra$
      or $\la z, \delta_2 \ra \ewarrow \la z \cdot a, \delta' \ra$
    \item $\la z, \pi x.\, \delta \ra \ewarrow \la z \cdot a, \delta' \ra$,
      if $\la z, \delta^x_r \ra \ewarrow \la z \cdot a, \delta' \ra$ for some
      $r \in \rigid$
    \item $\la z, \delta^* \ra \ewarrow \la z \cdot a, \gamma; \delta^* \ra$ if
      $\la z, \delta \ra \ewarrow \la z \cdot a, \gamma \ra$
  \end{enumerate}
  The set of final configurations \ewfinal{} is the smallest set such that
  \begin{enumerate}
    \item $\la z, \alpha? \ra \in \ewfinal$ if $e, w, z \models \alpha$,
    \item $\la z, \delta_1;\delta_2 \ra \in \ewfinal$
      if $\la z, \delta_1 \ra \in \ewfinal$ and $\la z, \delta_2 \ra \in \ewfinal$
    \item $\la z, \delta_1 \vert \delta_2 \ra \in \ewfinal$
      if $\la z, \delta_1 \ra \in \ewfinal$,
      or $\la z, \delta_2 \ra \in \ewfinal$
    \item $\la z, \pi x.\,\delta \ra \in \ewfinal$
      if $\la z, \delta^x_r \ra \in \ewfinal$ for some $r \in \rigid$
    \item $\la z, \delta^* \ra \in \ewfinal$
      \qedhere
  \end{enumerate}
\end{definition}
We also write $\ewarrow^*$ for the transitive closure of $\ewarrow$.
For a primitive action $a$, the interpreter may take a transition if $a$ is currently possible, after which the remaining program is the empty program \nil.
For a sequence of sub-programs $\delta = \delta_1; \delta_2$, the interpreter may take a transition following $\delta_1$, in which case the remaining program $\gamma; \delta_2$ is the remaining program $\gamma$ after taking the transition in $\delta_1$, concatenated by the unchanged program $\delta_2$.
Alternatively, it may take a transition following $\delta_2$ if $\delta_1$ is final in the current configuration, in which case the remaining program is simply the remaining program of $\delta_2$ after taking the transition.
In the case of nondeterministic branching $\delta_1 \vert \delta_2$, it may follow the transitions of the first or the second sub-program such that the remaining program is the remaining program of the taken sub-program.
For the nondeterministic pick operator $\pi x.\, \delta$, it may follow any transition that results from the program $\delta^x_r$, where $x$ is substituted by some ground term $r$.
Finally, for nondeterministic iteration $\delta^*$, the interpreter may take the same transitions as $\delta$ (i.e., continue with another iteration).

For the final configurations, atomic tests $\alpha?$ are final if $\alpha$ is satisfied in the current configuration.
The sequence of sub-programs $\delta_1; \delta_2$ is final if both sub-programs are final.
For nondeterministic branching, the program $\delta_1 \vert \delta_2$ is final if either sub-program is final.
Similarly, for $\pi x.\, \delta$, the program is final if it is final for some substitution of $x$.
Nondeterministic iteration $\delta^*$ is final, i.e., the interpreter may always decide to stop (and not continue with the next iteration).

The transition semantics of \dsg are similar to those of \esg and also similar to the action steps of \tesg.
The main difference is that the relation also depends on the epistemic state $e$ because tests may use the epistemic operators \belconnector and \know.

Following the transition semantics for a given program $\delta$, we obtain a set of \emph{program traces}:
\begin{definition}[Program Traces]
  ~\\
  Given an epistemic state $e$, a world $w$, and a trace $z$, the set $\|\delta\|^z_{e,w}$ of traces of program $\delta$ is defined as the following set:
  \[
    \|\delta\|^z_{e, w} =
    \{ z' \in \mathcal{Z} \mid
      \la z, \delta \ra \ewarrow^* \la z \cdot z', \delta' \ra
    \text{ and } \la z \cdot z', \delta' \ra \in \ewfinal \}
    \qedhere
  \]
\end{definition}

This transition semantics is similar to \esg and also similar to the action steps of \tesg.
Compared to \esg, this transition semantics also refers to the epistemic state $e$, as test formulas can also mention belief operators.
Additionally, in contrast to \esg and \tesg, it only allows a transition for an atomic action if the action is possible in the current state.
Furthermore, while \esg and \tesg allow infinite traces, we only allow finite traces, as we do not include temporal formulas in the logic.

Finally, 
we can define the semantics for \dsg formulas:
\begin{definition}[Truth of Formulas]\label{def:dsg-truth}
  Given an epistemic state $e$, a world $w$, and a formula $\alpha$,
  we define for every $z \in \mathcal{Z}$:
  \begin{enumerate}
    \item $e, w, z \models F\mleft(t_1, \ldots, t_k\mright)$ iff $w \mleft\lbrack F\mleft(t_1, \ldots, t_k\mright), z \mright\rbrack = 1$
    \item $e, w, z \models \bel{\alpha}{r}$
      iff $\forall d \in e:\: \norm\mleft(d, \stateset_\alpha, \stateset_\true, r\mright)$
    \item $e, w, z \models \left(t_1 = t_2\right)$ iff $t_1$ and $t_2$ are identical
    \item $e, w, z \models \alpha \wedge \beta$ iff $e, w, z \models \alpha$ and
      $e, w, z \models \beta$
    \item $e, w, z \models \neg \alpha$ iff $e, w, z \not\models \alpha$
    \item $e, w, z \models \forall x.\, \alpha$ iff $e, w, z \models \alpha^x_r$ for
      all $r \in \rigid$.
    \item $e, w, z \models \square \alpha$ iff $e, w, z \cdot z' \models \alpha$
      for all $z' \in \mathcal{Z}$
    \item $e, w, z \models [\delta]\alpha$ iff
     $e, w, z \cdot z' \models \alpha$
     for all $z' \in \|\delta\|^z_{e,w}$.
     \label{def:dsg-truth:program}
     \qedhere
  \end{enumerate}
\end{definition}

Note in particular that Item 2 states that the degree of belief in a formula is obtained by looking at the normalized weight of the possible worlds that satisfy the formula.

We write $e, w \models \alpha$ for $e, w, \la\ra \models \alpha$.
Also, if $\alpha$ is objective, we write $w, z \models \alpha$ for $e, w, z \models \alpha$ and $w \models \alpha$ for $w, \la\ra \models \alpha$.
Additionally, for a set of sentences $\Sigma$, we write $e, w, z \models \Sigma$ if $e, w, z \models \phi$ for all $\phi \in \Sigma$, and $\Sigma \models \alpha$ if $e, w \models \Sigma$ entails $e, w \models \alpha$ for every model $\left(e, w\right)$.

\subsection{Basic Action Theories}\label{sec:dsg-bat}
A \acf{BAT} defines the effects of all actions of the domain, as well as the initial state:
\begin{definition}[Basic Action Theory]
  Given a finite set of predicates $\mathcal{F}$ including \oi and $l$, a set $\Sigma$
  of sentences is called a \acf{BAT} over $\mathcal{F}$ iff
  $\Sigma = \Sigma_0 \cup \Sigma_\text{pre} \cup \Sigma_\text{post}$, where
  $\Sigma$ mentions only fluent predicates in $\mathcal{F}$ and
  \begin{enumerate}
    \item $\Sigma_0$ is any set of fluent sentences,
    \item $\Sigma_\text{pre}$ consists of a single sentence
      of the form $\square
      \poss(a) \equivspace \pi$, where $\pi$ is a fluent formula with free variable $a$,%
      \footnote{We assume that free variables are universally quantified from the outside, $\square$ has lower syntactic precedence than the logical connectives, and $[\cdot]$ has the highest priority, so that $\square \poss(a) \equiv \pi$ stands for $\forall a.\, \square \left(\poss(a) \equiv \gamma\right)$ and $\square \poss(a) \supset \left([a]F(\vec{x}) \equivspace \gamma_F\right)$ stands for $\forall a,\vec{x}.\, \square \left(\poss(a) \supset \left([a]F(\vec{x}) \equivspace \gamma_F\right)\right)$.}
    \item $\Sigma_\text{post}$ is a set of sentences, one for each fluent predicate $F \in \mathcal{F}$, of the form $\square \poss(a) \supset \left([a]F(\vec{x}) \equivspace \gamma_F\right)$, and where $\gamma_F$ is a fluent formula with free variables among $a$ and $\vec{x}$.
      \qedhere
  \end{enumerate}
\end{definition}
Given a \ac{BAT} $\Sigma$, we say that a program $\delta$ is a program over $\Sigma$ if it only mentions fluents and actions from $\Sigma$.

Note that the successor state axioms slightly differ from the successor state axioms in \autoref{sec:tesg-bat}, where they have the form $\square [a]F(\vec{x}) \equivspace \gamma_F$.
In contrast to before, the successor state axioms in \dsg \acp{BAT} only define the effects of an action if the action is currently possible and otherwise do not make any statement about the action effects.
This is necessary because we include $\poss(a)$ in the transition semantics (\autoref{def:dsg-trans}).
To understand why it is necessary, consider the following example: if $w, z \models \neg \poss(a)$, then by \subdefref{def:dsg-truth}{def:dsg-truth:program}, $w, z \models [a] \neg F$ is vacuously true for any $0$-ary fluent $F$ because there is no trace $z' \in \|a\|^z_{e,w}$.
This would be contradicting a successor state axiom $\square [a] F \equiv \gamma_F$ (unless $\square \gamma_F \equiv \neg \true$).
Restricting the successor state axiom to possible actions avoids this issue.%
\footnote{
  In \autoref{chap:timed-esg}, we used a different solution based on \parencite{classenPlanningVerificationAgent2013} by allowing an action transition even if the action is impossible and then augmenting the program by guarding each action with a test $\poss(a)?$.
  Here, we prefer the presented solution where the transition semantics only allows actions that are actually possible without augmenting the program, mainly because it will simplify the definition of bisimulation and subsequent proofs in \autoref{sec:bisimulation}.
}

\subsubsection{A Noisy Basic Action Theory}

\begin{figure}[htb]
  \centering
  \includegraphics[width=0.5\textwidth]{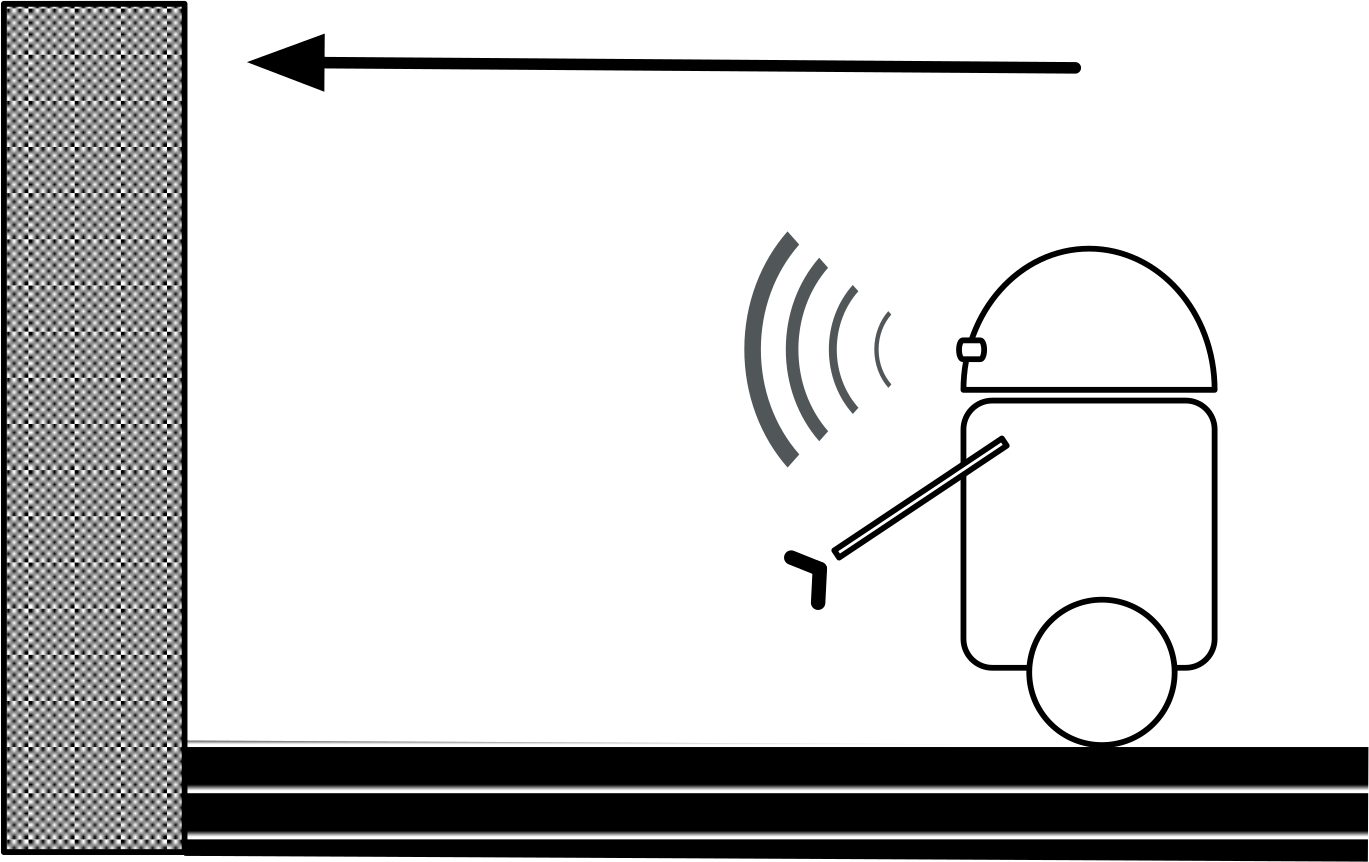}
  \caption[A robot driving towards a wall.]{A robot driving towards a wall~\cite{belleReasoningProbabilitiesUnbounded2017}, which is the same example as in \autoref{fig:robot-wall}.
  The robot can measure the distance to the wall with its action \sonar and it can move towards the wall with the action \move.
  Both actions are noisy: the sonar does not measure the exact distance and the \move action may move with further or shorter than intended.
}
  \label{fig:abstraction-robot-wall}
\end{figure}

We present a \ac{BAT} for a simple robotics scenario with noisy actions, inspired from \cite{belleReasoningProbabilitiesUnbounded2017}.
In this scenario, a robot moves towards a wall and it is equipped with a sonar sensor that can measure the distance to the wall, as shown in \autoref{fig:abstraction-robot-wall}.
A \ac{BAT} \sigmal defining this scenario may look as follows:
\begin{itemize}[left=0pt .. \parindent]
  \item
    A \move action is possible if the robot moves either one step to the back or to the front.
    A \sonar action is always possible:
    \begin{align*}
      \square \poss(a) \equivspace
      & \exists x,y \left(a = \move(x,y) \wedge \left(x = 1 \vee x = -1\right)\right)
      \vee \exists z \left(a = \sonar(z)\right)
    \end{align*}
  \item After doing action $a$, the robot is at position $x$ if $a$ is a \move action that moves the robot to location $x$ or if $a$ is not \move and the robot was at location $x$ before:
    \begin{align*}
      \square \poss(a) \supset \big(\lb a \rb \loc(x) \equivspace
        {}&\exists y,z \left(a = \move(y,z) \wedge \loc(l) \wedge x = l + z \right)
      \\
      & \vee \neg \exists y, z \left(a = \move(y,z)\right) \wedge \loc(x)\big)
    \end{align*}
  \item For the \sonar action, the likelihood that the robot measures the correct distance is $0.8$, the likelihood that it measures a distance with an error of $\pm 1$ is $0.1$.
    Furthermore, for the \move action, the likelihood that the robot moves the intended distance $x$ is $0.6$, the likelihood that the actual movement $y$ is off by $\pm 1$ is $0.2$:
    \begin{align*}
      \square l(a, u) \equivspace
      {}& \exists z \left( a = \sonar(z) \wedge \loc(x) \wedge u = \Theta(x, z, .8, .1)\right)
      \\
      &  \vee \exists x,y \left(a = \move(x, y) \wedge u = \Theta(x, y, .6, .2)\right)
      \\
      &  \vee \neg \exists x,y,z \left(a = \move(x,y) \vee a = \sonar(z)\right) \wedge u = .0
    \end{align*}
    where $\Theta(u,v,c,d) =
      \begin{cases} c & \text{ if } u = v \\ d & \text{ if } \vert u-v \vert = 1 \\ 0 & \text{ otherwise} \end{cases}$.
  \item  The robot cannot detect the distance that it has actually moved, i.e., any two actions $\move(x, y)$ and $\move(x, z)$ are o.i.:
    \[
      \square \oi(a, a') \equivspace a = a' \vee \exists x, y, z \left(a = \move(x, y) \wedge a' = \move(x, z)\right)
    \]
  \item Initially, the robot is three units away from the wall:
    \[
      \loc(x) \equivspace x = 3
    \]
\end{itemize}

Based on this \ac{BAT}, we define a program that first moves the robot close to the wall and then back:\footnote{The unary $\move(x)$ can be understood as abbreviation $\move(x) \eqdef \pi y\, \move(x, y)$, where nature nondeterministically picks the distance $y$ that the robot really moved (similarly for $\sonar()$).}
\begin{align*}
   &\sonar();
   \\
  & \gwhile \neg \know{\exists x \left(\loc(x) \wedge x \leq 2\right)}
  \gdo \move(-1); \sonar() \gdone;
  \\ &\gwhile \neg \know{\exists x \left(\loc(x) \wedge x > 5\right)}
  \gdo \move(1); \sonar() \gdone
\end{align*}

The robot first measures its distance to the wall and then moves closer until it knows that its distance to the wall is less than two units.
Afterwards, it moves away until it knows that is more than five units away from the wall.
As the robot's \move action is noisy, each \move is followed by \sonar to measure how far it is away from the wall.
One possible execution trace of this program may look as follows:
\begin{align}
  z_l = \langle &\sonar(3), \move(-1, 0), \sonar(3), \move(-1, -1),
  \nonumber \\ &
  \sonar(2), \move(-1, -1), \sonar(1), \move(1, 1),
  \nonumber \\ &
  \sonar(3), \move(1, 1),
  \sonar(2), \move(1, 1), \nonumber
  \\ &
  \sonar(4), \move(1, 0),
  \sonar(4), \move(1, 1), \sonar(6) \label{eqn:low-level-trace}
  \rangle
\end{align}
First, the robot (correctly) senses that it is three units away from the wall and starts moving.
However, the first \move does not have the desired effect: the robot intended to move by one unit but actually did not move (indicated by the second argument being $0$).
After the second \move, the robot is  at $\loc(2)$, as it started at $\loc(3)$ and moved successfully once.
However, as its sensor is noisy and it measured $\sonar(2)$, it believes that it could also be at $\loc(3)$.
For safe measure, it executes another \move and then senses $\sonar(1)$, after which it knows for sure that it is at most two units away from the wall.
In the second part, the robot moves back until it knows that it has reached a distance further than five units away from the wall.
As this simple example shows, the trace $z_l$ is already quite hard to understand.
While it is clear from the \ac{BAT} what each action does, the robot's intent is not immediately obvious and the trace is cluttered with noise and low-level details.

\subsubsection{An Abstract Basic Action Theory}
We present a second, more abstract \ac{BAT} for the same scenario but without noisy actions:
\begin{itemize}[left=0pt .. \parindent]
  \item After doing action $a$, the robot is at location $l$ if $a$ is the action $\goto(l)$ or if $a$ is no \goto action and the robot has been at $l$ before:%
    \footnote{For the sake of simplicity, we only allow the robot to go to \near or \far and omit the location \midpos.}
    \[
      \square \poss(a) \supset
      \big(\lb a \rb \at(l) \equivspace
      a = \goto(l) \vee \neg \exists x \left(a = \goto(x)\right) \wedge \at(l)\big)
    \]
  \item The action likelihood axiom states that the robot may only \goto the locations \near or \far and that the action is not noisy: 
    \begin{align*}
      \square l(a, u) \equivspace &(a = \goto(\near) \vee a = \goto(\far)) \wedge u = 1.0
      \\
              &\vee \neg (a = \goto(\near) \vee a = \goto(\far)) \wedge u = 0.0
    \end{align*}
  \item The agent can distinguish all actions:
    \[
      \square \oi(a, a') \equivspace a = a'
    \]
  \item Initially, the robot is in the middle:
    \[
      \at(l) \equiv l = \midpos
    \]
\end{itemize}

In the next section, we will show how we can connect the low-level \ac{BAT} \sigmal with the high-level \ac{BAT} \sigmah by using \emph{abstraction}.


\section{Bisimulation}\label{sec:bisimulation}
In this section, we define an abstraction of a low-level \ac{BAT} $\Sigma_l$ by a high-level \ac{BAT} $\Sigma_h$.
This will allow us to construct abstract \golog programs over the high-level \ac{BAT}, which are equivalent and can be translated to some program over the low-level \ac{BAT}.
We do so by mapping the high-level \ac{BAT} to the low-level \ac{BAT} by means of a \emph{refinement mapping}.
Based on the mapping, we can then define two notions of isomorphism: In \emph{objective isomorphism}, two states are isomorphic if they satisfy the same (objective) atomic formulas.
To deal with the epistemic state, we also introduce \emph{epistemic isomorphism}, which intuitively relates the probability of a high-level state to a probability of a set of low-level states.
These isomorphisms are \emph{local properties} in the sense that they relate fixed world and epistemic states respectively.
In order to extend this to a dynamic setting, we then define a notion of \emph{bisimulation}.
Intuitively, for every possible transition of the high-level program, there must be a corresponding step of the low-level program that simulates the high-level step, i.e., it results in a state that is again similar to the resulting high-level state, and vice versa.

For the sake of simplicity,\footnote{The technical results do not hinge on this, but allowing arbitrary epistemic states would make the main results and proofs more tedious. For the general case, we need to set up for every distribution on the high level a corresponding distribution on the low level and establish a bisimulation for each of those pairs.} we assume in the following that an epistemic state $e$ is always a singleton, i.e., $e_h = \{ d_h \}$ and $e_l = \{ d_l \}$.
In order to define an abstraction of $\bat_l$, we  translate the high-level \ac{BAT} $\Sigma_h$ into the low-level \ac{BAT} $\Sigma_l$ by mapping each high-level fluent of $\Sigma_h$ to a low-level formula of $\Sigma_l$, and every high-level action of $\Sigma_h$ to a low-level program of $\Sigma_l$:
\begin{definition}[Refinement Mapping]
  Given two basic action theories $\Sigma_l$ over $\mathcal{F}_l$ and $\Sigma_h$ over $\mathcal{F}_h$. The function $m$  is a \emph{refinement mapping} from $\Sigma_h$ to $\Sigma_l$ iff:
  \begin{enumerate}
    \item For every action $a(\vec{x})$ mentioned in $\Sigma_h$, $m\mleft(a\mleft(\vec{x}\mright)\mright) = \delta_a\mleft(\vec{x}\mright)$, where $\delta_a\mleft(\vec{x}\mright)$ is a Golog program over the low-level theory $\Sigma_l$ with free variables among $\vec{x}$.
    \item For every fluent predicate $F \in \mathcal{F}_h$, $m\mleft(F(\vec{x})\mright) = \phi_F\mleft(\vec{x}\mright)$, where $\phi_F\mleft(\vec{x}\mright)$ is a static formula over $\mathcal{F}_l$ with free variables among $\vec{x}$.
      \qedhere
  \end{enumerate}
\end{definition}

For a formula $\alpha$ over $\mathcal{F}_h$, we also write $m(\alpha)$ for the formula obtained by applying $m$ to each fluent predicate and action mentioned in $\alpha$.
For a trace $z = \la a_1, a_2, \ldots \ra$ of actions from $\Sigma_h$, we also write $m(z)$ for $\la m(a_1), m(a_2), \ldots \ra$.
For a program $\delta$ over $\Sigma_h$, the program $m(\delta)$ is the same program as $\delta$ with each primitive action $a$ replaced by $m(a)$ and each formula $\alpha$ replaced by $m(\alpha)$.

Continuing our example, we define a refinement mapping that maps \sigmah to \sigmal by mapping each high-level fluent to a low-level formula and each high-level action to a low-level program:
\begin{itemize}[left=0pt .. \parindent]
  \item The high-level fluent $\at(l)$ is mapped to a low-level formula by translating the distance to the locations \near, \midpos, and \far:
    \begin{align*}
      \at(l) \:\mapsto\:{}
      &l = \near \wedge \exists x \left(\loc(x) \wedge x \leq 2\right)
      \\
      & \vee l = \midpos \wedge \exists x \left(\loc(x) \wedge  x > 2 \wedge x \leq 5\right)
      \\
       &\vee l = \far \wedge \exists x \left(\loc(x) \wedge x > 5\right)
    \end{align*}
  \item The action $\goto$ is mapped to a program that guarantees that the robot reaches the right position:
    \begin{align*}
      \goto(x) \:\mapsto\:{}
       &
       \sonar();
       \\
       & \gif x = \near \gthen \\
       & \; \gwhile \neg \know{\exists x \left(\loc(x) \wedge x \leq 2\right)}
       \gdo \move(-1); \sonar() \gdone
       \\
       & \gelif x = \far \gthen \\
       & \; \gwhile \neg \know{\exists x \left(\loc(x) \wedge x > 5\right)}
       \gdo \move(1); \sonar() \gdone
       \\ &\gfi
    \end{align*}
\end{itemize}

To show that a high-level \ac{BAT} indeed abstracts a low-level \ac{BAT}, we first define a notion of isomorphism, intuitively stating that two states satisfy the same fluents:
\begin{definition}[Objective Isomorphism]\label{def:oiso}
  ~ \\
  We say $\left(w_h,z_h\right)$ is objectively $m$-isomorphic to $\left(w_l,z_l\right)$, written $\left(w_h,z_h\right) \oiso \left(w_l,z_l\right)$ iff
      for every atomic formula $\alpha$ mentioned in $\Sigma_h$:
      \[
        w_h, z_h \models \alpha \textrm{ iff } w_l, z_l \models m\mleft(\alpha\mright)
        \qedhere
      \]
\end{definition}

Additionally, because we need to relate degrees of belief, we need to connect the two \acp{BAT} in terms of epistemic states.
To do so, we define epistemic isomorphism as follows:
\begin{definition}[Epistemic Isomorphism]\label{def:eiso}
  ~ \\
  For every $(w_h, z_h) \in \states$ and $\stateset_l \subseteq \states$,
  we say that $\left(d_h,w_h,z_h\right)$ is epistemically $m$-isomorphic to $\left(d_l, \stateset_l\right)$, written $\left(d_l,w_h,z_h\right) \eiso \left(d_l,\stateset_l\right)$ iff
  for the partition $P = \stateset_l / \oicomp$, for each $\stateset_l^i \in P$ and $\left(w_l^i, z_l^i\right) \in \stateset_l^i$:
    \[
      \norm(d_h, \{ \left(w_h, z_h\right) \}, \wh{\true})
      = \norm(d_l, \stateset_l^i, \wl[^i]{\true})
      \qedhere
    \]
\end{definition}

The intuition of epistemic isomorphism is as follows: As the high-level state $\left(w_h, z_h\right)$ is more abstract than the low-level state $\left(w_l, z_l\right)$, multiple low-level states may be isomorphic to the same high-level state.
Therefore, each high-level state is mapped to a set of low-level states.
To be epistemically isomorphic, they must entail the same beliefs, therefore, the corresponding normalized weights must be equal.
However, we do not require the low-level states $\stateset_l$ to be observationally indistinguishable.
Indeed, since we will have a high-level action corresponding to many low-level actions, almost always low-level states will not be observationally indistinguishable.
Therefore, we first partition $\stateset_l$ according to \oicomp and then require the \norm over $\left(w_h, z_h\right)$ to be the same as the \norm over each member of the partition.

\begin{figure}[thb]
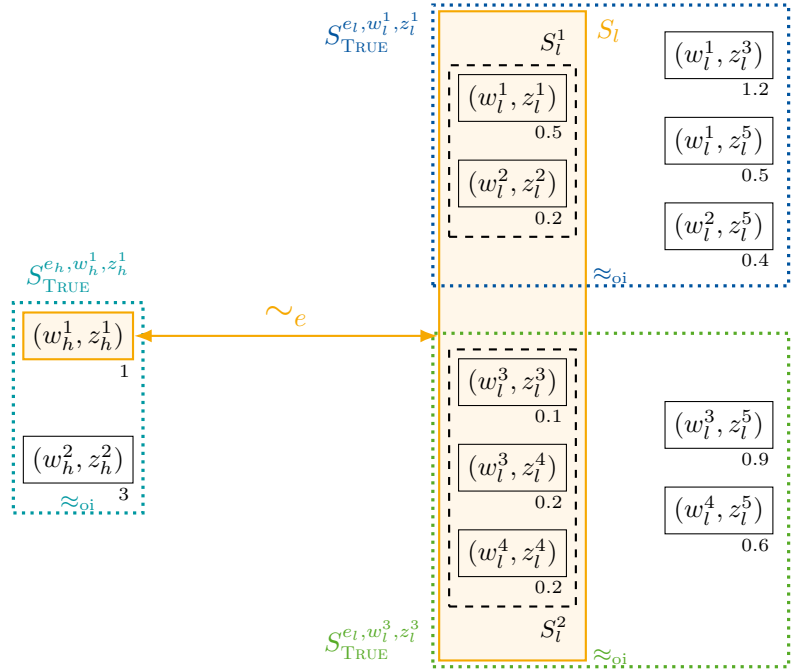

  \centering
  \includestandalone{figures/epistemic-isomorphism}
  \caption[An example for epistemic isomorphism.]{
    An example for epistemic isomorphism $(d_h, w_h^1, z_h^1) \eiso (d_l, S_l)$.
    Each node is labeled with the weight $d(w) \times l^*(w, z)$, e.g., $d_h(w_h^1) \times l^*(w_h^1, z_h^1) = 1$.
    Dotted boxes mark the equivalence classes wrt $\oicomp$, dashed boxes mark the members of the partition $P = S_l / \oicomp$.
  }
  \label{fig:epistemic-isomorphism}
\end{figure}

\autoref{fig:epistemic-isomorphism} illustrates epistemic isomorphism.
On the left-hand side, we have the high-level state $(w_h^1, z_h^1)$ and a second high-level state $(w_h^2, z_h^2)$ that is observationally indistinguishable from $(w_h^1, z_h^1)$.
On the right-hand side, we can see that the low-level states are partitioned by $\oicomp$ into two sets, $\stateset^{w_l^1, z_l^1}_\true$, which are the states compatible to $(w_l^1, z_l^1)$, and $\stateset^{w_l^3, z_l^3}_\true$, which are the states compatible to $(w_l^3, z_l^3)$.
Vertically aligned in the center is the set $S_l$, which is also partitioned into $S_l^1$ and $S_l^2$.
For both $S_l^1$ and $S_l^2$, the normalized weight is equal to the normalized weight of $(w_h^1, z_h^1)$, which is why $(d_h, w_h^1, z_h^1)$ is indeed epistemically isomorphic to $(d_l, S_l)$.
As an example, for $S_l^1$, we obtain:
\begin{align*}
  \norm(d_l, \stateset_l^1,  \wl[^1]{\true}) = &\frac{0.5 + 0.2}{0.5 + 0.2 + 1.2 + 0.5 + 0.4} \\
  = &\frac{1}{4} = \frac{1}{1 + 3} = \norm(d_h, \{ (w_h, z_h)\}, \wh{\true})
\end{align*}

Having established objective and epistemic isomorphisms, we can now define a suitable notion of bisimulation:
\begin{definition}[Bisimulation]
  \label{def:bisim}
  ~ \\
  A relation
  $B \subseteq \states \times \states$
  is an \emph{$m$-bisimulation between $\left(e_h, w_h\right)$ and $\left(e_l, w_l\right)$} if $\left(\left(w_h, z_h\right), \left(w_l, z_l\right)\right) \in B$ implies that
  \begin{enumerate}
    \item \label{def:bisim:oiso}
       $\left(w_h, z_h\right) \oiso \left(w_l, z_l\right)$,
    \item \label{def:bisim:eiso}
      $\left(d_h, w_h, z_h\right) \eiso \left(d_l, \left\{ \left(w_l', z_l'\right) \mid \left(\left(w_h, z_h\right), \left(w_l', z_l'\right)\right) \in B \right\} \right)$,
    \item \label{def:bisim:executability}
      $w_h \models \exec(z_h)$ and $w_l \models \exec(z_l)$,
    \item \label{def:bisim:high-action}
      for every high-level action $a$, if $w_h, z_h \models \poss(a)$, then there is $z'_l \in \|m(a)\|^{z_l}_{e_l,w_l}$ such that $((w_h, z_h \cdot a), (w_l, z_l \cdot z_l')) \in B$,
    \item \label{def:bisim:low-action}
        for every high-level action $a$, if there is $z'_l \in \|m(a)\|^{z_l}_{e_l,w_l}$, then $w_h, z_h \models \poss(a)$ and $((w_h, z_h \cdot a), (w_l, z_l \cdot z_l')) \in B$,
      \item \label{def:bisim:high-epistemic-state}
      for every $\left(w_h', z_h'\right) \oicomp \left(w_h, z_h\right)$
      with $d_h(w_h') > 0$ and $e_h, w_h' \models \exec(z_h')$,
      there is $\left(w_l', z_l'\right) \oicomp \left(w_l, z_l\right)$
      such that 
      $\left(\left(w_h', z_h'\right), \left(w_l', z_l'\right)\right) \in B$,
    \item \label{def:bisim:low-epistemic-state}
      for every $\left(w_l', z_l'\right) \oicomp \left(w_l, z_l\right)$
      with $d_l(w_l') > 0$ and $e_l, w_l' \models \exec(z_l')$,
      there is $\left(w_h', z_h'\right) \oicomp \left(w_h, z_h\right)$
      such that 
      $\left(\left(w_h', z_h'\right), \left(w_l', z_l'\right)\right) \in B$.
  \end{enumerate}

  We call a bisimulation $B$ \emph{definite} if $\left(\left(w_h, z_h\right), \left(w_l, z_l\right)\right) \in B$ and $\left(\left(w_h', z_h'\right), \left(w_l, z_l\right)\right) \in B$ implies $\left(w_h, z_h\right) = \left(w_h', z_h'\right)$.

We say that $\left(e_h,w_h\right)$ is bisimilar to $\left(e_l,w_l\right)$ relative to refinement mapping $m$, written $\left(e_h,w_h\right) \bisim \left(e_l,w_l\right)$, if and only if there exists a definite $m$-bisimulation relation $B$ between $\left(e_h, w_h\right)$ and $\left(e_l, w_l\right)$ such that $\left(\left(w_h, \la\ra\right), \left(w_l, \la\ra\right)\right) \in B$.
\end{definition}

The general idea of bisimulation is that two states are bisimilar if they have the same local properties (i.e., they are isomorphic) and each reachable state from the first state has a corresponding reachable state from the second state (and vice versa) such that the two successors are again bisimilar.
Here,
properties \ref{def:bisim:oiso}, \ref{def:bisim:eiso}, and \ref{def:bisim:executability} refer to static properties of $\left(w_h, z_h\right)$ and $\left(w_l, z_l\right)$.
While property \ref{def:bisim:oiso} directly establishes objective isomorphism of $\left(w_h, z_h\right)$ and $\left(w_l, z_l\right)$, property \ref{def:bisim:eiso} establishes epistemic isomorphism between $\left(w_h, z_h\right)$ and all states $\left(w_l', z_l'\right)$ that occur in $B$.
As usual in bisimulations, we also require that if we follow a high-level transition of the system, there is a corresponding low-level transition (and vice versa).
Here, such a transition may either be an action that is executed (properties \ref{def:bisim:high-action} and \ref{def:bisim:low-action}), or it may be an epistemic transition from the current state to another observationally indistinguishable state (properties \ref{def:bisim:high-epistemic-state} and \ref{def:bisim:low-epistemic-state}).

A \emph{definite} bisimulation is a bisimulation where no two high-level states are mapped to the same low-level state (note that the converse is allowed).
This is necessary when we want to show that high-level and low-level epistemic states entail the same beliefs: We will sum over all observationally indistinguishable states that satisfy some formula;
if we allow the same low-level state to be mapped to two different high-level states, then the sum over the high-level states will result in a different weight than the sum over the low-level states, as both high-level states contribute to the sum while the low-level state is considered only once, therefore entailing different degrees of belief.
In a sense, this captures the idea that the high-level state is more abstract than the low-level states: While each high-level state may be mapped to multiple low-level states, there cannot be two different abstract states for the same low-level state.

Our notion of bisimulation is similar to bisimulation for abstracting non-stochastic and objective basic action theories, as described by \textcite{banihashemiAbstractionSituationCalculus2017}.
In comparison, the notion of objective isomorphism (property \ref{def:bisim:oiso}) and reachable states via actions (properties \ref{def:bisim:high-action} and \ref{def:bisim:low-action}) are analogous, while epistemic isomorphism (property \ref{def:bisim:eiso}) and reachable states via observational indistinguishability (properties \ref{def:bisim:high-epistemic-state} and \ref{def:bisim:low-epistemic-state}) have no corresponding counterparts.

Given a corresponding $m$-bisimulation, we want to show that $\left(e_h, w_h\right)$ is a model of a formula $\alpha$ iff $\left(e_l, w_l\right)$ is a model of the mapped formula $m(\alpha)$.
To do so, we first show that this is true for static formulas, not considering programs.
In the second step, we will show that the high-level and low-level models induce the same program traces, which will then allow us to extend the statement to bounded formulas, which may refer to programs.
We start with static formulas:
\begin{theoremE}\label{thm:static-equivalence}
  Let $\left(e_h,w_h\right) \bisim \left(e_l,w_l\right)$ with definite $m$\hyp{}bisimulation $B$.
  For every static formula $\alpha$ and traces $z_h, z_l$ with $\left(\left(w_h, z_h\right), \left(w_l, z_l\right)\right) \in B$:
  \[
  e_h, w_h, z_h \models \alpha \text{ iff } e_l, w_l, z_l \models m\mleft(\alpha\mright)
  \]
\end{theoremE}
\begin{proofE}[normal][Proof Sketch]
  By structural induction on $\alpha$. The interesting case is $\alpha = \bel{\beta}{r}$. Let
  \begin{align*}
    \norm(d_h, \wh{\beta}, \wh{\true}) &= r_h
    \\
    \norm(d_l, \wl{\beta}, \wl{\true}) &= r_l
  \end{align*}
  We need to show that $r_h = r_l$.
  \\
  \textbf{$\leq$:}
  Let $\left(w_h^i, z_h^i\right) \in \wh{\beta}$.
  We can ignore those $(w_h^i, z_h^i)$ with $d_h(w_h^i) = 0$ because they do not contribute to $r_h$.
  By \subdefref{def:bisim}{def:bisim:high-epistemic-state}, there is a $\left(w_l^i, z_l^i\right)$ with $\left(\left(w_h^i, z_h^i\right), \left(w_l^i, z_l^i\right)\right) \in B$ and $\left(w_l^i, z_l^i\right) \oicomp \left(w_l, z_l\right)$.
  From \subdefref{def:bisim}{def:bisim:eiso} and \autoref{def:eiso}, we know that for each such $\left(w_h^i, z_h^i\right)$, $\left(w_h^i, z_h^i\right)$ is epistemically isomorphic to the union $\stateset_l$ of all bisimilar $\left(w_l', z_l'\right)$.
  Using the partition $P = S_l / \oicomp$, there is $S_l^i \in P$ with $(w_l^i, z_l^i) \in S_l^i$.
  It follows:
  \[
    \norm(d_h, \{(w_h^i, z_h^i)\}, \wh[^i]{\true})
    =
    \norm(d_l, \stateset_l^i, \wl{\true})
  \]
  As $B$ is definite, we can directly take the union of both sides to obtain the overall probability of $\wh{\beta}$:
  \begin{align*}
      \norm(d_h, \wh{\beta}, \wh{\true})
      = \norm(d_l, \bigcup_i \stateset_l^i, \wl{\true})
  \end{align*}
  Furthermore, by induction, for each $\left(w_l', z_l'\right) \in S_l^i$, it follows that $e_l, w_l' \models [z_l'] m(\beta)$ and therefore, $S_l^i \subseteq \wl{\beta}$.
  With that,
  \[
      \norm(d_l, \bigcup_i \stateset_l^i, \wl{\true})
      \leq
      \norm(d_l, \wl{\beta}, \wl{\true})
  \]
  Thus, $r_h \leq r_l$.
  \\
  \textbf{$\geq$:}
  For each $(w_l^i, z_l^i) \in \wl{\beta}$, there is a $(w_h^i, z_h^i)$ such that $((w_h^i, z_h^i), (w_l^i, z_l^i)) \in B$ and such that $(w_h^i, z_h^i)$ is epistemically isomorphic to the union of $\stateset_l$ of all bisimilar $(w_l', z_l')$.
  Let $P = S_l / \oicomp$ and $S_l^i \in P$ with $(w_l^i, z_l^i) \in S_l^i$.
  It can be shown that
  \[
    \norm(d_l, \bigcup_i \stateset_l^i, \wl{\true})
    =
    \norm(d_h, \bigcup_i \left\{\left(w_h^i, z_h^i\right)\right\}, \wh{\true})
  \]
  We can partition $\wl{\beta}$ into $\{ \stateset^1_{m(\beta)}, \stateset^1_{m(\beta)}, \ldots \}$ such that for each $i$, $\stateset^i_{m(\beta)} \subseteq S_l^i$.
  Clearly,
  \begin{align*}
    &\norm(d_l, \bigcup_i \stateset_{m(\beta)}^{i}, \wl{\true}) \nonumber
    \leq
    \norm(d_l, \bigcup_i \stateset_l^i, \wl{\true}) \nonumber
  \end{align*}
  Finally, by induction, $e_h, w_h^i \models [z_h^i] \beta$, thus $(w_h^i, z_h^i) \in \wh{\beta}$, and therefore $\bigcup_i \left\{\left(w_h^i, z_h^i\right)\right\} \subseteq \wh{\beta}$.
  We obtain:
  \[
    \norm(d_l, \wl{\beta}, \wl{\true})
    \leq
    \norm(d_h, \wh{\beta}, \wh{\true})
  \]
  Thus, $r_h \geq r_l$.
\end{proofE}
\begin{proofE}
  By structural induction on $\alpha$.
  \begin{itemize}
    \item Let $\alpha$ be an atomic formula. Then, since $\left(z_h, z_l\right) \in B$, it follows from \subdefref{def:bisim}{def:bisim:oiso}, that $\left(w_h, z_h\right) \oiso \left(w_l, z_l\right)$, and thus $w_h, z_h \models \alpha$ iff $w_l, z_l \models m \mleft(\alpha\mright)$.
    \item Let $\alpha = \beta \wedge \gamma$. The claim follows directly by induction and the semantics of conjunction.
    \item Let $\alpha = \neg \beta$. The claim follows directly by induction and the semantics of negation.
    \item Let $\alpha = \forall x.\, \beta$. The claim follows directly by induction and the semantics of all-quantification.
    \item Let $\alpha = \bel{\beta}{r}$.
      By definition, $e_h, w_h \models \bel{\beta}{r_h}$ iff
      \[
        \norm\mleft(d_h, \wh{\beta}, \wh{\true}, r_h\mright)
      \]
      Similarly, $e_l, w_l \models \bel{m(\beta)}{r_l}$ iff
      \[
        \norm\mleft(d_l, \wl{\beta}, \wl{\true}, r_l\mright)
      \]
      \underline{$r_h \leq r_l$}:
      For each $\left(w_h^i, z_h^i\right) \in \wh{\beta}$ with $d_h(w_h^i) > 0$ and $e_h, w_h^i \models \exec(z_h^i)$,
      by \subdefref{def:bisim}{def:bisim:high-epistemic-state}, there is a $\left(w_l^i, z_l^i\right)$ with $\left(\left(w_h^i, z_h^i\right), \left(w_l^i, z_l^i\right)\right) \in B$ and $\left(w_l^i, z_l^i\right) \oicomp \left(w_l, z_l\right)$.
      By \subdefref{def:bisim}{def:bisim:eiso},
      \[
        \left(d_h, w_h^i, z_h^i\right) \eiso (d_l, \underbrace{\left\{ \left(w_l', z_l'\right) \mid \left(\left(w_h^i, z_h^i\right), \left(w_l', z_l'\right)\right) \in B \right\}}_{=: \stateset_B} )
      \]
    Let $P$ be the partition $P = \stateset_B / \oicomp$ of $\stateset_B$. As $\left(w_l^i, z_l^i\right) \in \stateset_B$, there is a $\stateset_l^i \in P$ with $\left(w_l^i, z_l^i\right) \in \stateset_l^i$.
    By \autoref{def:eiso}:
    \[
      \norm(d_h, \{\left(w_h^i, z_h^i\right)\}, \wh[^i]{\true})
      =
      \norm(d_l, \stateset_l^i, \wl[^i]{\true})
    \]
    With $\left(w_l, z_l\right) \oicomp \left(w_l^i, z_l^i\right)$, it follows that $\wl[^i]{\true} = \wl{\true}$.
    Hence:
    \[\label{eqn:rh-leq-rl-wh-equal-sli}
      \norm(d_h, \{\left(w_h^i, z_h^i\right)\}, \wh[^i]{\true})
      =
      \norm(d_l, \stateset_l^i, \wl{\true})
    \]
    So far, we have only considered $\left(w_h^i, z_h^i\right) \in \wh{\beta}$ with $d_h(w_h^i) > 0$ and $e_h, w_h^i \models \exec(z_h^i)$.
    By definition of $\norm$, any $(w_h', z_h')$ with $d_h(w_h') = 0$ cannot add to $\norm$.
    Also, again by definition, for every $(w_h', z_h') \in \wh{\beta}$, $e_h, w_h' \models \exec(z_h')$.
    Therefore:
    \begin{gather}\label{eqn:zero-weight-states-do-not-add-anything}
      \norm(d_h, \bigcup_i \{ (w_h^i, z_h^i) \}, \wh{\true}) = \norm(d_h, \wh{\beta}, \wh{\true})
    \end{gather}
    Now, as $B$ is definite, it follows for each $i \neq j$ that $\stateset_l^i \neq \stateset_l^j$ and as $P$ is a partition, $\stateset_l^i \cap \stateset_l^j = \emptyset$.
    With this and with \autoref{eqn:rh-leq-rl-wh-equal-sli} and \autoref{eqn:zero-weight-states-do-not-add-anything}, it follows that
    \[
      \norm(d_h, \wh{\beta}, \wh{\true})
      = \norm(d_l, \bigcup_i \stateset_l^i, \wl{\true})
    \]

    We continue by showing the connection between all $\stateset_l^i$ and $\wl{\beta}$:
    For each $\left(w_l', z_l'\right) \in \stateset_l^i$, by definition of $\stateset_l^i$, we have $\left(w_l', z_l'\right) \oicomp \left(w_l, z_l\right)$.
    As $\left(\left(w_h^i, z_h^i\right), \left(w_l', z_l'\right)\right) \in B$, by \subdefref{def:bisim}{def:bisim:executability}, $e_l, w_l' \models \exec(z_l')$.
    Also, it follows by induction that $e_l, w_l', z_l' \models m(\beta)$.
    Thus, $\left(w_l', z_l'\right) \in \wl{\beta}$ and therefore, $\stateset_l^i \subseteq \wl{\beta}$.
    Therefore:
    \[
      \norm(d_l, \bigcup_i \stateset_l^i, \wl{\true})
      \leq
      \norm(d_l, \wl{\beta}, \wl{\true})
    \]
    We summarize:
    \begin{align*}
      r_h
      &= \norm(d_h, \wh{\beta}, \wh{\true})
      \\
      &= \norm(d_l, \bigcup_i \stateset_l^i, \wl{\true})
      \\
      &\leq \norm(d_l, \wl{\beta}, \wl{\true})
      \\
      &= r_l
    \end{align*}
    Thus, $r_h \leq r_l$.


      \medskip
      \underline{$r_l \leq r_h$}:
      For each $\left(w_l^i, z_l^i\right) \in \wl{\beta}$ with $d_l(w_l^i) > 0$ and $e_l, w_l^i \models \exec(z_l^i)$,
      as $(w_l^i, z_l^i) \oicomp (w_l, z_l)$, by \subdefref{def:bisim}{def:bisim:low-epistemic-state}, there is a $\left(w_h^i, z_h^i\right)$ with
      $\left(\left(w_h^i, z_h^i\right), \left(w_l^i, z_l^i\right)\right) \in B$ and therefore, by \subdefref{def:bisim}{def:bisim:eiso},
      \[
        \left(d_h, w_h^i, z_h^i\right) \eiso (d_l, \underbrace{\left\{ \left(w_l', z_l'\right) \mid \left(\left(w_h^i, z_h^i\right), \left(w_l', z_l'\right)\right) \in B \right\}}_{=: \stateset_B^i} )
      \]
    Let $P$ be the partition $P = \stateset_B^i / \oicomp$ of $\stateset_B^i$. As $\left(w_l^i, z_l^i\right) \in \stateset_B^i$, there is a $\stateset_l^i \in P$ with $\left(w_l^i, z_l^i\right) \in \stateset_l^i$.
      By \autoref{def:eiso},
      \begin{gather}\label{eqn:i-norms-are-equal}
        \norm(d_l, \stateset_l^i, \wl[^i]{\true})
        =
        \norm(d_h, \{ \left(w_h^i, z_h^i\right) \}, \wh[^i]{\true})
      \end{gather}
    Now, as $\left(w_h, z_h\right) \oicomp \left(w_h^i, z_h^i\right)$, it follows that $\wh[^i]{\true} = \wh{\true}$, similarly $\wl[^i]{\true} = \wl{\true}$.
    Therefore, we can also write \autoref{eqn:i-norms-are-equal} as
    \begin{gather}\label{eqn:i-norms-over-wl-are-equal}
      \norm(d_l, \stateset_l^i, \wl{\true})
      =
      \norm(d_h, \left\{\left(w_h^i, z_h^i\right)\right\}, \wh{\true})
    \end{gather}
    Now, suppose there is $j, k$ with $j \neq k$ such that $(w_h^j, z_h^j) = (w_h^k, z_h^k)$.
    Clearly, $\stateset_B^j = \stateset_B^k$.
    Also, $(w_l^j, z_l^j) \in \wl{\beta}$ and $(w_l^k, z_l^k) \in \wl{\beta}$,
    $(w_l^j, z_l^j) \oicomp (w_l, z_l)$, $(w_l^k, z_l^k) \oicomp (w_l, z_l)$,
    and therefore also $(w_l^j, z_l^j) \oicomp (w_l^k, z_l^k)$.
    Thus, $\stateset_l^j = \stateset_l^k$.
    As \autoref{eqn:i-norms-over-wl-are-equal} holds for each $i$, it follows that
    \begin{gather}\label{eqn:sum-norms-are-equal}
      \norm(d_l, \bigcup_i \stateset_l^i, \wl{\true})
      =
      \norm(d_h, \bigcup_i \left\{\left(w_h^i, z_h^i\right)\right\}, \wh{\true})
    \end{gather}
    \\
    Let $Q = \{ \stateset_{m(\beta)}^{1}, \stateset_{m(\beta)}^{2}, \ldots \}$ be the partition of $\wl{\beta} \cap \{ \left(w_l', z_l'\right) \mid d_l(w_l') > 0 \}$ such that
    $\stateset_{m(\beta)}^{i} \subseteq \stateset_l^i$.
    With \autoref{eqn:i-norms-over-wl-are-equal}, it directly follows that
    \begin{align}\label{eqn:sum-norms-l-is-leq}
      &\norm(d_l, \bigcup_i \stateset_{m(\beta)}^{i}, \wl{\true}) \nonumber
      \\
      &\leq
      \norm(d_l, \bigcup_i \stateset_l^i, \wl{\true}) \nonumber
      \\
      &=
      \norm(d_h, \bigcup_i \left\{\left(w_h^i, z_h^i\right)\right\}, \wh{\true})
    \end{align}
    By definition of \norm, any $(w_l', z_l')$ with $d_l(w_l') = 0$ cannot add to \norm, i.e.,
    \[
      \norm(d_l, \bigcup_i \stateset_{m(\beta)}^i, \wl{\true}) = \norm(d_l, \wl{\beta}, \wl{\true})
    \]
    With that, \autoref{eqn:sum-norms-l-is-leq} can be written as:
    \begin{gather}\label{eqn:l-norm-is-leq-sum-h-norms}
      \norm(d_l, \wl{\beta}, \wl{\true})
      \leq
      \norm(d_h, \bigcup_i \left\{\left(w_h^i, z_h^i\right)\right\}, \wh{\true})
    \end{gather}
    Finally, as $\left(\left(w_h^i, z_h^i\right), \left(w_l^i, z_l^i\right)\right) \in B$
    and $e_l, w_l^i, z_l^i \models  m(\beta)$,
    it follows by induction that $e_h, w_h^i, z_h^i \models \beta$.
    Therefore, with $\left(w_h^i, z_h^i\right) \oicomp \left(w_h, z_h\right)$,
    we have $\left(w_h^i, z_h^i\right) \in \wh{\beta}$.
    Hence:
    \[
      \norm(d_l, \wl{\beta}, \wl{\true})
      \leq
      \norm(d_h, \wh{\beta}, \wh{\true})
    \]
    Therefore $r_l \leq r_h$.

    With $r_h = r_l = r$, it follows that $e_h, w_h, z_h \models \bel{\beta}{r}$ iff $e_l, w_l, z_l \models \bel{m(\beta)}{r}$.
    \qedhere
  \end{itemize}
\end{proofE}

With \autoref{thm:static-equivalence}, we have established a \emph{static} equivalence between the high-level and the low-level states.
In the next step, we need to extend this to programs, i.e., non-static formulas of the form $[\delta] \alpha$.
In order to do so, using \autoref{thm:static-equivalence}, we first show that if $\left(e_h, w_h\right)$ is bisimilar to $\left(e_l, w_l\right)$, then $\left(e_h, w_h\right)$ and $\left(e_l, w_l\right)$ induce the same traces of a program $\delta$:
\begin{lemmaE}\label{lma:bisimulation-traces}
  Let $\left(e_h,w_h\right) \bisim \left(e_l,w_l\right)$ with $m$-bisimulation $B$, $\left(\left(w_h, z_h\right), \left(w_l, z_l\right)\right) \in B$, and $\delta$ be an arbitrary program.
  \begin{enumerate}
    \item If $z_l' \in \left\| m\mleft(\delta\mright) \right\|^{z_l}_{e_l, w_l}$ is a low-level trace, then there is a high-level trace $z_h' \in \left\| \delta \right\|^{z_h}_{e_h, w_h}$ such that $z_l' = m(z_h')$ and $\left(\left(w_h, z_h \cdot z_h'\right), \left(w_l, z_l \cdot z_l'\right)\right) \in B$.
    \item If $z_h' \in \left\| \delta \right\|^{z_h}_{e_h, w_h}$ is a high-level trace, then there is a low-level trace $z_l' \in \left\| m\mleft(\delta\mright) \right\|^{z_l}_{e_l, w_l}$ such that $z_l' = m(z_h')$ and $\left(\left(w_h, z_h \cdot z_h'\right), \left(w_l, z_l \cdot z_l'\right)\right) \in B$.
  \end{enumerate}
\end{lemmaE}
\begin{proof}[Proof Idea]
  By structural induction on $\delta$.
  For every static formula $\alpha$ that occurs in $\delta$, we can use \autoref{thm:static-equivalence} to show that $\alpha$ is satisfied by $(e_h, w_h, z_h)$ iff $m(\alpha)$ is satisfied by $(e_l, w_l, z_l)$.
  As tests and precondition axioms may only mention static formulas, the claim follows.
\end{proof}
\begin{proofE}
  ~
  \begin{enumerate}
    \item By structural induction on $\delta$.
      \begin{itemize}
        \item Let $\delta = a$ and thus $z_l' = \la m(a) \ra$. Then, by \subdefref{def:bisim}{def:bisim:low-action}, $w_h, z_h \models \poss(a)$, therefore $\la a \ra \in \left\| \delta \right\|^{z_h}_{e_h, w_h}$ and also $\left(\left(w_h, z_h \cdot a\right), \left(w_l, z_l \cdot z_l'\right)\right) \in B$.
        \item Let $\delta = \alpha?$. From $z_l' \in \left\| m\mleft(\delta\mright) \right\|^{z_l}_{e_l, w_l}$, it directly follows that $\la z_l, m(\alpha)? \ra \in \mathcal{F}^{e_l, w_l}$, $z_l' = \la\ra$ and $e_l, w_l, z_l \models m(\alpha)$.
          By \autoref{thm:static-equivalence}, it follows that $e_h, w_h, z_h \models \alpha$.
          Thus, $\la z_h, \alpha? \ra \in \mathcal{F}^{e_h, w_h}$, and therefore, for $z_h' = \la\ra$, we obtain $z_h' \in \left\| \delta \right\|^{z_h}_{e_h, w_h}$.
          Finally, as $z_h = z_l = \la\ra$ and $\left(\left(w_h, z_h\right), \left(w_l, z_l\right)\right) \in B$, it follows that $\left(\left(w_h, z_h \cdot z_h'\right), \left(w_l, z_l \cdot z_l'\right)\right) \in B$.
        \item Let $\delta = \delta_1; \delta_2$. By induction, for $z_l^1 \in \|m(\delta_1)\|^{z_l}_{e_l, w_l}$, there is $z_h^1 = \la a_1, \ldots, a_k \ra \in \|\delta_1\|^{z_h}_{e_h, w_h}$ with $z_l^1 = \la m(a_1), \ldots, m(a_k) \ra$ and $\left(z_h \cdot z_h^1, z_l, \cdot z_l^1\right) \in B$.
          Let $z_l^2 \in \| m(\delta_2)\|^{z_l \cdot z_l^1}_{e_l, w_l}$.
          It follows again by induction that there is $z_h^2 = \la a_{k+1}, \ldots, a_{n} \ra \in \|\delta_2\|^{z_h \cdot z_h^1}_{e_h, w_h}$  and such that $z_l^2 = \la m(a_{k+1}), \ldots m(a_n) \ra$ and $\left(z_h \cdot z_h^1 \cdot z_h^2, z_l \cdot z_l^1 \cdot z_l^2\right) \in B$.
        \item Let $\delta = \delta_1 \vert \delta_2$.
          Two cases:
          \begin{enumerate}
            \item Assume $z_l' \in \| m(\delta_1) \|^{z_l}_{e_l, w_l}$.
              Then, by induction, there is $z_h' \in \left\| \delta_1 \right\|^{z_h}_{e_h, w_h}$ such that $z_l' = m(z_h')$ and $\left(\left(w_h, z_h \cdot z_h'\right), \left(w_l, z_l \cdot z_l'\right)\right) \in B$.
            \item Assume $z_l' \in \| m(\delta_2) \|^{z_l}_{e_l, w_l}$.
              Then, by induction, there is $z_h' \in \left\| \delta_2 \right\|^{z_h}_{e_h, w_h}$ such that $z_l' = m(z_h')$ and $\left(\left(w_h, z_h \cdot z_h'\right), \left(w_l, z_l \cdot z_l'\right)\right) \in B$.
          \end{enumerate}
        \item Let $\delta = \pi x.\,\delta_1$ and $z_l' \in \|m(\delta)\|^{z_l}_{e_l, w_l}$ and so $z_l' \in \|m({\delta_1}^x_r)\|^{z_l}_{e_l, w_l}$ for some $r \in \rigid$.
          By induction, there is $z_h' \in \| {\delta_1}^x_r \|^{z_h}_{e_h, w_h}$ and therefore also $z_h' \in \| \pi x.\, \delta_1 \|^{z_h}_{e_h, w_h}$ such that $z_l' = m(z_h')$ and $\left(\left(w_h, z_h \cdot z_h'\right), \left(w_l, z_l \cdot z_l'\right)\right) \in B$.
        \item Let $\delta = \delta_1^*$ and $z_l' \in \|m(\delta_1^*)\|^{z_l}_{e_l, w_l}$.
          It is easy to see that $z_l'$ is the result of finitely many repetitions of $m(\delta_1)$,
          i.e., $z_l' = z_l^{(1)} \cdot \ldots \cdot z_l^{(n)}$ for some $n \in \naturals_0$ and where for all $i$, $z_l^{(i+1)} \in \| m(\delta_1) \|^{z_l \cdot z_l^{(1)} \cdot \ldots \cdot z_l^{(i)}}_{e_l, w_l}$.
          By sub-induction over $n$, we show that there is $z_h' = z_h^{(1)} \cdot \ldots \cdot z_h^{(n)} \in \|\delta\|^{z_h}_{e_h, w_h}$ such that $z_l^{(i)} = m(z_h^{(i)})$ and $\left(\left(w_h, z_h \cdot z_h'\right), \left(w_l, z_l \cdot z_l'\right)\right) \in B$.
          \begin{description}
            \item[Base case.] For $n = 0$ and thus $z_l' = \la\ra \in \| m(\delta_1) \|^{z_l}_{e_l, w_l}$, it is clear that $z_h' = \la\ra \in \| \delta \|^{z_h}_{e_h,w_h}$ and $m(z_h') = \la\ra = z_l'$ and so $\left(\left(w_h, z_h \cdot z_h'\right), \left(w_l, z_l, \cdot z_l'\right)\right) = \left(\left(w_h, z_h\right), \left(w_l, z_l\right)\right) \in B$.
            \item[Induction step.]
              Let $z_l' = z_l^{(1)} \cdot \ldots \cdot z_l^{(n+1)}$ such that for all $i \leq n$, $z_l^{(i+1)} \in \| m(\delta_1) \|^{z_l \cdot z_l^{(1)} \cdot \ldots \cdot z_l^{(i)}}_{e_l, w_l}$.
              Let $z_l^*$ denote $z_l^* = z_l^{(1)} \cdot \ldots \cdot z_l^{(n)}$ and so $z_l^{(n+1)} \in \| m(\delta) \|^{z_l \cdot z_l^*}_{e_l, w_l}$.
              By sub-induction, there is $z_h^* \in \| \delta \|^{z_h}_{e_h, w_h}$ such that $z_l^* = m(z_h^*)$ and $\left(\left(w_h, z_h \cdot z_h^*\right), \left(w_l, z_l \cdot z_l^*\right)\right) \in B$.
              As $z_l^{(n+1)} \in \| m(\delta) \|^{z_l \cdot z_l^*}_{e_l, w_l}$, it follows by induction that there is $z_h^{(n+1)} \in \| \delta_1 \|^{z_h \cdot z_h^*}_{e_h, w_h}$ such that $m(z_h^{(n+1)}) = z_l^{(n+1)}$ and $((w_h, z_h \cdot z_h^* \cdot z_h^{(n+1)}), (w_l, z_l \cdot z_l^* \cdot z_l^{(n+1)} )) \in B$.
              Hence, $z_h' = z_h^* \cdot z_h'' \in \| \delta \|^{z_h}_{e_h, w_h}$ and $\left(\left(w_h, z_h \cdot z_h'\right), \left(w_l, z_l  \cdot z_l'\right)\right) \in B$.
          \end{description}
      \end{itemize}
    \item By structural induction on $\delta$.
      \begin{itemize}
        \item Let $\delta = a$ and thus $z_h' = \la a \ra \in \left\| \delta \right\|^{z_h}_{e_h, w_h}$.
          Therefore, $w_h, z_h \models \poss(a)$ and thus, by \subdefref{def:bisim}{def:bisim:high-action}, there is $z_l' \in \left\| m(a) \right\|^{z_l}_{e_l, w_l}$ with $\left(\left(w_h, z_h \cdot z_h'\right), \left(w_l, z_l \cdot z_l'\right)\right) \in B$.
        \item Let $\delta = \alpha?$.
          From $z_h' \in \left\| \delta \right\|^{z_h}_{e_h, w_h}$, it directly follows that $\la z_h, a? \ra \in \mathcal{F}^{e_h, w_h}$, $z_h' = \la\ra$, and $e_h, w_h, z_h \models \alpha$.
          By \autoref{thm:static-equivalence}, it follows that $e_l, w_l, z_l \models m(\alpha)$.
          Thus, $\la z_l, m(\alpha)? \ra \in \mathcal{F}^{e_l, w_l}$, and therefore $z_l' = \la\ra \in \left\| m(\delta) \right\|^{z_l}_{e_l, w_l}$.
          Finally, as $z_h = z_l = \la\ra$ and $\left(\left(w_h, z_h\right), \left(w_l, z_l\right)\right) \in B$, it follows that $\left(\left(w_h, z_h \cdot z_h'\right), \left(w_l, z_l \cdot z_l'\right)\right) \in B$.
        \item Let $\delta = \delta_1; \delta_2$.
          By induction, for $z_h^1 \in \| \delta_1 \|^{z_h}_{e_h, w_h}$, there is $z_l^1 \in \| m(\delta_1) \|^{z_l}_{e_l, w_l}$ such that $z_l^1 = m(z_h^1)$ with $\left(\left(w_h, z_h \cdot z_h^1\right), \left(w_l, z_l \cdot z_l^1\right)\right) \in B$.
          Again by induction, for $z_h^2 \in \| \delta_2 \|^{z_h \cdot z_h^1}_{e_h, w_h}$, there is $z_l^2 \in \| m(\delta_2) \|^{z_l \cdot z_l^1}_{e_l, w_l}$ such that $z_l^2 = m(z_h^2)$ and $\left(\left(z_h \cdot z_h^1 \cdot z_h^2\right), \left(z_l \cdot z_l^1 \cdot z_l^2\right)\right) \in B$.
        \item Let $\delta = \delta_1 \vert \delta_2$.
          Two cases:
          \begin{enumerate}
            \item Assume $z_h' = \in \| \delta_1 \|^{z_h}_{e_h, w_h}$.
              Then, by induction, there is $z_l' \in \left\| m\mleft(\delta_1\mright) \right\|^{z_l}_{e_l, w_l}$ with $z_l' = m(z_h')$ and $\left(\left(w_h, z_h \cdot z_h'\right), \left(w_l, z_l \cdot z_l'\right)\right) \in B$.
            \item Assume $z_h' \in \| \delta_2 \|^{z_h}_{e_h, w_h}$.
              Then, by induction, there is $z_l' \in \left\| m\mleft(\delta_2\mright) \right\|^{z_l}_{e_l, w_l}$ with $z_l' = m(z_h')$ and $\left(\left(w_h, z_h \cdot z_h'\right), \left(w_l, z_l \cdot z_l'\right)\right) \in B$.
          \end{enumerate}
        \item Let $\delta = \pi x.\,\delta_1$ and $z_h' \in \|\delta\|^{z_h}_{e_h, w_h}$ and so $z_h' \in \|{\delta_1}^x_r\|^{z_h}_{e_h, w_h}$ for some $r \in \rigid$.
          By induction, there is $z_l' \in \| m({\delta_1}^x_r) \|^{z_l}_{e_l, w_l}$ and therefore also $z_l' \in \| m(\pi x.\, \delta_1) \|^{z_l}_{e_l, w_l}$ such that $z_l' = m(z_h')$ and $\left(\left(w_h, z_h \cdot z_h'\right), \left(w_l, z_l \cdot z_l'\right)\right) \in B$.
        \item Let $\delta = \delta_1^*$ and $z_h' \in \|\delta_1^*\|^{z_h}_{e_h, w_h}$.
          It is easy to see that $z_h'$ is the result of finitely many repetitions of $\delta_1$,
          i.e., $z_h' = z_h^{(1)} \cdot \ldots \cdot z_h^{(n)}$ for some $n \in \naturals_0$ and where for all $i$, $z_h^{(i+1)} \in \| \delta_1 \|^{z_h \cdot z_h^{(1)} \cdot \ldots \cdot z_h^{(i)}}_{e_h, w_h}$.
          By sub-induction over $n$, we show that there is $z_l' = z_l^{(1)} \cdot \ldots \cdot z_l^{(n)} \in \|m(\delta)\|^{z_l}_{e_l, w_l}$ such that $z_l^{(i)} = m(z_h^{(i)})$ and $\left(\left(w_h, z_h \cdot z_h'\right), \left(w_l, z_l \cdot z_l'\right)\right) \in B$.
          \begin{description}
            \item[Base case.] For $n = 0$ and thus $z_h' = \la\ra \in \| \delta_1 \|^{z_h}_{e_h, w_h}$, it is clear that $z_l' = \la\ra \in \| m(\delta) \|^{z_l}_{e_l,w_l}$ and $m(z_h') = \la\ra = z_l'$ and so $\left(\left(w_h, z_h \cdot z_h'\right), \left(w_l, z_l, \cdot z_l'\right)\right) = \left(\left(w_h, z_h\right), \left(w_l, z_l\right)\right) \in B$.
            \item[Induction step.]
              Let $z_h' = z_h^{(1)} \cdot \ldots \cdot z_h^{(n+1)}$ such that for all $i \leq n$, $z_h^{(i+1)} \in \| \delta_1 \|^{z_h \cdot z_h^{(1)} \cdot \ldots \cdot z_h^{(i)}}_{e_h, w_h}$.
              Let $z_h^*$ denote $z_h^* = z_h^{(1)} \cdot \ldots \cdot z_h^{(n)}$ and so $z_h^{(n+1)} \in \| \delta \|^{z_h \cdot z_h^*}_{e_h, w_h}$.
              By sub-induction, there is $z_l^* \in \| m(\delta) \|^{z_l}_{e_l, w_l}$ such that $z_l^* = m(z_h^*)$ and $\left(\left(w_h, z_h \cdot z_h^*\right), \left(w_l, z_l \cdot z_l^*\right)\right) \in B$.
              As $z_h^{(n+1)} \in \| \delta \|^{z_h \cdot z_h^*}_{e_h, w_h}$, it follows by induction that there is $z_l^{(n+1)} \in \| m(\delta_1) \|^{z_l \cdot z_l^*}_{e_l, w_l}$ such that $m(z_h^{(n+1)}) = z_l^{(n+1)}$ and $((w_h, z_h \cdot z_h^* \cdot z_h^{(n+1)}), (w_l, z_l \cdot z_l^* \cdot z_l^{(n+1)} )) \in B$.
              Hence, $z_l' = z_l^* \cdot z_l'' \in \| m(\delta) \|^{z_l}_{e_l, w_l}$ and $\left(\left(w_h, z_h \cdot z_h'\right), \left(w_l, z_l  \cdot z_l'\right)\right) \in B$.
              \qedhere
          \end{description}
      \end{itemize}
  \end{enumerate}
\end{proofE}

Note that \autoref{lma:bisimulation-traces} would not hold if $\delta$ contained interleaved concurrency. Intuitively, this is because for a high-level program such as $a_h^1 \| a_h^2$, the only valid high-level traces would be $\la a_h^1, a_h^2\ra$ and $\la a_h^2, a_h^1\ra$, i.e., one action is completely executed before the other action is started.
On the other hand, with $m(a_h^1) = a_l^1; a_l^2$ and $m(a_h^2) = a_l^3; a_l^4$, we may obtain interleaved traces such as $\la a_l^1, a_l^3, a_l^2, a_l^4\ra$, which does not have a corresponding high-level trace.
While a limited form of concurrency could be permitted by only allowing interleaved execution of high-level actions (i.e., each $m(a)$ must be completely executed before switching to a different branch of execution), we omit this for the sake of simplicity.

With \autoref{lma:bisimulation-traces}, we can extend \autoref{thm:static-equivalence} to bounded formulas:
\begin{theoremE}\label{thm:model-mapping}
  Let $\left(e_h,w_h\right) \bisim \left(e_l,w_l\right)$ with $m$-bisimulation $B$.
  For every bounded formula $\alpha$ and traces $z_h, z_l$ with $\left(z_h, z_l\right) \in B$:
  \[
  e_h, w_h, z_h \models \alpha \text{ iff } e_l, w_l, z_l \models m\mleft(\alpha\mright)
  \]
\end{theoremE}
\begin{proofE}[no proof end][Proof Idea]
  By structural induction on $\alpha$, similarly to \autoref{thm:static-equivalence}.
  For formulas of the form $\alpha = [\delta] \beta$, it can be shown with \autoref{lma:bisimulation-traces} that they induce the same traces, which allows us to apply \autoref{thm:static-equivalence} again.
\end{proofE}
\begin{proofE}
  By structural induction on $\alpha$.
  \begin{itemize}
    \item Let $\alpha$ be an atomic formula. Then, since $\left(z_h, z_l\right) \in B$, we know that $\left(w_h, z_h\right) \oiso \left(w_l, z_l\right)$, and thus $w_h, z_h \models \alpha$ iff $w_l, z_l \models m \mleft(\alpha\mright)$.
    \item Let $\alpha = \bel{\beta}{r}$.
      Same proof as in \autoref{thm:static-equivalence}.
    \item Let $\alpha = \beta \wedge \gamma$. The claim follows directly by induction and the semantics of conjunction.
    \item Let $\alpha = \neg \beta$. The claim follows directly by induction and the semantics of negation.
    \item Let $\alpha = \forall x.\, \beta$. The claim follows directly by induction and the semantics of all-quantification.
    \item Let $\alpha = [\delta] \beta$.
      \\
      \textbf{$\Leftarrow$:}
      Let $e_h, w_h, z_h \not\models [\delta]\beta$.
      There is a finite trace $z'_h \in \|\delta\|^{z_h}_{e_h,w_h}$ with $e_h, w_h, z_h \cdot z'_h \not\models \beta$.
      By \autoref{lma:bisimulation-traces}, there is $z_l' \in \|m(\delta)\|^{z_l}_{e_l, w_l}$ with $\left(z_h \cdot z_h', z_l \cdot z_l'\right) \in B$.
      By induction, $e_l, w_l, z_l \cdot z_l' \not\models \beta$, and thus $e_l, w_l, z_l \not\models [m(\delta)]m(\beta)$.
      \\
      \textbf{$\Rightarrow$:}
      Let $\left(e_l,w_l\right) \not\models [m\mleft(\delta\mright)] m\mleft(\beta\mright)$, i.e., there is a finite trace $z_l' \in \|m(\delta)\|^{z_l}_{e_l, w_l}$ with $e_l, w_l, z_l \cdot z_l' \not\models m(\beta)$.
      By \autoref{lma:bisimulation-traces}, there is a $z_h' \in \| \delta \|^{z_h}_{e_h, w_h}$ with $\left(z_h \cdot z_h', z_l \cdot z_l'\right) \in B$.
      By induction, $e_h, w_h, z_h \cdot z_h' \not\models \beta$ and thus $e_h, w_h, z_h \not\models [\delta]\beta$.
      \qedhere
  \end{itemize}
\end{proofE}

It directly follows that the high- and low-level models entail the same formulas after executing some program $\delta$:
\begin{corollaryE}\label{thm:mapping-satisfiability}
  Let $\left(e_h,w_h\right) \bisim \left(e_l,w_l\right)$.
  Then for any high-level Golog program $\delta$ and static high-level formula $\beta$:
  \[
    e_h,w_h \models [\delta]\beta
    \text{ iff }
    e_l,w_l \models [m\mleft(\delta\mright)] m\mleft(\beta\mright)
  \]
\end{corollaryE}
\begin{proofE}
  This is a special case of \autoref{thm:model-mapping} with $z_h = \la\ra, z_l = \la\ra, \alpha = [\delta]\beta$.
\end{proofE}

Hence, a bisimulation between $(e_h, w_h)$ and $(e_l, w_l)$ indeed establishes an equivalence between thew high-level and low-level model, as they produce the same program traces and satisfy the same formulas.
\section{Sound and Complete Abstraction}
In the previous section, we described properties of abstraction with respect to particular models $\left(e_h, w_h\right)$ and $\left(e_l, w_l\right)$.
However, we are usually more interested in the relationship between a high-level \ac{BAT} $\Sigma_h$ and a low-level \ac{BAT} $\Sigma_l$.
A first notion in that regards is a \emph{sound abstraction}, which intuitively states that for every low-level model of $\bat_l$, there exists a bisimilar high-level model of $\bat_h$:
\begin{definition}[Sound Abstraction]
  We say that $\Sigma_h$ is a \emph{sound abstraction of $\Sigma_l$ relative to refinement mapping $m$} if and only if for each model $\left(e_l, w_l\right) \models \know{\Sigma_l} \wedge \Sigma_l$, there exists a model $\left(e_h, w_h\right) \models \know{\Sigma_h} \wedge \Sigma_h$ such that $\left(e_h, w_h\right) \bisim \left(e_l, w_l\right)$.
\end{definition}
Notice that in addition to requiring that $(e_l, w_l)$ models $\bat_l$, we also require that the agent \emph{knows} $\bat_l$  (and similarly for $\bat_h$).
Therefore, we require the real world to have the same physical laws as that believed by the agent, which is fairly standard.
However, we do not require that the agent knows everything about the real world, nor do we require that everything the agent believes is also true in the real world.

We can show that conclusions by the high-level \ac{BAT} $\Sigma_h$ are consistent with the low-level \ac{BAT} $\Sigma_l$:
\begin{theoremE}\label{thm:sound-abstraction}
  Let $\Sigma_h$ be a sound abstraction of $\Sigma_l$ relative to mapping $m$.
Then, for every bounded formula $\alpha$, if $\know{\Sigma_h} \wedge \Sigma_h \models \alpha$, then $\know{\Sigma_l} \wedge \Sigma_l \models m(\alpha)$.
\end{theoremE}
\begin{proofE}
  Let $\know{\Sigma_h} \wedge \Sigma_h \models \alpha$.
  Suppose $\know{\Sigma_l} \wedge \Sigma_l \not\models m(\alpha)$,
  i.e., there is a model $\left(e_l, w_l\right)$ of $\know{\Sigma_l} \wedge \Sigma_l$ with $e_l, w_l \not\models m(\alpha)$.
  As $\Sigma_h$ is a sound abstraction of $\Sigma_l$, there is a model $\left(e_h, w_h\right)$ of $\know{\Sigma_h} \wedge \Sigma_h$ with $\left(e_h, w_h\right) \bisim \left(e_l, w_l\right)$.
  By \autoref{thm:model-mapping}, $e_h, w_h \not\models \alpha$.
  Contradiction to $\know{\Sigma_h} \wedge \Sigma_h \models \alpha$.
  Thus, $\know{\Sigma_l} \wedge \Sigma_l \models  m(\alpha)$.
\end{proofE}



While a sound abstraction ensures that any entailment of the high-level \ac{BAT} $\Sigma_h$ is consistent with the low-level \ac{BAT} $\Sigma_l$, $\Sigma_h$ may have less information than $\Sigma_l$, e.g., $\Sigma_h$ may consider it possible that some program $\delta$ is executable, while $\Sigma_l$ knows that it is not.
This leads to a second notion of abstraction:
\begin{definition}[Complete Abstraction]
  We say that $\Sigma_h$ is a \emph{complete abstraction of $\Sigma_l$ relative to refinement mapping $m$} if and only if for each model $\left(e_h, w_h\right) \models \know{\Sigma_h} \wedge \Sigma_h$,
  there exists a model $\left(e_l, w_l\right) \models \know{\Sigma_l} \wedge \Sigma_l$ such that $\left(e_h, w_h\right) \bisim \left(e_l, w_l\right)$.
\end{definition}

Indeed, if we have a complete abstraction, then $\Sigma_h$ must entail everything that $\Sigma_l$ entails:
\begin{theoremE}\label{thm:complete-abstraction}
  Let $\Sigma_h$ be a complete abstraction of $\Sigma_l$ relative to mapping $m$. Then, for every bounded formula $\alpha$, if $\know{\Sigma_l} \wedge \Sigma_l \models m(\alpha)$, then $\know{\Sigma_h} \wedge \Sigma_h \models \alpha$.
\end{theoremE}
\begin{proofE}
  Let $\know{\Sigma_l} \wedge \Sigma_l \models m(\alpha)$.
  Suppose $\know{\Sigma_h} \wedge \Sigma_h \not\models \alpha$, i.e., there is a model $\left(e_h, w_h\right)$ of $\know{\Sigma_h} \wedge \Sigma_h$ with $\left(e_h, w_h\right) \not\models \alpha$.
  As $\Sigma_h$ is a complete abstraction of $\Sigma_l$, there is a model $\left(e_l, w_l\right)$ with $e_l, w_l \models \know{\Sigma_l} \wedge \Sigma_l$ and $\left(e_h, w_h\right) \bisim \left(e_l, w_l\right)$.
  By \autoref{thm:model-mapping}, $e_l, w_l \not\models m(\alpha)$.
  Contradiction to $\Sigma_l \models m(\alpha)$.
  Thus, $\know{\Sigma_h} \wedge \Sigma_h \models \alpha$.
\end{proofE}

The strongest notion of abstraction is the combination of sound and complete abstraction:
\begin{definition}[Sound and Complete Abstraction]
  ~ \\
  We say that $\Sigma_h$ is a \emph{sound and complete abstraction} of $\Sigma_l$ relative to refinement mapping $m$ if $\Sigma_h$ is both a sound and a complete abstraction of $\Sigma_l$ wrt $m$.
\end{definition}

\begin{theoremE}\label{thm:sound-complete-abstraction}
  Let $\Sigma_h$ be a sound and complete abstraction of $\Sigma_l$ relative to refinement mapping $m$.
  Then, for every bounded formula $\alpha$, $\know{\Sigma_h} \wedge \Sigma_h \models \alpha$ iff $\know{\Sigma_l} \wedge \Sigma_l \models m(\alpha)$.
\end{theoremE}
\begin{proofE}
  Follows directly from \autoref{thm:sound-abstraction} and \autoref{thm:complete-abstraction} .
\end{proofE}

\begin{figure}[htb]
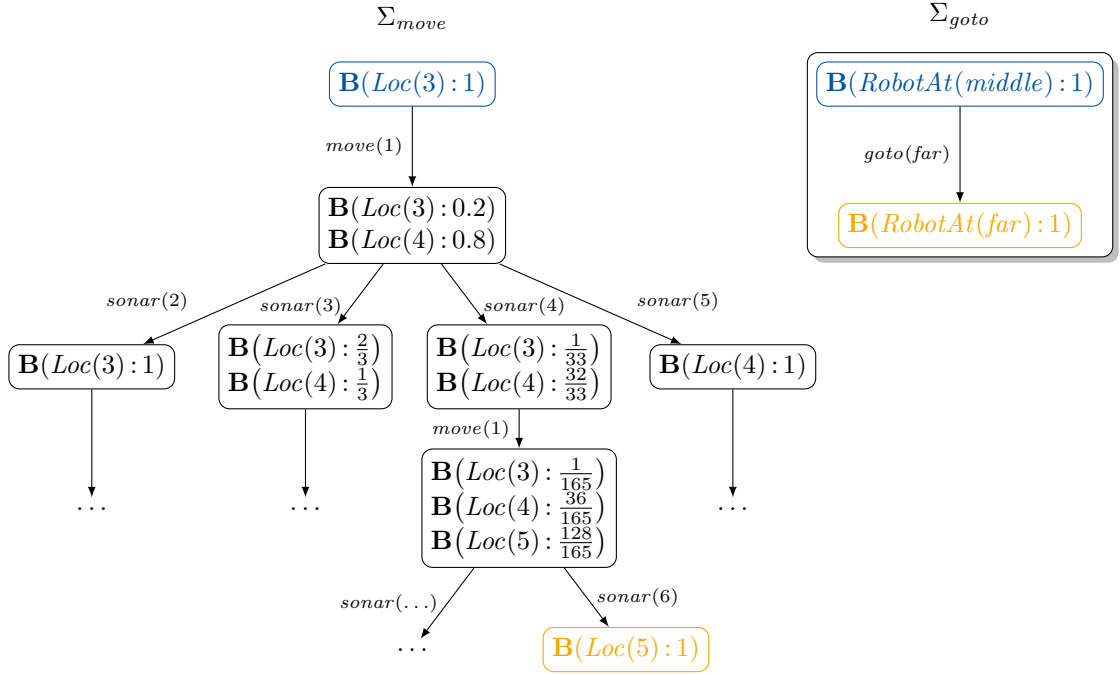

  \centering
  \includestandalone{figures/bisimulation}
  \caption[Bisimulation.]{
    Bisimulation for the running example, where sets of states are summarized by the belief that they entail.
    The single transition for \goto of the high-level \ac{BAT} is shown on the right.
    The agent knows that it is initially in the middle and after doing $\goto(\far)$, it is far away from the wall.
    Some corresponding transitions of the low-level \ac{BAT} are shown on the left:
    Initially, the agent knows that it is at distance $3$, which is a bisimilar state to the initial high-level state (blue).
    After $\move(1)$, the agent cannot distinguish  whether it actually moved or was stuck, so there are two possibilities: it can either still be at distance $3$ or at distance $4$.
    Eventually, it reaches a state where it knows that it is at distance $5$, which is again a bisimilar state to the corresponding high-level state (orange).
  }
  \label{fig:bisimulation}
\end{figure}

Coming back to our example, we can show that \sigmah is indeed a sound abstraction of \sigmal:
\newcommand*{\nonhfluents}{\ensuremath{\mathcal{F} \setminus \mathcal{F}_h}\xspace}
\begin{propositionE}\label{thm:sound-bat}
  \sigmah is a sound abstraction of \sigmal relative to refinement mapping $m$.
\end{propositionE}
\begin{proof}[Proof Sketch]
  Let $e_l, w_l \models \know{\sigmal} \wedge \sigmal$.
  We show by construction that there is a model $\left(e_h, w_h\right)$ with $e_h, w_h \models \know{\sigmah} \wedge \sigmah$ and $\left(e_h, w_h\right) \bisim \left(e_l, w_l\right)$.
  First, note that there may be multiple worlds $w_l'$ with $d_l(w_l') > 0$, which all need to be considered.
  However, from $e_l, w_l \models \know{\sigmal}$, it follows that $w_l' \models \sigmal$ for every $w_l'$ with $d_l(w_l') > 0$.
  Let $w_h \models \sigmah$ and let $e_h$ be an epistemic state such that $d_h(w_h) = 1$ and $d_h(w_h') = 0$ for every $w_h' \neq w_h$.
  Clearly, $e_h, w_h \models \know{\sigmah} \wedge \sigmah$.
  Now, let
  \[
    B_0 =
    \left\{ \left(\left(w_h, \la\ra\right), \left(w_l, \la\ra\right)\right) \right\} \cup
    \left\{ \left(\left(w_h, \la\ra\right), \left(w_l', \la\ra\right)\right) \mid d_l(w_l') > 0 \right\}
  \]
  Next, let
  \begin{align*}
    B_{i+1} = \big\{ \left(\left(w_h', z_h' \cdot a\right), \left(w_l', z_l' \cdot z_l''\right)\right) \mid
      &
      \left(\left(w_h', z_h'\right), \left(w_l', z_l'\right)\right) \in B_i,
      \\ &
      w_h', z_h' \models \poss(a), z_l'' \in \|m(a)\|^{z_l'}_{e_l, w_l}\big\}
  \end{align*}
  As $B$ only mentions a single high-level world $w_h$, it directly follows that $B$ is definite.
  It can be shown by induction on $i$ that $B = \bigcup_i B_i$ is an $m$-bisimulation between $\left(e_h, w_h\right)$ and $\left(e_l, w_l\right)$.
  Therefore, for each $e_l, w_l \models \know{\sigmal} \wedge \sigmal$, there is a $\left(e_h, w_h\right) \models \know{\Sigma_h} \wedge \Sigma_h$ with $(e_h, w_h) \bisim (e_l, w_l)$.
  Thus, \sigmah is a sound abstraction of \sigmal.
\end{proof}
\begin{proofE}
  Let $e_l, w_l \models \know{\sigmal} \wedge \sigmal$.
  We show by construction that there is a model $\left(e_h, w_h\right)$ with $e_h, w_h \models \know{\sigmah} \wedge \sigmah$ and $\left(e_h, w_h\right) \bisim \left(e_l, w_l\right)$.
  First, note that there may be multiple worlds $w_l'$ with $d_l(w_l') > 0$, which all need to be considered.
  However, from $e_l, w_l \models \know{\sigmal}$, it follows that $w_l' \models \sigmal$ for every $w_l'$ with $d_l(w_l') > 0$.
  Let $w_h \models \sigmah$ and let $e_h$ be an epistemic state such that $d_h(w_h) = 1$ and $d_h(w_h') = 0$ for every $w_h' \neq w_h$.
  Clearly, $e_h, w_h \models \know{\sigmah} \wedge \sigmah$.
  Now, let
  \[
    B_0 =
    \left\{ \left(\left(w_h, \la\ra\right), \left(w_l, \la\ra\right)\right) \right\} \cup
    \left\{ \left(\left(w_h, \la\ra\right), \left(w_l', \la\ra\right)\right) \mid d_l(w_l') > 0 \right\}
  \]
  Next, let
  \begin{align*}
    B_{i+1} = \big\{ \left(\left(w_h', z_h' \cdot a\right), \left(w_l', z_l' \cdot z_l''\right)\right) \mid
      &\left(\left(w_h', z_h'\right), \left(w_l', z_l'\right)\right) \in B_i,
      \\ &
      w_h', z_h' \models \poss(a), z_l'' \in \|m(a)\|^{z_l'}_{e_l, w_l}\big\}
  \end{align*}
  As $B$ only mentions a single high-level world $w_h$, it directly follows that $B$ is definite.
  We show by induction on $i$ that $B = \bigcup_i B_i$ is an $m$-bisimulation between $\left(e_h, w_h\right)$ and $\left(e_l, w_l\right)$.
  Let $\left(\left(w_h', z_h'\right), \left(w_l', z_l'\right)\right) \in B_i$.

  \medskip
  \noindent \textbf{Base case.}
  Note that $z_h' = z_l' = \la\ra$ by definition of $B_0$.
  We show that all criteria of \autoref{def:bisim} are satisfied:
  \begin{enumerate}
    \item By definition, $w_l' \models \sigmal$ and thus $w_l' \models \forall x (\loc(x) \equivspace x = 3)$.
      At the same time, $w_h' \models \sigmah$ and thus $w_h' \models \forall l \neg \at(l)$.
      Therefore, for all $l$, $w_h' \models \at(l)$ iff $w_l' \models m(\at(l))$ and thus, $\left(w_h', \la\ra\right) \oiso \left(w_l', \la\ra\right)$.
    \item By definition of $e_h$, $\norm(d_h, \{ \left(w_h, \la\ra\right) \}, \wh{\true}, 1)$.
      Also, by definition of $B_0$, $d_l(w_l') > 0$ iff $\left(\left(w_h, \la\ra\right), \left(w_l', \la\ra\right)\right) \in B$.
      Let $\stateset_l = \{ \left(w_l', \la\ra\right) \mid d_l(w_l') > 0 \}$.
      It directly follows that for each set $\stateset_l^i$ of the partition $\stateset_l / \oicomp$,
      $\norm(d_l, \stateset_l^i, \wlfull{e_l, [\stateset_l^{(i)}]}{\true}, 1)$.
      Thus, $\left(d_h, w_h, \la\ra\right) \eiso \left(d_l, \stateset_l\right)$.
    \item As $z_h' = z_l' = \la\ra$, it directly follows that $e_h, w_h' \models \exec(z_h')$ and $e_l, w_l' \models \exec(z_l')$.
    \item Let $w_h' \models \poss(a)$.
      Then, $a = \goto(l)$ for some $l \in \{ \near, \far \}$.
      As $e_l, w_l' \models \sigmal$, it follows for each such $l$ that there is some $z_l'' \in \|m(\goto(l))\|^{\la\ra}_{e_l, w_l'}$:
      \begin{itemize}
        \item For $l = \near$, $z_l'' = \la \sonar(), \move(-1, -1), \sonar() \ra$.
        \item For $l = \far$, \newline $z_l'' = \la \sonar, \move(1, 1), \sonar(), \move(1, 1), \sonar() \ra$.
      \end{itemize}
      By definition of $B_{i+1}$, we obtain $\left(\left(w_h', z_h' \cdot a\right), \left(w_l', z_l' \cdot z_l''\right)\right) \in B$.
    \item Let $z_l'' \in \|m(a)\|^{z_l'}_{e_l, w_l}$.
      Clearly, $a = \goto(l)$ for some $l \in \{ \near, \far \}$.
      By definition of \sigmah, it directly follows that $e_h, w_h', z_h' \models \poss(a)$.
      By definition of $B_{i+1}$, it also follows that $\left(\left(w_h', z_h' \cdot a\right), \left(w_l', z_l' \cdot z_l''\right)\right) \in B$.
    \item Let $\left(w_h'', z_h''\right) \oicomp \left(w_h', z_h'\right)$ with $d_h(w_h'') > 0$ and $e_h, w_h'' \models \exec(z_h'')$.
      As $d_h(w_h'') = 0$ for every $w_h'' \neq w_h'$ and $z_h' = \la\ra$, it directly follows that $\left(w_h'', z_h''\right) = \left(w_h', z_h'\right)$
      and thus
      $\left(\left(w_h'', z_h''\right), \left(w_l', z_l'\right)\right) \in B$.
    \item Let $\left(w_l'', z_l''\right) \oicomp \left(w_l', z_l'\right)$ with $d_l(w_l'') > 0$ and $e_h, w_l'' \models \exec(z_l'')$.
      Clearly, $z_l'' = z_l' = \la\ra$.
      By definition of $B_0$, $\left(\left(w_h, \la\ra\right), \left(w_l'', \la\ra\right)\right) \in B$.
  \end{enumerate}
  \medskip
  \noindent \textbf{Induction step.}
  \begin{enumerate}
    \item Let $z_h' = z_h'' \cdot a$ and $z_l' = z_l'' \cdot z_l'''$ for some $z_l''' \in \|m(a)\|^{z_l''}_{e_l, w_l'}$.
      By construction, $\left(\left(w_h', z_h''\right), \left(w_l', z_l''\right)\right) \in B$.
      By induction, $\left(w_h', z_h''\right) \oiso \left(w_l', z_l''\right)$.
      Furthermore, $w_h' \models \sigmah$ and $w_l' \models \sigmal$.
      As before, $a = \goto(l)$ with $l \in \{ \near, \far \}$.
      As $w_h' \models \sigmah$, for every $l'$, $e_h, w_h', z_h' \models \at(l')$ iff $l' = l$.
      Similarly, by definition of $m$ and $\sigmal$, for every $l'$, $e_l, w_l', z_l' \models m(\at(l'))$ iff $l' = l$.
      Thus, $\left(w_h', z_h'\right) \oiso \left(w_l', z_l'\right)$.
    \item Suppose
      \[
        \left(d_h, w_h', z_h'\right) \not\eiso (d_l, \underbrace{\left\{ \left(w_l'', z_l''\right) \mid \left(\left(w_h', z_h'\right), \left(w_l'', z_l''\right)\right) \in B \right\}}_{=:\stateset_B} )
      \]
      First, note that $\norm(d_h, \{ \left(w_h', z_h'\right) \}, \wh[']{\true}, 1)$.
      Therefore, there is a $\stateset_l^i \in \stateset_B / \oicomp$ and $(w_l^i, z_l^i) \in \stateset_l^i$ where $\norm(d_l, \stateset_l^i, \wl[^i]{\true}, 1) \neq 1$,
      i.e., there is $\left(w_l'', z_l''\right) \in \wl[^i]{\true} \setminus \stateset_l^i$ with $d_l(w_l'') \times l^*(w_l'', z_l'') > 0$.
      It follows that $\left(w_l'', z_l''\right) \oicomp \left(w_l^i, z_l^i\right)$ for some $\left(w_l^i, z_l^i\right) \in \stateset_l^i$.
      But then, by definition of \sigmal, $z_l''$ is the same as $z_l^i$, except a possibly different second parameter of each $\move(x, y)$ action.
      Also, $z_l^i = z_l^{i,1} \cdot z_l^{i,2}$, where $z_l^{i,2} \in \| m(a) \|^{z_l^{i,1}}_{e_l, w_l^i}$ for some action $a$.
      As $z_l'' \oisim[w_l''] z_l^i$, it follows that $z_l'' = z_l^{''1} \cdot z_l^{''2}$ with $z_l^{''2} \in \| m(a) \|^{z_l^{''1}}_{e_l,w_l''}$
      and $z_l^{''1} \oisim[w_l''] z_l^{i,1}$.
      But then, by induction, there is some $(w_h'', z_h^{''1})$ such that $((w_h'', z_h^{''1}), (w_l'', z_l^{''1})) \in B$,
      and therefore, by definition of $B$, also $((w_h'', z_h^{''1} \cdot a), (w_l'', z_l'')) \in B$.
      Contradiction to $\left(w_l'', z_l''\right) \in \wl[^i]{\true} \setminus \stateset_l^i$.
      It follows:
      \[
        \left(d_h, w_h', z_h'\right) \eiso (d_l, \underbrace{\left\{ \left(w_l'', z_l''\right) \mid \left(\left(w_h', z_h'\right), \left(w_l'', z_l''\right)\right) \in B \right\}}_{=:\stateset_B} )
      \]
    \item $w_h' \models \exec(z_h')$ and $w_l' \models \exec(z_l')$ directly follows by construction of $B$.
    \item Let $w_h', z_h' \models \poss(a)$.
      Then, $a = \goto(l)$ for some $l \in \{ \near, \far \}$.
      As $e_l, w_l' \models \sigmal$, it follows for each such $l$ that there is some $z_l'' \in \|m(\goto(l))\|^{z_l'}_{e_l, w_l'}$.
      By definition of $B_{i+1}$, it also follows that $\left(\left(w_h', z_h' \cdot a\right), \left(w_l', z_l' \cdot z_l''\right)\right) \in B$.
    \item Let $z_l'' \in \|m(a)\|^{z_l'}_{e_l, w_l}$.
      Clearly, $a = \goto(l)$ for some $l \in \{ \near, \far \}$.
      By definition of \sigmah, it directly follows that $e_h, w_h', z_h' \models \poss(a)$.
      By definition of $B_{i+1}$, it also follows that $\left(\left(w_h', z_h' \cdot a\right), \left(w_l', z_l' \cdot z_l''\right)\right) \in B$.
    \item Let $\left(w_h'', z_h''\right) \oicomp \left(w_h', z_h'\right)$ with $d_h(w_h'') > 0$ and $e_h, w_h'' \models \exec(z_h'')$.
      As $d_h(w_h'') = 0$ for every $w_h'' \neq w_h'$, it follows that $w_h'' = w_h'$.
      Furthermore, by definition of \sigmah, $z_h'' \oisim[w_h''] z_h'$ iff $z_h'' = z_h'$, therefore
       $\left(w_h'', z_h''\right) = \left(w_h', z_h'\right)$ and thus
      $\left(\left(w_h'', z_h''\right), \left(w_l', z_l'\right)\right) \in B$.
    \item Let $\left(w_l'', z_l''\right) \oicomp \left(w_l', z_l'\right)$ with $d_l(w_l'') > 0$ and $e_h, w_l'' \models \exec(z_l'')$.
      As $z_l'' \oisim[w_l''] z_l'$, the trace $z_l''$ must consist of the same actions as $z_l'$, except for a possibly different second parameter in each $\move(x, y)$.
      Furthermore, as \sigmah only contains the action \goto, the trace $z_l'$ only consists of mapped \goto actions, i.e., $z_l' \in \| m(\goto(l_1)); \ldots; m(\goto(l_n))\|^{\la\ra}_{e_l, w_l'}$
      We can split $z_l' = z_l^{'1} \cdot z_l^{'2}$ such that $z_l^{'2} \in \|m(\goto(l_n))\|^{z_l^{'1}}_{e_l, w_l'}$.
      Then, because of $z'' \oisim[w_l''] z_l'$, we can also split $z_l''$ such that $z_l'' = z_l^{''1} \cdot z_l^{''2}$, $z_l^{''1} \oisim[w_l'']  z_l^{'1}$ with $z_l^{''2} \in \|m(\goto(l_n')\|^{z_l^{''1}}_{e_l, w_l''}$.
      By induction, $((w_l', z_l^{'1}), (w_l'', z_l^{''1})) \in B$.
      Finally, as $z_l^{''2} \in \|m(\goto(l_n')\|^{z_l^{''1}}_{e_l, w_l''}$, it follows that $\left(\left(w_h', z_h'\right), \left(w_l'', z_l''\right)\right) \in B$ by definition of $B_{i}$.
  \end{enumerate}

  We conclude that $B$ is an $m$-bisimulation between $\left(e_h, w_h\right)$ and  $\left(e_l, w_l\right)$. Therefore, $\left(e_h, w_h\right) \bisim \left(e_l, w_l\right)$, and therefore $\sigmah$ is a sound abstraction of $\sigmal$.
\end{proofE}

Furthermore, the abstraction is also complete:
%

\begin{propositionE}
  \sigmah is a complete abstraction of \sigmal relative to refinement mapping $m$.
\end{propositionE}
\begin{proofE}[text proof={Proof Idea}]
  Let $e_h, w_h \models \know{\sigmah} \wedge \sigmah$.
  We show by construction that there is a model $\left(e_l, w_l\right)$ with $e_l, w_l \models \know{\sigmal} \wedge \sigmal$
  and $\left(e_h, w_h\right) \bisim \left(e_l, w_l\right)$.
  First, note that from $e_h, w_h \models \know{\sigmah}$ it follows that $d_h(w_h') > 0$ implies $w_h' \models \sigmal$.
  Now, for each $w_h^i$ with $d_h(w_h^i) > 0$, let $w_l^i$ be a world with $w_l^i \models \sigmal$ and such that $w_l^i$ is like $w_h^i$ for the high-level fluents, i.e., for every $F \not\in \mathcal{F}_l$ and every $z \in \traces$, $w^i[F, z] = w_h^i[F, z]$.
  Thus, $w^i$ is exactly like $w_h^i$ for every fluent not mentioned in \sigmal.
  We set $e_l(w_l^i) = e_h(w_h^i)$ and $e_l(w_l') = 0$ for every other world.
  Clearly, $e_l, w_l^i \models \know{\sigmal} \wedge \sigmal$.
  Now, let:
  \[
  B_0 = \{ \left(\left(w_h^i, \la\ra\right), \left(w_l^i, \la\ra\right)\right) \mid e_h(w_h^i) > 0 \}
  \]
  As before:
  \begin{align*}
    B_{i+1} = \big\{ \left(\left(w_h', z_h' \cdot a\right), \left(w_l', z_l' \cdot z_l''\right)\right) \mid
      &\left(\left(w_h', z_h'\right), \left(w_l', z_l'\right)\right) \in B_i,
      \\ &
      w_h', z_h' \models \poss(a), z_l'' \in \|m(a)\|^{z_l'}_{e_l, w_l}\big\}
  \end{align*}
  As each $w_l^i$ is like $w_h^i$, it follows that $B$ is definite.
  We can again show by induction on $i$ that $B$ is an $m$-bisimulation between $\left(e_h, w_h\right)$ and $\left(e_l, w_l\right)$. Therefore, $\left(e_h, w_h\right) \bisim \left(e_l, w_l\right)$ and thus, $\sigmah$ is a complete abstraction of $\sigmal$.
\end{proofE}

A \sigmah is a sound and complete abstraction of \sigmal relative to refinement mapping $m$, it follows with \autoref{thm:sound-complete-abstraction} that they entail the same (mapped) formulas.
Therefore, we can use \sigmah for reasoning and planning, e.g., we may write a high-level \golog program in terms of \sigmah and then use a classical \golog interpreter to find a ground action sequence that realizes the program.
To continue the example, we may write a very simple abstract program $\delta_h$ that first moves to the wall if necessary and then moves back:
\begin{algorithmic}
  \IIf{$\neg \at(\near)$}
    $\goto(\near)$
  \IEndIf;
  $\goto(\far)$
\end{algorithmic}
If the robot is initially not near the wall (as in our example), the following sequence is a realization of the program:
\[
  \la \goto(\near), \goto(\far) \ra 
\]
Note that this high-level trace is much simpler than the trace of the low-level program shown in \autoref{eqn:low-level-trace}.
At the same time, as \sigmah is a sound and complete abstraction of \sigmal, both traces are equivalent in the sense that the low-level trace results from translating the high-level program to the low-level \ac{BAT}.
Hence, for execution, this sequence may be translated to \sigmal by applying the refinement mapping $m$.
The translated program then takes care of stochastic actions and noisy sensors.

\section{Discussion}\label{sec:abstraction-discussion}
In this chapter, we have presented a framework for abstraction of probabilistic dynamic domains.
More specifically, in a first step, we have defined a transition semantics for \golog programs with noisy actions based on \ds, a variant of the situation calculus with probabilistic belief.
We have then defined a suitable notion of bisimulation in the logic that allows the abstraction of noisy robot programs in terms of a refinement mapping from an abstract to a low-level basic action theory.
As seen in the example, this abstraction method allows to obtain a significantly simpler high-level domain, which can be used for reasoning or high-level programming without the need to deal with stochastic actions.
Furthermore, for a user, the resulting programs and traces are much easier to understand, because they do not contain noisy sensors and actuators and are often much shorter.

While abstractions need to be manually constructed, future work may explore abstraction generation algorithms based on \cite{holtzenSoundAbstractionDecomposition2018,belleAbstractingProbabilisticModels2020}.
A further extension to our work might be to provide conditions under which we can modify the low-level program, with for example new sensors and actuators with different error profiles, but still show that the high-level program remains unmodified to achieve the intended high-level goal.

Interestingly, as the logics \ds and \es are fully compatible for non-probabilistic formulas not mentioning noisy actions \cite{belleReasoningProbabilitiesUnbounded2017} and abstraction allows to get rid of probabilistic formulas and noisy actions, we may construct \es programs that are sound and complete abstractions of \ds programs.
Therefore, if we have such an abstraction, it is entirely sufficient to write an abstract program that ignores all the probabilistic aspects of the domain and instead focuses on the high-level aspects of the reasoning task.
To actually execute the program on a robot, it can then be translated to a program of the low-level domain, which takes care of the stochastic actions and noisy sensors, which brings us a step towards closing the gap between high-level reasoning and plan execution.

\chapter{Conclusion}\label{chap:conclusion}

We summarize the main results of this thesis and then discuss possible future work.

\section{Summary}
While timing constraints and noisy actions are ubiquitous on real-world robotic systems, reasoning about actions usually expects a succinct description of the robot's capabilities that abstracts away timing aspects and uncertainty.
In this thesis, we have investigated several approaches towards bridging this gap between high-level reasoning systems and execution on a robot.
In the first part, we have taken into account the low-level platform components including their timing constraints with metric time.
\autoref{chap:timed-esg} provided the logical foundations by extending the logic \esg, a variant the situation calculus, with timed traces, real-valued clocks, and temporal logic.
We have seen that the resulting logic \tesg is a faithful extension of \esg, as basic action theories entail the same formulas in both logics.
This is a crucial property, because it allows us to use previously established results and apply them to \tesg, e.g., by combining \golog programs based on \tesg with planning~\parencite{classenPLATASIntegratingPlanning2012}.
At the same time, \tesg induces the same valid temporal formulas as \ac{MTL}.
As such, it can be seen as a faithful combination of reasoning about actions in the style of the situation calculus on the one hand and temporal properties in the style of \ac{MTL} on the other.

Building on top of \tesg, we have described two approaches to transform an abstract program into a platform-specific program that considers all platform constraints.
In both approaches, the platform components are modeled with \aclp{TA} with additional temporal formulas akin to \ac{MTL} that connect the abstract program with the robot self model.
In \autoref{chap:synthesis}, we have taken an approach based on \emph{synthesis}.
In this setting, the agent's actions are partitioned into actions controllable by the agent and actions controlled by the environment.
The synthesis problem is then to determine a realization of the program that is guaranteed to satisfy the specification independent of the environment's choices.
As we can model durative actions with \emph{start} actions under the agent's control and \emph{end} actions under the environment's control, this results in a program realization that can deal with actions whose durations are not known beforehand.
Additionally, exogenous events may also be modeled as environment actions, therefore the resulting controller is guaranteed to react to all exogenous events.
We have also described and evaluated an implementation of the approach.
While the tool is able to synthesize controllers, it does not scale well, partly due to the high complexity of the problem.
However, as it considers all possible environment choices, it is suitable for \emph{offline transformation}, at least with a limited scale: Given an abstract \golog program and a self model of the robot, we may determine a controller that executes the program in every possible scenario.
When executing the program, we then only need to execute the controller, which is able to react to all events as long as they are modeled by the program.

As the synthesis approach does not scale well, we have described a second approach based on some restricting assumptions.
Rather than executing a program with branches and loops, we focus on transforming a single plan, i.e., a sequence of actions.
Additionally, we assume that all actions are controllable by the agent.
These assumptions allow us to convert the program into a \ac{TA} and construct the product of the program automaton and the robot self model such that every accepting run on the automaton executes the program while satisfying all constraints.
To solve the transformation task, we can use the \ac{TA} verification tool \uppaal to determine a valid execution.
Due to the simplifications of the model and the sophisticated verification techniques implemented in \uppaal, this approach performs better than the first approach and scaled to plans with over 100 actions.
Therefore, it is suitable for \emph{online transformation}: Given an abstract \golog program, we can first determine a realization of the program and then transform the resulting plan during online execution such that all constraints are satisfied.
If an unexpected event occurs that renders the plan invalid, we may determine a new plan, transform it again, and then continue executing it.

Finally, in \autoref{chap:abstraction}, we have focused on \emph{uncertainty}.
In many robotics applications, uncertainty is present in the form of noisy sensors and effectors.
However, it is desirable to ignore stochastic aspects for reasoning tasks: For a developer, writing a program that incorporates stochastic actions is challenging and for the reasoner, determining a realization of the program is hard.
At the same time, when executing the program, these aspects need to be taken into account.
We therefore proposed to use \emph{abstraction} to deal with stochastic actions: In addition to the low-level basic action theory that includes noisy sensors and effectors, we model a second basic action theory that is an abstraction of the low-level theory and may be non-stochastic.
A \emph{refinement mapping} then maps high-level propositions and actions to low-level formulas and programs.
We have defined a suitable notion of \emph{bisimulation} that guarantees the equivalence between the two basic action theories.
Hence, we can use the high-level theory for writing a program and reasoning about actions and then translate the realization of the program to the low-level theory to deal with stochastic actions.

\section{Future Work}
For future work, it may be interesting to investigate the following aspects:
\begin{itemize}
  \item For the synthesis approach described in \autoref{chap:synthesis}, it may be promising to investigate techniques such as symbolic model checking~\parencite{larsenModelcheckingRealtimeSystems1995}, control structure analysis~\parencite{larsenUppaalStatusDevelopments1997}\todo{Double-check if this makes sense}, or symmetry reduction~\parencite{hendriksAddingSymmetryReduction2004} to improve the performance of the synthesis tool.
    These approaches have worked well for the \ac{TA} verification tool \uppaal, scaling well to large problems, even though these problems are quite difficult.
    Therefore, it seems reasonable to assume that they also result in significant performance benefits for the synthesis problem.
  \item A different approach towards improving the performance of the synthesis approach could be to restrict the constraint language.
    Rather than allowing full \ac{MTL}, it may be useful to consider weaker logics such as \mtlzi or time-bounded \ac{MTL}, where model checking has lower complexity~\parencite{ouaknineRecentResultsMetric2008}.
  \item While full \ac{MTL} is undecidable on infinite traces, Safety \ac{MTL} is decidable even on infinite traces.
    Therefore, restricting the constraint language to Safety \ac{MTL} may permit controller synthesis for non-terminating \golog programs.
  \item As the synthesis approach is capable of controlling the program against full \ac{MTL} and therefore allows an expressive temporal logic for constraints, we restricted the \acl{BAT} to a finite domain.
    While this may be suitable for many robotics applications, it may still be interesting to consider more expressive action representations, e.g., bounded action theories~\parencite{degiacomoBoundedSituationCalculus2016}.
  \item One assumption of the reachability approach in \autoref{chap:transformation-as-reachability-problem} is that all actions are controllable by the agent.
    This restriction was necessary to formalize the transformation problem as a reachability problem on \aclp{TA}.
    However, we may use a similar approach while allowing the environment to control some of the actions if we extend the approach to \emph{timed game automata}~\parencite{malerSynthesisDiscreteControllers1995}, which is also supported by \uppaal~\parencite{behrmannUPPAALTigaTimePlaying2007}. 
  \item In the abstraction framework described in \autoref{chap:abstraction}, so far we need to define the refinement mapping as well as the corresponding bisimulation manually.
    It would be interesting to do this algorithmically.
    A first step would be to verify the correctness of a given bisimulation between the high-level and low-level programs.
    In a second step, one could algorithmically check whether a bisimulation exists for a given refinement mapping.
    As this problem is related to the verification of belief programs~\parencite{liuProjectionProbabilisticEpistemic2022}, it can be expected that these problems are undecidable in general.
    In this case, it would be interesting to find expressive fragments that render those problems decidable.
\end{itemize}

\appendix

\chapter{Proofs}\label{chap:proofs}

\printProofs

\chapter{Contributions}\label{chap:contributions}
This appendix provides a list of publications by the author. 
\sloppy
\printbibliography[category=contributions,type=article,heading=subbibliography,title={Journal Publications}]
\printbibliography[category=contributions,type=inproceedings,notsubtype=workshop,heading=subbibliography,title={Conference Publications}]
\printbibliography[category=contributions,subtype=workshop,heading=subbibliography,title={Workshop Publications}]
\printbibliography[category=contributions,type=unpublished,heading=subbibliography,title={Poster Presentations}]

\sloppy
\printbibliography[notcategory=contributions]

\end{document}